\documentclass[a4paper,12pt,oneside,customfont,custombib,print,withindex]{PhDThesisPSnPDF}

\input{Preamble/preamble}

\newcommand{\thesistitle}{Intuitive Human-Robot Interfaces Leveraging on Autonomy Features for the Control of Highly-redundant Robots}
\newcommand{\thesistitleshort}{Intuitive Human-Robot Interfaces Leveraging on Autonomy Features}

\title{\thesistitle{}}

\subtitle{PhD Thesis}

\author{Davide Torielli}

\dept{DIBRIS}

\university{University of Genova}

\degreetitle{Doctor of Philosophy}

\subject{Human-Robot Control Interfaces} 

\ifdefineAbstract
\fi

\ifdefineChapter
\fi

\makeglossaries

\newacronym{dof}{DoF}{Degrees of Freedom}
\newacronym{tpo}{TPO}{TelePhysicalOperation}
\newacronym{ros}{ROS}{Robot Operating System}
\newacronym{imu}{IMU}{Inertial Measurement Unit}
\newacronym{emg}{EMG}{Electromyography}
\newacronym{vslam}{V-SLAM}{Visual-Simultaneous and Localization Mapping}
\newacronym{usb}{USB}{Universal Serial Bus}
\newacronym{vpu}{VPU}{Visual Processing Unit}
\newacronym{gui}{GUI}{Graphical User Interface}
\newacronym{api}{API}{Application Programming Interface}
\newacronym{sot}{SoT}{Stack of Tasks}
\newacronym{qp}{QP}{Quadratic Programming}
\newacronym{tpik}{TPIK}{Task Priority Inverse Kinematic}
\newacronym{vtr}{VTR}{Velocity Transmission Ratio}
\newacronym{ee}{EE}{End-effector}
\newacronym{rgb}{RGB}{Red Green Blue}
\newacronym{rgbd}{RGB-D}{Red Green Blue Depth}
\newacronym{hsv}{HSV}{Hue Saturation Value}
\newacronym{pcb}{PCB}{Printed Circuit Board}
\newacronym{cmos}{CMOS}{Complementary Metal-Oxide Semiconductor}
\newacronym{anova}{ANOVA}{ANalysis Of VAriance}
\newacronym{cpu}{CPU}{Central Processing Unit}
\newacronym{gpu}{GPU}{Graphical Processing Unit}
\newacronym{ram}{RAM}{Random Access Memory}
\newacronym{led}{LED}{Lighting Emitting Diode}
\newacronym{adl}{ADL}{Activities of Daily Living}
\newacronym{bldc}{BLDC}{Brushless DC electric motor}
\newacronym{bt}{BT}{Behavior Tree}
\newacronym{pid}{PID}{Proportional Integrative Derivative}
\newacronym{fsm}{FSM}{Finite State Machine}
\newacronym{ar}{AR}{Augmented Reality}
\newacronym{bomi}{BoMI}{Body-Machine Interfaces}
\newacronym{bci}{BCI}{Brain-Computer Interfaces}
\newacronym{eeg}{EEG}{electroencephalogram}
\newacronym{hal}{HAL}{Hardware Abstraction Layer}
\newacronym{cern}{CERN}{Conseil Européen pour la Recherche Nucléaire}
\newacronym{hhcm}{HHCM}{Humanoids and Human Center Mechatronics}
\newacronym{iit}{IIT}{Italian Institute of Technology}
\newacronym{sirslab}{SIRSLab}{Siena Robotics and Systems Lab}
\newacronym{friend}{FRIEND}{Functional Robot with dexterous arm and user-frIENdly interface for Disabled people}
\newacronym{toro}{TORO}{TOrque-controlled humanoid RObot}
\newacronym{moca}{MOCA}{MObile Collaborative robot Assistant}
\newacronym{edan}{EDAN}{EMG-controlled Daily AssistaNt}
\newacronym{puma}{PUMA}{Programmable Universal Machine for Assembly, or Programmable Universal Manipulation Arm}
\newacronym[description={Socio-physical Interaction Skills for Cooperative Human-Robot Systems in Agile Production~\cite{sophia}}]{sophia}{SOPHIA}{Socio-physical Interaction Skills for Cooperative Human-Robot Systems in Agile Production}
\newacronym[description={CONfigurable CollaborativE Robot Technologies~\cite{concert}}]{concert}{CONCERT}{CONfigurable CollaborativE Robot Technologies}
\newacronym[description={Robot Enabler for Load Assistive relaXation~\cite{relax}}]{relax}{RELAX}{Robot Enabler for Load Assistive relaXation}
\newacronym[description={Human-Robot Sensorimotor Augmentation~\cite{haria}}]{haria}{HARIA}{Human-Robot Sensorimotor Augmentation}

\begin{document}

\thispagestyle{empty}
\begin{figure}[h!]
 \centering
 \includegraphics[scale=0.25]{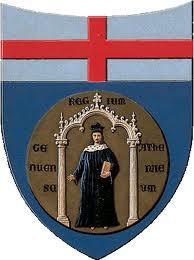} %
	\begin{center} 
		\Large
		{\textsc{University of Genova}}\\
		  \vspace{0.5em}
		  \large
	         \textsc{PhD Program in Bioengineering and Robotics}
	\end{center}
\end{figure}
\vspace{-1cm}

\begin{center}

		\LARGE
		\textbf{\thesistitle{}} \\
\end{center}

 	\begin{center} 
		by \\
		\vspace{0.5em}
		\textbf{Davide Torielli}\\
		\vspace{1em}

	\vspace{0.5cm}	
		\normalsize
		Thesis submitted for the degree of \textit{Doctor of Philosophy} ($36^\circ$ cycle) \\
	\vspace{0.5cm}	
		\normalsize
		January 2024\\ 
	\end{center}
	\vspace{0.4em}

\vfill

	\noindent {Nikos Tsagarakis} \hfill  {Supervisor}	
	\\
	\noindent {Luca Muratore} \hfill  {Supervisor}	
	\\
	\noindent {Paolo Massobrio}	\hfill  {Head of the PhD program}	
	\vspace{1em} \\

\vspace{2em}
\vfill
\begin{figure}[h!]
 \centering
 \includegraphics[height=1.2cm]{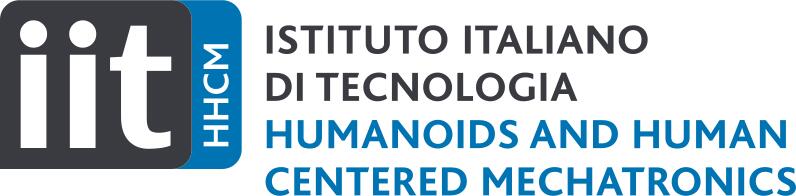}
 \hspace{20px}
 \includegraphics[height=1.2cm]{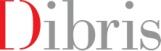}
\begin{center} 
	    \small{Humanoids and Human Centered Mechatronics lab, Istituto Italiano di Tecnologia\\}
		\small{Department of Informatics, Bioengineering, Robotics and Systems Engineering\\}
	\end{center}
\end{figure}

\frontmatter

\begin{dedication}

	\begin{center}
		\thispagestyle{empty}
		\vspace*{\fill}
		\begin{flushright}
			``Choose a job you love,\\
			and you will never have to work a day in your life'' \\
			\textit{(\href{https://quoteinvestigator.com/2014/09/02/job-love/}{An Old-Timer})}
			
		\end{flushright}
		\vspace*{\fill}
	\end{center}

\end{dedication}

\begin{declaration}

I hereby declare that except where specific reference is made to the work of 
others, the contents of this dissertation are original and have not been 
submitted in whole or in part for consideration for any other degree or 
qualification in this, or any other university. This dissertation is my own 
work and contains nothing which is the outcome of work done in collaboration 
with others, except as specified in the text.

\end{declaration}

\begin{acknowledgements}      

Once I read on a thesis that the author read that the acknowledgments should begin with a long list of names. Unfortunately, I have too many people I would like to add, and I am terrified that I will forget some of them. As a general rule, if you have ever been invited at \textit{Cabanetta}, consider you on this list.
Nevertheless, let's go by categories.
First acknowledgment to my parents and my whole family, for always supporting me. Thanks to my \textit{Sanfru}\footnote{The \href{https://maps.app.goo.gl/ZYEeYMxJZ58A6fFA7}{real one}, not the fake Camogli one.} friends, who probably will never read this. Thanks to \acrshort{hhcm} lab people, and all the other ones met in \acrshort{iit} and on the hiking trips, they were colleagues that seamlessly become friends. 
I met also wonderful people all around the world thanks to conferences and schools: I would like to not forget them neither.
Thanks to my supervisors, their guidance, ideas and support were awesome. 
Lastly \textit{and least}, thanks to myself, because I have reached the end of this adventure. Bravo, Tori.

\end{acknowledgements}

\begin{abstract}

The advancements in robotics have revealed the potential of complex robotic platforms, promising a wide spread of robotics technologies to help people in various scenarios, from industrial to households. 
To harness the capabilities of modern robots, it is of paramount importance to develop human-robot interaction interfaces that allow people to seamlessly operate them.
To address this challenge, traditional interface methods, such as remote controllers and keyboards, are going to be replaced by more intuitive communication means, that permit, for example, to command the robot through body gestures, and to receive feedback that extends the visual domain, such as tactile clues.
At the same time, the robot must be equipped with autonomous capabilities that relieve the operators in considering all the aspects of the task and of the robot motions, thus reducing their workload, decreasing the execution time of the task, and minimizing the possibility of failures.

This PhD thesis takes on these challenges by exploring and developing innovative human-robot interaction paradigms that focus on the key aspects of enabling intuitive human-robot communication, enhancing user's situation awareness, and incorporating different levels of robot autonomy. 

With the TelePhysicalOperation interface, the user can teleoperate the different capabilities of a robot (e.g., single/double arm manipulation, wheel/leg locomotion) by applying virtual forces on selected robot body parts.
This approach emulates the intuitiveness of physical human-robot interaction, but at the same time it permits to teleoperate the robot from a safe distance, in a way that resembles a \enquote{Marionette} interface.
The system is further enhanced with wearable haptic feedback functions to align better with the \enquote{Marionette} metaphor, and a user study has been conducted to validate its efficacy with and without the haptic channel enabled. Considering the importance of robot independence, the TelePhysicalOperation interface incorporates autonomy modules to face, for example, the teleoperation of dual-arm mobile base robots for bimanual object grasping and transportation tasks.

With the laser-guided interface, the user can indicate points of interest to the robot through the utilization of a simple but effective laser emitter device. 
With a neural network-based vision system, the robot tracks the laser projection in real time, allowing the user to indicate not only fixed goals, like objects, but also paths to follow.
With the implemented autonomous behavior, a mobile manipulator employs its locomanipulation abilities to follow the indicated goals. The behavior is modeled using Behavior Trees, exploiting their reactivity to promptly respond to changes in goal positions, and their modularity to adapt the motion planning to the task needs.
The proposed laser interface has also been employed in an assistive scenario. In this case, users with upper limbs impairments can control an assistive manipulator by directing a head-worn laser emitter to the point of interests, to collaboratively address activities of everyday life.

In summary, this research contributes to effectively exploiting the extensive capabilities of modern robotic systems through user-friendly human-robot interfaces. With the developed interfaces, the gap that still prevents a large adoption of robotic systems is further reduced.

\vspace{20px}


\end{abstract}

\tableofcontents

\listoffigures

\listoftables

\printglossary
\printglossary[nogroupskip, type=\acronymtype]

\mainmatter

\chapter{Introduction}\label{chap:intro}

\lettrine{H}{uman-robot} interaction is a broad field of study \enquote{dedicated to understanding, designing, and evaluating robotic systems for use by or with humans}~\cite{Goodrich2008}. This field has gained significant importance for the potential of deploying robots in human environments to support various tasks.
Humans interact with robots through human-robot interfaces, which are the apparatus with which users convey instructions, commands for the robot are generated, and information about the ongoing processes is delivered to the user. The design and effectiveness of these interfaces significantly impact the overall human-robot interaction experience, and are indeed important to fully employ the robotic potential in real-world contexts. 

While the roots of human-robot interaction extend deep into the past~\cite{Goertz1952}, its evolution is still very active today, in order to address the challenges posed by the advancement of robotic capabilities.
The continuous advancements in robotic mechatronic design have led to the creation of highly-redundant robots. These robots possess extensive motion capabilities, incorporating multiple arms for manipulation, wheels and legs for locomotion, and humanoids bodies to complement their movements. This has permitted to tackle increasingly complex locomanipulation tasks.
Indeed, the increasing robotic skills have permitted to enhance the range of scenarios where robots can be employed, inspiring researchers in exploring various applications like human-robot collaboration~\cite{VILLANI2018, Berg2020}, industrial manufacturing~\cite{Hoglund2018}, inspection and maintenance~\cite{Lu2017}, construction sites~\cite{Carra2018}, agriculture fields~\cite{Emmi2014}, post-disaster areas~\cite{Liu2013}, elderly care~\cite{Bardaro2022} rehabilitation~\cite{Colombo2018}, assistance~\cite{Petrich2022}, clinical therapies~\cite{Goodrich2013}, and surgery~\cite{Taylor2003}.

To allow users to fully leverage these high-capable systems, human-robot interfaces must enable an effortless and efficient interaction. Developments in this direction are crucial to surpass the current limitations and to foster the adoption of robotic systems in real-world applications, to assist human in executing demanding tasks, operating in hazardous environments and facilitating routine daily activities.

\section{Objectives}

The focus of this thesis is the exploration and development of novel human-robot interfaces, empowering users to effortlessly utilize modern robotic systems in various applications.

\noindent The main objectives pursued by this thesis consist in what follows:

\begin{itemize}

	\item \textbf{Develop intuitive interfaces}.
	One objective follows the recent trends toward the development of intuitive methods to express user's command to the robot~\cite{VILLANI2018}. Intuitiveness plays a crucial role since it allows the user to interact naturally with the robot, resembling a communication with other people. This allows to minimize the time necessary to learn how to operate the robot, and to reduce the effort when executing a task.
	To purse this, modern works are trying to overcome the limitations of traditional interaction means, like keyboards, remote controllers, joysticks, and teach pendants. In fact, such input devices are usually not intuitive at all, requiring efforts and time to learn them, possibly limiting the exploration of the high-capabilities of robots. Hence, the first objective is to develop new intuitive interfaces to enable the user to use his/her body, in particular his/her arms, to provide commands to the robot, in a very intuitive and natural way. 

	\item \textbf{Convey to the user relevant situational awareness information}. If the whole interface must be intuitive, it is also important to follow this characteristic when providing the user with situational awareness, which is fundamental to comprehending what happens during the human-robot interaction~\cite{Endsley2016}. Hence, another objective is to develop methods to deliver relevant information to the user maintaining the communication intuitive. In particular, the focus is on haptic feedback channels to complement simple visual feedback means.

	\item \textbf{Equip the robot with autonomy}.
	Another key objective is to furnish robots with a certain level of autonomy. Indeed, an interface can not be truly easy-to-use if the user must control every motion of a highly-redundant robot and manage every aspect of a task. At the same time, a fully autonomous robot is not achievable yet, since actual technologies can not reach the humans abilities in decision-making and in responding to unexpected situations~\cite{Gaofeng2023}. A suitable compromise must be addressed, which depends on the robot in use, on the task, and also on the user's preferences. 
	Thus, efforts are made to incorporate different robot autonomy behaviors, permitting the user to act in a shared-control and supervisory level~\cite{Selvaggio2021}.

\end{itemize}

The final objective is to integrate all these concepts and realize them by implementing a new set of human-robot software architectures, validating such methods in operating robotic systems in practical applications. The realization of such interfaces must result in versatile solutions, to be adaptable to various scenarios. At the same time, simplicity is a key feature taken into consideration to not overload the operator, to reduce the setup time, and to minimize the costs.

Discussions about the contributions of this thesis are given in the following Section~\ref{sec:intro:contrib}, while comprehensive details of how these objectives have been pursued are provided throughout the rest of it.

\section{Contributions}\label{sec:intro:contrib}

In this section, the main contributions of to the works presented in this thesis are introduced, highlighting their novel characteristics.

\subsection{TelePhysicalOperation: A \enquote{Marionette} Teleoperation Interface}
This contribution is about the ideation of the \acrfull{tpo} concept. The idea merges the classical teleoperation, hence controlling the robot from a distance, with the intuitiveness of the physical human-robot interaction. By applying virtual forces on the robot's body, the operator can control the robot as in a physical interaction employed for guiding/teaching tasks, but from a distance without any physical contact. The innovation lies in this novel approach of interacting with a robot, that resembles a \enquote{Marionette} interface since the operator pulls/pushes the robot with virtual ropes attached to his/her arms and to the selected robot's body parts. To realize such a concept, a lightweight and unobtrusive wearable interface has been designed to track the operator's arms movements which generate the virtual forces. This is complemented by the development of an appropriate software architecture, which enables this mode of interaction with different kinds of robots without necessitating the operator to know about their kinematic details~\cite{TPO}.
	
\subsection{Wearable Haptic-enabled TelePhysicalOperation }
In a real \enquote{Marionette} interface, the person can feel the tension of the rope when moving the puppet, intrinsically helping him/her in the motion control. To comply better with this metaphor, the TelePhysicalOperation interface has been enhanced with the addition of a haptic feedback channel. The novel vibrotactile wearable device, consisting in a bracelet and a finger ring, has been designed by \acrfull{sirslab}, University of Siena. In collaboration with this institution, the contributions regards the conceptualization of the haptic-enabled \acrshort{tpo} interface, the integration of the devices in the framework, and the design and evaluation of a user-study to validate the TelePhysicalOperation interface with and without the haptic functionality~\cite{TPO4}.

\subsection{Manipulability-Aware Shared Locomanipulation Motion Generation}
Another contribution regards the development of a manipulability-based locomanipulation shared control interface, enabling the generation of arm motions and mobile base motions of a mobile manipulator from a single input on the end-effector (e.g., the virtual force of the \acrshort{tpo} interface). While manipulability is usually employed as a secondary task to optimize the arm's shape, in this interface it is employed to balance the motion between the arm and the mobile base. Specifically, when the end-effector is in a region of low manipulability, it results that the arm is slowed down in favor of mobile base motions. The novelty also lies in the utilization of the \acrfull{vtr} measure, representing the length of the axes of the manipulability ellipsoid. This measure permits to consider the manipulability in the three principal directions, generating arm and mobile base motions accordingly to the manipulability level in different directions.
The method is versatile and applicable to any generic interface to facilitate the execution of locomanipulation tasks with mobile base manipulators. In this case, it has been integrated in the \acrfull{tpo} architecture~\cite{TPO2}.
	
\subsection{Bimanual Grasping and Transportation of Objects of Unknown Mass}
With this contribution, an interface has been developed for teleoperating dual-arm mobile robots for bimanual transportation tasks, incorporating a specific level of shared control. This allows the robot to autonomously regulate the grasping forces on the object being transported. Meanwhile, the operator can focus solely on providing directions for the object, without worrying about the generation of arm motions to adhere to the grasping constraint while transporting. 
To further facilitate this task, the robot can autonomously reach the object, grasp it, and estimate its weight. The weight estimation is necessary to compute the necessary grasping forces during transportation, to prevent the object from falling or being excessively squeezed.
This interface has been combined with the previously-mentioned manipulability-based locomanipulation motion generation and integrated in the \acrlong{tpo} architecture~\cite{TPO3}.

\subsection{A Laser-guided Interface Exploring Neural Network Perception and Behavior Trees Motion Generation}
Another contribution involves the development of a laser-guided human-robot interface designed for supervisory control, which enables the operator to effortlessly designate locations or objects using an inexpensive laser emitter device. 
The laser projection is detected utilizing the robot vision system. This system allows for rapid detection of the laser spot, employing a neural network that is more robust and faster with respect to previous solutions based on classical computer vision algorithms, with the only practical drawback of necessitating to train the network model, even if it is a one-time process.
The interface combines the laser spot detection with a robot motion planner based on \acrlong{bt}s (BTs). The reactivity of this solution well-combines with the responsiveness of the neural network model, ensuring that the user experiences no latency between the selection of a location and the robot motion. This permits not only the selection of a fixed location as a goal but also enables the design of a path which the robot is able to follow in real-time. 
Furthermore, thanks to the modularity of the \acrshort{bt}s, it is straightforward to modify the robot behavior according to the task and to the robot in use. For example, the \acrshort{bt}'s structure can be configured to let a mobile manipulator follow the goal only with the mobile base or engage both the body and the arm for a more complex locomanipulation task~\cite{LaserJournal}. 
	
\subsection{A Laser-guided Interface for Robotic Assistance}
The intuitiveness of the laser-guided interface is also employed in an assistive scenario, where users with upper limbs impairments interact with a robot to accomplish \acrfull{adl} tasks. 
While in the previous contribution the laser emitter was handheld, now it is worn on the user's head, permitting to control the robot with head movements which results in the laser being pointed to specific locations. With respect to previous state-of-the-art works which employ head movements interfaces, this approach offers users a simple, cost-effective, and comfortable input system that does not require a specific mapping between the input and the robot's motion. 
The perception layer that detects the laser spot employs the same neural network as before. However, a different control layer has been integrated to provide the user with two different control modalities: pose-based and velocity-based. 
The first employs standard tools to generate a collision-free trajectory toward the indicated goal. The second enables a more direct control by utilizing a paper keyboard placed in the environment. The keys can be selected by directing the laser at them to control the end-effector in the Cartesian space and to perform gripper actions. The two modalities well-combine without the necessity of additional inputs to switch between them since it is only necessary to project the laser in the appropriate location to utilize one or the other modality~\cite{LaserRal}.

\section{Robots and Software Employed}\label{sec:intro:tools}

A good engineer does not reinvent the wheel. Obviously, this thesis relies on previously developed software and hardware to implement and validate the works carried out. The most important systems utilized in the works discussed are listed in this section, while other important ones will be introduced in the relevant chapters.

One such kind of software is \acrfull{ros}~\cite{ROS}, the well-known open-source middleware that provides a variety of libraries and tools for building robotic applications, used extensively in all the human-robot interfaces developed to allow for a seamless communication between the various software and hardware modules.

In what follows, they are presented systems developed at the \acrfull{hhcm} lab, \acrfull{iit}, where the works discussed in this thesis have been carried out.

\subsection{CENTAURO Robot}\label{sec:intro:centauro}

\begin{figure}[H]
	\centering
	\includegraphics[width=0.5\linewidth]{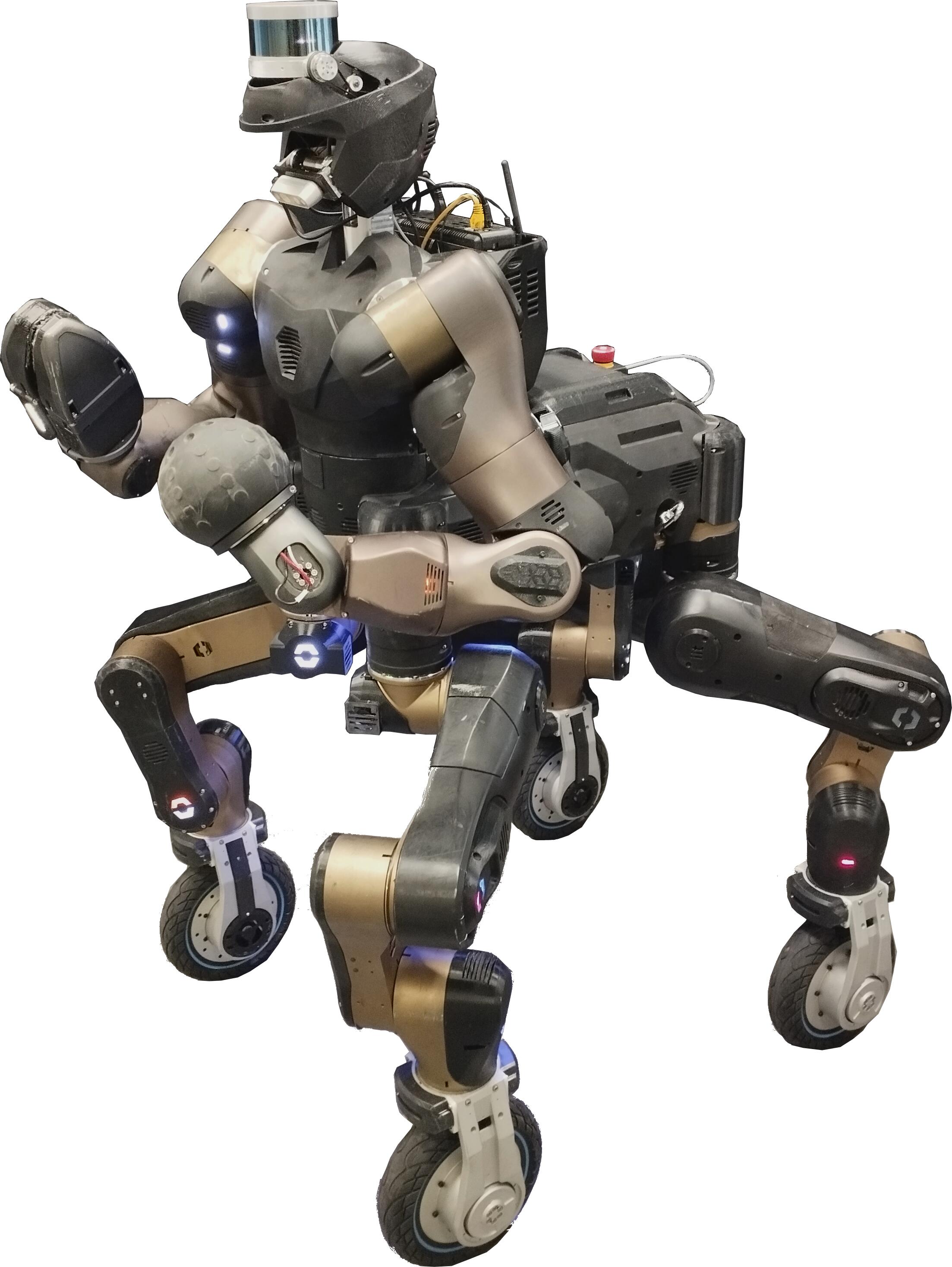}
	\caption[The CENTAURO robot]{The CENTAURO robot, a hybrid leg-wheel system with an anthropomorphic upper body, utilized in different validations in the works presented in this thesis.}
	\label{fig:centauro}
\end{figure}

The CENTAURO~\cite{Centauro, Centauro2} is a robotic platform which combines manipulation and locomotion abilities, developed within the \textit{CENTAURO} European Project~\cite{CentauroProject}. The system, shown in \figurename{}~\ref{fig:centauro} is composed by a quadruped body with wheels and a humanoid dual-arm torso.
Each leg has a 5-\acrshort{dof} spider-like kinematics that permits the positioning and orientation of the wheeled foot. The last actuator of the leg provides the steering motion to the wheel that has a dedicated additional motor for the rolling motion. 
The wheel module has evolved over the years, showing a thicker size with a \SI{0.078}{\meter} diameter in some robot configurations and a thinner size with a \SI{0.124}{\meter} diameter in other configurations.
The anthropomorphic-like arms are composed of a 3-\acrshort{dof} shoulder, a flexion-extension elbow, a yaw-pitch forearm, and a wrist that in some robot configurations has an additional yaw \acrshort{dof}, for a total of $6/7$ \acrshort{dof}. 
Also, the arm end-effectors are modular and interchangeable. In this thesis, two types of end-effectors have been exploited: a ball-shaped passive end-effector (mounted on the left arm in \figurename{}~\ref{fig:centauro}), and a 1-\acrshort{dof} beak-like gripper, the DAGANA end-effector, introduced in Section~\ref{sec:intro:dagana} (mounted on the right arm in \figurename{}~\ref{fig:centauro}).
Additional sensors, like LIDAR, cameras, force-torque sensors, can be present in the body of the robot; in some the works presented in this thesis, it has been exploited an Intel\textsuperscript{\textregistered} RealSense D435i\footnote{\url{https://www.intelrealsense.com/depth-camera-d435i/}} mounted on a pitch actuator on the robot head.

The CENTAURO has been designed to deliver high manipulation performance, with each arm able to manipulate up to \SI{10}{\kilo\gram} and different locomotion possibilities enabled by the hybrid leg-wheel system, permitting to face locomanipulation tasks of different kinds. The actuators are torque controlled, permitting to be compliant to external disturbances, a feature often necessary for critical manipulation tasks.

The locomanipulation abilities of this highly-redundant platform have permitted to validate in deep the functionalities of the human-robot interfaces developed. In particular, the robot has been utilized in the experimental validations of Chapter~\ref{chap:TPO}, Chapter~\ref{chap:TPOH}, Chapter~\ref{chap:tpoAuto}, and Chapter~\ref{chap:Laser1}.

\subsection{Robotic Manipulator}\label{sec:intro:arm}
\begin{figure}[H]
	\centering
	\includegraphics[width=0.55\linewidth]{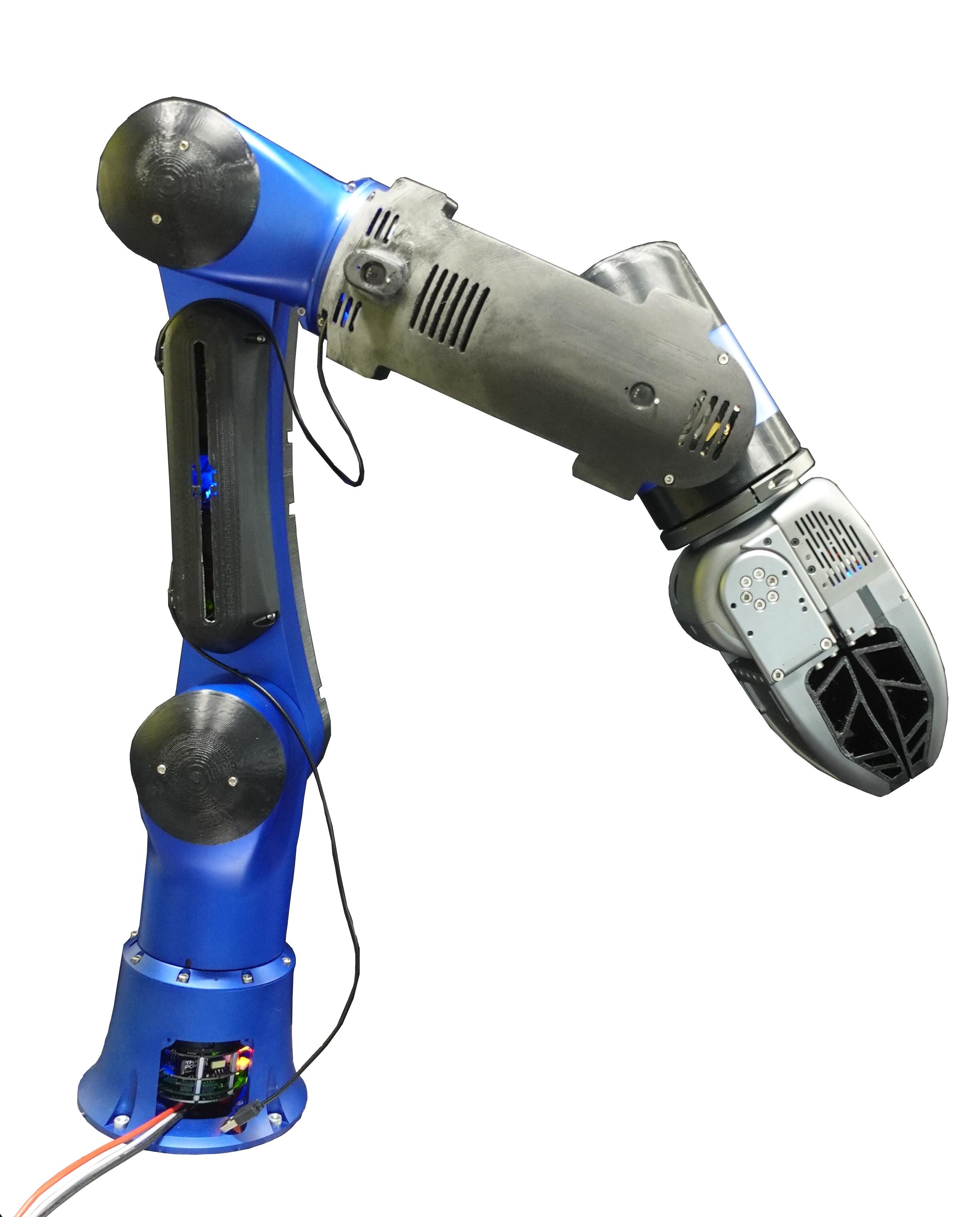}
	\caption[The robotic manipulator developed at \acrshort{hhcm} lab]{The robotic manipulator developed at the \acrshort{hhcm} lab, equipped with the DAGANA gripper.}
	\label{fig:inailarm}
\end{figure}

The robot in \figurename{}~\ref{fig:inailarm} is a lightweight,  small size, 6-\acrshort{dof} manipulator developed at the \acrshort{hhcm} lab. It is long \SI{1.02}{\meter} when full stretched, and weighs \SI{10.5}{\kilo\gram}. In the picture the DAGANA gripper, presented in Section~\ref{sec:intro:dagana}, is mounted as an end-effector.

The design of the manipulator consists in a 2-\acrshort{dof} shoulder, a 1-\acrshort{dof} elbow, and a 3-\acrshort{dof} wrist. The continuous payload capacity of the robot is \SI{4}{\kilo\gram}.
The robot is equipped with compliant interaction capabilities achieved through the use of torque-controlled actuators, which are realized by integrating \acrshort{bldc} motor units with harmonic reduction drives and torque sensors in the robot arm joints.

This robotic manipulator has been employed in the validations of the human-robot interface developed for the assistive scenario described in Chapter~\ref{chap:Laser2}.

\subsection{DAGANA End-Effector}\label{sec:intro:dagana}

\begin{figure}[H]
	\centering
	\includegraphics[height=3.5cm]{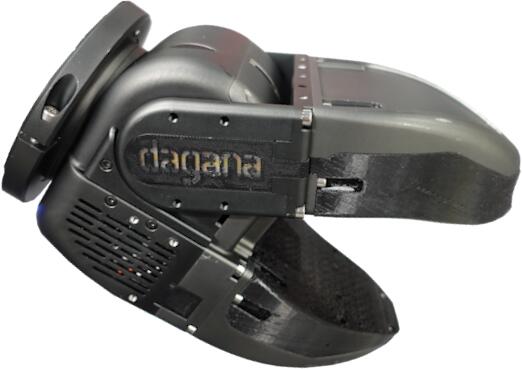}
	\hspace{15px}
	\includegraphics[height=3.5cm]{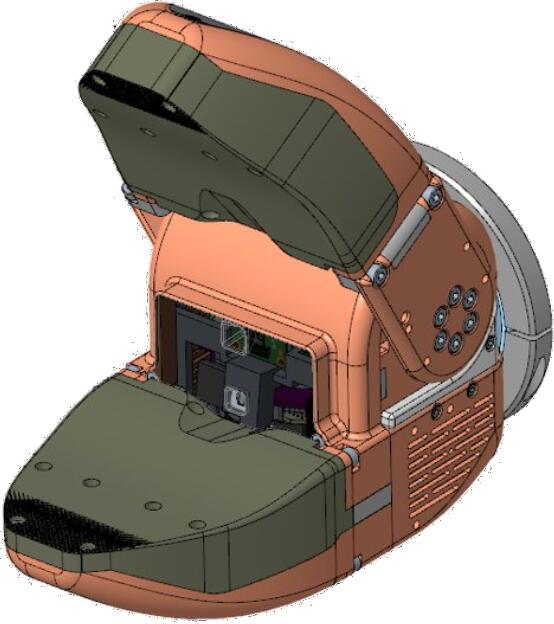}
	\hspace{25px}
	\includegraphics[height=3.5cm]{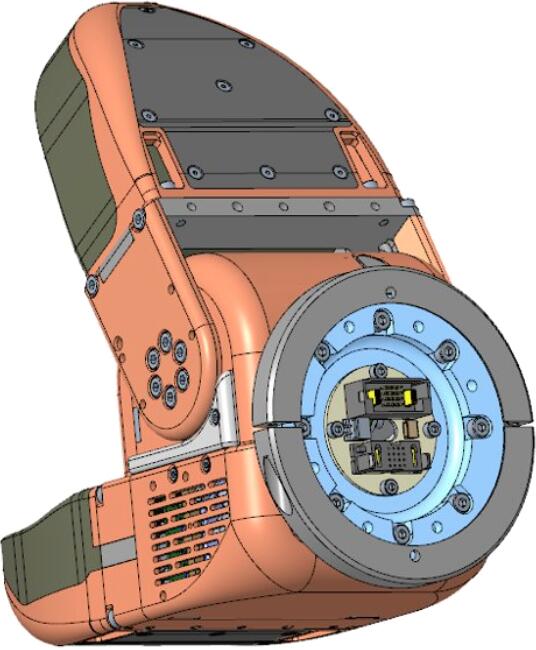}
	\caption[The DAGANA gripper]{The DAGANA gripper, a 1-\acrshort{dof} beak-like end-effector.}
	\label{fig:dagana}
\end{figure}

The DAGANA is a robotic end-effector developed at the \acrshort{hhcm} lab, consisted in a powerful and high capable modular grasping tool to be equipped on different robotics arms (\figurename{}~\ref{fig:dagana}). The design consists in a double jaw type gripper in which the bottom jaw is fixed while the top jaw is powered by the single actuator. The jaws are modular and can be changed easily with jaws of different shapes and material to accommodate the grasping of a particular tool or object. 
The DAGANA is also equipped some sensory devices: a \SI{2}{MegaPixel} \acrshort{cmos} camera, a proximity sensor of \SI{0.4}{\meter} range, a $6$-axis motion sensor, and an omnidirectional microphone. The devices and the electronics are integrated in the front side of the main gripper body.

The DAGANA gripper has been utilized as the grasping tool for both the CENTAURO and the robotic manipulator described in the previous sections. In particular, its grasping capacities have been employed in the experiments of Chapter~\ref{chap:TPOH}, Chapter~\ref{chap:Laser1}, and Chapter~\ref{chap:Laser2}.  

\subsection{XBot and CartesI/O Software Frameworks}\label{sec:intro:framework}
A powerful and high-skilled robot is useless without a suitable control system. Thus, to manage different robots like the ones presented, in the \acrshort{hhcm} lab it has been developed a real-time software architecture, the \textit{XBot} middleware, whose first version has been presented in~\cite{XBot}, and its second version in~\cite{XBot2}.
This middleware provides the \acrfull{hal} and the \acrlong{api}s (APIs) necessary to communicate with the hardware systems. 
Another tool is \textit{CartesI/O}~\cite{cartesio}, a Cartesian control software framework, based on the \textit{Open-SoT} library~\cite{Hoffman2017} which implements a \acrfull{sot} strategy to solve prioritized \acrfull{qp} problems applied to Cartesian control. 
The above-mentioned software nicely integrates with the \acrshort{ros} ecosystem, providing the infrastructure to communicate with the common robotics software tools. 

These software frameworks have been employed in all the works presented in this thesis; details about the utilization of their functionalities will be given where necessary. 

Further insights about \textit{XBot} and \textit{CartesI/O} can be found in the PhD thesis~\cite{MuratorePHD} and~\cite{LaurenziPHD}.

\section{Outline}
The structure of this thesis is divided as follows.

\noindent The introductory part, to which this chapter belongs, is composed by two other chapters:
\begin{itemize}
	\item Chapter~\ref{chap:soa} reviews the current state-of-the-art related to human-robot interfaces for operating a robotic system. The focus is on the technologies that address the key points addressed by this thesis: intuitive communication, haptic feedback, and robot autonomy. The chapter highlights limitations of previous works and includes brief discussions on relevant background, with references provided for in-depth explanations.
	
	\item Chapter~\ref{chap:chap3} presents and explains the meaning of intuitiveness, situation awareness, and robot autonomy related to human-robot interfaces. The main challenges are highlighted and their significance is discussed. The proposed interfaces, that will be detailed extensively in the other chapters, are introduced in consideration of the key features and challenges presented.
\end{itemize}	
	
\noindent The details of the works developed are explained dividing them into two main parts.

\noindent Part~\ref{part:one} regards the \acrlong{tpo} interfaces:
\begin{itemize}	
	\item Chapter~\ref{chap:TPO} introduces the \acrfull{tpo} interface. It explains the concept behind this \enquote{Marionette} kind of interface, and its advantages for remotely controlling robots. The chapter explains the software architecture and the wearable hardware interface utilized to track the operator's arm movements. Experimental validations are presented.  
	
	\item Chapter~\ref{chap:TPOH} presents an enhancement of the \acrlong{tpo} interface, which consists in the addition of a channel of haptic feedback to enhance the situation awareness for the user and to better align with the \enquote{Marionette} paradigm. The chapter explains the design of the wearable vibrotactile haptic devices employed to convey haptic feedback and details their integration in the \acrshort{tpo} framework. New experiments involving naive users are presented to validate the interface with and without the haptic activation.
	
	\item Chapter~\ref{chap:tpoAuto} provides insights about the robot autonomy functionalities of the \acrlong{tpo} interface, developed to enhance the human-robot interface and exploit highly-redundant robots more efficiently. The chapter details the manipulability-aware locomanipulation method, along with a framework designed to facilitate the teleoperation for bimanual grasping and transportation tasks. New, more complex experiments are presented to validate such enhancements.
\end{itemize}

Part~\ref{part:two} regards the laser-guided interfaces:
\begin{itemize}

	\item Chapter~\ref{chap:Laser1} introduces the laser-guided interface ideated to facilitate the control of robot by intuitively and effortlessly indicating points in the environment with an inexpensive laser device. The chapter presents the robot perception layer used to detect the laser projection in the environment, and how robot autonomy is incorporated to generate motions based on a \acrlong{bt} planner. Relevant experiments are showcased with a series of locomanipulation tasks.
	
	\item Chapter~\ref{chap:Laser2} employs the laser-guided interface to face the challenges of an assistive scenario, where users with impaired arms collaborate with a robotic manipulator to face \acrlong{adl} tasks. Given the context, the laser device is worn on the user's head, allowing intuitive command of the robot through head gestures. Different control modes are presented, featuring varying levels of robot autonomy. Relevant experimental validations are provided.
\end{itemize}
	
\noindent Finally, Chapter~\ref{chap:conclusion} draws the conclusions, summarizing the works developed and introducing potential future directions.

\section{Publications}
In this section the scientific publications resulting from the works presented in this thesis are listed.
These works have received funding from the European Union's Horizon 2020 research and innovation programme under grant agreement No. 871237 \acrfull{sophia}~\cite{sophia}, grant agreement No.\ 101016007 \acrfull{concert}~\cite{concert}, and grant agreement No.\ 101070292 \acrfull{haria}~\cite{haria}, and the Italian Fondo per la Crescita Sostenibile - Sportello \enquote{Fabbrica intelligente}, PON I\&C 2014 - 2020, project number F/190042/01-03/X44 \acrfull{relax}~\cite{relax}.

\begin{refsection}
	
\subsection{Journal Articles (Full Peer Review)}
\begin{enumerate}[label={[J.\arabic*]}]
	\item \fullcite{TPO}~\cite{TPO}
	\item \fullcite{Muratore2023}~\cite{Muratore2023}
\end{enumerate}

\subsection{Conference Articles (Full Peer Review)}
\begin{enumerate}[label={[C.\arabic*]}]
	\item \fullcite{TPO2}~\cite{TPO2}
	\item \fullcite{TPOIRIM}~\cite{TPOIRIM}
	\item \fullcite{TPO3}~\cite{TPO3}
\end{enumerate}

\subsection{Journal Articles Under Review (Full Peer Review)}
\begin{enumerate}[label={[j.\arabic*]}]
	\item \fullcite{LaserJournal}~\cite{LaserJournal}
	\item \fullcite{LaserRal}~\cite{LaserRal}
\end{enumerate}

\subsection{Conference Articles Under Review (Full Peer Review)}
\begin{enumerate}[label={[c.\arabic*]}]
	\item \fullcite{TPO4}~\cite{TPO4}
\end{enumerate}

\subsection{Articles To Be Submitted}
\begin{enumerate}[label={[s.\arabic*]}]
	\item \fullcite{Bertoni2023}~\cite{Bertoni2023}
\end{enumerate}

\end{refsection}

\chapter{State Of the Art}\label{chap:soa}

\lettrine{T}{his} thesis is dedicated to advancing the field of human-robot interaction interfaces, aiming to empower users to efficiently and effortlessly operate robotic systems.
The exploration of such interfaces is an ever-evolving area of interest in robotics, which has never ceased to be particularly intriguing for researchers. 

In this chapter, we will explore the relevant previous works within this realm, with the focus on those that tackled the main challenges central to this thesis: the exploration of intuitive methods for interacting with the robot (Section~\ref{sec:soa:hri}), the development of haptic feedback solutions to convey situational awareness to the user (Section~\ref{sec:soa:haptic}), and the incorporation of different levels of robot autonomy to enhance the overall human-robot interaction (Section~\ref{sec:soa:auto}).

\section{Intuitive Human-Robot Interaction Interfaces}\label{sec:soa:hri}

The growing complexity of robots and the effective exploitation of their enhanced capabilities can be tackled by the development of novel intuitive human-robot interfaces. 
These interfaces must minimize the operators' learning curve and enable a seamless communication with the robot to augment the effectiveness of operating a robot during the task execution~\cite{Gaofeng2023}.
The development of an interface that is easy-to-use is important to let the operator concentrate on the task instead on how to communicate his/her intentions to the robot. Traditional interaction means, like keyboards, joysticks, and teach pendants, are not suitable for this purpose due to the cognitive effort required from the operator to deal with a not-always-intuitive mapping between the input buttons and the robot behavior~\cite{VILLANI2018}.
Hence, modern interfaces have been conceived with the aim to overcome these issues by exploring more natural communication means, like directly employing the user's body movements.

\noindent In what follows, various works in these directions are presented.

\subsection{Leader-Follower Systems for Controlling Remote Robots}\label{sec:soa:leadFolInt}

To control robots, especially if they present a high number of \acrfull{dof}, conventional devices such as joysticks are often inadequate to manage their diverse motion capabilities. 
To face this challenge, many researchers have investigated human-robot interfaces based on a leader-follower scheme, where the leader is not a simple joystick but a robotic system with which that the operator physically interacts to command the remote follower robot~\cite{Hokayem2006}.

For example, interfaces can be composed by leader devices that are designed almost identical to the follower robot being controlled. 
In \cite{Takubo2006}, a humanoid robot is teleoperated by manipulating its smaller version, to apply destination poses rendering the feeling of manipulating a doll. 
A similar approach is explored in~\cite{Mae2017}, where a spider-like legged robot is controlled manipulating its mini-version, incorporating some robot autonomy features to maintain the robot's body balance.
With these interfaces, operators can generate robot motions in a way that resemble a physical guiding/teaching task but without physically interacting with the follower robot. Operating at a distance avoids potential safety issues related to the robot's strength or to the dangerous remote location where the robot is. 
While these methods are in general intuitive, since the user manipulates a similar robot which motions are replicated by the follower robot, it is evident that they lack flexibility, since their design is highly specific to the particular robot in use.

Pursuing a different direction, other works concentrate on the design of leader systems that may present different kinematics with respect to the robot to be controlled. 
For instance, the HUG interface~\cite{abi2018} consists of two commercial robotic manipulators serving as the user's input interface and the haptic feedback channel to command the humanoid robots \acrshort{toro}~\cite{Englsberger2014} and Justin~\cite{Vogel2021}.
Other examples employ exoskeleton-like devices to control various robots~\cite{Rebelo2012, Jo2013}. 
Due to the kinematic differences, a mapping is required to correlate the motions of the leader device to the motions of the follower robot, increasing the complexity of the interface. Furthermore, once established, this mapping is not always straightforward for the operator to understand, especially when the leader and the follower systems have a high amount of \acrshort{dof} and significant different structures.  

In general, these types of interfaces empower users to leverage the diverse motion capabilities of the robot, even allowing the control of very highly-redundant robots like bipedal humanoids~\cite{dafarra2022, Schwarz2023}. 
The development of very sophisticated and intriguing interfaces~\cite{Ishiguro2020} offers an immersive telepresence experience~\cite{Darvish23}. Nevertheless, these solutions usually burden the operator with cumbersome and expensive systems, compromising the intuitiveness and simplicity of the human-robot interface, some of the key objectives of all the contributions presented in this thesis.

\subsection{Body Motions and Gestures as Inputs for Human-Robot Interfaces}\label{sec:soa:bodyTrack}

The human body has a lot of \acrshort{dof} with which people communicate every day. This has inspired researchers to harness the potential and naturalness of body movements for controlling robots, resulting in the development of innovative \acrfull{bomi}~\cite{Casadio2012}.

Users' body motions and gestures have been tracked and mapped to specific robot commands through exploring a range of technologies like \acrfull{imu}~\cite{Stanton2012, Noccaro2017, Kourosh2019, Wu2019, Gasper2021}, \acrfull{emg}~\cite{Ajoudani2012,Hassan2020,Pagounis2020}, optical motion capture~\cite{nunez2018}, human skeleton tracking from camera images~\cite{Parnell2018}, and \acrfull{ar}~\cite{Tzafestas2006, Whitney2019, Zhou2020, Makhataeva2020, Fusaro23, Junling2023}.

One of the technologies to track the human body is an \acrshort{imu}-based suit worn by the operator. This kind of interface is flexible, since it does not limit the user in a confined space, and, usually, lightweight and comfortable to wear. 
Furthermore, it is modular: only the required amount of human body parts can be tracked, from a single arm~\cite{Noccaro2017} to the entire body to allow controlling all the aspects of highly-redundant robots~\cite{Stanton2012, Kourosh2019}. 
The work in~\cite{Wu2019} exploits a full body \acrshort{imu}-based suit combined with a human center of pressure model and a teleimpedance interface to control the locomotion and manipulation actions.
Teleimpedance control enriches the command sent to the remote robot by combing the master’s estimated position and the stiffness references obtained through \acrshort{emg} devices~\cite{Ajoudani2012}. With teleimpedance, an additional command channel is enabled to directly command the impedance of the follower robot in real time. A recent survey explores in depth the concept of teleimpedance and the related works~\cite{Peternel2022}.

Regarding myoelectric control with \acrshort{emg} devices, \cite{Ison2015} presents an in-depth study where it is demonstrated how individuals develop a muscle synergy that exhibits increased performance over days. Furthermore, these synergies are maintained after one week, showcasing the potential of such technology for efficient robot control that can generalize to new tasks.

Another interesting approach is the employment of \acrshort{ar} interfaces. The recent advancements in this field not only led to the commercialization of such devices for the general public, but they have also facilitated their application in robotics. 
The use of a visor enhances the user situation awareness, enabling to overlap the \acrfull{gui} to the real world to command and monitor a robot.
For example, \cite{Fusaro23} focuses on the task allocation of mixed human-robot teams, intuitively assisting and coordinating the human agent with an \acrshort{ar} device. Multiple surveys focus on the advantages of these technologies~\cite{Tzafestas2006, Makhataeva2020, Junling2023}.

A very important tool that humans possess are their hands. Many works have integrated various types of inputs, as the ones presented before, with hand gestures.
This combination expands the range of inputs available to the user, in a way that resembles the natural communication that people use in their everyday interactions. 
Nevertheless, wearable interfaces for tracking hands can be expensive, especially if multiple fingers' movements are tracked~\cite{Fang2015, Weber2016}. 
In contrast, visual methods can be employed by utilizing a camera pointed at the user's hand~\cite{Mi2016, Maric2016, Muratore2023}, but this approach limits the versatility as the user is constrained to stay in the field of view of the camera.
Interested readers can refer to the survey in \cite{Xia2019} for an in-depth presentation of hand motion inputs methods employed for interacting with robots.

Body tracking is of particular relevance for the concept of TelePhysicalOperation introduced in Chapter~\ref{chap:TPO} and discussed throughout the whole Part~\ref{part:one}. Indirectly, body gestures are also employed in the laser-guided interfaces of Part~\ref{part:two}.

\subsection{Physical Human-Robot Interaction for Robot Control}\label{sec:soa:phri}

While significant advancements have been made in human-robot interfaces, controlling the robot from a distance remains a challenging task. 
In contrast, a highly intuitive approach to control the robot is by physically interacting with it. By applying forces on different parts of the robot's body, it is possible, for instance, to drive a manipulator along the desired path avoiding static obstacles, or to teach it how to execute a manipulation task.

Physical human-robot interaction interfaces offer operators a method to collaborate with robots to accomplish a particular task~\cite{Akella1999,Kruger2009}. 
In~\cite{Erden2010}, a physically interactive control scheme is designed. The scheme recognizes the applied forces from the user and switches to an interactive control modality accordingly, allowing the user to guide the robot motions by physical contacts. 
\cite{Ficuciello2014} develops a physical interaction architecture which incorporate a stable impedance controller that exploit robot's redundancy to ensure a decoupled apparent inertia at the end-effector.
In \cite{Cherubini2016}, direct physical contacts between the human and the robot are investigated for a series of  cooperative assembly task. The approach alternates passive and active robot behaviors to reduce the operator workload and the risk of injuries. 

Physical interfaces have also been explored for mobile robots, as shown in~\cite{Kim2020}, where the \acrfull{moca} operates autonomously until the operator chooses to physically connect with the robot (through a mechanical admittance interface) to perform conjoined actions. The \acrshort{moca} robot is exploited again in~\cite{Lamon2020}, where haptic guidance is provided to physically drive the arm to grasp a desired object. Another example is \cite{Muratore2023}, where the mobile base of the \acrshort{relax} mobile manipulator can be driven by applying forces on the compliant low-impedance arm.

While physical interactions remain an interesting and intuitive approach to control a collaborative robot, such a direct interaction is not always feasible. There are cases in which the robot must be teleoperated from a remote location, or situations where a local robot can not be touched due to safety cons related to the robot strength. Additionally, issues related to the human perception of safety may emerge due to the physical interaction with the robot~\cite{Rubagotti2022}.

Physical human-robot interaction is the inspiration of the TelePhysicalOperation paradigm presented in Chapter~\ref{chap:TPO}.

\subsection{Vision and Laser-guided Human-Robot Interfaces}\label{sec:soa:laser}

There are many cases in which the operator's objective is to drive the robot toward specific points of interest for inspection or interaction~\cite{VILLANI2018}.
Instead of commanding the robot's motion to reach these locations, \enquote{selecting} points in the environment and allowing the robot to autonomously reach them is a very intuitive and effortless way to guide the robot toward the desired task. This mode of operation requires a certain robot's autonomy, a topic explored later in Section~\ref{sec:soa:auto}. Instead, in this section, the focus is about how to provide functionalities for the human-robot interfaces that permit users to input such targets.

Some solutions explore this idea with human robot interfaces based on vision guidance. For example, in \cite{Solvang08}, the robot can recognize the marks manually made by the human operator in a workpiece to perform some manufacturing jobs. 
Another work proposes an architecture to program the paths of industrial robots by using a commercial laser tracker, which provides a visual reference for the user and transmits the pose of the pointed location to the robot~\cite{Szybicki22}. 

Regarding laser-guided interfaces, more flexible solutions recognize the laser spot projected by a laser emitter by processing camera images. 
For example, a laser gesture interface is developed in \cite{Ishii2009}, allowing the user to utilize a simple laser pointer to select clusters of objects and issue commands to the mobile robot. The laser strokes are projected on the ground, and all operations involve planar robot motions which perform various actions on the object collections by pushing them. To detect the laser projections, a camera, fixed in the environment, provides 2D images that are processed with a narrow band-pass filter that passes only the laser light.

More elaborated perception solutions considers the detection of the 3D position of the laser projection. Usually, the first step is to detect the laser spot in the pixel coordinates of a 2D image gathered by an RGB camera~\cite{Zakaria2014}. For this purpose, common computer vision algorithms can be employed, like Hough transform for the detection of the spot as a circle~\cite{Krstinic2014} and motion analysis through Lucas–Kanade optical flow estimation~\cite{Krstinic2014, Wang2020} together with Kalman filter~\cite{Wang2020}.
To provide a goal to the robot, the detected spot in pixel coordinates must be converted into a 3D pose, which can be done by using stereo-vision techniques based on multiple \acrshort{rgb} cameras or a single \acrshort{rgbd} camera. 
The detection ability is provided to the robot such that objects of interest can be selected with the laser, creating, for example, the \enquote{clickable} world of \cite{Kemp2008, Nguyen2008}. Here, the mobile manipulator employs vision sensors (laser range finder and eye-in-hand camera) to detect the laser directed at objects to be grasped.
In~\cite{Yaxin2018}, a white disk piece is attached to objects functioning as the area where the laser spot must be projected to be recognized.

Although the experiments have demonstrated the validity of these methods for selecting objects, they may not be suitable for robust and fast continuous tracking. Typically, they are utilized to select the object once, extract the pose, and then command the robot toward the location of interest. Usually the processes can take some time, especially when extracting the laser position from a sequence of images instead of a single one.

Regarding a more flexible laser tracking, intended as a fast laser detection that can continuously follow a moving laser spot, in~\cite{Cardenas2020} the robot follows the laser without any specific object to be pointed, but the setup is very limited since the robot consists in a simple pan and tilt support for the stereo camera. Similarly to the previously mentioned works, the laser is detected by processing the images with classical operations, in this case beginning with filtering the \acrshort{hsv} values of the images.
In~\cite{Sprute2019}, a mobile base robot is able to follow the laser spot projected on the ground to delimit no-way areas. 

The idea of using a laser pointer is simple but effective and very intuitive when correctly implemented. An issue with the presented works is that the techniques employed to detect/track the laser are based on classical computer vision algorithms. 
Such methods may be not reliable on different background surfaces, since external factors like light conditions, and the different surfaces where the laser is projected, may influence the detection and may require a fine-tuning of detection's parameters, like the \acrshort{hsv} thresholds utilized to filter out specific colors' wavelength.
To overcome this, recent works have begun exploring neural network models to track the laser spot, like for the assistive human-robot interface of \cite{Zhong2019}, improved in the subsequent works of the same authors~\cite{Liu2021, Liu2023}. In general, since the simplicity and intuitiveness of laser-based interfaces, they have been explored in a wide range of assistive robotic works, a field discussed later in Section~\ref{chap2:soa:assistive}.

This section is related to the laser-guided human-robot interfaces discussed in Part~\ref{part:two}.

\subsection{Multimodal Human-Robot Interfaces}

Multimodal human-robot interfaces encompass systems that leverage multiple modalities for interacting with the robot. Such interfaces exploit a combination of different methods, including those highlighted in the previous sections, but exploring also others, like \acrlong{gui}s, speech recognition~\cite{Maurtua2017}, eye tracking~\cite{Kim2014}, emotion understanding~\cite{Lagomarsino2022} and free-form dialog~\cite{chatgpt}. 
With multimodal interfaces, by combining multiple input possibilities, the human-robot interaction can be evidently improved since the operator can choose the most appropriate way to interact with the robot based on the task requirements, on the nature of the command, and the capabilities of the robot itself.
 
The benefits of such interfaces are tangible in real-world applications. For example, a multimodal and modular interface is utilized for operations at \acrshort{cern}'s accelerators complex. The system allows controlling single and multiple mobile manipulators of different capabilities with varying levels of robot autonomy~\cite{Lunghi2019}.

A common input method employed in multimodal interfaces are vocal commands, which leverage on the natural affinity humans have for verbal communication~\cite{Mavridis2015}. While speech interpretation alone may suffice in certain contexts, like assistive and social robotics~\cite{Vargas2021}, in other scenarios, like industrial human-robot collaboration, vocal commands may be not sufficient or feasible at every stage of a task, but yet they prove valuable when integrated with other input modalities. 
For example, in \cite{Liu2020}, haptics, body motions and voice inputs are combined in a multimodal interface to control function blocks that trigger the execution of tasks. 
In \cite{Yongda2018}, an interface for teleoperating a manipulator integrates speech control with gesture control, necessary when the speech alone can not clearly describe part of the task, such as indicating a correct azimuth.
In \cite{Liu2018}, voice, hand motion and body posture data are collected and recognized using three different deep learning networks, and then merged together to control an industrial manipulator.

Employing a combination of inputs allows compensating the limits inherent in each source. For example, \acrfull{bci} can have a low accuracy and a limited number of commands options, but these drawbacks can be mitigated by incorporating other input modes like eye tracking, \acrshort{emg} signals, and body tracking~\cite{Dong2021}. 
However, a notable challenge, particularly evident with \acrshort{bci}s, is that an already complex interface is further elaborated resulting in a potentially very intricate overall system. 
Addressing the issues of high cost and bulkiness associated with some \acrshort{bci} devices, \cite{Kim2014} integrates a low-cost eye tracking device with \acrfull{eeg}-based \acrshort{bci}, a technology which is typically less invasive and less expensive.

In conclusion, the integration of multiple input modalities is crucial to provide a flexible interface to users and to establish an efficient human-robot interaction. Nevertheless, it is equally important to consider the additional complexity introduced by the diverse range of input possibilities, which may affect the usability and applicability of such interfaces. These considerations are taken into account in this thesis's contributions toward novel human-robot interfaces.
In-depth discussions about multimodal human-robot interfaces can be found in~\cite{Kirchner2019} and \cite{Hang2023}.

\subsection{Assistive Human-Robot Interfaces}\label{chap2:soa:assistive} 

A particular field where the intuitiveness of the human-robot interface is of paramount importance is the assistive field. Only by developing easy-to-use interfaces, assistive robots can really help people who need assistance in home-care scenarios, allowing users to interact with these robots for \acrfull{adl} tasks.  

Together with the development of assistive robots, substantial efforts have been spent in providing human-robot interfaces to enable individuals with disabilities to control these systems~\cite{Kyrarini2021}.
An example of such platforms is the \acrshort{friend} system, which integrates a wheelchair with a robotic manipulator, commanded with a multimodal interface that includes various input possibilities like speech, \acrshort{bci}, chin and head movements~\cite{Graser2013}. Another arm-equipped wheelchair is the \acrshort{edan} platform, controlled by the user with \acrshort{emg} signals in combination with a head-switch~\cite{Vogel2020, Vogel2021}. 

Recognizing that each daily activity presents unique challenges, other researchers have dedicated their efforts in developing task-specific interfaces, for example for assistive-feeding~\cite{Gordon2020, Bhattacharjee2020} and dressing~\cite{Zhang2019}. 
In~\cite{Gordon2020}, the assistive-feeding system is based on a learning framework to plan manipulation strategies to face previously unseen objects. In~\cite{Bhattacharjee2020}, a study about the autonomy level of the robot is performed, to explore the user preferences in the feeding task.

Preliminary investigations have also addressed more challenging activities. For instance, in~\cite{Oladayo2019}, simulated experiments are conducted to tackle the task of beard shaving. In~\cite{Hughes2021}, the research focuses on demonstrating hair detangling on wigs of different curliness. In \cite{Bilyea2023}, the authors modeled the interaction forces between humans and robotic manipulators when utilizing conventional tools commonly employed in various types of daily activities.

One key challenge in developing assistive human-robot interfaces is how to comply with the limited mobility of impaired users, maintaining as much as possible the intuitiveness. For example, interfaces based on handheld joysticks~\cite{Wang2012} or arm gestures~\cite{Esposito2021} may not be suitable for certain disabilities which affect the upper limbs. Similarly, verbal communication~\cite{Poirier2019} may not be feasible due to the user condition. 
To address these challenges, an option is to adapt commercial assistive technologies, such as sip-and-puff devices, for controlling the robot~\cite{Jain2019}. However, these devices have very limited control space dimensions (i.e., two inputs associated to sipping and puffing actions), thus requiring the user to frequently switch between several control modes increasing the cognitive workload and slowing down the task execution.

Other approaches explore non-conventional inputs to interact with the robot, like head gestures tracked by a camera mounted on the head of the person. For instance, in \cite{Kyrarini2019}, the head gestures are used to navigate a state machine that maps the various manipulator abilities (i.e., 6D Cartesian movements and opening/closing the gripper). 
Even if it is not always necessary to command the robot in this way since the interface allows the robot to learn the movements and to reproduce them autonomously, adapting to different conditions, the head gestures and the state machine navigation may lack intuitiveness.
Another solution entails utilizing a screen where a virtual keyboard is operated with a virtual cursor controlled through non-traditional means, such as head movements~\cite{Rudigkeit2020}. In this work, also ergonomics plays and important role since continuous head movements may cause heavy fatigue on the user. 

Similarly to head movements, eyes movements can also be utilized to control a virtual cursor present on a screen~\cite{Sunny2021} or on an \acrshort{ar} device~\cite{Sharma2022}. Another non-conventional input method is the tongue, utilized to press specific areas of a device installed in the mouth~\cite{Mohammadi2021}. 
Even if these interfaces have shown their advantages, they may be invasive and not inherently intuitive. They may pose challenges to users in terms of learning and understanding the mappings between their inputs and desired actions.

Additionally, numerous studies have focused on \acrlong{bci}~\cite{Gandhi2015}, which offer the potential to directly interpret the user intentions. However, the complexity and the cost associated with these brain interfaces pose significant challenges in their development and adoption in real-life scenarios.

Other works have explored laser pointers, since its ease of use in many scenarios as covered in the previous Section~\ref{sec:soa:laser}. Indeed, this kind of input has shown promising results when utilized for assistive human-robot interfaces, since it is inherently intuitive, and can be comfortably worn on the user head to accommodate people with upper limbs impairments. For example, the interface from \cite{Nguyen2008} has been employed for robotic assistance in \cite{Choi2008}. Here, an ear-worn laser pointer is used to point objects to be picked by a mobile manipulator. The laser spot is detected by filtering the laser color wavelength in the images coming from a camera.
In \cite{Gualtieri2017}, the laser guides a wheelchair equipped with a manipulator to reach and pick objects. While the laser is pulsating, the spot is detected by looking in a sequence of images for areas with changes in the intensity. This increases the robustness of the detection, but also augment the necessary detection time. 
Another wheelchair system is designed in~\cite{Wilkinson2021}, where a projector illuminates the object selected by the laser, to allow the user to confirm that the proper item has been correctly recognized before grasping it. A second laser is employed to confirm the choice. The two laser spots are detected by looking for two high-value areas in infrared images.
In \cite{Zhong2019}, a neural network is utilized to recognize the pose of objects and to detect the laser projected on them. As before, the objective is to grasp the selected object with the arm of a wheelchair manipulator. In another work, the same authors explore a handheld laser pointer by flashing it on different objects to command specific manipulating actions~\cite{Liu2021}.

The assistive scenario, faced with a laser-based human-robot interface, is taken into account in the work presented in Chapter~\ref{chap:Laser2}.

\section{Situation Awareness through Haptic Feedback in Human Robot Interfaces}\label{sec:soa:haptic}

An important aspect in human-robot interactions is to provide operators with a situational awareness of the commands issued, the robot's status, and the events unfolding at the robot's location~\cite{Endsley2016}. 
In this context, substantial efforts have been dedicated in developing interfaces that can transmit a heterogeneous range of sensory information to the user, which is often essential to work with high-\acrshort{dof} robots~\cite{Darvish23}.
Even when full telepresence immersion is not necessary, having a subset of such types of feedback is useful. Indeed, direct observation of the robot or the acquisition of information through a screen does not always guarantee sufficient insights about the task or the robot state~\cite{Moniruzzaman2022}.
Consequently, several works have coupled the remote robot control with a sensory feedback channel to supplement the visual domain, encompassing an array of stimuli, like haptic cues to leverage the human sense of touch~\cite{Dargahi2004, Billard2019, Youssef2023}.  

Various tactile devices have been proposed for diverse robotic systems and applications, like wheeled robots~\cite{Zhao2023}, surgical manipulators~\cite{pacchierotti2015cutaneous}, underwater vehicles~\cite{XIA2023}, drones~\cite{Macchini2020}, and humanoids~\cite{Schwarz2021,dafarra2022,baek2022}.

Haptic devices can deliver information in different ways to various operator's body parts. For example, in~\cite{Brygo2014}, a vibrotactile belt alerts the operator about the loss of balance of a bipedal robot.  In~\cite{baek2022}, an elaborated grounded haptic structure pulls the user's body when the robot encounters an obstacle. Although these solutions offer an engaging user experience, they are bulky and impractical in some scenarios.

Other devices exploit the sensing ability of the human hand~\cite{pacchierotti2017wearable}. In \cite{Pacchierotti2015}, a 3-\acrshort{dof} cutaneous haptic device provides feedback at the fingertip, with the possibility to attach such device to commercial grounded haptic interfaces. 
Several works explore wearable devices, overcoming the limitation of the grounded interfaces. Such devices can deliver different cutaneous cues (e.g., skin stretch, vibration, temperature), hence they are capable of transmitting diverse information to the user,  while remaining unobtrusive and flexible.
In~\cite{Macchini2020}, a glove device is developed to return vibrations in different locations of the user's hand corresponding to the six axes of motion of a quadrotor. 
Another haptic glove is developed in~\cite{Coppola2022}, which provides vibration stimuli on the fingertips to enable a tactile sense during object manipulation with the remote robot.
\cite{Katayama2020} establishes a \textit{shared haptic perception} connection between the human operator and the collaborative robot. The interface utilizes a wearable skin vibration finger sensor to let the robot recognize human actions and inform the human about the recognition of such actions.

In other studies, wearable devices are worn on the user's arm.
In~\cite{BAI2019}, vibrotactile patterns delivered on the user's hand and arm aid in guiding the robot avoiding obstacles. Similarly, a vibrotactile bracelet is developed in \cite{Bimbo2017}, to alert the user about the collision of the manipulator with the environment.

Nevertheless, it is still a challenge to develop a wearable haptic interface that monitor the motions of highly-redundant robots while maintaining a simple \acrshort{bomi}. 
Some solutions offer extensive inputs and feedback possibilities, but are cumbersome to wear~\cite{Brygo2014,Schwarz2021,dafarra2022,baek2022}; other approaches limit the flexibility with grounded haptic devices~\cite{Pacchierotti2015, pacchierotti2015cutaneous, BAI2019, Correa2022, Cheng2022, baek2022}, or with an external tracking system~\cite{nunez2018}; development of unobtrusive wearable interfaces are usually adopted to control robots with a low number of \acrshort{dof}~\cite{Vogel2011,Bimbo2017,Gasper2021,Katayama2020}.

Furthermore, an open challenge is how to avoid complicating the overall human-robot interface, while still providing relevant information. Indeed, accounting for multiple inputs from the user and feeding back different kinds of stimuli may easily compromise ergonomics and become impractical. 

Wearable haptic feedback is the focus of the work discussed in Chapter~\ref{chap:TPOH}.
\section{Robot Autonomy in Human-Robot Interfaces}\label{sec:soa:auto}

As the complexity of robotic systems has increased, improving their capabilities, additional challenges for managing such highly-redundant systems have been presented. 
While the development of intuitive input interfaces (as discussed in Section~\ref{sec:soa:hri}) addresses some of these challenges by enabling operators to easily control the complex robot capabilities, micromanaging every aspect of the human-robot interaction can become tedious for the operator and potentially escalate the difficulty of the task.

This issue is addressed by more intelligent human-robot interfaces, which assist the operator in controlling robots through the implementation of a certain level of robot autonomy.
These interfaces generally explore the concepts of \textit{shared control} or \textit{shared autonomy}~\cite{Selvaggio2021}, utilized in various applications like manufacturing~\cite{Pichler2017}, human-robot collaboration~\cite{Music2017}, undersea exploration~\cite{Brantner2021}, space~\cite{Schmaus2023}, surgical robotics~\cite{Payne2021}, and assistive robotics~\cite{Bengtson2020, Bustamante2022}. 

A common approach involves the incorporation of robot autonomy features that modify the operator's commands according to the task's constraints. For example, while the operator is commanding a trajectory to the robot, the input can be overridden to navigate around obstacles~\cite{Masone2018}, to avoid unsafe regions~\cite{broad2019}, or to keep the manipulator's end-effector pointed toward the object to be grasped~\cite{Farraj2016}.

When dealing with highly-redundant robots or a team of robots, the automatic generation of certain robot(s) motions become essential, as manually controlling all the \acrshort{dof} of the robot(s) can be very difficult, if not impossible.
For example, in \cite{Song2016}, an omnidirectional mobile manipulator is controlled combining user's inputs and robot's autonomous behaviors; while the user commands a specific trajectory, the robot, exploiting the omnidirectional base, autonomously maintains the heading orientation toward the target position.
With legged robots, as discussed in Chapter~\ref{sec:soa:leadFolInt}, sophisticated exoskeletons can be employed to retarget the user's body movements to the robot ones. If such complex interfaces are not a feasible option, it is the responsibility of the robot to autonomously generate walking motions from simpler user commands like directions and velocities~\cite{Penco2019}.
In \cite{Franchi2012}, a heterogeneous group of mobile robots navigates in cluttered environments avoiding collisions, following a leader robot which is the only one subject to the operator control.

Some approaches leverage on the concept of task priority~\cite{Nakamura1987}, which ensures that high-priority control objectives are satisfied without being affected by low-priority ones. Hence, the user's commands can be considered an objective to be satisfied alongside other objectives of different priorities. These additional objectives may include maintaining a good arm manipulability~\cite{Faroni2016}, avoiding joint limits~\cite{ORTENZI2018}, keeping the yaw attitude of floating platforms~\cite{Roger2021}, considering physical constraints of humanoids~\cite{Sentis2005}, and managing a cooperation of multiple robots~\cite{Simetti2019}.

Robot autonomy is necessary when the human-robot interface adopts a supervised control approach~\cite{Selvaggio2021}. Supervisory control reduces the operator's intervention to the specification of the essential task objectives, leveraging on the autonomy of the robot to plan and execute the necessary motions. 
An example is found in \cite{Cheong2021}, where the Pholus robot, a variant of the CENTAURO robot (Section~\ref{sec:intro:centauro}), can detect objects to manipulate and highlight them to the human operator. Once the operator selects the object and the desired action, the robot autonomously plans and executes the mobile manipulation motions necessary to fulfill the given instructions. Another example~\cite{Schmaus2023} proposes a system leveraging Action Templates to define the knowledge of handling objects, offering to the operator supervisory modes to facilitate the teleoperation of the robot.

Another aspect involves providing users with the flexibility to determine the extent of their intervention in the task and, correspondingly, the degree of robot autonomy. Studies in the field of assistive robotics demonstrate that too much robot autonomy is not always the best option~\cite{Kim2012, Bhattacharjee2020}. Indeed, the level of autonomy should be tailored not only to the specific task requirements but also to user preferences; the adaptation of the level of robot autonomy based on the human confidence in performing the tasks is a key challenge for future research~\cite{Music2017}. 
Furthermore, in delicate applications, like surgical robotics, the level of autonomy is a cause of concern due to ethical and legal considerations~\cite{Yang2017}.

Studies about shared control and shared autonomy features in human-robot interfaces are numerous. In the following sections, they are discussed previous research that exploit techniques similar to the ones employed in the works presented in this thesis.

\subsection{Manipulability Measure for Regulating Robot Motions}\label{sec:soa:manip}
While controlling a robot, analyzing the manipulability metric for the generation of robot motions is useful to maintain a good arm dexterity for example to manipulate objects efficiently. Anyway, for an operator, assessing the manipulability level of the arm is not always straightforward and can augment his/her burden. Therefore, the manipulability measure (or a similar metric), is typically considered automatically by the human-robot interface.

There are different definitions of manipulability~\cite{Patel2015}, with one of the most common characterizing it as a measure of the transmission ratio from the joint velocity to the end-effector velocity~\cite{Yoshi1985, Chiu1988}. 
Initially analyzed for single-chain robotic manipulators~\cite{Yoshi1984}, the manipulability has been investigated and extended for the case of dual-arm robots~\cite{Lee1989}, closed-loop chain robots~\cite{Bicchi2000}, leader-follower teleoperated systems~\cite{Torabi2018}, and mobile manipulators~\cite{Gardner2000, Bayle2003}. Having a good manipulability level is related to the avoidance of kinematic singularities, ensuring smooth motions of the end-effector. 

Analyzing a \textit{global} manipulability measure (or a similar metric that takes into account the kinematic singularities and/or joint limits) is valuable for the design of manipulators, since it provides a quantitative measure of their dexterity~\cite{Angeles2016}. 
For example, in \cite{Torabi2018}, challenges encountered in a leader-follower surgical teleoperation system are addressed. The leader system, operated by the user, often differs kinematically from the follower robot, making it challenging to convey information about the robot's dexterity to the leader system and, consequently, to the user. To tackle this issue, the manipulability index is extended to a \enquote{teleoperation manipulability index}, which considers singularities and joints limits of both leader and follower systems, to improve the design of the leader system.
In~\cite{Gardner2000}, manipulability is extended for a mobile manipulator, and it is investigated to optimize the placement of the $3$-\acrshort{dof} arm on the platform composed by two independent drive wheels. 

While controlling a robot, a \textit{local} manipulability measure is typically considered to avoid regions where the end-effector would operate with more difficulties, because of the vicinity to kinematic singularities.
Often, whole-body controllers consider the manipulability within a task priority strategy~\cite{Nakamura1987,Siciliano1991,Mansard2009}. This approach allows controlling the robot while considering multiple objectives, including those related to dexterity, like avoiding singularities~\cite{Dietrich2012} and preventing joint limits~\cite{Sentis2006}. 

In other works dealing with mobile manipulators, the manipulability is employed not only to optimize the end-effector pose, but also to generate reference velocities for the mobile base. 
The utilization of manipulability to generate mobile base velocities was first studied in the pioneering work~\cite{Yamamoto1992}, where simulation results show how mobile base velocities can be formulated to keep a \acrshort{puma}-like $3$-\acrshort{dof} arm in a region of good manipulability.  
Similar to \cite{Gardner2000}, other works propose new manipulability measures for mobile base robots, but, in these cases, the focus is on the control of the robot rather than on the design of the system.
In \cite{Bayle2003}, the addressed challenge is that the arm's manipulability can be low even if the extended manipulability of the entire mobile manipulator is high. To cope with this issue, a weight is introduced to give more importance to the fixed arm manipulability over to the mobile arm one (or vice versa), depending on the manipulation task executed. 
A similar approach is followed in \cite{Leoro2021}, where, instead, the product of the two different manipulability measures (fixed arm and mobile arm) is used. 
The work in \cite{Zhang2016} proposes a formulation based on quadratic programming with joint limits as a constraint and the manipulability as an objective to maximize.
In \cite{Wang2016}, the problem is extended to a dual-arm mobile platform, incorporating a \textit{virtual kinematic chain} to specify the common motion of the two arms. 

The manipulability is the basis of the shared locomanipulation interface introduced in Chapter~\ref{chap:tpoAuto}.

\subsection{Estimation of Parameters of Unknown Objects}\label{}

When in front of unknown objects, humans can rely on a very extraordinary sensor system (i.e.\ visual information combined with the sense of touch) to comprehend the objects' properties and hence determine the best strategy to manipulate them. For example, when transporting an object, people can adjust the grasping forces based on the tactile information perceived.
In contrast, for robotic systems, handling unknown objects is still an open challenge, related to the limited sensory system~\cite{Francomano2013} and reasoning capacity that are still far from the human capabilities. Even when a robot is teleoperated, conveying information about the object to the user requires the robot to first gather relevant information by estimating the object parameters. 

The problem of estimating the object parameters, such as inertia and mass, has been tackled in different ways by past works. 
In a survey~\cite{MAVRAKIS2020}, the authors  categorize different techniques into three main groups: (1) purely visual methods (e.g., \cite{Jiajun2015}), (2) exploratory methods involving basic interactions with the object (e.g., little strikes~\cite{Krotkov1997} or tilts~\cite{Yong1999}), and (3) fixed-object methods, where the object is fixed with respect to the end-effector(s), such as when it is already grasped~\cite{Marino2017}. In the context of this thesis we will focus on the third category.

Methods from the third category typically leverage on dynamic equations to estimate the object parameters using joints motions and end-effector wrenches (estimated from joints state or measured by a sensor).
Some studies focus mostly on estimating the inertia parameters of the robot model with an additional load rigidly attached to the end-effector, without necessarily utilizing these parameters for manipulation or transportation.
An early work is \cite{Olsen1985}, which delineates the equations to estimate the dynamics of the load based on the joints state, but without conducting any experiments. A more recent study is \cite{Farsoni2018}, which identifies the load's parameters exploiting a recursive total least squares method combined with force-torque measurements at the wrist.

One practical application that relies on the estimation of object parameters is the cooperative transportation with two or more robotics manipulators. In this scenario, the estimated parameters are integrated in the dynamics models employed for generating control references for the robots. Further discussions on this application are given in the following Section~\ref{sec:soa:bimanual}.

Estimating of object's parameters is considered in the bimanual grasping and transportation interface of Chapter~\ref{chap:tpoAuto}.

\subsection{Autonomy Features for Bimanual Manipulation}\label{sec:soa:bimanual}

In some situations a certain level of robot autonomy is necessary to accomplish a task that would otherwise be very difficult to face, such as maintaining the grasping on an object transported during dual-arm transportation task~\cite{Billard2019}. 

According to the survey~\cite{SMITH2012}, which follows the taxonomy of \cite{Surdilovic2010}, dual-arm manipulation can be divided into non-coordinated manipulation and coordinated manipulation. 
In the first, the two arms perform different tasks, while in the second the arms are dedicated to the same task. 
Among the coordinated classification, bimanual manipulation is defined as physically interacting with the same object, e.g., grasping and transporting a large and heavy load, which a single arm can not handle.
A slightly different taxonomy is reported in \cite{Krebs2022}, which considers everything as \enquote{bimanual}, but maintains the coordinated/uncoordinated classification. Within the coordinated category, a task can be loosely or tightly coupled. The tightly coupled classification is further specialized into asymmetric and symmetric, with the latter involving both arms focusing on the same role on the same object (e.g., transportation).
In the context of this thesis, we will use the generic term \enquote{bimanual} to encompass coordinated symmetric manipulation of an object performed with two robotic arms.

Different control laws have been proposed to address bimanual pick-and-place tasks, such as hybrid force/position control~\cite{Uchiyama1988, Benali2018}, impedance control~\cite{Wimbock2012, Lee2014}, and admittance control~\cite{Bjerkeng2014, Tarbouriech2019, Maolin2022}.
For example, \cite{Lee2014} presents an impedance control scheme for the dual-arm system, using the relative Jacobian to treat the two arms as a single manipulator. This simplifies the definition of desired impedance and trajectories.

The bimanual task is often approached through a cooperative task space formulation~\cite{Chiacchio1996}. This formulation extends the principles of single manipulators to the combined dual-arm system, using relative and absolute variables and solving the kinematic problem with Jacobian matrices~\cite{CACCAVALE2000, Adorno2010, Lee2012, Lee2014, Park2015, JAMISOLA2015}.
In \cite{Lee2012}, the redundancy of the bimanual robot is considered optimizing a cost function inspired by the human bimanual actions~\cite{Guiard1987}. The work implements coarse motion of one arm and fine motion of the other arm, employing task-compatibility indices~\cite{Chiu1988}.
In \cite{Shahbazi2017}, the grasping constraint is exploited to decompose the set of dynamics equations into two orthogonally decoupled sets~\cite{Aghili2005}. After determining the torque in the sub-manifold compatible with the constraint, the constraint force is analytically calculated, and distributed without requiring force sensors or other approximation methods. The experiments are conducted on a simulated planar bimanual model.

Instead of using the dual-arm kinematics directly, other approaches exploit a virtual kinematic chain to specify a set of constraints that define the coordinated motion of the two arms~\cite{Likar2014, Wang2016}.

Theoretical studies in bimanual manipulation have paved the way for various practical applications.
In \cite{Laghi2018}, a dual-arm system coordinates the motions of both arms  to manipulate an object, relying on the input from a single user's arm which provides position and impedance references. 
In \cite{Laghi2022}, direct teleoperation and autonomous control are combined to help the operator in reaching and manipulating objects with one or two robotic arms. As the user guides the robot towards the target, the system comprehends his/her intention according to the commanded trajectory, and adjusts the commands to achieve a suitable grasping pose with one or two arms, depending on the size of the object.
In \cite{Ozdamar2022}, a shared autonomy telemanipulation framework enables the individual and combined control of multiple robotic arms by incorporating user arm's impedance and inputs from a joystick. 

While the aforementioned works involve robotic systems composed by fixed manipulators, in \cite{Rakita2019}, the operator moves his/her arms to control a humanoid robot which mimics these movements with a level of shared control to facilitate the completion of the task. 
Leveraging on two-hand human manipulation data, a neural network is employed to infer which bimanual action is occurring. Based on the inferred action, a corresponding assistance mode is activated, contributing to a more efficient and easier control of the bimanual task for the operator.

The development of autonomy features for teleoperation of dual-arm robot for bimanual transportation tasks is a challenge taken into account in the works of Chapter~\ref{chap:tpoAuto}.

\subsection{Behavior Tree-based Motion Planning}\label{sec:soa:bt}

To handle the autonomous robot abilities, and to plan the robot motions, a motion planner is necessary. The planner must convert the high-level instructions from the user (e.g., a pose to reach) into low-level commands (e.g., joint references), possibly considering the external conditions sensed by the robot itself (e.g., the presence of obstacles). 

To address this matter, recent works have explored the \acrfull{bt} theory as an alternative to the classical \acrfull{fsm}s~\cite{BenAri2018}.
Originally, \acrlong{bt}s were born as a tool for planning activities of autonomous agents in computer games. In the last years, they have received an increasing interest also in the robotics field, mainly due to the key characteristics of transparency, reactiveness and modularity~\cite{Iovino2022}.

With \acrshort{bt}s, the robot behavior can be modeled as a tree structure with leaves representing the robot primitive actions, and other nodes and edges specifying their ordering and dependencies. By breaking down the robot behavior into smaller, more manageable components, the BT provides a transparent and modular way to program the autonomous actions of the system.
This particular structure has shown promising results across various types of robotic systems and applications, like  humanoids~\cite{Berenz2018}, swarm robots~\cite{Jones2018}, underwater vehicles~\cite{Sprague2018}, mobile manipulators~\cite{Colledanchise2019}, aircraft systems~\cite{Castano2019}, wheel-leg platforms~\cite{Deluca2023}, and collaborative robots~\cite{Fusaro23}.

In the context of this thesis, \acrshort{bt} are utilized to plan the motions of mobile manipulators according to the operator instructions.
Related to this matter, in \cite{Zhou2019}, \acrshort{bt}s models are automatically generated to perform a picking task with a dual-arm mobile manipulator according to the user's intention processed as audio inputs. 
The work in \cite{French2019} utilizes \acrshort{bt}s to represent high-level tasks to handle primitive actions learned from human demonstration, with the objective to perform household cleanings with a mobile manipulator.
In \cite{Jiang2018, Kim2018}, a software architecture supports reactiveness to changes in the dynamic environment thanks to the BT, adapting the robot behavior during a human-following task.
\acrshort{ros}2~\cite{ROS}, the well-known middleware for robotics, with its de facto standard navigation stack, Nav2~\cite{Macenski2020}, employs configurable \acrshort{bt}s to handle the modules implemented to plan the path of mobile and surface robots.
Many other applications are summarized in the survey~\cite{Iovino2022}.

In what follows, a brief background on the basics of \acrlong{bt}s is given, as an aid to follow better the description of the implemented \acrlong{bt}-based human-robot interface of Chapter~\ref{chap:Laser1}.

\subsubsection{Background on Behavior Trees}\label{sec:soa:bttheory}
\begin{figure}[H]
	\centering
	\includegraphics[width=0.9\linewidth]{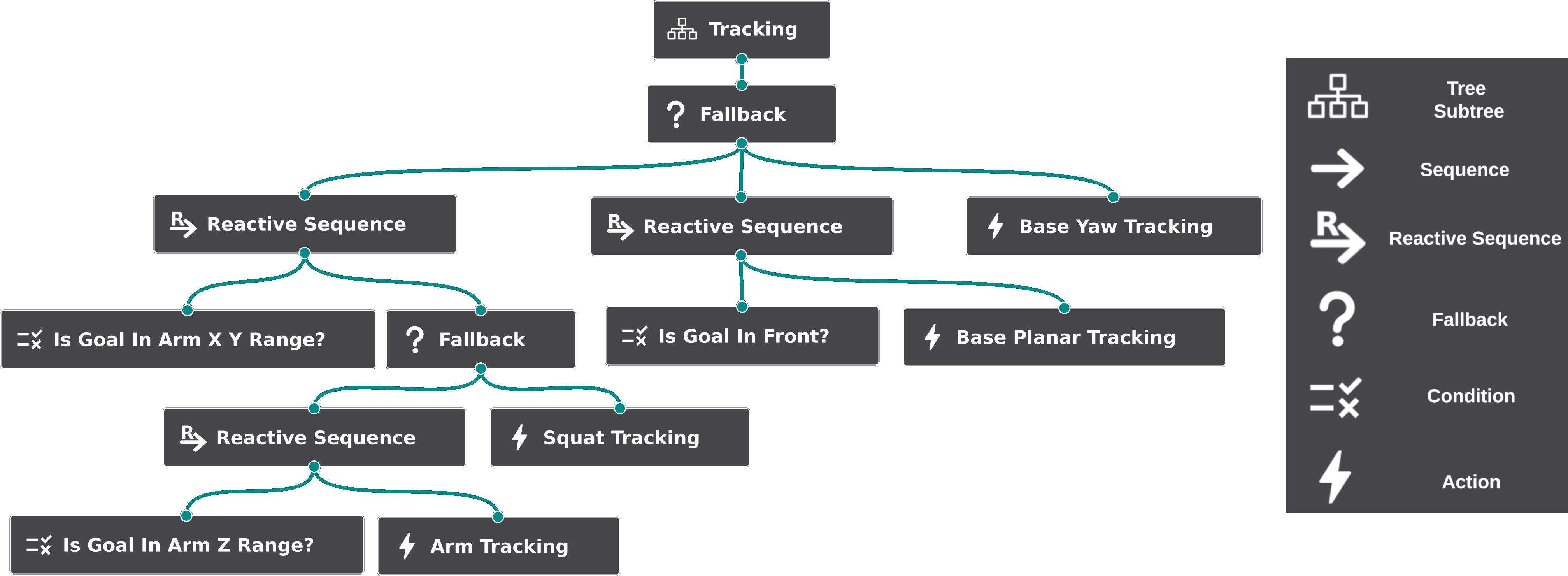}
	\caption[Example of Behavior Tree scheme]{Example of Behavior Tree scheme, utilized in the Chapter~\ref{chap:Laser1}.}
	\label{fig:soa:bt}
\end{figure}

Behavior Trees (\acrshort{bt}s) derive from the directed trees of the graph theory~\cite{Bondy2010}. They are composed by  nodes connected by edges in such a way that each \textit{child} node can have a single \textit{parent} node, but a parent can have multiple children. The direction of the \acrshort{bt} goes from the single \textit{root} node, toward the \textit{leaf} nodes, following the logic defined by the intermediate \textit{control flow} nodes. \figurename{}~\ref{fig:soa:bt} shows an example of BT taken from an application presented in Chapter~\ref{chap:Laser1}. 

The execution of a \acrshort{bt} begins at the root node with the generation of a \textit{tick} signal at a certain frequency. The \textit{tick} is propagated to the root's children which, in turn, will propagate it to their children (if any), according to the control flow. Each node is executed if, and only if, it receives the \textit{tick} signal from the parent. After receiving a tick, a child returns \texttt{success}, \texttt{failure}, or \texttt{running} to the parent.

The leaves of the \acrshort{bt} can be \textit{condition} or \textit{action} nodes. Essentially, a \textit{condition} is an \texttt{if} statement that returns only \texttt{success} of \texttt{failure}, but not \texttt{running}. An \textit{action} node usually models the execution of a command of the agent whose behavior is defined by the BT, and returns \texttt{success}, \texttt{failure}, or \texttt{running} based on its nature. 

In the classical formulation, \textit{control flow} nodes are limited to three types: \textit{sequence}, \textit{fallback} and \textit{parallel}, but with the increasing spread of \acrshort{bt} for various applications, custom control flow nodes have been added.
In the works presented in this thesis, two of the classical types (\textit{sequence} and \textit{fallback}) and \textit{reactive sequence} nodes are utilized. 

\noindent In what follows, the classical \textit{control flow} nodes and the \textit{reactive sequence} node are explained.

\begin{itemize}
	\item \textbf{\textit{Sequence}}. After receiving the tick from its parent, this node ticks the first children (graphically, the leftmost one). If the child succeeds, the \textit{sequence} ticks the following child. When all children succeed, the \textit{sequence} returns \texttt{success} as well. If one of the child fails, the \textit{sequence} returns \texttt{fail} and does not tick the children which follow the one that has failed. If one the child returns \texttt{running}, the sequence returns \texttt{running} as well, waiting for the tick of the subsequent loop to propagate it again to the same running child. If there are previous children that returned \texttt{success} before the running one, they are not ticked again in the following loops. 
	
	\item \textbf{\textit{Fallback}}. As the \textit{sequence} node, this node ticks one child at a time. Differently from the previous, as soon as one child succeeds, the \textit{fallback} returns \texttt{success}, without ticking the following children. Instead, if a child fails, the following child is ticked. Hence, the \textit{fallback} returns \texttt{failure} only when all the children until the last one fail. As for \textit{sequence}, a child returning \texttt{running} causes the \textit{fallback} to return \texttt{running} as well, ticking the same child (skipping the previous ones) in the following loop. 
	
	\item \textbf{\textit{Parallel}}. Differently from the previous control flow nodes, the \textit{parallel} node ticks all the children at the same time. The \textit{parallel} node returns \texttt{success} as soon a sufficient amount $M$ of children out of the total $N$ succeed, returns \texttt{failure} if more than $N-M$ children fail (because as a consequence $M$ \texttt{success} would be impossible), and returns \texttt{running} otherwise. 
	
	\item \textbf{\textit{Reactive Sequence}}. This node is similar to a \textit{sequence}, with the difference that, if a child returns \texttt{running}, at the next loop also the previous children are ticked again. If it happens that in the following loops one of the previous children fails, the child that had returned \texttt{running} is \enquote{reactively} halted. The node is particular useful, for example, to check for some \textit{conditions} (previous children) at every loop, while an \textit{action} (last children) is running. As soon as one of the \textit{conditions} is not met anymore, the running \textit{action} is halted.
\end{itemize}

\noindent The functions of these control flow nodes are summarized in \tablename{}~\ref{tab:soa:btFlow}.

\begin{table}[H]
	\newcommand{\specialcell}[2][c]{%
		\begin{tabular}[#1]{@{}c@{}}#2\end{tabular}}
	\newcommand{\specialRowSpace}{\vspace{15px}} %
	\centering
	\caption{Some of the control flow nodes of a \acrshort{bt} and their returning conditions.}
	\label{tab:soa:btFlow}
	\begin{tabularx}{\textwidth}{@{\hspace{-0.1\tabcolsep}} l @{\hspace{3\tabcolsep}} c @{\hspace{1.5\tabcolsep}} c @{\hspace{1.5\tabcolsep}} c @{\hspace{\tabcolsep}}}
		\toprule%
		& \centering \texttt{success} & \texttt{failure} & \texttt{running} \\
		
		\midrule
		
		\specialRowSpace
		\textbf{\textit{Sequence}} &
		\specialcell{All children \\ succeed} & 
		{\small \specialcell{One child fails}} & 
		{\small \specialcell{One child is running. At the next loop\\the running child is ticked again}}\\
		
		\specialRowSpace
		\textbf{\textit{Fallback}} & 
		\specialcell{One child \\ succeeds} & 
		{\small \specialcell{All children fail}} & 
		{\small \specialcell{One child is running. At the next loop\\the running child is ticked again}}\\
		
		\specialRowSpace
		\textbf{\textit{Parallel}} & 
		\specialcell{At least $M$ \\children succeed} & 
		{\small \specialcell{$>N-M$ \\children fail}} & 
		{\small \specialcell{In all the other cases}}\\
		
		\textbf{\specialcell{ \textit{Reactive} \\ \textit{Sequence}}}& 
		\specialcell{All children \\ succeed} & 
		{\small \specialcell{One child fails. \\Running child is halted}} & 
		{\small \specialcell{One child is running. At the next\\ loop the previous children are ticked}}\\
		
		\bottomrule
	\end{tabularx}
\end{table}

\noindent For an example about the flow of a BT, the BT in \figurename{}~\ref{fig:soa:bt} is detailed in Section~\ref{sec:laser:btflow}.

For more detailed explanations about Behavior Trees theory and their application to robotics, interested readers can refer to the survey~\cite{Iovino2022}, or the more extensive book~\cite{BTBook}.

\chapter{Characteristic and Challenges of Human-Robot Interfaces}\label{chap:chap3}

\lettrine{I}{n} the previous chapters the challenges addresses by this thesis have been outlined.
It has been emphasized that it is crucial to explore innovative human-robot interfaces to effectively harness the capabilities of modern robotic platforms.
This chapter discusses the related challenges in detail, and presents the contributions of this thesis highlighting how they respond to them.

\section{Key Features of Human-Robot Interaction}\label{sec:chap3:keys}

\begin{figure}[H]
	\centering
	\includegraphics[width=1\linewidth]{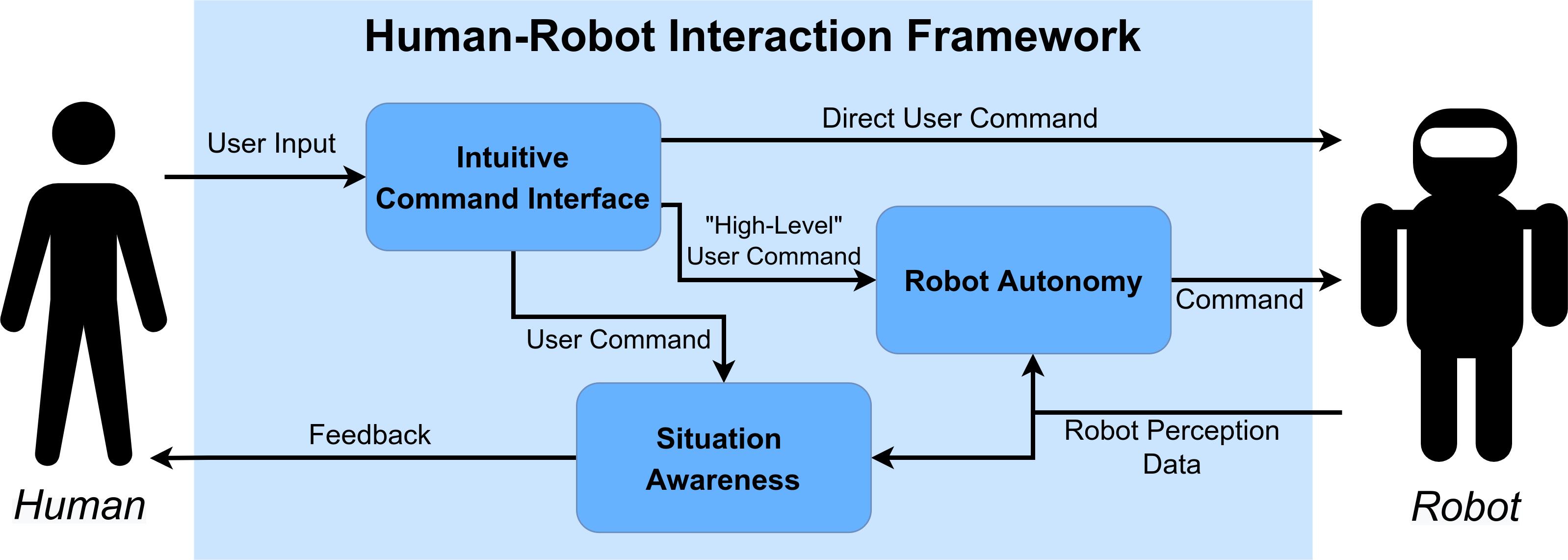}
	\caption[Human-robot interface overall scheme]{Scheme of the human-robot interaction framework highlighting the key features considered by this thesis.}
	\label{fig:wholeoutline}
\end{figure}

To facilitate the control of robotic systems, a variety of human-robot interfaces have been developed, as reported in Chapter~\ref{chap:soa}. 
It has been shown how researchers are looking for innovative solutions that prioritize the user-friendliness of the interface. The aim is to empower even non-expert users to control robots, from highly-redundant multi-limb robots to simpler manipulators.

The content of this thesis focuses on particular key challenges and features that a human-robot interface should present. By following the scheme in \figurename{}~\ref{fig:wholeoutline}, we can observe three interconnected blocks:

\begin{itemize}
	\item \textbf{Intuitive Command Interface}. The aim of this block is to provide to the user intuitive modes of sending inputs to the robot. The block collects these inputs and elaborate them to produce suitable commands (Section~\ref{sec:chap3:int}).
	
	\item \textbf{Situation Awareness}. This block delivers to the user informative feedback signals about the commands issued and about data coming from the robot (Section~\ref{sec:chap3:sit}).
	
	\item \textbf{Robot Autonomy}. The user commands, if the framework implements a certain level of robot autonomy, are further elaborated by this block, which generates appropriate commands for the robot to comply with the user requests, according to the sensed data coming from the robot itself (Section~\ref{sec:chap3:auto}). 
\end{itemize}

These three key themes are recurrent in studies about human-robot interactions. For example, the survey~\cite{Losey2018} identifies as key characteristics \textit{intent detection}, \textit{arbitration}, and \textit{feedback/communication}. The arbitration and feedback/communication themes are just other names of \textit{Robot Autonomy} and \textit{Situation Awareness}, since the first refers to \enquote{the division of control among agents when attempting to accomplish some task}, and the latter as the challenges inherent to communicate information to the user. Instead, the scope of intent detection is slightly different from the \textit{Intuitive Command Interface} theme of this thesis, since with the prior the focus is on \enquote{the need for the robot to have knowledge of some aspect of the human's planned action in order for the robot to appropriately assist toward achieving that action}, while with the latter challenges are about how to provide to the user effortless and easy-to-learn interfaces. Nevertheless, even within this two slightly different categories, common characteristics and challenges are evident.

In what follows, the details of such key features are discussed.

\subsection{Intuitive Command Interface}\label{sec:chap3:int}

Intuitiveness for human-robot interfaces refers to the ease with which users can understand and operate robotic systems without the need for extensive training or explicit instructions. An intuitive interaction is natural, easy to understand, and consistent with human expectations. 
If the communication is intuitive, the user can focus on the tasks instead of how to provide commands to the robot~\cite{VILLANI2018}.
By adhering such principles, users, even without specialized robotics expertise, can efficiently operate robots to accomplish tasks, thereby increasing their acceptance toward the adoption of robotic technologies in diverse applications, ranging from industrial settings to everyday life~\cite{Murphy2019}.  

There are various challenges linked to the development of intuitive human-robot interfaces.
Robots, especially highly-redundant ones, may necessitate a wide range of input commands to control their various capabilities. 
Hence, the input system delivered to the operator can become very complex, leading to the design of sophisticated exoskeletons or full-body suits to track the operator's movements. 
Even if these interfaces offer users the flexibility to exploit the full range of robot capabilities, there is a potential reduction in intuitiveness. This happens because users require to learn a complicated mapping between functions of the input devices and the diverse robot motions. 
Moreover, complex input systems typically entail higher costs and longer setup time, factors  undesirable in real-world contexts.
  
Another challenge lies in ensuring adaptability to the individual user. Users may have different preferences, or they may even have some physical limitations, thus requiring particular attention in designing interfaces suitable for diverse needs. 

Simultaneously, the interface should be adaptable to different robots and tasks, enabling users to employ the same interface across various applications without the need to learn specific control methods for each robot and task.

To address these challenges, one of the most interesting solutions is to enable communication between humans and robots through human body movements~\cite{LiuWang2018}.
This approach can offer an easy-to-learn and easy-to-use interface, that permits, for example, to command the robot's locomanipulation skills using natural body movements, or indicating objects or locations of interest through gestures.
This mode of interaction overcomes the limitations of traditional control-boxes and joysticks, which involve less intuitive mappings between the input and the robot behavior.
Nevertheless, particular attention must be taken into account when developing interfaces based on the user body movements, since the complexity of the input system can grow, mirroring the complexity of the robot itself, and falling in the above-mentioned category of complex exoskeletons or full body suits.
 
A highly intuitive method of controlling a robot is through physical human-robot interaction, particularly used for teaching or guiding the robot in collaborative tasks. This approach typically requires no learning time: the robot can be simply guided to desired locations by applying forces on its body.
However, physical interactions are not always possible, for example when the robot is located remotely, or when safety issues impose limitations on the robot strength and speed. Nonetheless, given its natural intuitiveness, this mode of interaction can be source of inspiration to develop intuitive human-robot interfaces, exemplified by the \acrlong{tpo} interface discussed in this thesis.

Overall, focusing on the intuitiveness is important to continue exploring new directions to develop human-robot interfaces that strike the best compromise between simplicity and possibility to control all the necessary robot capabilities.
As we will see, the contributions of this thesis consider the intuitive use of motions of the human body to command even a highly-redundant robot, while prioritizing simplicity, low-cost, and ease of learning.

\subsection{Situation Awareness}\label{sec:chap3:sit}

In human-robot interfaces, it is important to let the operator \enquote{understand what is happening}~\cite{Dosstantos2020}, by adding situation awareness characteristics to the interface, to deliver important information to the user through a feedback channel~\cite{Endsley2016}.

Indeed, the integration of feedback mechanisms plays a crucial role in improving the interface's effectiveness since it permits to communicate to the user information about the commands they have issued, the current state of the robot, and the progress of the task.
An effective situation awareness enables users to plan the subsequent actions accordingly to the received information. A clear feedback aids the comprehension of ongoing processes, which is fundamental in scenarios where users may not have a technical background or extensive training in robotics. 

In any case, it is crucial to account for the right trade-off between incorporating a variety of feedback information and maintaining the simplicity of the interface, a direction considered by the works presented in this thesis.

Particular attention must be taken in providing only the relevant information, to not overwhelm the user with useless data, and deliver them in the most comprehensible way possible. 
In the field of telepresence~\cite{Darvish23}, innovative interfaces are explored by developing full suits that deliver a heterogeneous range of feedback, including tactile, auditory, and visual cues, realizing a very immersive experience for the user.
However, these interfaces can be very complex, and may burden the operator with an excessive amount of information to handle, some of which that may not be necessary at every moment.
The situation awareness capabilities of the interface is strictly related to the overall intuitiveness: too little information may leave the user unaware of the situation, while too much, irrelevant, information, may confuse the user and divert the attention from the important aspects to monitor during a task.

Furthermore, the practical implementation of such technologies should be modular: different levels of details should be provided to different users, according to their expertise and role in the interaction with the robot. 
 
\subsubsection{Visual Feedback}\label{sec:chap3:visual} 
 
The most common method to deliver situational information is through visual feedback, realized in different ways~\cite{Moniruzzaman2022}. 

When in proximity to the robot, users can observe its actions to monitor its status and assess the progress of the task. This type of feedback comes for free since does not require additional means. However, direct observation of the robot may not always offer sufficient information, and users might miss crucial details since there is no way to direct their attention toward crucial aspects to consider. This is particular critical if users lack the expertise to identify the key details to consider in the robot's actions.

An important mean to deliver visual information to the operator are screens, where a multitude of data can be displayed. Screens are beneficial to deliver additional information when the operator control a local robot, and they are indispensable when teleoperating a robot at a remote location. 
With such a mean, data from some of the robot's sensors can be directly presented to the user, such as images from a robot camera to observe the robot's surroundings. Various types of \acrfull{gui} can be utilized to convey multiple pieces of information in a wide variety of methods, specializing the view according to the needs.

Nevertheless, relying solely on visual feedback from a \acrshort{gui} may be not sufficient. This is particular true with highly-redundant robots and complex tasks, where the numerous details to consider can overwhelm the operator's visual sense. 
Furthermore, certain information, like the interaction forces between the robot and the environment, can not be easily and naturally perceived through the visual sense alone.
Indeed, human-robot interfaces typically integrate other types of feedback along the visual ones, with one of the most common being haptic feedback, discussed in Section~\ref{sec:chap3:haptic}.

\subsubsection{Haptic Feedback}\label{sec:chap3:haptic}

Haptic feedback is a technology that leverage on the human sense of touch to communicate information~\cite{Hannaford2016}.  
The introduction of haptic feedback into human-robot interfaces augments the range of information delivered to the user, complementing visual and other forms of information. 

By leveraging on a different human sense, certain types of information are conveyed in a more natural way. For instance, when the robot physically interacts with the environment, it is natural for the operator to perceive these interaction forces through tactile stimuli, as the tactile sense is typically engaged during physical interactions with the environment or objects.

Haptic feedback can be also particular useful to alert operators. By delivering tactile sensations, the operator's attention can be captured better compared to signaling a warning on a screen.

The field of haptics is vast. According to the directions taken by this thesis, we will not deal with haptic grounded devices, as these interfaces can limit the operator movements falling into the category of cumbersome control interfaces, which this thesis's contribution aims to surpass. 
Instead, we will focus wearable haptic devices, which offer a more versatile approach. These devices well comply with the possibility of commanding the robot through user body motions, as they utilize the same mean, the user's body, to deliver information. By exploiting into the sense of touch, haptic feedback provides a valuable tool for conveying critical information in a way that is intuitive and complementary to other forms of sensory feedback.

\subsection{Robot Autonomy}\label{sec:chap3:auto}

The incorporation of a certain level of robot autonomy in human-robot interfaces is crucial for several reasons. Ideally, it would be awesome to have a robot able to perform any kind of task in a full autonomous manner. Realistically, the operation of robots still necessitates of a human in the loop. This is unavoidable because, as today, robots do not consistently demonstrate capabilities comparable to human ones in terms of comprehending the task's context, planning adequate behaviors, and promptly adapting to unforeseen inconveniences that may happen~\cite{Selvaggio2021}.

Still, modern robots are equipped with sensors and intelligence that allow them to efficiently act in a semi-autonomous manner. A certain level of robot autonomy is almost always incorporated in human-robot interfaces, \enquote{due to the great benefits of combining the human intelligence with the higher power/precision abilities of robots}~\cite{Gaofeng2023}. 

With the implementation of robot autonomy features, one primary benefit is the reduction of the cognitive load on the human operator. 
With highly-redundant robots, it is often binding to delegate part of the control to the robot itself, since manual control of every robot motion would demand a substantial operator effort, potentially increasing the difficulty of a task. 
In some scenarios like assistive ones, where users may present limited physical abilities, autonomy is fundamental to enable a natural integration of assistive robotic systems into the everyday life of individuals.
In general, by offloading challenging, repetitive, and routine objectives to the robot's autonomy, the human operator can concentrate on the most important parts of the task related to the higher-level decision-making aspects. This not only enhances the overall efficiency of the system but also makes it more user-friendly.

A human-robot interface should also present flexibility in balancing between autonomy and user control, incorporating a variable level of robot autonomy. This adaptability enables the human operator to intervene or adjust the robot's actions when necessary, as a consequence of unforeseen events, limitations in the robot autonomy, and user's preferences about task execution. 

Different applications present different needs related to the level of robot autonomy. A very banal example is the implementation of inverse kinematic algorithms that permit the user to command the manipulator's end-effector at a Cartesian level rather than controlling each individual joint. 
This autonomy is also typically expanded with automatic planning solutions to satisfy the user's commands while simultaneously addressing other control objectives, like preventing kinematic singularities or avoiding obstacles.

The more users are granted a higher level of control, thus reducing their responsibilities in the robot management, the more the robot requires independence and abilities to accurately fulfill the user requests. Hence, more challenges are presented in the development of a suitable control architecture, which necessitates the implementation of additional software modules. 
For instance, the user may issue a high-level command like indicating an object to manipulate. In this case, the robot must understand what the user is pointing, move to reach the object, and interact with the object in the manner intended by the user.

Ultimately, incorporating autonomy for the robot is related to the ease of use of the interface, and it is crucial to consider in the development of novel solutions. As it will be presented, the works of this thesis address such considerations, implementing different levels of robot autonomy depending on the interface adopted in the considered applications.

\section{\thesistitleshort}

The main focus of this thesis is to develop innovative human-robot interfaces which address the key features presented in the previous Section~\ref{sec:chap3:keys}.
By facing the challenges introduced by the operation of modern robotic systems, the human-robot interfaces explored connect the human to the robot permitting to operate the various robot capabilities with ease. 

To address various scenarios and robots, multiple interfaces have been developed that can be classified into two main categories:

\begin{itemize}
	\item \textbf{\acrfull{tpo}} interfaces, which enable a \textit{direct control}~\cite{Gaofeng2023} of the robot in an intuitive manner by means of operator's arms movements. The architecture is enriched with autonomy functionalities in a shared control fashion, to allow the control of highly-redundant robots and the execution of complex tasks (Part~\ref{part:one}). 
	
	\item \textbf{Laser-guided} interfaces, a kind of \textit{supervisory control}~\cite{Gaofeng2023}, which permit the operator to manage the robot by intuitively point to locations in the environment with a laser emitter device. This relieves the operator from controlling directly the motions of the robot, which are instead taken into consideration by the autonomy implemented in the interfaces (Part~\ref{part:two}). 
	
\end{itemize}

\noindent The following sections highlight the advantages of these interfaces considering the key themes addressed.

\subsection{TelePhysicalOperation}\label{sec:chap3:tpo}
 
\subsubsection{A ``Marionette'' Teleoperation Interface}
The \acrfull{tpo} concept, that will be presented in Chapter~\ref{chap:TPO}, enables a remote robot control but keeping the intuitiveness of a physical human-robot interaction, permitting to operate at a safe distance with ease. The operator applies virtual forces on the robot body, and the interface generates robot motions to comply to such forces, resembling a physical human-robot interaction but without any real contact.

The interaction with the robot is similar to a \enquote{Marionette} interface: the operator, by moving his/her arms, pulls and pushes the robot body part selected, like a link of a robot arm to control the manipulation ability, or the pelvis of a legged platform to control the locomotion ability. 
Multiple virtual forces can be applied at the same time: for example, by selecting two different links of the same kinematic chain, it is possible to shape its form exploring its redundancy.

The method abstracts the particular kinematic characteristics of the robot. Since the inputs are virtual forces, differently from other approaches, there is no need for a mapping between the different kinematics of the input system (the user's body, or an exoskeleton-like device) and the robot. Hence, the operator has not to internalize a mapping between inputs and robot motions that can be not so intuitive due to the different kinematics. By eliminating the need of such mapping, the interface becomes flexible and easily adaptable to different robots. 

The method is also different from a classic Cartesian velocity end-effector control, since it allows not only to command the end-effector movements directly (by applying a virtual force), but also allows the application of multiple forces on different links of the arm to shape it exploiting its redundancy.

The generation of the virtual forces is based on the user's arms motions, monitored with lightweight tracking cameras worn on each of operator's wrist. The \acrshort{tpo} Suit is comprising essentially of these cameras, eventually wearing only one camera if commanding a single virtual force at a time is sufficient. 
The suit is comfortable to wear, cheap, and easy to set up, and maintain the freedom of movement since it is not tethered to any other unit of the system, thanks to a wearable pocket-size PC which enables a wireless communication with the robot. 
This results in a versatile wearable system, not bound to a particular robot as can occur with certain exoskeleton-like solutions, yet capable of controlling highly-redundant robots.
Furthermore, there is no the necessity of many wearable devices which may complicate the interface, and that are present in other motion tracking solutions like \acrshort{imu}-based suits.

\subsubsection{Wearable Haptic-enabled TelePhysicalOperation}
The \acrlong{tpo} architecture provides a number of feedback mechanisms to provide information to the user. One such feature is the visualization of the virtual forces applied to the robot, displayed as arrows on the robot kinematic model through a \acrshort{gui}. Even if a local user can have a direct feedback by looking at the robot, this functionality may help in understanding the given inputs, and can be utilized according to individual preferences.  

Another communication channel incorporated in the \acrshort{tpo} framework is a haptic feedback, as it will be detailed in Chapter~\ref{chap:TPOH}.
In a physical human-robot interaction and in a real \enquote{Marionette} interface, the user can feel the counteracting forces while interacting, indirectly aiding him/her in commanding the robot or marionette.
Recognizing the importance of delivering natural haptic feedback to the operator for controlling a robot from a distance, wearable haptic vibrotactile devices have been designed and integrated in the \acrshort{tpo} Suit.

This enhancement delivers tactile feedback according to the application of the \acrshort{tpo} virtual forces, resembling the feeling of the increasing tension of the rope of the real \enquote{Marionette} interface. 
The haptic devices permit also the transmission of other tactile stimuli, related, for example, to the interaction of the robot's end-effectors with the environment. 
Additionally, control inputs are also incorporated in the devices as buttons, expanding the user possibilities.  

The intuitiveness, ease-of-use, and flexibility of the \acrlong{tpo} interface, along with the simplicity of the \acrshort{tpo} Suit, are maintained by developing the haptic devices based on a lightweight and unobtrusive design. 

\subsubsection{Enhancing TelePhysicalOperation with Robot Autonomy Features}

The TelePhysicalOperation concept enables a direct control of the robot maintaining a high level of intuitiveness. Nevertheless, especially for highly-redundant robots, we have stressed the fact that it is not ideal to make the operator taking care of every aspect of the robot's motion. In fact an effective human-robot interface should equip the robot with an adequate level of autonomy.
In the first implementation of the \acrshort{tpo} interface, that will be detailed in Chapter~\ref{chap:TPO}, some autonomy features are considered to manage more efficiently manipulation and locomotion abilities. For example, a wheel-leg platform can be moved by virtually pushing its body instead of controlling each single leg. 

In Chapter~\ref{chap:tpoAuto}, more interesting autonomy features will be introduced. These modules, integrated into the \acrshort{tpo} architecture, facilitate the execution of tasks with highly-redundant robots.

The first feature is a manipulability-aware shared locomanipulation motion generation method utilized to facilitate the execution of telemanipulation tasks with mobile manipulators. The manipulability measure of the robot arm end-effector is explored to make the mobile base actively coordinate its motions with the manipulator motions in order to assist the end-effector reaching the manipulation target location with a good level of dexterity.

The operator can exclusively apply a virtual force to the end-effector, while the underlying architecture smoothly distributes this input toward the mobile base when the manipulability of the arm is low.
This autonomy feature ensures that the manipulability of the arm is autonomously maintained at a desired level, an important factor to avoid kinematic singularities that may be challenging for the operator to address manually while commanding the manipulator motions.
Concurrently, the mobile base is autonomously activated when the arm in a region of low manipulability, ensuring compliance with the virtual force from the operator who is guiding the end-effector toward a direction. 
In conclusion, the operator is still in control of the most relevant aspects of the robot motion, i.e.\ the end-effector positioning, but he/she has not to consider when mobile base motions are required, for example to reach distant targets. 

Another autonomy feature addresses the challenges inherent in operating a robot for bimanual object transporting. Instead of requiring to generate arm motions, the operator can just command the transported object directions. It is the interface which takes the responsibility of generating arms' motions, according to the operator's input and the grasping constraint (i.e., transporting the load without dropping it and without squeezing it too tightly). 
This feature permits to accomplish a task that would otherwise be very difficult or even impossible. Firstly, generating coordinated motions of the two robot's arms can be tedious for the user. More importantly, regulating the motions of the arms complying with the grasping constraint can be very complex, even if some feedback information, like haptic information, is delivered.

This feature is integrated with the previous shared locomanipulation motion generation, which commands the mobile base according to the arms manipulability, further relieving the operator in considering another aspect of the robot motion during the object transportation. 
To complete this bimanual transportation interface, the robot is equipped with the ability to autonomously grasp the object and estimate the object's mass, which is necessary to regulate the grasping forces accordingly.

\subsection{Laser-guided Interfaces}

The idea of the laser-guided interface, that will be discussed in Chapter~\ref{chap:Laser1}, revolves around the utilization of a laser emitter device to let the operator indicate points of interest to the robot by projecting the laser. 
Hence, the operator does not control directly the robot, but instead he/she commands target locations to be reached, paths to follow, or object of interest to interact with. This way of interacting with the robot is inherently intuitive, since it resembles the natural human communication style, i.e.\ pointing at a location or an object to indicate them as a target. 

The laser projection is detected by the robot through a vision system based on a neural network. This solution has proven to be robust, hence improving the adaptability of the interface to different scenarios, and fast, ensuring in practice a tracking of the laser spot, resulting in minimal latency between the user's request to reach a dynamic location and the robot's movement. 

The laser interface, because of its characteristic, requires the operator to be in the same location of the robot. Nevertheless, a safe distance can be maintained, hence leveraging on the full robot power without any safety issue. Another characteristic of the interface is that the laser projection itself serves as a visual feedback; a screen is not necessary to let the operator understand which is the target location that he/she is selecting.

As a supervisory control interface, this laser-guided interface demands a higher level of robot autonomy with respect to a direct control interface. Specifically, the robot must autonomously reach the indicated target in a suitable manner, which may involve exploiting locomanipulation capabilities or the avoidance of obstacles. Depending on the application addressed, highly-redundant robots and assistive manipulators, different approaches to implement such autonomy have been developed, as presented in the following sections.

\subsubsection{An Intuitive Tele-collaboration Interface Exploring Laser-based Interaction and Behavior Trees}

As mentioned earlier, the laser-guided interfaces allows the user to control the robot by indicating target locations, requiring a certain level of robot autonomy to generate motions in pursuit of these locations. 
In Chapter~\ref{chap:Laser1}, it will be detailed a solution based on a \acrfull{bt} planner, especially useful for highly-redundant robots.

In this approach, the robot's capabilities, like arm, gripper, and body actions, are embedded in the \acrshort{bt} as action nodes that run according to the logic flow of the \acrshort{bt}. 
This solution offers advantages in terms of modularity and reactivity. The modularity allows to easily define and specialize the structure of the \acrshort{bt} according to the mission needs and robot capabilities, by reusing the nodes and subtrees already implemented. 
The reactivity ensures that the robot's behavior promptly responds to changes in the conditions present in the \acrshort{bt}, activating the correct robot capability. This aligns with the reactiveness of the laser detection, enabling a prompt response to follow the dynamic movements of the laser spot.
 
Consequently, the proposed interface is effortless to use thanks to its natural intuitiveness and to the fact that the robot can generate motions autonomously, such that even highly-redundant robots can be managed with ease.
The operator does not have to worry about the specific robot abilities, as the robot autonomy generates the motions according to the \acrshort{bt} structure.
Finally, the modularity of such solution permits to easily adapt the automatic behavior to the mission and to robot in use, eventually providing flexibility in choosing the best strategy according to user preferences.

\subsubsection{A Laser-guided Interface for Robotic Assistance}

Based on the laser-guided interface, it has been developed a human-robot interface to deal with robotic assistance scenarios, which will be detailed in Chapter~\ref{chap:Laser2}. 

The application involves the utilization of an assistive robotic manipulator to aid individuals with upper limbs impairments in performing \acrfull{adl} tasks.
Considering such a context, instead of guiding the robot by pointing to locations with a handheld laser device, the laser emitter is worn on the head, enabling to intuitively control the robot through head movements that result in the laser to be directed at the environment.
 
As in the previous laser-guided interface, this mode of human-robot interaction is highly intuitive, since it is natural: people usually direct their heads to address a particular location or object. Furthermore, as previously, users can directly have an immediate visual feedback about the command given to the robot. 

Regarding how robot motions are generated, this is dependent on which one of the two control modalities is utilized, each leveraging different levels of robot autonomy.  
With the first modality, a motion planner is utilized to generate and execute an end-effector collision-free trajectory toward the location pointed by the laser. 
Instead, the second modality allows a more direct control of the robot, enabling the command of end-effector Cartesian velocities and gripper actions. To enable this, a paper keyboard is present in the environment, whose keys correspond to particular motions (e.g., Cartesian directions). By selecting one of the key pointing the laser onto it, the correspondent robot motion is commanded.

With these two control modes, the user is able to choose the best strategy according to the task and to his/her preferences. The selection of modes is seamless, since it is sufficient to point the laser; there is no the necessity of additional inputs, which feasibility can be limited in applications where users have impaired mobility.

\part{TelePhysicalOperation}\label{part:one}
\chapter{A ``Marionette'' Teleoperation Interface}\label{chap:TPO}

\begin{figure}[H]
	\centering
	\includegraphics[width=0.9\linewidth]{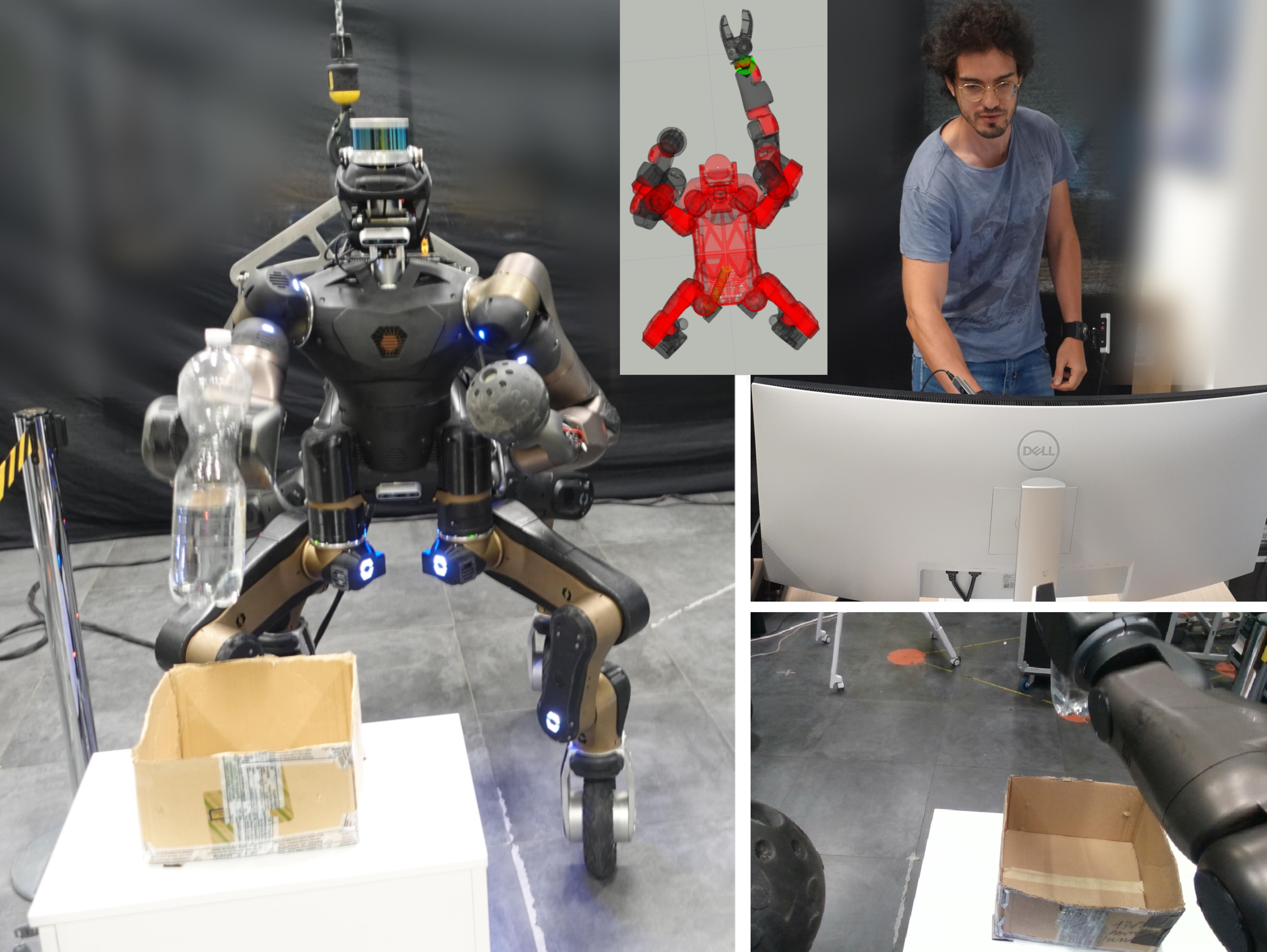}
	\caption[TelePhysicalOperation interface use-case example]{The author of this thesis is teleoperating the CENTAURO robot with the TelePhysicalOperation interface. By applying a force with his right arm on the right end-effector of the robot, he is pushing down the arm to place the bottle in the box. The virtual force is displayed as a green arrow in the RViz visualization.}
	\label{fig:tpo:firstphoto}
\end{figure}

\lettrine{A}{s} stated in the previous chapters, one of the key challenges to further advance robotics technology is the development of intuitive human-robot interfaces.
In this chapter, it is presented the \acrfull{tpo} interface, a teleoperation paradigm which enables the remote control of robots keeping the intuitiveness of a human-robot physical interaction (\figurename{}~\ref{fig:tpo:firstphoto}). In what follows, the concept of \acrlong{tpo} is presented, the system architecture is explained, and validation experiments are reported.\\

\noindent This chapter is based on the following article:\\
\fullcite{TPO}~\cite{TPO}

\section{TelePhysicalOperation Concept}\label{sec:tpo:concept}
\begin{figure}[H]
	\centering
	\includegraphics[width=\linewidth]{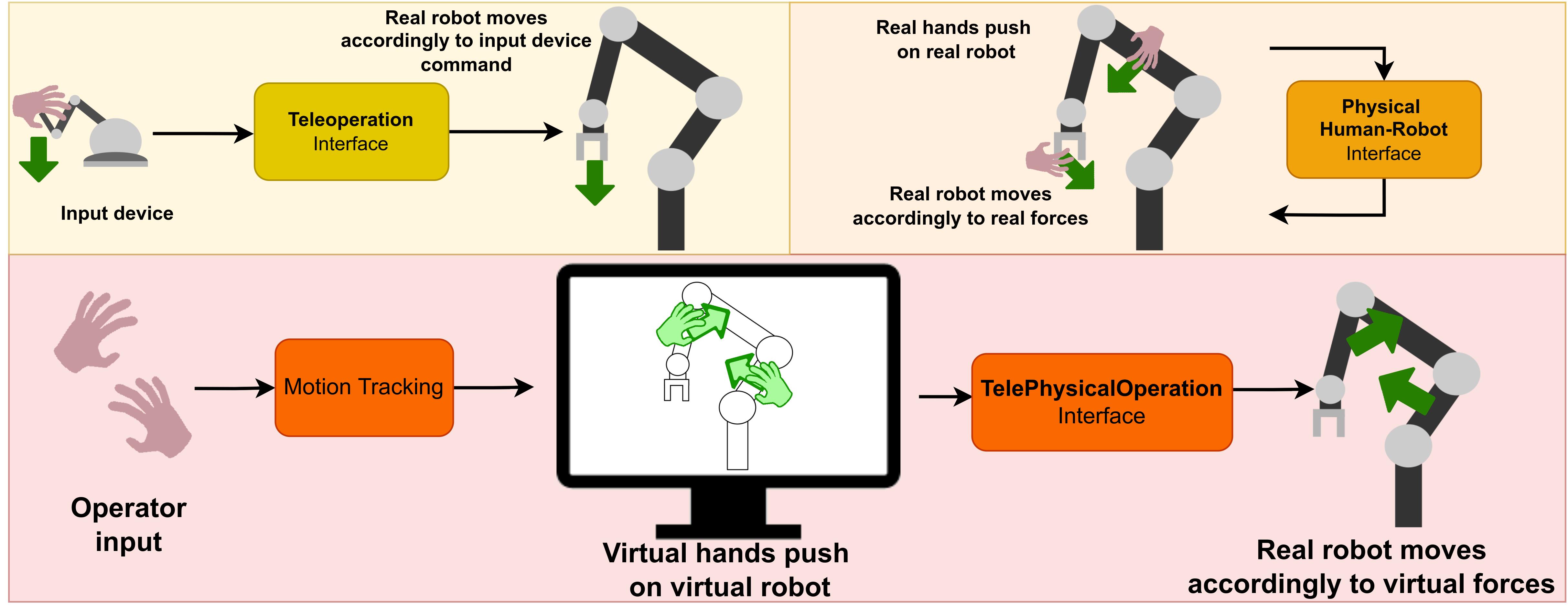}
	\caption[Concept of TelePhysicalOperation]{Concept of TelePhysicalOperation. Above, schematics of the traditional teleoperation and the physical human robot interaction interfaces. Below, the scheme of TelePhysicalOperation, derived by the combination of the two above controlling interfaces.}
	\label{fig:TPOConcept}
\end{figure}

The overall concept of \acrlong{tpo} is schematically illustrated in \figurename{}~\ref{fig:TPOConcept}. As it can be seen in the upper part of the scheme, the TelePhysicalOperation paradigm emerges from the blending of the teleoperation interface and the physical human-robot interface. The first is traditionally used to control a remote robot, thus permitting to maintain a safe distance from it and from the potential dangerous environment where the robot is operating. Works in this field have been listed in Section~\ref{sec:soa:leadFolInt} and Section~\ref{sec:soa:bodyTrack}. The second is a more intuitive way to collaborate with a robot since a physical interaction is possible, for example for a teaching or a guiding task, even if this can introduce some safety issues. Physical human-robot interfaces have been introduced in Section~\ref{sec:soa:phri}.

The rationale idea of the TelePhysicalOperation approach is that, by combining these two kind of interfaces, a generic and intuitive non-physical human-robot interaction interface can be provided, that can be suitable for controlling either a remote robot (teleoperation) or a collaborative robot (collaboration).
The TelePhysicalOperation way of interacting with the robot is inspired by how a user physically interacts with a collaborative robot during, for example, the teaching or the guiding phase of the collaborative task. When the operator applies forces on the robot body on multiple contact points, he/she has the possibility to precisely shape the pose of the robot end-effector as well as regulate other motions permitted by the available robot redundancy to accomplish the task.

To enable a virtual physical human-robot interaction, the \acrlong{tpo} concept relies on the application of virtual forces applied by the operator to different control points along the kinematic chain(s) of the robot. 
Therefore, only \textit{virtual} contacts are established between the operator and the real robot. As in a physical human-robot interaction, the remote/collaborative robot responds to these virtual forces regulating accordingly its motions. This resembles the compliant motion response that the operator expects when he/she physically interacts with the robot. 
Furthermore, by applying multiple virtual forces on the same kinematic chain, it is still possible to exploit the kinematic redundancy to shape the robot as necessary.

\begin{figure}[H]
	\centering
	\includegraphics[width=\linewidth]{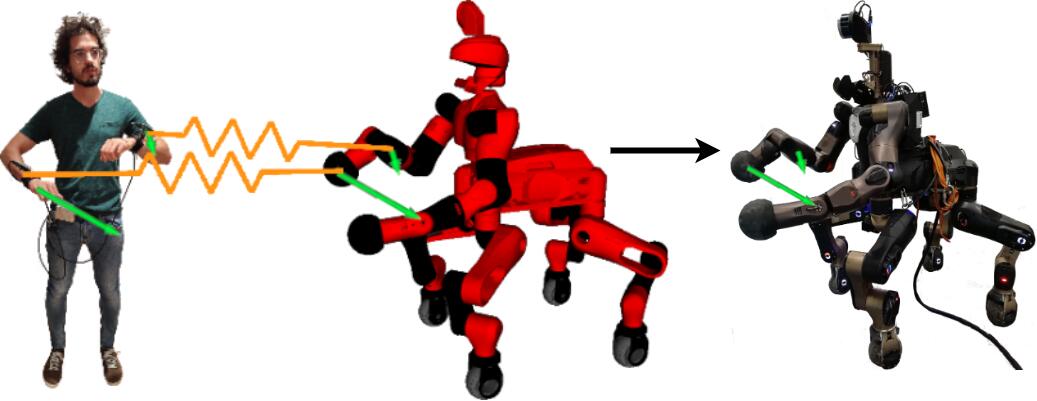}
	\caption[TelePhysicalOperation: a \enquote{Marionette} type interaction]{The TelePhysicalOperation concept follows a \enquote{Marionette} type interaction: the operator pulls and pushes virtual ropes that are attached to the selected robot body parts.}
	\label{fig:TPOHumanRobot}
\end{figure}

This TelePhysicalOperation idea approximates the \enquote{Marionette} motion generation principle, as illustrated in \figurename{}~\ref{fig:TPOHumanRobot}.
The virtual forces are generated by virtual ropes attached to one end to the operator wrists and to the other end to the selected control points of the robot body. The control points can be any parts of the robot, like the links of an arm or the body of a mobile base. The operator, by moving his/her arms, pulls and pushes the virtual ropes, as the ropes of a real \enquote{Marionette}. The virtual forces generated by the ropes are utilized to drive the motions of the robot accordingly. To monitor the movements of the operator arms, appropriate tracking devices are utilized, as detailed in Section~\ref{sec:tpo:tposuit}.

\noindent The main contributions of the proposed interface are summarized below:
\begin{itemize}
	
	\item The TelePhysicalOperation allows controlling the robot from a distance through operator's arms movement. As discussed in Section~\ref{sec:soa:bodyTrack}, using body motions is an intuitive method of operating a robot.
	Furthermore, the proposed interface negotiates the safety issues of the physical human-robot interaction while maintaining its intuitiveness, deriving from the natural way of guiding the robot through physical contacts, as seen in the previous studies of Section~\ref{sec:soa:phri}.
	This can be particularly useful in the cases where a high power collaborative robot may not permit to physically interact with it due to safety regulations.
	
	\item Differently from a standard end-effector based teleoperation, TelePhysicalOperation permits to control multiple body segments of the robot allowing to regulate both the end-effector motions and the motion of different links along the kinematic chain. This permits to control more precisely the shape of the chain, eventually exploiting the redundant \acrfull{dof} effectively.
	
	\item The method can abstract the particular kinematic characteristics of the robot. Indeed, it is not necessary to have a potential complex hardware interface whose \acrshort{dof} must be mapped to the possible different \acrshort{dof} of the robot kinematic chain, as done in other works presented in Section~\ref{sec:soa:leadFolInt}. The only input device required is a system that can track the motions of the operator's arms, needed to generate the virtual forces. This allows for a fast adaptation of the system to different kinds of robots.
	
\end{itemize}

\section{TelePhysicalOperation Realization}\label{sec:tpo:realization}

\begin{figure}[H]
	\centering
	\includegraphics[width=\linewidth]{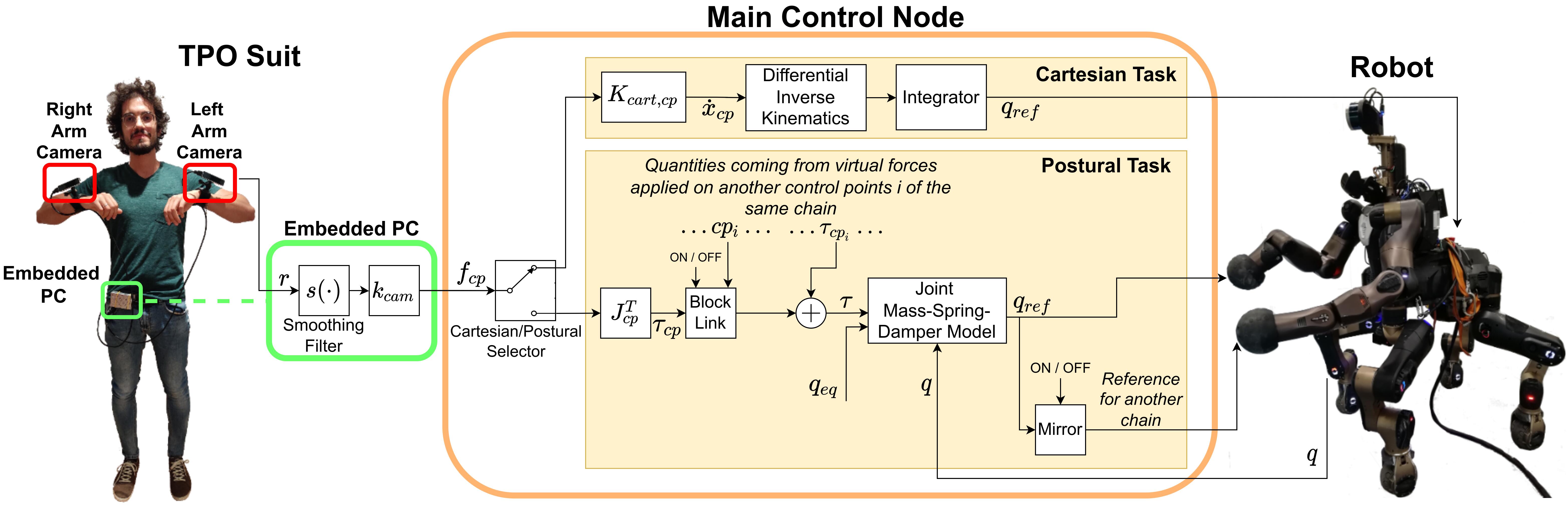}
	\caption[TelePhysicalOperation architecture scheme]{The scheme of the architecture of TelePhysicalOperation. On the left, the operator is wearing the TPO Suit, which provides a virtual force $\boldsymbol{f}_{\mathit{cp}}$ to the Main Control Node. The Main Control Node is in charge of processing the virtual force based on the task (Cartesian/postural) associated with the control point chosen. On the right, the joint references $\boldsymbol{q}_{\mathit{ref}}$ are sent to the robot.}
	\label{fig:controllerschemeModes}
\end{figure}

The TelePhysicalOperation realization relies on two main components as depicted in \figurename{}~\ref{fig:controllerschemeModes}:

\begin{itemize}
	\item The \acrlong{tpo} Suit (\acrshort{tpo} Suit), an effective low-cost motion tracking wearable system, in charge of monitoring the movements of the operator's arms and of providing the requested inputs under the form of virtual forces $\boldsymbol{f}_{\mathit{cp}}$ to the Main Control Node. This component is presented in Section~\ref{sec:tpo:tposuit}.
	
	\item The Main Control Node, which is responsible for handling the virtual forces $\boldsymbol{f}_{\mathit{cp}}$ received by the \acrshort{tpo} Suit and deriving the corresponding motions that must be generated by the robot due to the application of the virtual forces. This component is detailed in Section~\ref{sec:tpo:control}.
\end{itemize}

It is worth noticing that the two main components are independent to each other. The motion tracking delivered by the wearable \acrshort{tpo} Suit can be utilized as an input for other frameworks; the Main Control Node can accept input from any other hardware (like an \acrshort{imu} suit, or a traditional input controller) or software (like a potential field generator to avoid obstacles).
The potential exploitation of the architecture for other applications is facilitated also by the fact that it is built on top of the well-know \acrshort{ros} middleware~\cite{ROS}.

The TPO Main Control Node runs on a pilot PC which handles the data received and communicate with the robot. This communication, is accomplished through the utilization of the XBot architecture~\cite{XBot2}.

The TelePhysicalOperation architecture delivers some visual feedback features to render the robot's representation with \textit{RViz}, the \acrshort{ros}-standard kinematic visualizer, while displaying the virtual forces as arrows. Further details are given in Section~\ref{sec:tpo:visual}. More functionalities related to the situation awareness are explored with haptic feedback and integrated in the \acrshort{tpo} framework as discussed in Chapter~\ref{chap:TPOH}.

Furthermore, autonomy features are integrated to assist the operator in executing the task.
A simple one is represented by the Differential Inverse Kinematics block of \figurename{}~\ref{fig:controllerschemeModes}, which generates appropriate joint references from a Cartesian velocity reference considering the kinematic of the robot in use, e.g., a highly-redundant quadruped like the CENTAURO. This is handled by the \textit{CartesI/O} Control Framework~\cite{cartesio}, just incorporated in the \acrshort{tpo} architecture.
\textit{CartesI/O} handles also the automatic execution of trajectories to reach specific end-effector Cartesian poses, recorded during the teaching phase of the experiment presented in Section~\ref{sec:tpo:expteach}.
Contributions of this thesis involves instead other autonomy functionalities. 
Some of them are presented in this chapter, i.e., the \textit{Blocking Link} and the \textit{Mirroring Motion} features, displayed in \figurename{}~\ref{fig:controllerschemeModes} and described in Section~\ref{sec:tpo:auto}.
More elaborated autonomy features that have been developed and integrated in the TelePhysicalOperation architecture are discussed in depth in the Chapter~\ref{chap:tpoAuto}.

\subsection{TelePhysicalOperation Suit} \label{sec:tpo:tposuit}

\begin{figure}[H]
	\centering
	\includegraphics[width=0.7\linewidth]{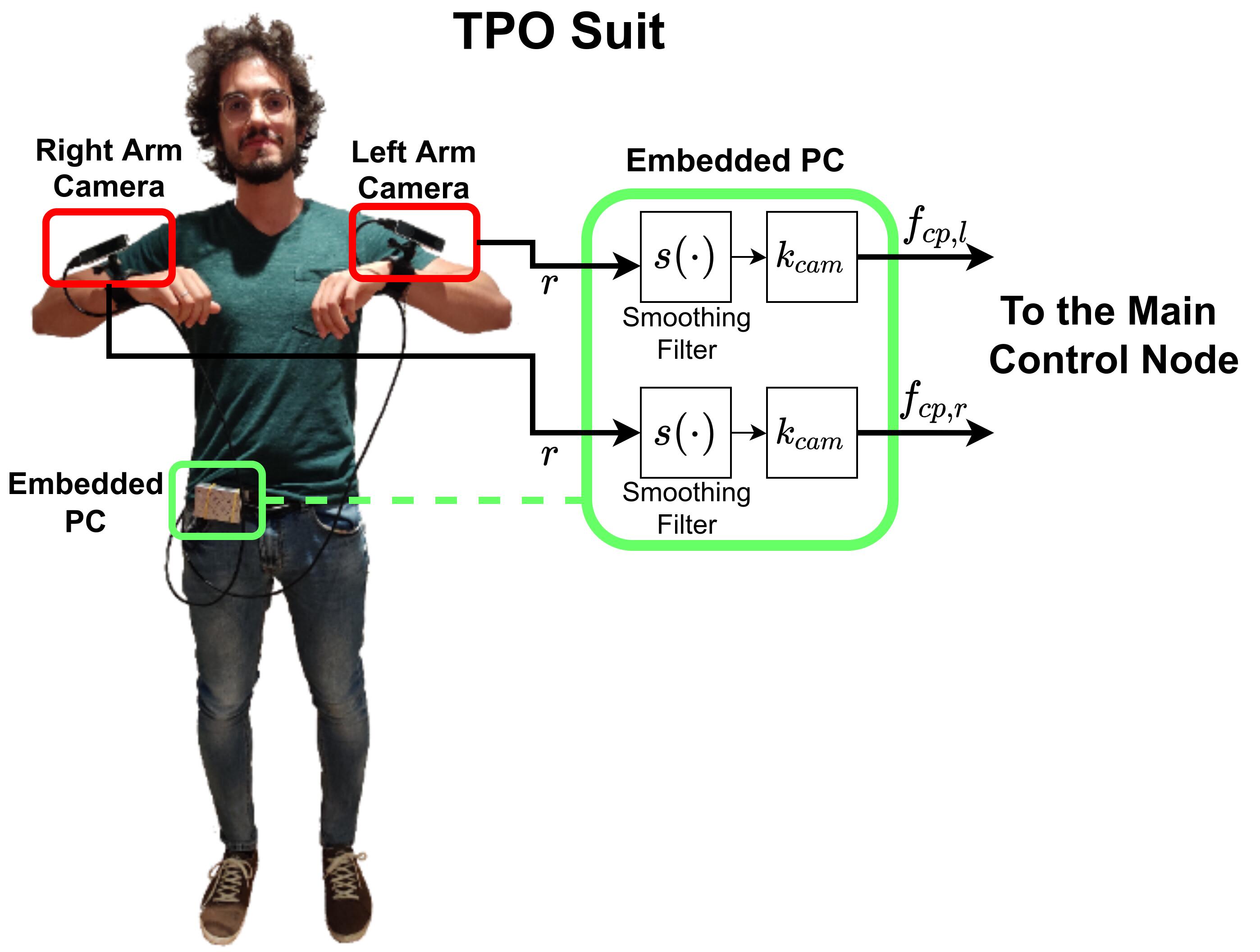}
	\caption[TelePhysicalOperation Suit]{The TelePhysicalOperation Suit is a lightweight wearable interface that permits to track the operator's arm movements by utilizing tracking cameras. From the tracking of the user's arms, the Suit computes the virtual forces $\boldsymbol{f}_{\mathit{cp}}$ and delivers them to the Main Control Node.}
	\label{fig:tpo:tposuit}
\end{figure}

As stated previously, the virtual forces to be applied to the remote robot body parts are generated from the motion of the operator's arms. To track these movements a lightweight and low cost motion capture solution has been realized, based on \acrfull{vslam} tracking cameras. 
The developed TPO Suit, schematized in \figurename{}~\ref{fig:tpo:tposuit}, is composed of:

\begin{itemize}
	\item Two\footnote{It's worth noticing that a single camera can suffice if inputs from only one user's arm is necessary for a particular task, as in the ones presented in Chapter~\ref{chap:tpoAuto}.} \textit{Intel\textsuperscript{\textregistered} RealSense Tracking Camera T265}\footnote{\href{https://dev.intelrealsense.com/docs/intel-realsensetm-visual-slam-and-the-t265-tracking-camera}{https://dev.intelrealsense.com/docs/intel-realsensetm-visual-slam-and-the-t265-tracking-camera}, discontinued product} that the operator wears on the left and right wrists with appropriate mounting wrist straps\footnote{\href{https://gopro.com/it/it/shop/mounts-accessories/hand-plus-wrist-strap/AHWBM-002.html}{https://gopro.com/it/it/shop/mounts-accessories/hand-plus-wrist-strap/AHWBM-002.html}}.
	\item A \textit{Raspberry Pi 4 Model B}\footnote{\href{https://www.raspberrypi.com/products/raspberry-pi-4-model-b/}{https://www.raspberrypi.com/products/raspberry-pi-4-model-b/}}, a fully functional computer with highly compact dimensions allowing the user to wear it comfortably.
	\item A compact battery bank, which provides power to the above components.
\end{itemize} 

The job of the worn computer is to handle the tracking cameras. In particular, it synchronizes the data coming from the cameras, computes the virtual forces, and forwards them through the network to the pilot PC where the Main Control Node is running.
The cameras are connected to the Raspberry Pi through \acrshort{usb} ports, while the Raspberry Pi can communicate with the pilot PC through by a Wi-Fi connection, thus eliminating the need of any tethering between the operator who wears the \acrshort{tpo} Suit and the rest of the system. 
It's worth noticing that the tethering can be set-up anyway in the case it is not necessary for the user to move too distant from the pilot PC. 

The T265 Camera is a \acrshort{vslam} tracking device composed of two fish-eye lens, an \acrshort{imu} and a \acrfull{vpu}. The technology running on the T265 camera exploits a combination of images and \acrshort{imu} data to track the movements of the camera itself, providing position, velocity and acceleration. All the algorithms are processed by the \acrshort{vpu}, allowing for low latency, efficient power consumption and easy integration.
Its small size, $(108 \, \times \, 25\ \times \, 13)$mm, and its light weight (\SI{55}{\gram} the single camera, \SI{114}{\gram} including the adapted mounting wrist strap) make it very comfortable to wear.
The T265 Camera and other tracking cameras are usually mounted on small mobile robots, especially drones, to track their position, but, to the best of the author knowledge, this technology was never been utilized in any relevant works to track parts of the human body. 
 
By mounting the two cameras on the operator's wrists, it is possible to track the motion of the arms with respect to a reference point in any direction, by reading the provided position. Hence, at every instant, the position of the two wrists with respect to their correspondent reference points are utilized to generate a pair of virtual forces that are applied to two different locations of the robot body. 

The computer of the \acrshort{tpo} Suit receives the cameras' data at a specific frequency of \SI{100}{\hertz}. For each camera connected, its actual pose is gathered as a transformation matrix $^{O}\boldsymbol{T}_{i} \in \mathbb{R}^{4\times 4}$. This matrix describes the transformation from the camera origin frame $O$ to the frame where the camera is at the moment, with the origin frame located in the point where the camera was switched on. Let's consider a reference frame $\mathit{ref}$ that is set during the initialization phase of the teleoperation and that can be reset at any time by the operator, which represents the \enquote{zero} position that generates a virtual force of zero magnitude. The transformation from the camera origin frame to the reference frame is denoted as $^{O}\boldsymbol{T}_{\mathit{ref}}$.
Each time that the camera data arrives, the relative transformation $^{\mathit{ref}}\boldsymbol{T}_i$ from the reference frame $\mathit{ref}$ to the actual frame is computed by a simple matrix multiplication:

\begin{equation}
	^{\mathit{ref}}\boldsymbol{T}_i = ^{O}\boldsymbol{T}_{\mathit{ref}}{}^{-1} ~ ^{O}\boldsymbol{T}_{i}
\end{equation}
where $^{-1}$ denotes the inverse of the transformation matrix.

From $^{\mathit{ref}}\boldsymbol{T}_i$ we extract the translation component $\boldsymbol{r} \in \mathbb{R}^{3\times 1}$, and associate it with the elongation of the virtual spring to compute the virtual force $\boldsymbol{f}_{\mathit{cp}} \in \mathbb{R}^{3\times 1}$ (\figurename{}~\ref{fig:tpo:tposuit}) as:

\begin{equation}\label{eq:tpo:f_cp}
	\boldsymbol{f}_{\mathit{cp}} = k_{\mathit{cam}} ~ s(\boldsymbol{r})
\end{equation}
where $k_{\mathit{cam}}$ is a positive gain representing the stiffness of the virtual spring; $s(\cdot)$ represents a simple filter to smooth out erratic movements of the operator arms, and a small spherical \enquote{buffer space} around the reference frame which cuts to zero all the movements inside it. 

The vector $\boldsymbol{f}_{\mathit{cp}}$ describes the virtual force applied by one input on the selected robot part, i.e., the control point $\mathit{cp}$. 
When multiple input sources are utilized, multiple virtual forces are computed, e.g., considering the \figurename{}~\ref{fig:tpo:tposuit} two virtual forces, $\boldsymbol{f}_{\mathit{cp,l}}$ and $\boldsymbol{f}_{\mathit{cp,r}}$ are generated from the left and right camera, respectively, on two different control points. The virtual forces gathered from the inputs are aggregated and sent through the network to the main control node through \acrshort{ros} topics.
The architecture built permits to switch on the fly where the virtual forces must be applied, i.e., the control points $\mathit{cp}$, by means of \acrshort{ros} services. 

In this chapter we are referring of the above-described version of the \acrshort{tpo} Suit.
In the Chapter~\ref{chap:TPOH}, the interface is enhanced by integrating wearable vibrotactile devices to add a haptic feedback channel, and additional input buttons to expand the user's possibilities.

\subsection{TelePhysicalOperation Main Control Node} \label{sec:tpo:control}

\begin{figure}[H]
	\centering
	\includegraphics[width=0.8\linewidth]{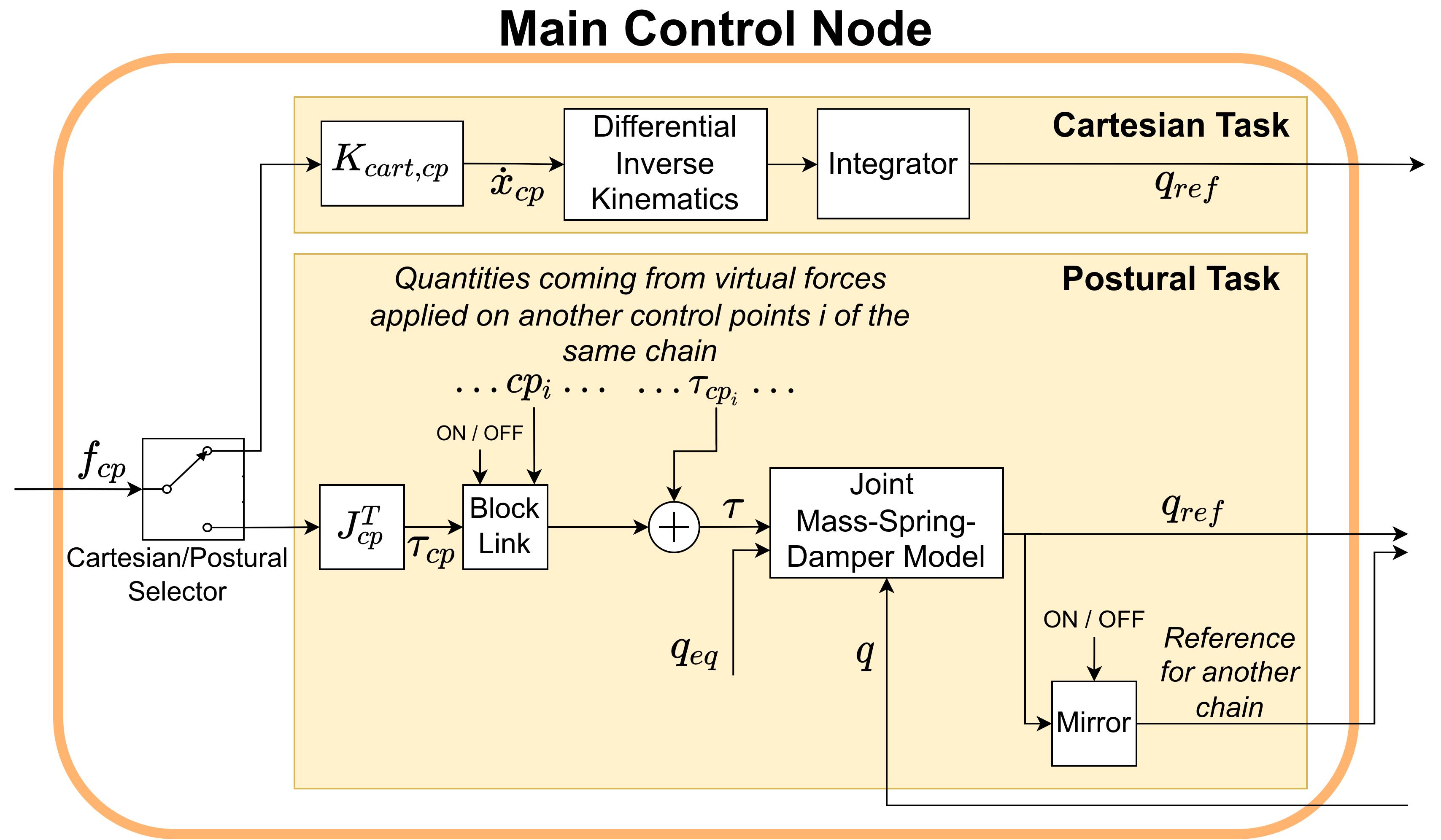}
	\caption[TelePhysicalOperation main dode]{The TelePhysicalOperation Main Control Node is in charge of generating robot motions from the virtual forces $\boldsymbol{f}_{\mathit{cp}}$ provided on specific control point $\mathit{cp}$.}
	\label{fig:tpomainnode}
\end{figure}

The TPO Main Control Node of \figurename{}~\ref{fig:tpomainnode} handles the incoming virtual forces $\boldsymbol{f}_{\mathit{cp}}$ and generates the joint command references $\boldsymbol{q}_{\mathit{ref}}$ for the robot. Depending on which robot body part the virtual force is applied to, it can be processed in two ways:

\begin{itemize}
	\item \textbf{Postural motion generation} (Section~\ref{sec:tpo:pos}). The virtual force is processed for a \textit{Postural} task, i.e., a task that considers the joint-level of the robot. A joint-level admittance control law is utilized to generate the joint command references $\boldsymbol{q}_{\mathit{ref}}$. Multiple virtual forces can be applied to different control points (i.e.\ links) of the same kinematic chain, generating motions that comply to all the virtual forces applied.
	
	\item \textbf{Cartesian motion generation} (Section~\ref{sec:tpo:vel}). The virtual force is utilized for a \textit{Cartesian} task, by processing it to generate a Cartesian linear velocity reference for the control point where the virtual force is applied. Hence, an inverse kinematic process is involved to compute the joint command references $\boldsymbol{q}_{\mathit{ref}}$ from the Cartesian velocity reference.
	
\end{itemize}

\subsubsection{Postural Motion Generation} \label{sec:tpo:pos}

To obtain the references of the postural motion task, the virtual force $\boldsymbol{f}_{\mathit{cp}}$ is considered as a force acting on the control point $\mathit{cp}$ selected at a location along the $N$-joint kinematic chain of the robot. 
From $\boldsymbol{f}_{\mathit{cp}}$, the resulting torques on the chain's joints $\boldsymbol{\tau}_{\mathit{cp}} \in \mathbb{R}^{N\times 1}$ are computed : 

\begin{equation}\label{eq:tpo:taocp}
	\boldsymbol{\tau}_{\mathit{cp}} = \boldsymbol{J}_{\mathit{cp}}^T ~ \boldsymbol{f}_{\mathit{cp}}
\end{equation}
where $^T$ denotes the transpose of a matrix, and $\boldsymbol{J}_{\mathit{cp}} \in \mathbb{R}^{3\times N}$ is the linear Jacobian matrix of the control point chosen, i.e., the matrix such that its product with the derivative of the configuration vector results in the Cartesian linear velocity of $\mathit{cp}$. 
It is worth noticing that, if the control point is not on the last link of the robot chain, the virtual force will not influence the joints \textit{after} the chosen link because the final columns of the Jacobian are filled with zeros. This is exactly what happens in a physical human-robot interaction when the user guides the collaborative robot by touching it on a link which is not the last one.

In case that more than one virtual force is acting on different control points on the same kinematic chain, their contributions will be summed to a total joint torque $\tau$: 

\begin{equation}\label{eq:taochain}
	\boldsymbol{\tau} = \sum_{\mathit{cp} \; \in \; \mathit{chain}} \boldsymbol{\tau}_{\mathit{cp}}
\end{equation}
If the virtual forces are acting on different kinematic chains (e.g., on two different arms) each $\boldsymbol{\tau}$ is handled separately.

For each controlled kinematic chain, having derived the corresponding total joint torque due to the application of the virtual forces, the joint reference motion is computed by considering a joint mass-spring-damper model (\figurename{}~\ref{fig:controllerschemeModes}) as follows:

\begin{equation} \label{eq:tpo:control}
\begin{gathered}
	\boldsymbol{\ddot{q}}_{\mathit{ref}}(t) = \boldsymbol{M}_{tpo}^{-1} \big ( \boldsymbol{K}_{tpo} (\boldsymbol{q}_{eq} - \boldsymbol{q}(t)) - \boldsymbol{D}_{tpo} \boldsymbol{\dot{q}}_{\mathit{ref}}(t-1)+\boldsymbol{\tau} \big) \\
	\boldsymbol{\dot{q}}_{\mathit{ref}}(t) = \boldsymbol{\dot{q}}_{\mathit{ref}}(t-1) + \boldsymbol{\ddot{q}}_{\mathit{ref}}(t) ~ \Delta t \\
	\boldsymbol{q}_{\mathit{ref}}(t) = \boldsymbol{q}_{\mathit{ref}}(t-1) + \boldsymbol{\dot{q}}_{\mathit{ref}}(t) ~ \Delta t
\end{gathered}
\end{equation}
where $\boldsymbol{q}_{\mathit{ref}}(t) \in \mathbb{R}^{N\times 1}$ is the joint position reference vector; $\boldsymbol{M}_{tpo}, \boldsymbol{K}_{tpo}, \boldsymbol{D}_{tpo} \in \mathbb{R}^{N\times N}$ are diagonal matrices of the mass, stiffness and damping parameters of the joint mass-spring-damper model; $\boldsymbol{q} \in \mathbb{R}^{N\times 1}$ is the current position of the joints; $\boldsymbol{q}_{eq} \in \mathbb{R}^{N\times 1}$ is the equilibrium set point where a stiffness greater than zero will drag the joints; $\Delta t$ is the time interval between two consecutive control loops.

The parameters of the mass-spring-damper model can be set experimentally to regulate the motion behavior of the individual joints that are subject to the torques derived by \eqref{eq:taochain}. As an example, the diagonal elements of the stiffness $\boldsymbol{K}$ can be set to zero to eliminate the returning elastic torque towards the equilibrium set point of the joints. 
Instead, by increasing these elements, the joints will elastically tend to go back to their $\boldsymbol{q}_{eq}$ set point if the virtual force is not sufficiently strong to win against this returning motion.

\subsubsection{Cartesian Motion Generation} \label{sec:tpo:vel}

The postural based motion control of Section~\ref{sec:tpo:pos} provides full flexibility to the operator enabling to control the robot at the joint space based on the control points selected on the robot body. 
This kind of motion control lets the user regulate as needed the end-effector pose as well as the available kinematic redundancy if it exists. 
In some situation though, the possibility to regulate the motions of the remote robot at the task (Cartesian) space may be required, and it may be more appropriate to facilitate the execution of some tasks. 
For example, the operator can apply a virtual force on the control point set in the base of a wheeled mobile robot, with the intention to move the whole mobile base. 
Another challenging Cartesian task can be the control of the pose of a quadruped pelvis by selecting a control point on the pelvis and driving its pose using the virtual force. Such kind of tasks would be challenging if not impossible to be carried out through postural motion generation as they involve motions of the wheels or the several legs joints that contributes, for example, to the (\enquote{squatting}) motion of the pelvis.

To realize the functionality of regulating specific links of the robot body in the corresponding task space, the TelePhysicalOperation method provides a Cartesian motion generation interface. 
After selecting a control point, from the applied virtual force $\boldsymbol{f}_{\mathit{cp}}$, it is derived a velocity reference $\boldsymbol{\dot{x}}_{\mathit{cp}} \in \mathbb{R}^{3\times 1}$ as follows:

\begin{equation}\label{eq:x_cp}
    \boldsymbol{\dot{x}}_{\mathit{cp}} = \boldsymbol{K}_{\mathit{cart,cp}} ~\boldsymbol{f}_{\mathit{cp}}
\end{equation}
where $\boldsymbol{K}_{\mathit{cart,cp}} \in \mathbb{R}^{3\times 3}$ is a diagonal matrix of gains based on the Cartesian task specific for the control point $\mathit{cp}$.

This Cartesian motion generation interface permits also to limit the possible Cartesian directions to comply with the physical constraints of the robot or the specific requirements of the Cartesian task. This can be done by putting the correspondent elements in the diagonal of $\boldsymbol{K}_{\mathit{cart,cp}}$ equal to zero, for example to control the mobility of a planar mobile robot, which can not follow a velocity along the $z$ axis.

From the resulting Cartesian velocity reference $\boldsymbol{\dot{x}}_{\mathit{cp}}$, a joint reference $\boldsymbol{q}_{\mathit{ref}}$ is derived with an inverse kinematic process.

\subsection{Visual Feedback Tools}\label{sec:tpo:visual}

\begin{figure} [H]
	\centering
	\includegraphics[width=0.24\linewidth]{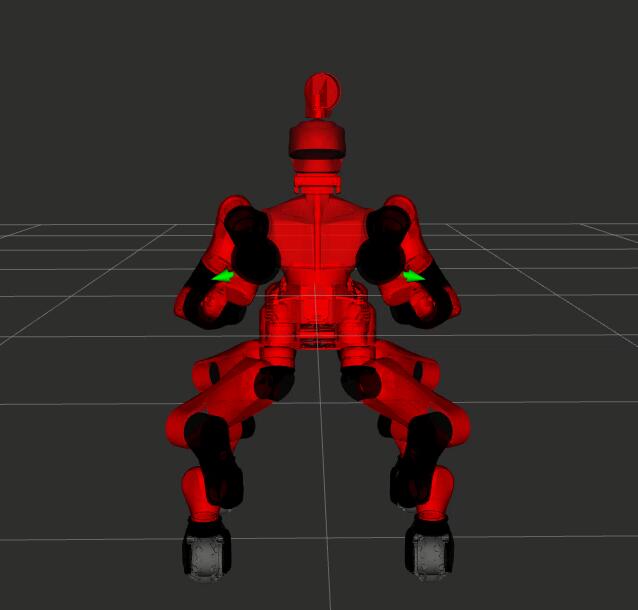}
	\includegraphics[width=0.24\linewidth]{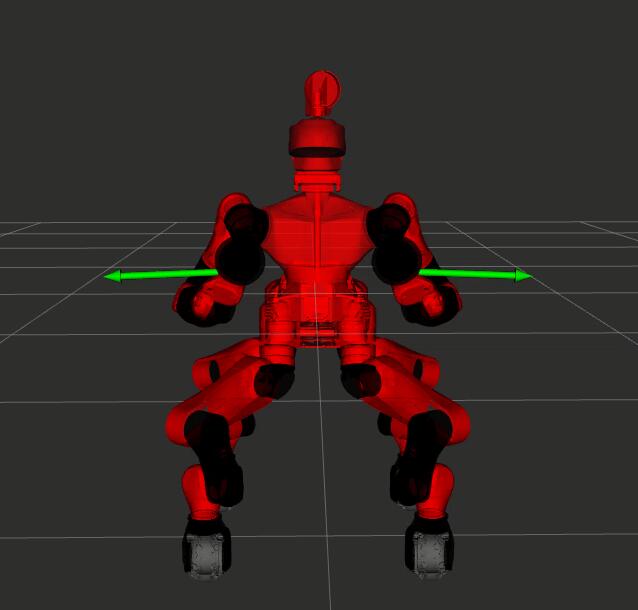}
	\includegraphics[width=0.24\linewidth]{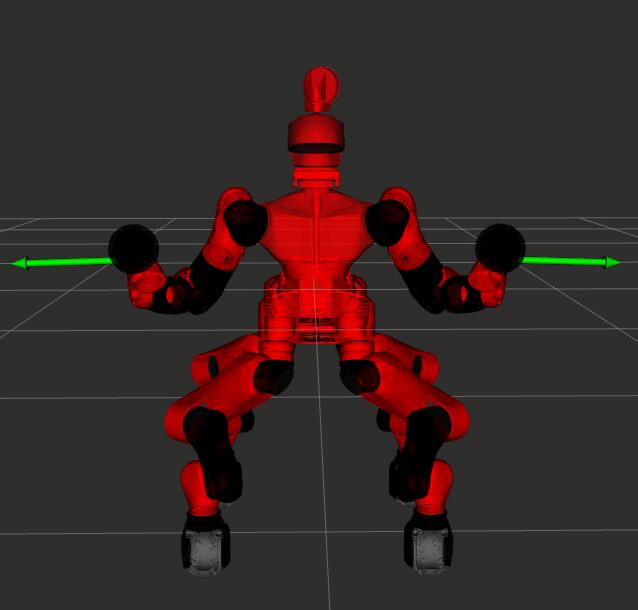}
	\includegraphics[width=0.24\linewidth]{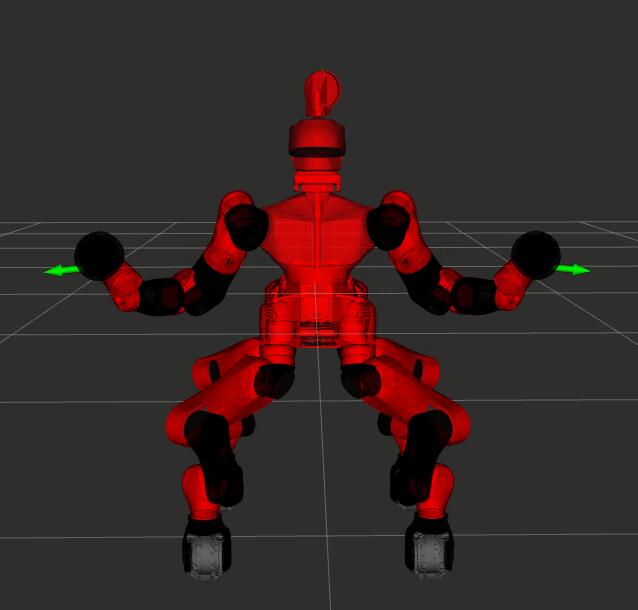}\\
	\vspace{15px}
	\includegraphics[width=0.24\linewidth]{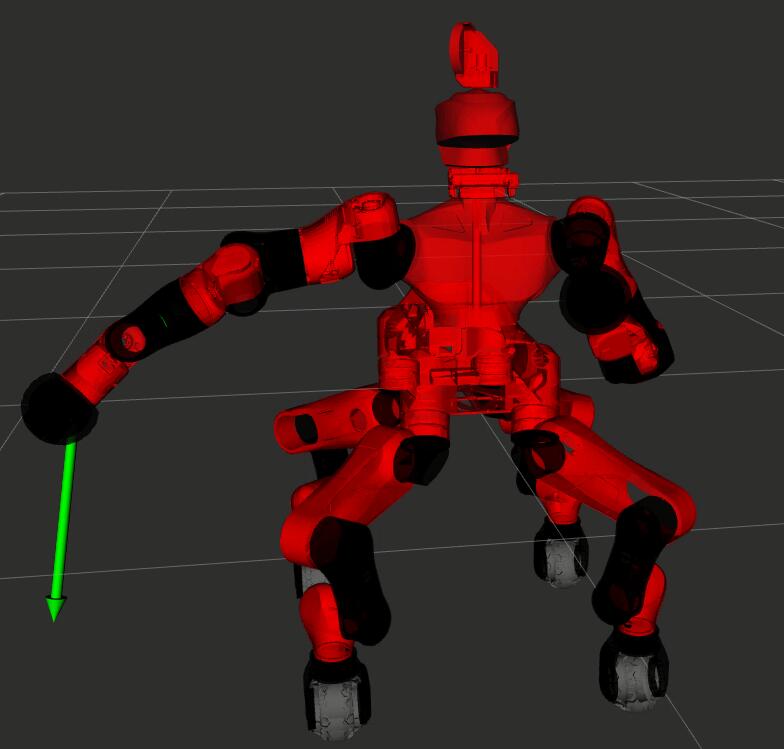}
	\includegraphics[width=0.24\linewidth]{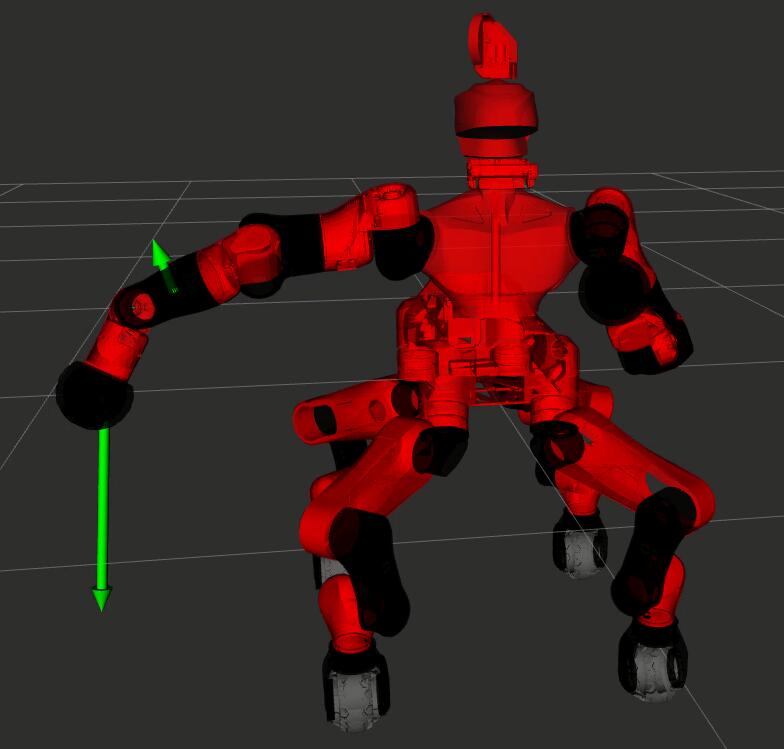}
	\includegraphics[width=0.24\linewidth]{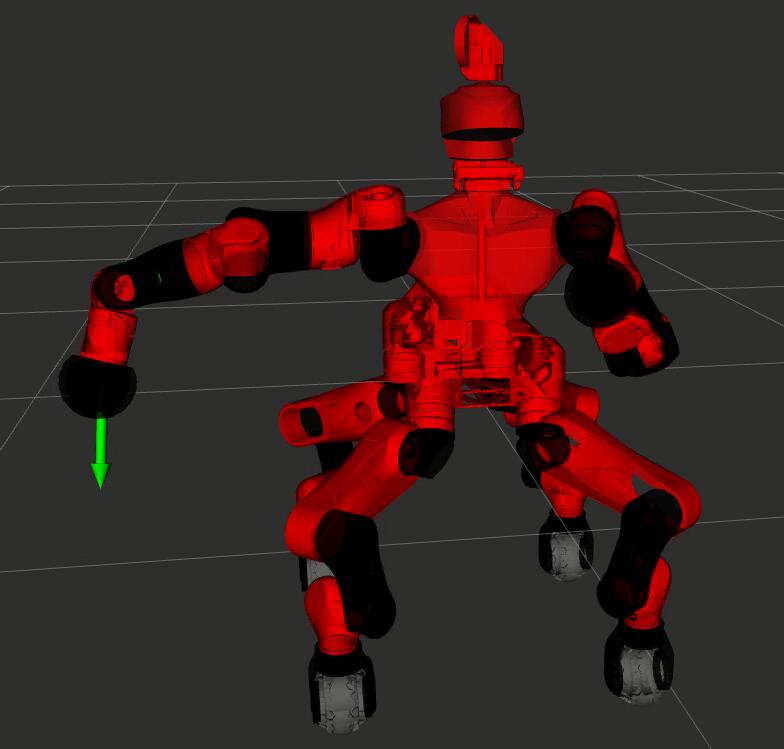}
	\includegraphics[width=0.24\linewidth]{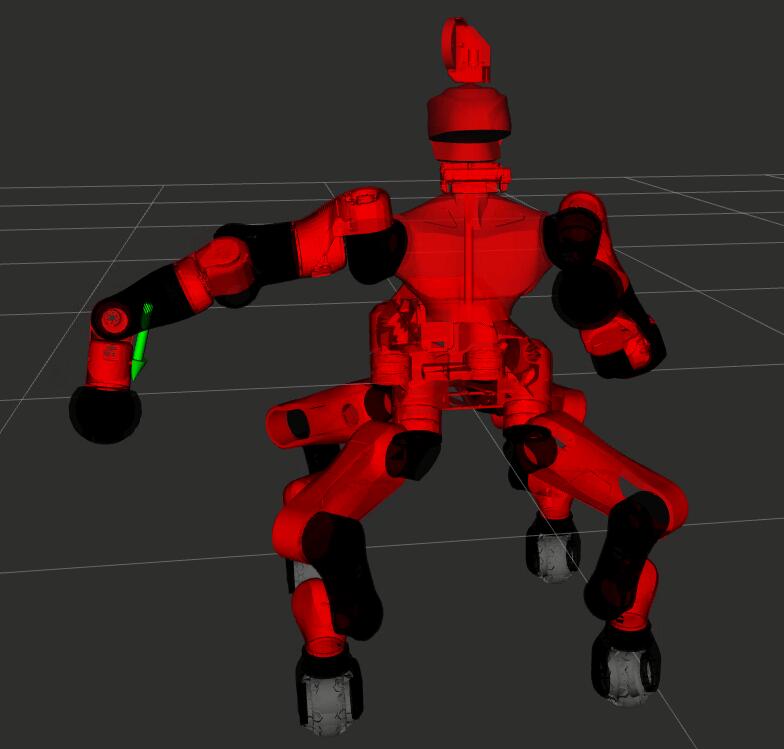}\\
	\vspace{15px}
	\includegraphics[width=0.24\linewidth]{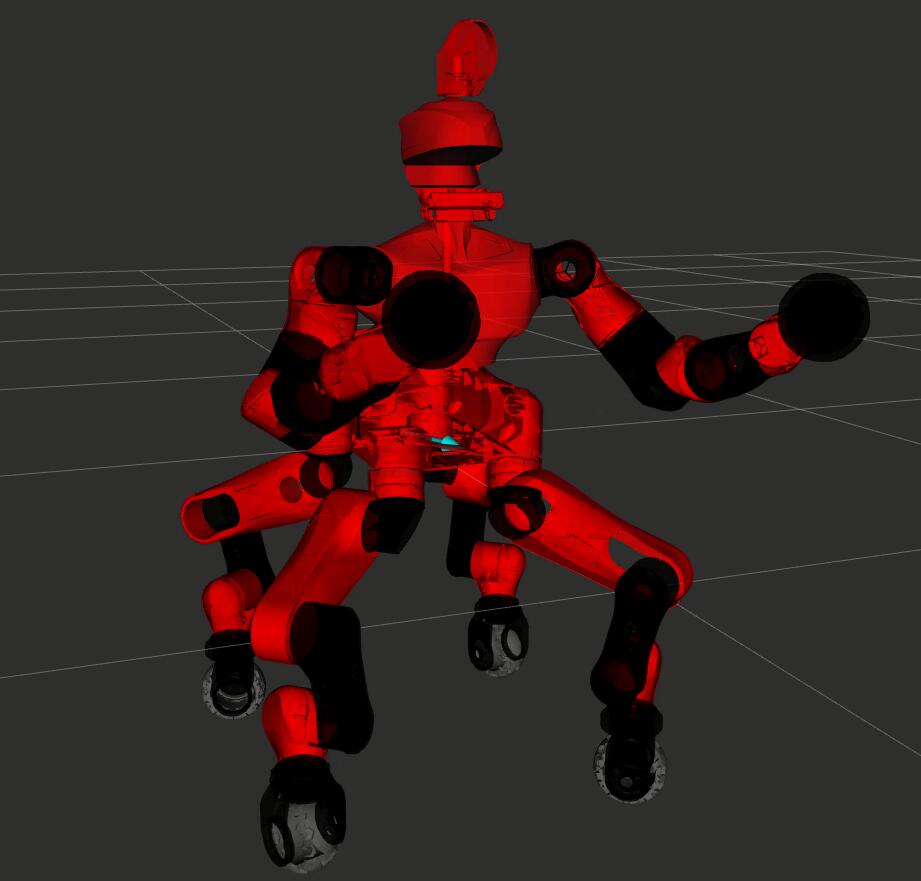}
	\includegraphics[width=0.24\linewidth]{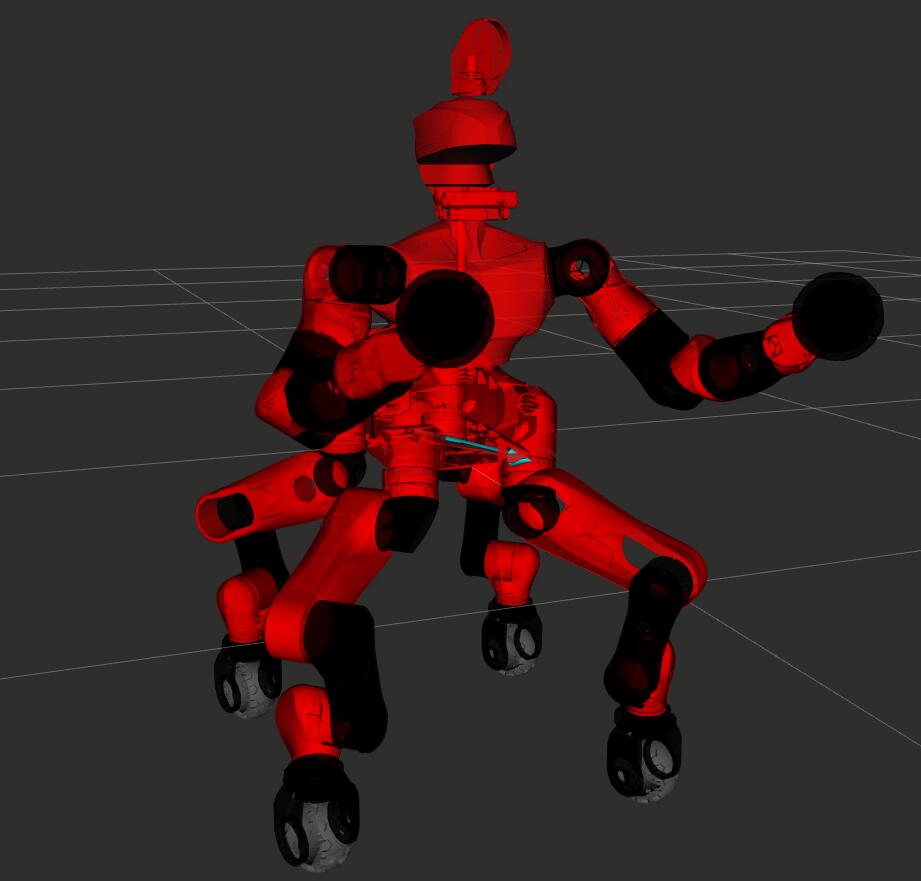}
	\includegraphics[width=0.24\linewidth]{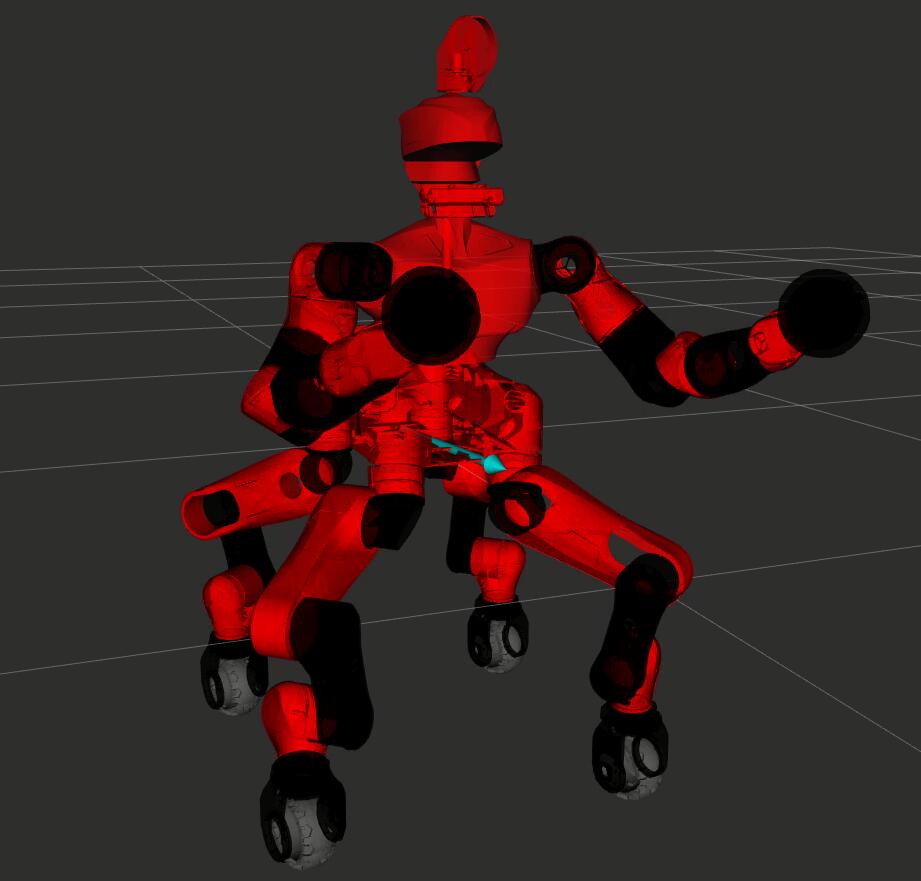}
	\includegraphics[width=0.24\linewidth]{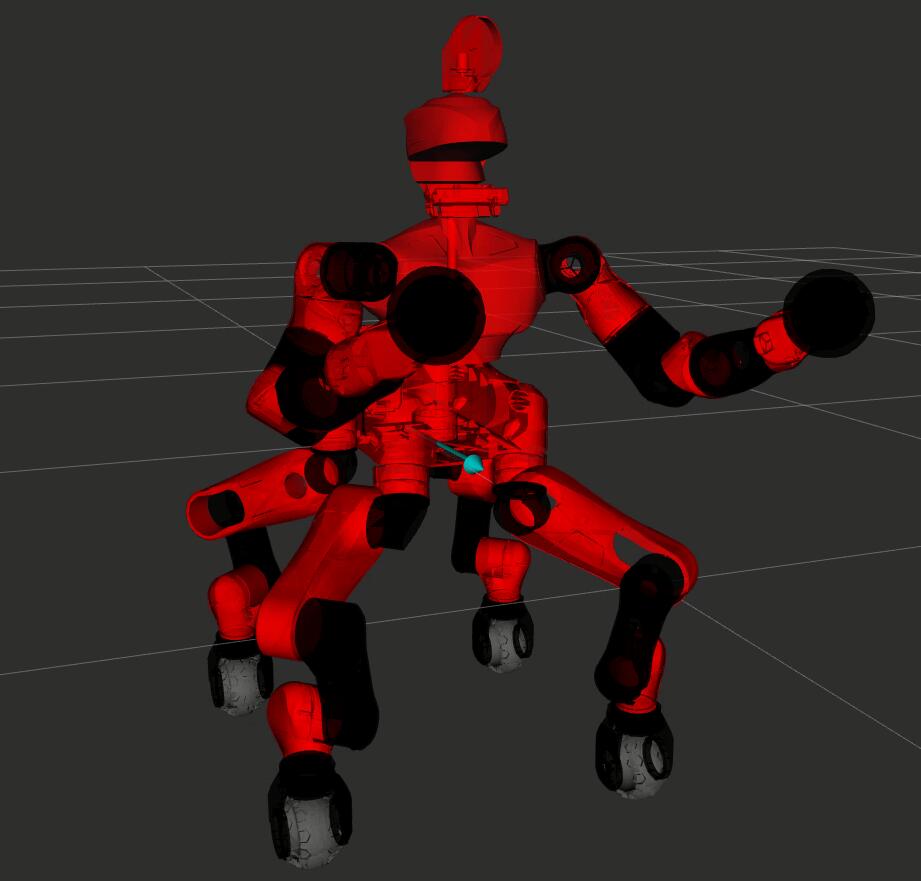}
	\caption[Visual tepresentation of TPO virtual forces]{Sequences that show the CENTAURO robot in the RViz visualizer, with arrows representing the magnitude and directions of the \acrshort{tpo} virtual forces applied by the operator.}
	\label{fig:tpo:rvizArrows}
\end{figure}

To complete the \acrshort{tpo} architecture, graphical information has been added to the human-robot interface to assist the operator. In \figurename{}~\ref{fig:tpo:rvizArrows} the kinematic model of the CENTAURO robot is shown in the RViz visualizer. The arrows visually represent the magnitude and directions of the virtual forces applied to different parts of the robot's body by the operator. A longer arrow represents a higher virtual force magnitude resulting from a more substantial displacement of the operator's wrist with respect to the reference frame $\mathit{ref}$.
The color of the arrow conveys the teleoperation status: yellow indicates deactivated teleoperation (not shown in the figure), while green (for postural task, like the arms, first and second rows) and light blue (for Cartesian task, like the body, last row) signifies activated teleoperation, implying that the robot is moving under the influence of the operator's arm movements.

The sequences in \figurename{}~\ref{fig:tpo:rvizArrows} are extracted from the experiments presented in Section~\ref{sec:tpo:exp}, in particular from the Section~\ref{sec:tpo:arms}, Section~\ref{sec:tpo:btn}, and Section~\ref{sec:tpo:loco}, for the first, second and third rows, respectively. 
In the first row, the operator is applying two virtual forces on the robot end-effectors. In the second, two virtual forces are applied to the same robot's arm. In the third row, a virtual force is applied on the robot's body to push it forward, consequently moving the entire platform.

\subsection{Toward Robot Autonomy Features}\label{sec:tpo:auto}

As stated in the introductory chapters, together with intuitiveness the human-robot interface should also present some autonomy features to help the operator in controlling the complexity of the robot.
For the mentioned reasons, autonomy features have been developed and integrated in the TelePhysicalOperation architecture. While more advanced functionalities are presented in the dedicated Chapter~\ref{chap:tpoAuto}, in this section two of them are presented: the \textit{Blocking Link} feature, and the \textit{Mirroring Motion} feature (Section~\ref{sec:tpo:blocklink} and Section~\ref{sec:tpo:mirror}).

\subsubsection{The Blocking Link Feature}\label{sec:tpo:blocklink}

As explained previously, more virtual forces can be applied to different parts of the same kinematic chain, generating joint motions that comply to such forces.
According to \eqref{eq:tpo:taocp} and \eqref{eq:taochain}, each virtual force applied on a control point will contribute to command only the joint of the chain from the chain's root up to the last joint before the control point, which is what it would happen in a real physical human-robot interaction. This results from the fact that, if the selected control point is not on the last link of the chain, the columns of the Jacobian relative to the joints \textit{after} the control point are filled with zero. 
However, when two virtual forces are applied, both of them contribute to the motions of the firsts joints of the chain. 
There are situations in which the operator may want to move only specific joints in the middle of the robot chain without influencing the position of the first joints of the chain. This functionality is accomplished by activating the \textit{Blocking Link} feature.

Let's consider two virtual forces applied on different control points of a chain, and denote them with $\boldsymbol{f}_A$, and $\boldsymbol{f}_B$, applied on the control points $A$ and $B$ positioned in two different links of the chain. Let's assume that $A$ is on a robot's link nearer to the root of the chain with respect to the link where the control point $B$ is. With the \textit{Blocking Link} feature, $\boldsymbol{f}_B$ will generate motions only for the joints between the links of $A$ and $B$ control points. Hence, the joints present between the chain's root and the link of $A$ can be moved only by $\boldsymbol{f}_A$.

For example on the CENTAURO robot, one may want to move precisely the wrist joint while keeping the shoulder and elbow joints fixed, as it can be seen in some sequences of an experiment that will be presented later (second, third and fourth image of \figurename{}~\ref{fig:TPOExpButton}). In the images, the virtual force applied on the end-effector is influencing only the wrist joint, because of the presence of another virtual force on the robot forearm, which is the only virtual force that can influence the shoulder and elbow joints. This permits to bend only the wrist joint to avoid more precisely the obstacle.

\subsubsection{The Mirroring Motion Feature}\label{sec:tpo:mirror}

For a dual-arm manipulation system, there are situations in which it is useful to teleoperate the two twin arms in a specular and symmetrical manner. Hence, there would not be the necessity of a second input on the second arm since the necessary motion can be provided by a single virtual force on the first arm that can be mirrored on the second arm.

With TelePhysicalOperation, it can be activated the so-called \textit{Mirroring Motion} feature to enable this functionality. With this feature, the virtual forces applied on a control point of an arm are also applied on the correspondent control point of the other arm, symmetrically (respect to a plane perpendicular to the floor). 

A showcase of this feature is visible in the demonstration with the CENTAURO robot presented in the next section (shown in the two bottom images of \figurename{}~\ref{fig:TPOExpArms}). There, the force applied on the right end-effector by the user right arm has been mirrored to the left end-effector.

In another experiment the feature has been activated during a teleoperated bimanually transportation of a load (as shown in the last image of \figurename{}~\ref{fig:TPOExpBox}). In this particular experiment, the mirroring helped in placing the box in the final location. Nevertheless, it is important to notice that, for bimanual tasks of this nature, more advanced and specialized solutions have been developed, as discussed in Chapter~\ref{chap:tpoAuto}.

\section{TelePhysicalOperation Experimental Validations}\label{sec:tpo:exp}

Experimental validations about the \acrlong{tpo} concepts presented in this chapter have been conducted with the CENTAURO robot, briefly introduced in Section~\ref{sec:intro:centauro}. The experiments demonstrate the flexibility in controlling different capabilities of the CENTAURO robot, from the manipulation to the locomotion.

During these experiments the operator applies virtual forces to different robot body locations (i.e., the control points) to guide the robot and complete different tasks. In the \acrlong{tpo} configuration employed, the control points are selected by a second operator, according to the instruction provided by the first one. It is important to note that the need for a second operator has been eliminated with the improvements discussed in the Chapter~\ref{chap:TPOH}, and anyway reduced through the development of the autonomy features presented in Chapter~\ref{chap:tpoAuto}.
 
The parameters of the joint mass-spring-damper model present in \eqref{eq:tpo:control} have been experimentally tuned to adjust the sensitivity of the robot generated motions. The virtual spring stiffness of the cameras $k_{\mathit{cam}}$ has been experimentally set to \SI{1.8}{\newton\meter^{-1}} based on the level of sensitivity in the motion produced that felt comfortable by the human operator.

All the following experiments have been recorded, resulting in the video available at \href{https://youtu.be/dkBmbTyO_GQ}{https://youtu.be/dkBmbTyO\_GQ}.

\subsection{Latency Computation}

\begin{figure}[H]
	\begin{minipage}[c]{.60\linewidth}
		\centering
		\includegraphics[width=\linewidth]{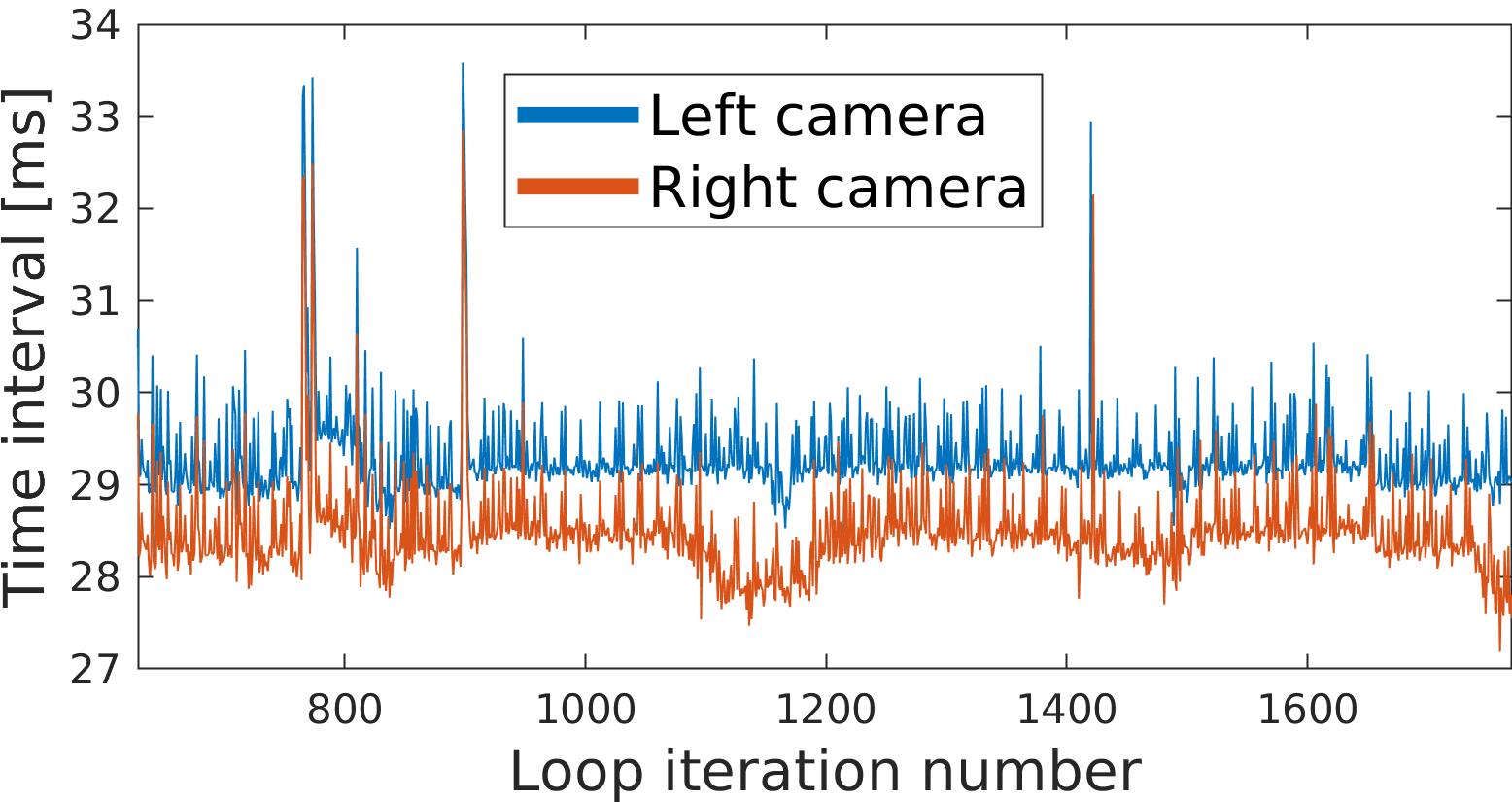}
	\end{minipage}
	\hspace{10px}
	\begin{minipage}[c]{.4\linewidth}%
		\vspace{-17px}
		\begin{tabular}{lcc}
			& \begin{tabular}[c]{@{}c@{}}{\small Mean}\\ {{\small [ms]}}\end{tabular} & \begin{tabular}[c]{@{}c@{}}{\small Std Dev}\\ {{\small [ms]}}\end{tabular} \\
			\toprule
			{\small  \textcolor{matlabBlue}{Left cam}}  & {\small 29.30} & {\small 0.47} \\
			{\small  \textcolor{matlabRed}{Right cam}} & {\small 28.48} & {\small 0.50} \\          
			\bottomrule                                             
		\end{tabular}
	\end{minipage}
	\hfill
	\caption[Latency introduced by the TelePhysicalOperation framework]{Latency introduced by the TelePhysicalOperation framework. The time interval is calculated by measuring the interval from when the data is polled from each camera to when the command references are sent to the robot. Considering \figurename{}~\ref{fig:controllerschemeModes}, from the leftmost side of \enquote{Embedded PC} block to the rightmost side of \enquote{Main Control Node} block.}
	\label{fig:TPOlatency}
\end{figure} 

It has been conducted a study regarding the latency introduced by the TelePhysicalOperation framework, where the embedded PC connected with the cameras was communicating with the pilot PC through \mbox{Wi-Fi} connection. In \figurename{}~\ref{fig:TPOlatency} results are shown. The latency for each camera is calculated by measuring the time from when the camera data is acquired, to when the joint commands are transmitted to the robot. Referring to \figurename{}~\ref{fig:controllerschemeModes}, this interval begins at the leftmost side of the \enquote{Embedded PC} block and ends at the rightmost side of the \enquote{Main Control Node} block.
On average, the latency has resulted to be less than \SI{30}{\milli\second}. 
The slight difference between the two cameras is due to the fact that the data from the right camera is gathered before the data from the left one.

\subsection{Robot's Arms Workspace Exploration}\label{sec:tpo:arms}
\begin{figure}[H]
	\centering
	\includegraphics[width=.49\linewidth,keepaspectratio]{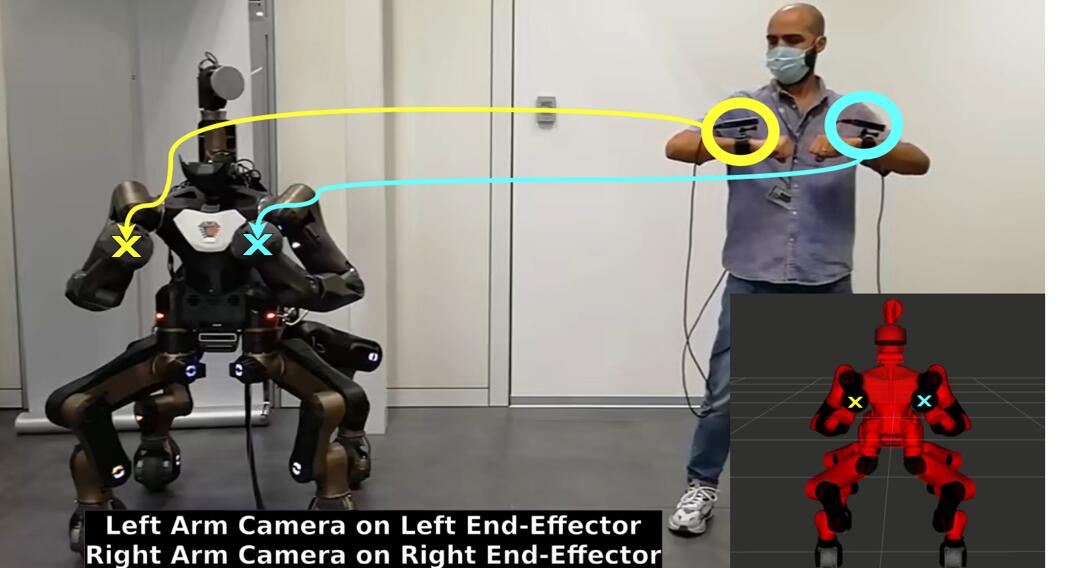}
	\includegraphics[width=.49\linewidth,keepaspectratio]{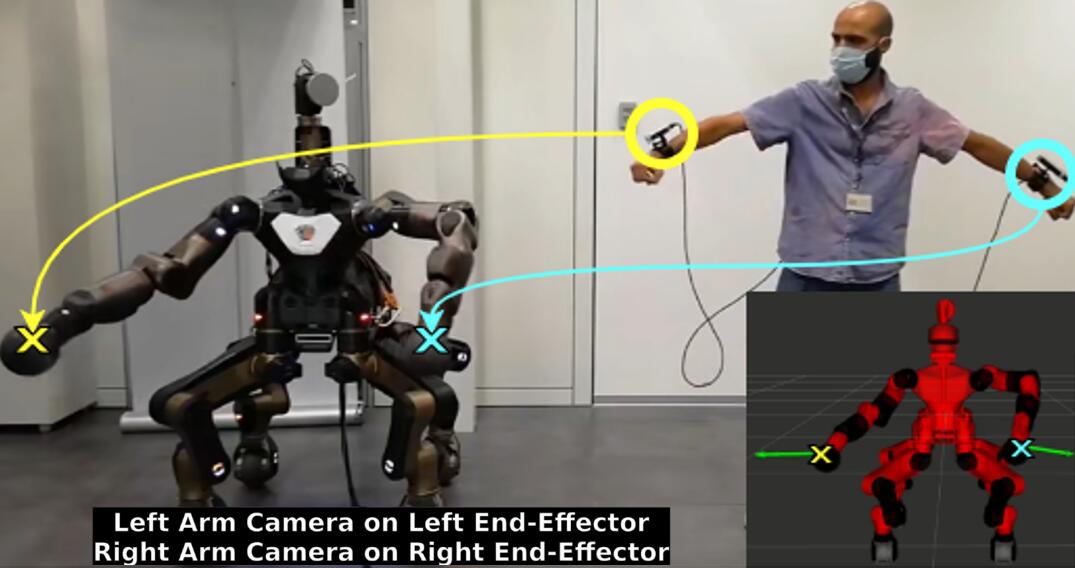}
	\\
	\vspace{5px}
	\includegraphics[width=.49\linewidth,keepaspectratio]{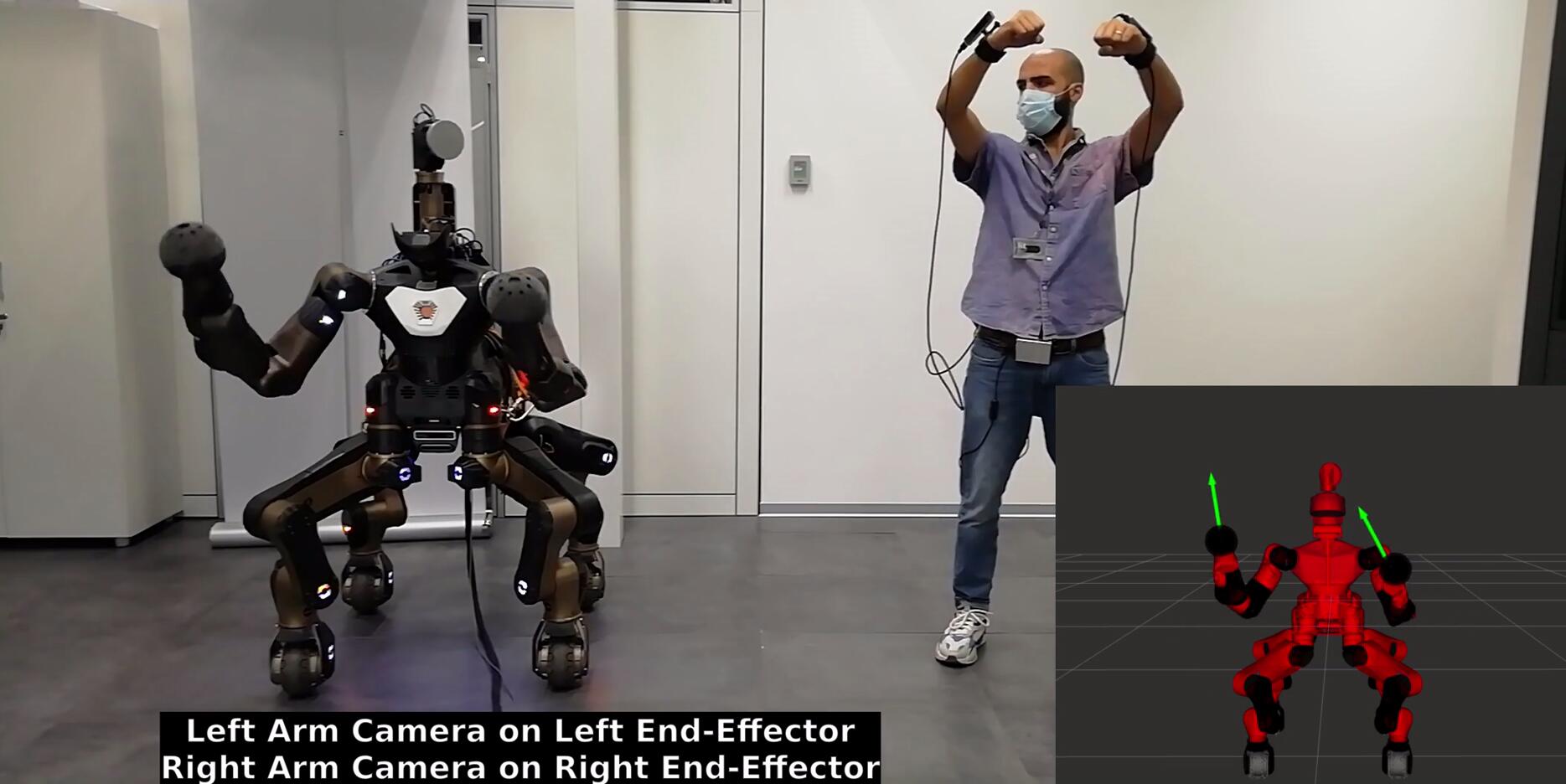}
	\includegraphics[width=.49\linewidth]{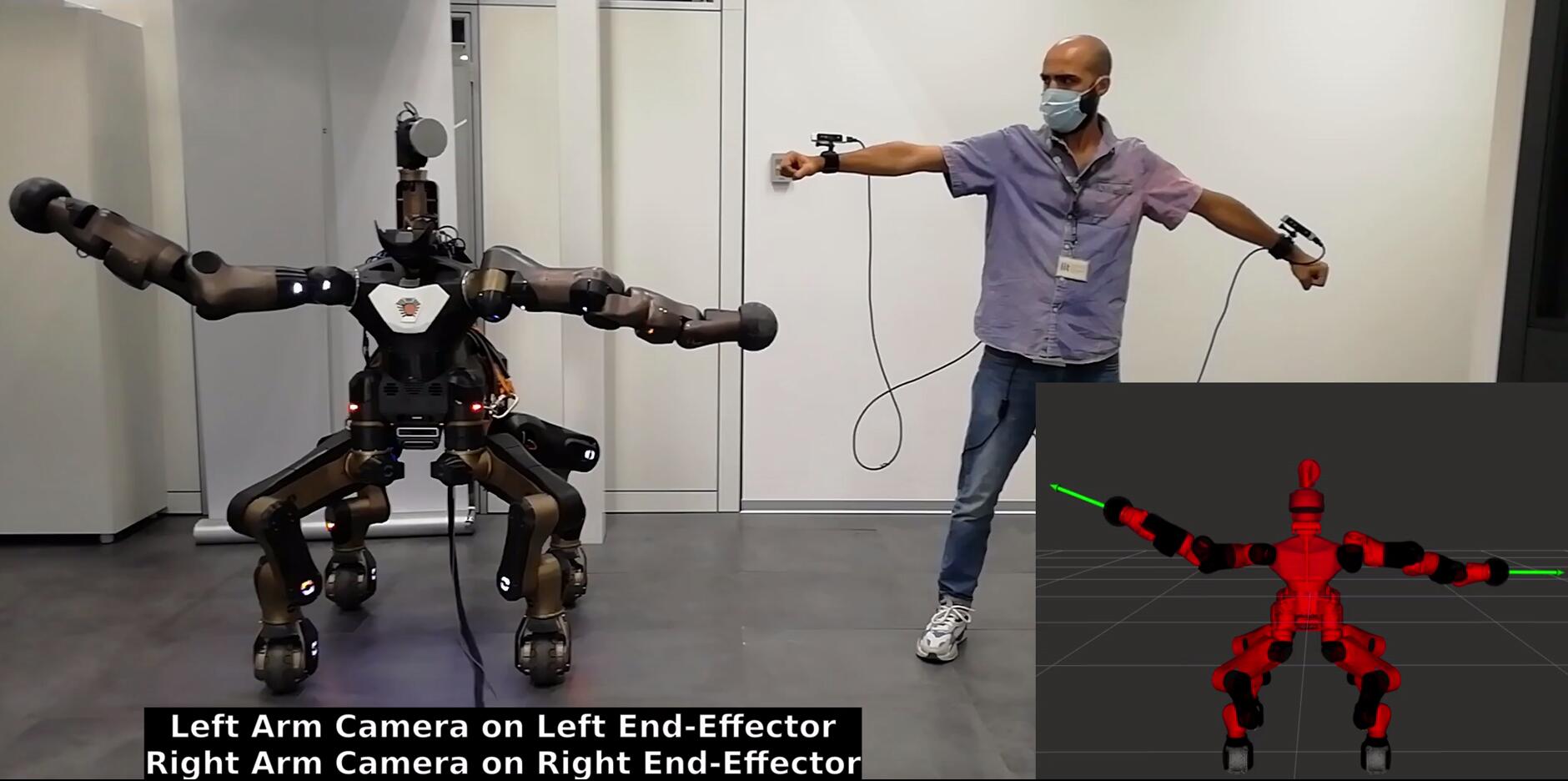}\\
	\vspace{5px}
	\includegraphics[width=.49\linewidth,keepaspectratio]{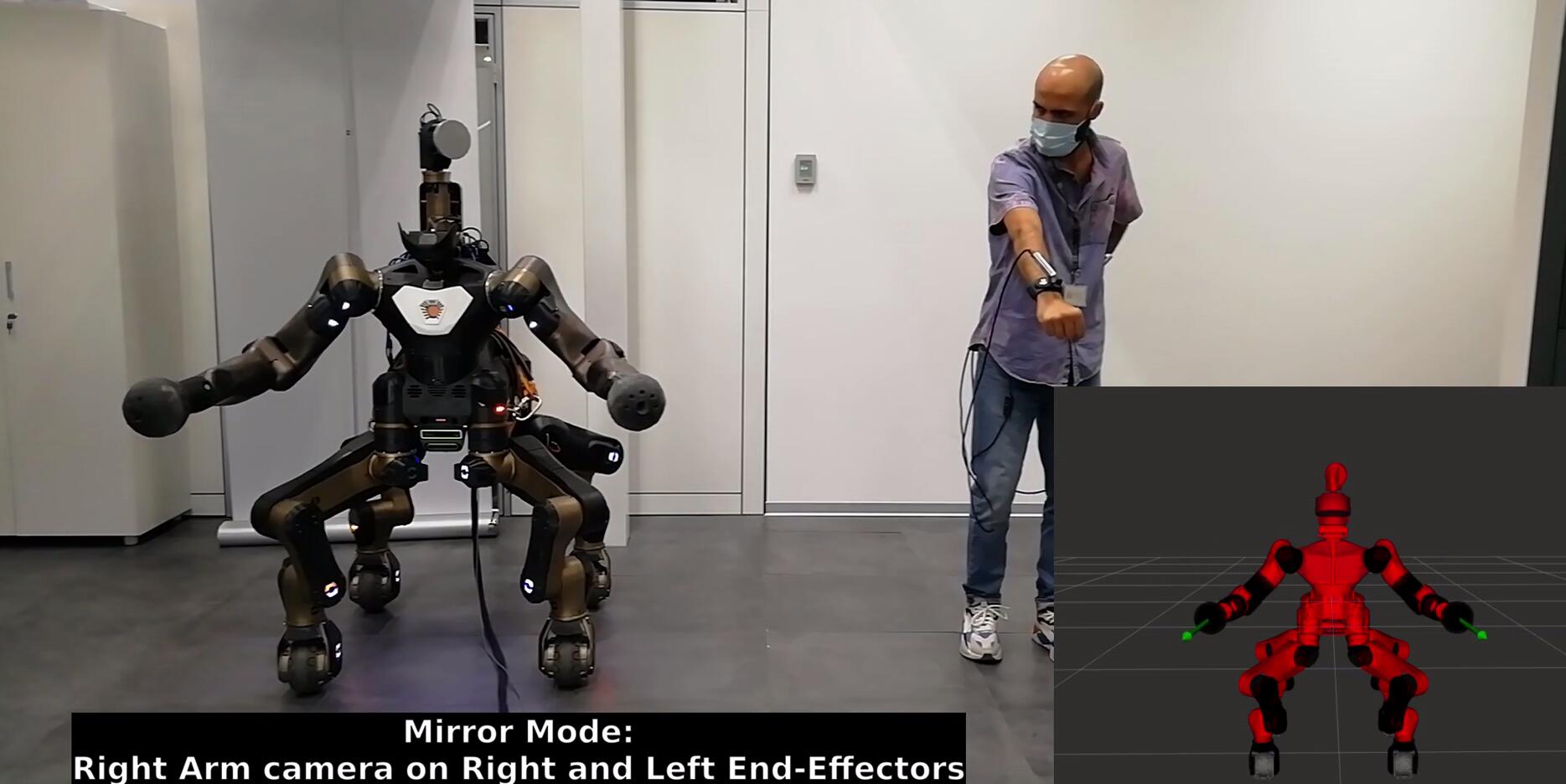}	
	\includegraphics[width=.49\linewidth,keepaspectratio]{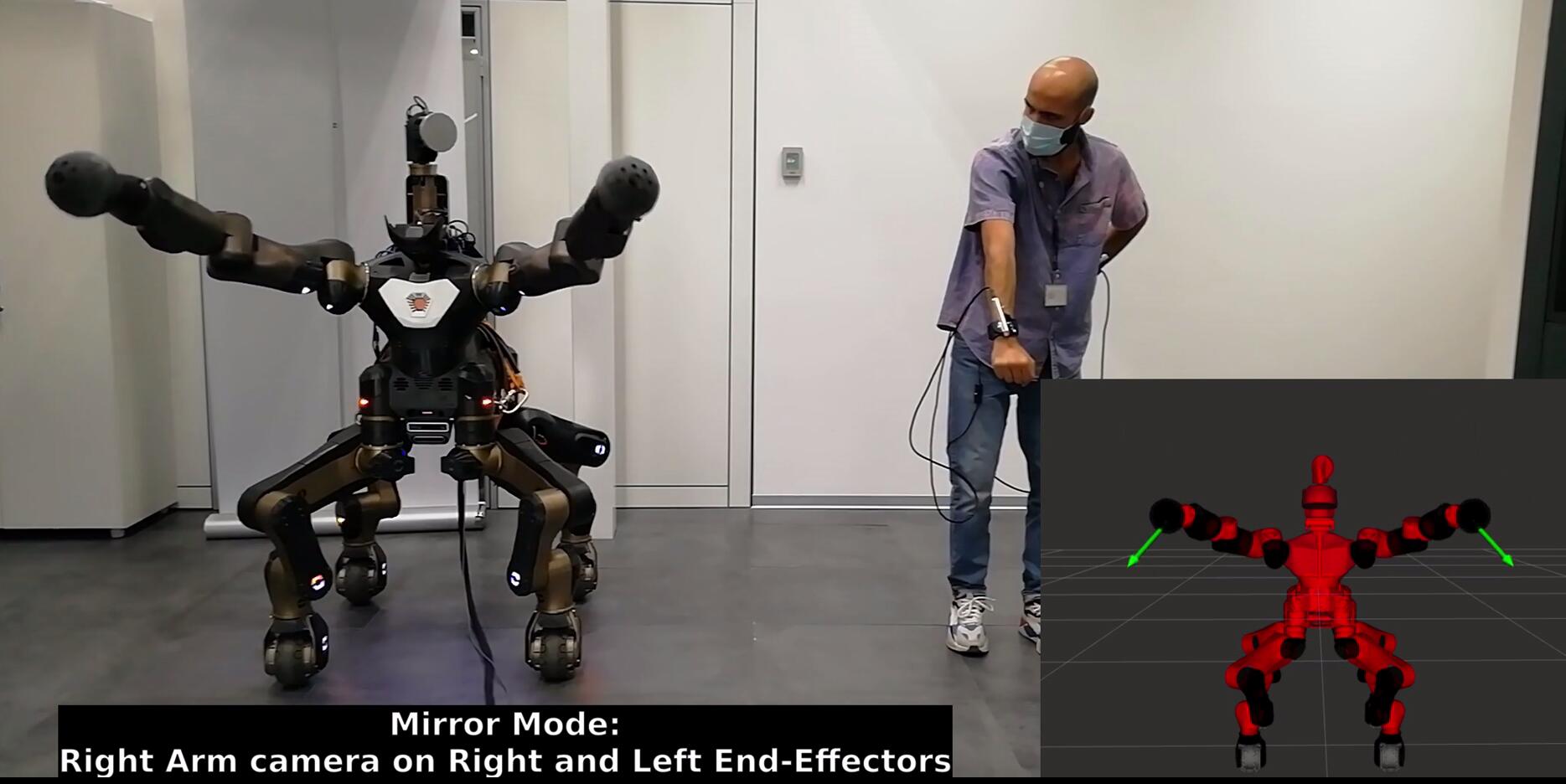}	
	\caption[TelePhysicalOperation: CENTAURO's arms workspace exploration]{CENTAURO's arms are moved by the application of virtual forces at the end-effectors with the TelePhysicalOperation interface. In first four images, each operator's arm is applying a virtual force on the respective robot end-effector. In the two bottom images, the \textit{Mirroring Motion} feature is activated: with one arm the operator is applying two specular virtual forces on both robot end-effectors.}
	\label{fig:TPOExpArms}
\end{figure}

This experiment showcases how the arms of CENTAURO are commanded within their workspace. Sequences of these motions are visible in \figurename{}~\ref{fig:TPOExpArms}. In the first four images, the operator applies virtual forces on the robot left and end-effector with his left and right arm, respectively.
In the two bottom images, the \textit{Mirroring Motion} feature, explained in Section~\ref{sec:tpo:mirror}, is utilized: the operator move his right arm to command two specular virtual forces on both robot end-effectors.

It can be observed that the TelePhysicalOperation interface results to not match postures between the robot and human operator. This is an expected  behavior that is intrinsic to the employed \enquote{Marionette} based control. 
Indeed, this approach controls the robot motions as guided by the virtual forces generated by the operator's arms, rather than mimicking the posture of the operator.

\subsection{Teaching End-effector Poses}\label{sec:tpo:expteach}
\begin{figure}[H]
	\centering
	\includegraphics[width=.49\linewidth]{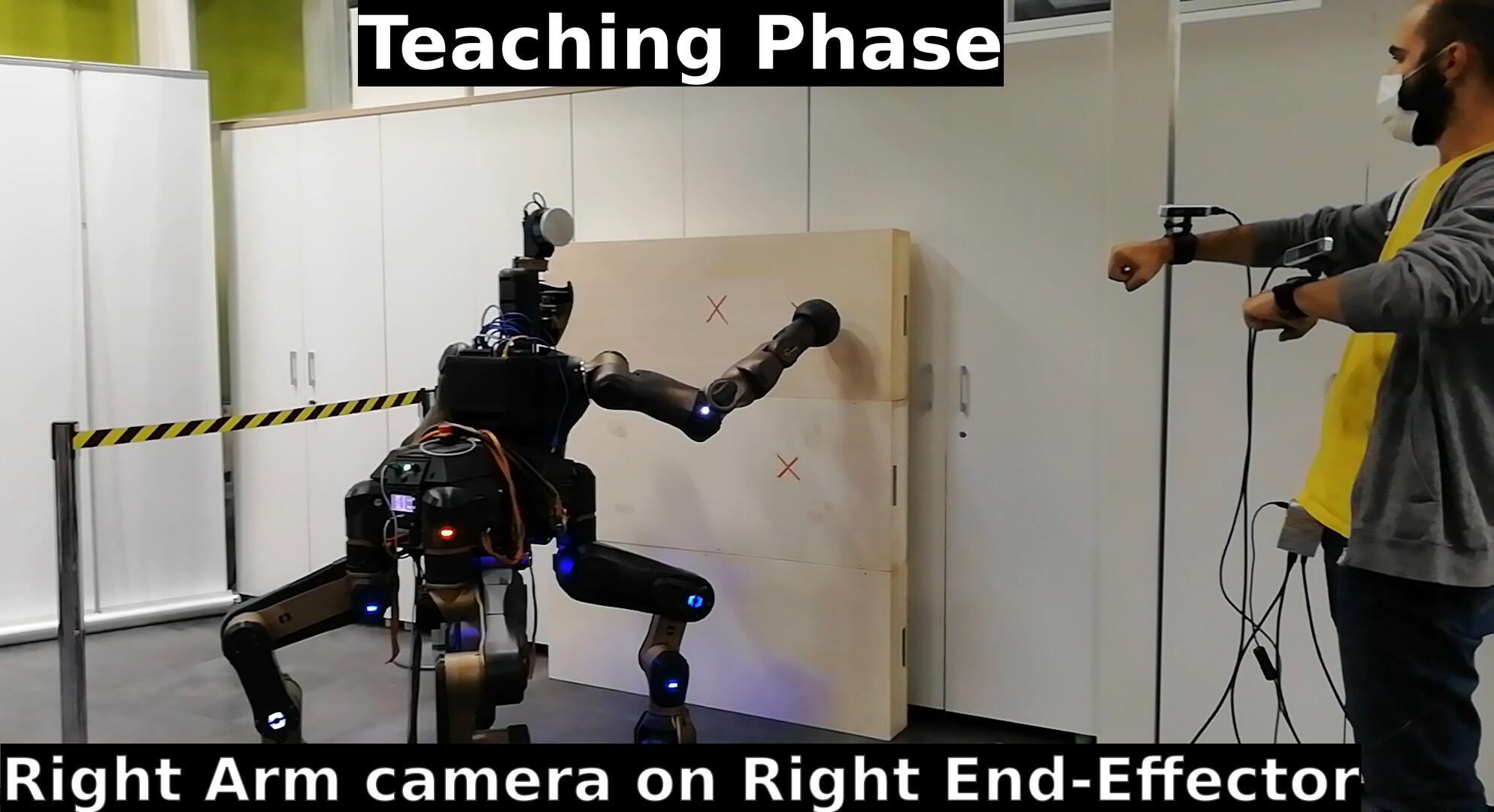}
	\includegraphics[width=.49\linewidth,trim={6cm 0 0 0}, clip]{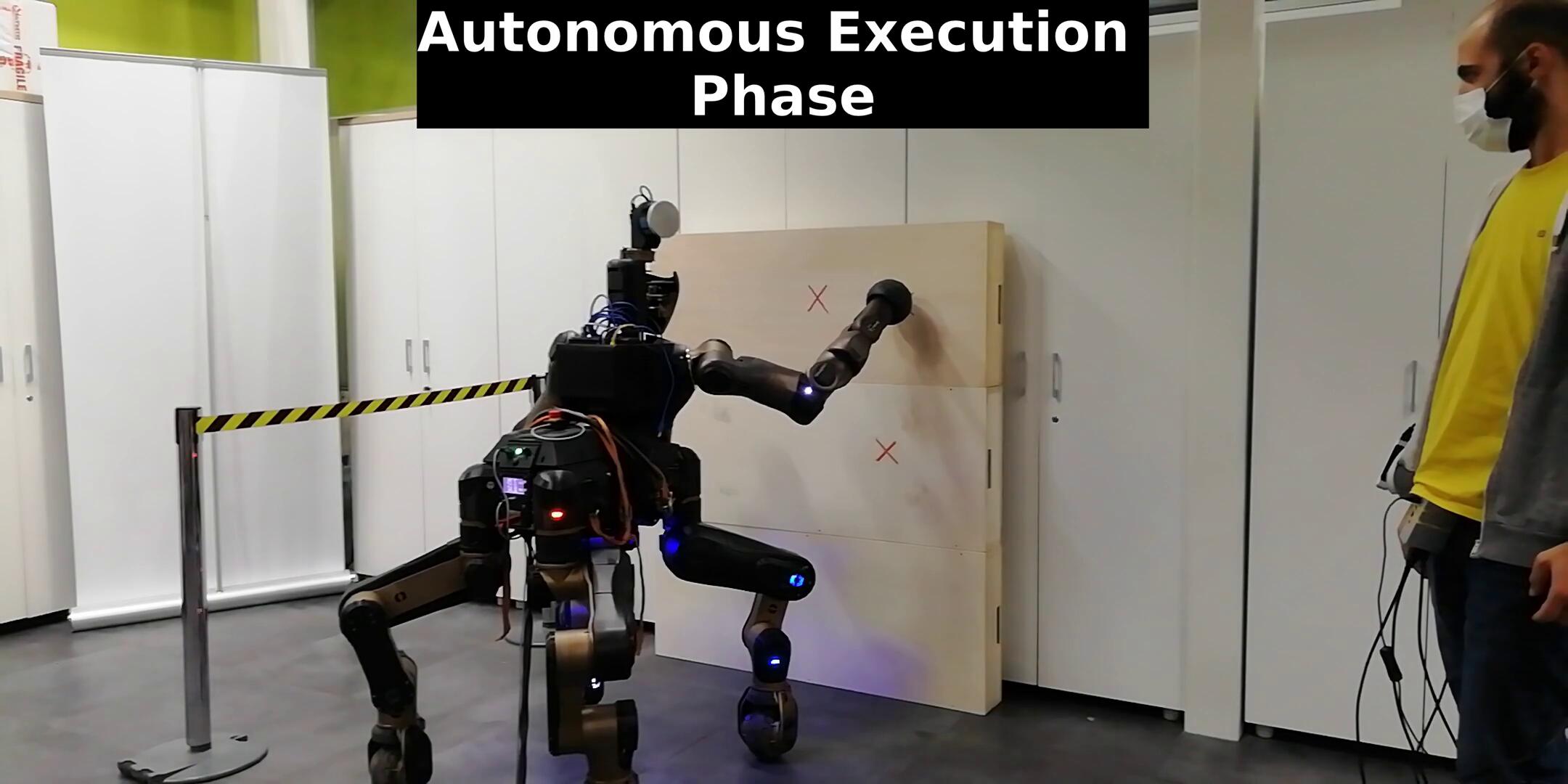}
	\caption[TelePhysicalOperation: teaching end-effector poses experiment]{The TelePhysicalOperation interface is exploited for a remote collaboration. In the first image, the operator remotely guides the robot specific locations. In the second image, the robot executes autonomously a trajectory which follows the recorded locations.}
	\label{fig:TPOExpTeaching}
\end{figure}

In this experiment a remote collaboration between the operator and the robot is performed, as shown in \figurename{}~\ref{fig:TPOExpTeaching}. In this demonstration the operator uses the TelePhysicalOperation interface to move the remote robot's end-effector to a number of workspace locations, identified by red crosses. Once the robot has reached these locations, the Cartesian pose of the end-effector are recorded. 
In a second phase, the robot is commanded such that it performs a trajectory which follows the previously recorded positions. 

This demonstration in principle resembles the teaching phase of a human-robot collaboration but without the need to physically interact with the robot body to guide its end-effector. 
Indeed, the operator, using the proposed interface, can perform the teaching/demonstration phase from remote similarly as when the robot is guided through direct physical interaction.

\subsection{Single Arm Button Reaching}\label{sec:tpo:btn}
\begin{figure}[H]
	\centering
	\includegraphics[width=.49\linewidth,keepaspectratio]{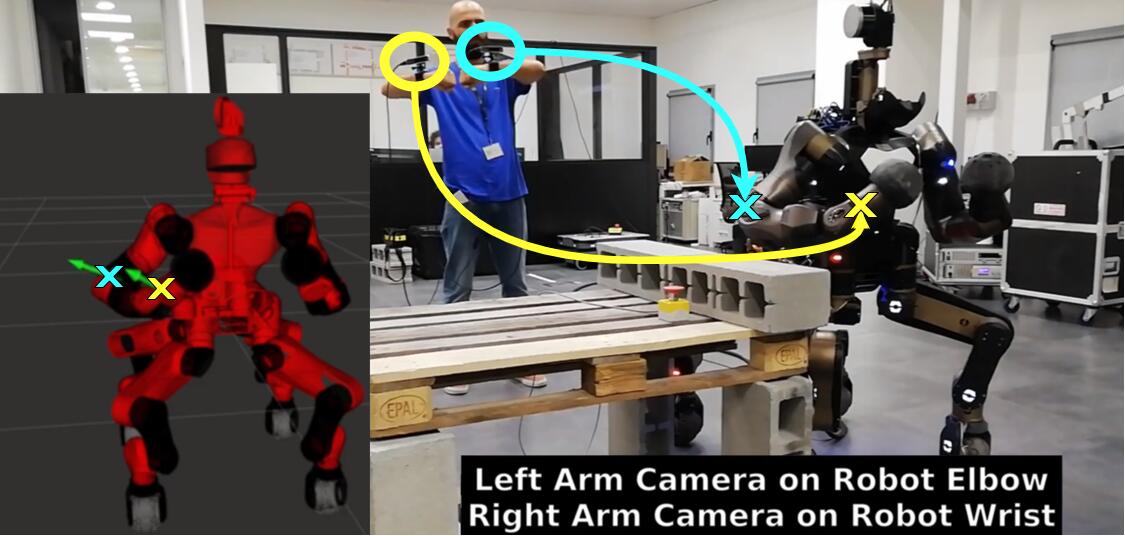}
	\includegraphics[width=.49\linewidth,keepaspectratio]{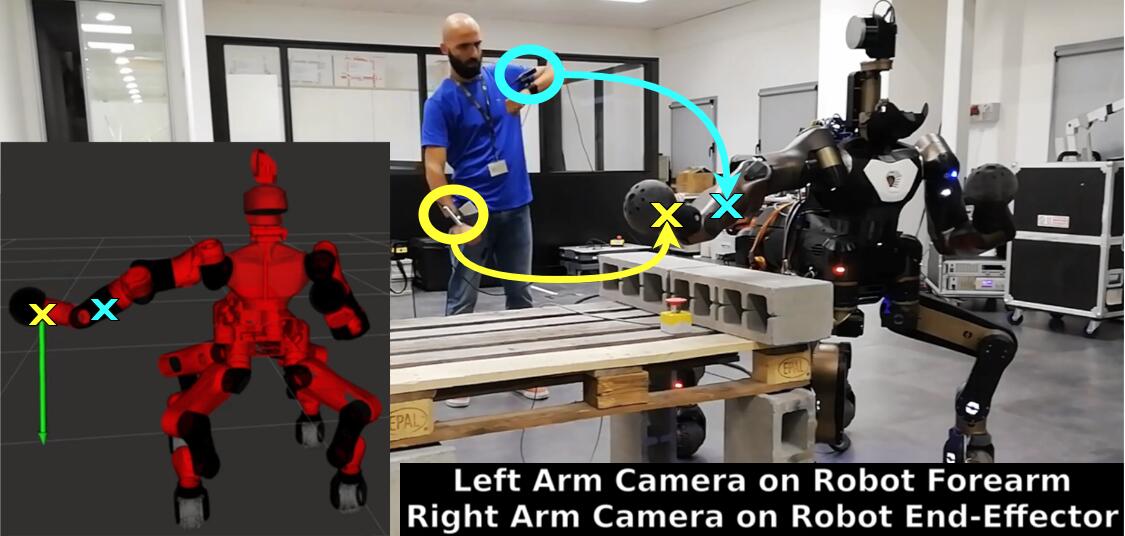}\\
	\vspace{5px}
	\includegraphics[width=.49\linewidth,keepaspectratio]{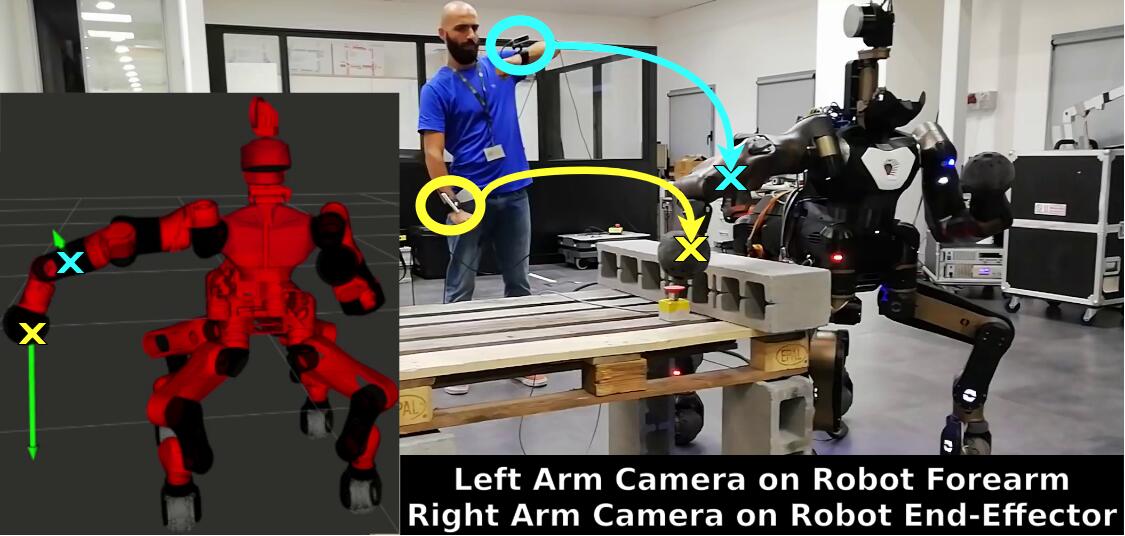}
	\includegraphics[width=.49\linewidth,keepaspectratio]{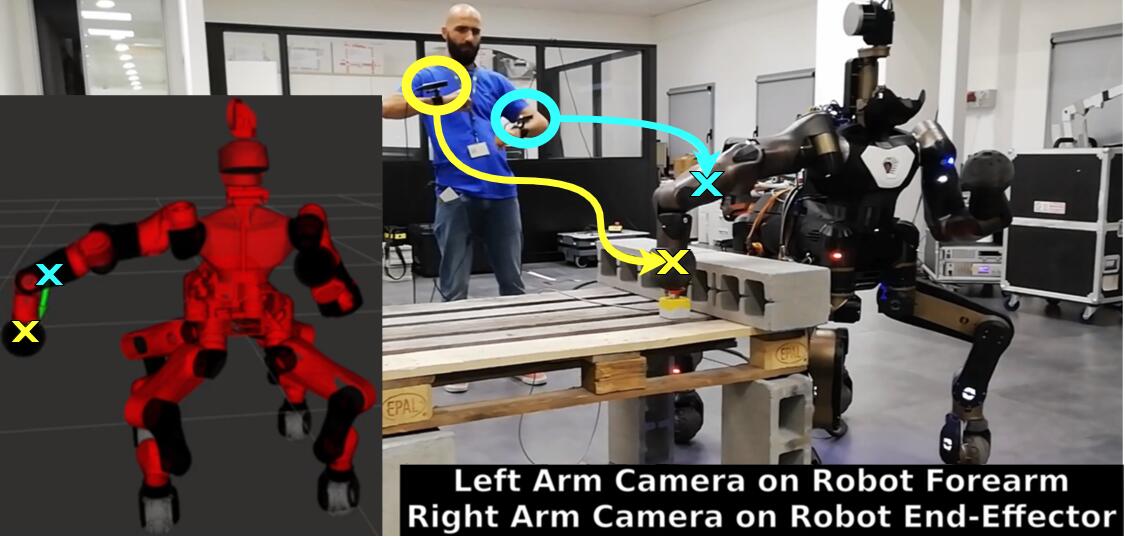}
	\caption[TelePhysicalOperation: button reaching experiment]{In the button experiment, the goal is to teleoperate the robot right arm toward a button to be pressed, while avoiding some obstacles (the bricks). With the TelePhysicalOperation interface, the operator shapes the robot arm by the application of two virtual forces in different links of the arm, to avoid the obstacle and reach the goal.}    
	\label{fig:TPOExpButton}
\end{figure}

In this experiment, shown in \figurename{}~\ref{fig:TPOExpButton}, the operator must command the robot's right arm to reach a button to be pressed while avoiding the brick obstacles.
This demonstration shows how, with the TelePhysicalOperation interface, is possible to apply different virtual forces on the same robot's arm in order to reach the desired location while adjusting the arm shape to avoid the obstacles.

In this case, the \textit{Blocking Link} feature, explained in Section~\ref{sec:tpo:blocklink}, is employed to exploit particular joints to avoid the obstacles.
In the top sequences of \figurename{}~\ref{fig:TPOExpButton}, the operator applies virtual forces to activate the shoulder and elbow joints, permitting to go over the obstacle. In the bottom sequences, the operator bends the robot's wrist to reach the button from above.

\begin{figure}[H]
	\centering
	\includegraphics[width=\linewidth]{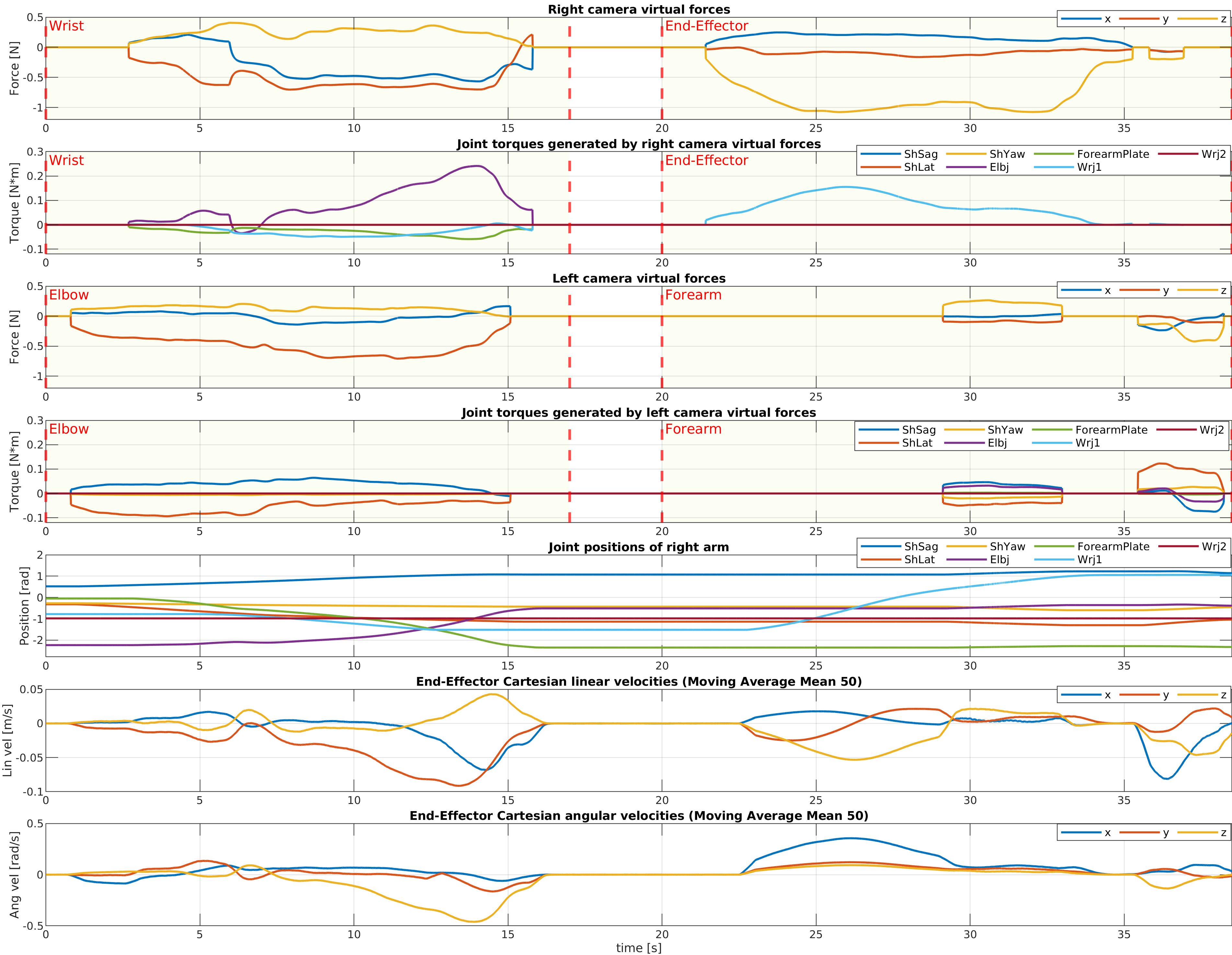}
	\caption[TelePhysicalOperation: button reaching experiment plots]{Plots for the button experiment. For each virtual force $\boldsymbol{f}_{\mathit{cp}}$ generated by the right and left user's arm (first and third plots, \eqref{eq:tpo:f_cp}), joint torques $\boldsymbol{\tau}_{\mathit{cp}}$ are computed (second and fourth plots, \eqref{eq:tpo:taocp}). The control points where the virtual forces are applied are indicated by the areas delimited by the vertical red dashed lines (i.e.\ \textit{Wrist}, \textit{Elbow}, \textit{End-Effector}, and \textit{Forearm}).
		In the fifth plot, the joint positions of the robot's arm are displayed.
		In the two plots at the bottom, the end-effector linear and angular Cartesian velocities (filtered to improve the visualization) are shown.
	}
	\label{fig:TPOExpButtonPlot}
\end{figure}

The plots in \figurename{}~\ref{fig:TPOExpButtonPlot} show the relevant data gathered during this experiment. 
In the first and third plots, the right and left camera inputs are depicted, respectively. These are the virtual forces $\boldsymbol{f}_{\mathit{cp}}$ of \eqref{eq:tpo:f_cp}, computed by filtering (with $s(\cdot)$) and scaling (with $k_{\mathit{cam}}$) the camera positions $\boldsymbol{r}$.
For each camera input, the second and fourth plots show the joint torques $\boldsymbol{\tau}$ generated with \eqref{eq:tpo:taocp} and modified according to the \textit{Blocking Link} feature. It can be seen that this feature nullifies the contribution of the right user's arm to the torque command of some joints in certain intervals. 
This happens because the other virtual force (generated from the left user's arm) is applied in an ancestor link of the arm kinematic chain. 
As visible in the second plot, this permits to the user to generate torques with his right arm only for the \textit{Elbj}, \textit{ForearmPlate}, \textit{Wrj1}, and \textit{Wrj2} joints in the time interval [$t=0s$, $t=16s$], and only for the \textit{Wrj1} and \textit{Wrj2} joints in the time interval beginning at $t=20s$ and finishing at the end of the experiment. It is also visible that, in the fourth plot, the user's left arm does not generate motions for the mentioned joints. This is intrinsic to the use of virtual forces: as explained in Section~\ref{sec:tpo:pos}, joints that follow the control point in the kinematic chain are not influenced since the relative columns of the Jacobian matrix are filled with zeros.

The last three plots show the state of the robot's right arm during the experiment: joint position, linear Cartesian velocity of the end-effector, and angular Cartesian velocity of the end-effector.
Please note that the linear and angular Cartesian velocities (in the two bottom plots) have been computed from the derivation of the sensed robot joint positions, hence they have been post-processed with a moving average filter to improve the visualization.

\subsection{Locomotion and Pick \& Place Box}\label{sec:tpo:loco}
\begin{figure}[H]
	\centering
	\includegraphics[width=1\linewidth, trim={7cm 0 0cm 0cm},clip]{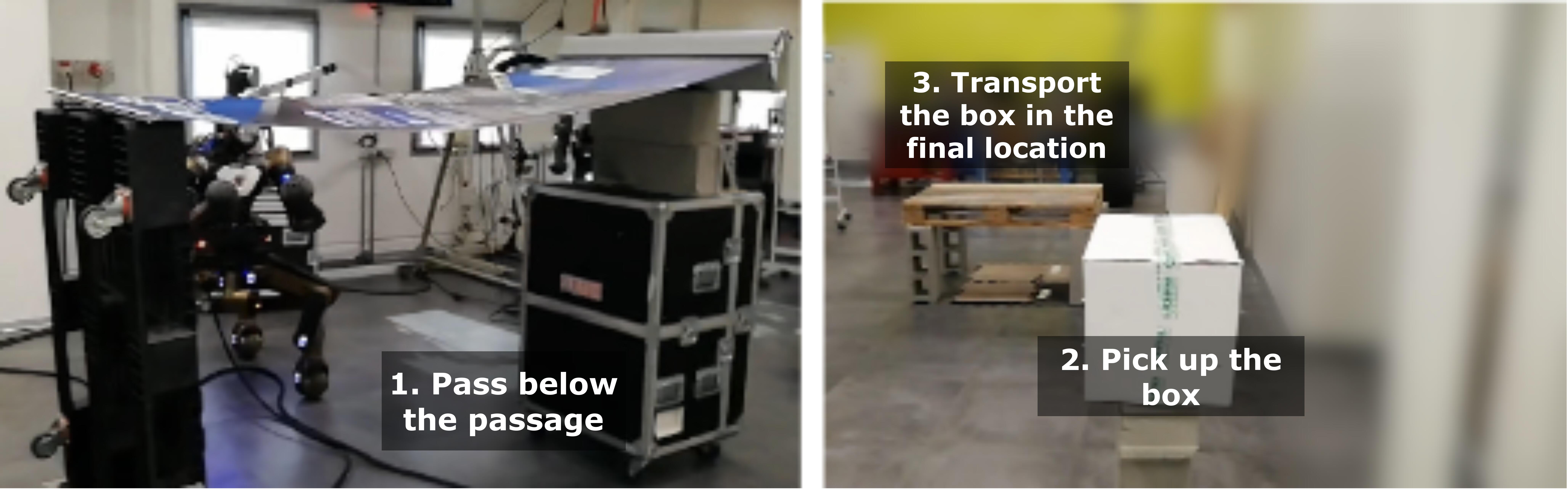}
	\caption[TelePhysicalOperation: locomotion and pick \& place experimental setup]{The setup of the locomotion and pick \& place box experiment. The operator must teleoperate the robot below a passage, commanding the body to squat down. Then, a box must be picked up, transported, and placed onto the final location.}
	\label{fig:TPOExpBoxInit}   
\end{figure}

In this experiment, a complex environment is set up as shown in \figurename{}~\ref{fig:TPOExpBoxInit}, where the locomotion and manipulation abilities of the CENTAURO robot are required to accomplish the task. 
Firstly, the CENTAURO must be teleoperated through a low passage, which imposes the necessity to command the robot to perform a \enquote{squatting} motion.
Then, a large object requiring both robot's arm to be manipulated must be bimanually picked up from one location, and placed in another site. 

\begin{figure}[H]
	\centering
	\includegraphics[height=0.24\linewidth,trim={5cm 0 0 0cm},clip]{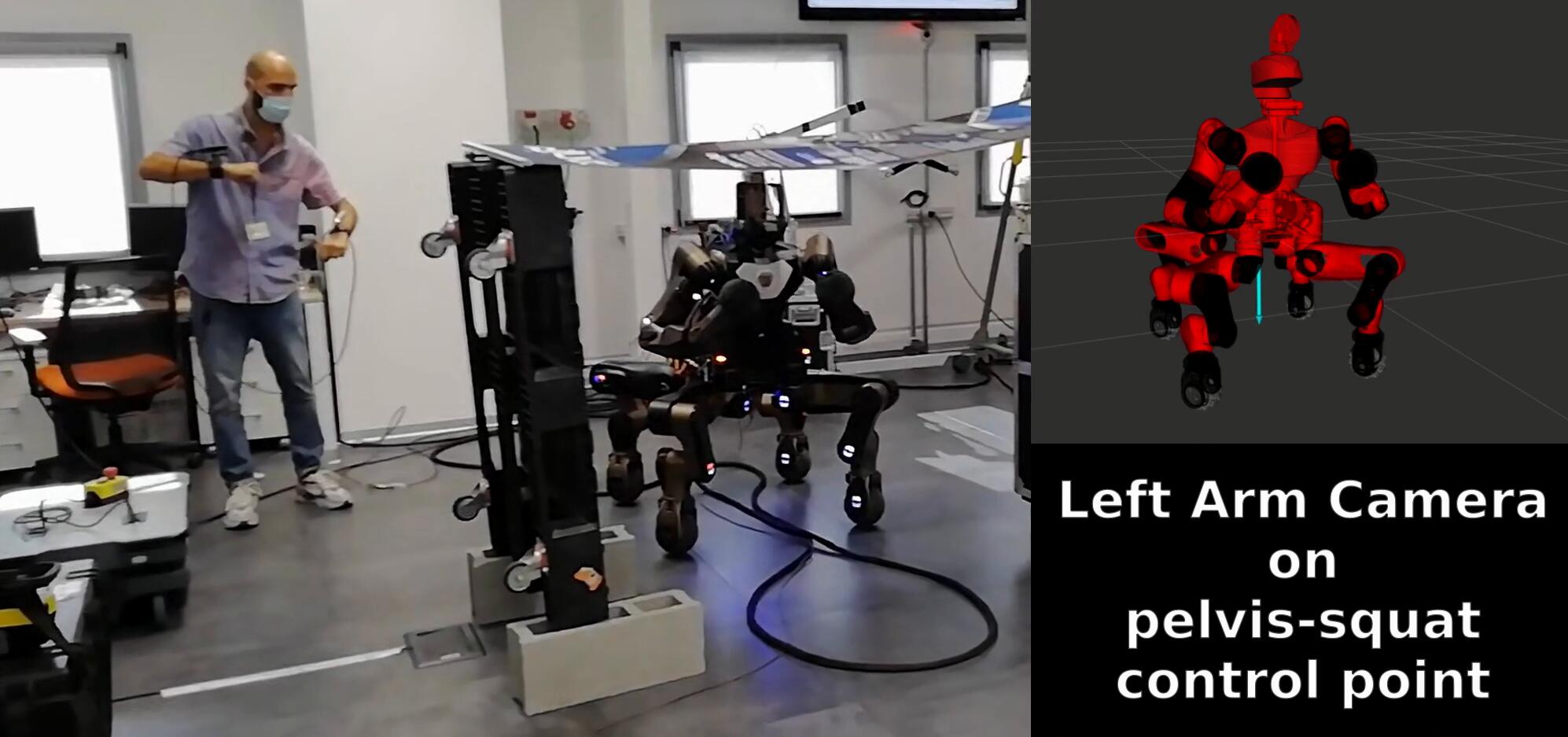}
	\includegraphics[height=0.24\linewidth,trim={2cm 0 0 0cm},clip]{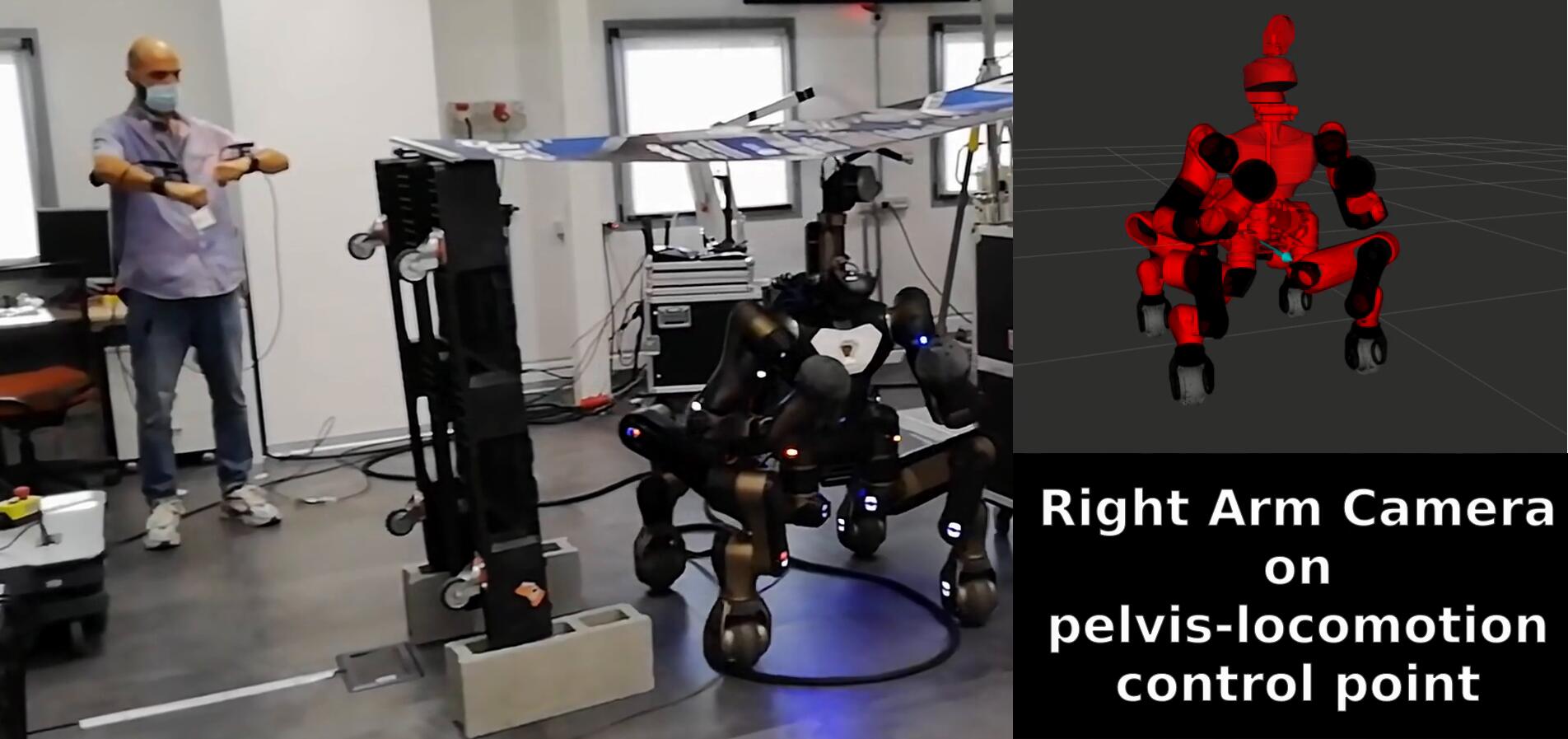}\\
	\vspace{5px}
	\includegraphics[height=0.24\linewidth,trim={7cm 0 0 0cm},clip]{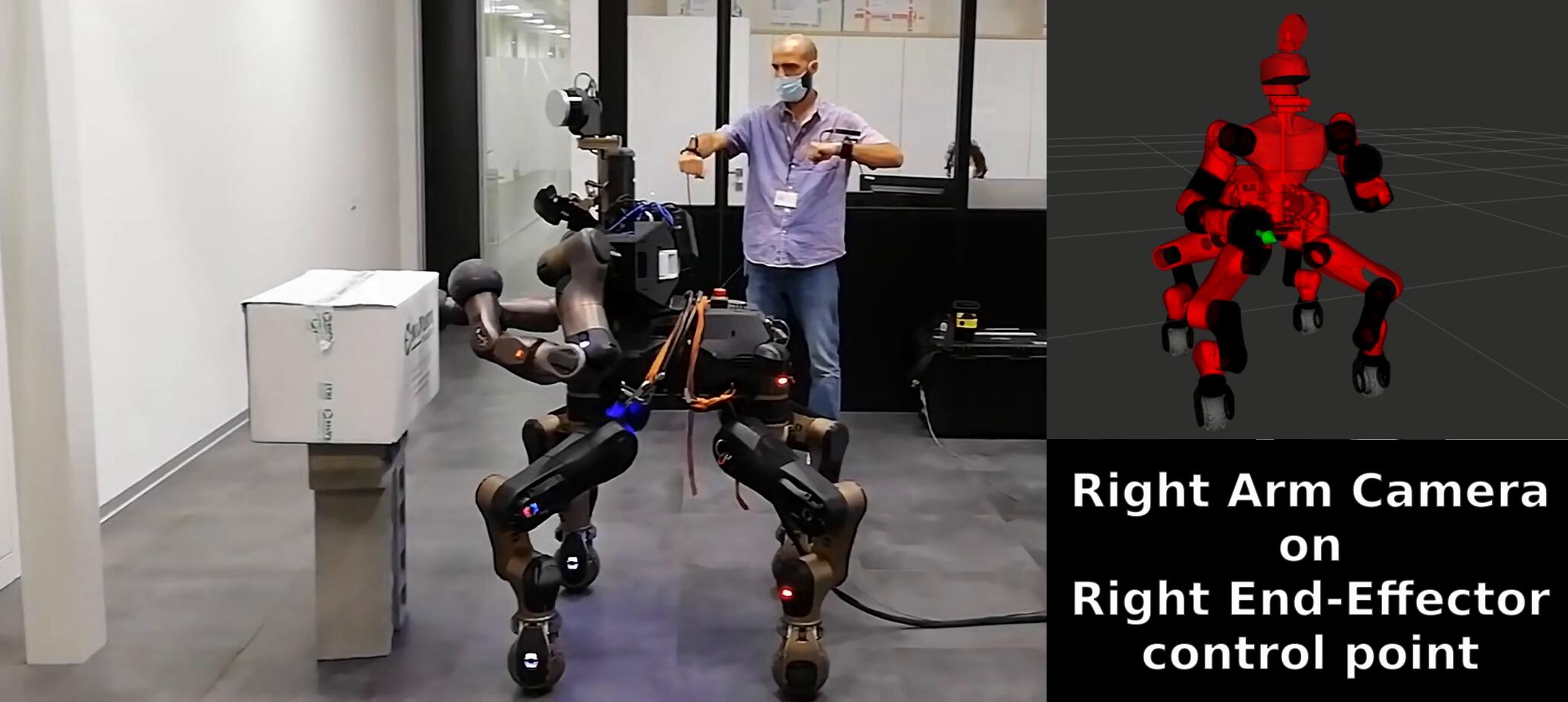}
	\includegraphics[height=0.24\linewidth,trim={6cm 0 0 0cm},clip]{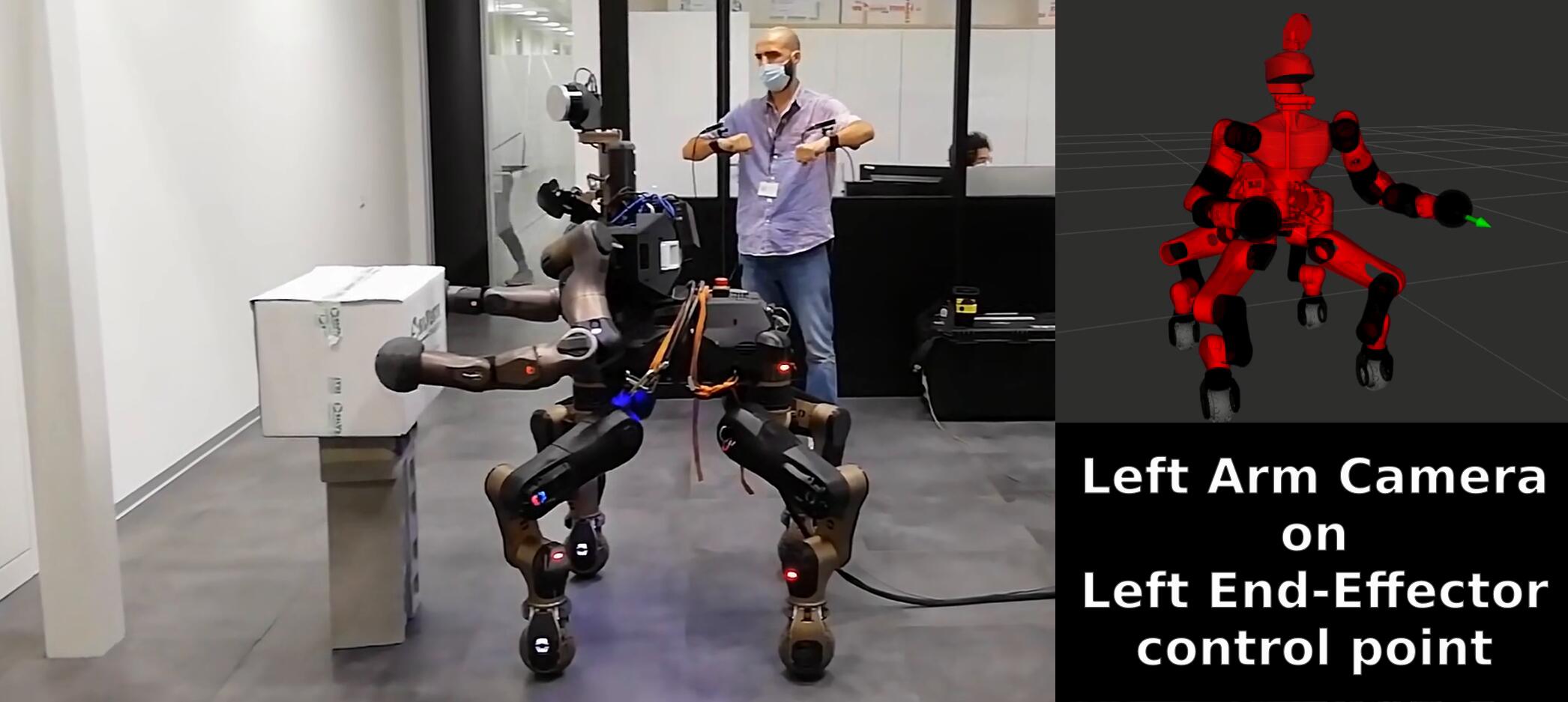}\\
	\vspace{5px}
	\includegraphics[height=0.20\linewidth,trim={12cm 0 0 0cm},clip]{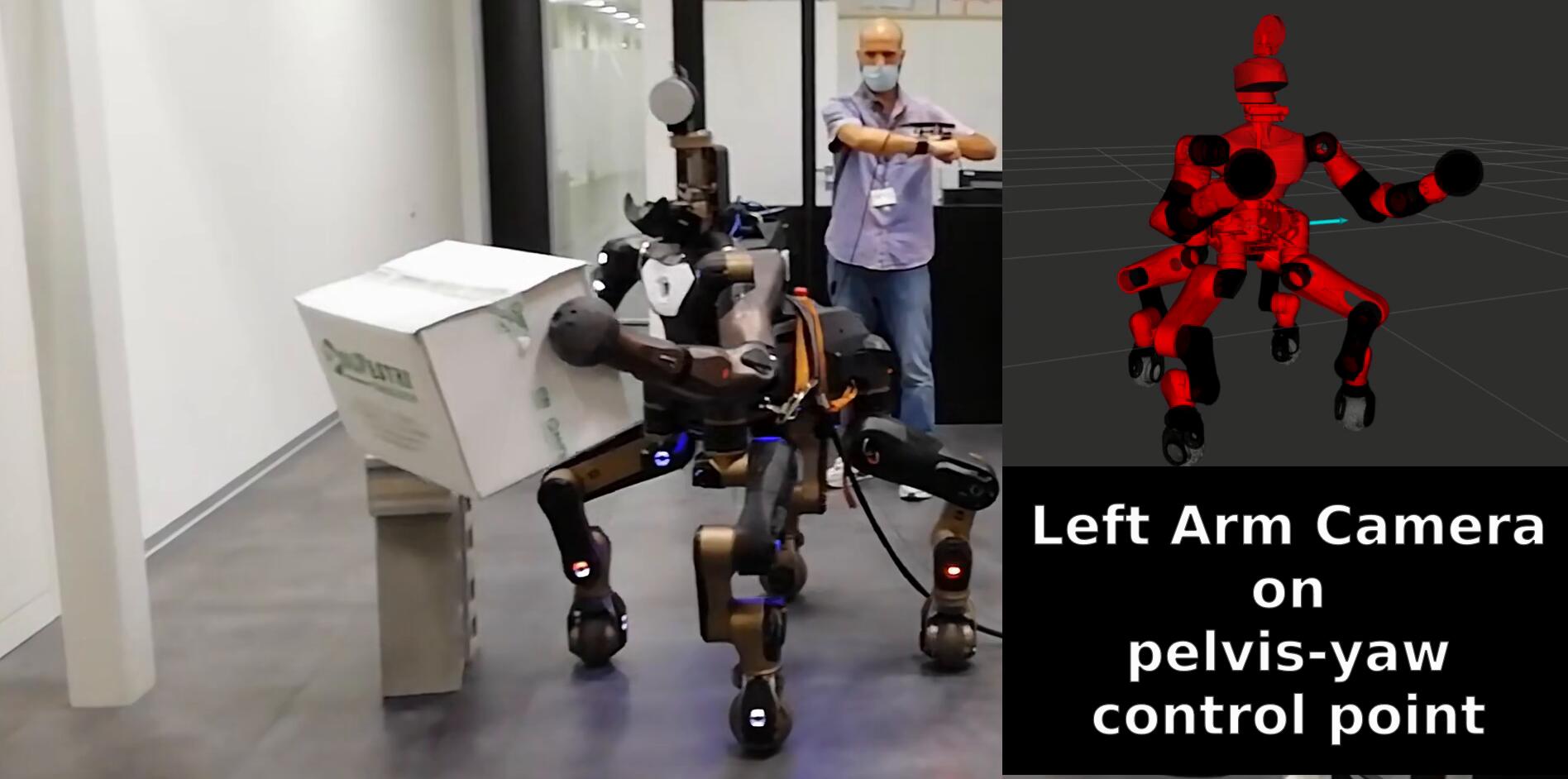}
	\includegraphics[height=0.20\linewidth,trim={20cm 0 0 0cm},clip]{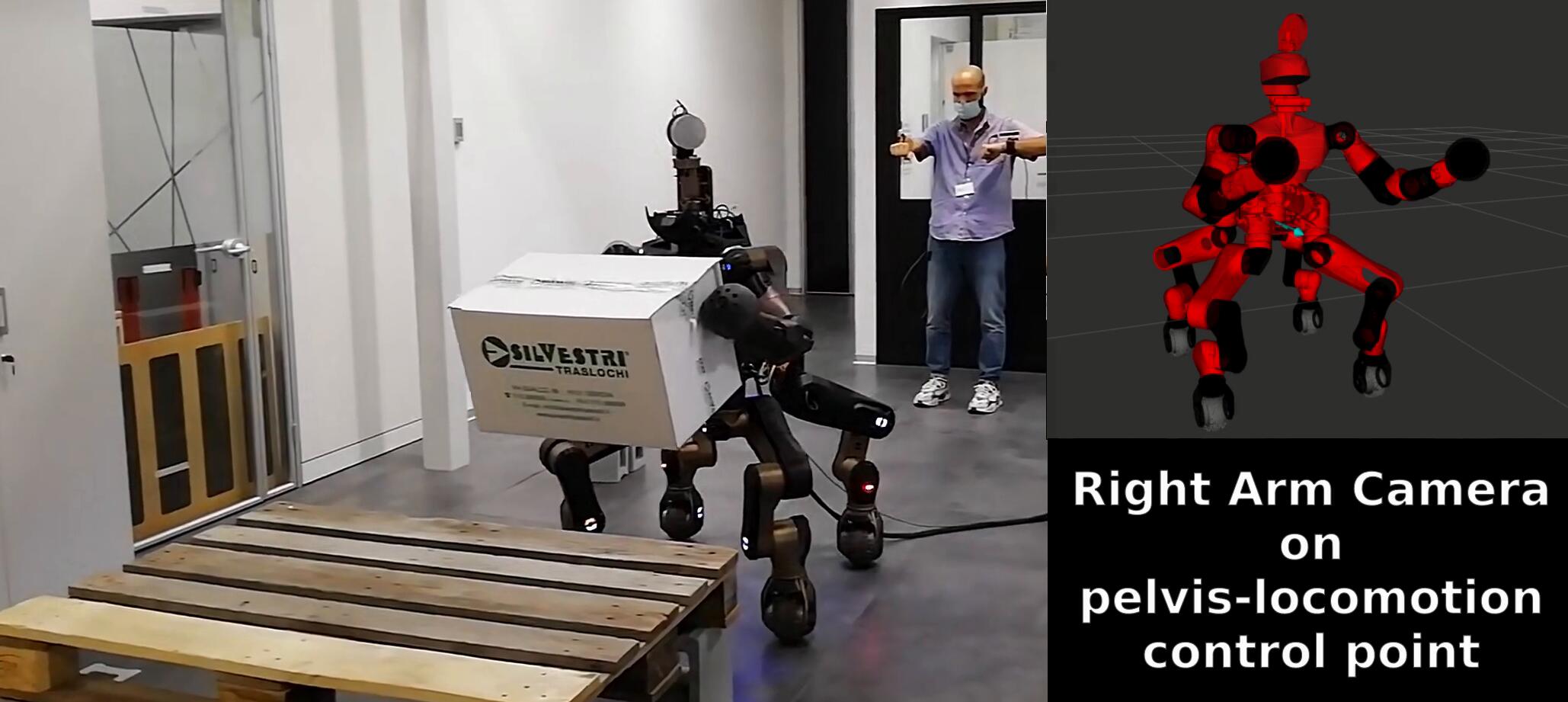}
	\includegraphics[height=0.20\linewidth,trim={20cm 0 0 0cm},clip]{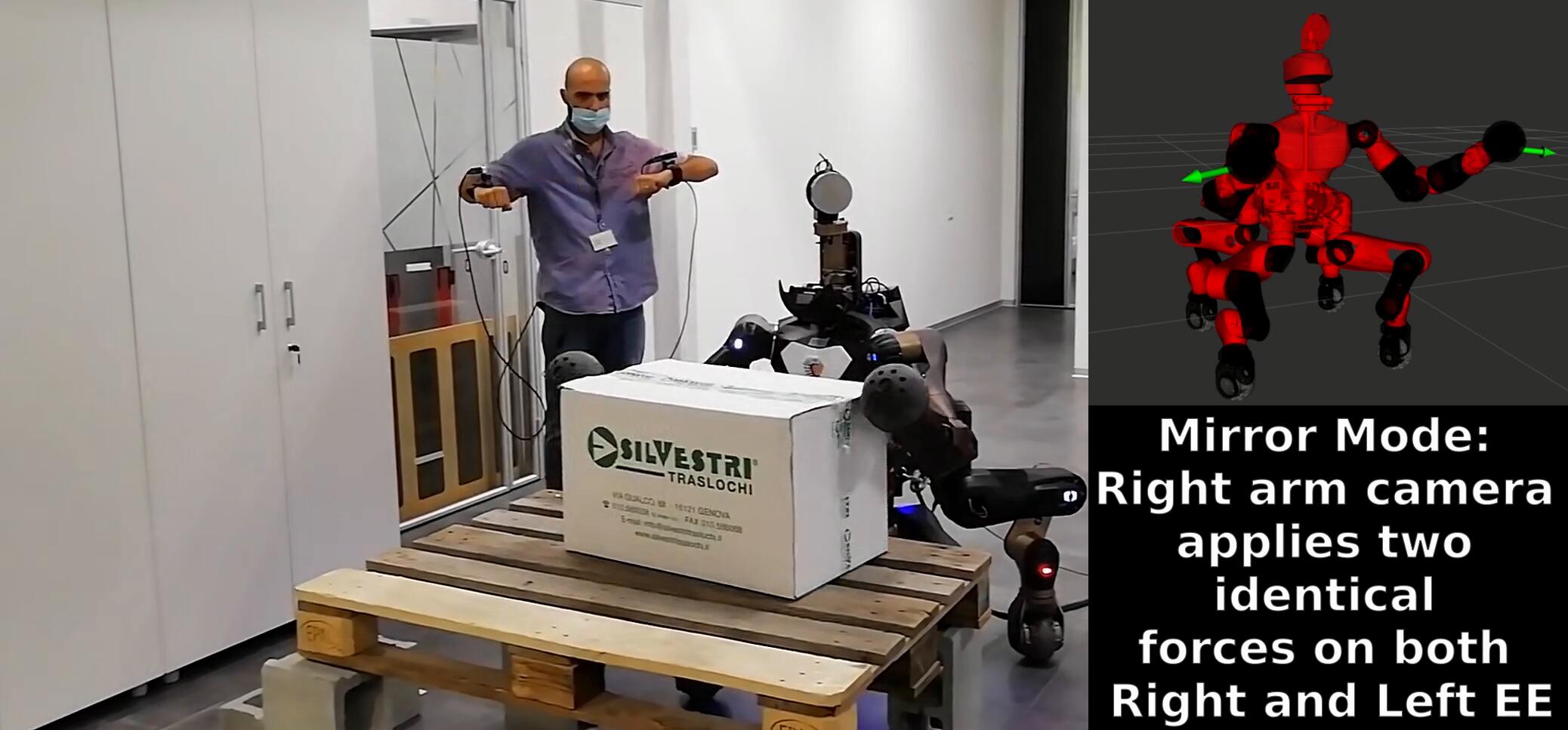}
	\caption[TelePhysicalOperation: locomotion and pick \& place experiment sequences]{The three phases of the locomotion and pick \& place box experiment: in the top row, CENTAURO robot is teleoperated to pass below the low passage to reach the box; in the second row, the two arms of the robot are positioned on the box sides; in the third row, the box is transported and placed in the final location.}    
	\label{fig:TPOExpBox}
\end{figure}

Some sequences of the experiment are shown in \figurename{}~\ref{fig:TPOExpBox}, each row representing different phases. In the first image, the user applies a virtual force directed to the ground on the robot body, to virtually push down the robot, generating a squatting motion. Then, in the second image, the robot passes below the narrow passage, according to another virtual force from the operator which pushes the robot forward. 
In the second row, the user controls the two robot arms to place the two end-effectors in a suitable grasping position for the object. In the third row, other robot body motions are commanded to transport the box toward the final location.
In the last image, the final placement of the box is facilitated by the \textit{Mirroring Motion} feature, with which the operator applies two symmetrical virtual forces on the two robot's end-effectors by employing only his right arm. This feature aids in coordinating the arms motions for the bimanual placement. 

\begin{figure}[H]
	\centering
	\includegraphics[width=\linewidth]{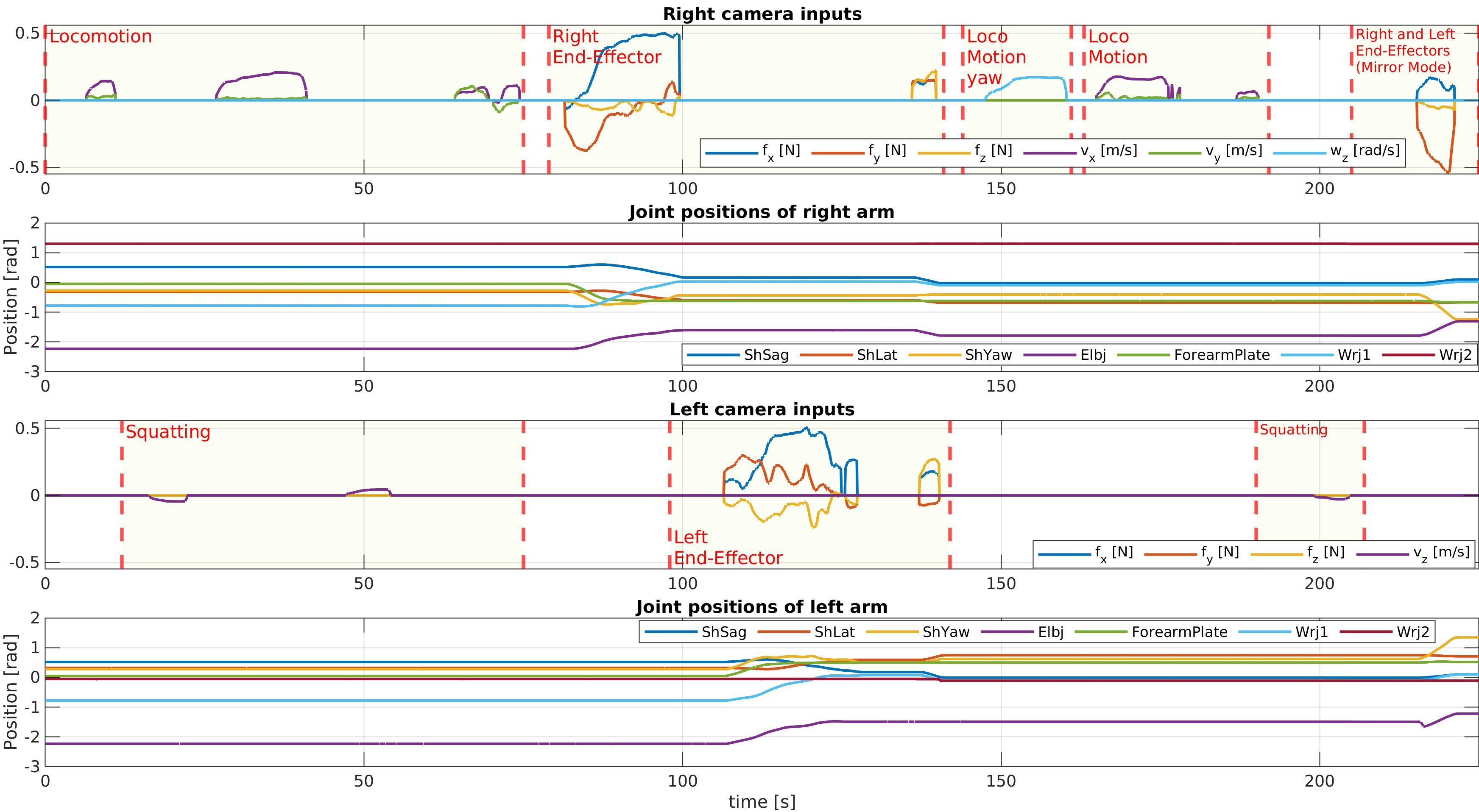}
	\caption[TelePhysicalOperation: locomotion and pick \& place experiment plots]{Plots for the locomotion and pick \& place box experiment. In the first and third plots, the camera inputs are shown. The different phases in which different control points are chosen are highlighted by the areas delimited by the vertical dashed red lines. For \textit{Left} and \textit{Right End-Effector} control points, the inputs generate postural motions \eqref{eq:tpo:taocp}. For \textit{Locomotion}, \textit{Locomotion yaw}, and \textit{Squatting} control points, the inputs generate Cartesian motions \eqref{eq:x_cp}. The joint positions of the right and left arm varying during the experiment are visible in the second and fourth plots, respectively.}
	\label{fig:TPOExpBoxPlot}
\end{figure}

Plots relative to the experiment are shown in \figurename{}~\ref{fig:TPOExpBoxPlot}. The first and third plots show the input from the user's arms tracked by the cameras. Differently from the previous experiments, in this case the virtual forces are used to generate not only postural motions by selecting control points on the robot arms (Section~\ref{sec:tpo:pos}), but also Cartesian motions by selecting control points on the robot body (Section~\ref{sec:tpo:vel}). The latter are employed to command the robot to move in the room (\textit{Locomotion} control point), squat up and down (\textit{Squatting control point}), and rotate on itself (\textit{Locomotion yaw} control point).
The second and fourth plots show the joint position of the two arms. At the very end of these plots, it is possible to see that the joint positions are following symmetrical directions, because of the activation of the \textit{Mirroring Motion} feature.

\section{Conclusions}\label{sec:tpo:conclusions}

In this chapter, the TelePhysicalOperation interface has been presented as a novel approach to improve the classical robot teleoperation by exploring virtual human-robot interactions. The intuitiveness of the interface relies on an approach that resembles controlling the remote robot in a \enquote{Marionette} based manner. By applying virtual forces on the robot body, the operator can effectively teleoperate the remote robot as in a physical human-robot interaction interface, exploiting both the robot manipulation and locomotion capabilities. 
Different number of virtual forces can be applied on different robot body parts, hence providing more possibilities than a classical Cartesian teleoperation interface that permits to command only end-effector directions (Section~\ref{sec:tpo:concept}).

To track the user's arm motion, the \acrshort{tpo} Suit, a light weight and comfortable wearable solution composed by tracking cameras, has been employed (Section~\ref{sec:tpo:tposuit}). 
A control architecture has been developed to realize the proposed concept, and to handle the virtual forces according to the control point chosen (Section~\ref{sec:tpo:control}).

Simple visualization tools (Section~\ref{sec:tpo:visual}) and particular autonomy features (Section~\ref{sec:tpo:auto}) have been integrated to aid the user in operating the robot with the TelePhysicalOperation interface.

Validations with the CENTAURO robot have been presented, demonstrating good efficacy and ability of the operator to command the execution of a number of tasks. In particular, it is shown how both the manipulation and locomotion ability of the robot can be controlled with ease, by the application of multiple virtual forces on different robot body parts, employing both postural and Cartesian motion generation. Furthermore, the validations revealed the utility of the integrated autonomy features that helped the operator in particular tasks (Section~\ref{sec:tpo:exp}).

In the following Chapter~\ref{chap:TPOH}, the TelePhysicalOperation architecture will be enhanced focusing on providing more situational awareness to the operator, by the development of wearable haptic devices. These devices also expand the input possibilities for the operator with additional input buttons. This addresses the limitation consisting in the necessity of a second operator to execute some instructions according to the main operator, like pausing the robot control or changing the control point.

While this chapter already presented some autonomy features, other more interesting autonomy functionalities will be explored in Chapter~\ref{chap:tpoAuto}, to help the operator with the robot mobility, the end-effector dexterity, and the bimanual transportation of a load.

\chapter{Wearable Haptic-enabled TelePhysicalOperation}\label{chap:TPOH}

\begin{figure}[H]
	\centering
	\includegraphics[width=0.9\linewidth]{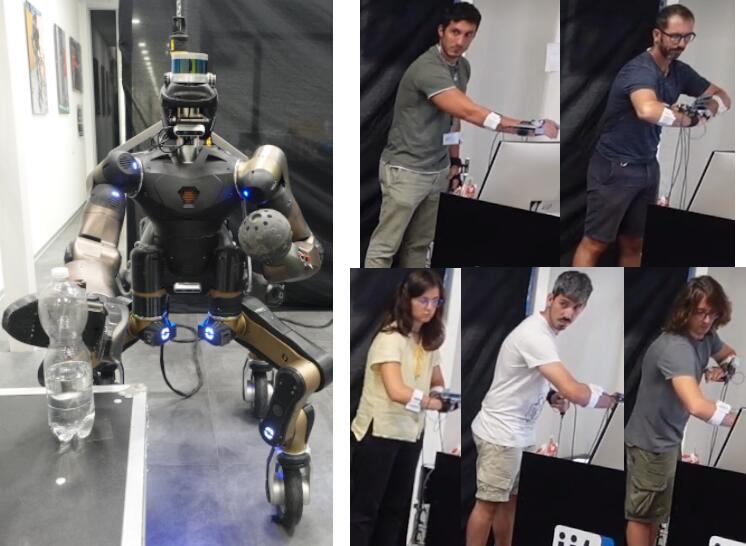}
	\caption[Users Controlling the CENTAURO with the haptic-enabled \acrshort{tpo}]{Some of the users involved in the experimental validations are teleoperating the CENTAURO robot with the haptic-enabled TelePhysicalOperation interface. The images are extracted from the first task of the mission presented in Section~\ref{sec:tph:exp}.}
	\label{fig:tph:firstphoto}
\end{figure}

\lettrine{I}{n} human-robot interfaces, as highlighted in the introductory chapters of this thesis, a key theme revolves around providing situation awareness to enable the operator to better comprehend the ongoing events. 
This chapter revolves around this matter, by presenting an enhancement of the \acrlong{tpo} interface, which involves the integration of a haptic feedback channel through the development of a wearable vibrotactile device.

New validation experiments are presented, employing the TelePhysicalOperation interface with and without the haptic enhancement (\figurename{}~\ref{fig:tph:firstphoto}). 

The development of the haptic interface and the integration of the vibrotactile devices in the TelePhysicalOperation architecture are the result of a collaborative effort between the lab where this thesis has been conducted, \acrshort{hhcm} at \acrlong{iit}, and another lab, the \acrshort{sirslab} at the University of Siena.\\  

\noindent This chapter is based on the following article:\\
\fullcite{TPO4}~\cite{TPO4}
\section{Wearable Haptics for the TelePhysicalOperation Interface}\label{sec:tpoh:intro}

\begin{figure}[H]
	\centering
	\includegraphics[width=1\linewidth]{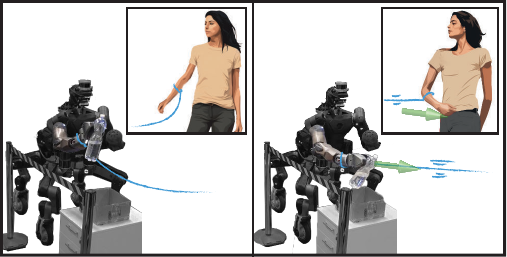}
	\caption[Haptics-enabled \acrlong{tpo} concept]{The concept of the haptics-enabled \acrlong{tpo}. In the left time-frame, the blue rope, representing the \acrlong{tpo} connection, is not under tension: the robot stands still. In the right time-frame, the user is drawing her arm closer to her body increasing the tension of the blue rope. Hence, the robot right arm, connected to the user's right arm, is pulled, following the user's movement. At the same time the user can perceive the force feedback of pulling through the wearable sensorimotor interface.}
	\label{fig:tph:intro}
\end{figure}

In general, while teleoperating, haptic feedback is very useful and sometimes fundamental to understand the task's and robot's status, since it is not always possible to understand what it is happening at the robot's side through visual feedback alone, whether direct or via cameras. 
Hence, as confirmed in previous works (Section~\ref{sec:soa:haptic}), the importance of delivering information through haptic feedback has been acknowledged in this contribution as well.

In Chapter~\ref{chap:TPO}, the \acrfull{tpo} interface has been presented as a way of teleoperating robots resembling a physical human-robot interaction through the application of virtual forces, as in a \enquote{Marionette} interaction interface.
In a physical human-robot interaction (Section~\ref{sec:soa:phri}) and in a real \enquote{Marionette} interface, the tactile sensation of counteracting forces generated by applying force on the robot's body serves as feedback for the user. This helps in comprehending what is happening and planning the subsequent commands accordingly. In contrast, with the TelePhysicalOperation interface, there was no such sensation since the absence of any contact or physical mean between the user and the robot.

This chapter presents how this limitation has been addressed, by the development and integration of a wearable haptic interface. 
This interface augments the \acrshort{tpo} interface with a haptic feedback channel and new control inputs, improving the user's experience.
While in the original implementation of the \acrshort{tpo} interface the user had only limited visual feedback of the robot and its workspace, with this integration cutaneous haptic cues are exploited to provide him/her with additional information. In addition, push buttons are integrated into the designed devices to allow the user to send additional control inputs through the \acrshort{tpo} framework, eliminating the necessity of the second operator for example to change the control point $\mathit{cp}$ where the virtual force is applied.
The resulting novel haptic-enabled \acrlong{tpo} interface allows the full sensorimotor interaction~\cite{prattichizzo2021human} between the human and the robot, providing flexibility to be adapted to different teleoperation scenarios.

Considering the trade-off between interface complexity and usability, the simplicity of the  original \acrshort{tpo} Suit is maintained, adopting a technology which is lightweight and unobtrusive.
The wearable haptic vibrotactile interface consists in a ring and an armband managed by a control board and powered by a battery worn on the hand. Both the ring and the armband  can deliver vibration and skin indentation stimuli. The ring is are also endowed with two small push buttons. The user wears two of such interfaces: one for each upper limb.
In the proposed implementation, skin indentation stimuli are sent to the user to increase his/her awareness of (1) the control command delivered to the robot and (2) the robot-environment interaction. Vibrations stimuli are used as acknowledgments about the execution of the command associated with a button. While the benefits of vibratory acknowledgments and force feedback in teleoperation have been shown in previous works~\cite{pacchierotti2015cutaneous, Franco2019}, here the haptic cues are specifically tailored to the \acrshort{tpo} system. In particular, it has been designed a new skin indentation feedback strategy that conveys the sensation of pulling the virtual ropes. Whenever a robot's part is being controlled by the operator and is actually moving, the corresponding operator's forearm is squeezed proportionally to the entity of the control action (\figurename{}~\ref{fig:tph:intro}).

\noindent In summary, the main contributions of the work presented in this chapter are:
\begin{itemize}

	\item The development of a haptic feedback strategy designed for the \enquote{Marionette}-inspired \acrfull{tpo} framework.
	
	\item The hardware development of a sensorimotor interface combining input (buttons) and feedback (haptic rings and armbands) devices and its integration in the \acrfull{tpo} system.
	
	\item The evaluation of the overall interface, with and without haptics, involving naive operators who teleoperated the CENTAURO robot, to accomplish a locomanipulation mission. Subjective participants' opinions are collected to evaluate the effectiveness of the haptic-enabled interface with respect to the original interface.
\end{itemize}

\section{Wearable Haptic Interface}\label{sec:tpoh:detail}
The developed wearable haptic interface is based on the \acrlong{tpo} paradigm, and it has been incorporated in the \acrshort{tpo} Suit, presented in Section~\ref{sec:tpo:tposuit}. In what follows, details about the hardware design of the devices (Section~\ref{sec:tpoh:hardware}), and their integration in the \acrshort{tpo} architecture (Section~\ref{sec:tpoh:integration}) are given.

\subsection{Haptic Interface Hardware Design}\label{sec:tpoh:hardware}
\begin{figure}[H]
	\centering	
	\includegraphics[width=0.8\linewidth]{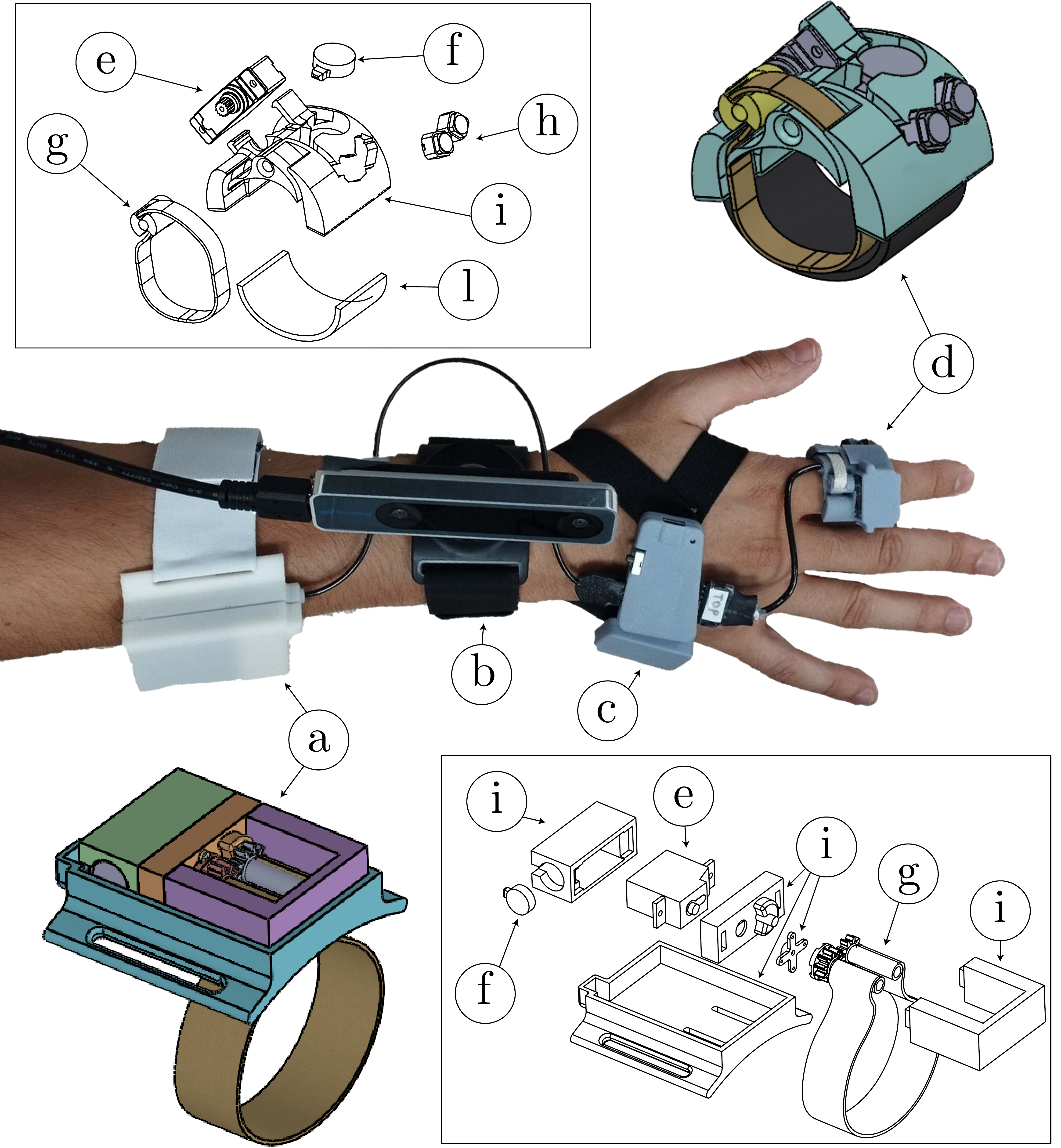}
	\caption[Haptic interface hardware details]{Details of the hardware of the haptic devices. (a) forearm haptic interface, (b) tracking camera, (c) \acrfull{pcb} and battery, (d) ring haptic interface, (e) servomotor, (f) vibromotor, (g) folding mechanism, (h) buttons, (i) structural plastic parts, (l) elastic band.}
	\label{fig:tpoh:hapticinterface}
\end{figure}

In addition to the exploitation of tracking cameras to track the operator's arms movement as in the original \acrshort{tpo} Suit, the wearable sensorimotor interface sends inputs to the \acrshort{tpo} system through buttons and generates haptic feedback to the user in the form of skin indentation and vibratory stimuli. 
Other than the tracking cameras, with the new interface the user wears an armband on each forearm, and a ring on each index finger (\figurename{}~\ref{fig:tpoh:hapticinterface}). 
Both the armbands and the ring can deliver skin indentation stimuli through a fabric belt that is folded and unfolded by a servomotor (Hitech servomotors, models HS-5035HD\footnote{\href{https://hitecrcd.com/products/servos/discontinued-servos-servo-accessories/discontinued-digital-servos/hs-5035hd-digital-ultra-nano-servo/product}{https://hitecrcd.com/products/servos/discontinued-servos-servo-accessories/discontinued-digital-servos/hs-5035hd-digital-ultra-nano-servo/product}} and HS-53\footnote{\href{https://hitecrcd.com/products/servos/analog/micromini/hs-53/product}{https://hitecrcd.com/products/servos/analog/micromini/hs-53/product}}), and vibrations produced by coin vibration motors (Precision Microdrives Model No. 310-103.004, 10mm Vibration Motor\footnote{\href{https://catalogue.precisionmicrodrives.com/product/310-103-10mm-vibration-motor-3mm-type}{https://catalogue.precisionmicrodrives.com/product/310-103-10mm-vibration-motor-3mm-type}}). The rings are also endowed with two push buttons each (Alpsalpine SPEH110100\footnote{\href{https://tech.alpsalpine.com/e/products/detail/SPEH110101/}{https://tech.alpsalpine.com/e/products/detail/SPEH110101/}}). 
The control board is a custom-designed board based on the Espressif ESP\footnote{\href{https://www.espressif.com/en/products/socs/esp32}{https://www.espressif.com/en/products/socs/esp32}}, built to ensure a stable wireless connection and to physically connect all the components needed to program the microcontroller, drive the motors, read the buttons, and recharge the battery in a small amount of space. The result is a lightweight interface: \SI{104}{\gram}, excluding the \SI{114}{\gram} of the camera.
The working principle of the skin indentation devices is based on~\cite{pacchio-hring}, with some improvements. In particular, as shown in \figurename{}~\ref{fig:tpoh:hapticinterface}, both the forearm and the ring haptic devices use only one motor to drive the fabric belt: in the forearm this has been implemented with opposing gears, while in the ring with a winch mechanism. 
The structure has been designed to be compact and adaptable to the physical characteristics of different users. The forearm haptic interface is adjustable with a Velcro fastener, and the ring has an interchangeable elastic band.

\subsection{Haptic Interface Integration}\label{sec:tpoh:integration}
\begin{figure}[H]
	\centering
	\includegraphics[width=\linewidth]{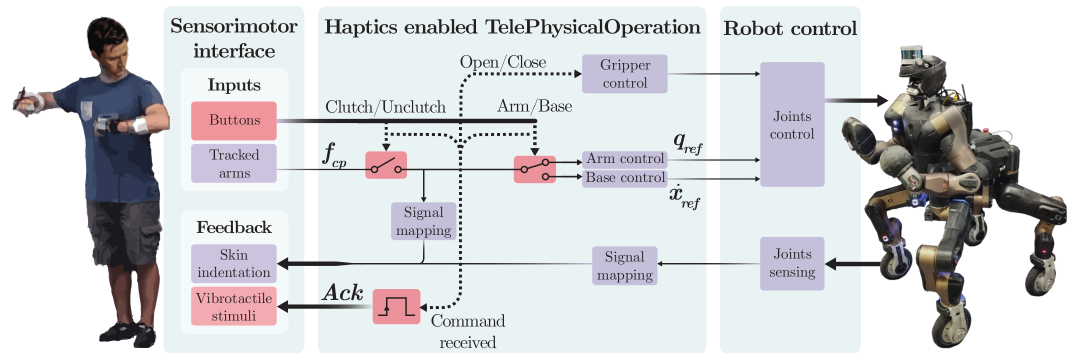}
	\caption[Logical scheme of the haptic-enabled \acrlong{tpo} interface]{Block scheme highlighting the relevant parts of the haptics-enabled \acrlong{tpo} framework. The functionalities of the buttons of the ring devices are indicated with the light pink blocks.}
	\label{fig:tpoh:architecture}
\end{figure}

An overall scheme of the haptic-enabled \acrfull{tpo} interface is depicted in \figurename{}~\ref{fig:tpoh:architecture}. The choice of showing the CENTAURO robot in the scheme is not casual, since it is the robot employed in the experimental validations. Indeed, the mapping between haptic stimuli and delivered information is tailored to this robot and its configuration chosen for the mission, as it will be clear soon. This mapping is summarized in \tablename{}~\ref{tab:map_feedback}. 

\begin{table}[H]
	\centering
	\caption{Mapping of the haptic interface feedback.}
	\label{tab:map_feedback}
	\begin{tabularx}{\linewidth}{l c c}
		
		\toprule
		& Squeeze Feedback & Vibration Feedback \\
		\midrule
		
		\textbf{Right Forearm} & R. virtual force magnitude & R. toggle activation acknowledgment \\
		\textbf{Left Forearm} & L. virtual force magnitude & L. toggle activation acknowledgment \\
		\textbf{Right Finger} & Gripper grasping force & Toggle gripper acknowledgment \\
		\textbf{Left Finger} & L. \acrshort{ee} external force magnitude & Arm/base \textit{cp} change acknowledgment \\
		\bottomrule
	\end{tabularx}
\end{table}

The user, when moving his/her arm to virtually pull/push the robot, experiences a squeezing (indentation) sensation on the forearm generated by the worn armband proportionally to the magnitude of the applied \acrshort{tpo} virtual force. 
In the version of the \acrshort{tpo} interface presented in Chapter~\ref{chap:TPO}, the user could not feel the resistance caused by the application of a virtual force on the robot, which constituted a difference with respect to the application of an actual force during a physical human-robot interaction. Now, the human-robot connection is physically perceived by the user through the haptic feedback, while still keeping the human in a remote position.

It is known the importance of allowing users to feel external forces acting on the robot's end-effector during manipulation tasks~\cite{pacchierotti2017wearable, Billard2019, Coppola2022}. For this reason, these dynamics are mapped with the ring interface. Skin indentation stimuli are delivered on the index fingers, accordingly to the robot's end-effectors interaction with its surroundings. To account for both, prehensile and non-prehensile manipulation actions, we envisage a robot setup in which the right arm is equipped with the DAGANA gripper (Section~\ref{sec:intro:dagana}) and the left arm with a passive ball-shaped end-effector.
Thus, the user's right finger is squeezed based on the grasping force applied by the gripper on a grasped object, while the left finger is squeezed proportionally to the contact force between the robot left end-effector and the environment.

\begin{table}[H]
	\centering
	\caption{Mapping of the haptic interface buttons.}
	\label{tab:map_button}
	\setlength{\tabcolsep}{13pt} %
	\begin{tabularx}{0.60\linewidth}{l c}
		
		\toprule
		& Functionality \\
		\midrule
		
		\textbf{Right button 1} & Right virtual force activation  \\
		\textbf{Left button 1} & Left virtual force activation  \\
		\textbf{Right button 2} & Open/Close gripper  \\
		\textbf{Left button 2} & Arm/base control point change \\
		\bottomrule
	\end{tabularx}
\end{table}

Another improvement introduced in this work is the addition of input buttons embedded in the rings, which functions are summarized in \tablename{}~\ref{tab:map_button}. In the previous version of \acrshort{tpo} (Chapter~\ref{chap:TPO}), the operator could not independently control certain functionalities, like activating and deactivating the teleoperation, changing the control point (e.g., control the robot base instead of the arm), or opening/closing the robot gripper. The solution adopted was the involvement of a second operator who executed these additional commands from a computer according to the first operator's vocal instructions. 
In the new system, thanks to push buttons, the presence of the second operator is not necessary anymore. Furthermore, exploiting the functionalities of the developed haptic devices, two different vibration patterns on the forearm and on the fingers are generated to acknowledge the operator about the correct execution of the functionality associated with the pressed button. A single short vibration signaled the engagement of the functionality, conversely, a double short vibration signaled the disengagement of the functionality. For example, referring to \tablename{}~\ref{tab:map_feedback}, the \enquote{toggle activation acknowledgment} vibrates once when the user activates the teleoperation, and vibrates twice when the user deactivates the teleoperation. %

It is worth noticing that the mapping between signals and feedback stimuli generation and between additional inputs and buttons has been chosen according to what was deemed more intuitive considering the control paradigm and the employment of the interface in a specific locomanipulation mission described in Section~\ref{sec:tph:exp}.
For example, since the user commands the robot through his/her arms, the feedback related to the virtual force is delivered to the forearms. Since humans usually interact with the environment with their hands, the stimuli related to the interactions between robot's end-effectors and the environment are delivered to the rings.
Nevertheless, the software architecture is flexible and different choices can be made. At the current stage the numerous amount of possible mappings have not been yet evaluated, but this is planned in future developments.

\section{Haptic-enabled TelePhysicalOperation Experimental Validations}\label{sec:tph:exp}

\begin{figure}[H]
	\centering
	\includegraphics[width=0.48\linewidth]{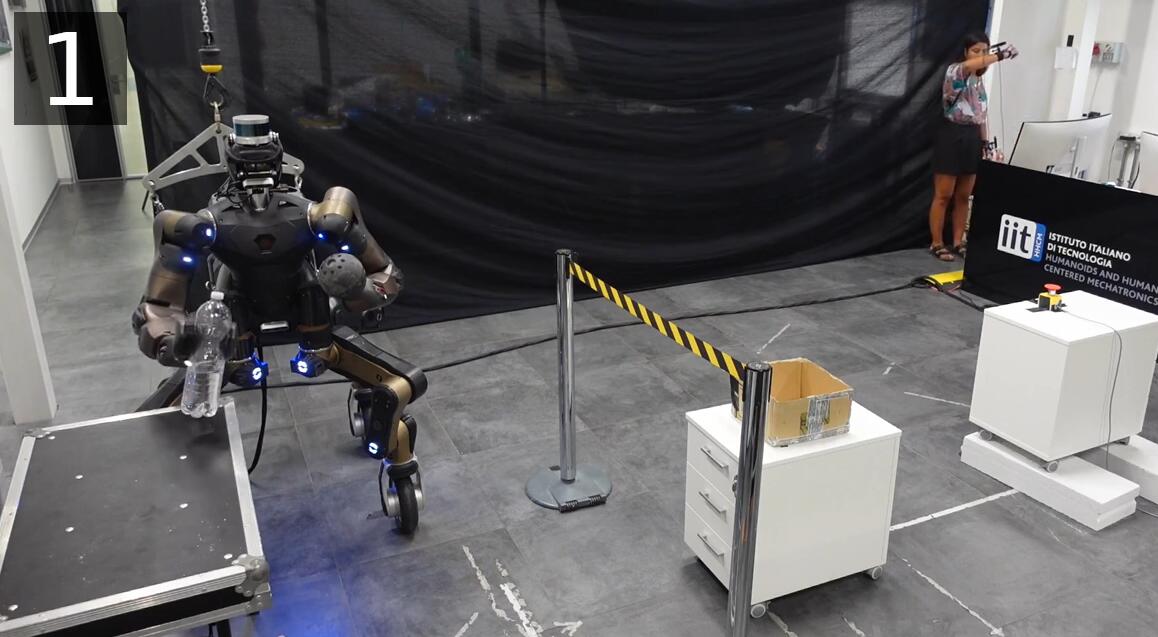}
	\includegraphics[width=0.48\linewidth]{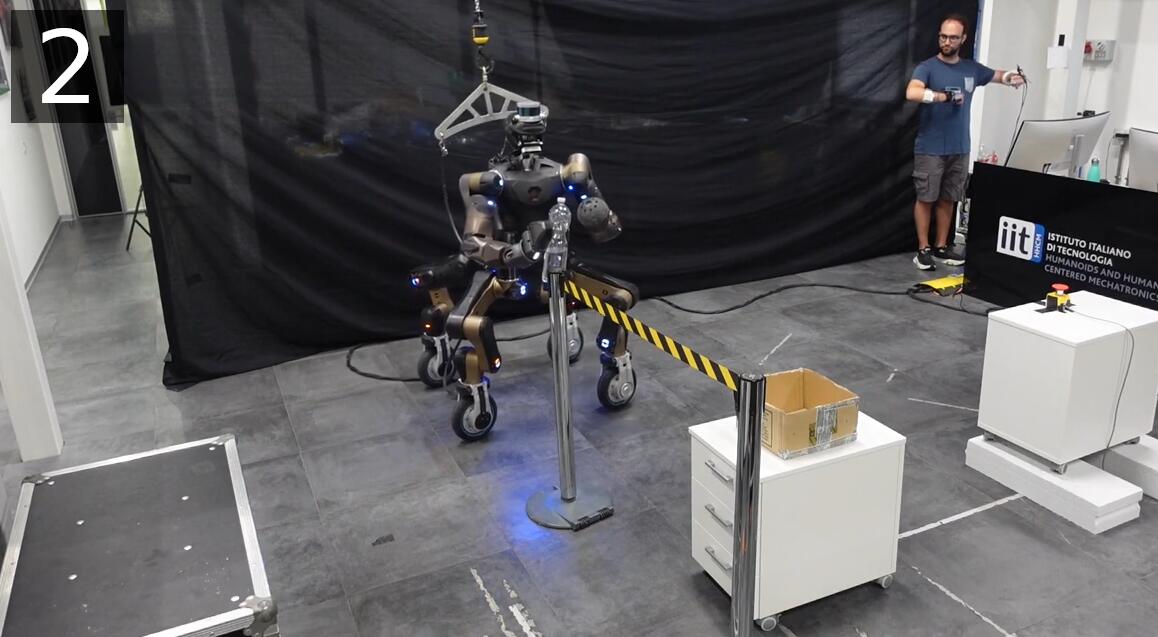}\\
	\vspace{3px}
	\includegraphics[width=0.48\linewidth]{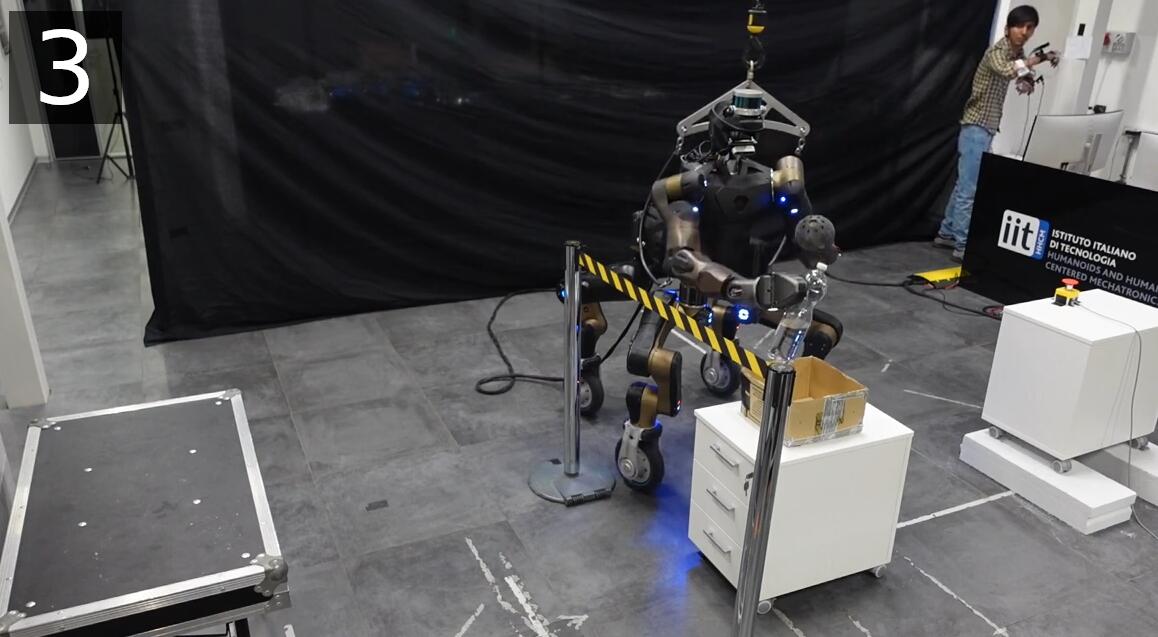}
	\includegraphics[width=0.48\linewidth]{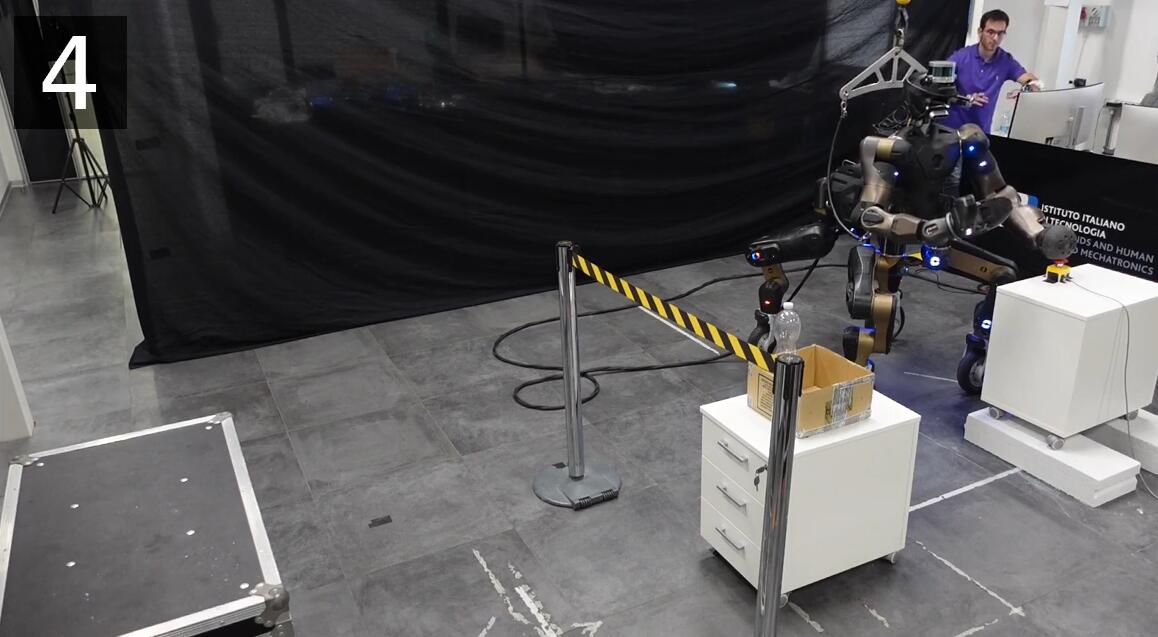}
	\caption[Haptic-enabled \acrshort{tpo} experiment sequences]{Some participants are teleoperating the CENTAURO robot with the TelePhysicalOperation interface in the various tasks of the mission. From top to bottom, left to right: (1) the bottle is picked up with the robot's right gripper; (2) the robot is guided toward the box avoiding an obstacle; (3) the bottle is placed in the box; (4) the button is pressed with the robot's left end-effector.}
	\label{fig:tpoh:exppeople}
\end{figure}

In this section, experimental validations are presented (\figurename{}~\ref{fig:tpoh:exppeople}). Naive users are involved in a mission where the locomanipulation capabilities of the CENTAURO robot (Section~\ref{sec:intro:centauro}) must be exploited. The mission has been conducted to demonstrate the efficacy of the \acrlong{tpo} interface with and without the haptic enhancement.

Highlights of the experiments are shown in the video available at the following link: \url{https://youtu.be/295qqHSaNhY}. The raw videos of all the trials with all the participants is available at \url{https://youtu.be/VTiB6I_fIWo}.

\subsection{Experiment Description}
\begin{figure} [H]
	\centering
	\includegraphics[width=0.8\linewidth]{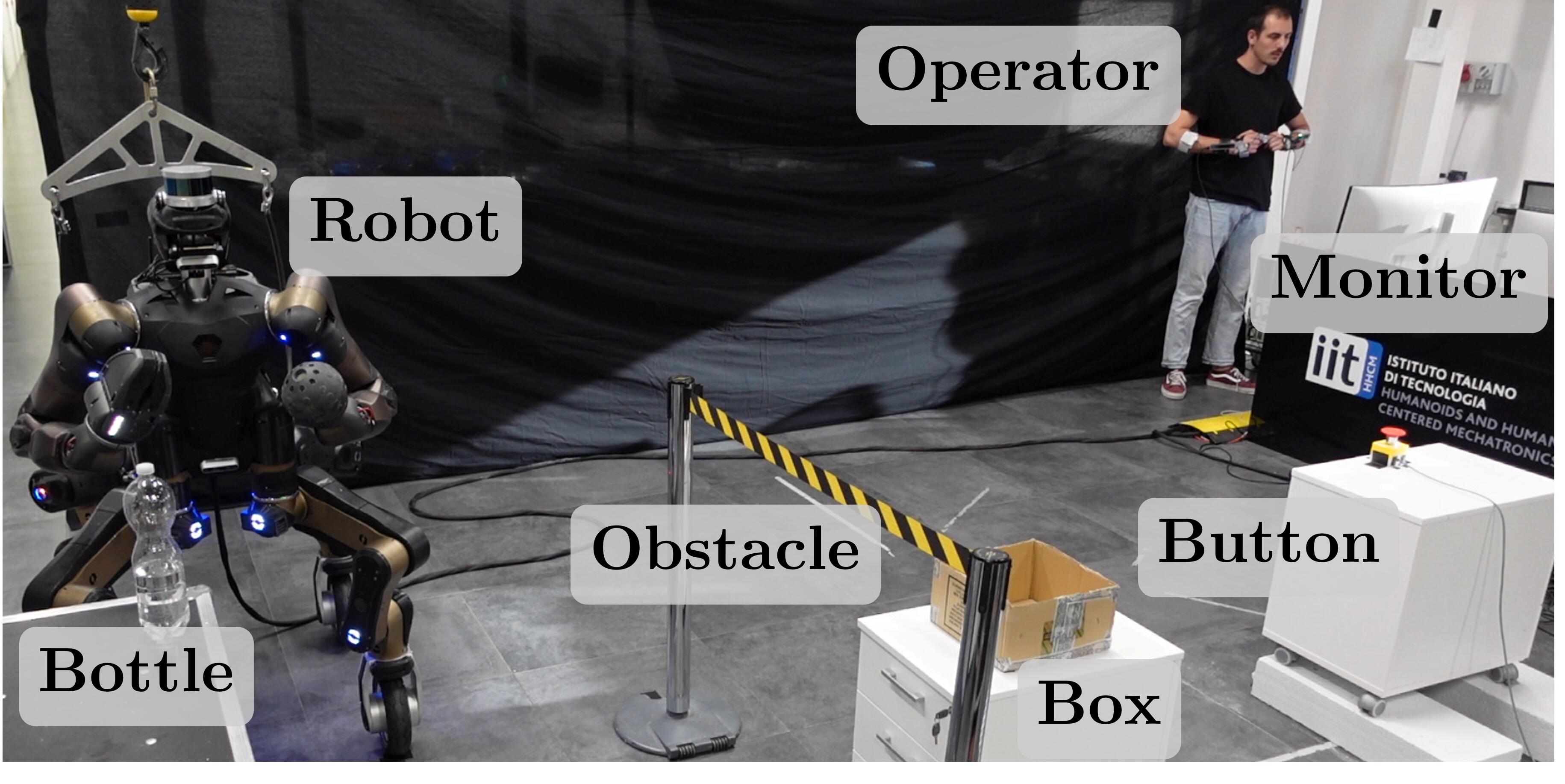}\\
	\vspace{5px}
	\includegraphics[width=0.7\linewidth]{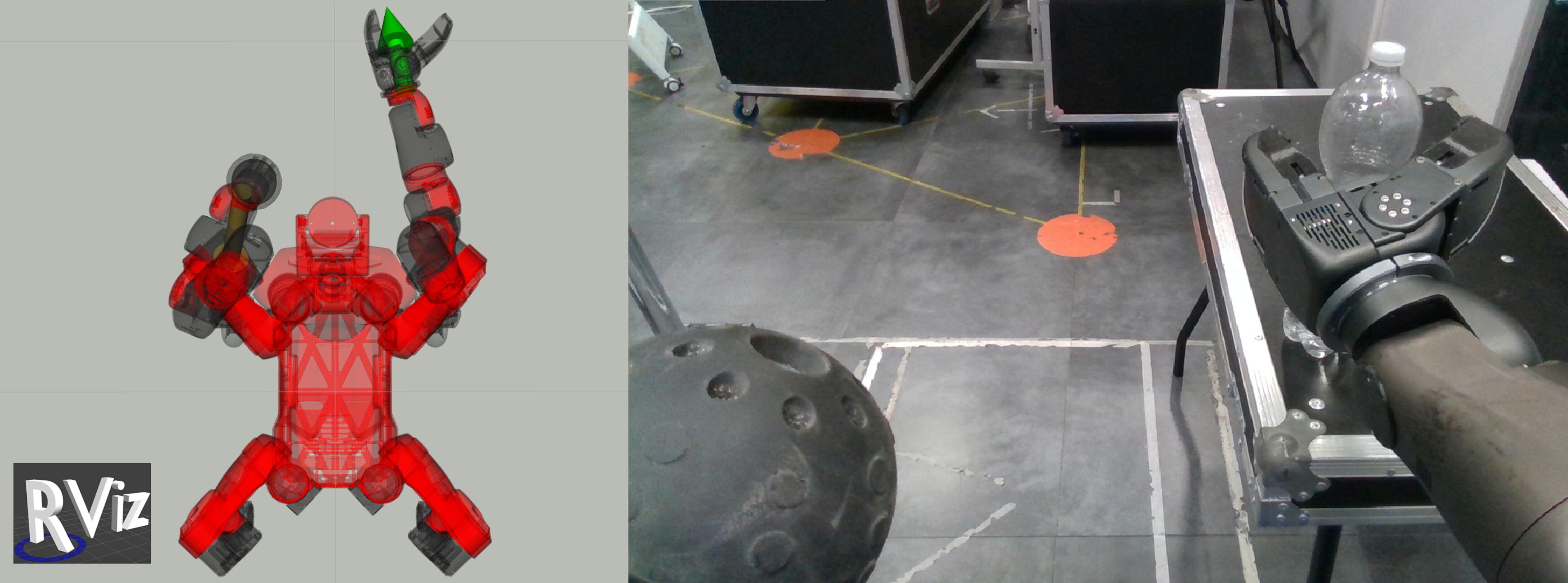}
	\caption[Haptic-enabled \acrshort{tpo} experimental setup]{The experimental setup of the mission. At the top, the scenario chosen with the important elements highlighted. At the botttom, the monitor in front of the operator shows the virtual forces on the robot model (left) and the output of the robot's head camera (right).}
	\label{fig:tpoh:setup}
\end{figure}

The setup where the user teleoperates the robot is shown in \figurename{}~\ref{fig:tpoh:setup}.
The operator controls the robot standing in front of a monitor, with the freedom to look at it or directly at the robot at any time as she/he prefer. The monitor, as shown in the bottom part of \figurename{}~\ref{fig:tpoh:setup}, shows an RViz environment with the robot twin and arrows representing the virtual forces applied to the robot parts (a visual feedback tool presented in Section~\ref{sec:tpo:visual}), and the CENTAURO head camera output.
During the experiments, the operator has the control on the CENTAURO manipulation (left/right arms and the right gripper) and wheeled mobility abilities. 

The mission involves a series of manipulation tasks positioned at different locations to be completed sequentially, as depicted in the sequences of \figurename{}~\ref{fig:tpoh:exppeople}, where some participants are operating the robot at the different locations.
At the beginning, the robot is in front of a table where a bottle must be picked up with the right gripper (top left of \figurename{}~\ref{fig:tpoh:exppeople}). Then, avoiding an obstacle (top right), the robot must be guided to another location to place the bottle inside a box (bottom left). Lastly, the robot must be moved close to an emergency button that must be pressed with the left end-effector (bottom right). A single task is considered failed if the bottle falls off or if the robot hits an object (the obstacle, the table, or one of the drawers supporting the box and the button), but the rest of the mission can continue (if the bottle drops, it is manually placed in the robot gripper to allow the continuation of the other tasks). The stopwatch counting the time of the mission starts as soon the operator activates the teleoperation for the first time and ends when the emergency button at the last location is correctly pressed.

\subsection{Specification of the TPO Interface Utilized in the Experiments}
The user controls the robot with the \acrlong{tpo} interface generating virtual forces from his/her arm motions as explained in Chapter~\ref{chap:TPO}.
The control points $\mathit{cp}$ where it is possible to apply the virtual forces are limited to the right and left end-effectors and the robot base. This means that the user can control the two robot arms and the mobile base motion in the $x-y$ plane.
The right arm of the user is always connected to the right end-effector of the robot, while the left arm of the user can switch between the left end-effector and the base of the robot.
Besides generating virtual forces with free arm movements, the user has the possibility to exploit $4$ additional input commands to: activate/deactivate the control (one for each arm), open/close the gripper, and switch the left arm control point. 
The way these inputs are commanded depends on the condition (\textit{A}, \textit{B}, \textit{C}) used in the experimental comparison:

\begin{itemize}
	\item \textbf{Condition \textit{A}.} The operator has no haptic feedback, nor push buttons and necessitates a second operator to command the $4$ additional inputs described above. The second operator gives vocal acknowledgments after she/he executes the requested command (\enquote{Ok} word). This condition corresponds to the original version of the \acrshort{tpo} interface (Chapter~\ref{chap:TPO}).
	
	\item \textbf{Condition \textit{B}.} The operator can use the push buttons but she/he does not receive any haptic feedback. The buttons are mapped as in \tablename{}~\ref{tab:map_button}.
	
	\item \textbf{Condition \textit{C}.} The operator has the full wearable interface available, including buttons and haptic feedback. Buttons are mapped as in condition \textit{B}, while the feedback are mapped as in \tablename{}~\ref{tab:map_feedback}
\end{itemize}

\noindent \tablename{}~\ref{tab:map_inputs} summarizes how the user commands the $4$ additional inputs in the three different conditions. Please note that for conditions \textit{B} and \textit{C} this table is equivalent to \tablename{}~\ref{tab:map_button}.

\begin{table}[H]
	\centering
	\caption{Mapping of the inputs for the $3$ modalities.}
	\label{tab:map_inputs}
	\setlength{\tabcolsep}{6pt} %
	\begin{tabularx}{.88\linewidth}{l c c}
		\toprule
		& Vocal Input (A) & Button Input (B, C)\\ 
		\midrule
		\textbf{Right virtual force activation} & \enquote{Right} & Right button 1    \\
		\textbf{Left virtual force activation} & \enquote{Left} & Left button 1   \\
		\textbf{Open/Close gripper} & \enquote{Gripper} & Right button 2    \\
		\textbf{Left EE/Base control point change} & \enquote{Change} & Left button 2   \\
		\bottomrule
	\end{tabularx}
\end{table}

\subsection{Experiment Details and Questionnaires}
The experiments involved $12$ participants, $10$ males, $2$ females, with age ranging from $25$ to $39$, in average $29.83\pm4.41$, voluntarily participating after having signed an informed consent.
Each experiment lasted about $2$ hours and consisted in a \textit{training phase} followed by a \textit{testing phase}. The training phase lasted around $40$ minutes during which the experimenter taught the participant how to control the robot and let the participant try the mission in the 3 different conditions. Furthermore, the device squeezing range has been set up according to the user sensibility and physical characteristics. 
In the testing phase, each participant executed the whole mission $6$ times, $2$ for each condition. The order of the conditions has been permuted following a certain sequence for each participant (e.g., \textit{ABC}, \textit{ACB}, \textit{BAC}, etc.). Immediately after the two trials for each condition, participants were asked to fill in the NASA-TLX questionnaire referred to the tested condition. At the end of all the $6$ trials, participants assigned the weights to the NASA-TLX factors, and compiled two additional questionnaires. 
The first questionnaire was about the wearable interface and was composed of $16$ questions formulated as $5$-level Likert items with answers varying
from \enquote{Strongly disagree} to \enquote{Strongly agree}, similarly to~\cite{casalino2018operator}.
The second questionnaire compared the three conditions one-vs-one (i.e., \textit{A}vs\textit{B}, \textit{A}vs\textit{C}, \textit{B}vs\textit{C}) and was formulated as a 7-point linear scale going from a certain condition to another, similarly to~\cite{maderna2022flexible}. 
For example, for each question of the questionnaire \textit{A}vs\textit{B}, the possible answers were \enquote{Strongly A}, \enquote{A}, \enquote{Slightly A}, \enquote{Neutral}, \enquote{Slightly B}, \enquote{B}, \enquote{Strongly B}.
The statements in the first questionnaire were customized to the specific application presented in this paper, while the statements in the second one include the whole SUS~\cite{brooke1996sus} and two additional questions taken from the USE Questionnaire~\cite{lund2001measuring}.

\section{Haptic-enabled TelePhysicalOperation Experiment Results Discussions}\label{sec:tpoh:disc}

\begin{table}[H]
	\centering
	\caption[Haptic-enabled \acrshort{tpo}: generic results from the user validation]{Experimental results considering all the trials of the three different conditions.}
	\label{tab:tpohresults}
	\setlength{\tabcolsep}{13pt} %
	\begin{tabularx}{0.70\linewidth}{l c c c}
		\toprule
            & \multicolumn{2}{c}{Completion Time [s]} & Task Failures \\
		  & All & Last Two Trials &  \\
		\midrule
		\textbf{\textit{A}} & $168.5\pm28.0$ & $150.1\pm28.1$ & 2 \\
		  \textbf{\textit{B}} & $178.5\pm2.1$ & $172.5\pm13.4$ & 1 \\
            \textbf{\textit{C}} & $182.6\pm17.2$ & $148.2\pm6.0$ & 1 \\
		\bottomrule
	\end{tabularx}
\end{table}

\begin{figure} [H]
	\centering
	\includegraphics[width=0.9\linewidth]{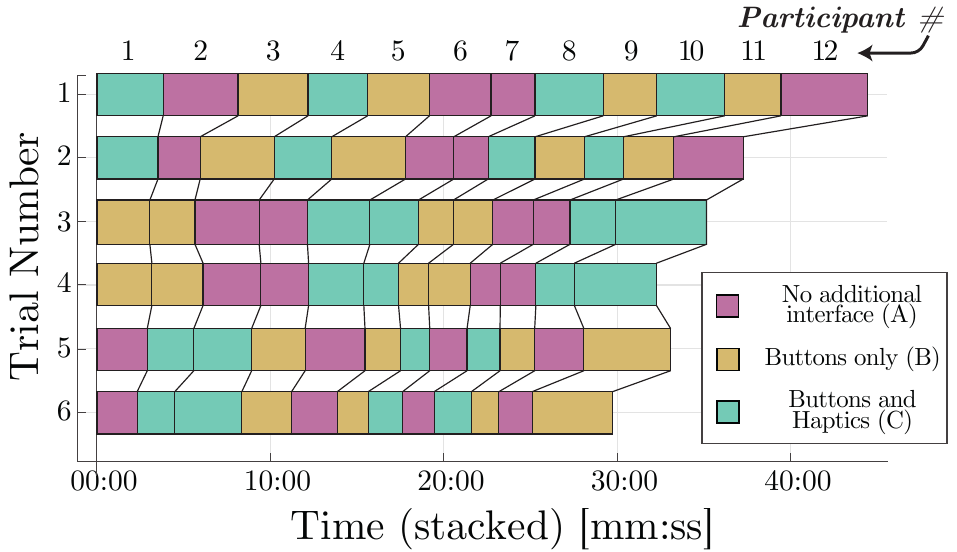}
	\caption[Haptic-enabled \acrshort{tpo}: completion times]{Completion times of each trial for all the participants, where the conditions \textit{A}, \textit{B}, \textit{C} have been highlighted in different colors.}
	\label{fig:tpoh:times}
\end{figure}

Obtained results about the completion time averaged among all participants and about the number of total failures are reported in \tablename{}~\ref{tab:tpohresults}. 
No statistical significance was found among the completion time values in the different conditions (using a one-way repeated measure \acrshort{anova}, $p > .05$). 
Anyway, a clear trend emerges considering all trials and all users, independently of the condition: the cumulative completion time improved from the first to the last trial, as shown in \figurename{}~\ref{fig:tpoh:times}. 
This is the reason why in \tablename{}~\ref{tab:tpohresults} also the completion time employed in the last two trials is reported. In the table we can see that the condition \textit{C} performs slightly better than the other two, but still without statistical significance.
In general, each participant improved his/her performance in terms of completion time along the trials, as displayed in \figurename{}~\ref{fig:tpoh:times}. 
The occurred failures are only a few and happened while grasping the bottle or in cases where the robot collided with other objects in the scene.

\begin{figure}[H]
	\centering
	\includegraphics[width=\linewidth]{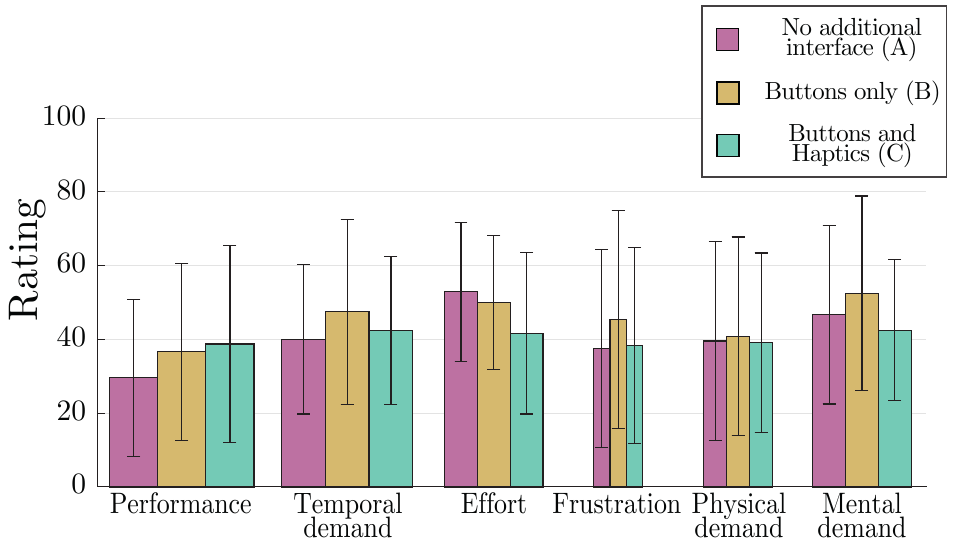}
	\caption[Haptic-enabled \acrshort{tpo}: NASA-TLX Ratings of the conditions]{NASA-TLX Ratings of the three conditions. Overall workload is depicted in \tablename{}~\ref{tab:tpoh:nasaow}}
	\label{fig:nasatlx}
\end{figure}
\begin{table}[H]
	\centering
	\caption[Haptic-enabled \acrshort{tpo}: NASA-TLX overall workload results]{Experimental results considering all the trials of the three different conditions.}
	\label{tab:tpoh:nasaow}
	\begin{tabularx}{0.41\linewidth}{l c}
		\toprule
		& NASA-TLX Overall Workload\\
		\midrule
		\textbf{\textit{A}} & $42.0\pm17.4$ \\
		\textbf{\textit{B}} & $47.25\pm19.1$\\
		\textbf{\textit{C}} & $43.02\pm17.6$\\
		\bottomrule
	\end{tabularx}
\end{table}

The results for the single weighted categories of NASA-TLX are plotted in \figurename{}~\ref{fig:nasatlx} according to the guidelines in~\cite{hart1988development}, while the overall workload (OW) is shown in \tablename{}~\ref{tab:tpoh:nasaow}. 
Slight differences are visible in favor of the condition \textit{A} for the performance (PE), weighted more than the other categories, and in favor of the condition \textit{C} for effort (EF) and mental demand (MD). 
Anyway, there was no statistically significant difference between the means of the ratings provided for the different interfaces (one-way repeated measures \acrshort{anova}, $p > .05$). 
The fact that all conditions show comparable results in terms of workload indicates that adding an extra feedback channel with respect to the visual one does not generate an additional burden on the users. The obtained overall workload is in line with the average values found for \enquote{Robot Operation} tasks reported in previous works~\cite{grier2015high,prattico2021towards}.

\begin{figure}[H]
	\centering
	\includegraphics[width=.95\linewidth]{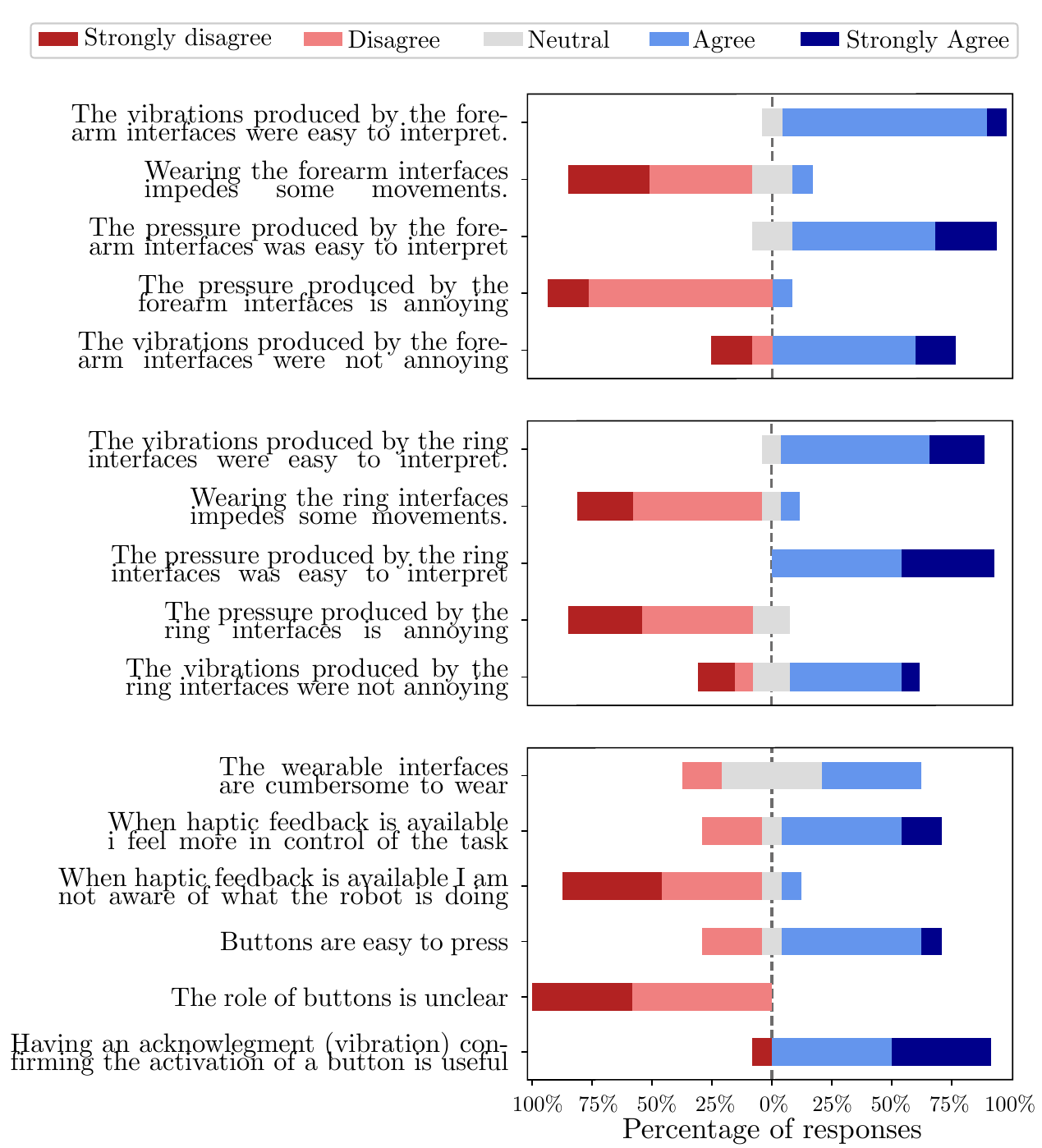}	
	\caption[Haptic-enabled \acrshort{tpo}: Likert questionnaire results about the haptic devices.]{The results of the first Likert questionnaire about the haptic devices.}
	\label{fig:tpoh:likert}
\end{figure}

The results of the first questionnaire, valuating in general the wearable haptic devices and their integration, are reported in \figurename{}~\ref{fig:tpoh:likert}.
In particular, users assessed the features of the single devices, forearm (top plot) and ring (middle plot), and of the whole haptic interface (bottom plot). The results reflect positive assessments of the haptic interface, with users commending its effectiveness and the benefits of incorporating such haptic stimuli in the interface.

\begin{table}[H]
	\centering
	\caption[Haptic-enabled \acrshort{tpo}: questions of second questionnaire]{The $12$ questions of the second questionnaire. Numbers in red indicate the negative ones.}
	\label{tab:tpoh:questions}
	\begin{tabularx}{\textwidth}{ c  l }
		\toprule
		\textbf{1.}     & I think that I would like to use this system frequently.\\
		{\color{bittersweet}\textbf{2.}} & I found this system unnecessarily complex.\\
		\textbf{3.} & I thought this system was easy to use.\\
		{\color{bittersweet}\textbf{4.}} & I think that I would need assistance to be able to use this system.\\
		\textbf{5.} & I found the various functions in this system were well integrated.\\
		{\color{bittersweet}\textbf{6.}} & I thought there was too much inconsistency in this system.\\
		\textbf{7.} & I would imagine that most people would learn to use this system very quickly.\\
		{\color{bittersweet}\textbf{8.}} & I found this system very cumbersome/awkward to use.\\
		\textbf{9.} & I felt very confident using this system.\\
		{\color{bittersweet}\textbf{10.}} & I needed to learn a lot of things before I could get going with this system.\\
		\textbf{11.} & I can recover from mistakes quickly and easily\\
		\textbf{12.} & I can use it successfully every time.\\
		\bottomrule
	\end{tabularx}
\end{table}
\vspace{-5px}

\begin{figure}[H]
	\centering
	\includegraphics[width=\linewidth]{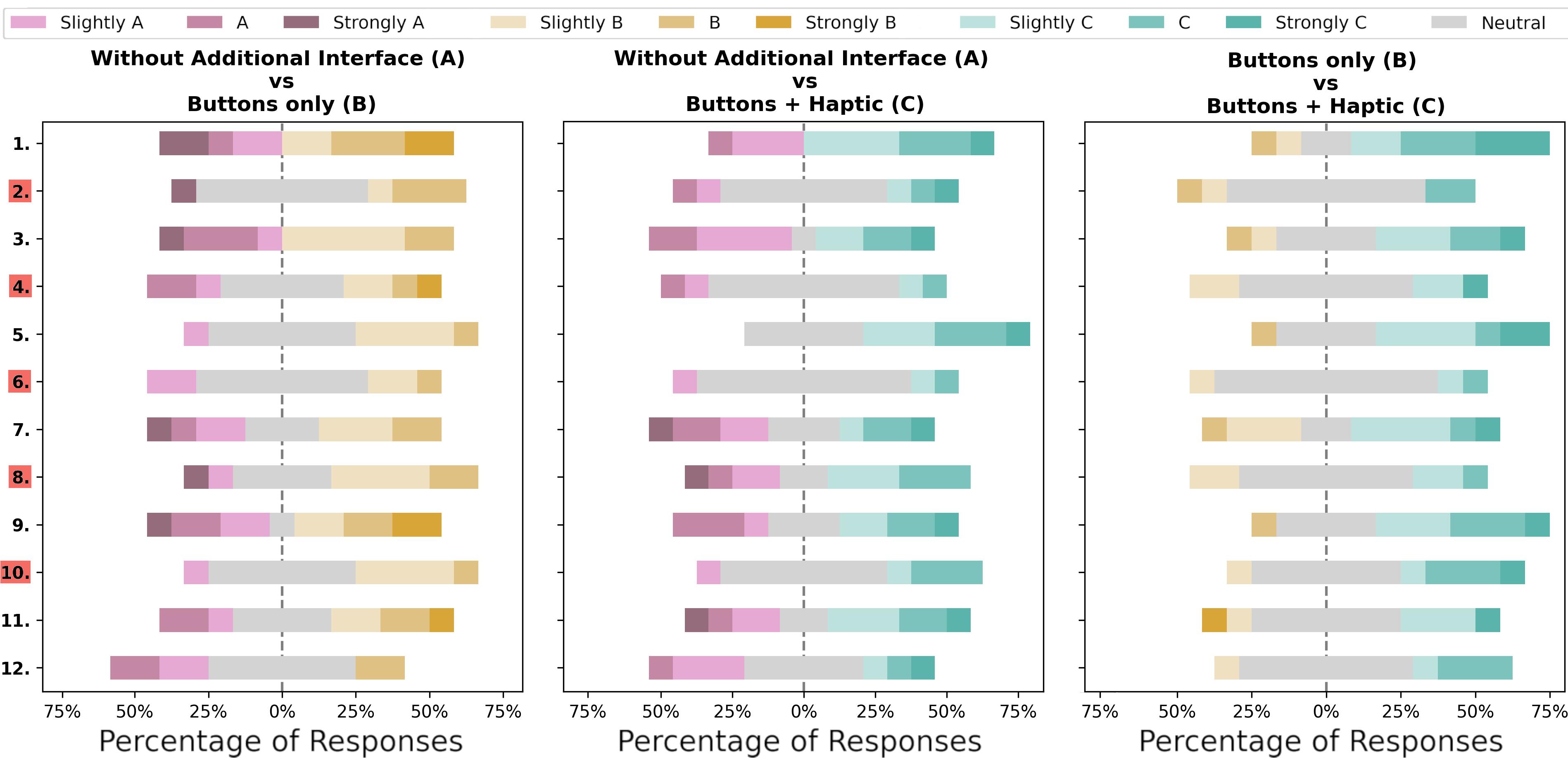}
	\caption[Haptic-enabled \acrshort{tpo}: Likert comparison between the conditions]{The results of the second questionnaire, comparing the $3$ conditions against each other. The labels of the negative questions are highlighted in red.}
	\label{fig:tpoh:avsbvsc}
\end{figure}

The questions of the second questionnaire are shown in \tablename{}~\ref{tab:tpoh:questions}, while the results, averaged among all participants, are shown in the one-vs-one plots of \figurename{}~\ref{fig:tpoh:avsbvsc}. Please note that some questions address a negative aspect (2, 4, 6, 8, 10) hence the relative answers represent a negative outcome for the selected condition. 

\begin{figure}[H]
	\centering
	\includegraphics[width=\linewidth]{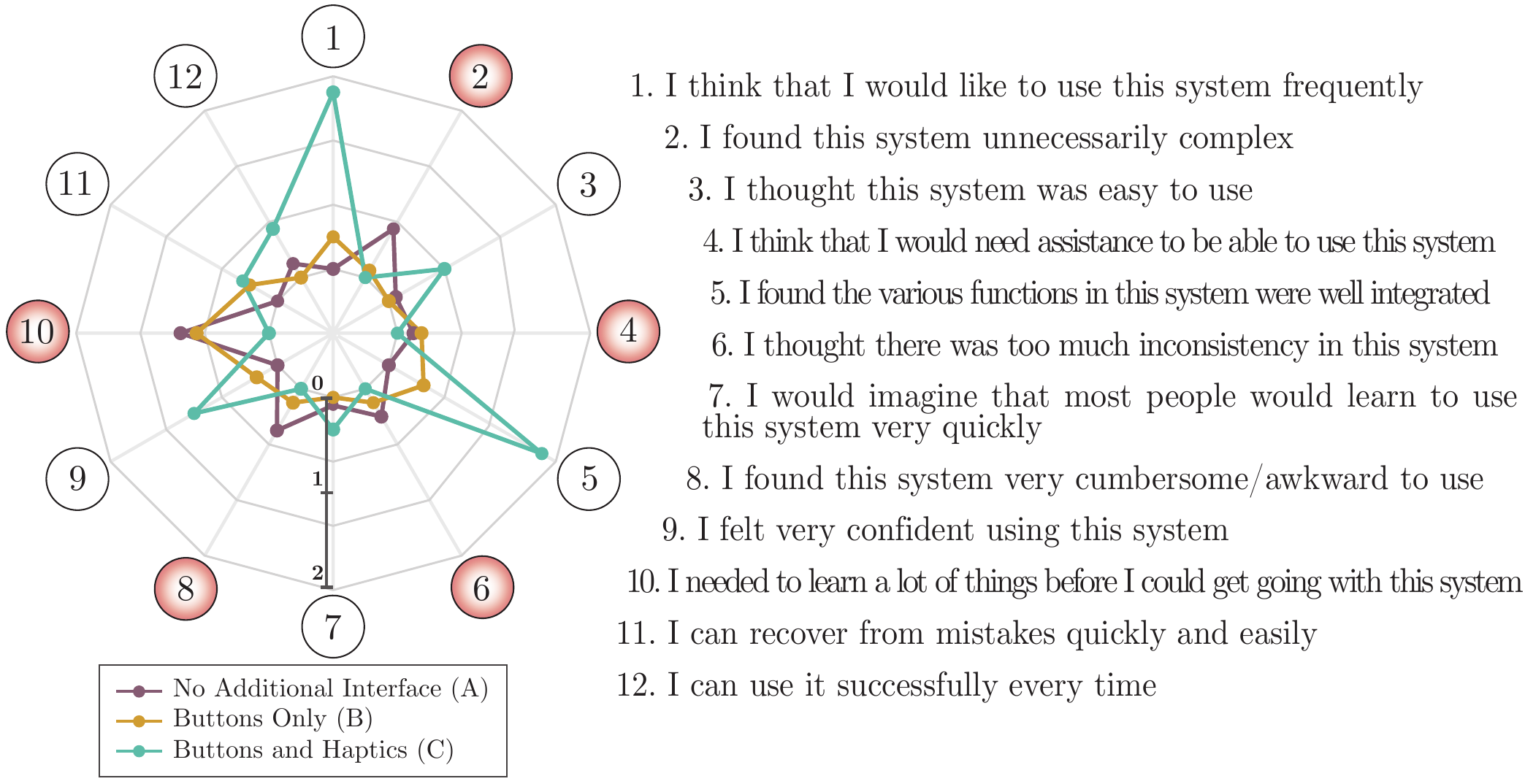}	
	\caption[Haptic-enabled \acrshort{tpo}: spider plot with score extracted from the Likert comparison between the conditions]{The results of the second questionnaire, comparing the $3$ conditions all together in an unique spider plot. Questions of \tablename{}~\ref{tab:tpoh:questions} are reported again for convenience. Data is gathered from the second questionnaire as in \figurename{}~\ref{fig:tpoh:avsbvsc}, but presented differently, computing a score for each question. The outer ring represents a score of $2$, while the inner one a score of $0$. Points from the negative questions (red circles) have been reversed to consider always a high score as a positive rating for a condition.}
	\label{fig:tpoh:avsbvsc_spider}
\end{figure}

The results of the second questionnaire have also been used to compute a comprehensive score for each question resulting in the plot of \figurename{}~\ref{fig:tpoh:avsbvsc_spider}. 
The score is computed as explained in what follows.
Considering the $7$-point linear scale going from a certain condition to the other, for each question of the three one-vs-one comparisons, no score has been assigned if the participant selected the middle point (\enquote{Neutral} answer, shown in gray in \figurename{}~\ref{fig:tpoh:avsbvsc}).
For the positive questions ($1$, $3$, $5$, $7$, $9$, $11$, $12$) a score of $+1$, $+2$ or $+3$ is given to the condition preferred depending on the point of the linear scale chosen. 
On the contrary, for the negative questions ($2$, $4$, $6$, $8$, $10$), the score of $+1$, $+2$ or $+3$ has been given to the condition opposed to the selected point, to always consider a spike in the spider plot as a positive rating. Indeed, please note that, differently from \figurename{}~\ref{fig:tpoh:avsbvsc}, here a spike always reflects a positive outcome.
For each question, the resulting score shown in the spider plot is the average among all participants. Hence, for each question, each condition can have a score ranging from 0 to 6, since within the three comparisons each condition compares twice. 

In general, we can notice the preferences for the condition \textit{C} against the two others, (as visible from the results of questions $1$, $9$, $11$, $12$), considering it only slightly more complicated (as per questions $2$, $3$, $7$, $8$), hence feeling the necessity to have more time to learn it (questions $4$, $10$), but judging the haptic interface well integrated (questions $5$, $6$). 
Interestingly, the condition with only the buttons available (\textit{B}) is not always preferred over the condition without the additional interface (\textit{A}). This is also demonstrated from the results of the NASA-TLX overall workload of \tablename{}~\ref{tab:tpoh:nasaow}. 
According to some users' free answers, their preference toward the condition \textit{A} was due to the availability of the vocal commands, since they did not have to use their hands, nor to learn the buttons mapping, to command the additional inputs when necessary. On the contrary, other users, experienced with handheld controller devices (video games, remote controlled drones, robot teleoperation), strongly preferred the buttons with respect to the vocal commands, because they found them more efficient, responsive, and able to give them more independence.
It is important to notice that the vocal commands were directed to a human operator who, in all cases, perfectly understood the commands and rapidly reacted to them. In the same situation, a speech recognition software might have been less responsive and accurate.
Moreover, the second operator always gave an auditory feedback after the execution of a command, which instead is not present after pressing a button in the condition \textit{B}. Instead, in the condition \textit{C}, a vibration feedback acknowledging the correct execution of the command associated with a button is present, and it has been considered very important to have by the participants. Anyway, the conditions \textit{B} and \textit{C} have both the evident advantage of not necessitating the second operator to execute the additional inputs, at the minimal cost of additional control buttons.

\section{Conclusions}

In Chapter~\ref{chap:TPO}, the concept of \acrfull{tpo} has been presented and validated.
In this chapter, an evolution of the \acrshort{tpo} framework has been explored, integrating additional sensing and feedback functionalities by means of a wearable haptic interface, composed of forearm and ring devices to be worn on each arm of the user.
By mapping the forearm squeezing feedback on the virtual forces commanded with the \acrshort{tpo} interface, the users have a direct sensation about the virtual forces applied on the robot, consistent with the \enquote{Marionette} metaphor of the TPO interface (Section~\ref{sec:tpoh:intro}).

The developed haptic interface has been designed to be lightweight and unobtrusive, following the characteristic of the previous \acrshort{tpo} Suit. Haptic feedback and additional inputs have been mapped considering its integration in the \acrshort{tpo} control paradigm and the validation of the interface in a specific locomanipulation mission.
Other than providing feedback about the virtual forces applied on the robot, other squeezing forces on the index fingers alert the user about external forces experienced by the robot end-effectors. The additional input buttons have been added to the ring devices to expand the user's range of inputs, surpassing the previously necessary vocal interaction with a second operator (Section~\ref{sec:tpoh:detail}).

Twelve participants have been involved in validating the \acrshort{tpo} way of control both with and without the proposed haptic interface. The mission consisted in a series of tasks where the manipulation and locomotion abilities of the CENTAURO robot had to be exploited (Section~\ref{sec:tph:exp}).
 
A NASA-TLX and two linear scale questionnaires have been distributed to let the naive users judge the interface in three different conditions. 
Results show a positive outcome for the devices integrated in the interface. Furthermore, comparing the haptics-enabled \acrshort{tpo} with the conditions without haptic feedback, results show a general subjective preference for the first, considering only a slight increase in complexity to wear the enhanced interface (Section~\ref{sec:tpoh:disc}).

Future plans related to this work will go toward a more in-depth study of the single feedback and its effect on the TelePhysicalOperation and other teleoperation interfaces, as well as toward a validation and comparison of other potential mapping configurations between the sensorimotor functionalities and the robot actions and interactions with the environment.

\chapter{Enhancing TelePhysicalOperation with Robot Autonomy Features}\label{chap:tpoAuto}

\begin{figure} [H]
	\centering
	\includegraphics[width=0.8\linewidth]{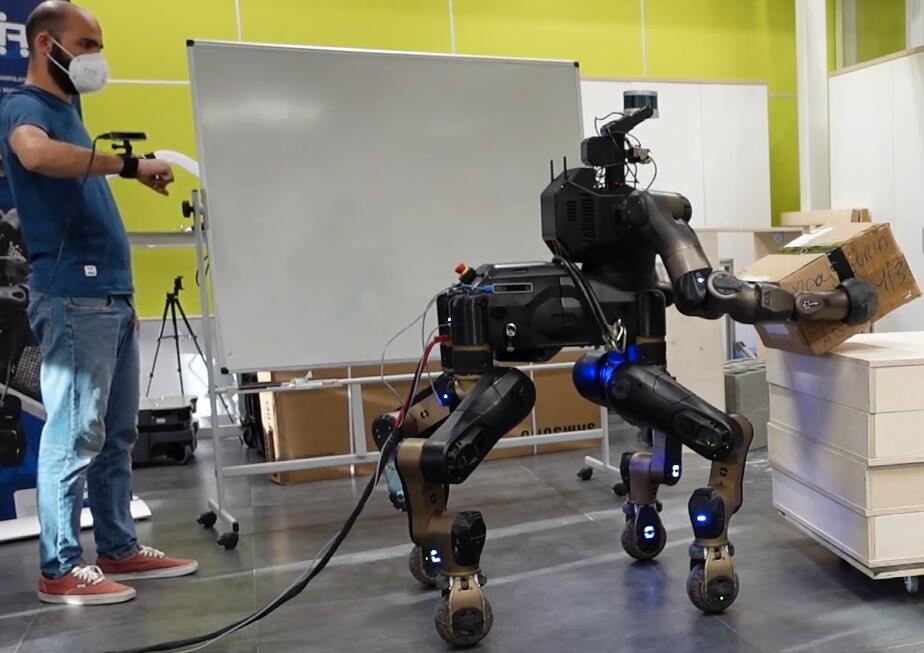}
	\caption[CENTAURO robot teleoperated for a bimanually trasporting mission]{The CENTAURO robot is being teleoperated to bimanually transport a box, exploiting autonomy features to help the operator in accomplish the task. This particular experiment is described in Section~\ref{sec:tpo2:box}.}
	\label{fig:tpo23:firstCentauro}
\end{figure}

\lettrine{I}{n} Chapter~\ref{chap:chap3} it has been stressed the fact that human-robot interfaces must equip the robot with a level of autonomy to relieve the operator from controlling every aspect of the robot motion and of the task. This chapter faces this matter by exploring new robot autonomy functionalities, integrated in the TelePhysicalOperation architecture introduced in Chapter~\ref{chap:TPO}, to help the operator in controlling the various capability of highly-redundant robotic systems during tasks like bimanual transportation (\figurename{}~\ref{fig:tpo23:firstCentauro}).\\

\noindent This chapter is based on the following articles:\\
\fullcite{TPO2}~\cite{TPO2}\\
\fullcite{TPO3}~\cite{TPO3}

\section{Introduction}

In the Chapter~\ref{chap:TPO}, the TelePhysicalOperation interface has been introduced as a new intuitive paradigm for the control of robotic systems. 
Alongside its intuitiveness, it has been shown that a certain amount of robot autonomy is useful and sometimes fundamental to facilitate the human-robot interaction, especially with highly-redundant robots and complex tasks. For this reason, some autonomy features were already incorporated in the first version of the TelePhysicalOperation interface (Section~\ref{sec:tpo:auto}). 
In this chapter, following the previous works about the exploration of different levels of robot autonomy (Section~\ref{sec:soa:auto}), more interesting autonomy functionalities, developed within the shared control paradigm~\cite{Selvaggio2021}.
These methods are integrated in the TelePhysicalOperation architecture, enhancing the overall interface.

\subsection{A Shared Control Locomanipulation Interface} 
One of the developed robot autonomy functionality is a shared control locomanipulation motion generation interface, designed for mobile manipulator robots.
When the operator commands a mobile manipulator for locomanipulation tasks without any help from the robot in the generation of motions, he/she must take care of both its manipulation and locomotion abilities.  
Usually this is achieved by commanding the arm and the base separately. This is not advisable because it would augment the operator burden in the sense that he/she has to continuously switch between the two control modality, possibly increasing also the execution time of the task. 
An alternative is the employment of input devices designed to control at the same time the two systems (the arm and the mobile platform). The problem is that this increases the complexity of the interface and may augment even more the operator workload, since he/she must control at the same time systems with very different characteristics.

Instead of following these solutions, it is more efficient to equip the robot with a certain amount of independence, to share the control with the operator in the generation of some of the necessary motions. 
This direction is followed by the developed manipulability-aware shared locomanipulation interface, detailed in Section~\ref{sec:tpo2:manipControl}. 
This method allows the operator to control the robot's locomanipulation ability by providing commands only for the robot end-effector, while the interface generates arm and base motions maintaining the manipulability measure of the end-effector at a certain level. The manipulability is a measure generally used to understand the vicinity to kinematic singularities, exploited for both the design and control of robots, as briefly recapped in Section~\ref{sec:soa:manip}.

\subsection{A Shared Control Bimanual Transportation Interface}\label{sec:tpo3:intro}
Another autonomy feature developed regards the teleoperation for bimanual transportation of a load. One of the main challenges of this task is that the operator is not aware of the forces applied on the object to be transported. Hence, it is difficult to control the two arms in a coordinate way in order to transport the object safely. It may happen that the commanded motions result in the robot not applying enough grasping forces, making the object slip and fall, or in the robot applying very high grasping forces, using more effort than necessary to handle the object. 
In Chapter~\ref{chap:TPO}, the TelePhysicalOperation interface has been validated with a bimanual pick-and-place operation of a box, where it was active the \textit{Mirroring Motion} feature, which mirrored the input for the robot arm on the other one to help the operator in moving the robot arms coordinately. Even if this helped the user in coordinating the robot's arms to place the load, for a proper bimanual transportation this is not sufficient, and more specific features are required.
In Chapter~\ref{chap:TPOH}, situation awareness in the form of haptic feedback was delivered to the user. Although tactile feedback may help in understanding the grasping forces applied on the load, this may be not enough to allow a safe transportation. Even with highly realistic haptic feedback, there is no guarantee that the operator would be able to coordinate the two robot arms' precisely as necessary. In any case, this would augment his/her cognitive effort. Furthermore, it is challenging to have a good force feedback fidelity while maintaining the simplicity of the wearable interface. 

For these reasons, for such a task, it is better to provide the robot with specific autonomy features to handle correctly the transported load. Previous works have faced this challenge implementing different solutions (Section~\ref{sec:soa:bimanual}).
Following the same direction, in Section~\ref{sec:tpo3:methods}, a bimanual grasping and transportation interface is detailed, developed to facilitate the teleoperation of pick-and-place operations of objects. With this interface, the robot is able to regulate the arm motions accordingly to the required amount of grasping force, which depends on the estimated object mass. Simultaneously, the robot accepts user commands about the directions in which the objects must be transported.
This allows the operator to focus solely on commanding the object velocities, without worrying about the robot's arms motions necessary to hold the object without loosing contact or causing damage due to excessive forces.
This function is combined with the manipulability-aware shared locomanipulation feature to transparently distribute the operator's command related to the velocity of the bimanually grasped object both on the arms and on the mobile base.

\section{Manipulability-aware Shared Locomanipulation}\label{sec:tpo2:manipControl}

\begin{figure} [H]
	\centering
	\includegraphics[width=1\linewidth]{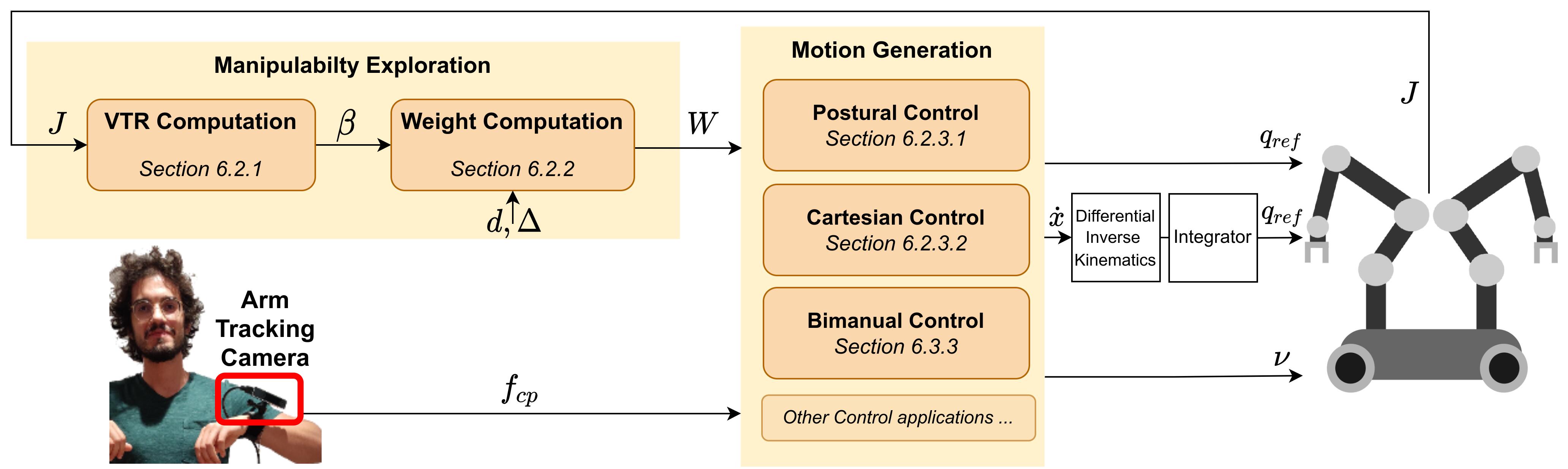}
	\caption[TPO: manipulability-aware shared locomanipulation scheme]{Logic scheme of the manipulability-aware shared locomanipulation interface. On the left side, from the current robot Jacobian $\boldsymbol{J}$, a weight $\boldsymbol{W}$ is computed from the \acrshort{vtr} $\beta$. This weight can be exploited by different control laws, which, from the operator command $\boldsymbol{f}_{\mathit{cp}}$, can generate motions for the arms $\boldsymbol{q}_{\mathit{ref}}$ and the mobile base $\boldsymbol{\nu}$.}
	\label{fig:TPO2generalscheme}
\end{figure}
 
With the TelePhysicalOperation interface presented in Chapter~\ref{chap:TPO}, the operator can choose to control the arm or the base of the robot separately with one of his/her arms; in alternative he/she can control both at the same time by generating two virtual forces with both his/her arms. It is now presented a new possibility with the employment of the manipulability-aware shared locomanipulation interface, schematically depicted in \figurename{}~\ref{fig:TPO2generalscheme}. This method has been integrated into the TelePhysicalOperation framework, hence, in what follows, we will consider for clarity \acrshort{tpo} virtual forces $\boldsymbol{f}_{\mathit{cp}}$ as the operator's command. However, as it will be evident, methods are general enough to be employed with any kind of teleoperation interface and input, like a Cartesian end-effector reference velocity.

\begin{figure}[H]
	\centering
	\includegraphics[width=0.3\linewidth]{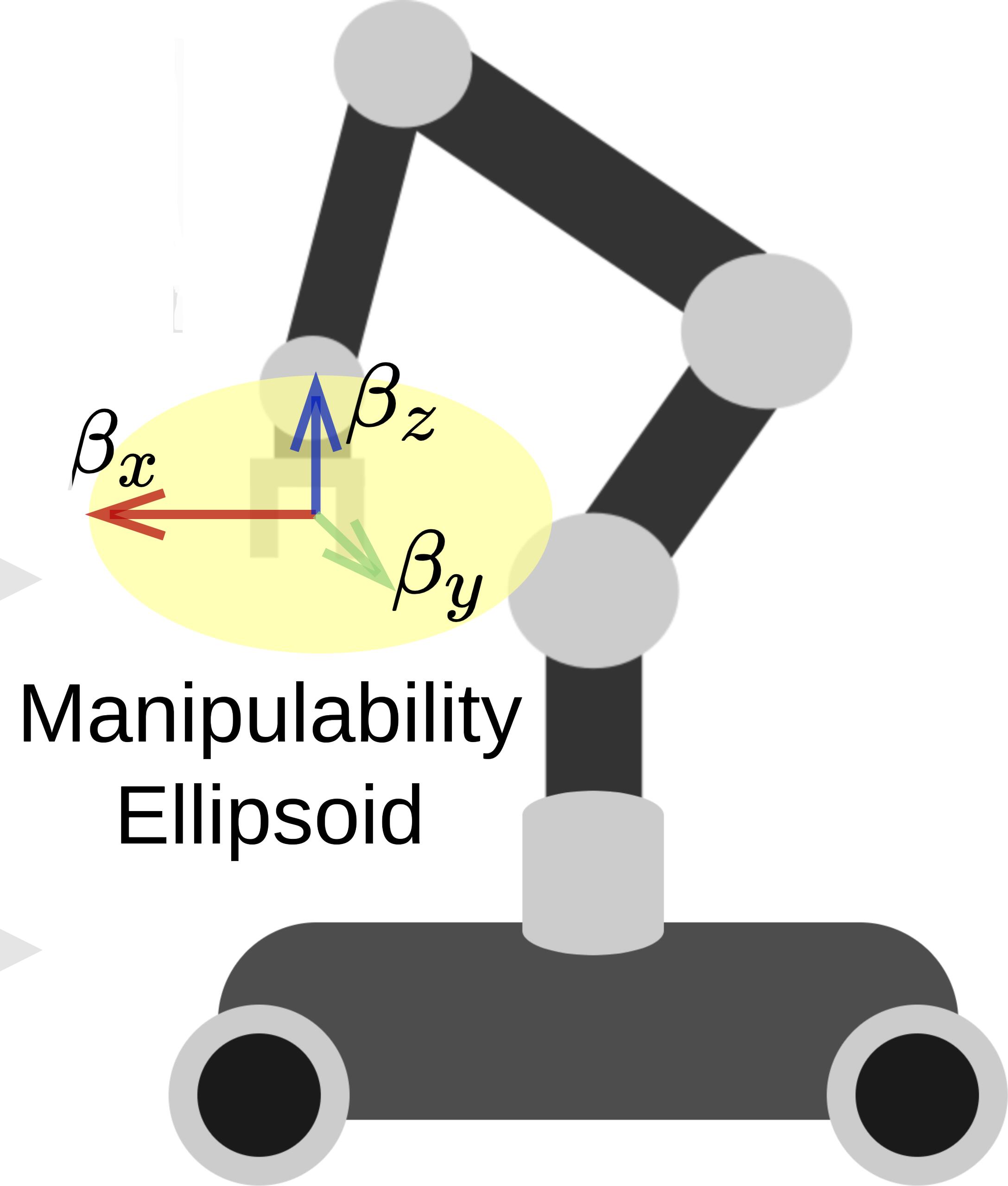}
	\caption[Manipulability ellipsoid and its axes (\acrshort{vtr})]{The end-effector manipulability ellipsoid with its three axes, called the \acrfull{vtr}.}
	\label{fig:rvizmanipsingle}
\end{figure}

Based on the manipulability measure of robot arm end-effector, the virtual force applied by the operator is used to generate motions for both the arm and the mobile base, in a shared control fashion.
The method explores the linear manipulability ellipsoid and its axes in the three principal Cartesian directions $\beta_x,\beta_y,\beta_z$, called \acrfull{vtr}, represented in \figurename{}~\ref{fig:rvizmanipsingle}.

When the end-effector reaches a postural configuration in which one (or more) of the axis' length of the manipulability ellipsoid is under a threshold, the component (or components) of the virtual force along this axis (or axes) is scaled down. At the same time, mobile base velocities are generated in this direction (or directions).
The result is that, in any case, the user can operate the end-effector to reach the desired target, with the mobile base starting to contribute to the motion only in the direction of the low manipulability and only when necessary to ensure that the manipulability of the end-effector does not decrease beyond a defined threshold.
Consequently, the operator's workload is lowered since he/she has not to command mobile base motions by switching the control point.

It is important to notice that different thresholds for the three axes can be set. This flexibility allows choosing the directions in which it is necessary to have the arm more dexterous, and accordingly regulating the mobile base motions in these specific directions.
The proposed approach can not generate movements in unwanted directions. The manipulability measure is used to scale the user's inputs, so the directions will be always under the control of the operator. In other words, the arm does not move autonomously in a direction just to augment its manipulability.

For simplicity, rotational motions of the end-effector are neglected. Hence, the manipulability measure is referred to the linear motions of the end-effector, and the generated scaled arm velocities and mobile base velocities are always considered linear velocities.

\noindent In summary this method presents the following features and contributions:

\begin{itemize}
	\item It generates blended motions of the mobile base and the manipulator from the end-effector reference inputs according to the manipulability condition. This prevents the robotic arm to reach configurations with low manipulability levels at the end-effector while the robotic arm is teleoperated. 
	\item Even when the arm motions are scaled down to preserve the manipulability, the operator can still guide the end-effector toward the desired goal because the mobile base follows the imposed directions. 
	\item The mobile base is never used if it is not necessary, i.e.\ if the arm remains in a region of good manipulability.
	\item The linear manipulability ellipsoid investigated by this work is considered in the three different Cartesian directions, i.e., $\hat{x}$, $\hat{y}$, $\hat{z}$. Consequently, different thresholds can be set for each direction, determining when the mobile base velocities start to be generated. This allows the selection of particular directions in which the manipulator must maintain higher manipulability, resulting to more frequent activation of the base along these directions.
\end{itemize}
\noindent In what follows, the mathematical details of the interface will be presented.

\subsection{Manipulability's Virtual Transmission Ratio (VTR)}
\begin{figure}[H]
	\centering
	\includegraphics[width=\linewidth]{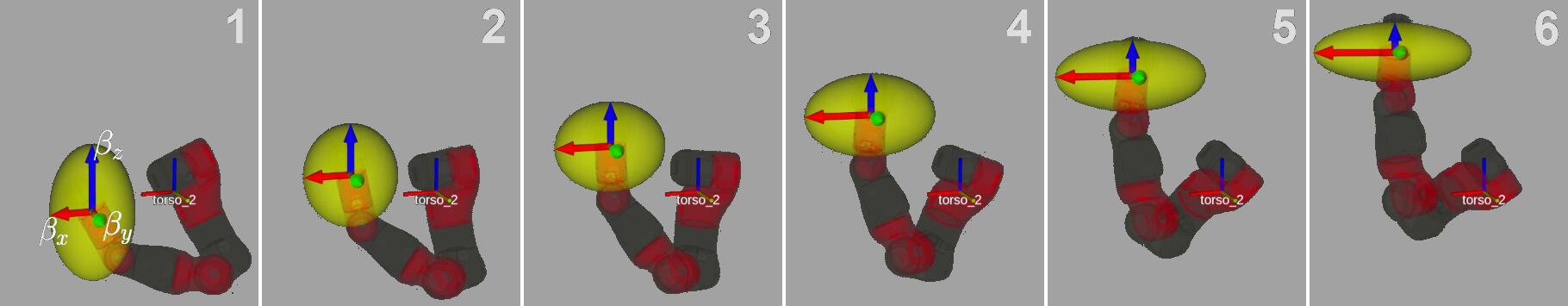}
	\caption[Example of changes in the manipulability measure for the CENTAURO arm]{The \acrshort{vtr} along the principal directions $\hat{x}, \hat{y}, \hat{z}$ is changing while the CENTAURO arm is moving upward. The red, green and blue arrows are the axes of the manipulability ellipsoid (in yellow). They represent the \acrshort{vtr} along the principal directions. The \textit{torso\_2} frame is shown as a reference.}
	\label{fig:tpo2:rvizmanip}
\end{figure}

There are many measures to evaluate the robot manipulability and dexterity~\cite{Patel2015}. In this work, the definition of the (linear) manipulability ellipsoid from~\cite{Yoshi1985} is considered:

\begin{equation}
	\boldsymbol{u}^T ~ (\boldsymbol{J}\boldsymbol{J}^T)^{-1} ~ \boldsymbol{u} \leq 1\\	
\end{equation}
where $\boldsymbol{u} \in \mathbb{R}^{3\times 1}$ is a generic vector; $\boldsymbol{J} \in \mathbb{R}^{3\times N}$ is the linear Jacobian matrix of the end-effector, i.e., the matrix such that its product with the derivative of the N-joints configuration vector results in the Cartesian linear velocity of the end-effector. The geometric space represented by this formula is the yellow ellipsoid in \figurename{}~\ref{fig:tpo2:rvizmanip}.

\noindent The manipulability ellipsoid is strictly related to the arm singularities and to the \acrshort{vtr} $\beta$:
\begin{equation}
	\beta = (\boldsymbol{u}^T ~ (\boldsymbol{J}\boldsymbol{J}^T)^{-1} ~ \boldsymbol{u})^{-\frac{1}{2}}\\
\end{equation}
Geometrically, the resulting scalar $\beta$ is the distance along the vector $\boldsymbol{u}$ between the center of the ellipsoid and the surface of the ellipsoid~\cite{Chiu1988}.
If the manipulator is seen as a mechanical transformer with joints velocity as input and Cartesian velocity as output, this ratio describes how \enquote{much} joints velocity is necessary to achieve a desired Cartesian velocity along $\boldsymbol{u}$~\cite{Chiu1988}. The thinner the ellipsoid is in a certain direction $\boldsymbol{u}$, the lower the \acrshort{vtr} is in this direction. This means that moving the end-effector in this direction not only would require a lot of manipulator effort, but also will make the manipulator always closer to a kinematic singularity; both are cases that are not desirable. 

In this work the idea is to distribute the end-effector motion between the arm and the mobile base motions according to the \acrshort{vtr} level. When the command for the end-effector (as received by the human operator) has a Cartesian direction in which the \acrshort{vtr} of the robotic arm is under a certain threshold, the motion is \textit{gradually} distributed to the mobile base maintaining the \acrshort{vtr} of the end-effector above the defined threshold.
Without loss of generality, the three principal axes $\hat{x}, \hat{y}, \hat{z}$ are considered as the directions of interest, thus the \acrshort{vtr} along them is computed:

\begin{equation}\label{eq:beta_xyz}
	\begin{gathered}
		\beta_x = ([1 ~ 0 ~ 0] ~ (\boldsymbol{J}\boldsymbol{J}^T)^{-1} ~ [1 ~ 0 ~ 0]^T)^{-\frac{1}{2}}\\
		\beta_y = ([0 ~ 1 ~ 0] ~ (\boldsymbol{J}\boldsymbol{J}^T)^{-1} ~ [0 ~ 1 ~ 0]^T)^{-\frac{1}{2}}\\
		\beta_z = ([0 ~ 0 ~ 1] ~ (\boldsymbol{J}\boldsymbol{J}^T)^{-1} ~ [0 ~ 0 ~ 1]^T)^{-\frac{1}{2}}\\
	\end{gathered}
\end{equation}

\noindent These equations can be rewritten as:

\begin{equation}\label{eq:beta_i}
		\beta_i = \big ([(\boldsymbol{J}\boldsymbol{J}^T)^{-1}]_{i,i}\big )^{-\frac{1}{2}}
			\qquad \text{for } i=x,y,z
\end{equation}
where $[(\boldsymbol{J}\boldsymbol{J}^T)^{-1}]_{i,i} \text{ for } i=x,y,z$ indicates the first, second, and third element of the diagonal of $(\boldsymbol{J}\boldsymbol{J}^T)^{-1}$.
\figurename{}~\ref{fig:tpo2:rvizmanip} shows an example of the VTR along the selected three directions, varying accordingly to the CENTAURO arm movements. For example, it can be observed how the $\beta_z$ value (represented by the blue arrow) is decreasing, indicating the difficulty of the end-effector to move along this direction in specific postures.

\subsection{VTR-based Weight Computation}
\begin{figure}[H]
	\centering
	\includegraphics[width=.6\linewidth]{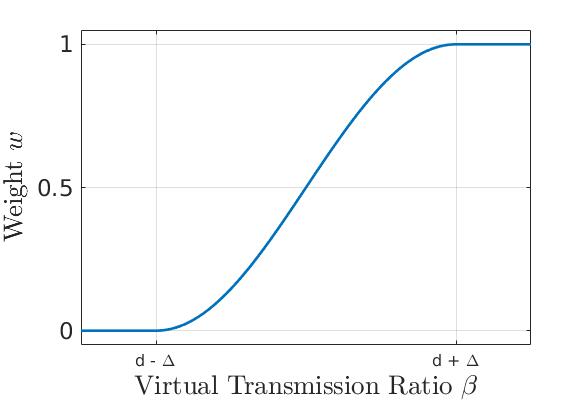}
	\caption{The sigmoid function of \eqref{eq:tpo2:weight} used to compute the weight}
	\label{fig:tpo2:manipSigmo}
\end{figure}

Having derived the \acrshort{vtr} $\beta_i ~ \text{for $i = x, y, z$}$ along the three principal axes, a weight $w_i \in [0,1]$ is computed.
\begin{equation}\label{eq:tpo2:weight}
	w_i = 
	\begin{cases}
		1 & \text{if $\beta_i \geq d_i + \Delta_i$}  \\
		\xi(\beta_i) & \text{if $ d_i - \Delta_i < \beta_i < d_i + \Delta_i$}  \\
		0 & \text{if $\beta_i \leq d_i - \Delta_i$}
	\end{cases}
\end{equation}
where $d_i + \Delta_i$ is an upper threshold over which the \acrshort{vtr} is said to be sufficiently high; $d_i - \Delta_i$ is a lower threshold under which the \acrshort{vtr} is too low; $\xi(\beta_i)$ is a sigmoid function introduced to smooth the transition. \eqref{eq:tpo2:weight} is represented in \figurename{}~\ref{fig:tpo2:manipSigmo}. 
For the sake of notation, we will refer in the following paragraphs to the diagonal matrix of weights $\boldsymbol{W} \in \mathbb{R}^{3\times3}$:
\begin{equation}
	\boldsymbol{W} =
	\begin{pmatrix}
		w_x & 0 & 0 \\
		0 & w_y & 0 \\
		0 & 0  & w_z
	\end{pmatrix}
\end{equation}
and its dual $\boldsymbol{I} - \boldsymbol{W}$, where $\boldsymbol{I} \in \mathbb{R}^{3\times3}$ is the identity matrix.

It is important to notice that the numerical values assumed by the \acrshort{vtr} depend on the specific kinematic characteristic of the manipulator in use. Therefore, the thresholds $d_i +\Delta_i$ should be tuned accordingly to the robot being utilized. To have more robot-independent parameters, it is possible to explore the utilization of a normalized version of the \acrshort{vtr}, by scaling the values accordingly to the maximum value achievable. This procedure is related to the concept of \textit{normalized manipulability}~\cite{Patel2015}, since the \acrshort{vtr} $\beta$ employed in this work is based on the classical manipulability~\cite{Yoshi1985}.

\noindent The computed weights can be exploited in different ways as explained in the next sections.

\subsection{VTR-based Weighted Motion Generation}\label{sec:tpo2:app}

\begin{figure}[H]
	\centering
	\includegraphics[width=0.9\linewidth]{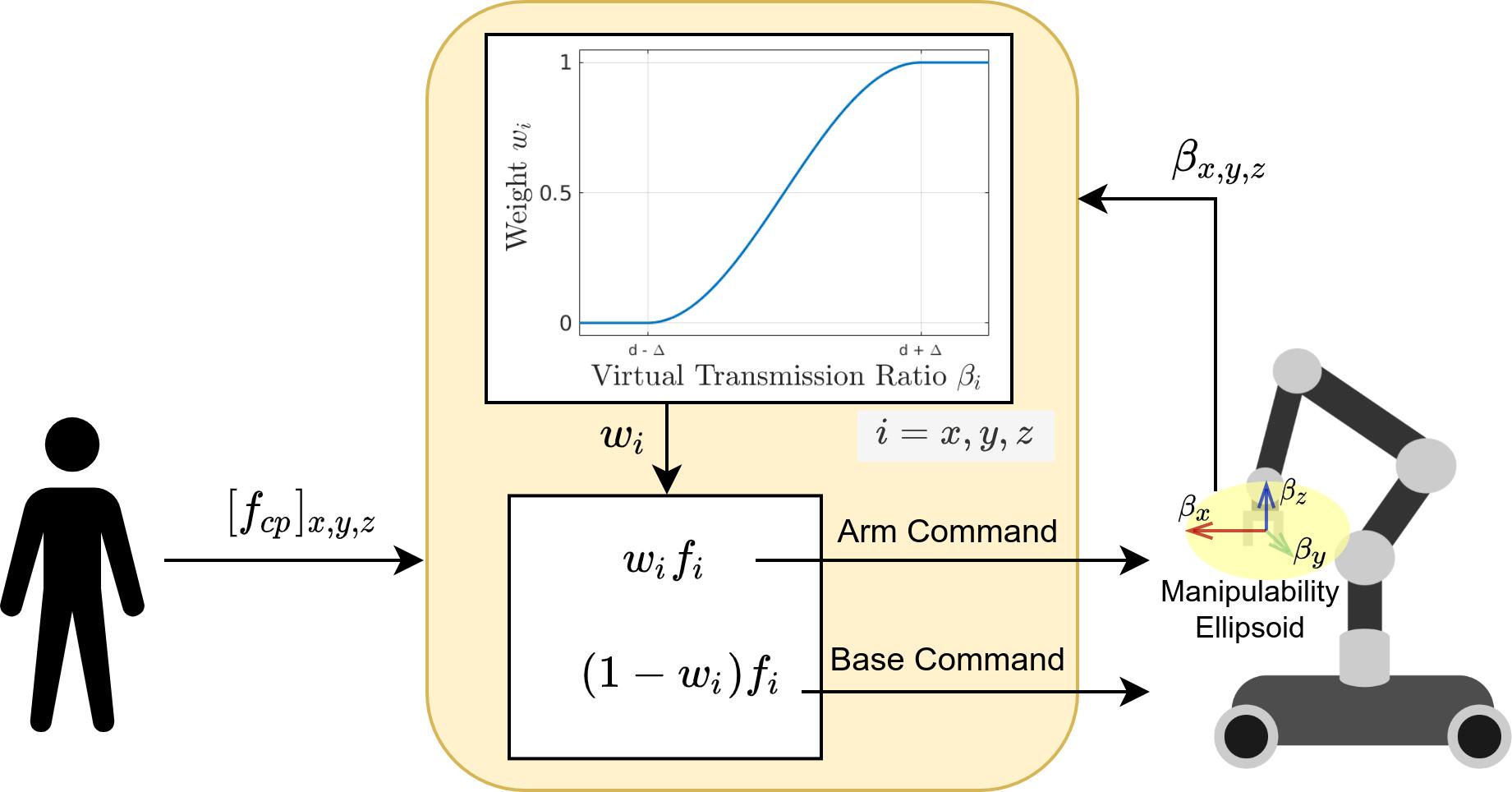}
	\caption[VTR-based weighted motion generation]{Scheme representing the VTR-based weighted motion generation. On the left, the user's input (here shown as a \acrshort{tpo} virtual force $[\boldsymbol{f}_{\mathit{cp}}]_{x,y,z}$) is decomposed into the three linear Cartesian components. On the right, from the VTR $\beta_i$, a weight $w_i$ is computed with a sigmoid function. The components of the user's input are then scaled according to the weight, generating both arm and mobile base motions.}
	\label{fig:manipsimpleschemev2}
\end{figure}

The purpose of the computed weight $\boldsymbol{W}$ is to adjust the input commands from the operator scaling their linear Cartesian components $x, y, z$ by the weights $w_i \text{ for } i=x,y,z$. 
As schematized in \figurename{}~\ref{fig:manipsimpleschemev2}, for each Cartesian components, when the \acrshort{vtr} is over the threshold $d_i + \Delta_i$, it results $w_i=1$, and for the component $i$ only arm commands are generated. 
When the \acrshort{vtr} is below the threshold $d_i - \Delta_i$, it results $w_i=0$, and along the direction $i$ only mobile base velocities are commanded.
When the \acrshort{vtr} is between the thresholds, both arms command and mobile base commands are generated, scaling the user's input according to the sigmoid function $\xi(\beta_i)$.
 
Depending on the utilized control method (\textit{Motion Generation} block of \figurename{}~\ref{fig:TPO2generalscheme}), the arm commands may refer to different robot motions as detailed in the next sections.

\subsubsection{VTR-based Weighted Postural Motion Generation}\label{sec:tpo2:apptorque}

Let's consider the control law in \eqref{eq:tpo:control} presented in the Chapter~\ref{chap:TPO} to teleoperate an N-joints manipulator, reported here for clarity:

\begin{equation} \label{eq:tpo2:control}
	\boldsymbol{\ddot{q}}_{\mathit{ref}}(t) = \boldsymbol{M}_{\mathit{tpo}}^{-1} \big ( \boldsymbol{K}_{\mathit{tpo}} (\boldsymbol{q}_{\mathit{eq}} - \boldsymbol{q}(t)) - \boldsymbol{D}_{\mathit{tpo}} \boldsymbol{\dot{q}}_{\mathit{ref}}(t-1)+\boldsymbol{\tau}_{\mathit{cp}} \big)
\end{equation}
where $\boldsymbol{q}_{\mathit{ref}}(t) \in \mathbb{R}^{N\times 1}$ is the joint position reference vector; $\boldsymbol{M}_{\mathit{tpo}}, \boldsymbol{K}_{\mathit{tpo}}, \boldsymbol{D}_{\mathit{tpo}} \in \mathbb{R}^{N\times N}$ are diagonal matrices of the mass, stiffness, and damping parameters of the joint mass-spring-damper model;  $\boldsymbol{q} , \boldsymbol{q}_{\mathit{eq}} \in \mathbb{R}^{N\times 1}$ are the current position of the joints and the equilibrium set point where a stiffness greater than zero will drag the joints; $\boldsymbol{\ddot{q}}_{\mathit{ref}}$ is integrated twice to obtain the joint command $\boldsymbol{q}_{\mathit{ref}}$.

In this work the approach explained in the Chapter~\ref{chap:TPO} is extended by deriving the torques $\boldsymbol{\tau}_{\mathit{cp}}$ adding in the formula \eqref{eq:tpo:taocp} the weight $\boldsymbol{W}$ to scale the virtual force $\boldsymbol{f}_{\mathit{cp}}$:

\begin{equation}\label{eq:tpo2:taoWTPO}
	\boldsymbol{\tau}_{\mathit{cp}} = \boldsymbol{J}_{\mathit{cp}}^T ~ \boldsymbol{W} \boldsymbol{f}_{\mathit{cp}}
\end{equation}
where $\boldsymbol{f}_{\mathit{cp}} \in \mathbb{R}^{3\times1}$ is the virtual force computed from the displacement of the operator arm gathered from a motion tracking device \eqref{eq:tpo:f_cp}.

Dually, the reference velocity command $\boldsymbol{\nu} \in \mathbb{R}^{3\times1}$ for the mobile base is computed as:
\begin{equation}\label{eq:tpo2:nuWTPO}
	\boldsymbol{\nu} = \boldsymbol{K}_{\nu} (\boldsymbol{I} - \boldsymbol{W}) \boldsymbol{f}_{\mathit{cp}}
\end{equation}
where $\boldsymbol{K}_{\nu} \in \mathbb{R}^{3\times3}$ is a diagonal matrix of gains to maintain the consistency between the different physical quantities. 
In practice, when the \acrshort{vtr} in one or more of the three principal Cartesian directions is lowering down toward the threshold, the corresponding elements in $\boldsymbol{W}$ change from $1$ to $0$. This would result in the virtual force scaled down to compute the torque \eqref{eq:tpo2:taoWTPO}, and scaled up to compute the base velocity $\boldsymbol{\nu}$ \eqref{eq:tpo2:nuWTPO}.
Hence, the arm motion will be limited along the direction of low \acrshort{vtr}, and the mobile base motion will compensate with a proper velocity. If in some directions the \acrshort{vtr} is above the threshold, the respective component of the manipulator reference is not modified, and the base will not move in these directions.

Note that the weight $\boldsymbol{W}$ will not affect the returning elastic component of the joint mass-spring-damper model of \eqref{eq:tpo2:control}. Therefore, with a proper setting of the diagonal elements of $\boldsymbol{K}_{\mathit{cp}}$, the arm can be further helped to recover from a posture where the \acrshort{vtr} is low thanks to the returning elastic element that brings the arm back toward the $\boldsymbol{q}_{\mathit{eq}}$ equilibrium, which should be set accordingly to result in a posture of the arm with good manipulability.

\subsubsection{VTR-based Weighted Cartesian Motion Generation}\label{sec:tpo2:appVel}
Consider now the more general case, in which a control law generates Cartesian velocities $\boldsymbol{\dot{x}}$ for the manipulator, instead of joint position references as in the postural motion generation of Section~\ref{sec:tpo2:apptorque}. 
For example, the TelePhysicalOperation architecture provides such generation with the formula \eqref{eq:x_cp}, explained in Section~\ref{sec:tpo:vel}.

These velocities $\boldsymbol{\dot{x}}$ can be scaled to generate arm and mobile base commands according to the \acrshort{vtr} similarly as the case of the postural control:
\begin{equation}\label{eq:tpo2:xDotWTPO}
	\begin{gathered}
		\boldsymbol{\dot{x}}^{*} = \boldsymbol{W} \boldsymbol{\dot{x}} \\
		\boldsymbol{\nu} = (\boldsymbol{I} - \boldsymbol{W}) \boldsymbol{\dot{x}}
	\end{gathered}
\end{equation}

Similarly to the case of the postural control, this will result in the robotic arm slowing down towards the directions in which the VTR is below the threshold if the desired end-effector linear Cartesian velocity $\boldsymbol{\dot{x}}$ has some non-null components in these directions. Dually, the velocity of the mobile base will increase towards these directions.

This weighted Cartesian motion generation is employed in combination with the bimanual transporting control law presented in Section~\ref{sec:tpo3:transport}.

\subsection{Visual Feedback Tools}

\begin{figure} [H]
	\centering
	\includegraphics[width=0.24\linewidth]{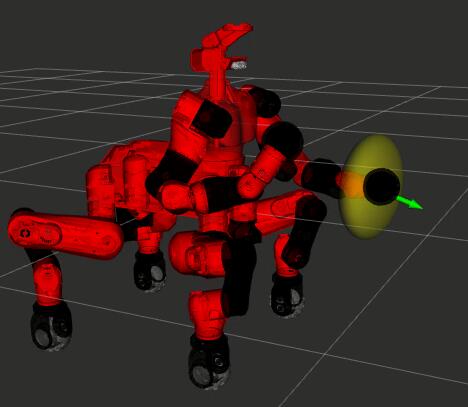}
	\includegraphics[width=0.24\linewidth]{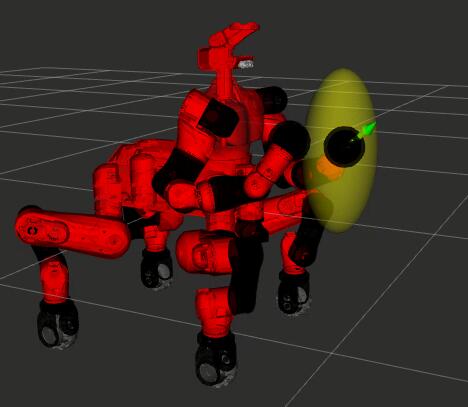}
	\includegraphics[width=0.24\linewidth]{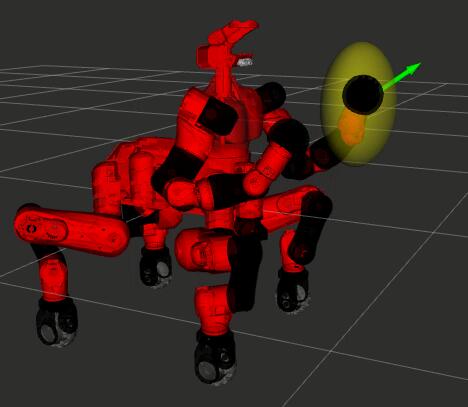}
	\includegraphics[width=0.24\linewidth]{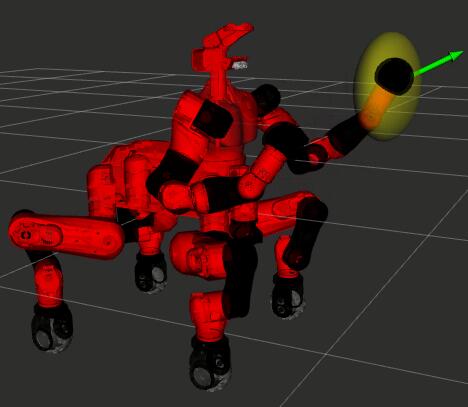}
	\caption[Visual representation of manipulability ellipsoid]{Sequences that show the CENTAURO robot in the RViz visualizer, with arrows representing the \acrshort{tpo} virtual forces applied by the operator, and in yellow the manipulability ellipsoid of the left end-effector.}
	\label{fig:tpo2:rvizManipExp}
\end{figure}

A small addition to the visual feedback information presented in Section~\ref{sec:tpo:visual} are shown in \figurename{}~\ref{fig:tpo2:rvizManipExp}. Since the exploitation of the manipulability level, there is the possibility to add to the robot model in RViz also an ellipsoid representing the manipulability, dynamically modifying its shape according to the joints positions of the arm of the robot. Even if the manipulability measure is taken into consideration by the autonomy features developed, having a visual guide about its state can be useful to understand what it is happening to the robot.

\section{Bimanual Grasping and Transportation of Objects of Unknown Mass}\label{sec:tpo3:methods}

\begin{figure}[H]
	\centering
	\includegraphics[width=1\linewidth]{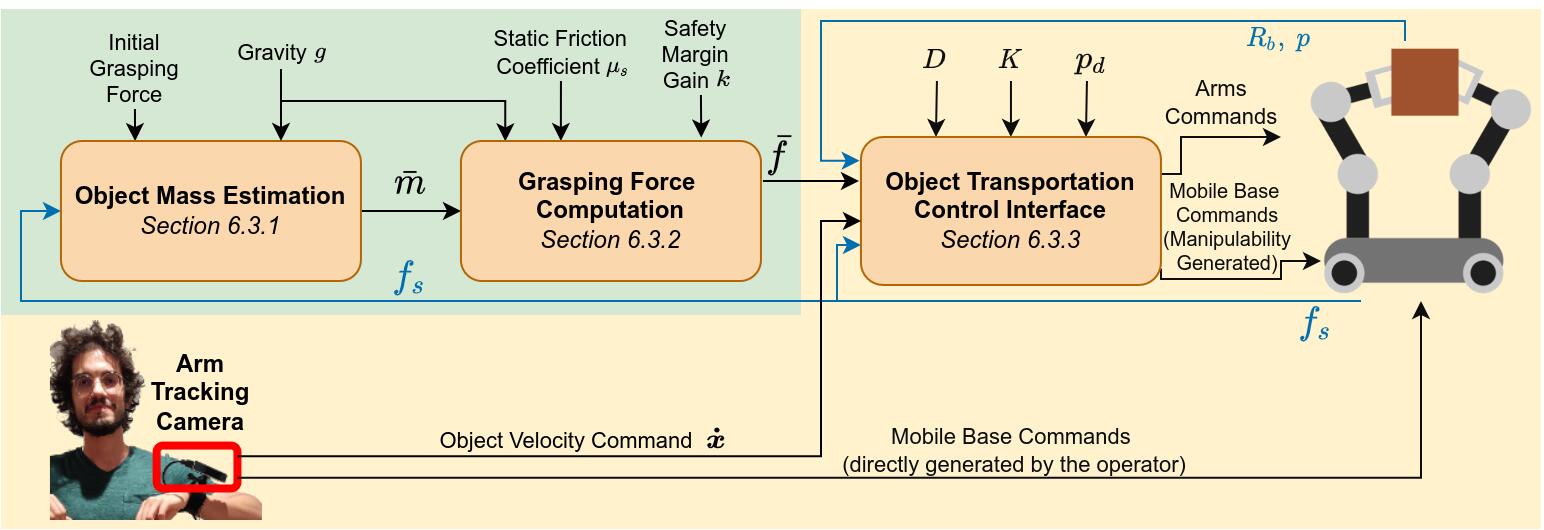}
	\caption[TPO: bimanual grasping and transportation framework scheme]{Logic scheme of bimanual grasping and transportation framework. In an initial phase (green area), the object mass $\bar{m}$ and the necessary grasping force $\boldsymbol{\bar{f}}$ are calculated. During the object transportation (yellow area), the operator can command just object velocities $\boldsymbol{\dot{x}}$ while the robot generates motions considering the grasping constraint and the manipulability level of the arms, to transport safely the load toward the given directions. Blue arrows distinguish the feedback signals.}
	\label{fig:tpo3:generalscheme}
\end{figure}

As mentioned in the introductory Section~\ref{sec:tpo3:intro}, bimanual transportation of an object can be a difficult task to perform by the operator alone without any help from the robot to coordinate the motions of the arms as necessary to transport the object. 
This section presents how this challenge has been faced and how the methods are integrated in the \acrlong{tpo} architecture.
In particular, a shared control method is developed to allow the robot to firstly autonomously grasp the object and secondly transport it according to user inputs by regulating the grasping forces based to the estimated mass of the object (\figurename{}~\ref{fig:tpo3:generalscheme}).

With this method, the user comfortably considers directly the grasped object as a reference point for the commanded velocities, instead of commanding directly the two robot arms. In particular, the user inputs are desired object velocities that can be generated from the TelePhysicalOperation architecture, in particular with the Cartesian motion generation as explained in Section~\ref{sec:tpo:vel}.
The interface does not necessitate of force/torque sensors on the robot arms, since the force sensed at the end-effectors is estimated from the actuator torques with a force estimation module based on the residual method~\cite{Haddadin2017}.

It is assumed that the nature of the object allows the robot to grasp and transport it by establishing contacts and applying forces on two opposing sides and by considering only linear forces (i.e., no torques).
At the user request, the robot is able to autonomously reach the object to be lifted. The pose of the object is detected and tracked by the robot vision system based on an \textit{ArUco} marker\footnote{\href{http://wiki.ros.org/aruco_detect}{http://wiki.ros.org/aruco\_detect}} placed on the object. From the object pose, two end-effector goals are established at the box sides, and Cartesian velocities for each arm are generated. While the robot is moving toward the object, based on the \acrshort{vtr} of each end-effector, the Cartesian velocities are balanced between each arm and the mobile base using the VTR-based weighed motion generation presented previously in Section~\ref{sec:tpo2:appVel}. The final velocity commanded to the mobile base is the average of the two $\boldsymbol{\nu}$ computed from the weight of each arm. Once the arms have reached the object, the following phase begins.

The grasping and transportation interface functionality can be divided in three modules, as depicted in \figurename{}~\ref{fig:tpo3:generalscheme}:

\begin{itemize}
 \item \textbf{Object Mass Estimation}. After autonomously reaching the object with the two arms, the robot grasps it by applying an initial level of grasping force and lifts it up. In this phase the initial level of grasping force must be put sufficiently high by the operator to lift the object firmly. Once the object is lifted from its support, its mass is estimated on the basis of the sensed forces at the end-effectors, as detailed in Section~\ref{sec:tpo3:massEst}.
 
 \item \textbf{Grasping Force Computation}. Based on the estimated mass, a required grasping force is computed. 
 This information is used to modulate the arm motions as required to maintain the grasp on the object while transporting it as commanded by the operator, without applying extended forces, which could damage the object or the robot itself.
 The procedure resembles what humans do in front of an object of unknown mass, i.e., they modulate the force necessary to keep the object after its mass is estimated during the lifting phase.
 The grasping force computations are detailed in the Section~\ref{sec:tpo3:forceOptComp}.
 
 \item \textbf{Object Transportation Control Interface}. During the object transportation (yellow area \figurename{}~\ref{fig:tpo3:generalscheme}), the operator commands just object velocities while the robot generates arms and mobile base motions considering the grasping constraint and the manipulability level of the arms (Section~\ref{sec:tpo2:appVel}), to transport safely the load toward the desired directions. Please note that, according to the scheme, the user can also directly command mobile base velocity for large displacements of the whole robot. This interface is detailed in Section~\ref{sec:tpo3:transport}.
 
\end{itemize}

\subsection{Object Mass Estimation}\label{sec:tpo3:massEst}

\begin{figure}[H]
	\centering
	\includegraphics[width=0.8\linewidth]{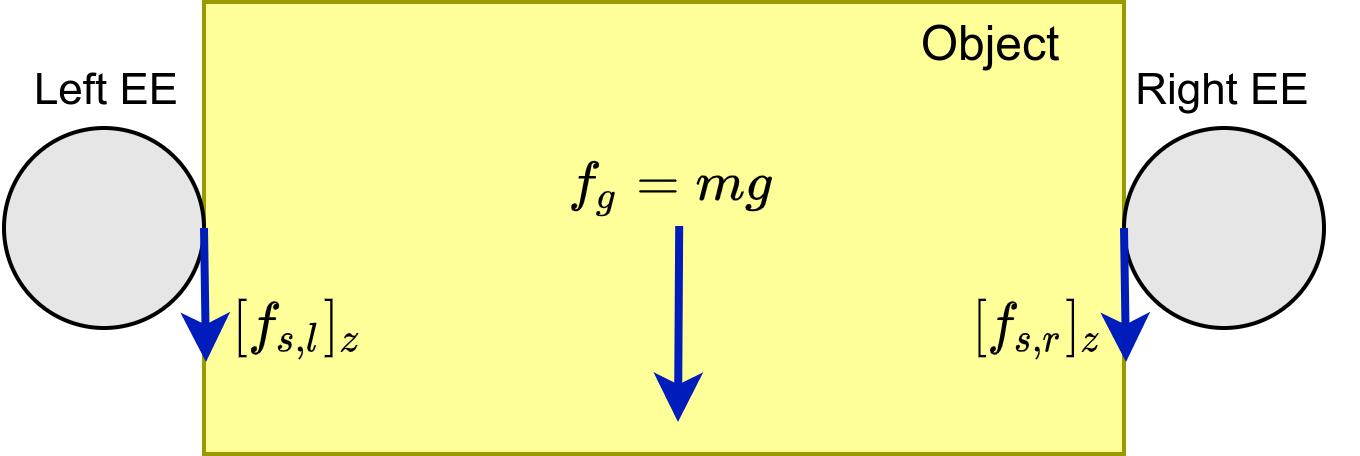}
	\caption{Forces involved in the mass estimation formulas of Section~\ref{sec:tpo3:massEst}.}
	\label{fig:tpo3:boxMassEstimationScheme}
\end{figure}

The proposed bimanual transportation framework involves the estimation of the object mass. The following dynamic equations are based on Coulomb's law of friction exploited similarly in previous works~\cite{Francomano2013, Morita2018}. 

A scheme of the forces involved in the formulas is depicted in \figurename{}~\ref{fig:tpo3:boxMassEstimationScheme}.
Taking into account the object frame with the $\hat{z}$ axis perpendicular to the ground, it is assumed that the contact points where the robot end-effectors apply forces to the object are at the same height on the $\hat{z}$ axis (in blue in the scheme of \figurename{}~\ref{fig:tpo3:boxMassEstimationScheme}). Furthermore, it is considered that a sufficiently high level of initial grasping force is applied in the first phase to permit to lift firmly the object. In such conditions, after the object has been lifted from its support, in absence of slip and external disturbances, the gravitational force $f_g$ is equal to the sum of the two end-effectors sensed forces along the $\hat{z}$ direction:

\begin{equation}
	f_g = [\boldsymbol{f}_{s,l}]_z + [\boldsymbol{f}_{s,r}]_z
\end{equation}
where $[\boldsymbol{f}_{s,l}]_z$ and $[\boldsymbol{f}_{s,r}]_z$ are the $\hat{z}$ axis component of the sensed forces of the left and right end-effectors. 
From Newton's second law applied to gravity, $f_g = m g$, the estimated mass $\bar{m}$ is computed as: 

\begin{equation}\label{eq:mass_est}
	\bar{m} = \dfrac{1}{g} \Big(  [\boldsymbol{f}_{s,l}]_z + [\boldsymbol{f}_{s,r}]_z \Big)
\end{equation}
where $g$ is the gravity acceleration. The mass estimation is performed by averaging the estimated mass values computed from a certain number of sensed force samples, in order to decrease the influence of the noise of the force estimation.

\subsection{Grasping Force Computation}\label{sec:tpo3:forceOptComp}

\begin{figure}[H]
	\centering
	\includegraphics[width=0.8\linewidth]{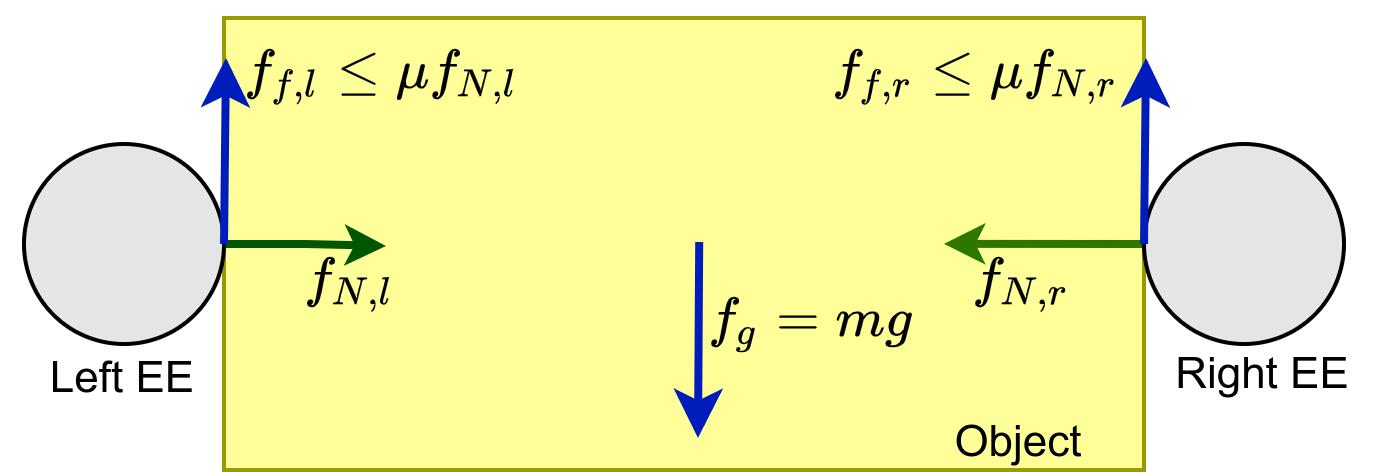}
	\caption{Forces involved in the grasping force computation formulas of Section~\ref{sec:tpo3:forceOptComp}.}
	\label{fig:tpo3:optimalGraspingForceScheme}
\end{figure}

\figurename{}~\ref{fig:tpo3:optimalGraspingForceScheme} shows the relevant quantities involved in the computation of the grasping force.
After estimating the mass of the object, the method computes the required grasping force that permits to maintain a stable grasp by counterbalancing the gravity force $f_g = m g$ with the friction forces at the contact points:

\begin{equation}
	f_{f,l} + f_{f,r} = mg
\end{equation}
where $f_{f,l}$ and $f_{f,r}$ are the Coulomb friction forces generated by the normal grasping forces $f_{N,l}$ and $f_{N,r}$ applied by the two robot end-effectors. Given the Coulomb friction law $f_{f} \leq \mu_s f_N$, with $\mu_s$ the static friction coefficient, it can be obtained what follows:

\begin{equation}
	\begin{gathered}
		\mu_s f_{N,l} + \mu_s f_{N,r} \geq f_{f,l} + f_{f,r} = mg \\
		f_{N,l} + f_{N,r} \geq \mu_s^{-1}mg 
	\end{gathered}
\end{equation}
where it is assumed that the friction coefficient is the same at both contact points. Given the above, the required grasping force $\bar{f}$ for each end-effector can be computed as:

\begin{equation}\label{eq:tpo3:optForce}
	\bar{f} = k ~ \Big(\dfrac{\bar{m} g}{2 \mu_s } \Big)
\end{equation}
where $k > 1$ is a safety margin gain used to take into account errors such as in the measurement of the sensed forces $\boldsymbol{f}_{s}$, and any uncertainties in the static friction coefficient $\mu_s$.
Having computed $\bar{f}$, the robot arms regulate the grasping forces applied by the end-effectors on the object from the initial force level (used during the lifting phase) to the required grasping force level $\bar{f}$.

\subsection{Object Transportation Control Interface}\label{sec:tpo3:transport}

\begin{figure}[H]
	\centering
	\includegraphics[width=0.8\linewidth]{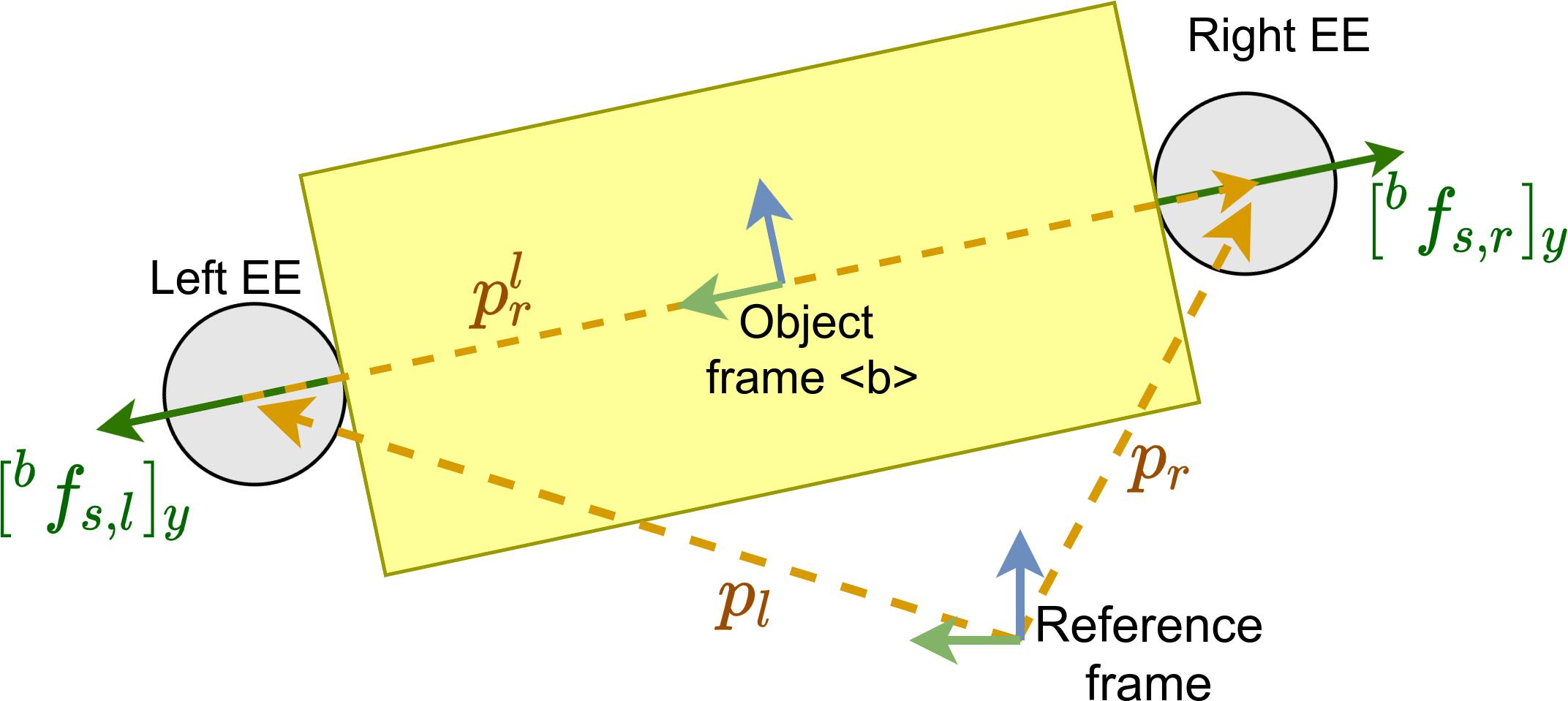}
	\caption[Relevant vectors involved in the object transportation control interface of Section~\ref{sec:tpo3:transport}]{Scheme of the relevant vectors involved in the object transportation control interface formulas of Section~\ref{sec:tpo3:transport}.}
	\label{fig:tpo3:boxControlLawSchme}
\end{figure}

Once the object is grasped and the required grasping force $\bar{f}$ is applied, the operator can control the robot by directly commanding the velocity of the bimanual grasped object while the object is held in an autonomous manner by the proposed interface. As a visual aid, a scheme is depicted in \figurename{}~\ref{fig:tpo3:boxControlLawSchme}.

From \cite{Tarbouriech2019}, which uses the principle of virtual sticks of \cite{Uchiyama1988}, and the cooperative task representation of \cite{Chiacchio1996}, it is computed the Cartesian linear velocity commands $\boldsymbol{\dot{x}}_j \in \mathbb{R}^3, \text{with } j=l,r$ for the left and right end-effectors with the following admittance control formula:

\begin{equation}\label{eq:tpo3:coopLaw}
	\boldsymbol{\dot{x}}_j = \boldsymbol{\dot{x}} + \boldsymbol{D}^{-1} \big (\boldsymbol{R}_{b} (^{b}\boldsymbol{f}_{s,j} - ^{b}\boldsymbol{f}_{d,j}) + \boldsymbol{K} (\boldsymbol{p}_{d,j} - \boldsymbol{p}_{j} )\big )
\end{equation}
where $\boldsymbol{\dot{x}}$ is the object linear velocity command imposed by the operator; $\boldsymbol{D}, \boldsymbol{K} \in \mathbb{R}^{3\times 3}$ are diagonal matrices of damping and stiffness; $\boldsymbol{f}_{s,j}, \boldsymbol{f}_{d,j} \in \mathbb{R}^{3\times 1}$ are the estimated and the desired force at the $j$ end-effector; $\boldsymbol{p}_{j}, \boldsymbol{p}_{d,j} \in \mathbb{R}^{3\times 1}$ are the actual position and the desired position of the $j$ end-effector; $\boldsymbol{R}_{b} \in \mathbb{R}^{3\times 3}$ is the rotation matrix from object frame \mbox{$<\!b\!>$} to the reference frame. The $\hat{y}$ axis of the object frame \mbox{$<\!b\!>$} lies in the line which connects the two end-effectors (\figurename{}~\ref{fig:tpo3:boxControlLawSchme}); hence the rotation matrix $\boldsymbol{R}_{b}$ is necessary because the required grasping force $\bar{f}$ is referred to the frame $<\!b\!>$.

\noindent The desired force $\boldsymbol{f}_{d,j}$ for each end-effector $j$ is set as: 

\begin{equation}\label{eq:tpo3:f_d}
	\begin{gathered}
		^{b}\boldsymbol{f}_{d,l} = 
		\begin{bmatrix}[^{b}\boldsymbol{f}_{s,l}]_x, & \bar{f}, & [^{b}\boldsymbol{f}_{s,l}]_z \end{bmatrix}^T \\
		^{b}\boldsymbol{f}_{d,r} = 
		\begin{bmatrix}[^{b}\boldsymbol{f}_{s,r}]_x, & -\bar{f}, & [^{b}\boldsymbol{f}_{s,r}]_z \end{bmatrix}^T \\
	\end{gathered}
\end{equation} 
where $\bar{f}$ is the required grasping force of \eqref{eq:tpo3:optForce} applied along the direction normal to the contact surfaces of the object. The signs of the $y$ component of the desired forces $^{b}\boldsymbol{f}_{d,l}$ and $^{b}\boldsymbol{f}_{d,r}$ are set accordingly to the chosen reference frame. Instead, the $x$ and $y$ components are set equal to the respective components of the sensed force.

Furthermore, to help in maintaining a stable grasp, a simple leader-follower concept is employed, where one arm (e.g., the right arm $r$) follows the leader one (e.g., the left arm $l$). Therefore, in \eqref{eq:tpo3:coopLaw} it is considered the Cartesian linear position error of the end-effectors, where the desired positions $\boldsymbol{p}_{d,l}$ and $\boldsymbol{p}_{d,r}$ of the left and right end-effectors are: 

\begin{equation}
	\begin{gathered}
		\boldsymbol{p}_{d,l} = \boldsymbol{p}_{l} \\
		\boldsymbol{p}_{d,r} = \boldsymbol{p}_{l} + \boldsymbol{p}_{r}^{l}(t_0)	
	\end{gathered}
\end{equation}
where $\boldsymbol{p}_{l} \text{ and } \boldsymbol{p}_{r}$ are the actual position of the left and right end-effectors; $\boldsymbol{p}_{r}^{l}(t_0)$ is the pose of the right end-effector with respect to the left one at the time $t_0$, which is the instant in which the box is grasped, and the user begins to teleoperate the robot. So, the elastic element $\boldsymbol{K}$ of \eqref{eq:tpo3:coopLaw} induce motions to maintain the right end-effector in position with respect to the left one as it was at the instant $t_0$.

As stated previously, the object velocity command $\boldsymbol{\dot{x}}$ can be generated with any kind of teleoperation interface. In the context of this thesis, it is employed the TelePhysicalOperation Cartesian motion generation explained in Section~\ref{sec:tpo:vel}. 
Furthermore, following the VTR-based weighted Cartesian motion generation of Section~\ref{sec:tpo2:appVel}, the $\boldsymbol{\dot{x}}$ can be pre-processed to compute its scaled version $\boldsymbol{\dot{x}}^*$ with \eqref{eq:tpo2:xDotWTPO}, with the weight computed
from the robot arm in the worst condition of manipulability. Subsequently, $\boldsymbol{\dot{x}}^*$ is used in place of $\boldsymbol{\dot{x}}$ in the control law of \eqref{eq:tpo3:coopLaw}, and mobile velocity commands $\boldsymbol{\nu}$ are generated as well.

\section{Autonomy-enhanced TelePhysicalOperation Experimental Validations}\label{sec:tpo23:exps}

Experimental validations about the methods presented in this chapter have been conducted with the CENTAURO robot, briefly introduced in Section~\ref{sec:intro:centauro}. 

In Section~\ref{sec:tpo2:buttons}, the manipulability-aware shared locomanipulation interface, presented in Section~\ref{sec:tpo2:manipControl}, is validated. 
In Section~\ref{sec:tpo3:estimation}, the mass estimation of a bimanual grasped object (Section~\ref{sec:tpo3:massEst}) is evaluated. 
In Section~\ref{sec:tpo2:box}, the manipulability-aware shared locomanipulation interface and the bimanual grasping and transporting interface of Section~\ref{sec:tpo3:methods} are combined for a bimanual transportation mission. 
A more difficult challenge is presented in Section~\ref{sec:tpo3:box}, where the bimanually grasped object must be placed into a container and the container itself transported to the final location.

In all these experiments, the employment of the haptic interface of Chapter~\ref{chap:TPOH} is not considered as it is not relevant for the validation of the proposed robot autonomy features.

Videos about the experiments of Section~\ref{sec:tpo2:buttons} and Section~\ref{sec:tpo2:box} are available at \href{https://youtu.be/7YqfVn8XvNk}{https://youtu.be/7YqfVn8XvNk}; while the video about the experiment of Section~\ref{sec:tpo3:box} is available at \href{https://youtu.be/7s1n0weq5To}{https://youtu.be/7s1n0weq5To}.

\subsection{Reaching Specific Locations with the End-Effector}\label{sec:tpo2:buttons}

\begin{figure}[H]
	\centering
	\includegraphics[width=0.85\linewidth]{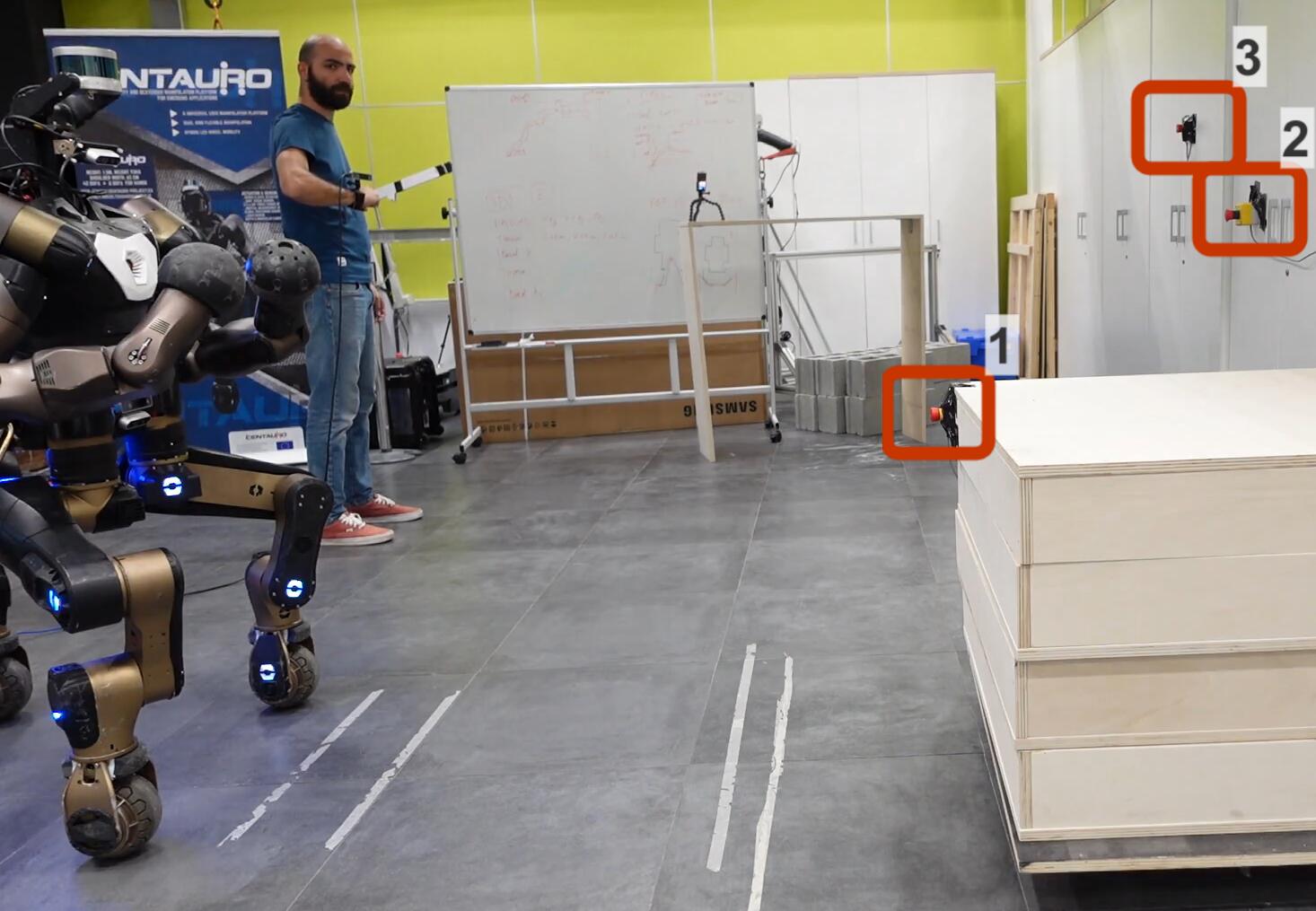}
	\caption[Autonomy enhanced \acrshort{tpo}: reaching locations experiment]{In this experiment the robot is teleoperated with the \acrshort{tpo} interface enhanced with the manipulability-aware shared locomanipulation to reach and press the three buttons situated in the environment.}
	\label{fig:tpo2:ExpThreeBtn}
\end{figure}

This experiment validates the manipulability-aware shared locomanipulation motion generation, presented in Section~\ref{sec:tpo2:manipControl}. It consists in a reaching task where the robot is teleoperated to press with the left end-effector some buttons located in three different positions of the environment (\figurename{}~\ref{fig:tpo2:ExpThreeBtn}).
Given the distances between the three locations and their different heights, reaching them requires a combination of both manipulator and mobile base motions.

The robot is controlled with the \acrshort{tpo} interface, using a single input deriving from the user's right arm motion. This input generates a virtual force applied to the left end-effector of the robot as explained in Section~\ref{sec:tpo:pos}. The user input is restricted to a single arm to resemble a situation where the utilization of two inputs (i.e., both arms), which is anyway possible with the \acrshort{tpo} interface, is not desirable because it would augment the operator's workload.

In such conditions, the task is executed with two different modalities: with and without the manipulability-aware motion generation method. When it is not used, the operator must repeatedly change the control point between the arm and the mobile base since it has not the possibility to control both at the same time. When controlling the mobile base, the virtual force applied generates a Cartesian motion as detailed in Section~\ref{sec:tpo:vel}.
Instead, when the method is used, the operator does not need to change the control point from the left end-effector since the underlying architecture generates both arm and mobile base motions according to the manipulator \acrshort{vtr} level, as explained in Section~\ref{sec:tpo2:apptorque}. 
This permits, as shown later, to decrease the execution time, and also to maintain a good arm dexterity.

\begin{figure}[H]
	\centering
	\includegraphics[width=0.49\linewidth]{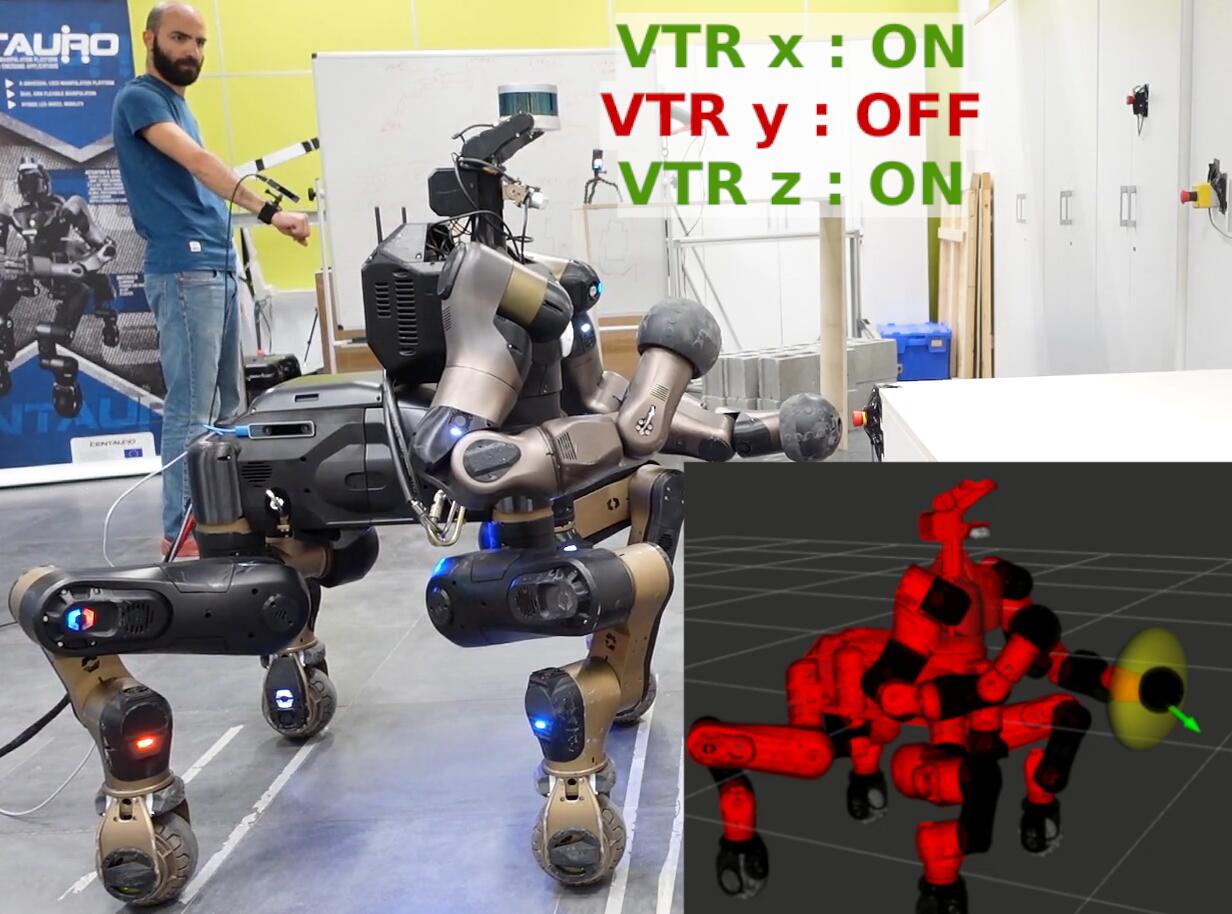}
	\includegraphics[width=0.49\linewidth]{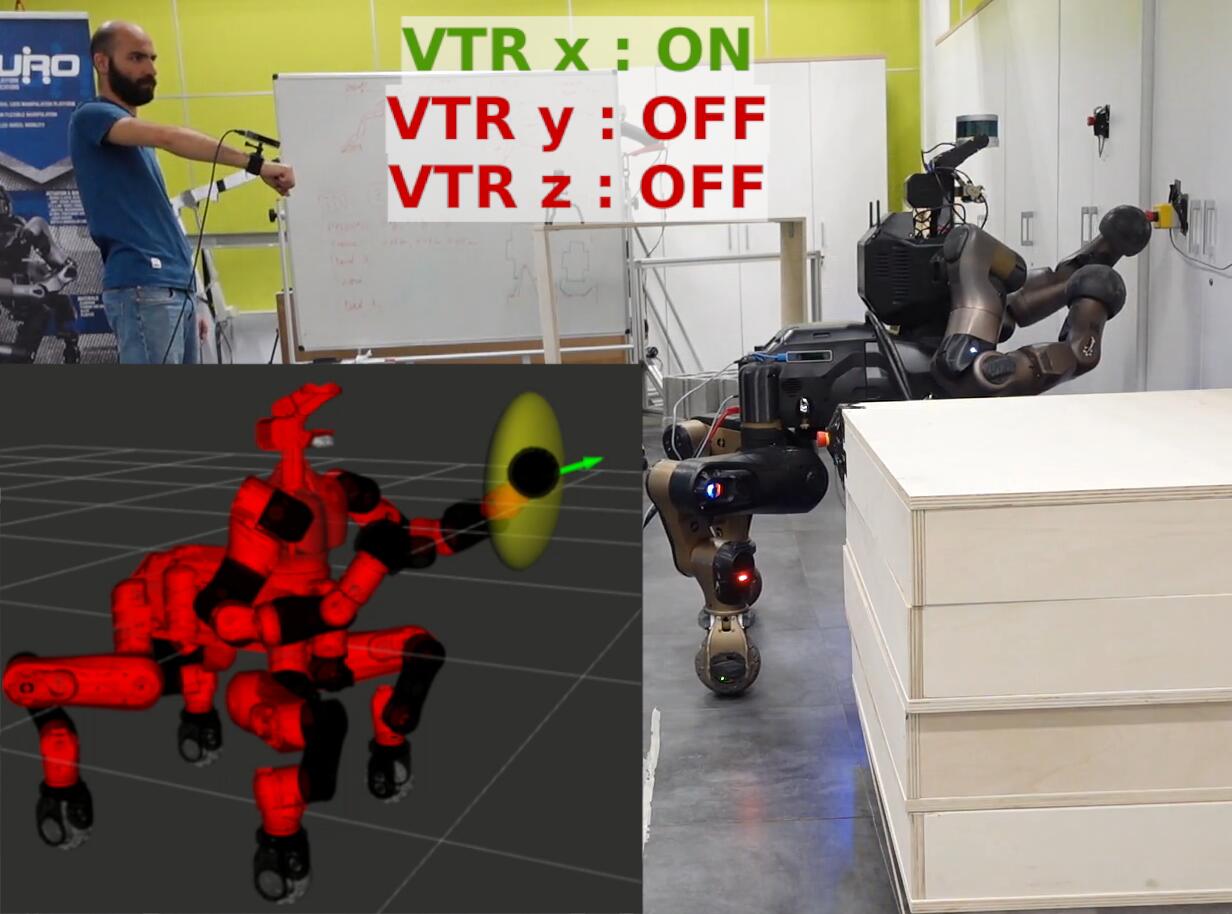}\\
	\includegraphics[width=0.49\linewidth]{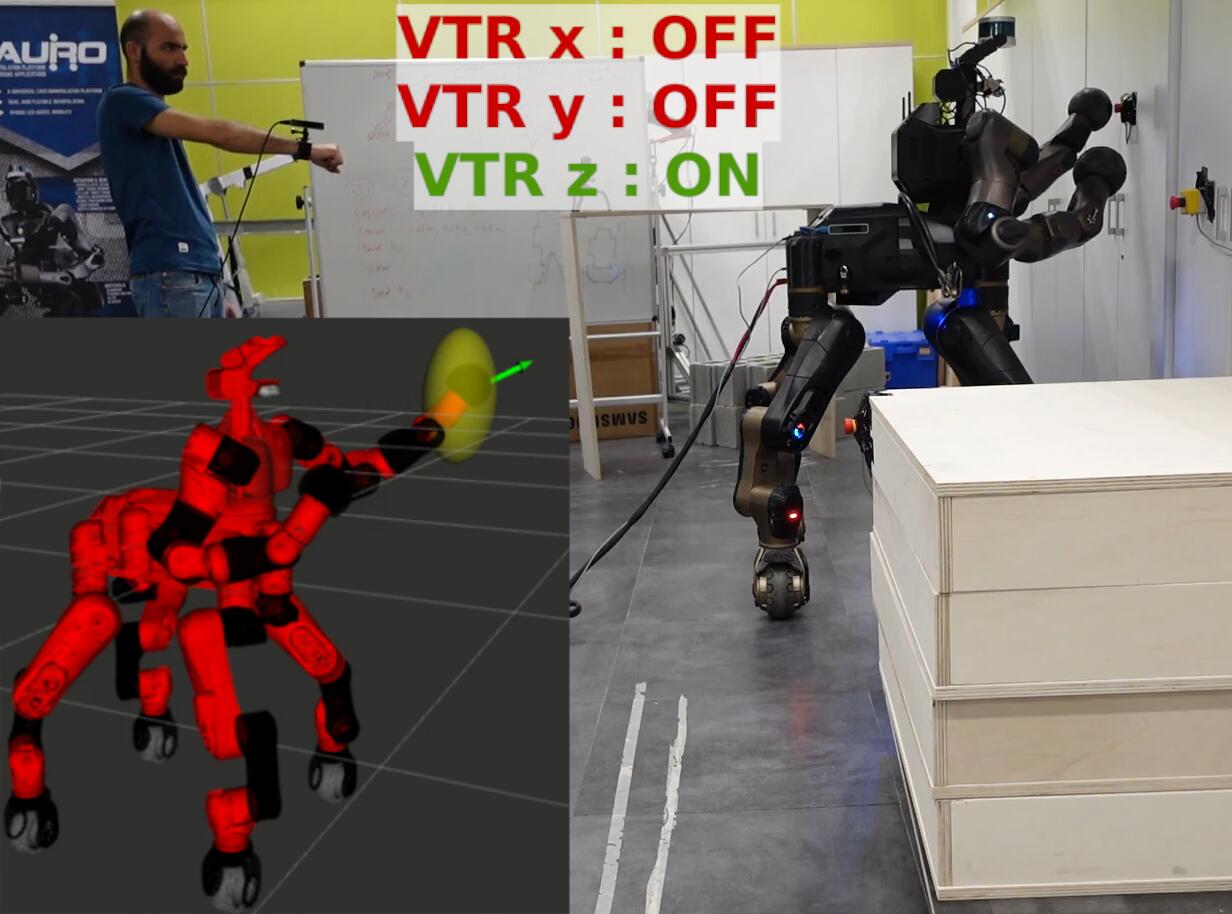}
	\caption[Autonomy enhanced \acrshort{tpo}: reaching locations experiment sequences]{Sequences of the reaching experiment employing the manipulability-aware motion generation. In the legends, it is indicated in which directions the manipulability-aware motion generation (\acrshort{vtr}) is active during the teleoperation. The robot kinematic visualization (\textit{RViz}) shows the input directions commanded by the user (green arrow) and the manipulability ellipsoid of the robot left end-effector (yellow shape).}
	\label{fig:tpo2:ExpThreeBtnDetail}
\end{figure}

Some sequences of the task executed with the manipulability-aware motion generation are depicted in \figurename{}~\ref{fig:tpo2:ExpThreeBtnDetail}. The legends in the images show when the manipulability-aware motion generation is active (\enquote{VTR ON}) in one particular direction. This happens when the \acrshort{vtr} is below the $d + \Delta$ threshold, which causes the robot to generate mobile base motions in that direction. This is particularly useful, for example, to reach the last button put at certain height. The robot increases the pelvis height with a \enquote{squatting up} motion, permitting the end-effector to press the button maintaining a good arm manipulability. 

\begin{figure}[H]
	\centering
	\includegraphics[width=1\linewidth]{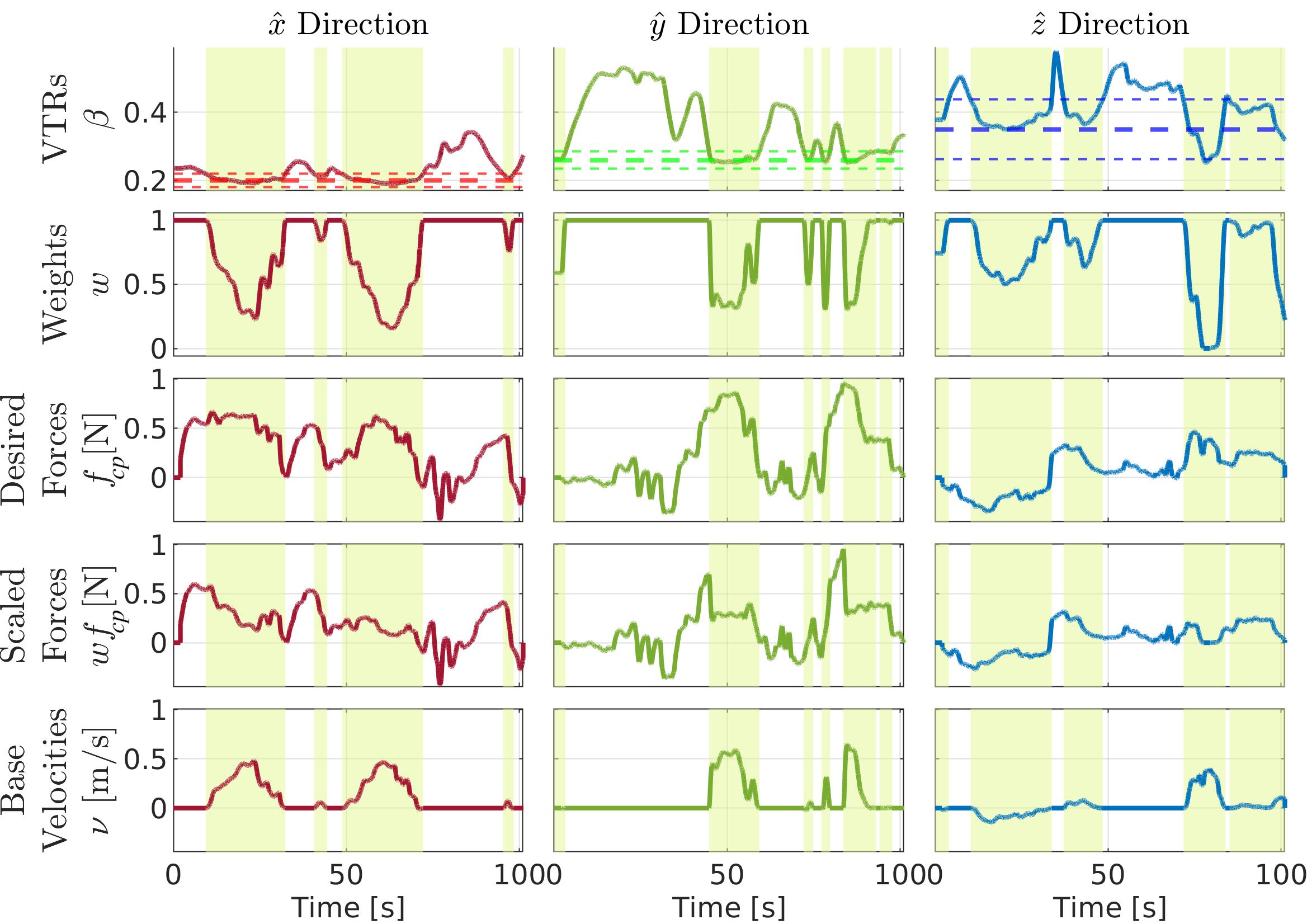}
	\caption[Autonomy enhanced \acrshort{tpo}: reaching locations experiment plots]{Plots of the reaching experiment employing the manipulability-aware motion generation. Each column shows the data for a particular direction component, $\hat{x}, \hat{y}, \hat{z}$. From top to bottom row they are shown: (1) the \acrshort{vtr}, with the thresholds margins $d-\Delta, d, d+\Delta$ represented by dashed horizontal lines; (2) the weights computed from the \acrshort{vtr}; (3) the \acrshort{tpo} virtual forces applied on the left end-effector by the operator; (4) the weighted version of such virtual forces; (5) the weighted mobile base velocities \eqref{eq:tpo2:nuWTPO}. The colored regions represents where the \acrshort{vtr} is below the given threshold $d + \Delta$.}
	\label{fig:TPO2ExpThreeBtnPlot}
\end{figure}

In \figurename{}~\ref{fig:TPO2ExpThreeBtnPlot}, the relevant quantities of the manipulability-aware motion generation are illustrated. The columns represent the three principal axes $\hat{x}, \hat{y}, \hat{z}$, so the data in each row is divided in the $x, y, z$ components. In the top row, the \acrshort{vtr} measure (\eqref{eq:beta_i}), represented together with the threshold margins $d-\Delta, d, d+\Delta$ (dashed lines), governs the activation of the manipulability-aware motion generation. The highlighted areas represent the time intervals when the \acrshort{vtr} is below the $d+\Delta$ threshold, which triggers the activation of the weight $w$ (represented in the second row) according to \eqref{eq:tpo2:weight}. In the third row, the \acrshort{tpo} virtual forces applied on the left end-effector generated by the user (\eqref{eq:tpo:f_cp}) are shown. According to the weight computed, these virtual forces are scaled to generate the joint torques of the arm $\boldsymbol{\tau}_{\mathit{cp}}$ with \eqref{eq:tpo2:taoWTPO} and the mobile base velocity with $\boldsymbol{\nu}$ \eqref{eq:tpo2:nuWTPO}. The scaled virtual forces $wf_{\mathit{cp}}$ and the mobile base velocity $\nu$ are plotted in the last two bottom plots, respectively. It can be observed in the highlighted areas how the desired forces are scaled down in favor of the mobile base velocities, following the weight $w$ computed from the \acrshort{vtr}.

\begin{figure}[H]
	\centering
	\includegraphics[width=1\linewidth]{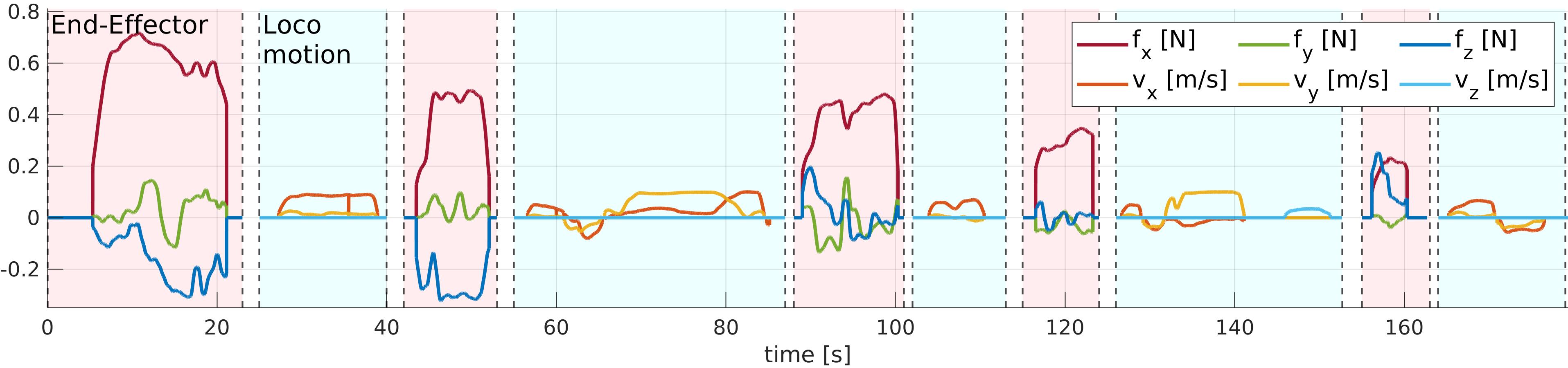}\\
	\includegraphics[width=1\linewidth]{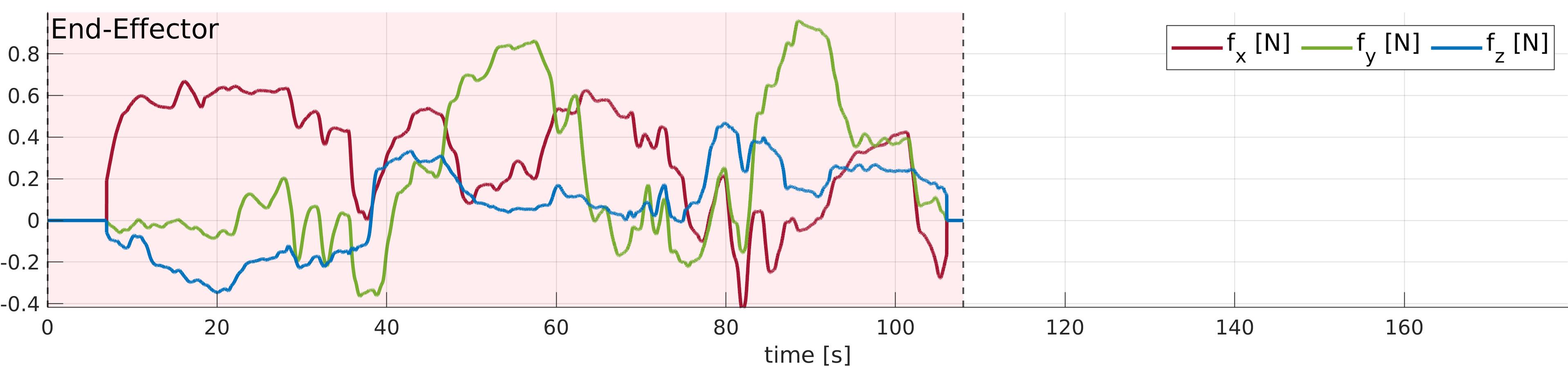}
	\caption[Autonomy enhanced \acrshort{tpo}: reaching locations experiment plots modalities comparison]{Plots of the reaching experiment, comparing the user inputs when the manipulability-aware motion generation is not employed (top plot) and when it is (bottom plot). In the top plot, it can be noticed how the operator has to switch repeatedly between the \textit{End-Effector} control point and the \textit{Locomotion} control point, increasing dead times and the overall execution time. }
	\label{fig:TPO2ExpThreeBtnDual}
\end{figure}

Finally, \figurename{}~\ref{fig:TPO2ExpThreeBtnDual} shows a comparison of the executions of the experiment without (top plot) and with (bottom plot) the manipulability-aware motion generation.

In the top plot, the user must generate both arm and mobile base motions, repetitively switching the control point from the \textit{End-Effector} to the \textit{Locomotion}. In the plot it is shown the virtual force $\boldsymbol{f}_\mathit{cp}$ (\eqref{eq:tpo:f_cp}) and the mobile base command (\eqref{eq:x_cp}) generated with the \acrshort{tpo} interface for the \textit{End-Effector} and \textit{Locomotion} control points, respectively.
Instead, in the bottom plot, the operator can just command virtual forces on the \textit{End-Effector} because the proposed method generates mobile base motions when the end-effector is in a region of low manipulability; permitting in any case to reach the locations that are initially outside the reachable workspace of the end-effector.

In the first modality, noticeable delays occur during the operation due to the repetitive changes of the control points. This results in the operator performing more actions, putting more effort, and slowing the execution of the task with respect to the second modality. Furthermore, there is no guarantee that the arm will remain a region of high manipulability, a measure which is not always easily accountable by the operator by watching the robot. This can further augment the operator's effort and the execution time due to the added difficulties of controlling the end-effector when the manipulability is low.

\subsection{Object Mass Estimation Validation}\label{sec:tpo3:estimation}
\begin{figure}[H]
	\centering
	\includegraphics[width=0.6\linewidth]{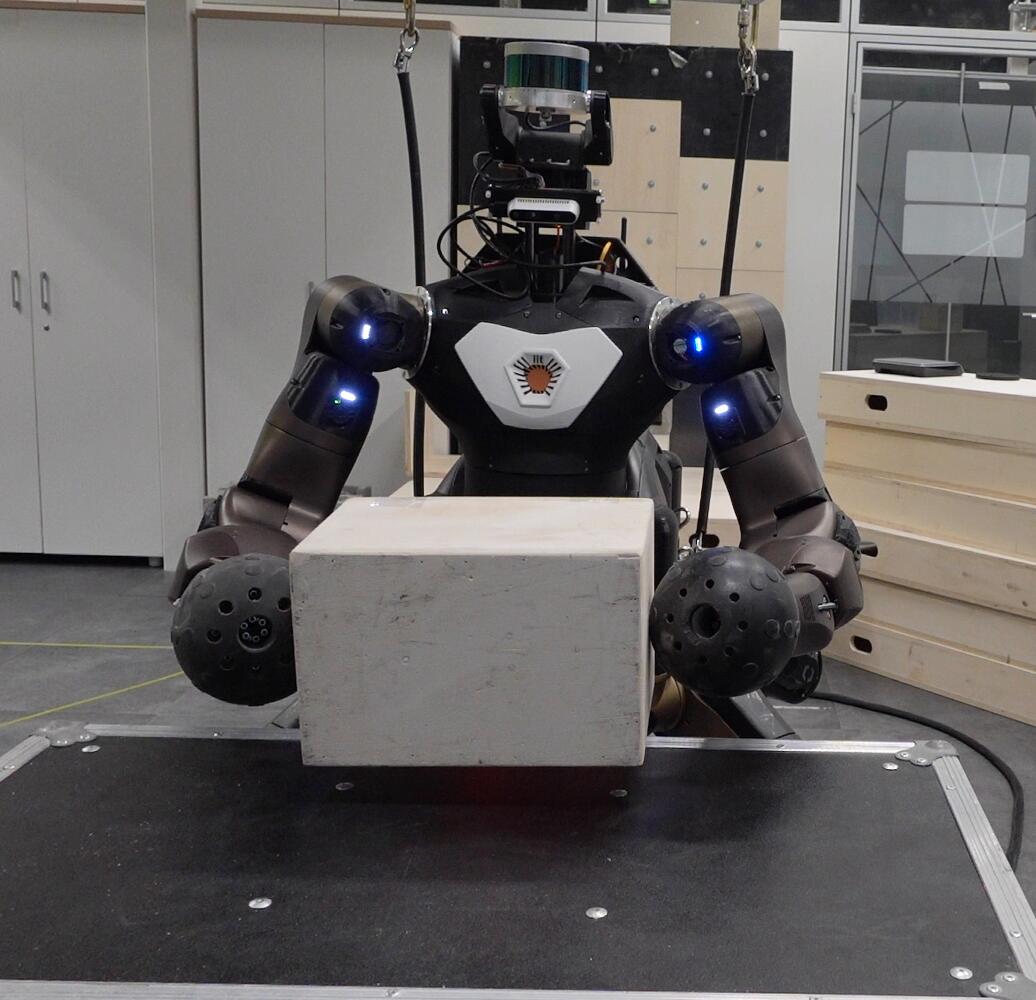}
	\caption[CENTAURO robot estimating the mass of a block]{The CENTAURO robot in one of the trial to estimate the mass of a wooden block.}
	\label{fig:mass_est}
\end{figure}

The bimanual grasping and transporting interface of Section~\ref{sec:tpo3:methods} includes a component to estimate the mass of the bimanually grasped object, detailed in Section~\ref{sec:tpo3:massEst}. This procedure is evaluated estimating the mass of a wooden block to assess its precision (\figurename{}~\ref{fig:mass_est}). 
In this experiment, the CENTAURO robot arms are placed at the sides of a wooden block object of \SI[separate-uncertainty = true]{1.958(2)}{\kilo\gram}. 
An initial grasping force is reached to allow the object to be firmly lifted. Three different initial grasping forces are chosen (\SI{45}{\newton},  \SI{40}{\newton}, and \SI{35}{\newton}) to evaluate the effect of the choice of the initial grasping force on the estimation of the mass of the object. All the three values are sufficiently high to lift the object safely without any slippage during the mass estimation. As soon as the lifting phase is completed, the mass of the object is estimated through \eqref{eq:mass_est}. The procedure has been performed $5$ times for each initial grasping force, for a total of $15$ trials.

\begin{figure}[H]
	\centering
	\includegraphics[width=1\linewidth]{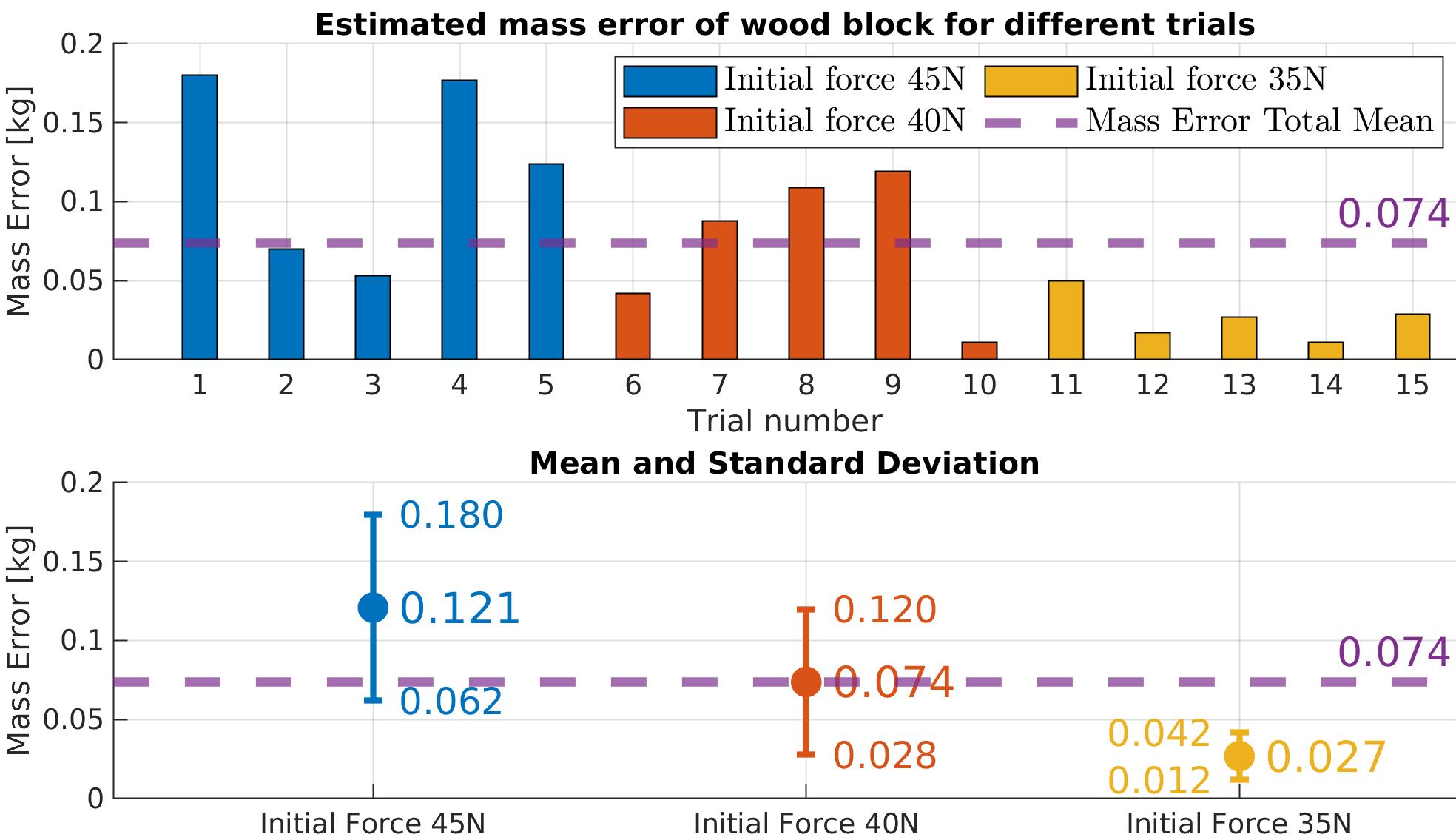}
	\caption[Results of the mass estimation of a bimanually grasped block]{Results over $15$ trials of the mass estimation of the wooden block. In the top plot, the estimation error of each trial is displayed; the three different colors represent the different initial grasping forces set (\SI{45}{\newton},  \SI{40}{\newton}, and \SI{35}{\newton}). In the bottom plot, the mean and the standard deviation for initial grasping force are shown. The purple dashed lines represent the overall error mean among all the $15$ trials.}
	\label{fig:mass_est_plot}
\end{figure}

The results of the mass estimation trials are shown in \figurename{}~\ref{fig:mass_est_plot}. It can be observed that the initial grasping force only slightly influences the mass estimation error. Interestingly, the estimated mass error is lower when the initial grasping force is lower. This may be due to the kinematic differences caused by the application of different amounts of forces due to the unmeasured elasticity in the robot structure, despite the encoders registering the same robot configuration used by the force estimation module to estimate the forces from the joint torques in the arms.
The results do not indicate a significant influence of the choice of the initial grasping force in the mass estimation, verifying the robustness of the approach at different levels of initial grasping forces.
The average error over all the trials has resulted to be \SI{0.074}{\kilo\gram}, with a standard deviation of \SI{0.057}{\kilo\gram}, resulting in a relative standard error of $3.769\%$.

\subsection{Box Approaching and Transporting}\label{sec:tpo2:box}

\begin{figure}[H]
	\centering
	\includegraphics[trim={0 6cm 0 0},clip, width=0.49\linewidth]{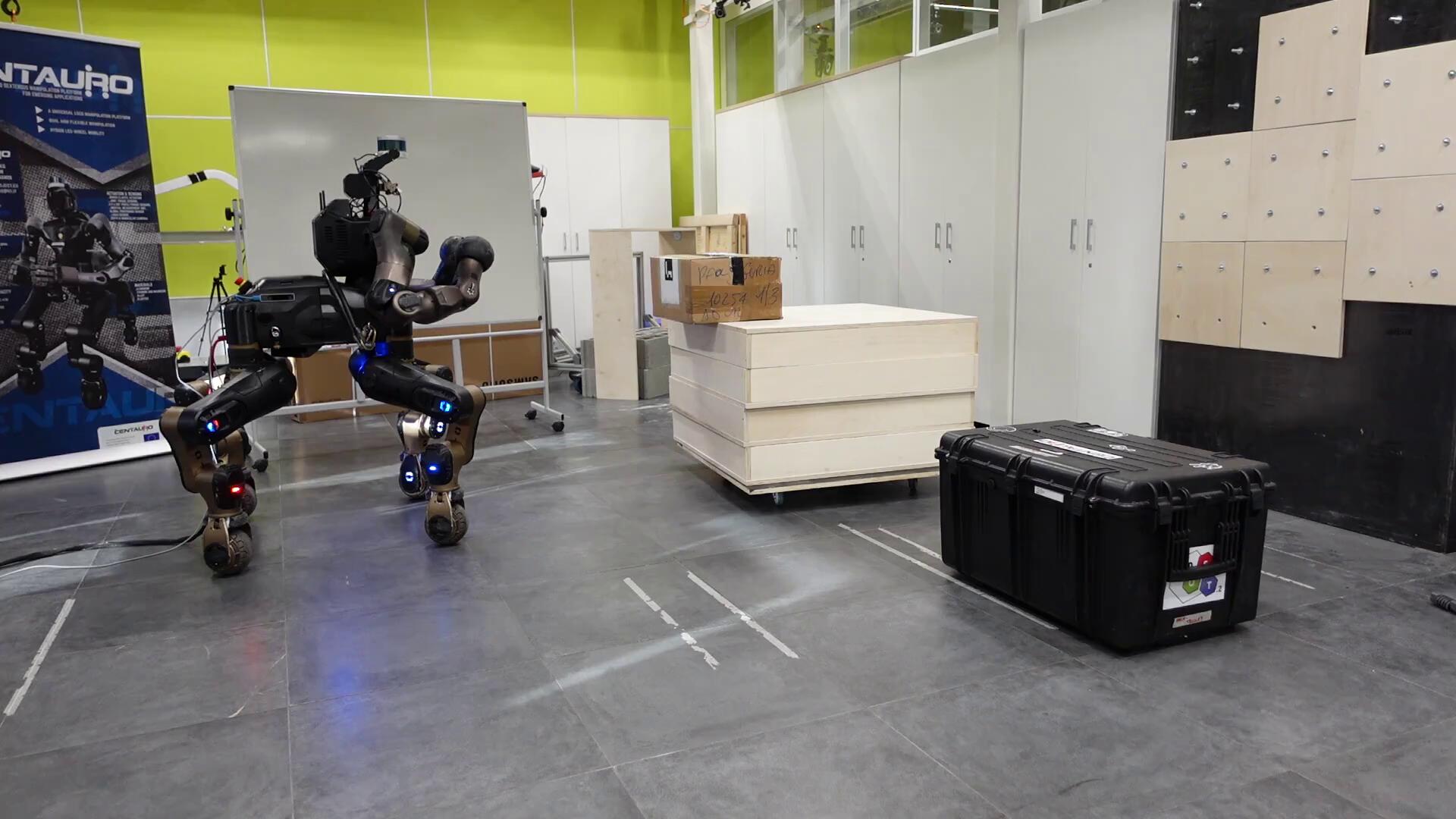}
	\includegraphics[trim={0 6cm 0 0},clip, width=0.49\linewidth]{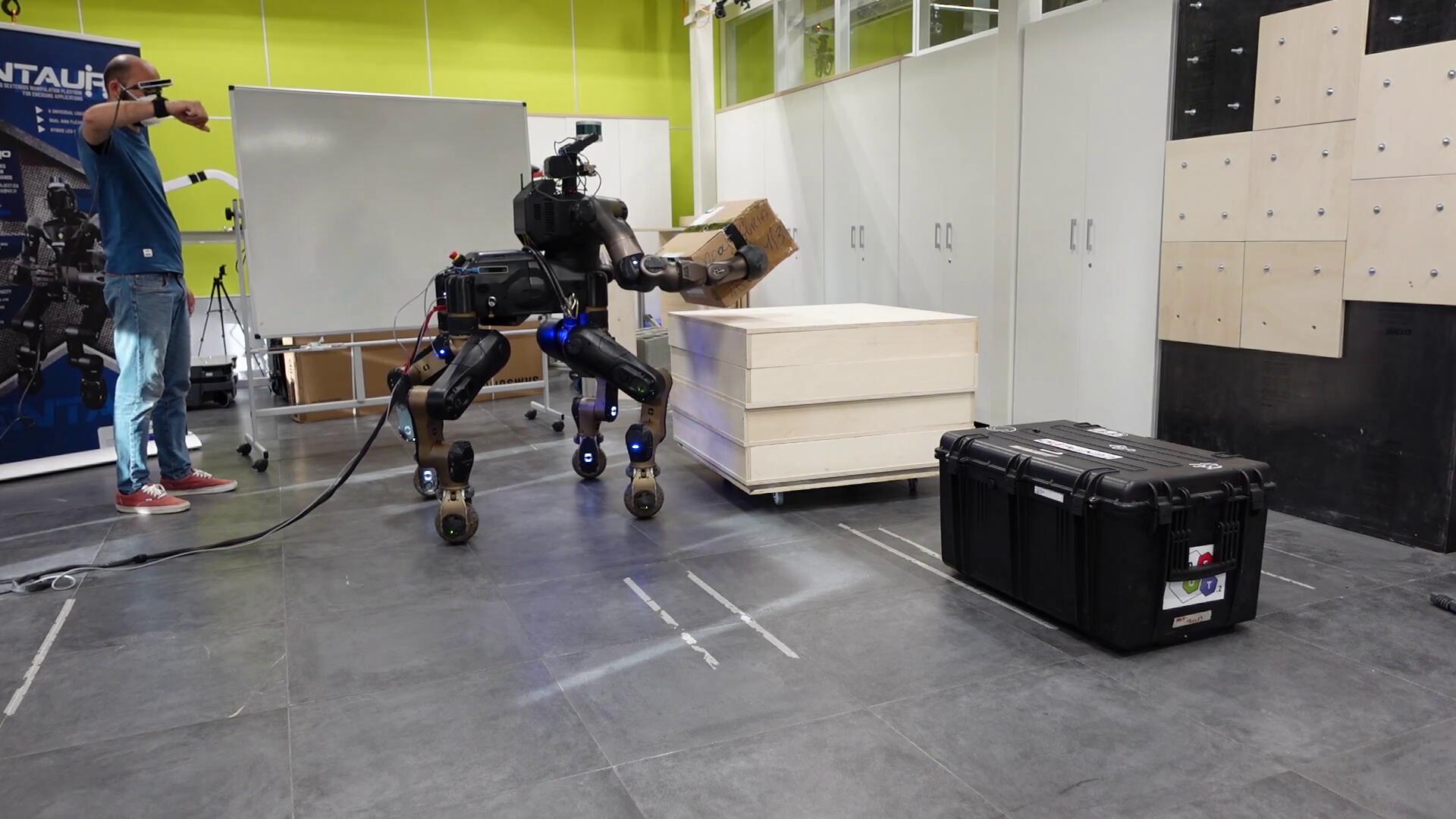}
	\caption[Autonomy enhanced \acrshort{tpo}: box approaching and transporting experiment]{The two phases of the experiment. On the right, the robot approaches autonomously the box to grasp it. On the left, the user operates the robot to transport the box.}
	\label{fig:TPO2ExpBoxGeneric}	
\end{figure}

This experiment validates the bimanual grasping and transporting interface of Section~\ref{sec:tpo3:methods} in an experiment where a box must be grasped and transported bimanually in a specific location. The experiment is composed of two phases as shown in \figurename{}~\ref{fig:TPO2ExpBoxGeneric}. 
In the first phase the robot autonomously approaches the object and grasps it with the two arms. In the second phase the user operates the robot to transport the object to the final location. 

\subsubsection{Autonomous Approaching and Grasping Phase}
\begin{figure}[H]
	\centering
	\includegraphics[width=0.49\linewidth]{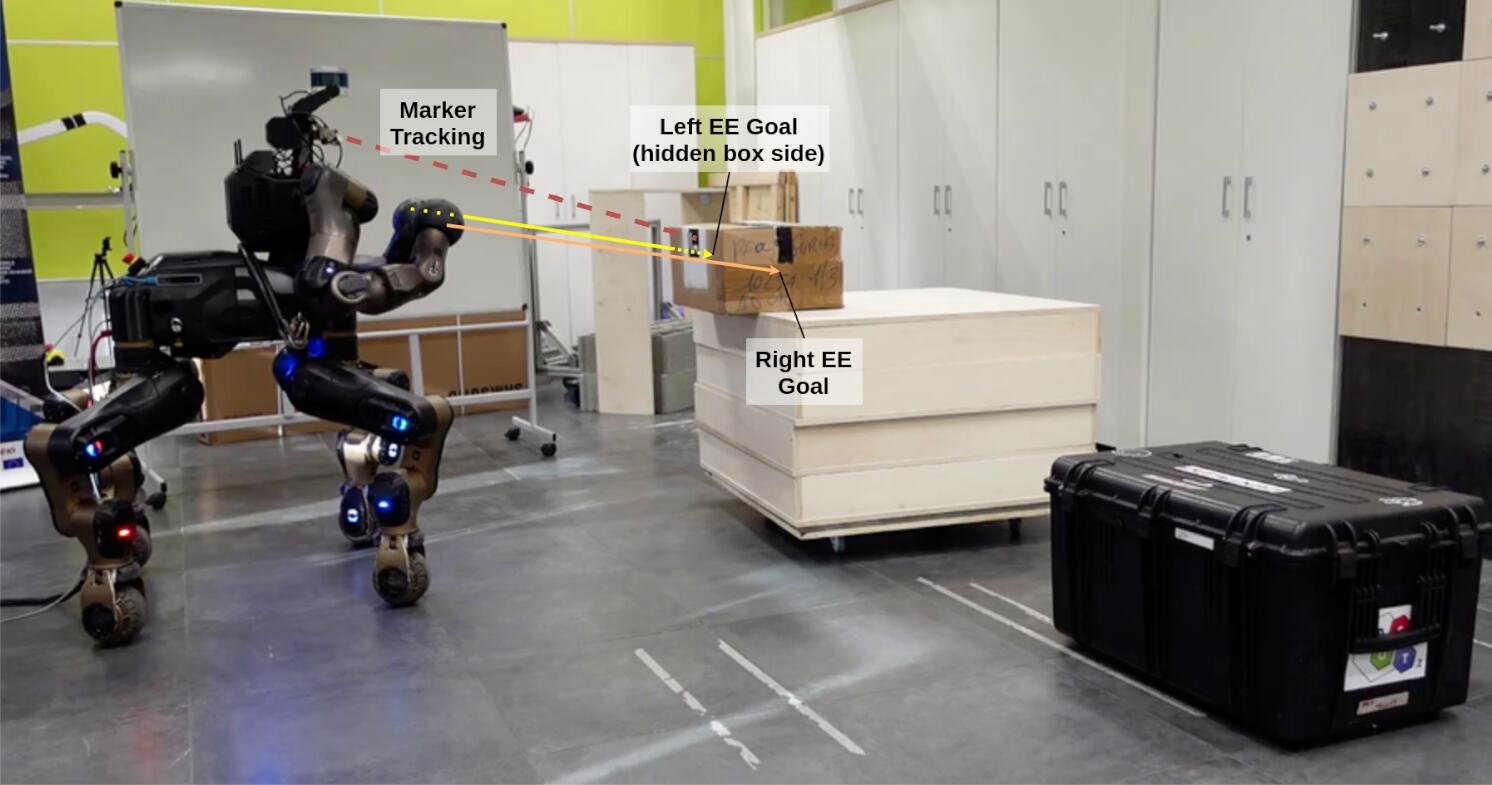}
	\includegraphics[width=0.49\linewidth]{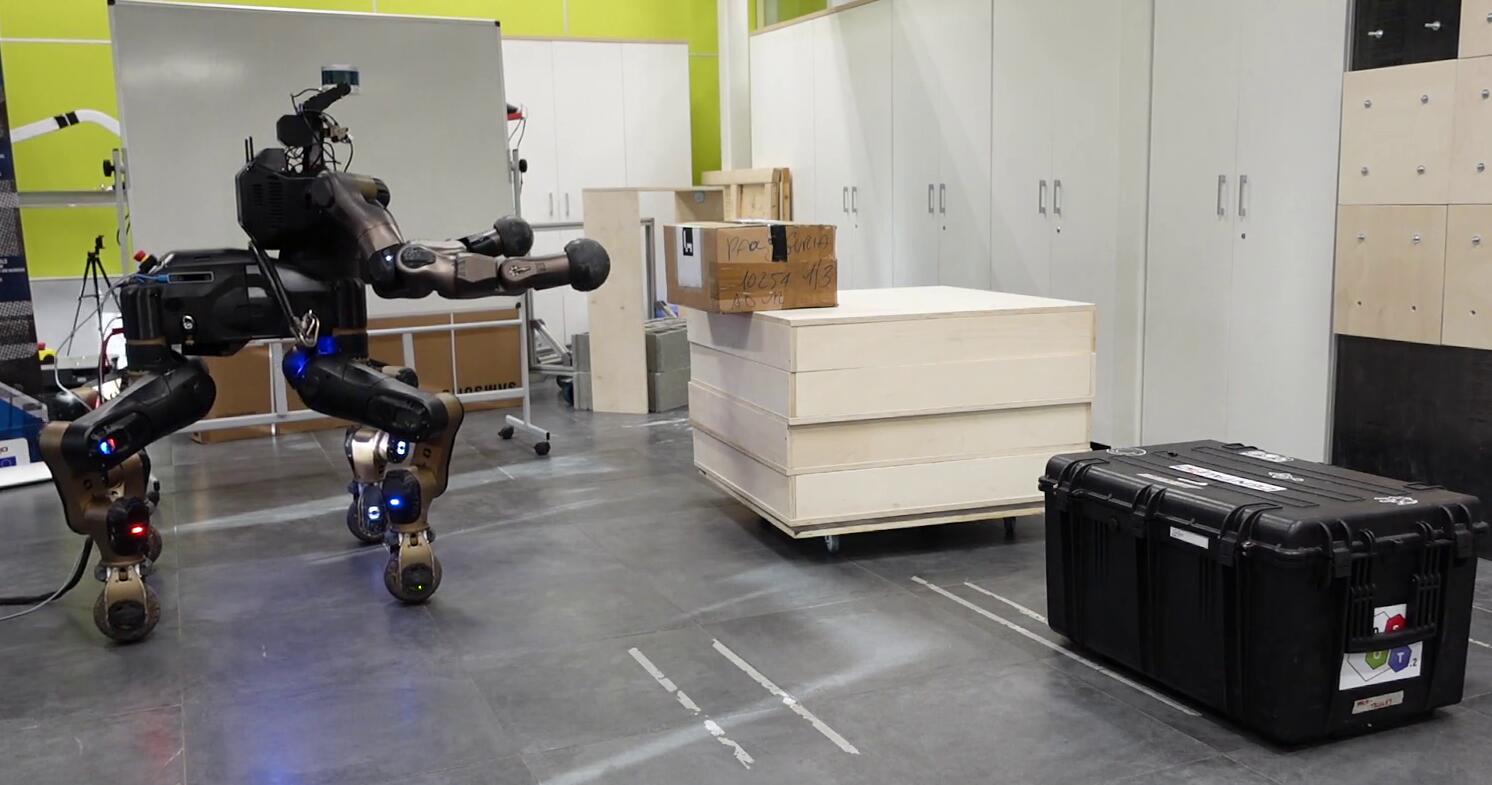}\\
	\vspace{5px}
	\includegraphics[width=0.49\linewidth]{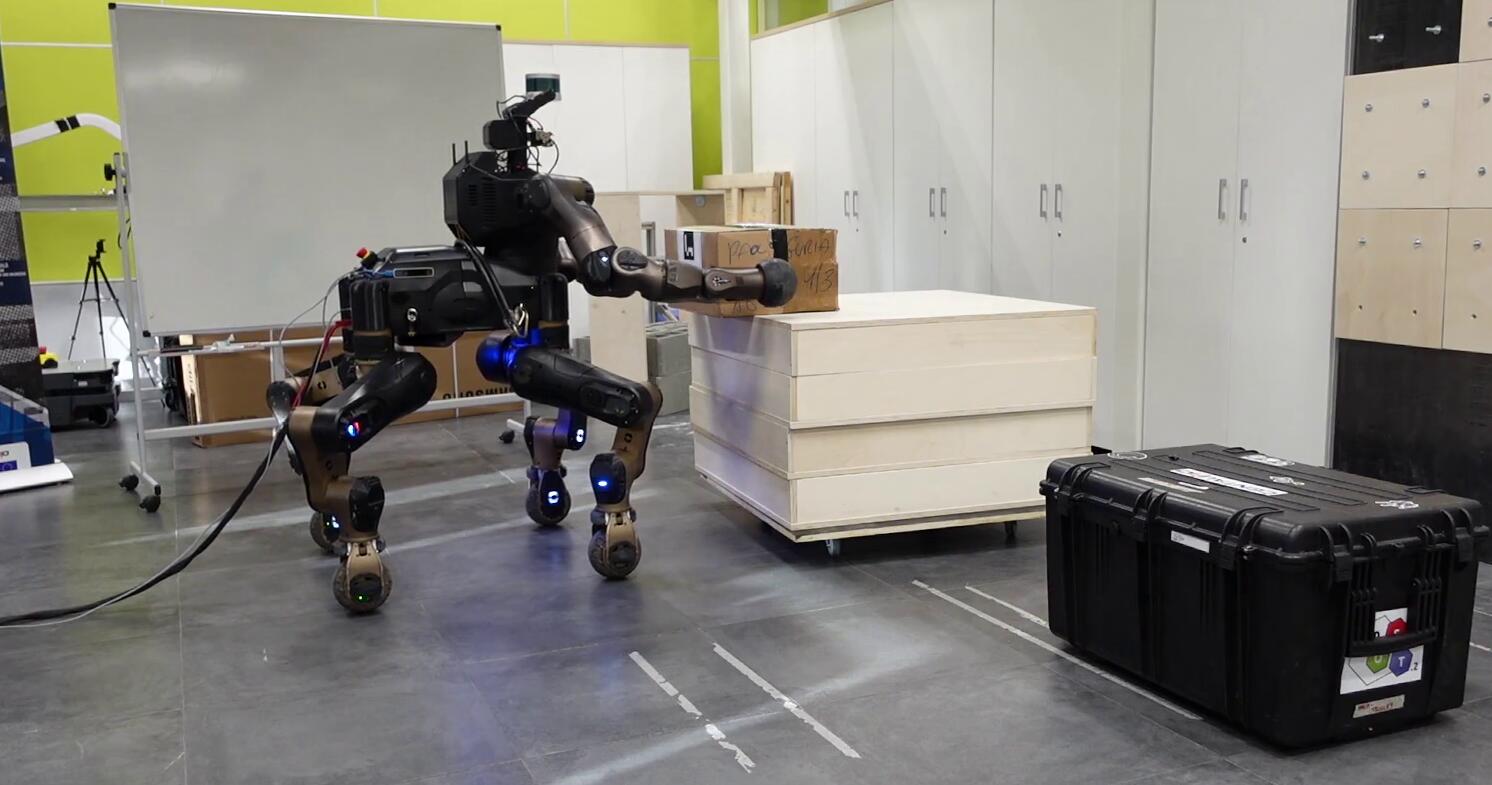}
	\includegraphics[width=0.49\linewidth]{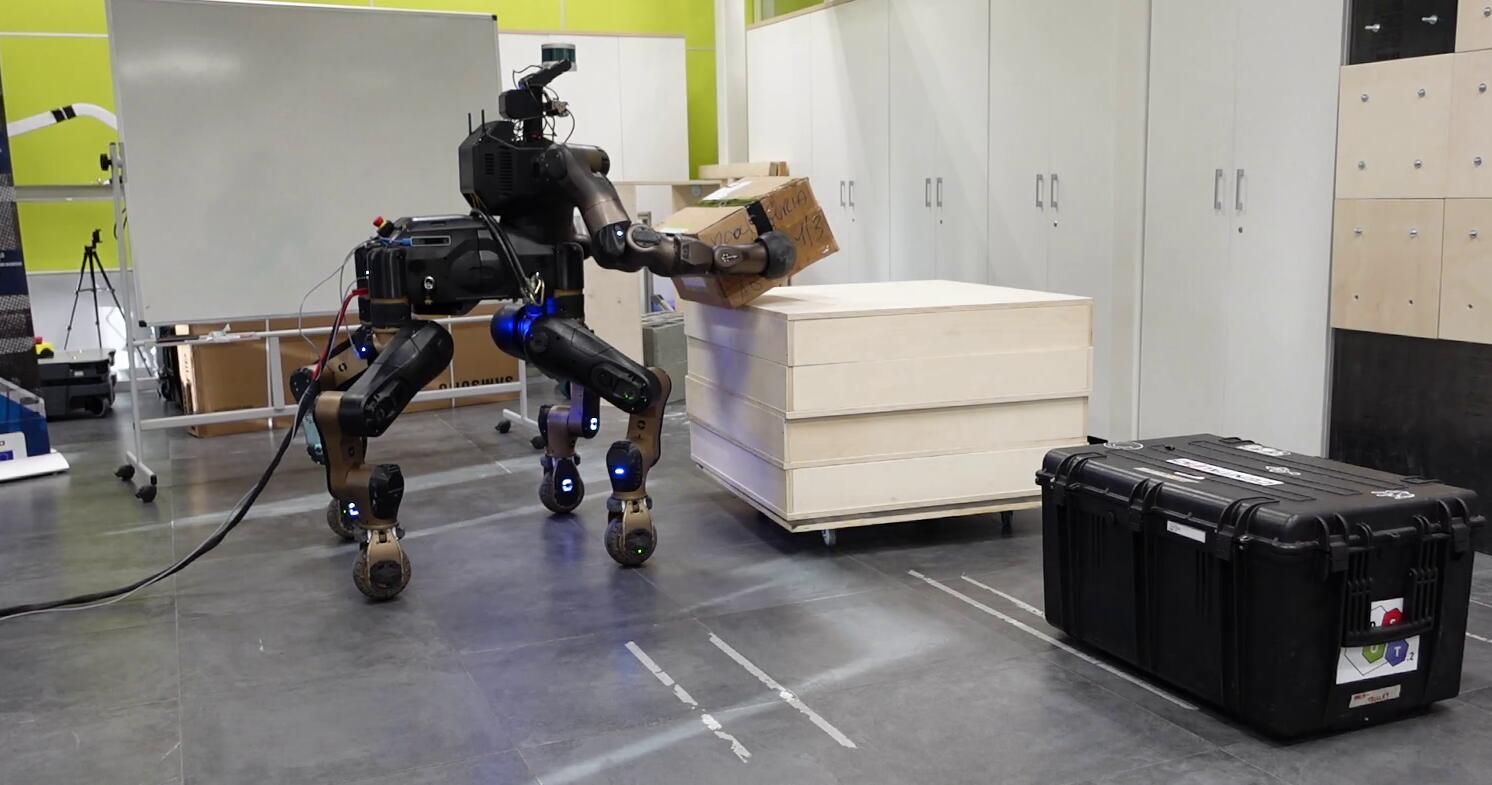}
	\caption[Autonomy enhanced \acrshort{tpo}: box approaching and grasping phase]{In the first phase of the experiment, the robot autonomously approaches the box by mean of its vision system and lifts it up, computing the grasping forces necessary for a safe teleoperated transportation that will happen in the second phase.}
	\label{fig:tpo2:box_approach_exp}
\end{figure}

Sequences of the approach and grasping phase are depicted in \figurename{}~\ref{fig:tpo2:box_approach_exp}. The robot, by tracking the box with its vision system, autonomously approach the object and lift it up.
The vision system relies on a \acrshort{rgb} camera placed on the robot head, and on a \textit{ArUco} marker detection algorithm, which updates the box pose with respect to the robot while the robot is moving forward. For the pose detection, the size of the box is assumed to be known. 

During the approaching, each end-effector is driven by two goals set-points: with the first the two arms are enlarged, while the second set-points are located at the box sides. From these goals, Cartesian velocities for each arm are generated through an inverse kinematic process. At the same time the manipulability-aware motion generation (Section~\ref{sec:tpo2:appVel}) is activated when the \acrshort{vtr} of the end-effectors is below the threshold, activating the mobile base motions which are necessary because initially the box is outside the arms reachable workspace. The final velocity commanded to
the mobile base is the average of the two $\boldsymbol{\nu}$ computed from the weight $\boldsymbol{W}$ of each arm.

Once the robot has reached the sides of the object with the two end-effectors, the arms are closed toward the object until the initial grasping force is reached. At this point, the box is lifted, its mass estimated, and the required grasping force is derived as explained in Section~\ref{sec:tpo3:methods}.

\begin{figure}[H]
	\centering
	\includegraphics[width=0.78\linewidth]{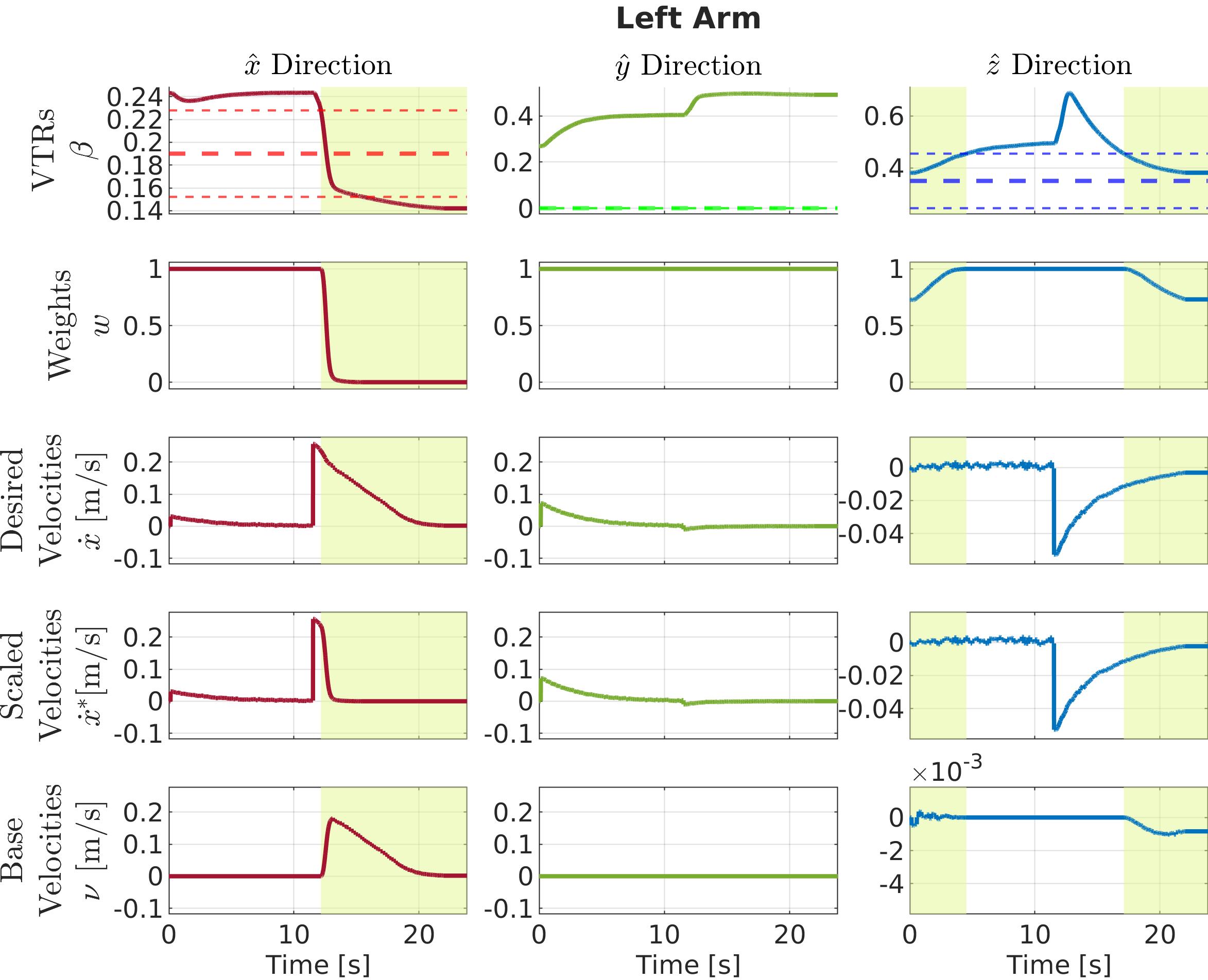}\\
	\vspace{5px}
	\includegraphics[width=0.78\linewidth]{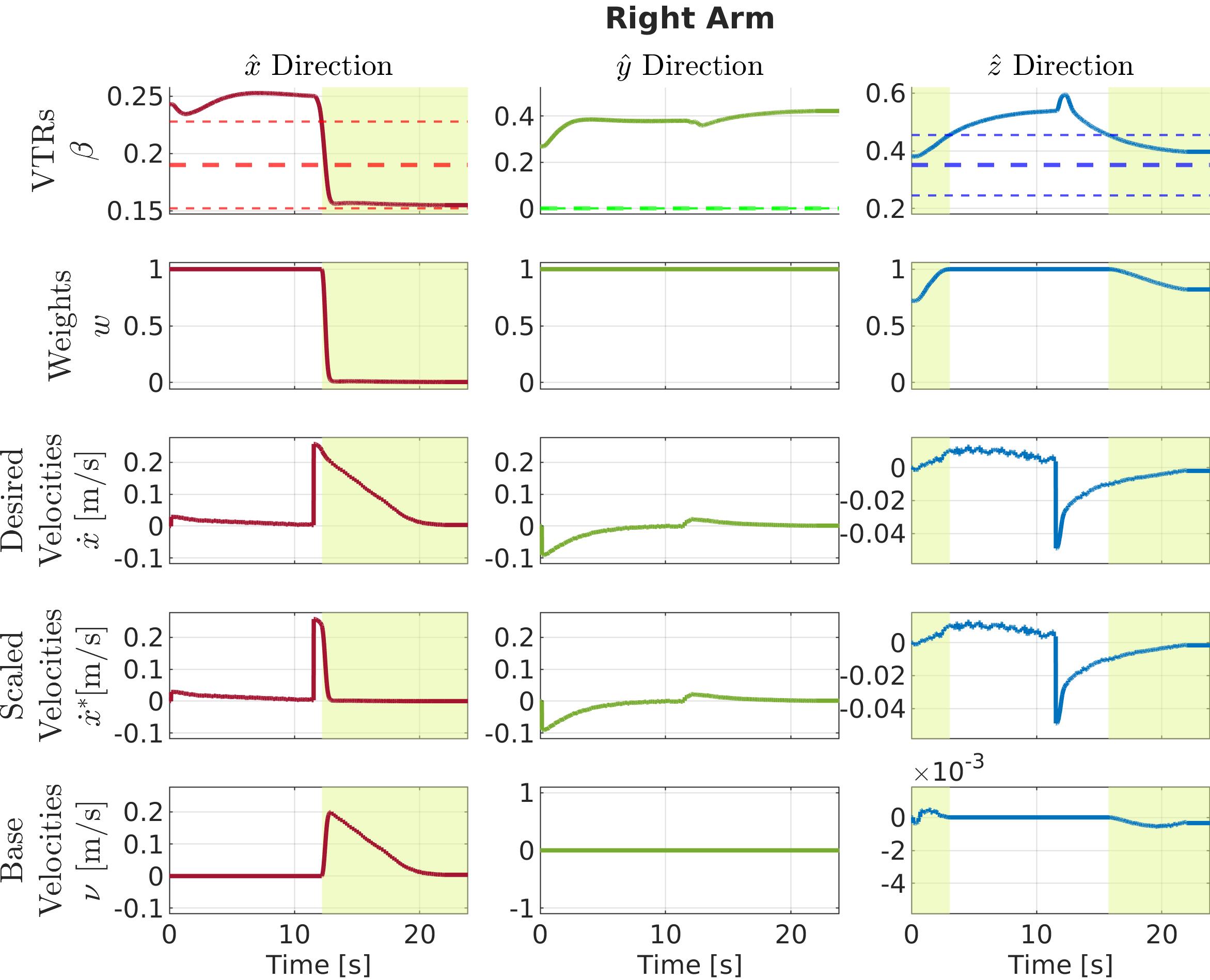}
	\caption[Autonomy enhanced \acrshort{tpo}: box approaching and grasping phase plots]{Plots for the box approaching task. For each arm, each column shows the data for a particular direction, $\hat{x}, \hat{y}, \hat{z}$. From top to bottom rows, in each column there are shown the $x, y, z$ components of: (1) the \acrshort{vtr} with the thresholds margins $d-\Delta, d, d+\Delta$ represented by the dashed horizontal lines; (2) the weight computed from the \acrshort{vtr}; (3) the desired end-effector Cartesian velocities computed from the given end-effector goals; (4) the scaled desired end-effector Cartesian velocities; (5) the weighted mobile base velocities. The colored regions represents where the \acrshort{vtr} is below the given threshold $d+\Delta$.}
	\label{fig:box_approach_manip}
\end{figure}

\figurename{}~\ref{fig:box_approach_manip} shows plots relative to the approaching action, whose disposition is similar to the one of \figurename{}~\ref{fig:TPO2ExpThreeBtnPlot}. The first series of plot refers to the left arm of the robot, the other series to the right arm of the robot. 
For each arm, the columns represent the three principal axes $\hat{x}, \hat{y}, \hat{z}$, so the data in each row is divided in the $x, y, z$ components. In the top row, the \acrshort{vtr} measure (\eqref{eq:beta_i}), represented together with the threshold margins $d-\Delta, d, d+\Delta$ (dashed lines), governs the activation of the manipulability-aware motion generation. 
The highlighted areas represent the time intervals when the \acrshort{vtr} is below the $d+\Delta$ threshold, which triggers the activation of the weight $w$ (represented in the second row) according to \eqref{eq:tpo2:weight}. 
In the third row, the desired end-effector Cartesian velocities $\boldsymbol{\dot{x}}^*$ computed to reach the two goal set-points are show. It is visible that first the arm are enlarged (note the component on the $\hat{y}$ axis), and then moved forward to reach the box sides.
According to the weights $w_i$ and the \eqref{eq:tpo2:xDotWTPO}, these Cartesian velocities are scaled as shown in the fourth row, and, consequently, mobile base velocity $\nu$ generated as shown in the fifth row. It can be observed in the highlighted areas how the desired end-effector velocities are diminished in favor of the mobile base velocities.
Note that the noise visible in the first part of the $z$ component is related to the uncertainties of the vision tracking system. 

It's worth noticing the differences of these plots with the ones of \figurename{}~\ref{fig:TPO2ExpThreeBtnPlot}, relative to the reaching locations experiment (Section~\ref{sec:tpo2:buttons}). In particular, looking at the third and fourth rows, now the end-effector input is a Cartesian velocity, while in the previous experiment it was a \acrshort{tpo} virtual force. Consequently, different methods are used for the manipulability-aware motion generation: the previous experiment is an example of \textit{postural} motion generation (Section~\ref{sec:tpo2:apptorque}), while this experiment is an example of \textit{Cartesian} motion generation (Section~\ref{sec:tpo2:appVel}).

Concerning the \acrshort{vtr} activation, in this approaching phase of the experiment, the threshold along the $\hat{y}$ direction is set to zero, to prevent the motion of the mobile base laterally, which may cause loss of the marker tracking. 
This puts in evidence the flexibility of the approach: thanks to the decomposition of the manipulability analysis into the three directions, the generation of motions can be better adapted to the specific task. This would not be possible by using the classical \textit{scalar} manipulability measure of \cite{Yoshi1985}.

\subsubsection{Teleoperated Transporting Phase}
\begin{figure}[H]
	\centering
	\includegraphics[width=0.49\linewidth]{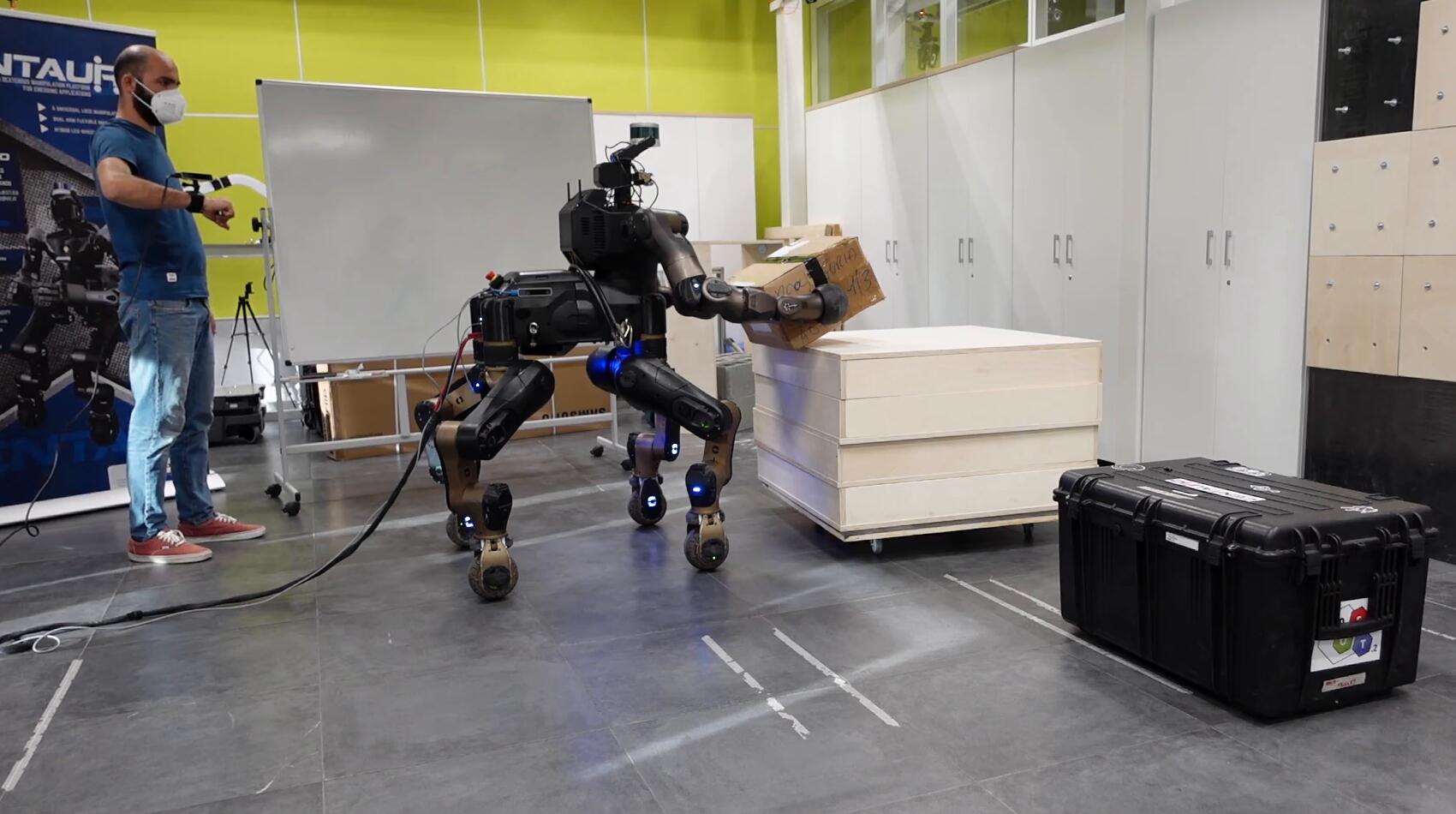}
	\includegraphics[width=0.49\linewidth]{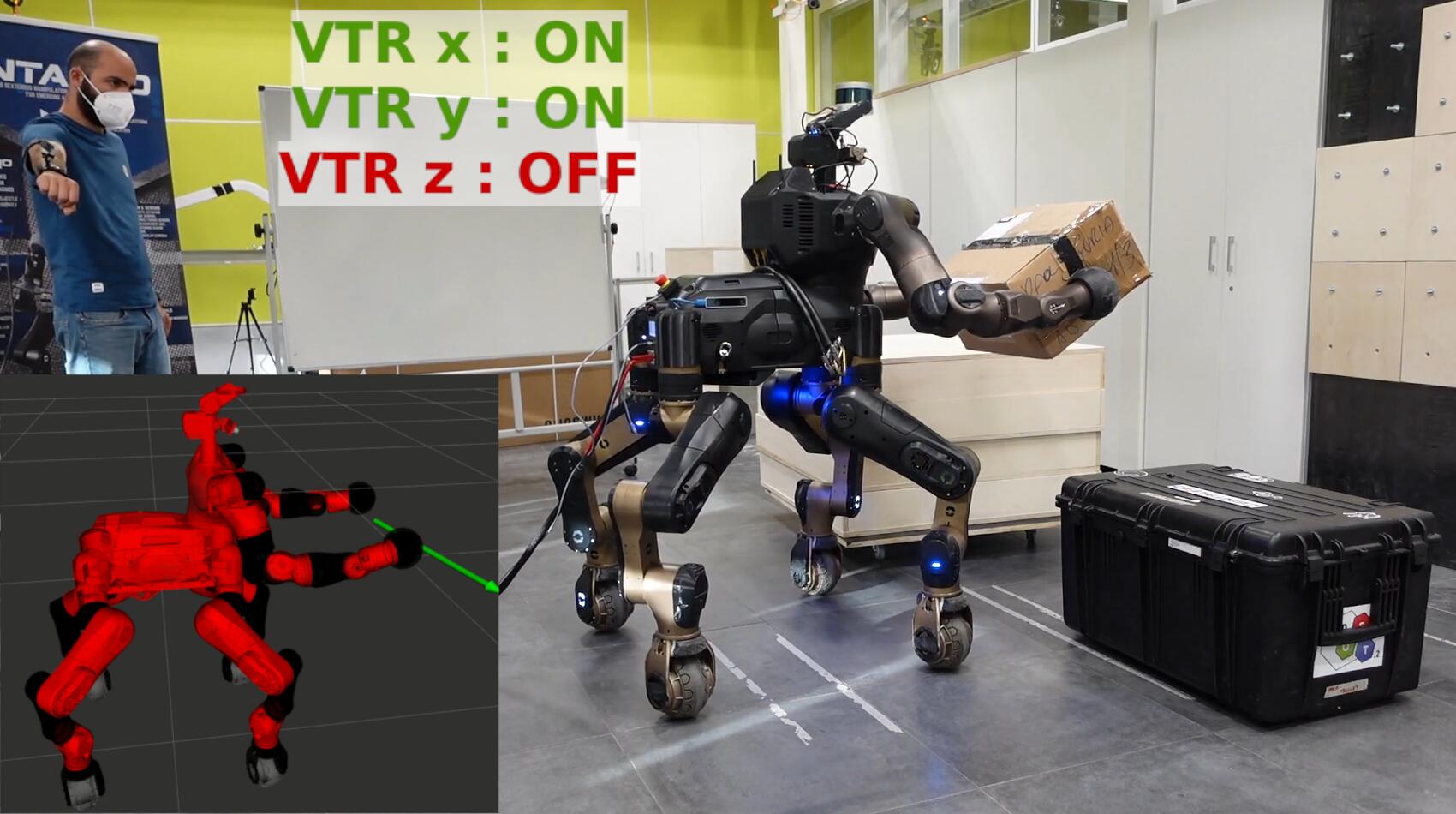}\\
		\vspace{5px}
	\includegraphics[width=0.49\linewidth]{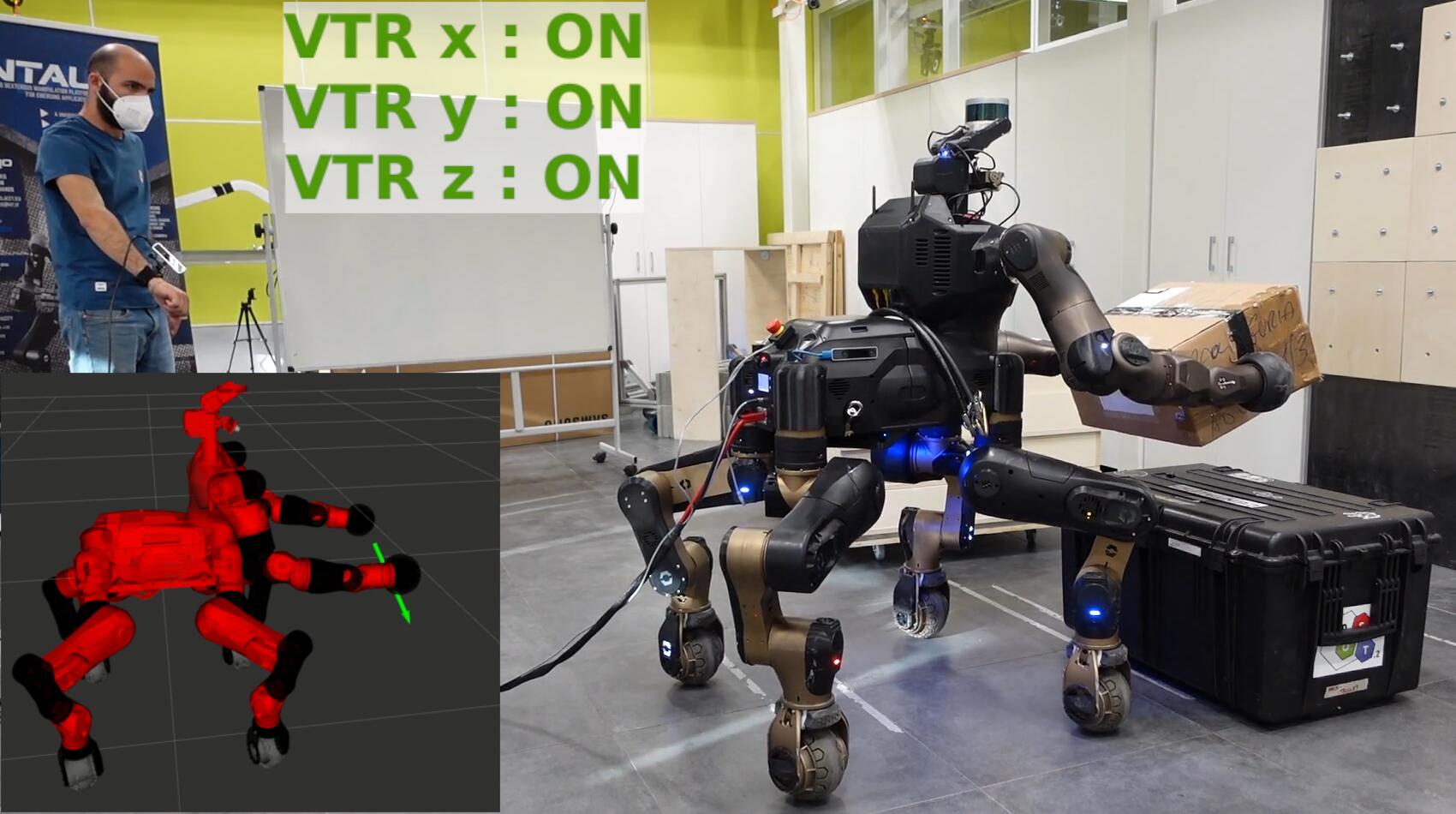}
	\includegraphics[width=0.49\linewidth]{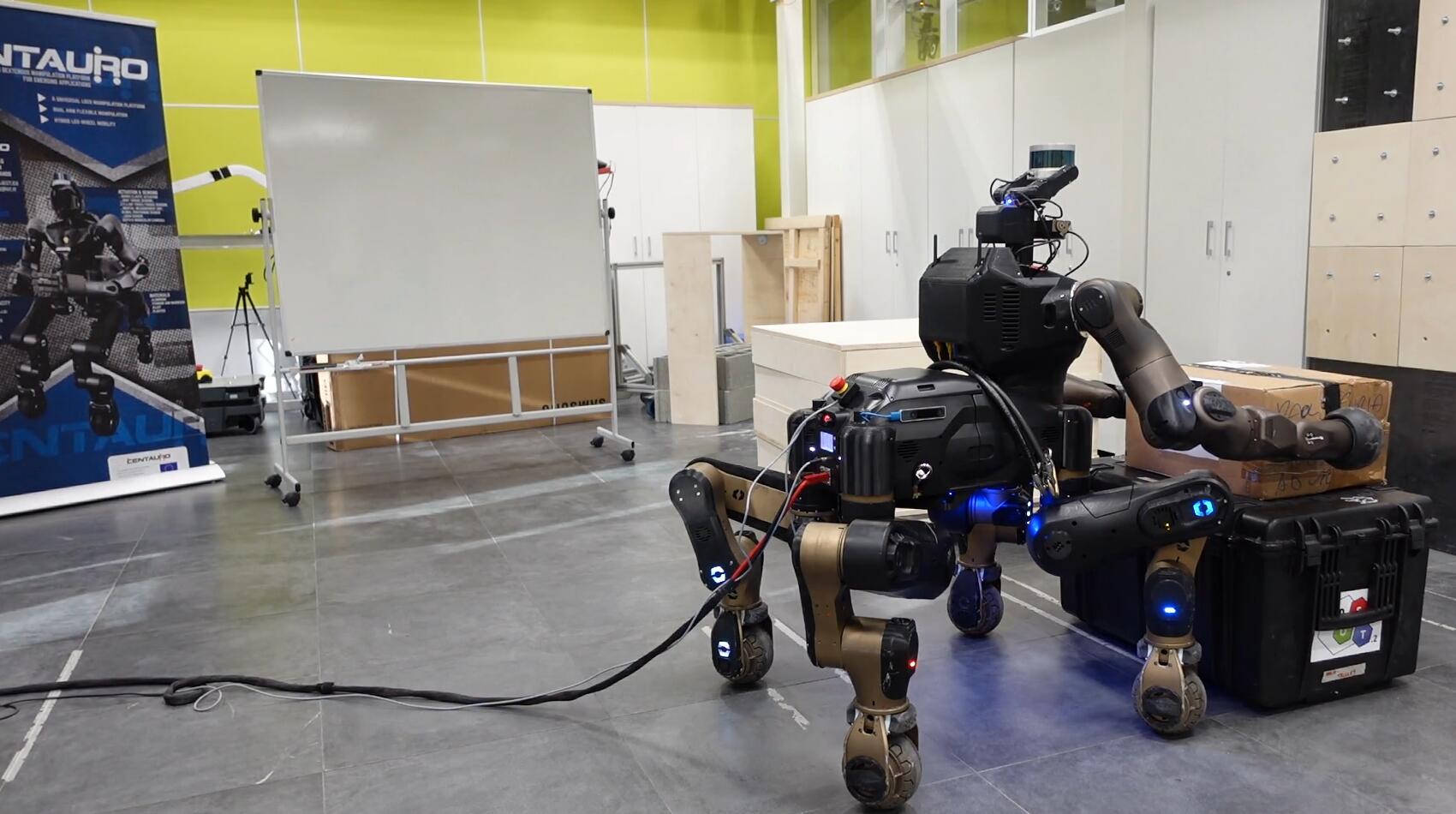}
	\caption[Autonomy enhanced \acrshort{tpo}: box transportation phase]{Sequences from the box transportation phase. From top to bottom, left to right: (1) the user begins to teleoperate the robot; (2) the \acrshort{vtr} is below the threshold on the $x$ and $y$ directions, making the mobile base moving in these directions according to the user commanded box velocity; (3) the \acrshort{vtr} is below the threshold also on the $z$ direction, making the robot to squat down too; (4) the box has been placed in the final location.}
	\label{fig:TPO2ExpBoxTransport}
\end{figure}

Once the robot has grasped the box, the user takes the control to transport it toward the final location as shown in \figurename{}~\ref{fig:TPO2ExpBoxTransport}. 
The robot generates motions according to the object transportation control interface explained in Section~\ref{sec:tpo3:transport}. At the final location, an additional command is issued by the operator to release the object and conclude the mission.

\begin{figure}[H]
	\centering
	\includegraphics[width=1\linewidth]{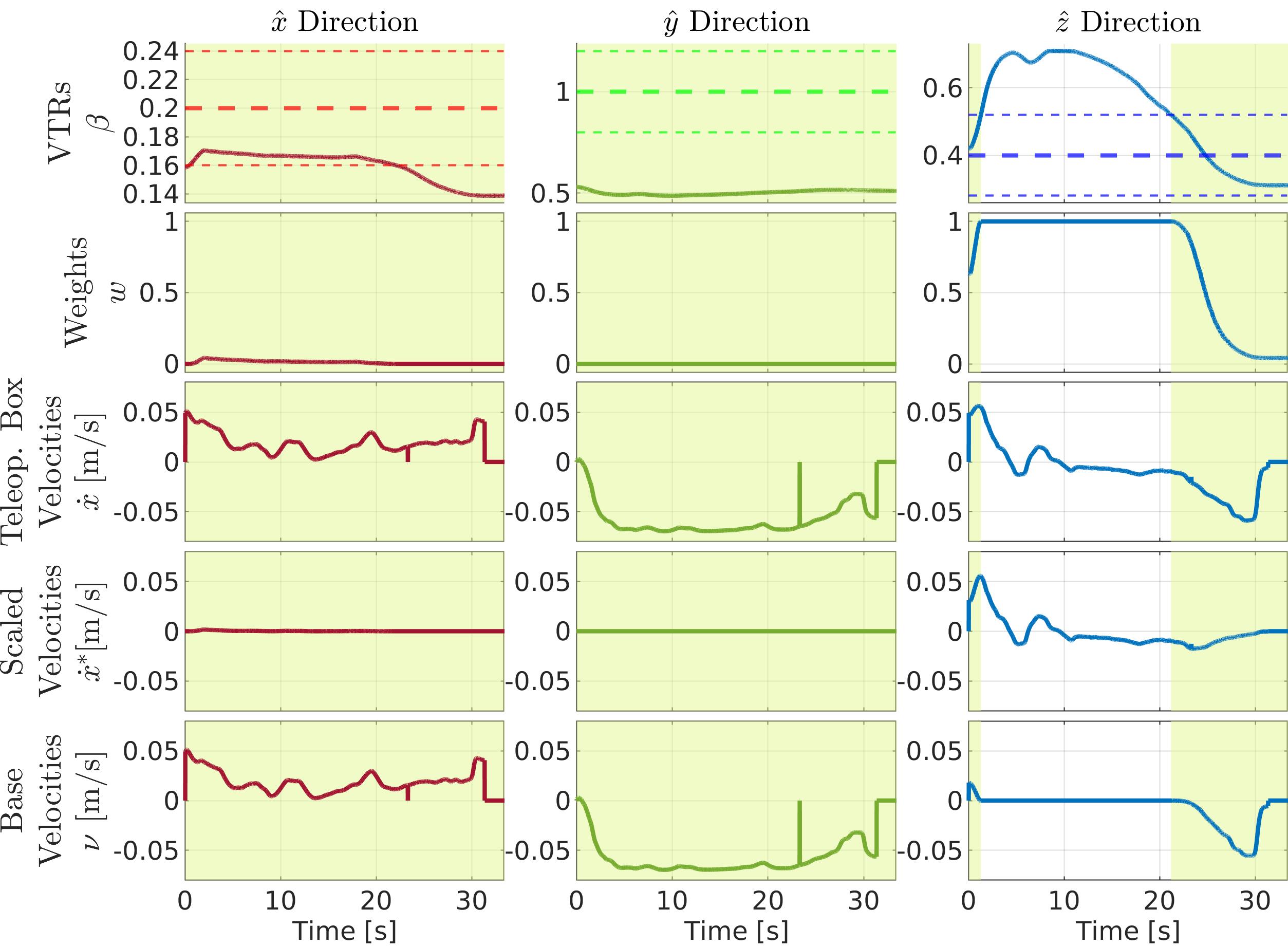}
	\caption[Autonomy enhanced \acrshort{tpo}: box transportation phase plots]{Plots for the box transportation phase. The data is relative to the arm which has the lowest \acrshort{vtr} at each instant. Each column shows the data for a particular direction, $\hat{x}, \hat{y}, \hat{z}$. From top to bottom rows, in each column they are shown the $x, y, z$ components of: (1) the \acrshort{vtr}, where the thresholds margins $d-\Delta, d, d+\Delta$ are represented by the dashed horizontal lines; (2) the weight computed from the \acrshort{vtr}; (3) the box Cartesian velocities commanded by the operator; (4) the scaled box velocities; (5) the weighted mobile base velocities. The colored region represents where the \acrshort{vtr} is below the given threshold $d+\Delta$.}
	\label{fig:box_transport_manip}
\end{figure}

The plots of \figurename{}~\ref{fig:box_transport_manip} show similar data as in \figurename{}~\ref{fig:box_approach_manip}, but now the third row shows the box velocities generated by the operator, instead of the automatically generated velocities computed to reach the goals set-points.
The desired box velocities are generated by the operator's right arm with the Cartesian motion generation of the \acrshort{tpo} interface (\eqref{eq:x_cp}). This reference is then scaled with the VTR-based Cartesian motion generation (\eqref{eq:tpo2:xDotWTPO}) according to the weight computed from the robot arm with in the worst condition of manipulability. Indeed, the first two rows of \figurename{}~\ref{fig:box_transport_manip} show, for each direction, data relative to the arm with the lowest \acrshort{vtr}. Consequently, mobile base velocities (bottom row) are generated. The scaled velocities $x^*$ are used as input for the object transportation control formula (\eqref{eq:tpo3:coopLaw}).
The overall result is that the arms and the body of the robot move cooperatively to follow the desired object directions keeping the grasping.

Concerning the \acrshort{vtr} thresholds, high values are set for $\hat{x}$ and $\hat{y}$ directions, to prevent significant motions in the arms along these directions to help to maintain the grasp of the box. In this way the velocity commands sent by the operator along $\hat{x}$ and $\hat{y}$ mainly result in the generation of mobile base velocities, effectively enabling the safe transportation of the grasped object to the final location.

\subsection{Transporting Objects of Different Masses}\label{sec:tpo3:box}

\begin{figure}[H]
	\centering
	\includegraphics[width=0.85\linewidth]{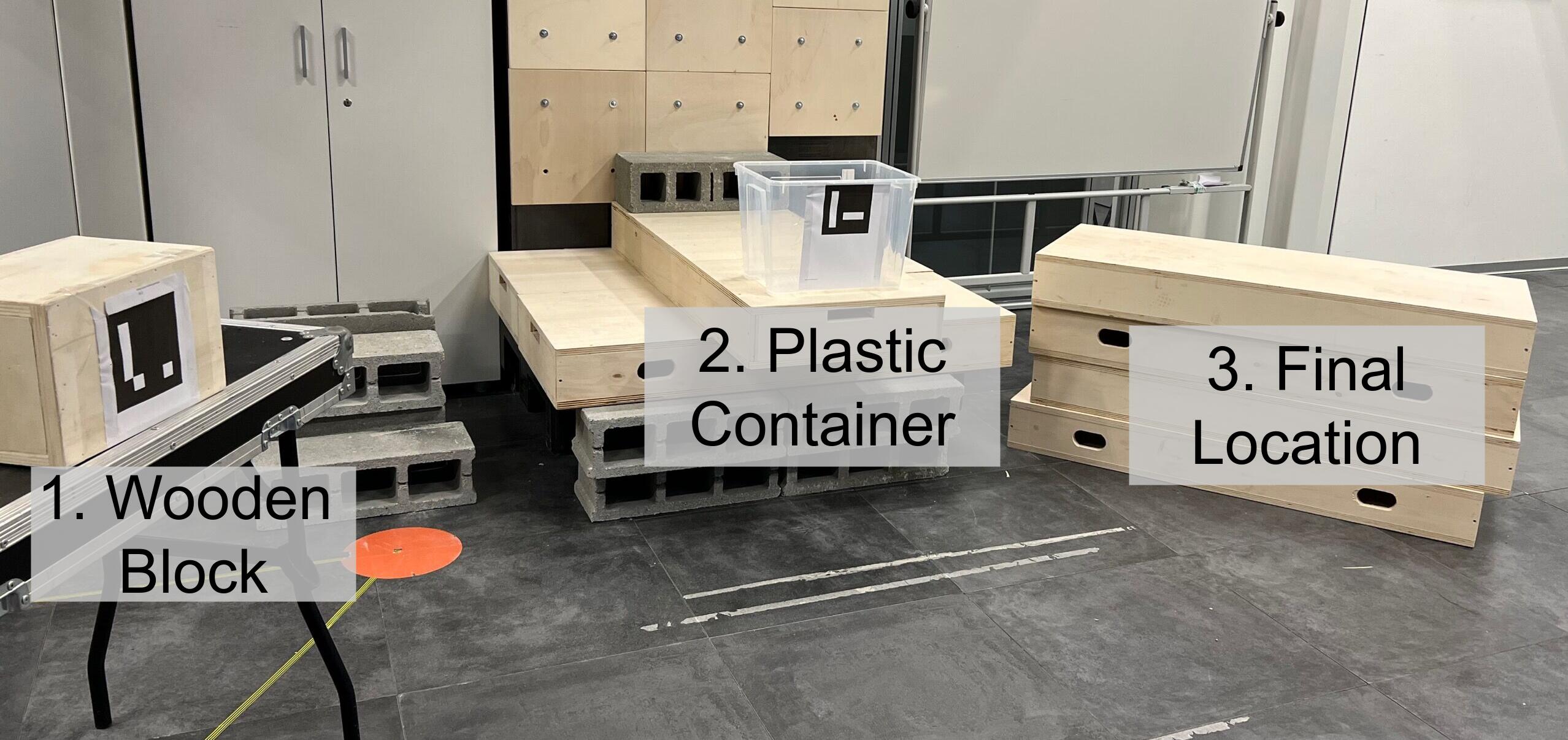}
	\caption[Autonomy enhanced \acrshort{tpo}: trasporting objects of different masses experimental setup]{The setup of the experiment about trasporting objects of different masses. The wooden block (1) must be placed inside the container (2). Then, the container with the block inside must be transported to the final location (3). The visible \textit{ArUco} markers permit the automatic approaching of the objects.}
	\label{fig:tpo3:box_exp_setup}
\end{figure}

This experiment is similar to the previous one in Section~\ref{sec:tpo2:box}. To further validate the bimanual grasping and transportation interface presented in Section~\ref{sec:tpo3:methods}, in this experiment a more challenging scenario (shown in \figurename{}~\ref{fig:tpo3:box_exp_setup}) is set. The objective is to pick a wooden block, transport it inside a plastic container, pick the container with the wooden block, and transport them to the final location. 
As in the previous experiment, the approaching and grasping phase of the two objects is performed autonomously by the robot, while the manipulability-aware motion generation explained in Section~\ref{sec:tpo2:manipControl} is exploited throughout the whole mission, with the weight $\boldsymbol{W}$ computed from the robot arm in the worst condition of manipulability. The approaching of the block and of the container is performed by detecting the \textit{ArUco} markers placed on objects.

\begin{figure}[H]
	\centering
	\includegraphics[height=0.24\linewidth,trim={0cm 0 5cm 2cm},clip]{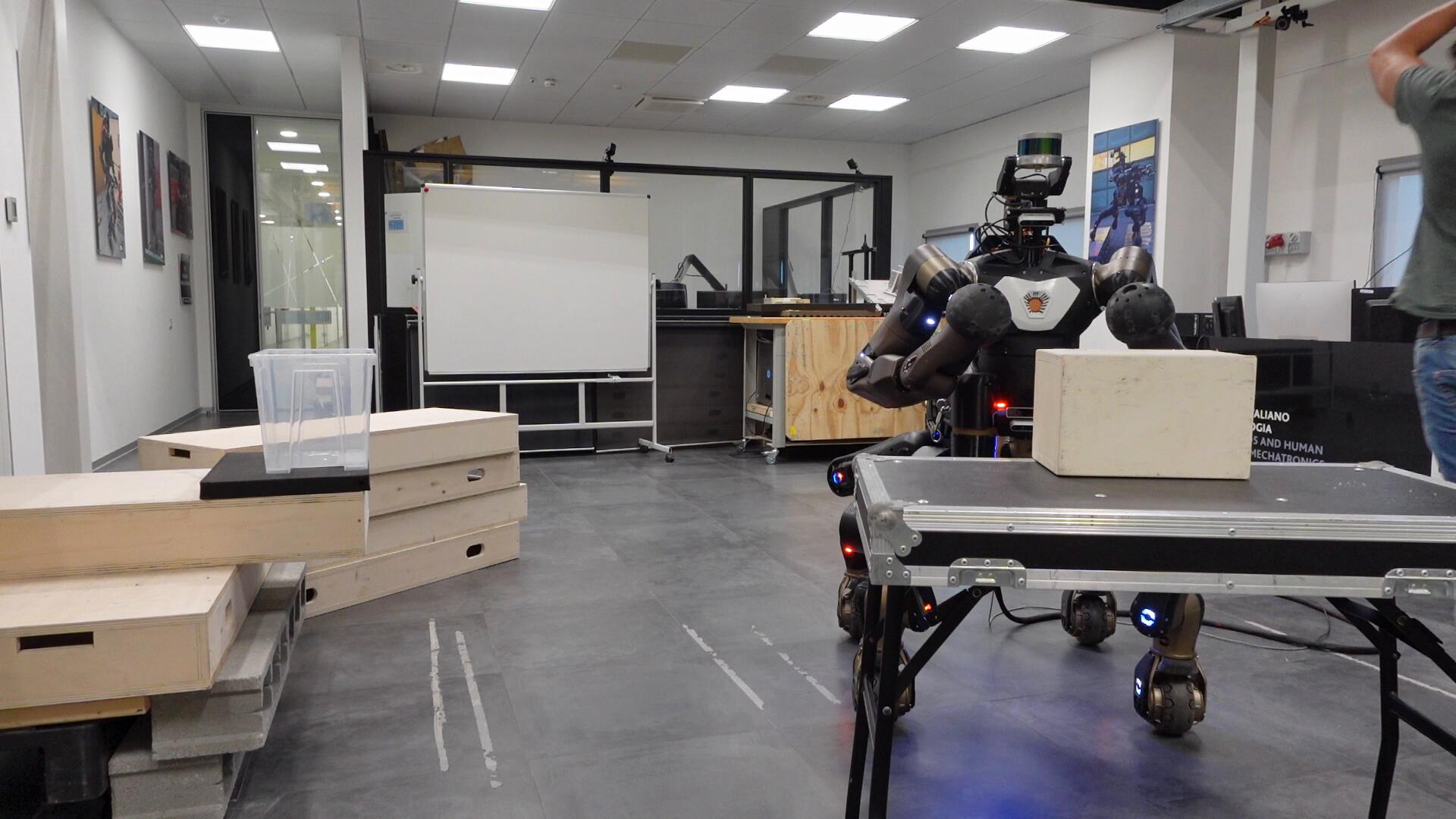}
	\includegraphics[height=0.24\linewidth,trim={0cm 0 5cm 2cm},clip]{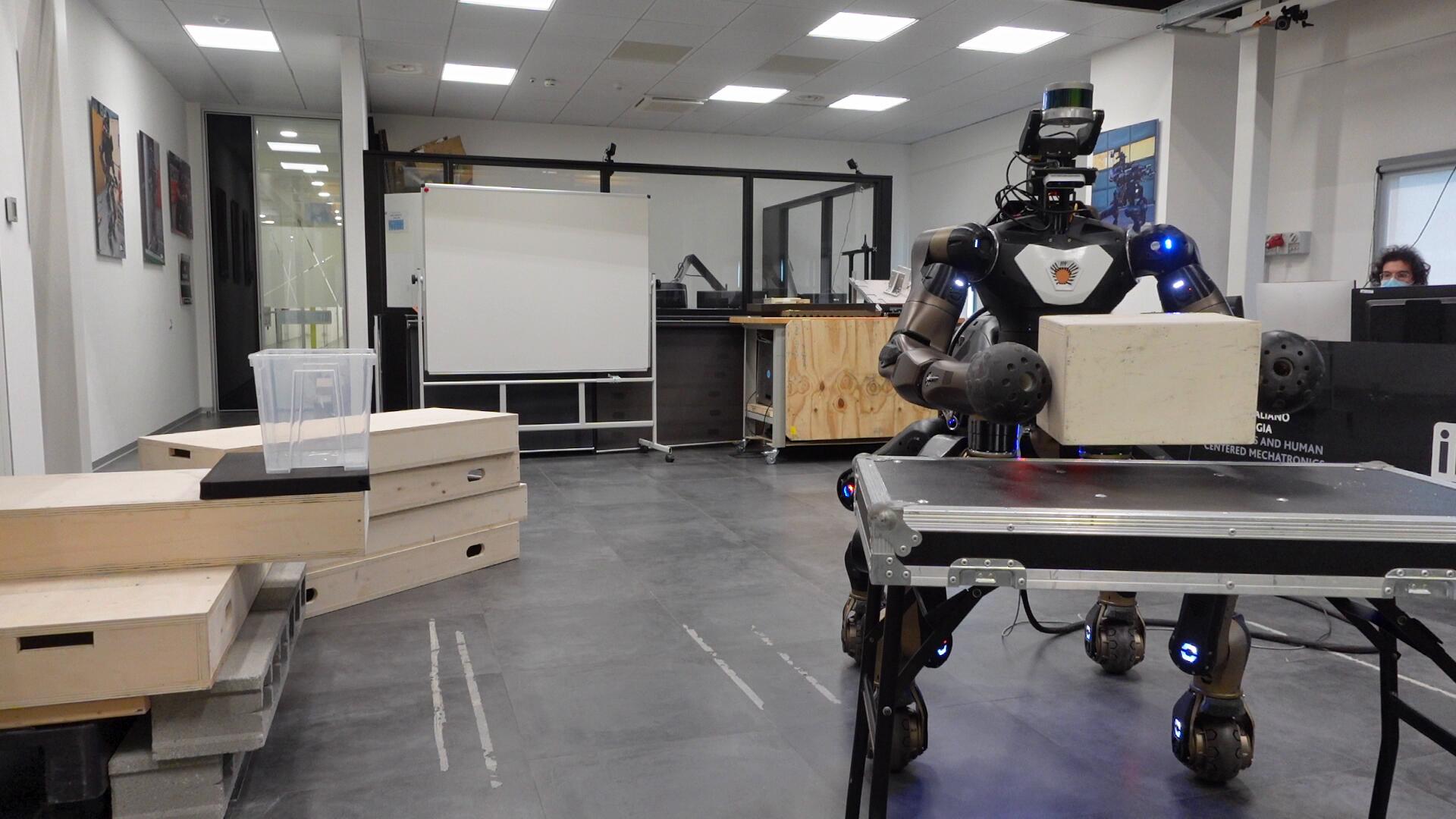}\\
	\vspace{2px}
	\includegraphics[height=0.24\linewidth,trim={10cm 0 12cm 3cm},clip]{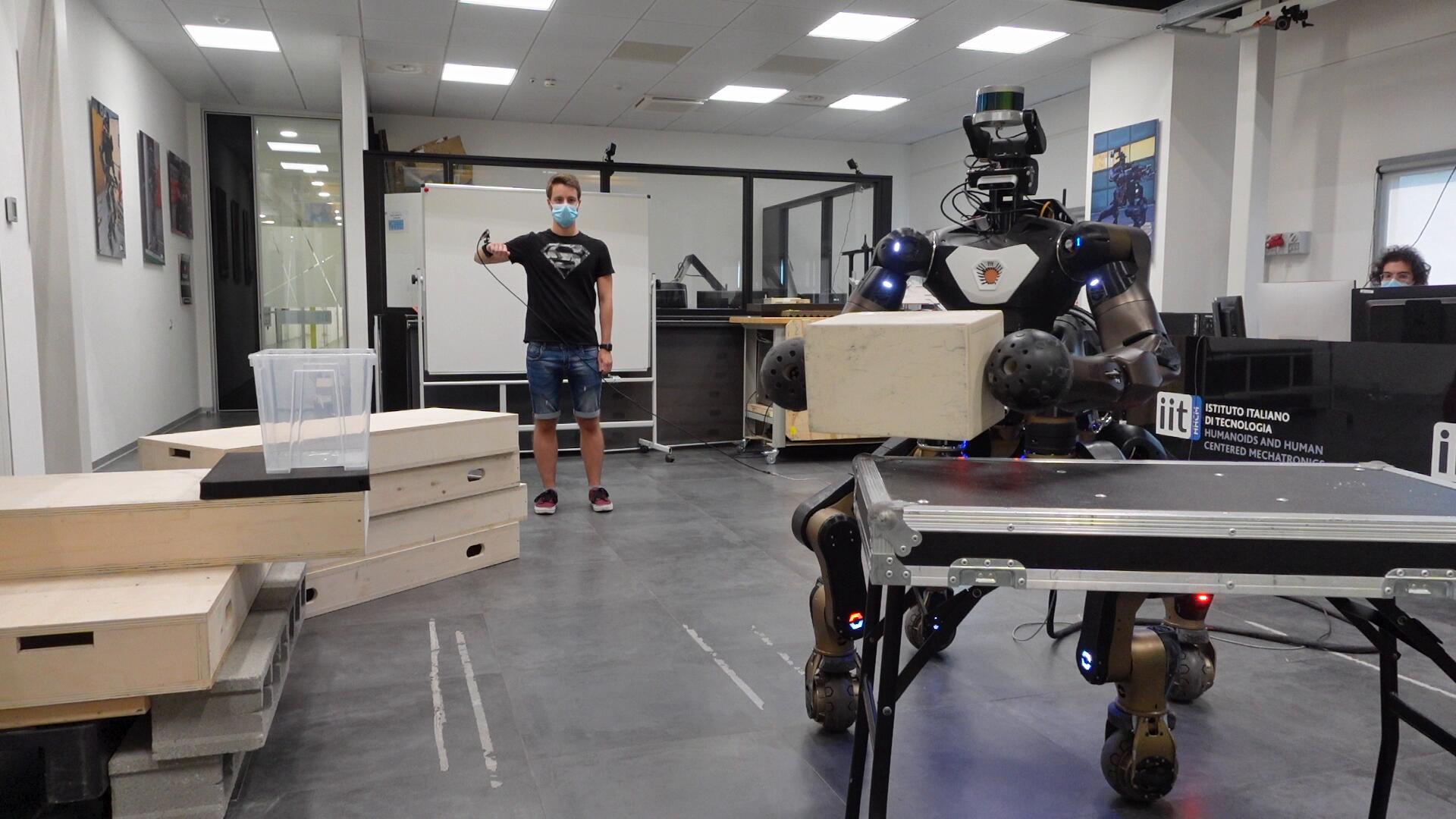}
	\includegraphics[height=0.24\linewidth,trim={10cm 0 12cm 3cm},clip]{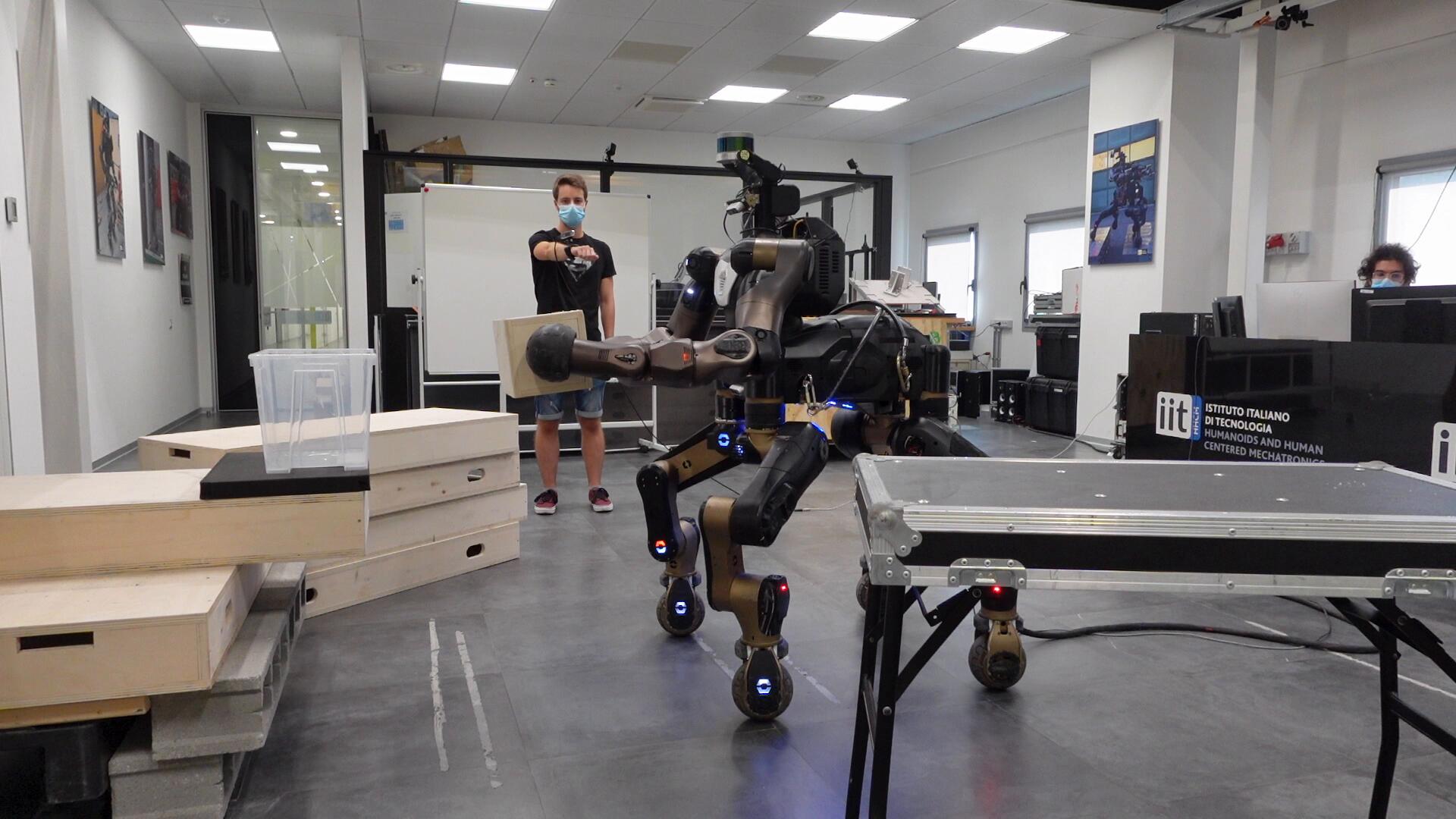}
	\includegraphics[height=0.24\linewidth,trim={10cm 0 12cm 3cm},clip]{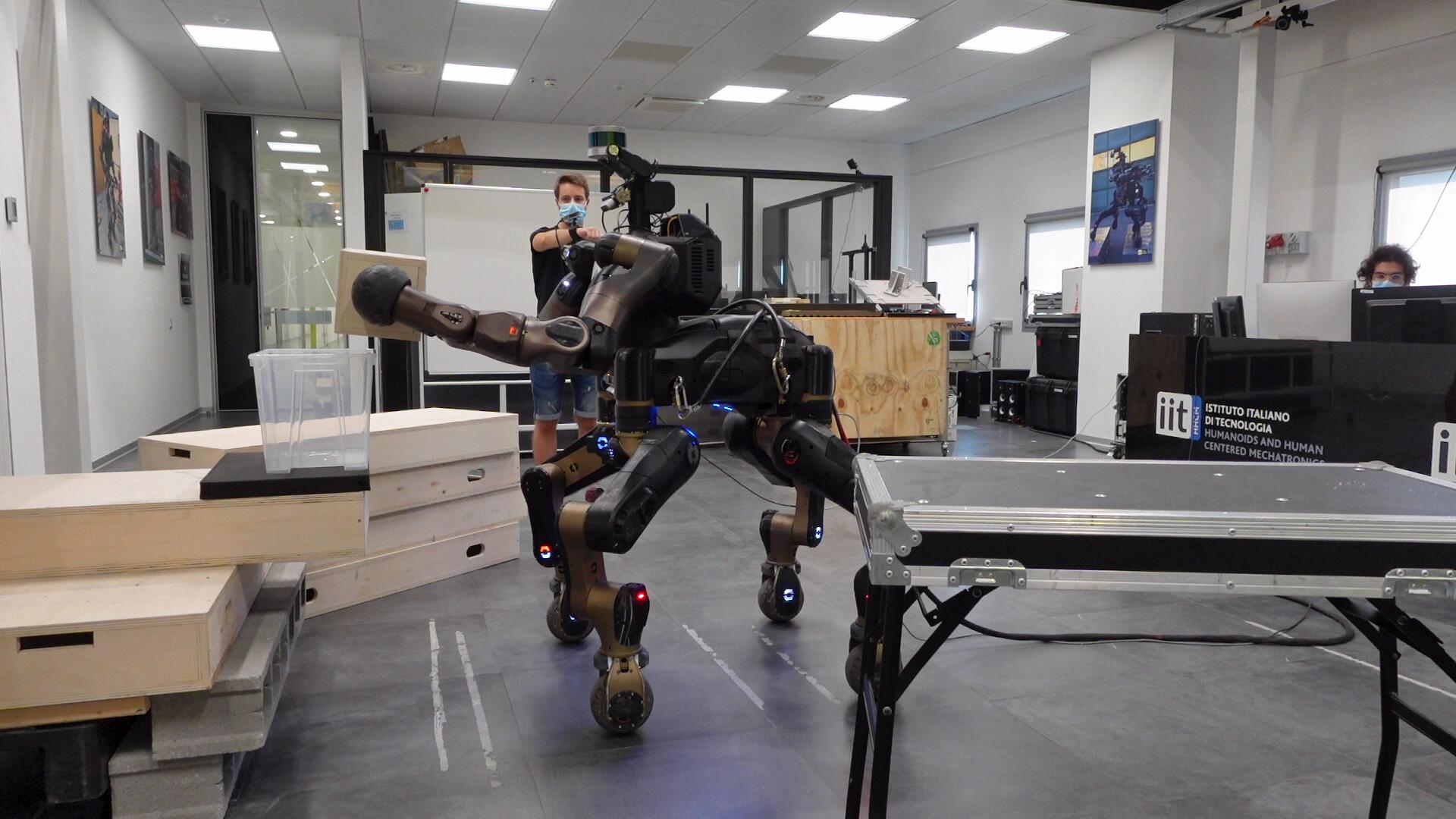}\\
		\vspace{2px}
	\includegraphics[height=0.24\linewidth,trim={10cm 0 12cm 3cm},clip]{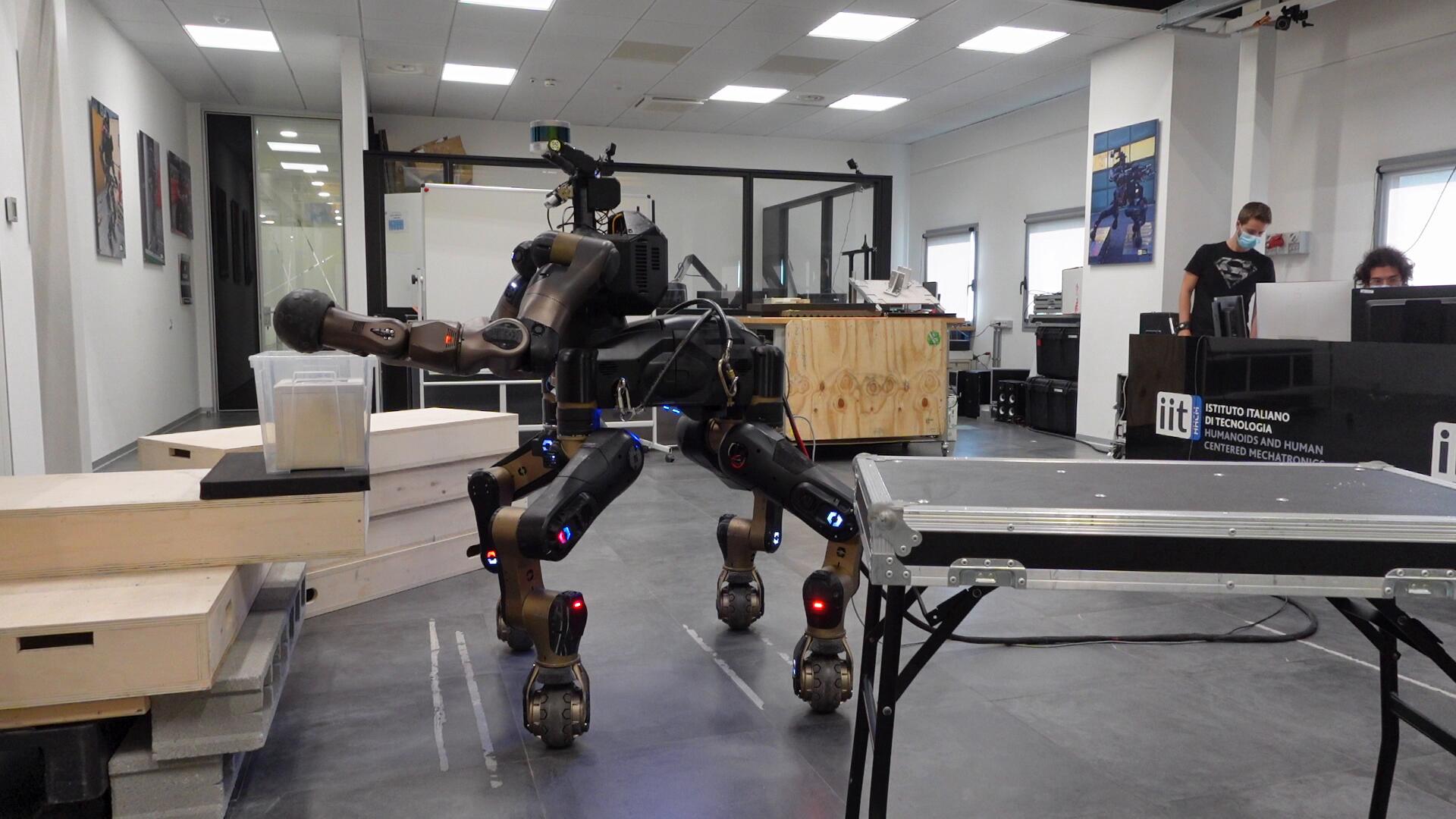}
	\includegraphics[height=0.24\linewidth,trim={10cm 0 12cm 3cm},clip]{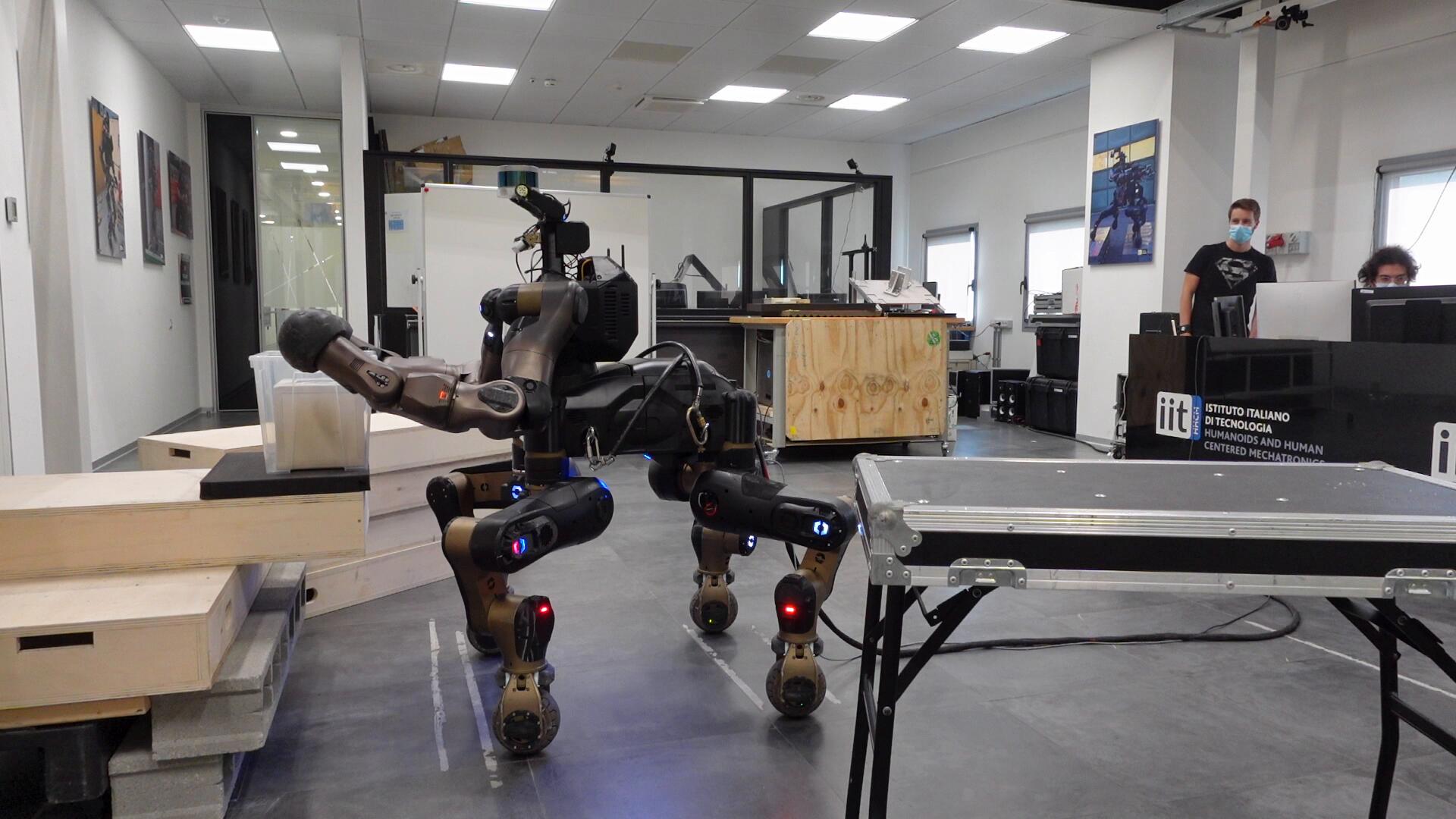}
	\includegraphics[height=0.24\linewidth,trim={10cm 0 12cm 3cm},clip]{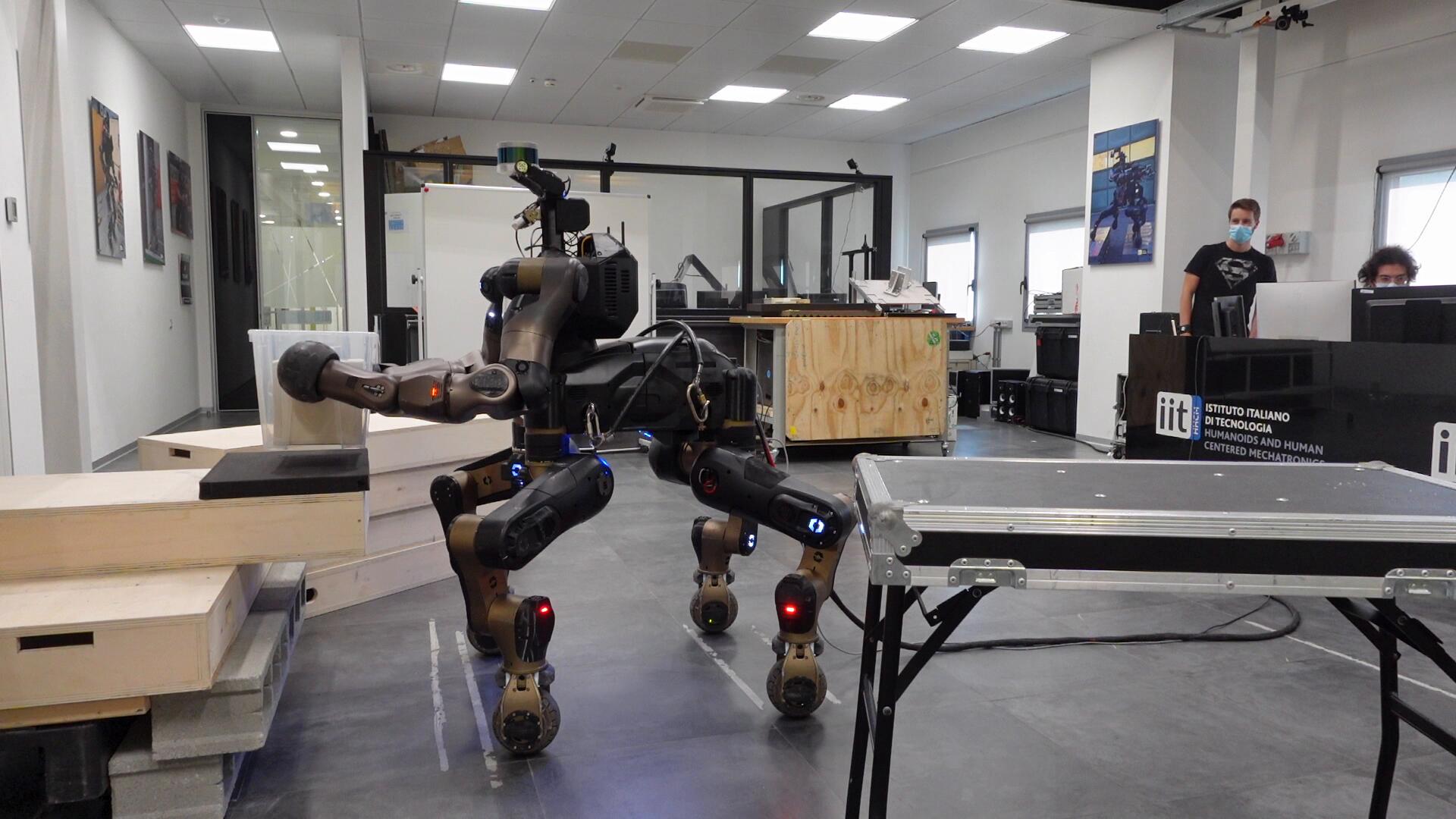}\\
		\vspace{2px}
	\includegraphics[height=0.24\linewidth,trim={10cm 0 12cm 3cm},clip]{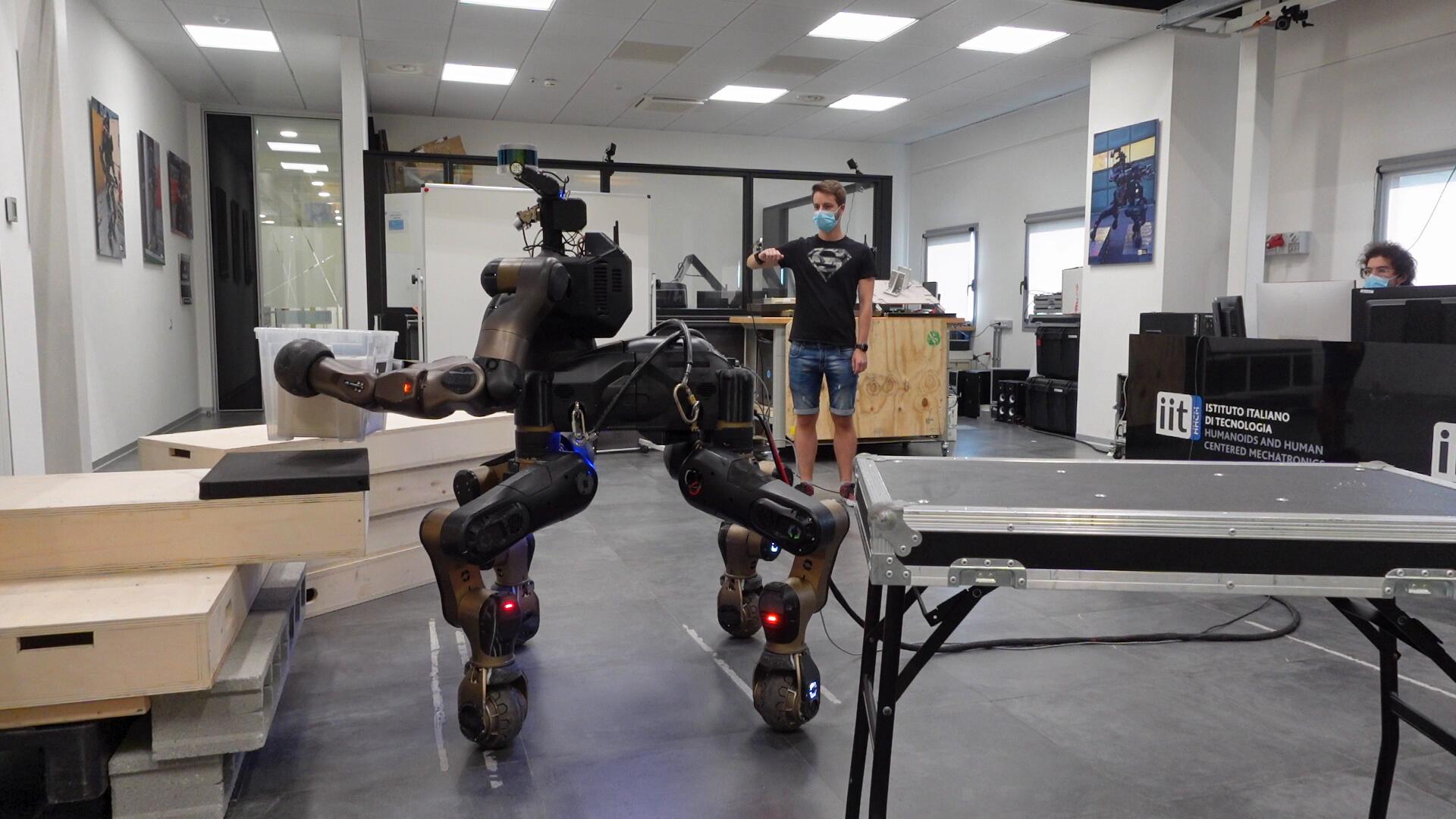}
	\includegraphics[height=0.24\linewidth,trim={10cm 0 12cm 3cm},clip]{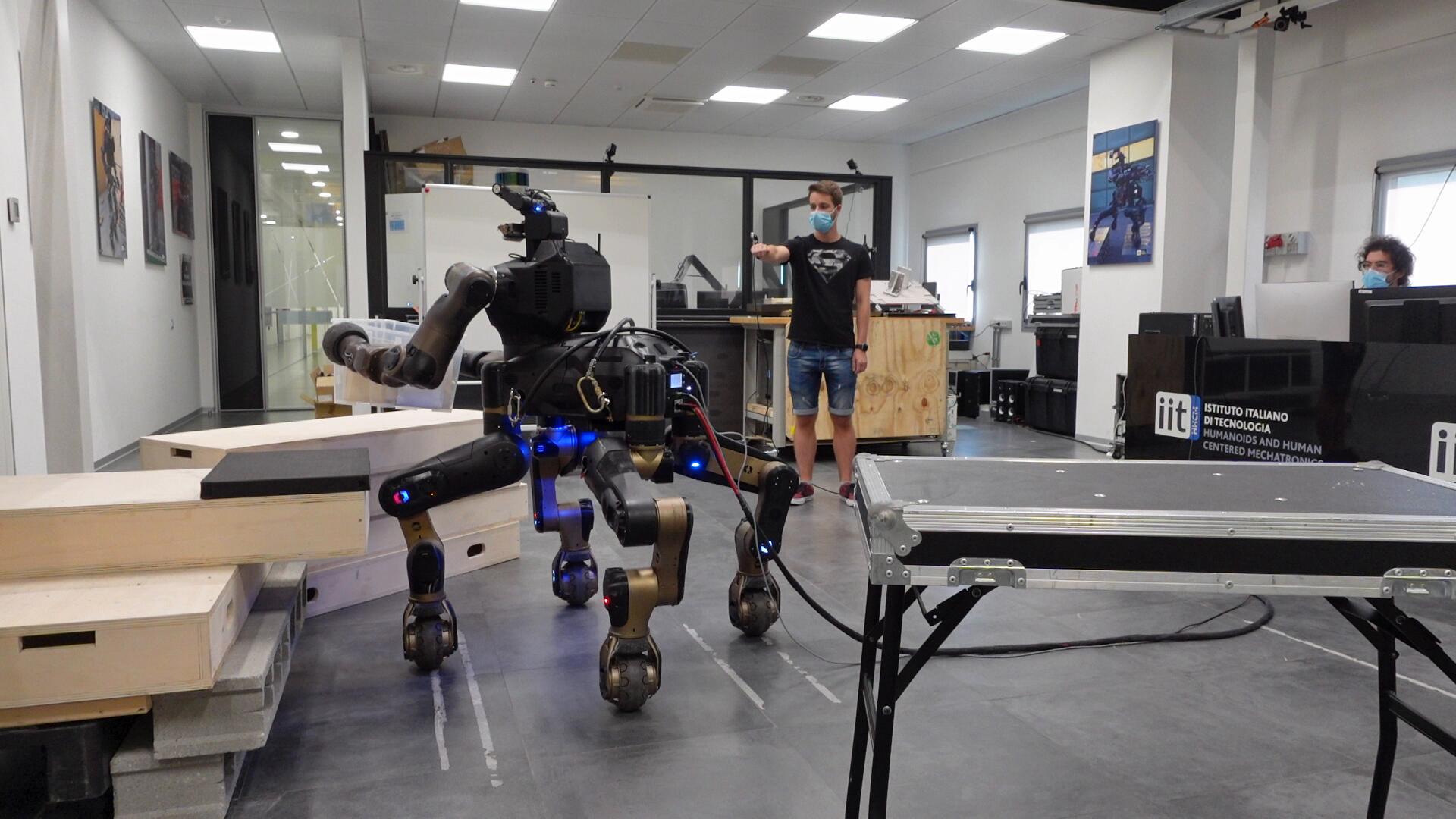}
	\includegraphics[height=0.24\linewidth,trim={10cm 0 12cm 3cm},clip]{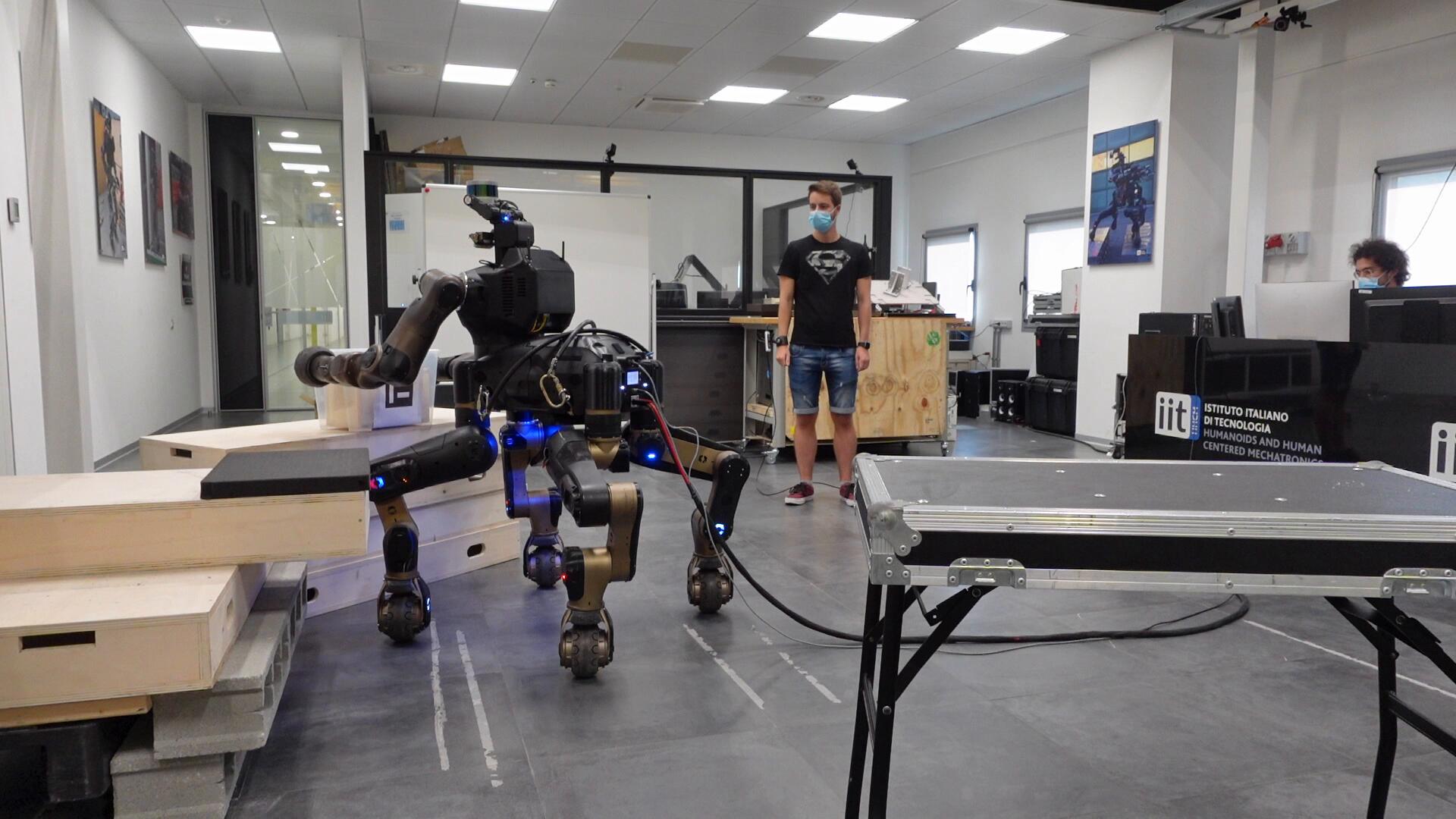}\\
	\caption[Autonomy enhanced \acrshort{tpo}: trasporting objects of different masses experiment sequences]{Sequences of the experiment about transporting object of different masses. From the top to the bottom row: (1) the CENTAURO robot autonomously approaches and grasps the block; (2) the user is teleoperating the robot to put the block inside the plastic container; (3) the robot autonomously grasps the plastic container; (4) the operator transports the container to the final location.}
	\label{fig:tpo3:box_exp_frames}
\end{figure}

The important moments of the experiment are visible in \figurename{}~\ref{fig:tpo3:box_exp_frames}.
In the first row, the robot autonomously reaches and grasps the object: once the two arms are placed to the sides of the block, this is picked up with an initial grasping force set to \SI{45}{\newton} applied to lift it firmly for the mass estimation phase. The robot estimates the mass of the object and derives the required grasping force $\bar{f}$, that is used to maintain the grasp during the transportation.
In the second row, the user teleoperates the robot to move the object inside the container. For large displacements of the mobile base, the user has the option to directly control the mobile base, as seen in the beginning of this phase. As the robot approaches the container, the user switches to commanding object velocities to place the object. The robot responds to these commands while maintaining the grasping force at the computed level.
In the third row, the autonomous pick-up and grasping force computation procedures are re-executed for the container with the block inside, hence dealing with a load of different mass.
In the fourth row, the robot is teleoperated again to transport the container to the final location.

By dealing with objects of different masses, the flexibility of the approach in adapting online to the different weights is demonstrated, permitting the transportation of multiple objects. 
During this experiment, the two object masses estimated with \eqref{eq:mass_est} resulted in \SI{2.111}{\kilo\gram} for the wooden block and \SI{2.671}{\kilo\gram} for the container with the block inside, with respect to a real value of \SI[separate-uncertainty = true]{1.958(2)}{\kilo\gram} and \SI[separate-uncertainty = true]{2.630(4)}{\kilo\gram}. 
For simplicity, the static friction coefficient $\mu_s$ has been set equal for the material pairs end-effector-wood and end-effector-plastic to $0.6$. The required grasping forces from $\bar{f}$ \eqref{eq:tpo3:optForce} resulted to be \SI{24.16}{\newton} and \SI{30.57}{\newton} for the two objects, considering a safety margin gain $k$ of $1.4$. In the control law of \eqref{eq:tpo3:coopLaw}, the parameters have been experimentally tuned to adjust the sensitivity of the robot generated motions: the diagonal elements of $\boldsymbol{D}$ and $\boldsymbol{K}$ have been set to $2500$ and $200$ respectively.
The rotation matrix $\boldsymbol{R}_b$ was updated at each control loop accordingly to the position of the two end-effectors, considering the frame \mbox{$<\!b\!>$} as a frame with the $\hat{y}$ axis lying in the line of conjunction of the two contact points.

\begin{figure}[H]
	\centering
	\includegraphics[width=0.8\linewidth]{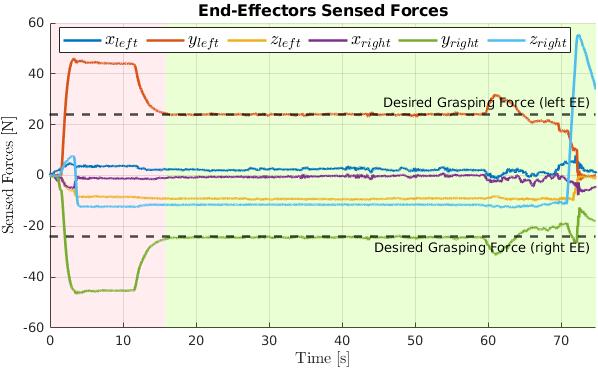}\\
	\vspace{20px}
	\includegraphics[width=0.8\linewidth]{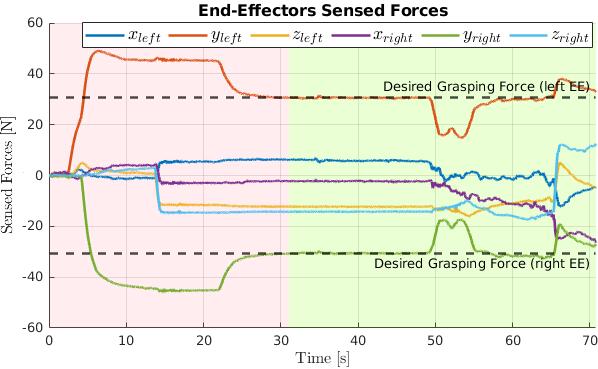}
	\caption[Autonomy enhanced \acrshort{tpo}: transporting objects of different masses experiment plots for \acrshort{ee}s sensed forces]{Left and right end-effectors sensed forces during the objects picking and transportation experiment. The top plot is referred to the wooden block, the bottom plot to the plastic container. The two colored areas in each plot represent the different phases: in the first pink area, the mass is estimated and the grasping force $\bar{f}$ is computed; in the second green area, the operator is controlling the robot. While teleoperating, the interface takes care of maintaining the grasping force to the desired value (represented by the black dashed horizontal line).}
	\label{fig:box_sensedforces}
\end{figure}

\begin{figure}[H]
	\centering
	\includegraphics[width=1\linewidth]{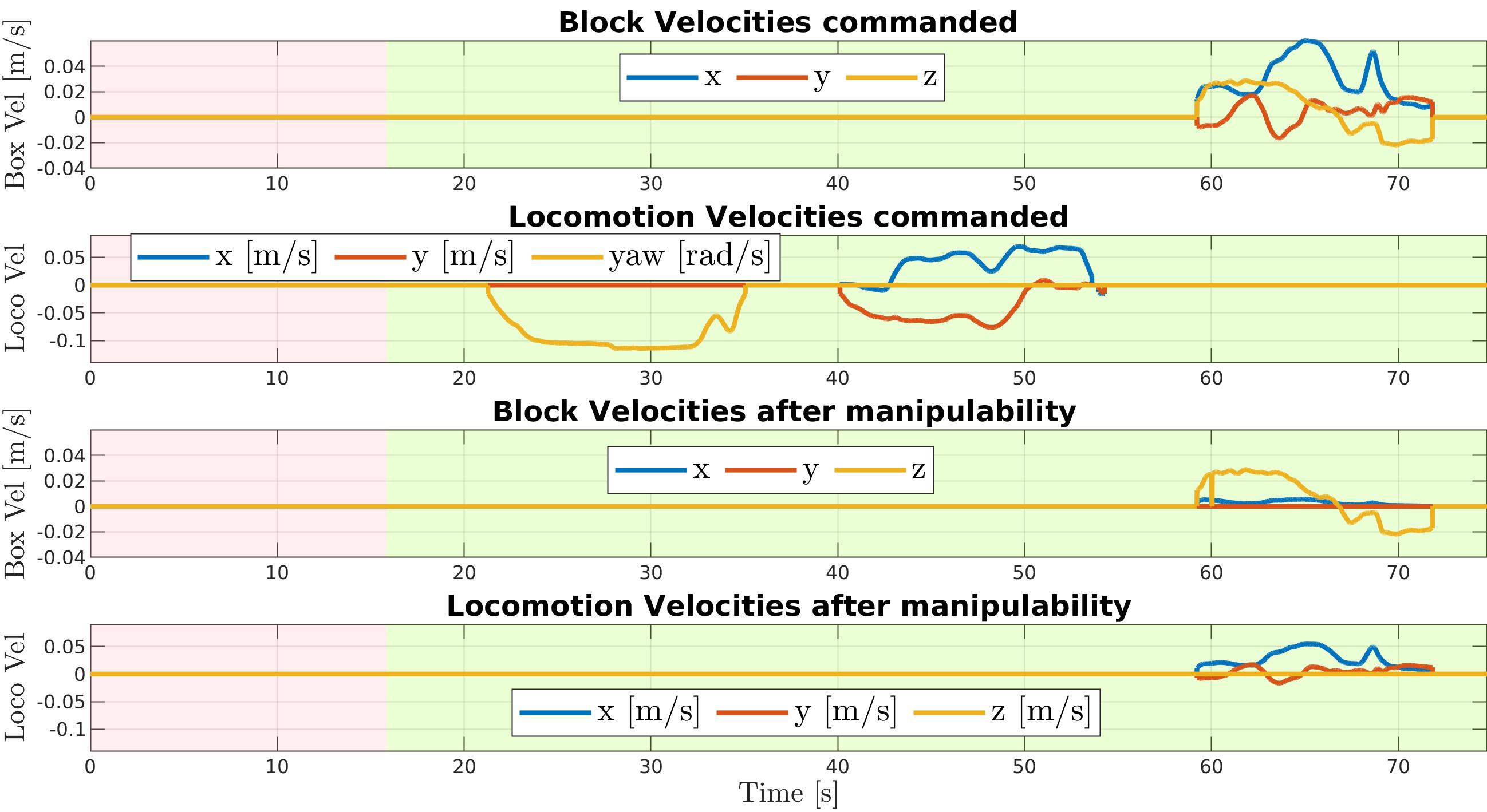}\\
	\vspace{20px}
	\includegraphics[width=1\linewidth]{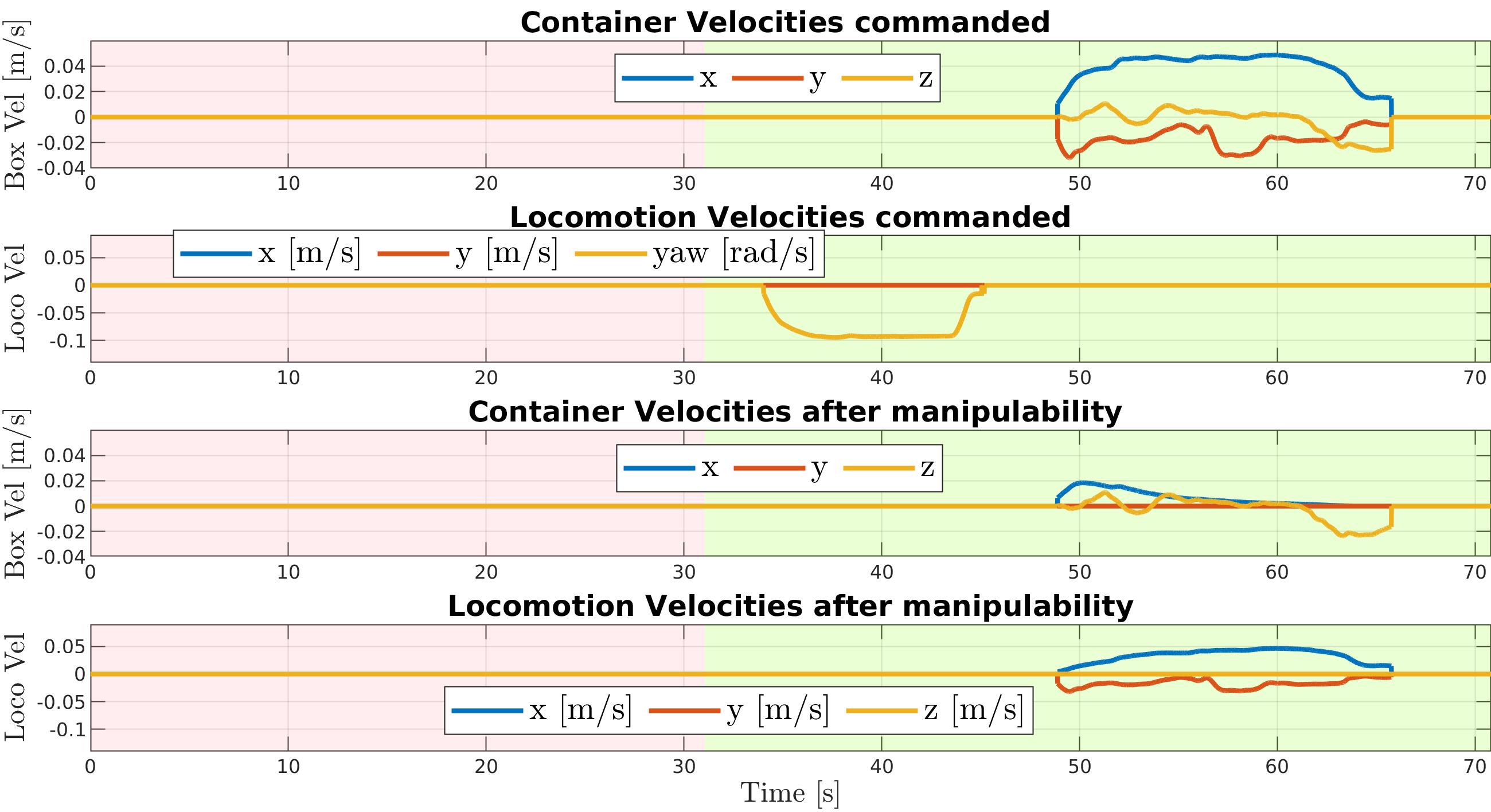}
	\caption[Autonomy enhanced \acrshort{tpo}: transporting objects of different masses experiment plots for user input data]{Inputs data for the objects picking and transportation experiment. Top and bottom groups of plots represent the data gathered during the pick and place sub-task of the wooden block and the plastic container respectively. The colored areas represent the different phases: in the first pink area, the mass is estimated and the grasping force $\bar{f}$ is computed; in the second green area, the operator is controlling the robot. In the first and second row they are displayed the operator commands for the object and for the robot locomotion, respectively. Accordingly to the arm manipulability level, the object velocities commanded (first row) are scaled down (third row) and locomotion velocity commands generated (fourth row).}
	\label{fig:box_inputs}
\end{figure}

\begin{figure}[H]
	\centering
	\includegraphics[width=1\linewidth]{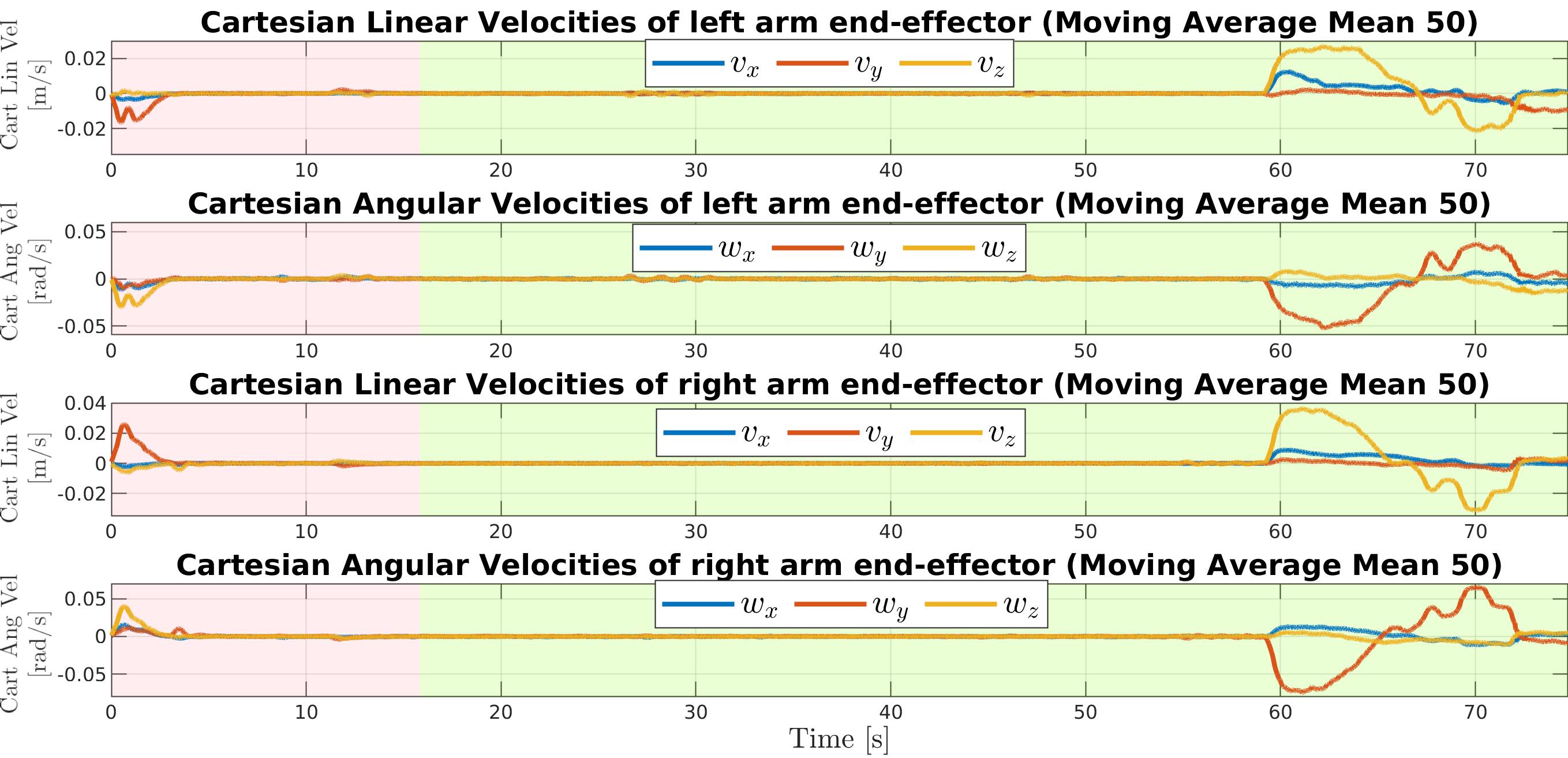}\\
	\vspace{20px}
	\includegraphics[width=1\linewidth]{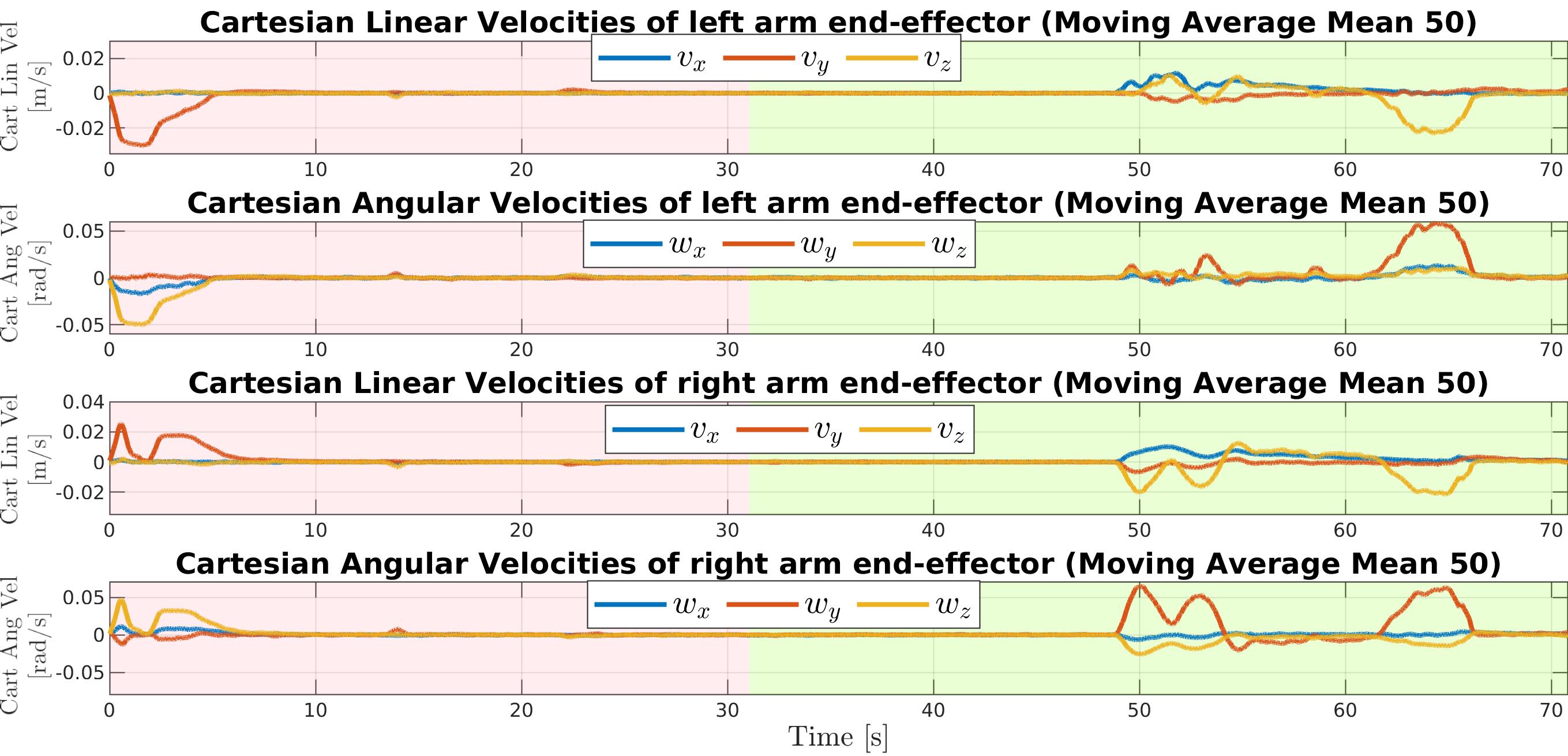}
	\caption[Autonomy enhanced \acrshort{tpo}: trasporting objects of different masses experiment plots for \acrshort{ee}s Cartesian Velocities]{End-effectors Cartesian velocities during the objects picking and transportation experiment. Top and bottom groups of plots represent the data gathered during the pick and place sub-task of the wooden block and the plastic container respectively. The colored areas represent the different phases: in the first pink area, the mass is estimated and the grasping force $\bar{f}$ is computed; in the second green area, the operator is controlling the robot. The data is computed from the sensed joint positions, hence filtered with a moving average to improve the visualization, reducing the noise.}
	\label{fig:box_jointCartVel}
\end{figure}

The relevant plots are shown in \figurename{}~\ref{fig:box_sensedforces}, \figurename{}~\ref{fig:box_inputs} and \figurename{}~\ref{fig:box_jointCartVel}. These figures follow certain similar conventions. The group of plots at the top refers to the grasping and transportation of the wooden block, whereas the group of plots at the bottom refers to the grasping and transportation of the plastic container with the wooden block inside.
All plots present two different colored areas: in the first, pink, area, the mass estimation and the grasping force computation are performed, while in the second, green, area the robot is teleoperated. 

In \figurename{}~\ref{fig:box_sensedforces}, the sensed forces at the two robot end-effectors $^{b}\boldsymbol{f}_{s,l}$ and $^{b}\boldsymbol{f}_{s,r}$ (rotated according to the object frame $<\!b\!>$) are shown. It can be noticed that the grasping forces ($y_{\mathit{left}}$ and $y_{\mathit{right}}$) in the first colored area reach the initial grasping force set (\SI{\pm45}{\newton}); whereas in the second colored area they are kept as much as possible equal to the required grasping force value $\bar{f}$ (\SI{\pm24.16}{\newton} for the block and \SI{\pm30.57}{\newton} for the container with the block inside). At the very end of the top plot, an increase in the sensed forces can be noticed (especially $z_{\mathit{right}}$). This is caused by contacts between the block and the container while the block is being placed. Other smaller deviations of the grasping forces from the desired references are due to the user object velocity commands, shown in the first rows of \figurename{}~\ref{fig:box_inputs}. 

In \figurename{}~\ref{fig:box_inputs}, the first two rows of each plot represent the user inputs. In particular, the first row shows the desired velocity for the transported object (block or container), while the second row shows the velocity command issued directly to the robot base. It can be noticed that, when object velocities are commanded, according to the manipulability level of the arms, these are scaled down (third row), and the locomotion velocities are generated (fourth row). 

In \figurename{}~\ref{fig:box_jointCartVel}, the Cartesian velocities of the end-effectors are plotted. Please note that these Cartesian velocities have been computed from the derivation of the robot joint positions, hence they have been post-processed with a moving average filter to reduce the noise and to improve the visualization for the reader.

\section{Conclusions}\label{sec:tpo23:conclusions}
This chapter has explored the key theme of robot autonomy, a crucial part in human-robot interfaces to enable a seamless interaction. The methods have been integrated in the \acrlong{tpo} architecture presented in Chapter~\ref{chap:TPO}, to enhance the interface with new robot autonomy modules. 

One of this autonomy module is the manipulability-aware shared locomanipulation motion generation, utilized to generate combined manipulation and locomotion motions in a shared control fashion while teleoperating a mobile manipulator.
The method distributes the operator's input between the manipulator and the mobile base according to the manipulability level of the end-effector. 
The approach does not involve the classical scalar manipulability measure, but the \acrfull{vtr} in the three principal directions.
This means that, when the end-effector's manipulability is low in certain directions, the user's input in such directions gradually generates mobile base velocities instead of arm motions.
Different thresholds can be set for different directions according to the task requirements. 
This results in the system generating more mobile base motions in these directions to maintain high end-effector's manipulability along them, while pursuing user's commands (Section~\ref{sec:tpo2:manipControl}). 

Another autonomy functionality has addressed the challenges in the teleoperation of robots for bimanually transporting objects of unknown mass. 
In the approaching phase, the robot is able to reach and lift the object autonomously. Similarly to how humans handle objects of unknown mass, the robot derives the required grasping force by estimating the object's mass once it has been lifted.
During the teleoperated transporting phase, the operator commands object directions without worrying about the arms motions. Indeed, the robot autonomously regulates the amount of grasping forces to prevent the object from falling and to not exert unnecessary high forces, while following the user's imposed directions (Section~\ref{sec:tpo3:methods}). 

These functionalities have been evaluated in a number of different tasks, involving the teleoperation of the CENTAURO robot.
Such validations have shown how these autonomy modules can assist the human operator, reducing his/her burden and improving the task's performance.
For example, by scaling the robot arm and mobile base motions transparently to the human operator, the need to switch between manipulator and mobile base control modes is eliminated. 
Furthermore, it has been demonstrated how safe object transportation tasks can be accomplished, combining the autonomous generation of arms and mobile base motions from the user's input about the object directions (Section~\ref{sec:tpo23:exps}).

Future works in these directions can explore more complex challenges like dual-arm non-symmetric tasks, or scenarios with additional constraints on arms motions, like keeping a desired object orientation.
More dynamic tasks can be addressed, such as the case of adding more objects inside a container while the robot is transporting it. Other kinds of bimanual tasks can be studied, e.g., for manipulating heavy tools that need at the same time to interact and to apply forces to the environment to accomplish the goal.

\part{Laser-guided Robot Control}\label{part:two}
\chapter{An Intuitive Tele-collaboration Interface Exploring Laser-based Interaction and Behavior Trees}\label{chap:Laser1}

\begin{figure}[H]
	\centering
	\includegraphics[width=0.7\linewidth]{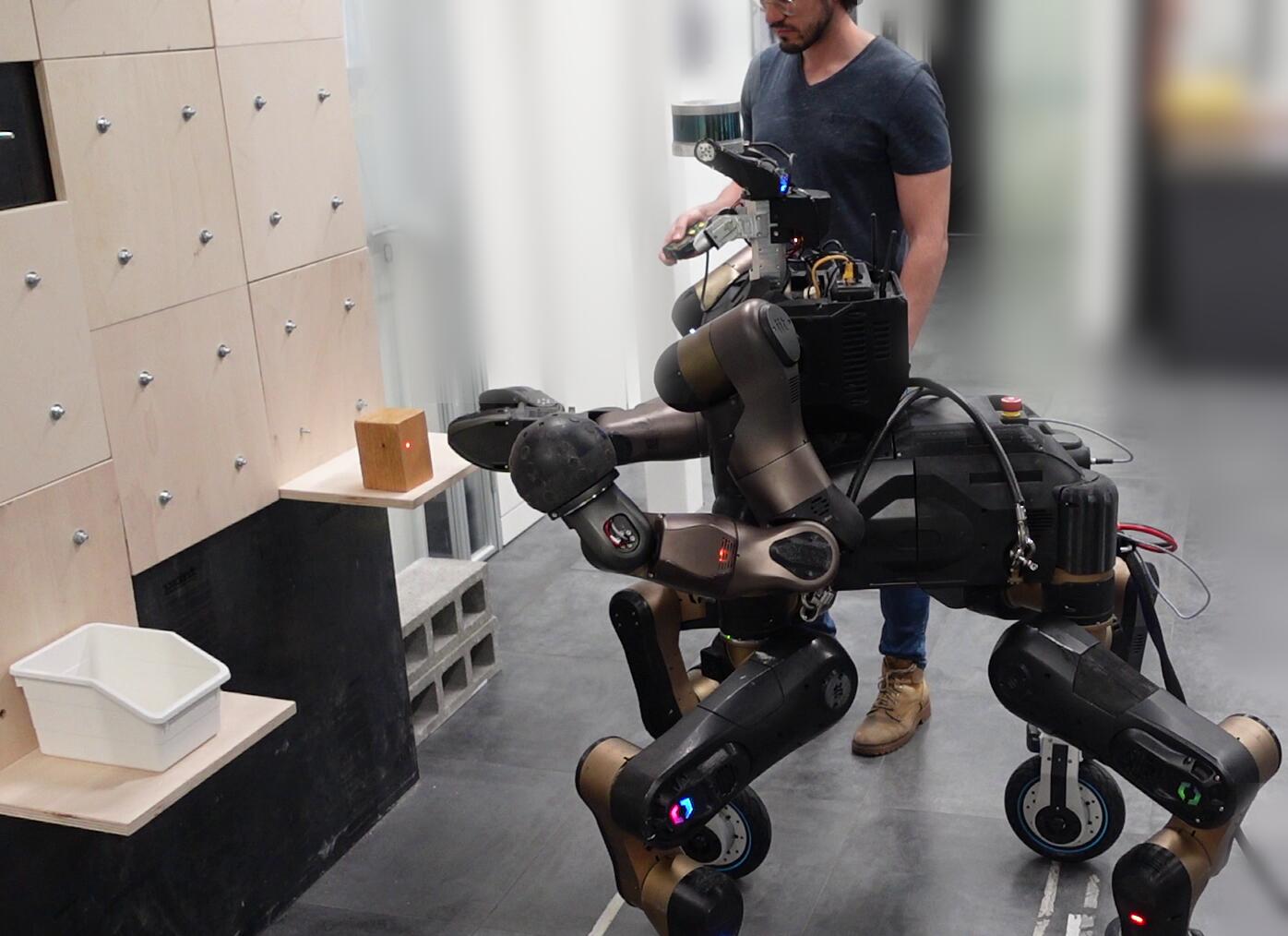}
	\caption[Laser-guided interface use case]{With the developed interface, the operator can effortlessly point the laser to guide to robot toward the object to grasp. This experiment is described in Section~\ref{sec:laser:grasp}.}
	\label{fig:laser:first}
\end{figure}

\lettrine{C}{onsidering} the key challenge of developing innovative, easy-to-use human-robot interfaces for commanding high capable robots, this chapter presents a laser-guided interface to enable a kind \textit{supervisory control}~\cite{Selvaggio2021} of a robot, hence intrinsically requiring an amount of robot autonomy (\figurename{}~\ref{fig:laser:first}).

This laser-guided interface is also employed in the next Chapter~\ref{chap:Laser2}, in the context of an assistive scenario.\\

\noindent This chapter is based on the following work:\\
\fullcite{LaserJournal}~\cite{LaserJournal}
\section{Introduction}\label{sec:laser:intro}

In Chapter~\ref{chap:chap3} we have seen that with the advent of increasingly highly-redundant robotic systems, there is a growing need for human-robot interaction interfaces and methods that must be intuitive and easy-to-learn. 
Following this challenge, this chapter presents a human-robot interface based on a very simple and effortless tool to provide commands to the robot: a laser emitter. This follows previous works which explored similar methods based on this input mean, as mentioned in Section~\ref{sec:soa:laser}.

The developed interface provides a visual servoing interaction that allows the human operator to effortlessly command even a highly-redundant robot. By pointing the laser to location of interest in the robot workspace, the user can instruct the robot to move to the selected location to execute an action.
This way of collaborating with the robot is very natural, mirroring how people naturally point to locations or objects of interest with their arms to communicate their intentions.

This kind of interface belongs to the field of \textit{supervisory control}~\cite{Gaofeng2023}, since the user does not command directly the robot motions as, for example, in the \acrlong{tpo} interface (Part~\ref{part:one}). Indeed, the user provides target locations that the robot must be able to reach someway, relying more on the robot intelligence to complete the task. To address this, the intuitiveness of the laser approach is combined with a certain level of robot autonomy, which permits the user to focus on the task rather than on how to deal with the complexity of generating the appropriate motions of the highly-redundant robot.

\section{Laser-guided Human-Robot Interface}\label{sec:laser:concept}

The architecture of the interaction interface employs a neural network-based laser spot detection and a \acrfull{bt}-based motion planner to reactively follow the point of interest indicated by the user. 

The visual guidance of the interface is based on the perception of the laser projected on a surface, provided by the robot's \acrfull{rgbd} camera. The neural network, fine-tuned for the application of detecting the laser's projection, extracts the 2D pixel coordinates from the \acrshort{rgb} images. These coordinates are then combined with the depth images to extract the laser's spot 3D position with respect to the robot reference frame. Details are given in Section~\ref{sec:laser:perception}.

While the goal is being tracked by the vision system, the robot exploits its own capabilities based on the plan implemented with the \acrshort{bt}. 
Given a target location indicated by the laser pointer, the robot is able to track it by exploiting its locomotion and manipulation skills accordingly to the \acrshort{bt} structure. As explained in Section~\ref{sec:soa:bttheory}, the \acrlong{bt} is a planning tool composed essentially by action nodes that are executed based on control flow and conditions nodes. Coming from computer games industry, its features have been recently employed in robotics, as recapped in the state-of-the-art Section~\ref{sec:soa:bt}. 
In this application, the action nodes of the BT model the various capabilities of the robot (such as locomotion, gaze, arm, and grasping actions), which are executed based on the conditions which monitor, for example, the distance between the robot and the goal indicated by the laser. 
The details about the \acrshort{bt}-based planner are given in Section~\ref{sec:laser:control}.

The combination of the fast detection of the laser projection and the responsiveness of the \acrshort{bt} equips the robot with the capacity to reactive adapt to changes in the goals as indicated by the user. Hence, it is possible not only to indicate a fixed goal but also a dynamic path that the robot is able to follow in real-time. 
While the laser spot is being tracked, the BT, based on its flow, runs the robot actions.
For example, for a manipulation task, the robot can approach a distant goal with appropriate locomotion abilities until such goal is near enough to be reached with manipulator motions. During such actions, if the goal is moved outside the arm's workspace again, the manipulation action is promptly halted in favor of the locomotion action. This management of locomanipulation motions is transparently addressed by the \acrshort{bt}. 

The architecture is flexible and modular, permitting to create with no effort new missions adapted to the capabilities of the robot in use.
Indeed, the developed interface is suitable to command different kind of robotic systems. 
When interacting with mobile robots, the user can point to environment locations to guide the robot through a desired path, repositioning continuously the laser if it is necessary because of a distant target or due to obstacles.
Similarly, when interacting with collaborative robot manipulators, the user can precisely direct the end-effector to a particular location of the robot workspace while also guiding the robot to avoid potential obstacles by indicating intermediate path points on the way to the final end-effector position. 
Another benefit is that the operator can effectively accommodate for errors of the final position of the tool control point (due, for example, to uncertainties in the robot kinematics/dynamics and perception). Indeed, the operator can compensate for inaccuracies by pointing the laser slightly off the desired position to eliminate any introduced errors in the end-effector position with respect to the task location of interest.

Furthermore, regarding the situation awareness, operators have a direct feedback of the goal issued to the robot through their visual perception of the laser projection, without the necessity of an additional mean like a display.

In summary, the main features and contributions of this laser-guided human-robot interface are:
\begin{itemize}
	\item The development of an intuitive method that enables a human operator to effortlessly guide even a highly-redundant robot by indicating point of interests in the environment with a laser pointer device. 
	
	\item The integration of a perception layer that detects the laser spot using a neural network solution. This results to be a faster and more robust solution compared to common computer vision techniques, permitting to promptly detect goal position changes, effectively enabling the real-time tracking of the laser spot.
	
	\item The modeling of the robot's behavior utilizes the \acrlong{bt} structure, offering a modular and reactive planning framework for the execution of robot actions based on goals received through the laser pointer interface. The implemented modules of the \acrshort{bt} enable the definition of new missions adaptable to the robot's capabilities with minimal effort. 
\end{itemize}
\section{Perception Layer}\label{sec:laser:perception}
\begin{figure}[H]
	\centering
	\includegraphics[width=1\linewidth]{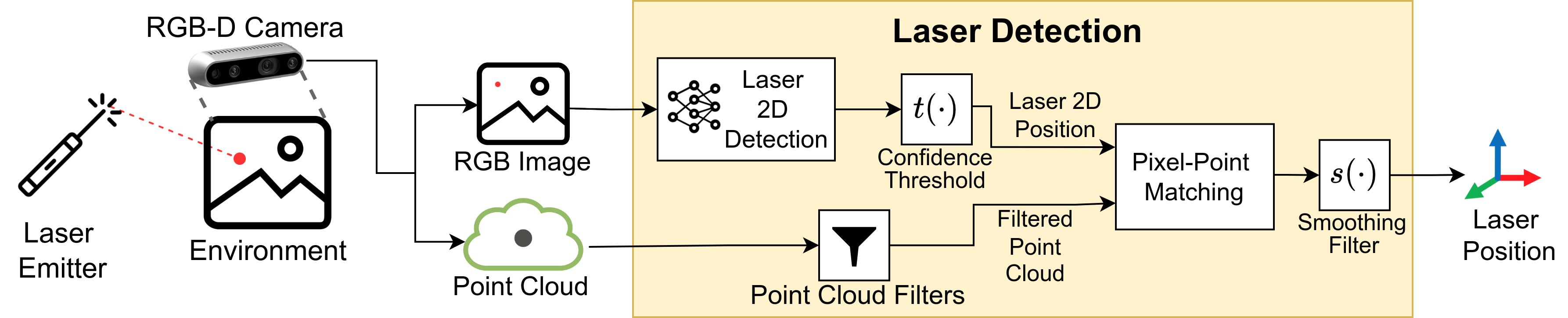}
	\caption[Laser detection logic scheme]{Logic scheme representing the perception layer. The \acrshort{rgbd} camera provides \acrshort{rgb} images and point clouds aligned to each other. The neural network extracts the 2D position of the laser spot, which is then matched with the point cloud to extract the 3D position of the laser spot.}
	\label{fig:laser:visionScheme}
\end{figure}

In this section the laser spot detection pipeline is explained, following the scheme of \figurename{}~\ref{fig:laser:visionScheme}. 
In summary, the input provided is the data from a \acrshort{rgbd} camera consisted in \acrshort{rgb} images and depth images (i.e., point clouds) aligned to each other. 
The \acrshort{rgb} image is processed by a neural network to extract the pixel coordinates of the laser spot. The extracted 2D coordinates are then matched against the correspondent point of the cloud, to output the 3D position of the laser spot.

\subsection{Laser Spot 2D Detection}
As mentioned before, the laser 2D position is gathered from \acrshort{rgb} images which are the given as input to a neural network.
Common state-of-the-art pre-trained neural networks are employed after fine-tuning them with a custom dataset containing images with the red laser projected against different surfaces.
The dataset consists in $385$ \acrshort{rgb} images with a resolution of $1280$x$720$ pixels, captured from an Intel Realsense D435 camera, which is the same camera used in the experiments. Each image has been manually annotated marking the bounding box of the laser spot, if present in the image. The annotated images are available at \cite{laserdataset}. The dataset has been divided into training subset ($80\%$) and validation subset ($20\%$). 
The framework employed for training and classification is PyTorch~\cite{pytorch}.
They have been considered three different state-of-the-art neural network models suitable for generic image classification and object detection. The first model is a Faster R-CNN with a ResNet-50-FPN backbone~\cite{fasterrcnn}. The other two are YOLOv5~\cite{yolov5} models, specifically the small and large versions. 
The fine-tuning has been performed on a system equipped with an \textit{Intel Core i9-10980XE} (18 core, 36 thread) \acrshort{cpu}, \textit{NVIDIA GeForce RTX 3090} \acrshort{gpu}, and \SI{64}{\giga B} \acrshort{ram}. Additional training specifications are shown in \tablename{}~\ref{tab:training}. The trained models are available at~\cite{laserModel}.

\begin{table}[H]
	\centering
	\caption[Laser detection fine-tuning parameters]{Parameters of the fine-tuning process of the neural network models.}
	\label{tab:training}
	\begin{tabularx}{.83\linewidth}{l c c c}
		
		\toprule
		Model & Epochs & Batches & Training Time \\
		\midrule
		
		\textbf{Faster R-CNN}~\cite{fasterrcnn} & 40  & 8 & 16min 9s \\
		\textbf{YOLOv5 small}~\cite{yolov5} & 200 & 8 & 16min 30s \\
		\textbf{YOLOv5 large}~\cite{yolov5} & 200 & 8 & 43min 47s  \\
		\bottomrule
	\end{tabularx}
\end{table}

Differently from the training part, in all the experiments conducted the inference has been run on a laptop equipped with an \textit{Intel Core i5-10300H} (4 core, 8 thread) \acrshort{cpu}, an \textit{NVIDIA GeForce GTX 1650s} \acrshort{gpu}, and \SI{16}{\giga B} \acrshort{ram}.
During the inference, the neural network provides multiple bounding boxes which possibly contain a laser spot. The bounding box with the highest confidence is considered, and, if the confidence is above a predetermined threshold, the center of the bounding box is picked as the 2D pixel corresponding to the laser spot. Nonetheless, since the laser spot is very small with respect to the whole image, it is irrelevant which pixel inside the bounding box is taken.

\subsection{Extracting the Laser Spot 3D position}

In the lower branch of the scheme of \figurename{}~\ref{fig:laser:visionScheme}, a cascade of point cloud filters is exploited to remove points which are of no interest, to reduce the amount of data to deal with. For example, points which are outside the robot workspace~\cite{pointcloudFilter} and points that correspond to the robot body~\cite{robotBodyFilter} can be eliminated.

In the \textit{Pixel-Point Matching} node, the matching of the 2D pixel and the 3D point is a trivial passage, since the \acrshort{rgb} image and the point cloud are provided aligned by the camera adopted. Indeed, given the whole point cloud, a specific point can be accessed using the pixel coordinates as indexes for the point cloud structure. 
To be sure that the \acrshort{rgb} and the point cloud data are synchronized, the \textit{Pixel-Point Matching} node checks their timestamps and discards the results if they are too different. 
Furthermore, to filter out possible disturbances and user erratic movements, there is a final smoothing filter $s(\cdot)$ to remove the high frequencies of the laser position changes, to provide a less chattering reference to the robot. 

\subsection{Laser Spot Detection Validation}
\begin{figure}[H]
	\centering
	\includegraphics[width=0.32\linewidth]{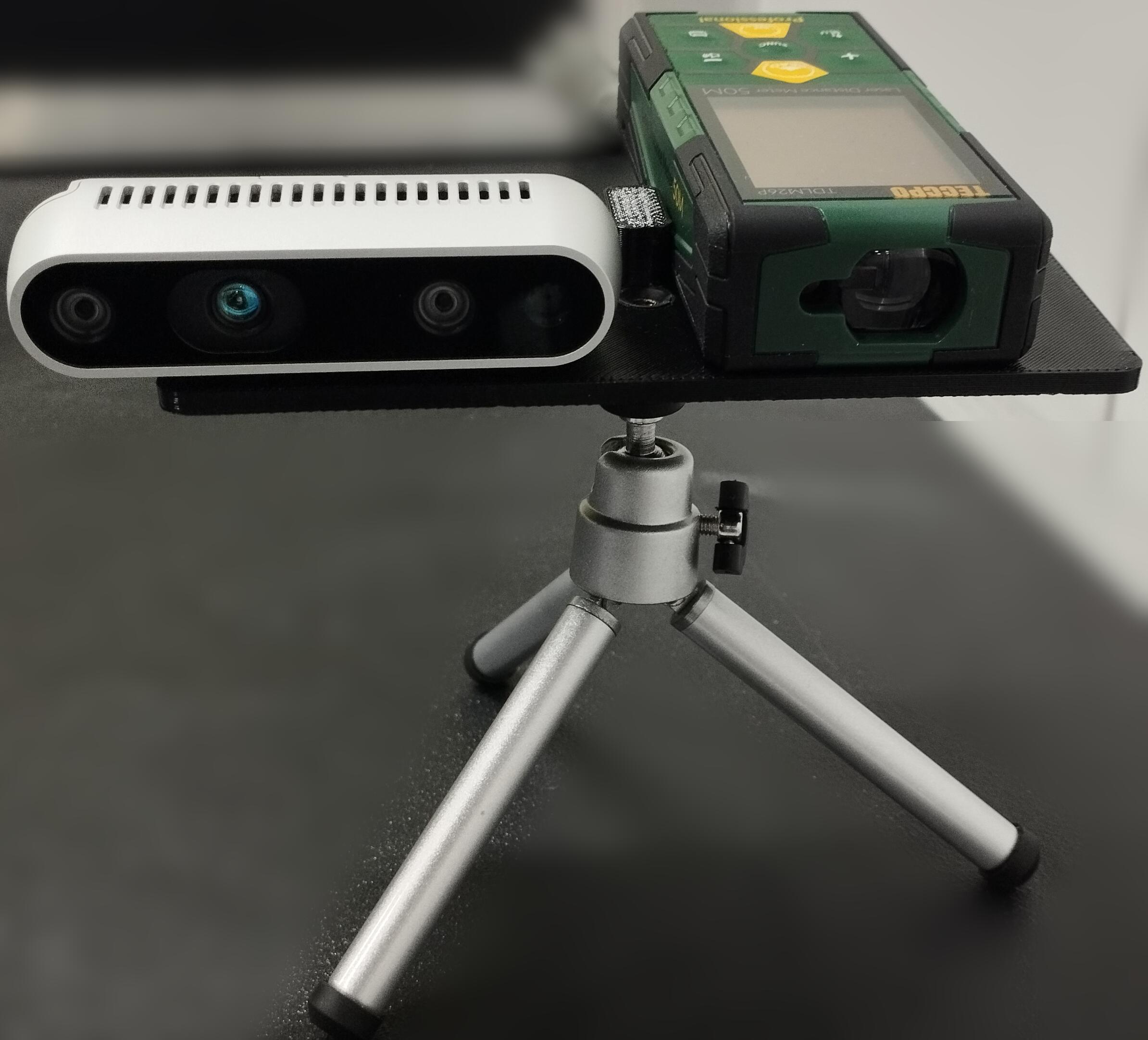}
	\includegraphics[width=0.32\linewidth]{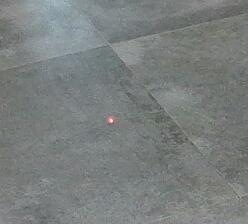}
	\includegraphics[width=0.32\linewidth]{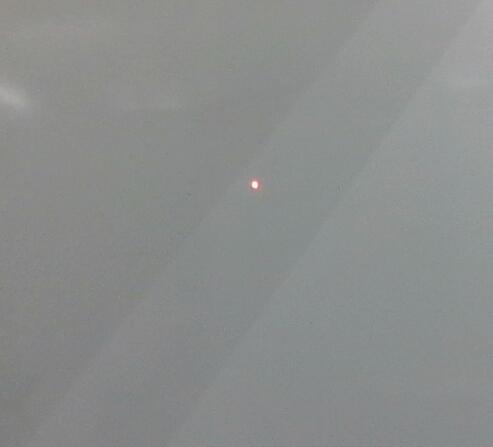}
	\caption[Setup for th laser spot detection evaluation]{The setting for evaluating the laser spot detection. On the left, the support for the depth camera and the distantiometer; on the center, the laser projected on the floor surface; on the right, the laser projected on the whiteboard surface.}
	\label{fig:LaserTest}
\end{figure}

For such a simple detection task, using a neural network may seem exaggerated, and classical computer vision methods might seem more appropriate. For example, as seen in some previous works presented in Section~\ref{sec:soa:laser}, the red spot can be detected by filtering the \acrshort{hsv} image to highlight only the correct color wavelength, and then some blob detection algorithm can be employed, like circle Hough transforms. The problem is that the dot is very small compared to the rest of the elements in the image, hence it can be easily mixed up with noise. To overcome this, a heavy filtering is necessary, resulting in fine-tuning parameters adapted only for the actual light-conditions and environments. 
Indeed, the color of the laser spot as seen by the camera can be slightly different when projecting the laser on surfaces of different materials and colors: this makes very difficult to generalize the parameters to filter out the elements on the image based on the HSV values. To increase the robustness some works process a sequence of images instead of a single one, but this significantly increase the detection time, and can be not sufficient as well.
It can also be noticed that algorithms that merge predictions and measurements like the Kalman filter or the particle filter are not helpful in this case, since in general the laser dot movements do not follow any pattern that can be predicted over time. 

Differently, as experienced in the conducted validations, the exploited neural network techniques resulted to be both fast and sufficiently robust for our cases.
In such validations, the laser spot positions provided by the laser detection framework are compared to the distance measures given by a commercial laser distantiometer. All the three neural networks explored (Faster R-CNN, YOLOv5 small, and YOLOv5 large) are considered for the laser 2D position detection part. 
A support is designed to have a known fixed relative position between the \acrshort{rgbd} camera and the distantiometer, as shown in the first image of \figurename{}~\ref{fig:LaserTest}.
The laser is pointed to two different surfaces, a gray-tiled floor and a whiteboard, as exemplified in the second and third images of \figurename{}~\ref{fig:LaserTest}. 
Three different distances are considered: short, medium and far, approximately measuring \SI{0.5}{\meter}, \SI{1}{\meter}, and \SI{2}{\meter}. Since the distantiometer provides only a scalar distance (not a 3D position), to have comparable data, a distance is computed as well from the extracted laser spot position.

\begin{figure}[H]
	\centering
	\includegraphics[width=0.88\linewidth]{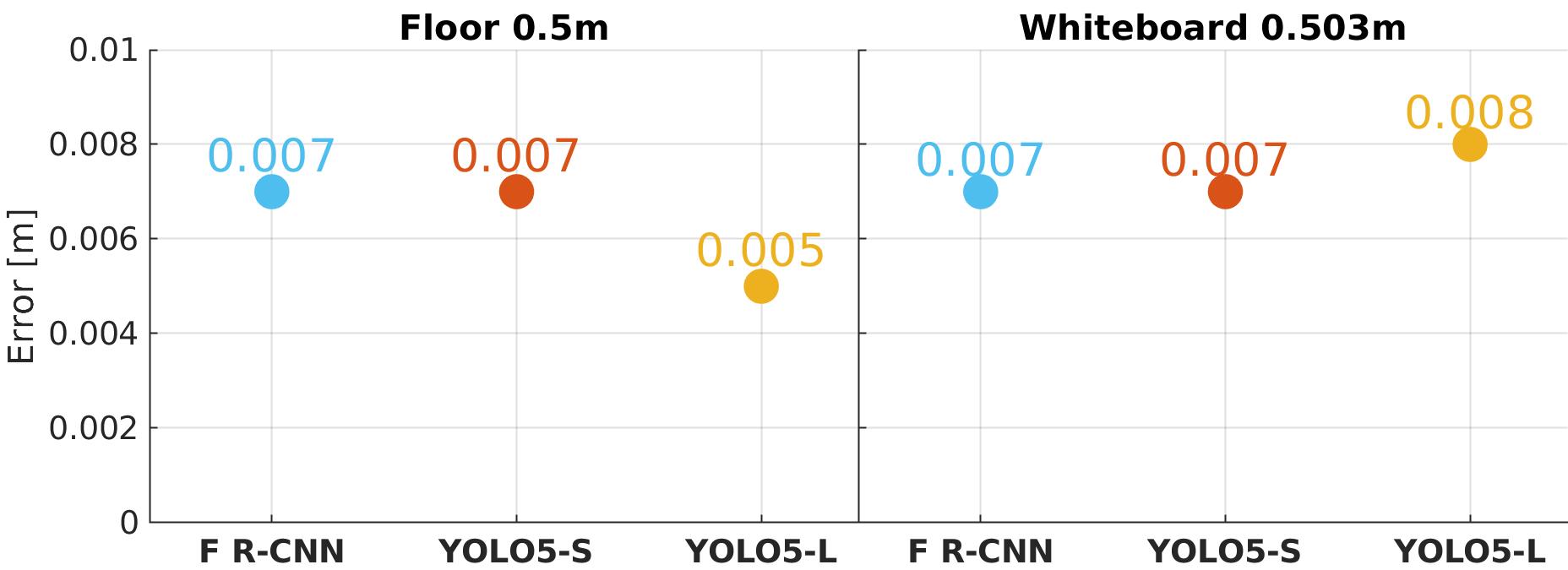}\\
	\vspace{20px}
	\includegraphics[width=0.88\linewidth]{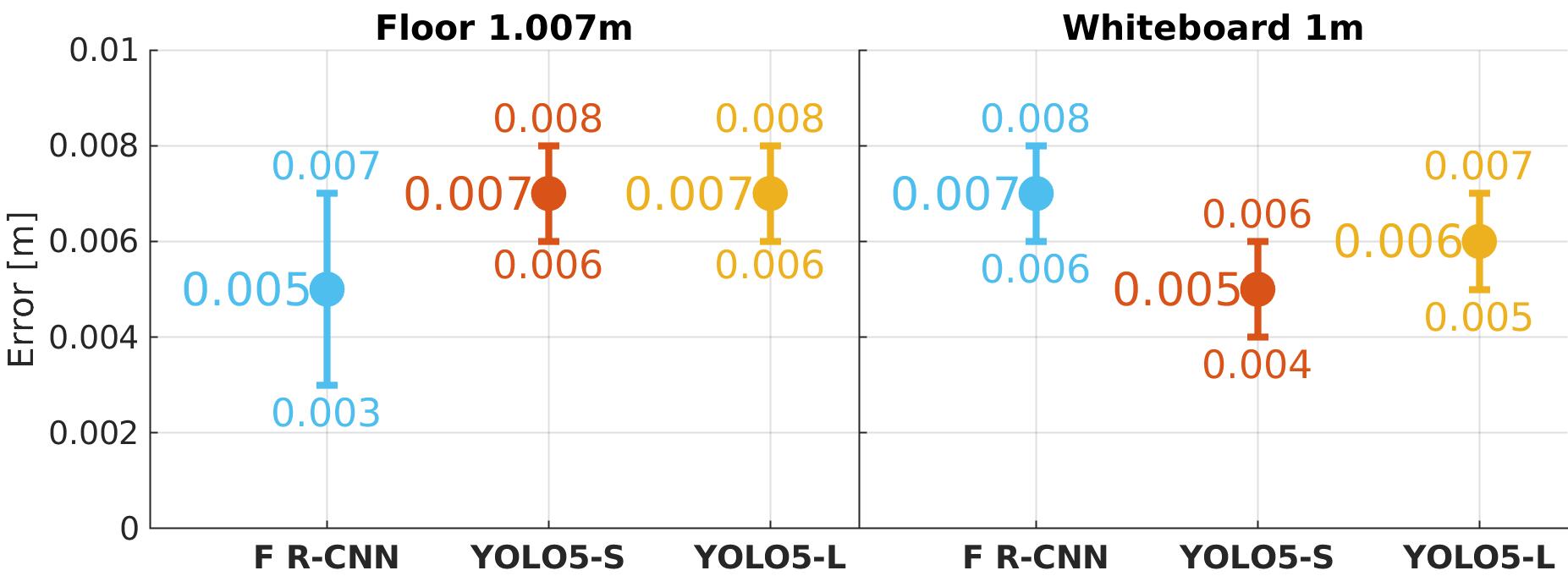}\\
	\vspace{20px}
	\includegraphics[width=0.88\linewidth]{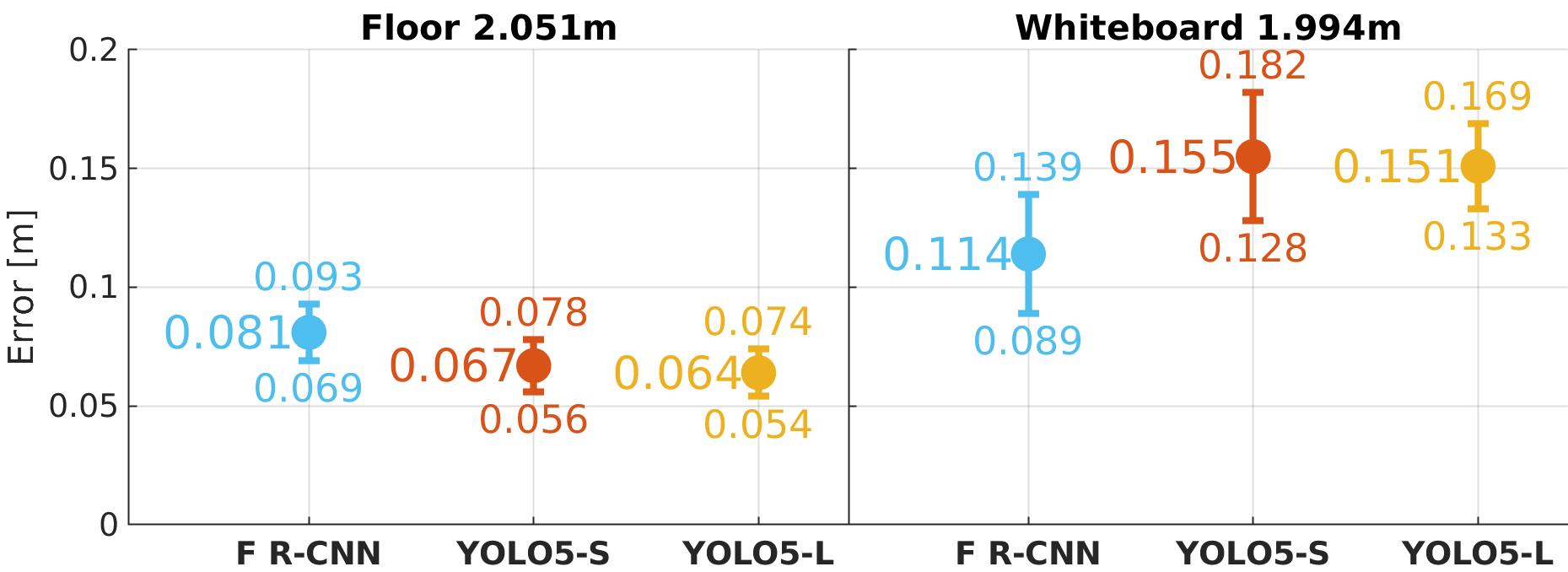}
	\caption[Laser detection validation: position errors]{Errors between the distantiometer data (in the titles) and the distance computed from the extracted laser position. The three different neural network models are compared with two different projection surfaces at three different distances.}
	\label{fig:LaserTestResultPrecision}
\end{figure}

As shown in \figurename{}~\ref{fig:LaserTestResultPrecision}, in terms of precision, the three neural network models are similar. As one may expect, with all models the precision and the accuracy decrease with the distance.
Since the laser projection was still well visible at \SI{2}{\meter}, and the bounding box still correctly positioned, the reducing precision probably comes from the Pixel-Point matching block, where the depth precision of the \acrshort{rgbd} camera plays an important role.

\begin{figure}[H]
	\centering
	\includegraphics[width=0.8\linewidth]{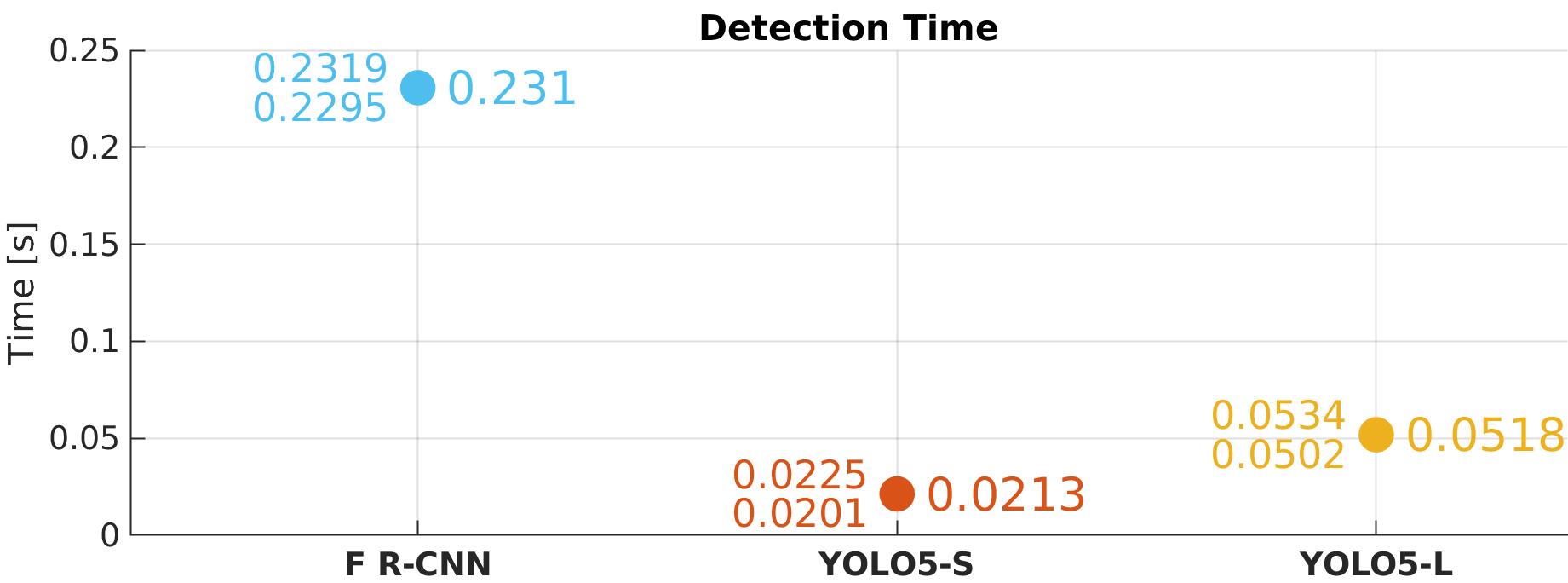}
	\caption[Laser detection validation: detection time]{Comparison of the time necessary to the neural networks to extract the laser 2D position from a single image.}
	\label{fig:LaserTestResultTime}
\end{figure}

Regarding the detection time, only the extraction of the 2D pixel coordinates is considered, since the computational time of the pixel-point matching block is neglectable, and anyway common to all the three neural network models.
In terms of this metric, the YOLOv5 models result better, as shown in \figurename{}~\ref{fig:LaserTestResultTime}. The plot refers to the time necessary to provide a result from a single image. In the case of the YOLOv5 small, it can be noticed that the detection frequency reached the frequency at which the images are captured by the camera used in the experiments ($30Hz$).
Instead, the Faster R-CNN model is the slowest, requiring approximately ten times more time than the YOLOv5 small. Nevertheless, all models are fast enough in the detection, enabling in practice a \textit{tracking} of the laser, allowing the robot to reactively follow it.
 
Regarding robustness, the confidence of false positives is sufficiently lower than the one of the true positives. This has permitted to set a robust confidence threshold value. The more relevant false positives consist in the incorrect detection of small objects that reflect light, like small screws, or that emit light, like \acrshort{led}s. 
From these tests and from the others conducted with the robot, it was observed that the slowest model (Faster R-CNN) is more robust in the sense that the difference between the confidence of true and false positives is greater. The fastest model (YOLOv5-S) is the one which with this difference is smaller. 
The observed results about precision, speed, and robustness were expected, given the characteristics of two-stage and single-stage detectors, to which Faster R-CNN and YOLO belong, respectively~\cite{Zaidi2022}.

The software of the laser spot detection pipeline, the trained neural network models, and the annotated image dataset utilized for training are available open-source at \href{https://github.com/ADVRHumanoids/nn_laser_spot_tracking}{https://github.com/ADVRHumanoids/nn\_laser\_spot\_tracking}~\cite{laserDetection}.

\section{Motion Generation Layer}\label{sec:laser:control}

\begin{figure}[H]
	\centering
	\includegraphics[width=1\linewidth]{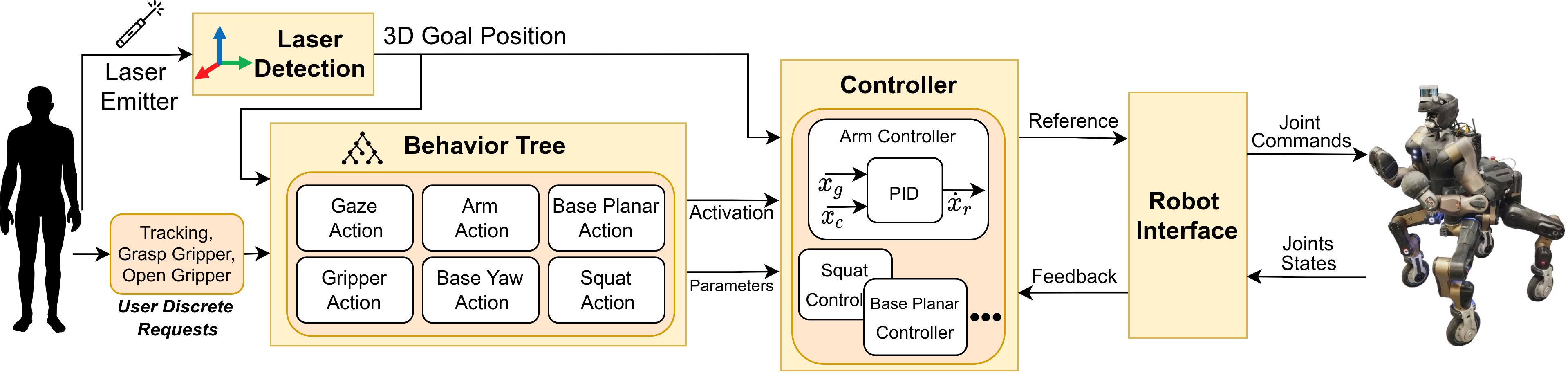}
	\caption[Laser-guided interface general scheme]{Logic Scheme representing the motion generation layer. Based on user inputs and the \acrshort{bt} structure, the actions relative to the robot abilities are activated. The Controller block generates references for such actions, commanded through the Robot Interface.}
	\label{fig:laser:controlScheme}
\end{figure}

This section details the control layer which generates robot motions from the user's input commands.
The scheme of such layer is depicted in \figurename{}~\ref{fig:laser:controlScheme}. 
In the considered context, the goal is commanded by the user through pointing the laser device, but, in general, it can be provided by any other source.

The robot locomanipulation capabilities are modeled with action modules, resulting in actions like Arm action, Base action, and Squat action, represented inside the \acrlong{bt} block of \figurename{}~\ref{fig:laser:controlScheme}.
Based on the position of the goal with respect to the robot, these action modules (modeled as leaf action node of the \acrshort{bt}) are activated and deactivated, following the flow of the \acrshort{bt}.
The activation of each module is linked to a specific function of the Controller block, which, expect for the gripper actions, is a \acrshort{pid} controller which generates a reference velocity $\boldsymbol{\dot{x}}_r$ based on the laser position and the parameters of the action module. For gripper actions, a simple discrete open/close command is issued.

For the sake of the clarity, in what follows, we will refer to the locomanipulation capabilities of the CENTAURO, the robotic platform employed in the experimental validations presented. 
Thus, the action modules employed consider CENTAURO-like capabilities, such as single arm manipulation, grasping with a 1-\acrshort{dof} gripper, wheeled locomotion, and particular body-legged movements like squatting.
Nevertheless, the concepts and the framework are general and adaptable to any kind of robot.

\subsection{Behavior Tree Based Planner}\label{sec:laser:bt}

As anticipated, the execution of the locomanipulation abilities of the robot are modeled through action module \acrshort{bt} nodes, detailed in Section~\ref{sec:laser:btactionmodule}, whose activation is governed by the flow of the \acrlong{bt}, as explained in Section~\ref{sec:laser:btflow}.

The \acrshort{bt} is implemented with the \texttt{BehaviorTree.cpp} library~\cite{behaviorTree}, integrated with the Groot2~\cite{groot2} graphical interface for editing, monitoring and visualizing the BT structure. The communication among the various software modules happens through \acrshort{ros}~\cite{ROS}.

\subsubsection{Behavior Tree Action Module}\label{sec:laser:btactionmodule}
\begin{figure}[H]
	\centering
	\includegraphics[width=0.55\linewidth]{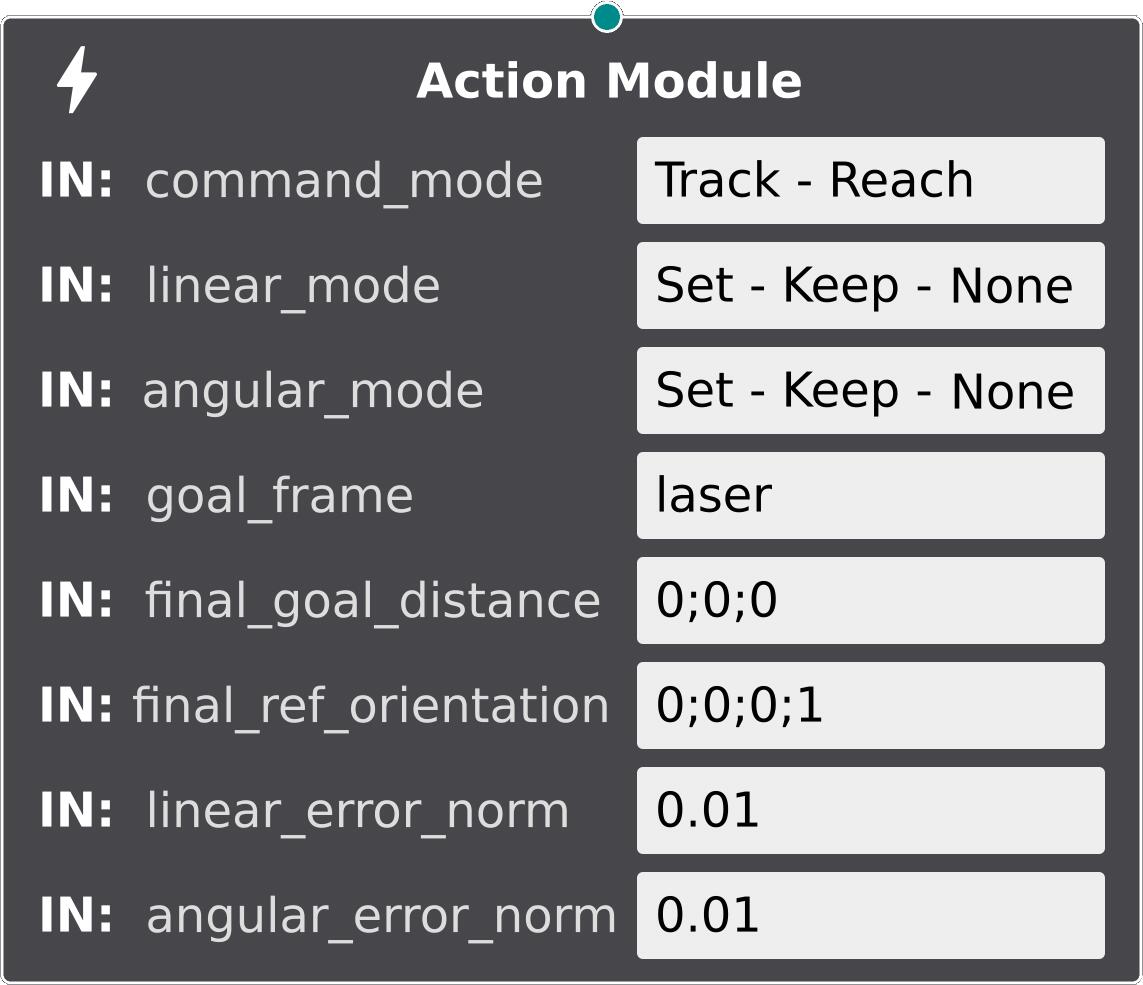}
	\caption[BT action module]{A generic Action Module, with all the settable parameters.}
	\label{fig:laser:nodeBT}
\end{figure}

An action module is a BT leaf node associated with a particular robot capability, like Arm action, Base action, Squat action, and so on. 
Once triggered by the tick of the \acrshort{bt}, this action requests to the Controller of \figurename{}~\ref{fig:laser:controlScheme} to activate a relative robot capability, i.e., it requests to start generating a command reference and sending it to the Robot Interface. As visible in \figurename{}~\ref{fig:laser:nodeBT}, some parameters, explained later, can be specified in the request. Hence, the purpose of the action module is to activate the reaching or tracking of a goal with the relative robot ability.

The \acrshort{bt} action node maintains its state to \texttt{running} until the Controller responds with a specific signal, whose nature depends on the module. 
Upon receiving the signal, the node returns \texttt{success}. If any failures occur (e.g., bad communication with the Controller) the node returns \texttt{failure}.
Typically, the signal acknowledges that the goal has been reached (i.e., the norm of the error pose is under a given threshold). This allows, for example, to structure the flow of the \acrshort{bt} in order to prevent the simultaneous execution of two concurrent actions, such as the Base Planar Action and the Base Yaw Action.
However, in some cases, as the Gaze Tracking action employed in \figurename{}~\ref{fig:laser:mainBT}, the signal may just confirm the initiation of the tracking. This is useful for Gaze Tracking since, in the case of the CENTAURO robot, the relative controller commands only the pitch actuator of the \acrshort{rgbd} camera's support.
This movement can be safely executed throughout the whole mission, even in conjunction with other motions, as it does not interfere with any other robot ability. 

If the \acrshort{bt} node is halted because of the control flow of the BT (e.g., the goal pose has changed, and another action must be executed), it sends an abort request to the Controller, to stop the generation of the reference and the execution of the relative robot motions. 

This way of communication between the action modules the Controller is implemented through \textit{ROS actions}\footnote{\href{https://wiki.ros.org/actionlib}{https://wiki.ros.org/actionlib}}, which enable a long-lasting two-way client/server communication, permitting to stop the execution of the Controller reference generation when the \acrshort{bt} node is halted\footnote{It is important to keep in mind the difference between \textit{action modules}, which are the node of the \acrshort{bt} protagonists of this section, and \textit{ROS actions}, which are the communication mean utilized by the action modules.}.

As anticipated, each action module can be specialized with parameters as needed, to attach them with the request sent to the Controller. These parameters can be adjusted to tailor the robot's behavior to different tasks, and they can be set easily using the Groot2 graphical user interface. The list of available parameters is shown in Figure~\ref{fig:laser:nodeBT}, and explained in what follows.

\begin{itemize}
	\item \texttt{command\_mode} can be set to either \texttt{Track} or \texttt{Reach}. When \texttt{Track} is selected, the Controller verifies the goal position at each iteration of the control loop, allowing for reactive adjustments to any changes in the goal position. On the other hand, when \texttt{Reach} is chosen, the Controller checks the goal position only at the time the action is requested, and proceeds to reach this fixed position. This option can be particularly useful for moving the robot relative to its own position, as in the case of \textit{Arm Forward} and \textit{Arm Backward} arm movements of the \textit{Gripper Grasp} subtree of \figurename{}~\ref{fig:laser:graspBT}.
	
	\item \texttt{linear\_mode} and \texttt{angular\_mode} possible options are \texttt{Set}, \texttt{Keep}, and \texttt{None}, which are used to control the Cartesian task handled by the Controller. With \texttt{Set} mode, the given linear/angular position (i.e.\ the goal to track) is pursued. With \texttt{Keep} mode, the robot maintains its current linear/angular position relative to the link associated with the task. For example, this is an option that is utilized to keep fixed the orientation of the end-effector while performing the \textit{Arm Forward} and \textit{Arm Backward} movements of the \textit{Gripper Grasp} subtree. 
	With \texttt{None} mode, the linear/angular part of the Cartesian task is deactivated, allowing more freedom for the active part. 
	It must be noticed that defining both \texttt{linear\_mode} and \texttt{angular\_mode} as \texttt{None} has no sense since the task would be completely deactivated.
	
	\item \texttt{goal\_frame} is the name of the goal to track/reach, for example the laser spot. This parameter can be set equal to a link of the robot when combined with \texttt{command\_mode == Reach} to command the link to move relative to its own position.
	
	\item \texttt{final\_goal\_distance} and \texttt{final\_ref\_orientation} can be specified to define an offset from the 3D pose of the \texttt{goal\_frame} as the target position, hence they are utilized when \texttt{linear\_mode} and \texttt{angular\_mode} are \texttt{Set}, respectively. For example, this parameter has been utilized to position the mobile base of the robot behind the laser spot to allow enough space for the end-effector to reach it. Please note that both Euler angles and quaternions (as shown in the figure) can be used to provide the \texttt{final\_ref\_orientation}.

	\item \texttt{linear\_error\_norm} and \texttt{angular\_error\_norm} parameters determine the desired final error norms that the robot should achieve in order to consider the goal as reached.
	
\end{itemize}

Note that not all these parameters are utilized for every action module. For example, to govern the yaw tracking, the \texttt{linear} parameters are meaningless.

\subsubsection{Behavior Tree Control Flow}\label{sec:laser:btflow}
\begin{figure}[H]
	\centering
	\includegraphics[width=1\linewidth]{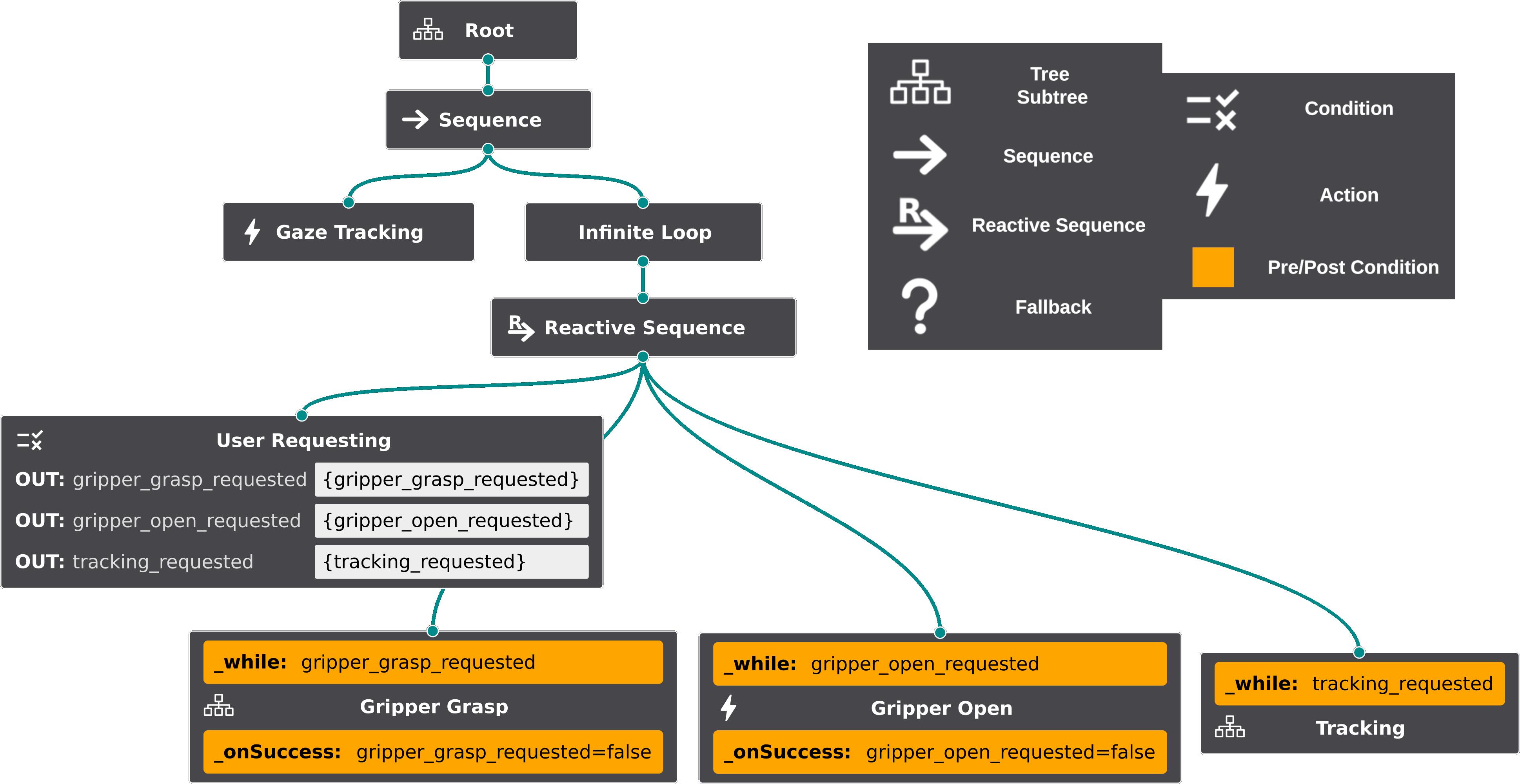}
	\caption[Laser-guided interface: top-level \acrshort{bt}]{The top-level \acrshort{bt} implemented. While the Gaze Tracking action is always active to follow the laser with the camera, the subtrees Gripper Grasp, Gripper Open, and Tracking are activated/halted according to the User Requesting condition.}
	\label{fig:laser:mainBT} 
\end{figure}

The implemented top-level tree is shown in \figurename{}~\ref{fig:laser:mainBT}. In an initialization phase, the Gaze Tracking node activates the homonym robot ability, to continuously follow the laser spot with the robot head. 
In the Infinite Loop, a custom node that let the children run forever, the \acrlong{bt} reacts to the user's input thanks to the Reactive Sequence and to the \texttt{\_while} and \texttt{\_onSuccess} pre- and post-conditions. 
The pre- and post-conditions are not part of the standard \acrshort{bt} theory, but they are anyway implemented in the utilized \texttt{BehaviorTree.cpp} library. Their purpose is to help in developing the BT and improving the readability, assuming that the condition script is kept basic. 

Once triggered by the \acrshort{bt} tick, the User Requesting node (leftmost BT leaf) checks if the user has issued any discrete request (\figurename{}~\ref{fig:laser:controlScheme}), sets the correspondent \texttt{\_requested} variable in the BT \enquote{memory} (i.e., the so-called blackboard), and reset any other request. 
These variables are checked by the pre-conditions \texttt{\_while}, and the relative subtree is activated, if requested. In the subtrees with the \texttt{\_onSuccess} post-condition, the user's request is reset if the subtree returns \texttt{success}. This prevents the execution of the subtree multiple times. 
Thanks to the tree reactivity, the user can also interrupt any subtree at any time by issuing a negative request. Furthermore, if a new request is received, any other subtree in execution is halted before activating the correct subtree. This is useful, for example, to let the user request a grasping while guiding the robot with the laser. The grasping request halts the tracking of the laser spot and the \acrshort{bt}'s flow switches to the subtree Gripper Grasp to grasp the object previously pointed. 

\begin{figure}[H]
	\centering
	\includegraphics[width=1\linewidth]{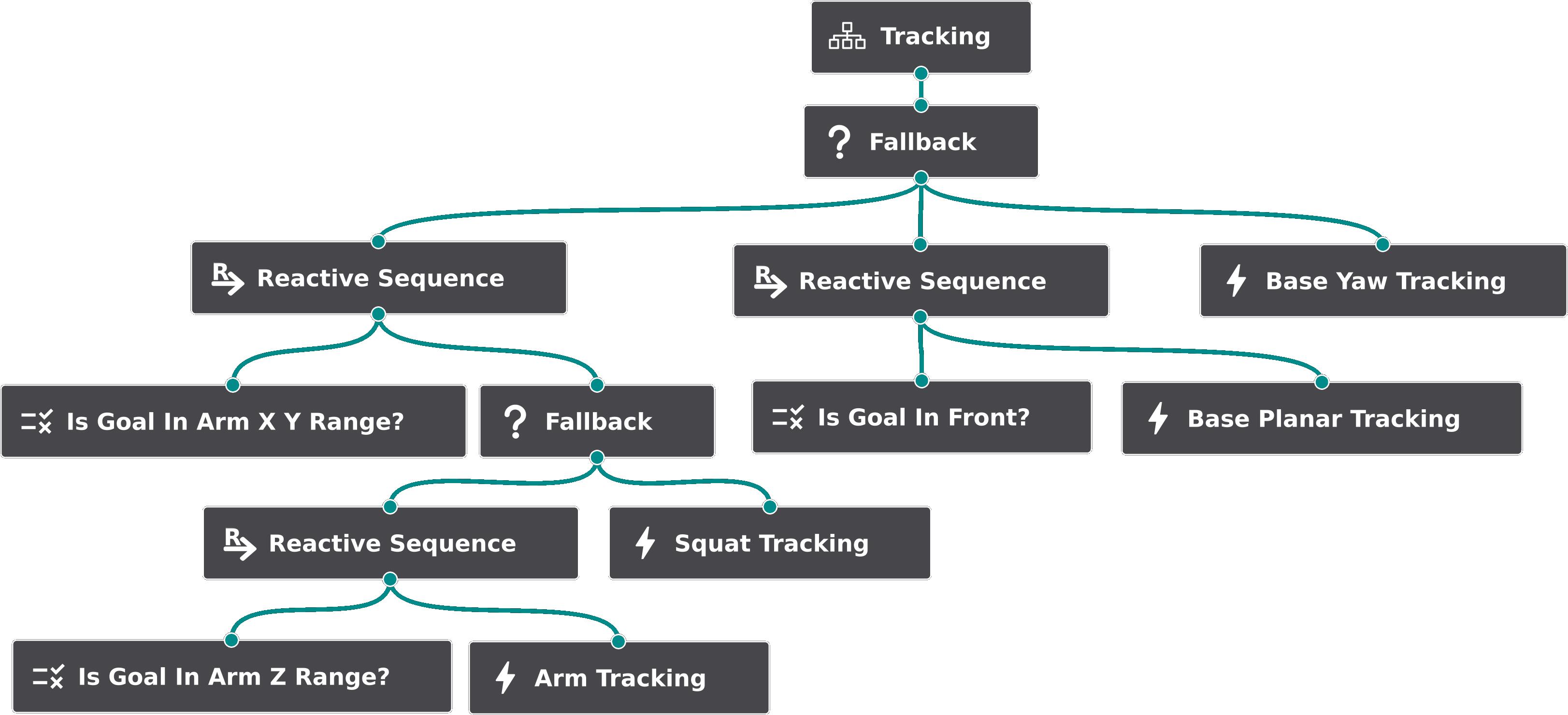}
	\caption[Laser-guided interface: tracking \acrshort{bt}]{The Tracking subtree, used to activate the most suitable robot ability to track the goal. Thanks to the reactivity of the BT, the actions are deactivated and activated promptly as soon the goal position changes.}
	\label{fig:laser:trackingBT}
\end{figure}

The Tracking subtree in the rightmost BT leaf of \figurename{}~\ref{fig:laser:mainBT} is expanded in \figurename{}~\ref{fig:laser:trackingBT}. Note that the parameters shown in \figurename{}~\ref{fig:laser:nodeBT} are not shown to avoid overwhelming the visualization. 

To understand this subtree behavior, as an example, we can imagine that the robot must reach a goal far from the end-effector reaching and not aligned with its head. In this case, both the conditions \textit{Is Goal in Arm X Y Range?}\ and \textit{Is Goal in Front?}\ return \texttt{failure} so the top level fallback executes the rightmost leaf, \textit{Base Yaw Tracking}. Indeed, the intention is to align the robot to the goal before doing anything else. In this way, it is easier to track the laser spot with the camera, and it is more natural to drive the robot toward it. With the CENTAURO robot, this is also necessary because the wheels modules are not omnidirectional.
As soon as the heading error is reduced under a threshold, the condition \textit{Is Goal in Front?}\ changes its return statement to \texttt{success}. This halts the Base Yaw Tracking and triggers the execution of the \textit{Base Planar Tracking} action, which commands the robot toward the goal at a predefined distance settable with the BT node parameters. 
As soon as the goal is reachable by the arm (left or right, depending on the one considered in the mission), the \textit{Is Goal in Arm X Y Range?}\ returns \texttt{success}, shifting the BT execution to the left part of the subtree. In this part, a logic is present to squat up or down the robot (\textit{Squat Tracking}) if the goal is too low or too high to be reached by the end-effector (\textit{Is Goal in Arm Z Range?}). As soon also \textit{Is Goal in Arm Z Range?}\ is satisfied, arm commands are issued with the \textit{Arm Tracking} action module. 
Thanks to the exploitation of the reactive sequences, at any instant, as soon as one condition is not met anymore, the BT changes promptly its state to govern the most suitable capability of the robot. This is really useful to track continuously the goal in the best way possible. 

Regarding the \textit{Is goal in arm X Y Z range?}\ conditions, to define what is reachable by the end-effector, a 3D space in front of the robot has been set considering the kinematic properties of the arm.

\begin{figure}[H]
	\centering
	\includegraphics[width=0.7\linewidth]{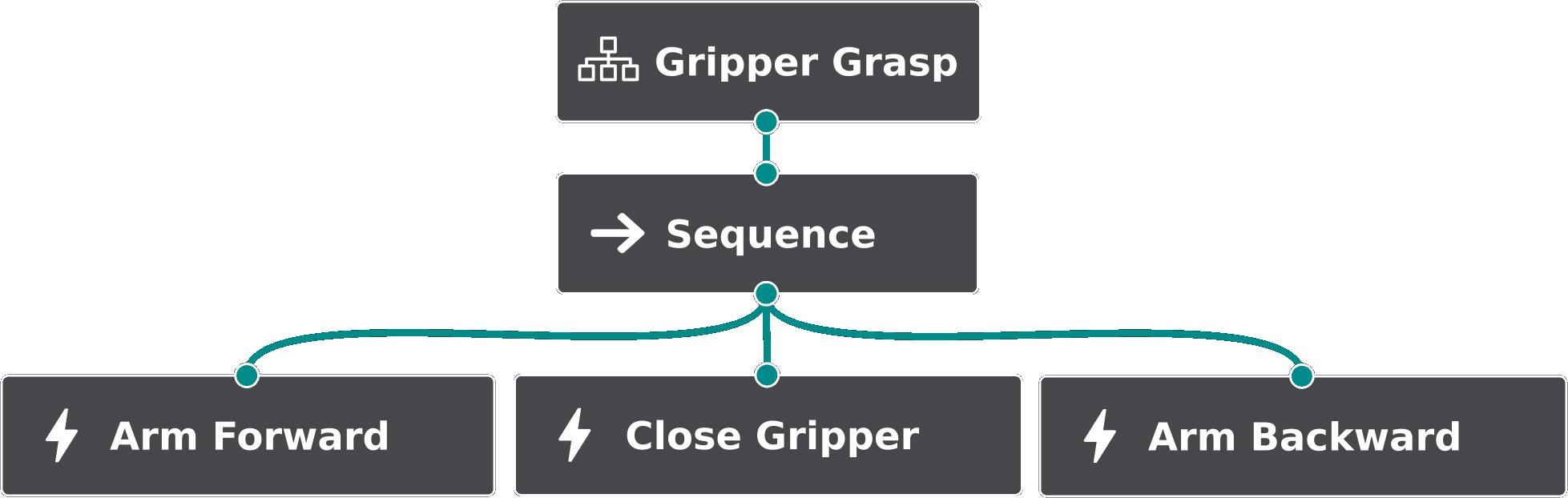}
	\caption[Laser-guided interface: grasping \acrshort{bt} scheme]{The Gripper Grasp subtree. To correctly grasp an object, the arm is driven forward a bit before closing the gripper. The object is then lifted up by retracting the arm.}
	\label{fig:laser:graspBT}
\end{figure}

In \figurename{}~\ref{fig:laser:graspBT}, the Gripper Grasp subtree is shown. The \textit{Arm Forward} action slightly moves the arm toward the object before the \textit{Close Gripper} action activates. The \textit{Arm Backward} action lifts the object by moving the arm up and back of a few centimeters. 

The \textit{Gripper Open} subtree is a single \acrshort{bt} node which opens the gripper to release an object.

The presented \acrshort{bt} of \figurename{}~\ref{fig:laser:mainBT} is generic enough and shows flexibility to different locomanipulation missions. Indeed, thanks to the implemented nodes, it is straightforward to adapt the \acrshort{bt} structure with different control flows and parameters to define diverse behaviors according to the mission and to the robot. 
For example, if only locomotion capabilities are necessary, like to guide a mobile platform toward a path, it can be employed the \acrshort{bt} of \figurename{}~\ref{fig:laser:trackingBT} after removing the left part related to the manipulation, as it will be presented in the experiment of Section~\ref{sec:laser:loco}.

\subsection{Controller and Robot Interface}
The Controller of \figurename{}~\ref{fig:laser:controlScheme} is in charge of generating command references for the Robot Interface based on the capability requested by the \acrlong{bt}. Indeed, each action module of the BT, when activated, requests to the Controller to execute the relative task (e.g., Arm Controller, Squat Controller, Base Planar Controller and so on). 
For mostly of the action modules represented in \figurename{}~\ref{fig:laser:controlScheme}, the controller works in the Cartesian space. In particular, each controller is a \acrshort{pid}-type controller that generates a Cartesian velocity reference $\boldsymbol{\dot{x}}_r$ to drive the specific control point of the task $\boldsymbol{x}_c$ (e.g., the right end-effector for the right arm task, the pelvis for the locomotion tasks) to the indicated \texttt{goal\_frame} point $\boldsymbol{x}_g$, considering the optional parameters of the action module (e.g., the offsets \texttt{final\_goal\_distance} and \texttt{final\_ref\_orientation}).

For the Gaze Action, the controller acts in the joint space, hence utilizing a \enquote{Postural} task,  which drives the pitch joint of the CENTAURO's head to a desired joint position, to keep the goal (i.e., the laser spot) in the vertical center of the \acrshort{rgbd} camera image. 

Another controller acting in the joint space is the one relative to the Gripper Open/Close Actions, which open and close the gripper having as goal the joint position related to the opened and closed gripper's state.

It is worth to notice that, in general, more controllers can be activated to exploit more robot capabilities at the same time, if they are not related to concurrent motions, like Base Planar and Base Yaw Actions.

The output of the Controller, a joint or a Cartesian reference, is then handled by the Robot Interface block which communicates with the robot employing the CartesI/O and Xbot2 frameworks, briefly introduced in Section~\ref{sec:intro:framework}. 
A stack of task is defined for the robot, including Cartesian tasks (for manipulation and locomotion abilities), Postural tasks (to regulate the pitch for the Gaze Action, and to open/close the gripper), and safety constraints such as the ones to limit the joint positions/velocities.

\section{Experimental Validations}\label{sec:laser:exp}

In what follows, experimental validations about the laser-guided interface are presented. Scenarios include manipulation and locomotion missions where the user command the robot to the location of interest through the laser emitter device. The robot employed in the experiment is the CENTAURO platform, introduced in Section~\ref{sec:intro:centauro}. In the pick-and-place experiment of Section~\ref{sec:laser:grasp}, the robot is equipped with the DAGANA gripper (Section~\ref{sec:intro:dagana}) in one of the arm.
For the detection of the laser, the neural network chosen among the studied ones is the YOLOv5-L model, for a good compromise between speed and robustness. 
In all the experiments, different versions of the \acrshort{bt} of \figurename{}~\ref{fig:laser:mainBT} are employed, according to the task needs.

The demonstrations presented are available in the video available at the following link: \href{https://youtu.be/emHXO4L6OX0}{https://youtu.be/emHXO4L6OX0}.

\subsection{End-Effector Waypoints Reaching Experiment}\label{sec:laser:expreach}
\begin{figure}[H]
	\centering
	\includegraphics[width=0.70\linewidth, keepaspectratio]{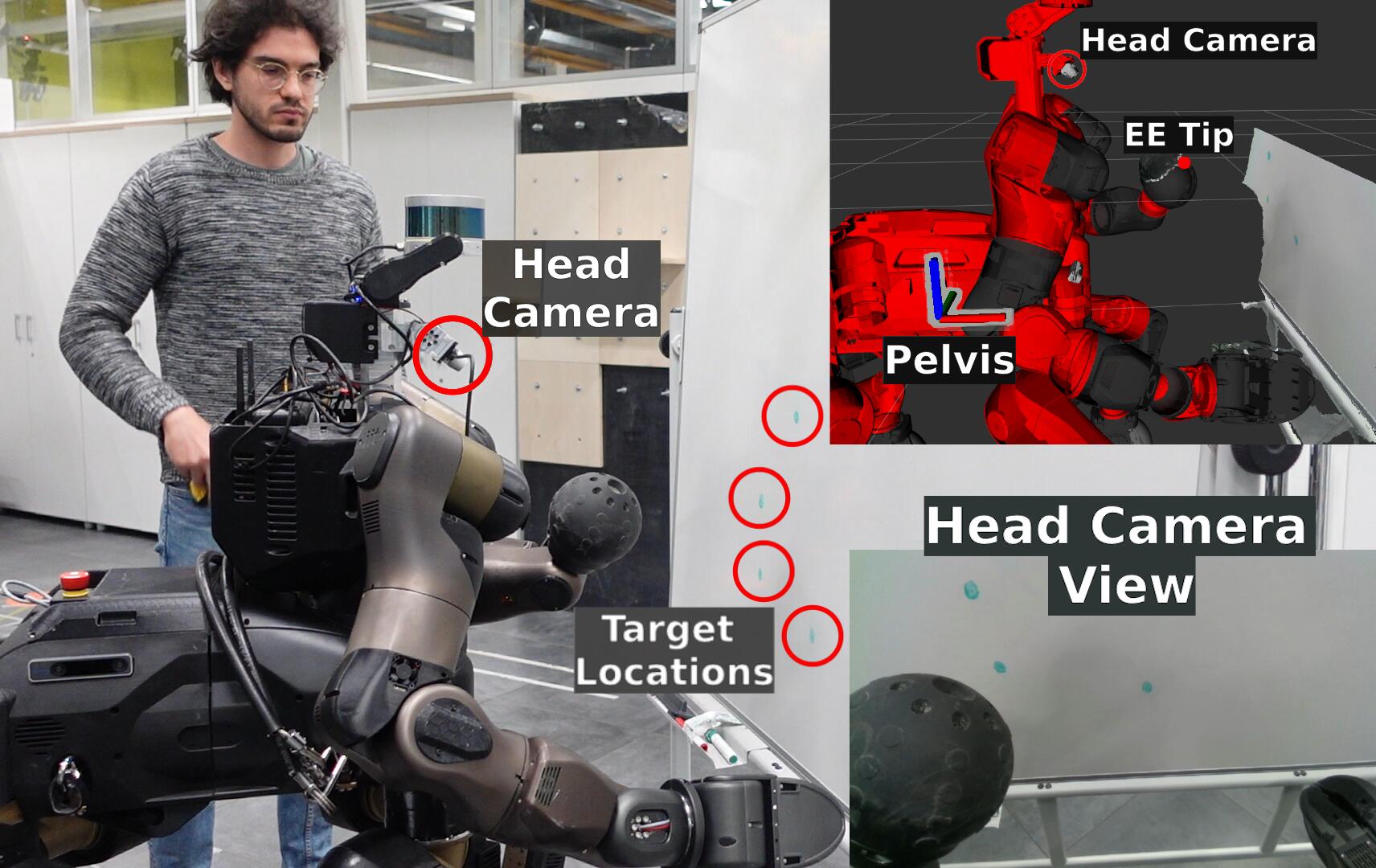}\\
	\vspace{3px}
	\includegraphics[trim={3cm 0 0 0},clip,width=0.49\linewidth, keepaspectratio]{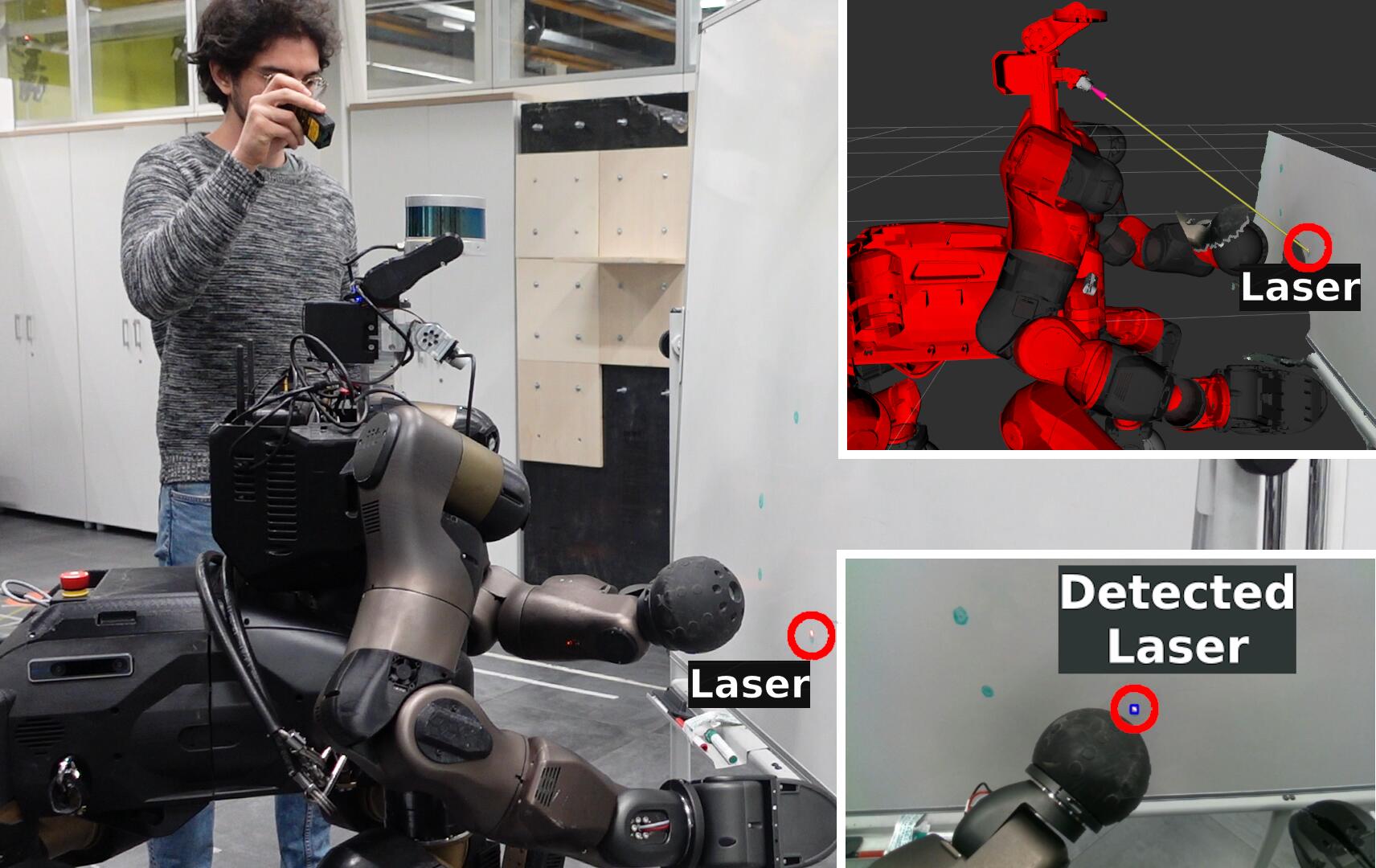}	
	\includegraphics[trim={3cm 0 0 0},clip,width=0.49\linewidth, keepaspectratio]{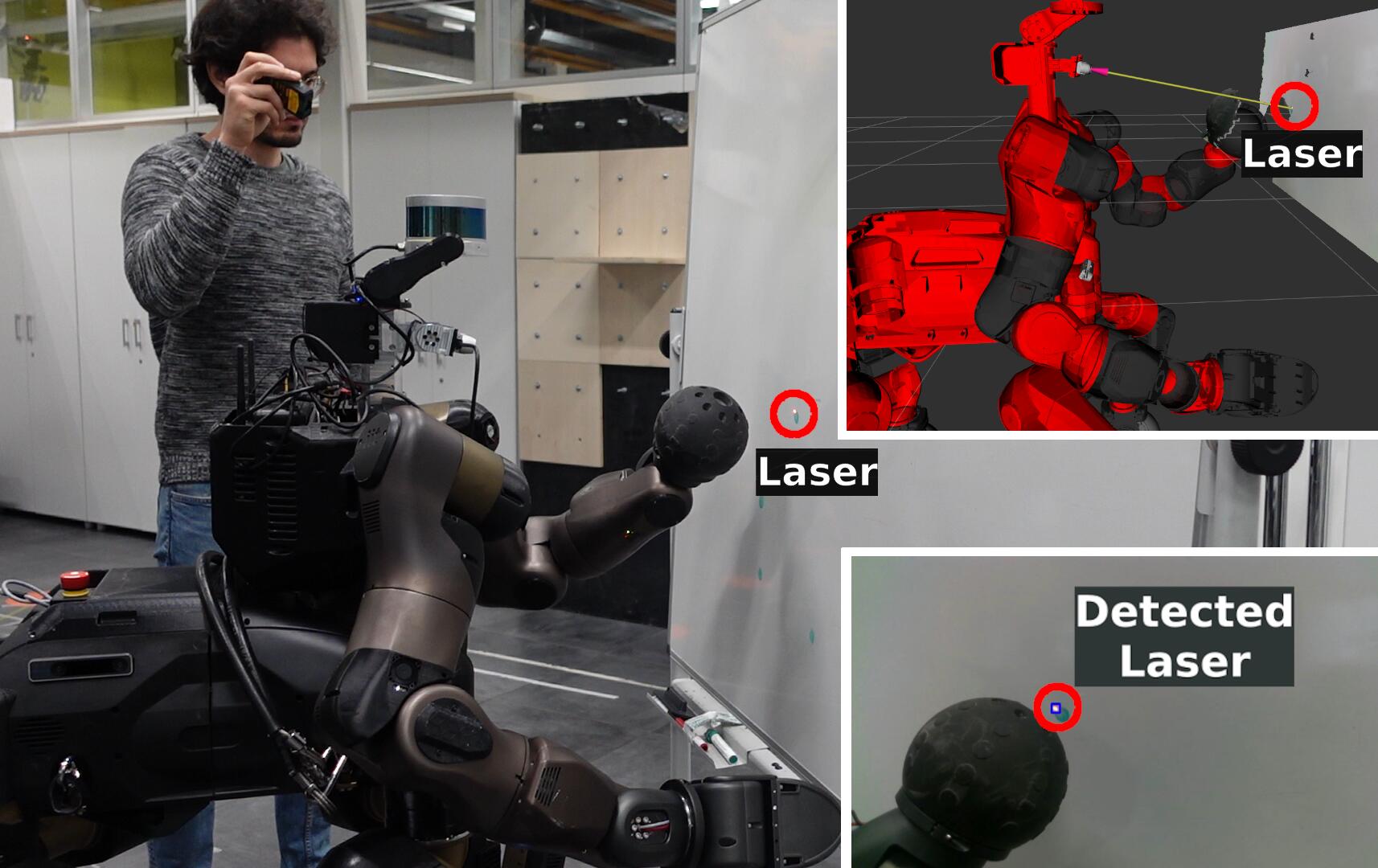}
	\caption[Laser-guided interface: end-effector waypoints reaching experiment]{In the end-effector waypoints reaching experiment, the operator guides the tip of the robot's left end-effector to the four target locations by pointing the laser to them.}
	\label{fig:laser:leftArmPhoto}	
\end{figure}

In this experiment, the operator guides the robot left arm through four different waypoints, by directing the laser in such locations. The setup and sequences of the experiment are visible in \figurename{}~\ref{fig:laser:leftArmPhoto}.

\begin{figure}
	\centering
	\includegraphics[width=0.6\linewidth]{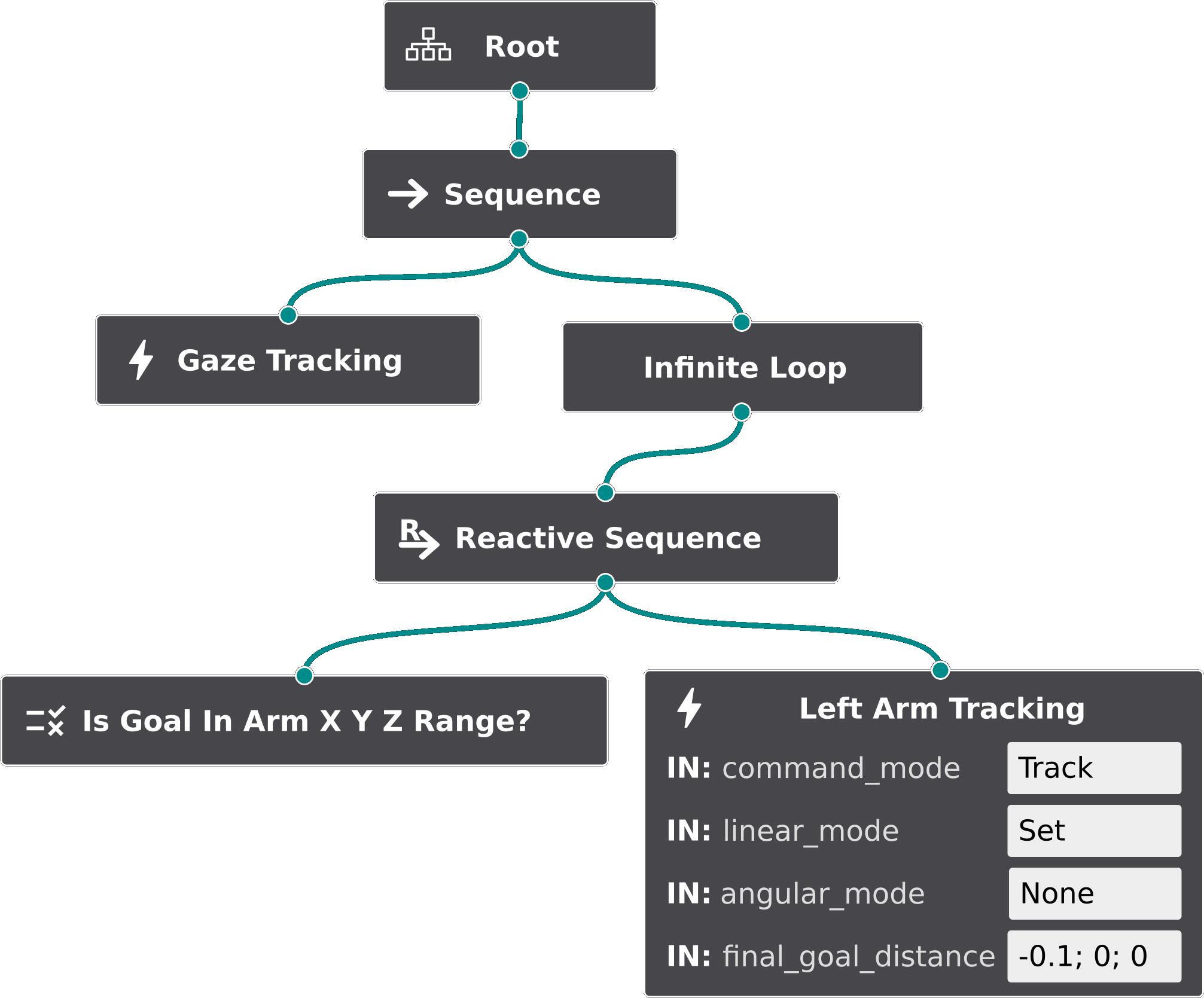}
	\caption[Laser-guided interface: end-effector waypoints reaching experiment BT]{The BT implemented for the end-effector waypoints reaching experiment. The robot's gaze always follows the laser spot. The left arm, if the laser spot is in its workspace, moves toward it by keeping a distance of \SI{0.1}{\meter} on the $\hat{x}$-axis.}
	\label{fig:leftArmBt}
\end{figure}

The robot behavior is driven according to the \acrshort{bt} shown in \figurename{}~\ref{fig:leftArmBt}. This \acrshort{bt} is a simplified version of the BT presented in Section~\ref{sec:laser:btflow}, since only the Left Arm and Gaze action modules are necessary.
 
The logic of such \acrshort{bt} is simple: while the \textit{Gaze Tracking} action keeps as much as possible the laser spot in the robot's camera field of view, the left arm tracks the position of the laser, but only if it is in the goal's arm workspace, a condition checked by the node \textit{Is Goal in Arm X Y Z Range?}. If the condition is not met, the robot stay still, since in this case no locomotion capabilities are modeled in the \acrshort{bt}.
As displayed in the relevant \acrshort{bt} parameters of the \textit{Left Arm Tracking} action module, the robot is instructed to follow the laser by maintaining the end-effector's tip at a \SI{0.1}{\meter} along the $\hat{x}$-axis (\texttt{final\_goal\_distance == -0.1;0;0}), which is the axis pointing out from the end-effector, hence included to not hit the whiteboard. Furthermore, the Cartesian angular part of the task is neglected (\texttt{angular\_mode == None}), meaning that the orientation of the end-effector frame is not controlled, since of no interest in this case.

\begin{figure}
	\centering
	\includegraphics[width=1\linewidth]{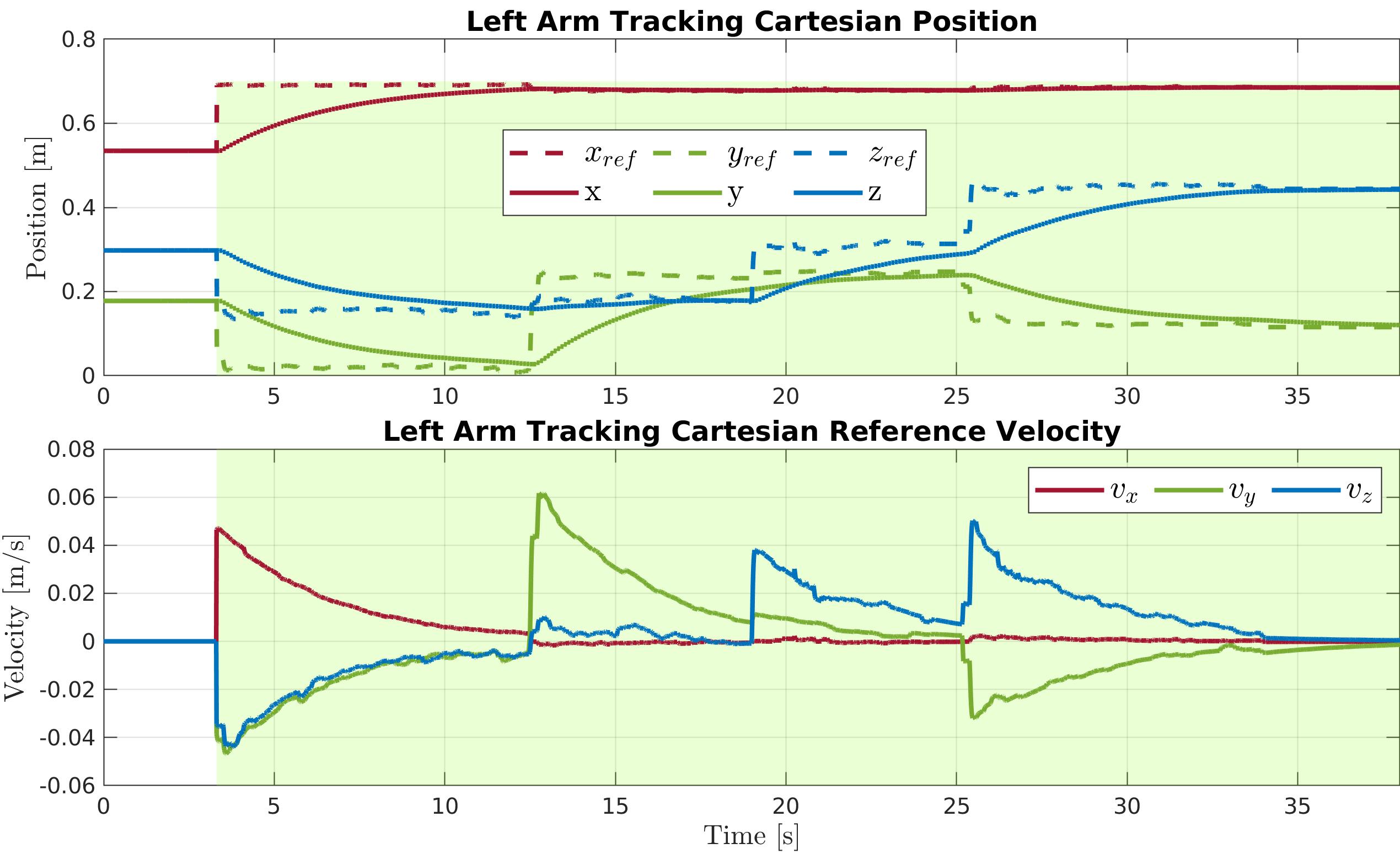}
	\caption[Laser-guided interface: end-effector waypoints reaching experiment plots]{Plots for the end-effector waypoints reaching experiment. Highlighted green areas show when the \textit{Left Arm Tracking} action is active (it is always active since the goal is always in the end-effector workspace). In the top plot, dashed lines represent the laser position with respect to the robot torso frame (including the \SI{0.1}{\meter} offset in the $\hat{x}$-axis); the continuous lines represent the left end-effector tip position. In the bottom plot, the  Cartesian velocity references generated by the left arm controller are displayed.}
	\label{fig:laser:leftArmPlot}
\end{figure}

Relevant plots are shown in \figurename{}~\ref{fig:laser:leftArmPlot}. The green areas show when the \textit{Left Arm Tracking} action is \texttt{running}. It can be noticed that it is always active since the condition \textit{Is Goal in Arm X Y Z range?}\ always returns \texttt{success}, which means that the laser spot stays in the left arm workspace. It is worth mentioning that also the Gaze Action is always active, as modeled by the \acrshort{bt}, even if no relative data is shown in the plots.
In the top plot, the dashed lines represent the laser spot position with respect to the robot torso frame (with the \SI{0.1}{\meter} offset included), as moved by the operator. The continuous lines represent the end-effector tip position, varying while the robot is moving.
In the bottom plot, the Cartesian velocity references $\boldsymbol{\dot{x}}_r$ are shown, generated by the \acrshort{pid} controller accordingly to the tracking error.
 
By following the plots, it can be observed how the velocity references promptly change (four times) as soon the user points the laser to another one of the four waypoints. This demonstrates the responsiveness of the perception layer, made possible thanks to the speed in the detection of the laser by the neural-network model employed.

\subsection{Locomotion Tracking Experiment}\label{sec:laser:loco}
\begin{figure}[H]
	\centering
	\includegraphics[width=0.65\linewidth, keepaspectratio]{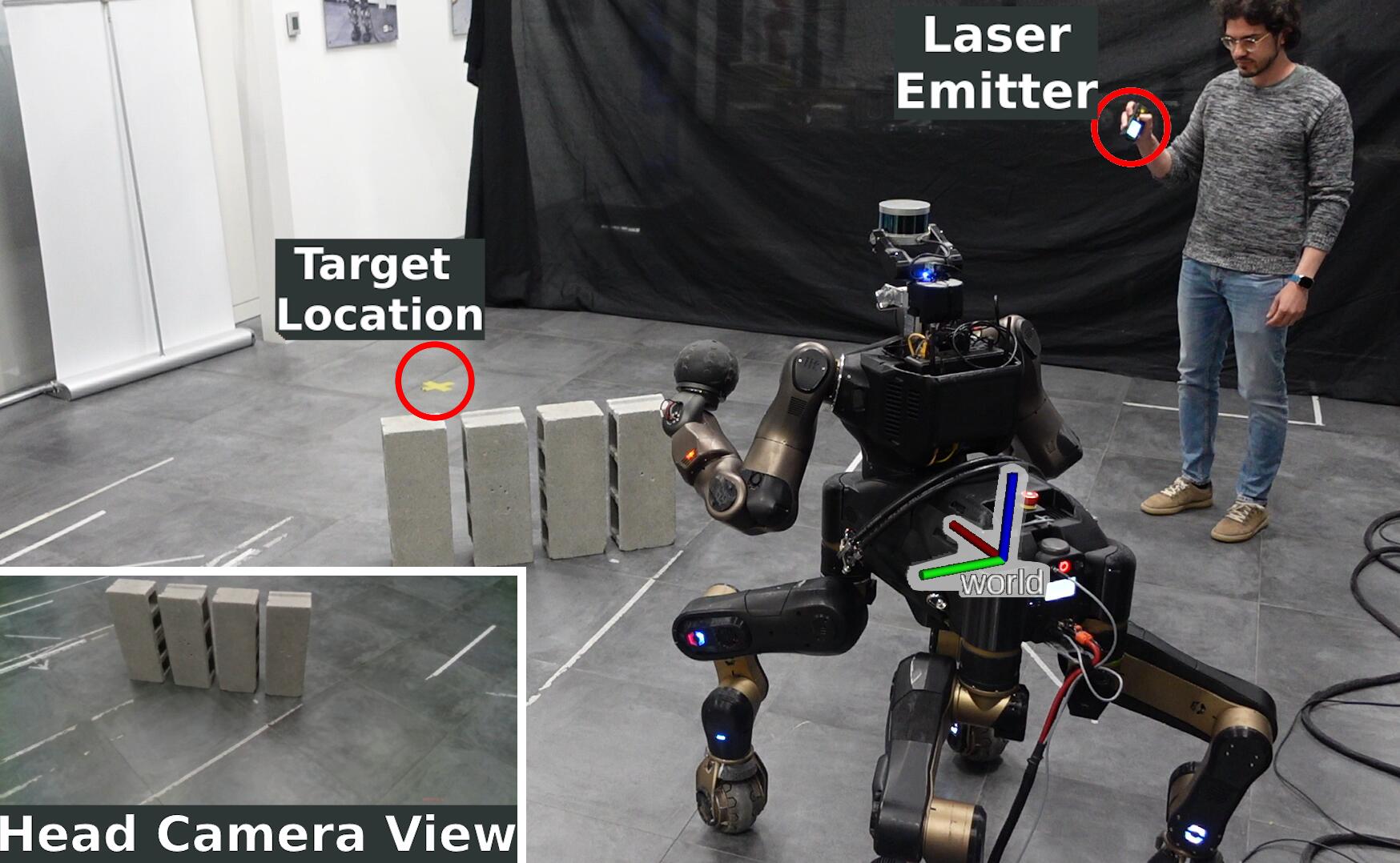}\\
	\vspace{6px}
	\includegraphics[width=0.32\linewidth, keepaspectratio]{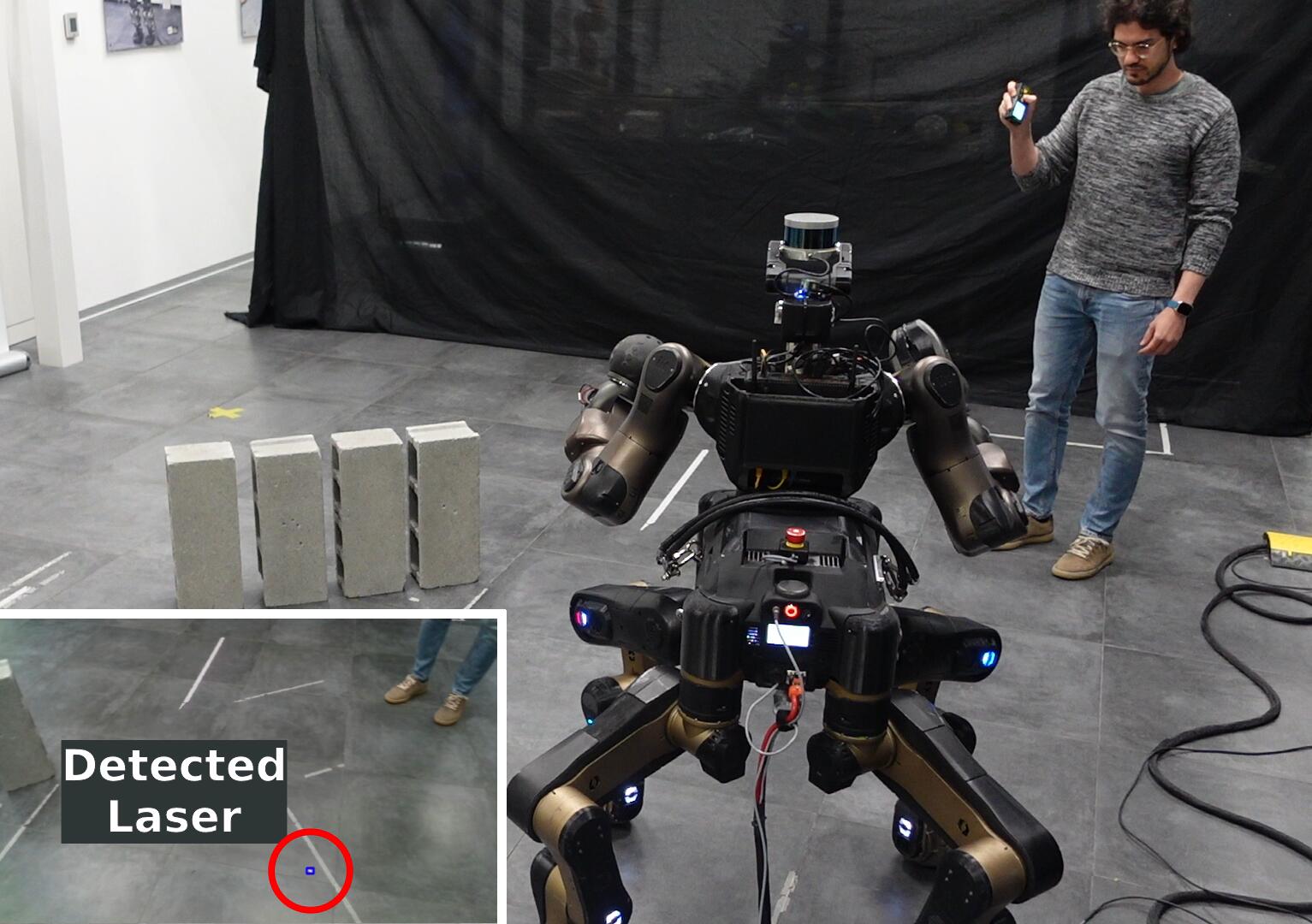}	
	\includegraphics[width=0.32\linewidth, keepaspectratio]{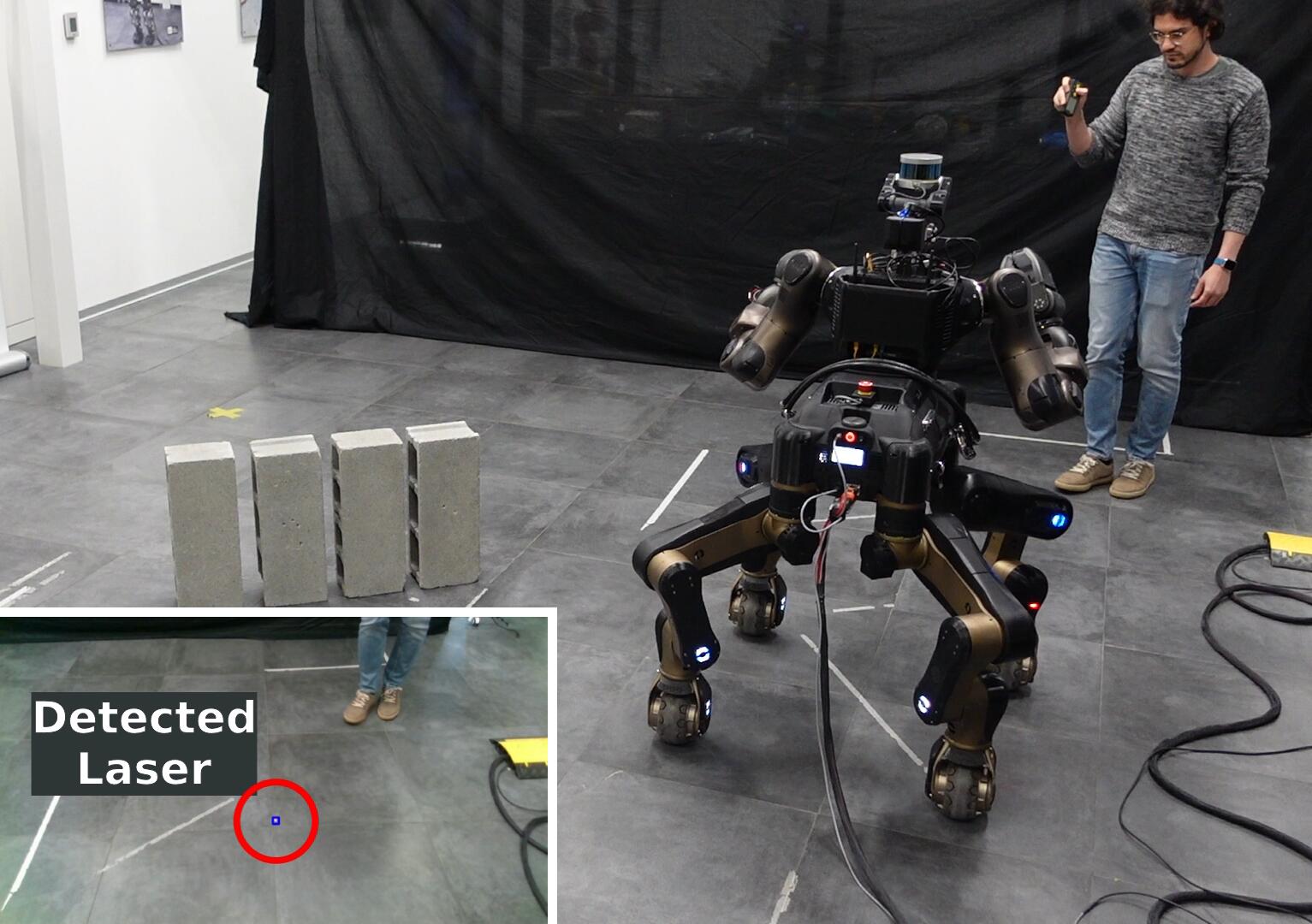}
	\includegraphics[width=0.32\linewidth, keepaspectratio]{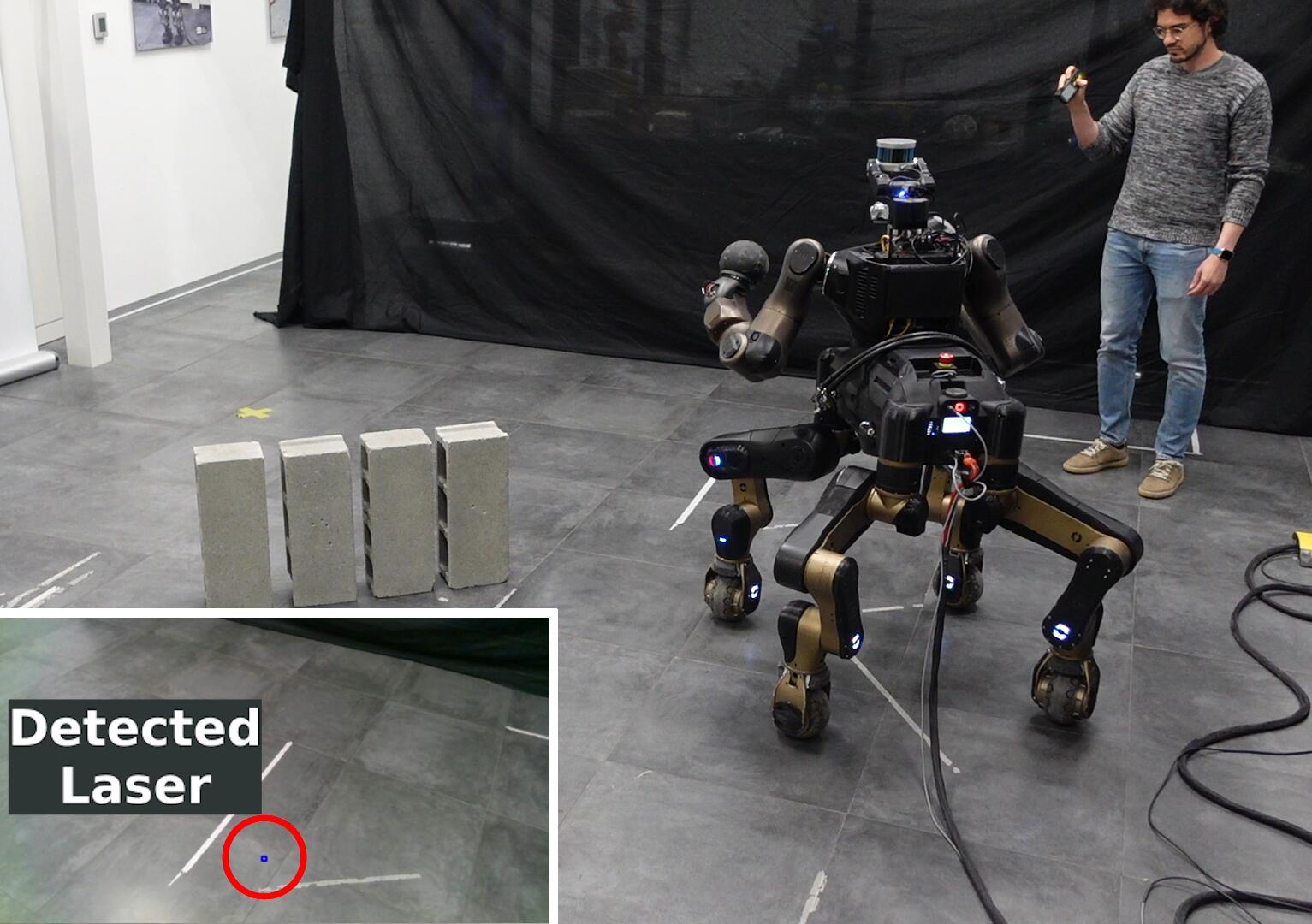}\\
	\vspace{3px}
	\includegraphics[width=0.32\linewidth, keepaspectratio]{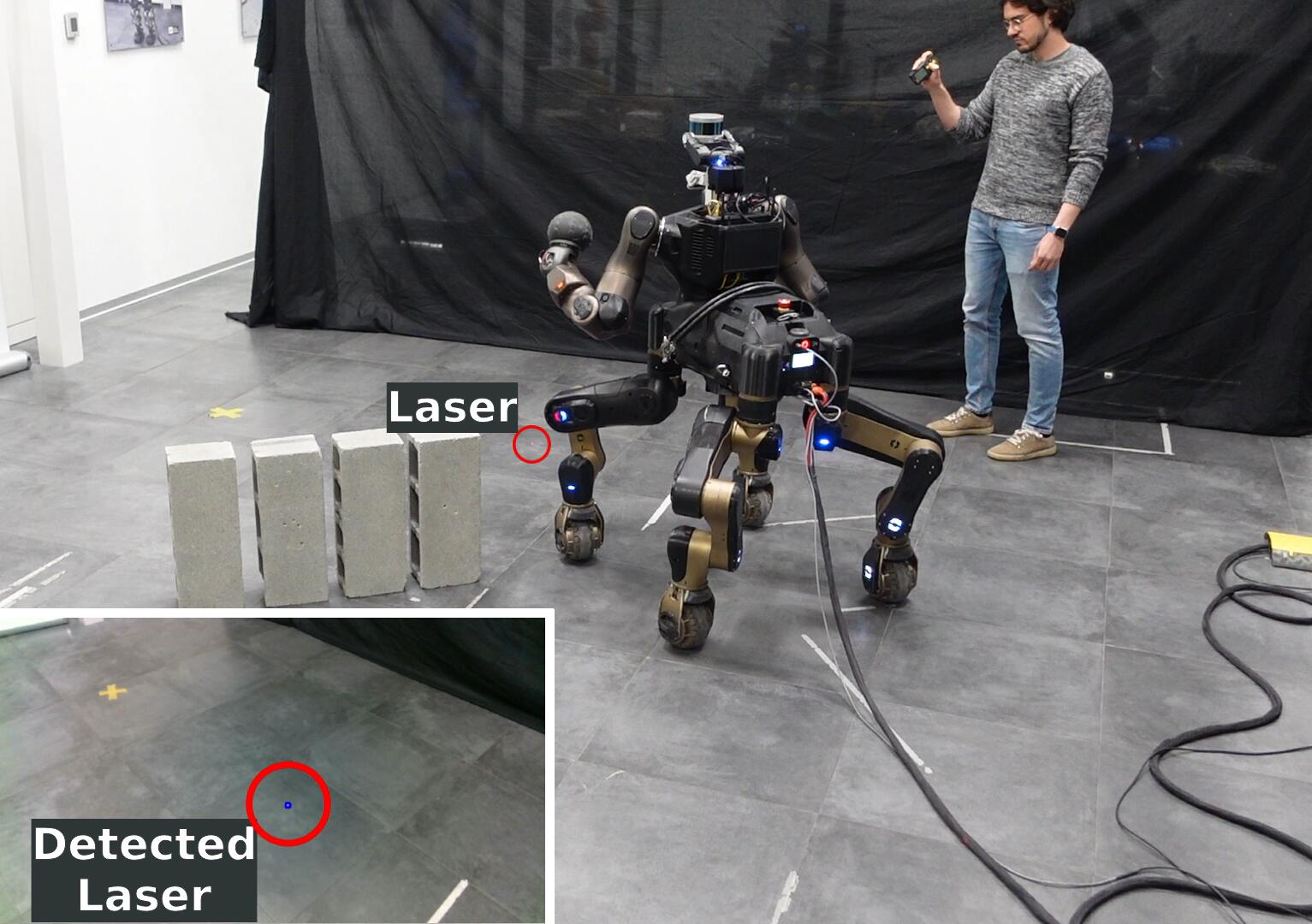}	
	\includegraphics[width=0.32\linewidth, keepaspectratio]{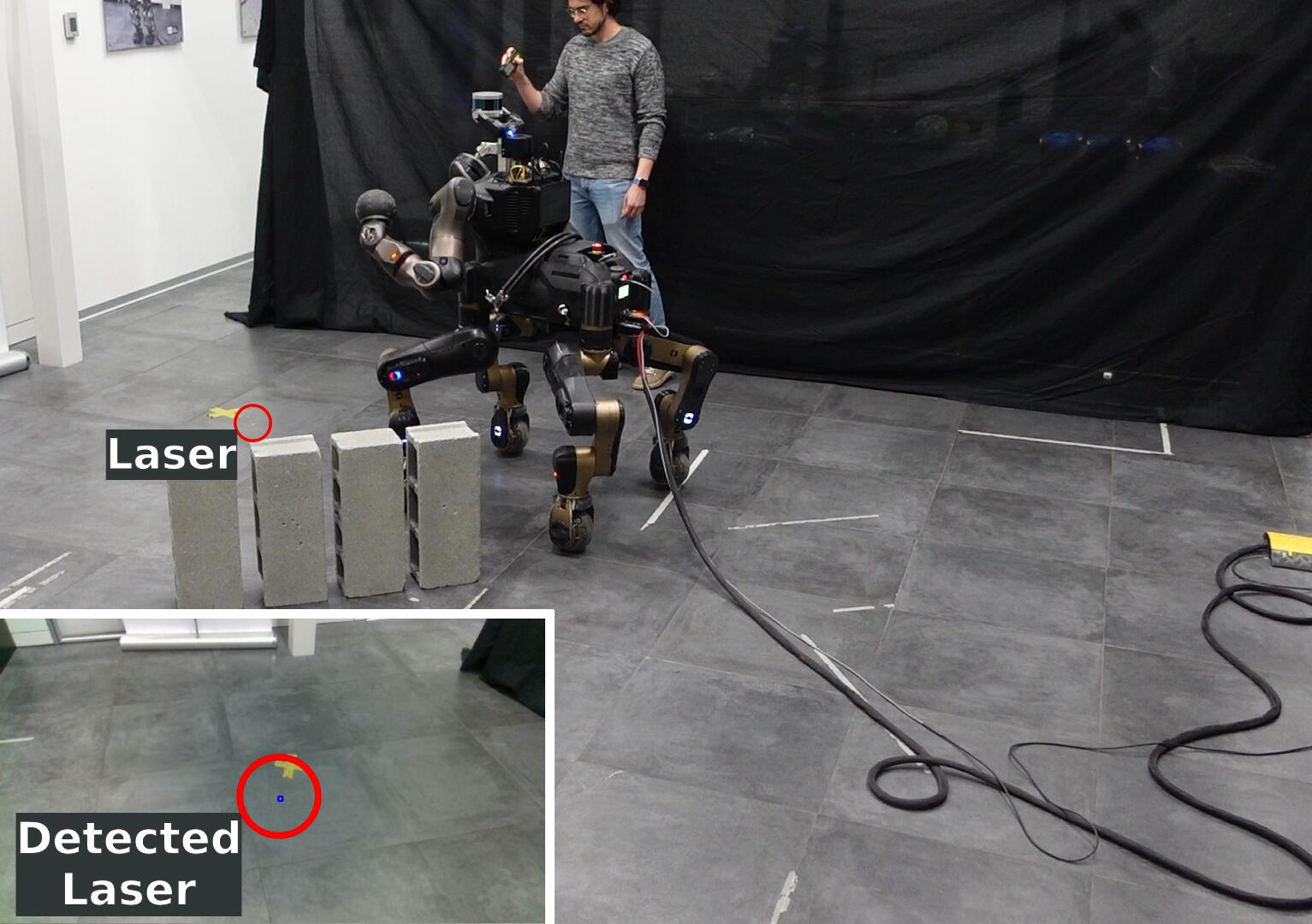}
	\includegraphics[width=0.32\linewidth, keepaspectratio]{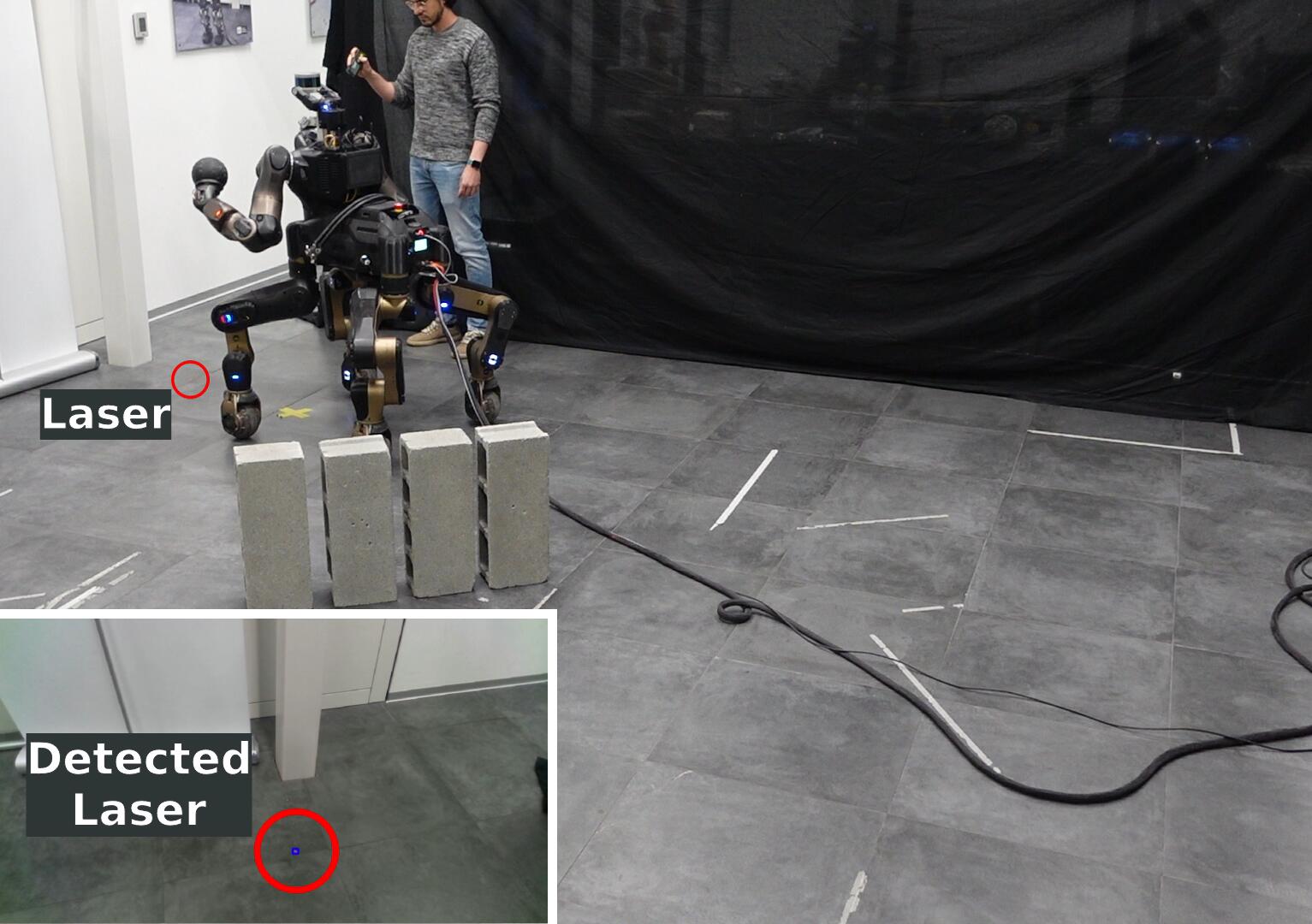}
	\caption[Laser-guided interface: locomotion tracking experiment]{In the locomotion tracking experiment, the operator guides the robot through a path, to reach a target location avoiding the obstacles in the middle of the room. Note that at the end the laser is pointed forward the target location since the interface considers the laser spot with a \SI{1}{\meter} offset as a goal to maintain it in the camera field of view.}
	\label{fig:laser:locoPhoto}	
\end{figure}

\begin{figure}
	\centering
	\includegraphics[width=0.8\linewidth]{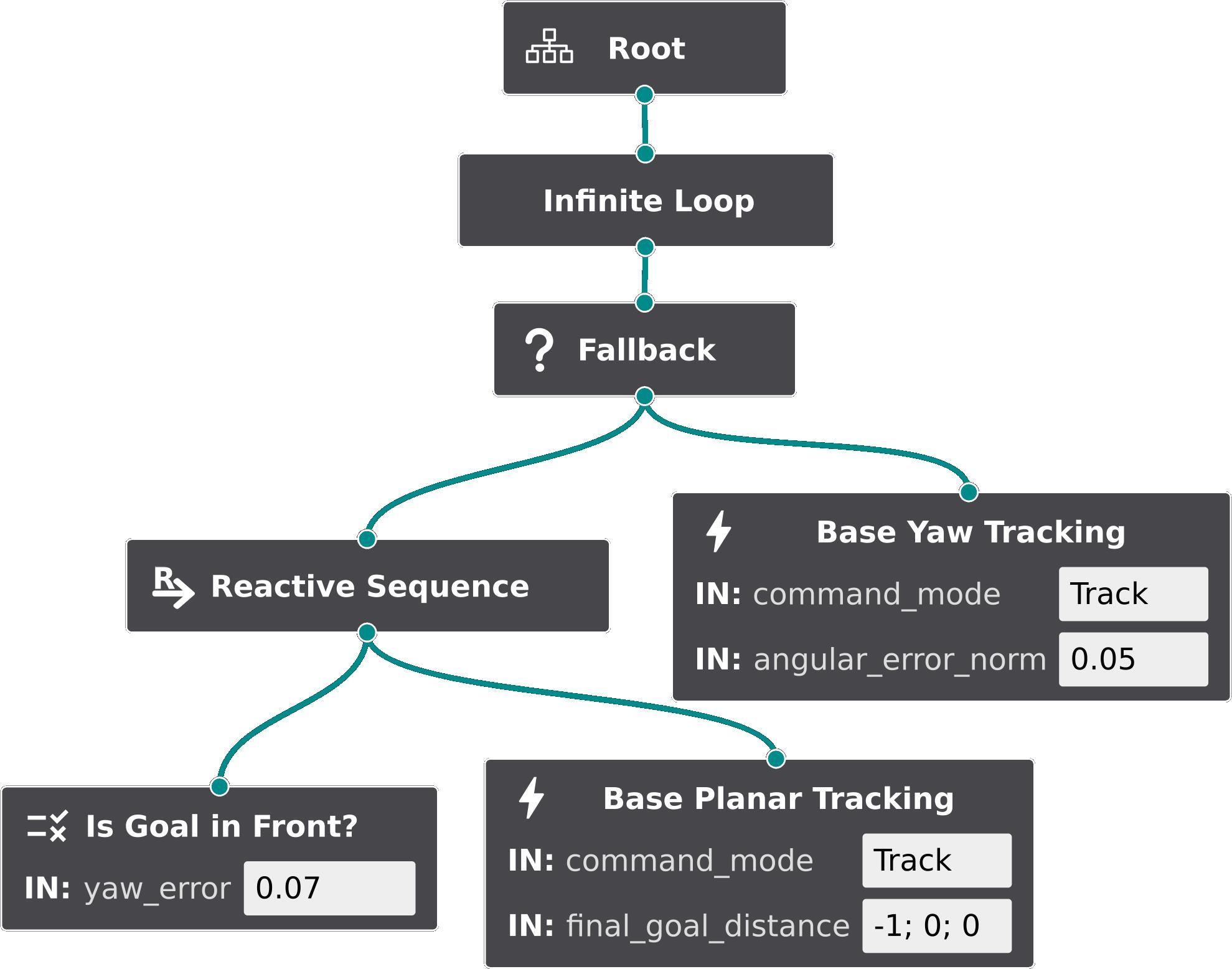}
	\caption[Laser-guided interface: locomotion tracking experiment bt]{The BT used for the locomotion tracking experiment. In the rightmost leaf, the robot aligns itself to the goal with the \textit{Base Yaw Tracking}. Once the goal is in front, the robot moves toward it with the \textit{Base Planar Tracking}. As soon as the goal is not in front anymore, the reactive sequence halts the \textit{Base Planar Tracking} and activates the \textit{Base Yaw Tracking} again.}
	\label{fig:laser:locoBt}
\end{figure}

In this experiment, the operator guides the robot to reach a target location through a path indicated by the laser,  avoiding the obstacles in the middle of the room (\figurename{}~\ref{fig:laser:locoPhoto}). 
The BT exploited is a simplified version of the whole BT presented in Section~\ref{sec:laser:btflow}. Indeed, since no manipulation is necessary, the \acrshort{bt} only includes the locomotion capabilities of the CENTAURO, as shown in \figurename{}~\ref{fig:laser:locoBt}. By following the control flow of such \acrshort{bt}, the robot firstly keeps itself aligned with the laser with the \textit{Base Yaw Tracking} action module, and, when the condition \textit{Is Goal in Front?}\ is satisfied, the robot moves toward to the goal.
The \texttt{angular\_error\_norm} parameter of the \textit{Base Yaw Tracking} action module and the \texttt{yaw\_error} parameter of the \textit{Is Goal In Front?}\ node are set in order to align the robot with the laser spot considering a certain \enquote{cone} threshold. 
In the \textit{Base Planar Tracking} node, the parameters are chosen to track the laser spot maintaining a distance of \SI{1}{\meter} from it (\texttt{final\_goal\_distance == -1;0;0}), to keep it in the camera view, considering that the camera is mounted on the robot head. Hence, in the final sequence of \figurename{}~\ref{fig:laser:locoPhoto}, the user is pointing the laser forward the marked target location, making the robot to stop above it as wanted.
It is worth noting that the Gaze Tracking Action is not present in the \acrshort{bt} since the robot's camera can be pointed always to the floor. 

\begin{figure}
	\centering
	\includegraphics[width=1\linewidth]{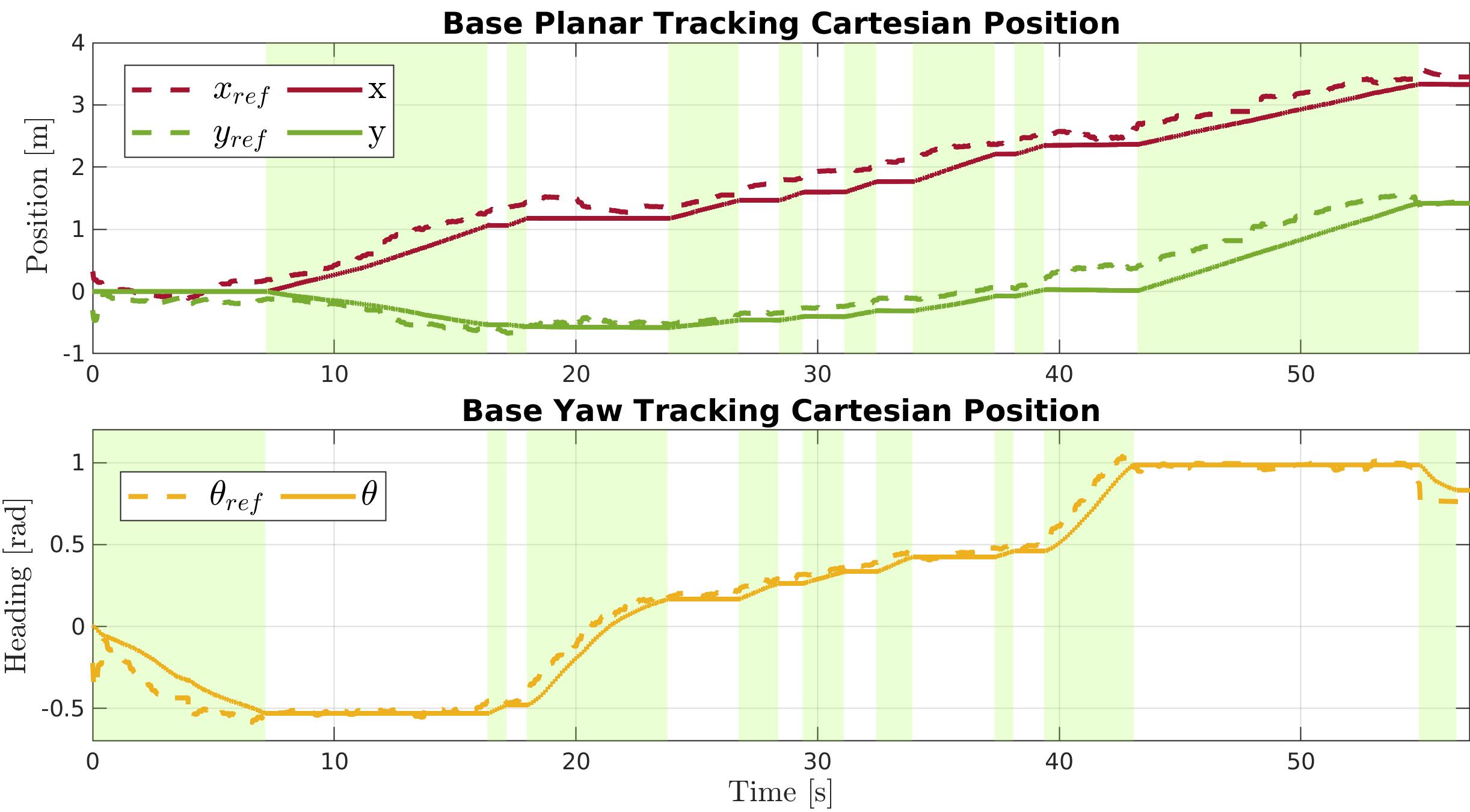}
	\caption[Laser-guided interface: locomotion tracking experiment plots]{Plots for the locomotion tracking experiment. Colored areas highlight the alternating activation of the two Actions \textit{Base Planar Tracking} (top plot) and \textit{Base Yaw Tracking} (bottom plot). The dashed lines and the continuous lines represent the laser spot and the robot pelvis position with respect to the world frame, respectively. The angles $\theta_{\mathit{ref}}$ and $\theta$ are the azimuth angles of the laser spot and of the robot pelvis with respect to the $\hat{x}$-axis of the world frame, respectively.}
	\label{fig:laser:locoplot}
\end{figure}

The plots of \figurename{}~\ref{fig:laser:locoplot} show the trajectories of the robot pelvis frame (continuous line) and of the laser (dashed lines), including the offset of \SI{1}{\meter} along the $\hat{x}$-axis. The data is represented with respect to the world frame (top image of \figurename{}~\ref{fig:laser:locoPhoto}), which is the frame where the robot pelvis frame is at the beginning of the experiment, with the $\hat{x}$-axis pointing in front of the robot, and the $\hat{z}$-axis perpendicular to the floor ($\hat{x}, \hat{y}, \hat{z}$ axis are in red, green, and blue, respectively).
In the bottom plot, the $\theta$ and $\theta_{\mathit{ref}}$ angles  are the azimuth angles of the pelvis frame's origin and of the laser's spot with respect to the $\hat{x}$-axis of the global frame, respectively.

As in the plots of the previous experiment in Section~\ref{sec:laser:expreach}, the highlighted green areas show when the Base Planar and Base Yaw Actions are active. It can be observed that the two nodes are active during alternating intervals. Indeed, this depends on the \textit{Is Goal In Front?}\ condition node, which checks the alignment of the robot with the laser, i.e., the difference between $\theta$ and $\theta_{\mathit{ref}}$.
With respect to the previous experiment, it can also be noticed that now the path of the laser spot is continuous instead of composed by a finite number of waypoints. By tracking such a dynamic goal, this further shows the responsiveness of the perception layer, together with the responsiveness of the \acrshort{bt}-based motion generation layer. Furthermore, it is demonstrated how the robot can be guided not only with discrete waypoints, but also with a continuous path, enabling in practice a tracking of the laser spot with the various robot capabilities.

\subsection{Locomanipulation Pick-and-place Experiment}\label{sec:laser:grasp}
\begin{figure}[H]
	\centering
	\includegraphics[height=0.44\linewidth, keepaspectratio]{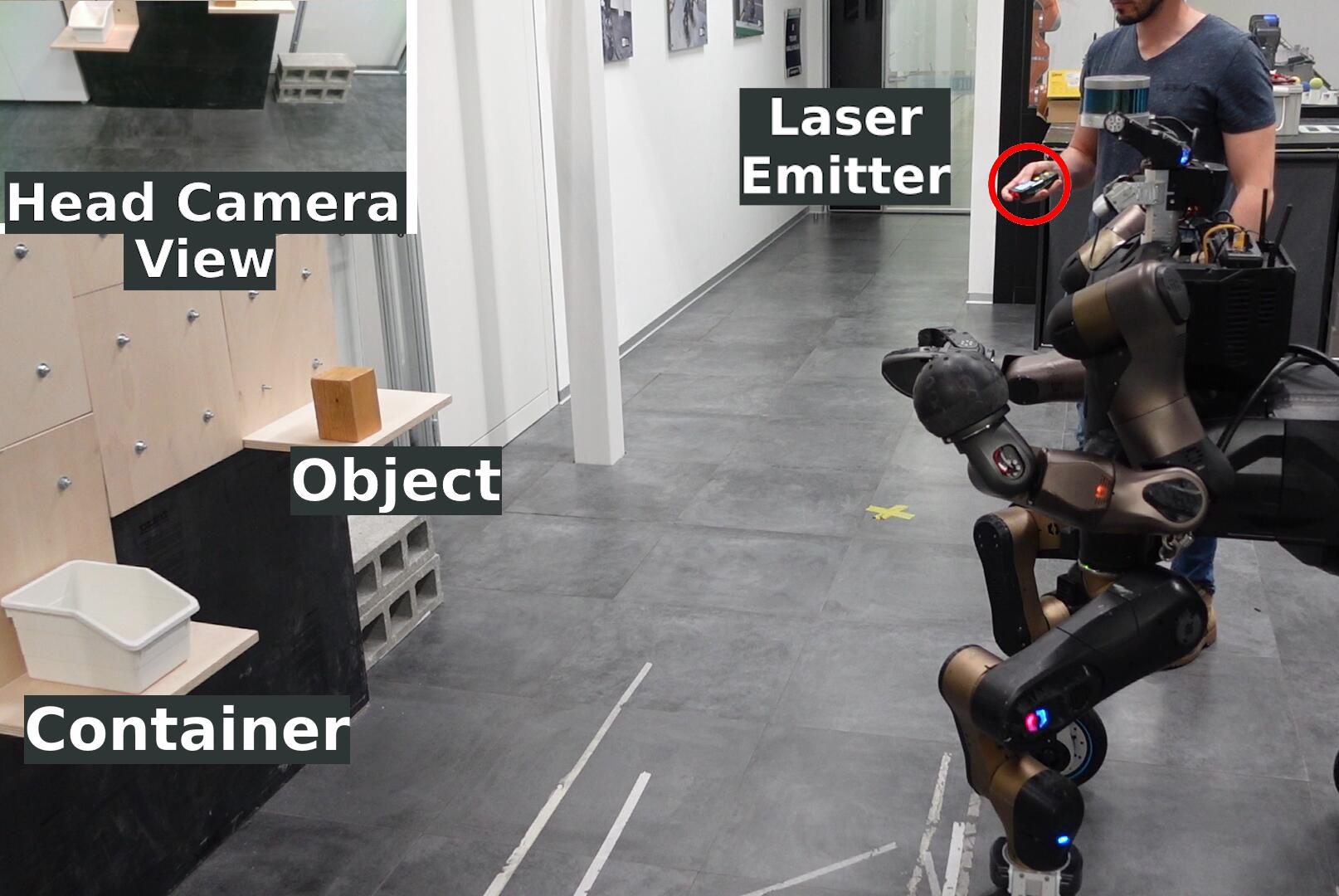}\\
	\vspace{6px}
	\includegraphics[height=0.31\linewidth, keepaspectratio]{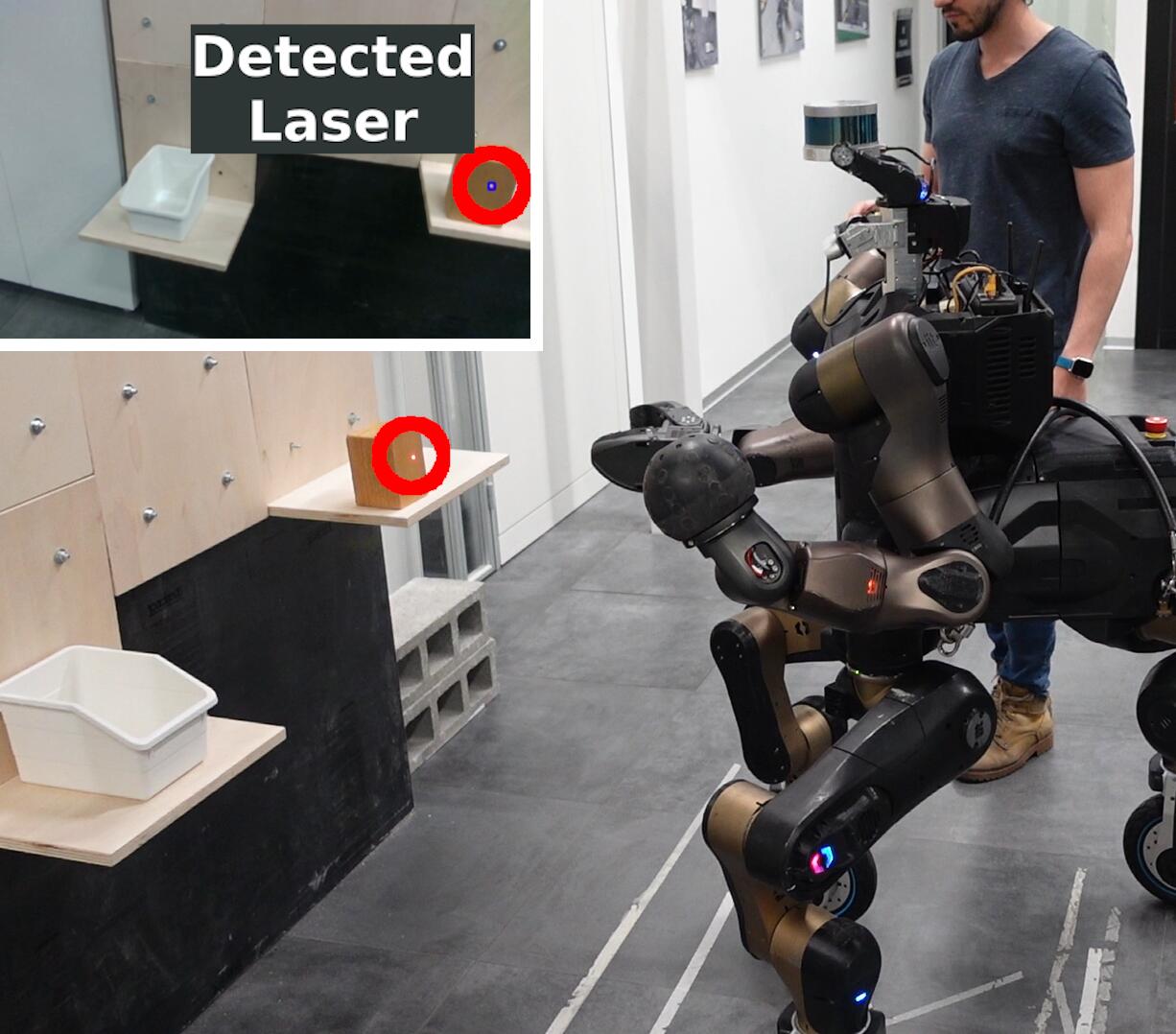}
	\includegraphics[height=0.31\linewidth, keepaspectratio]{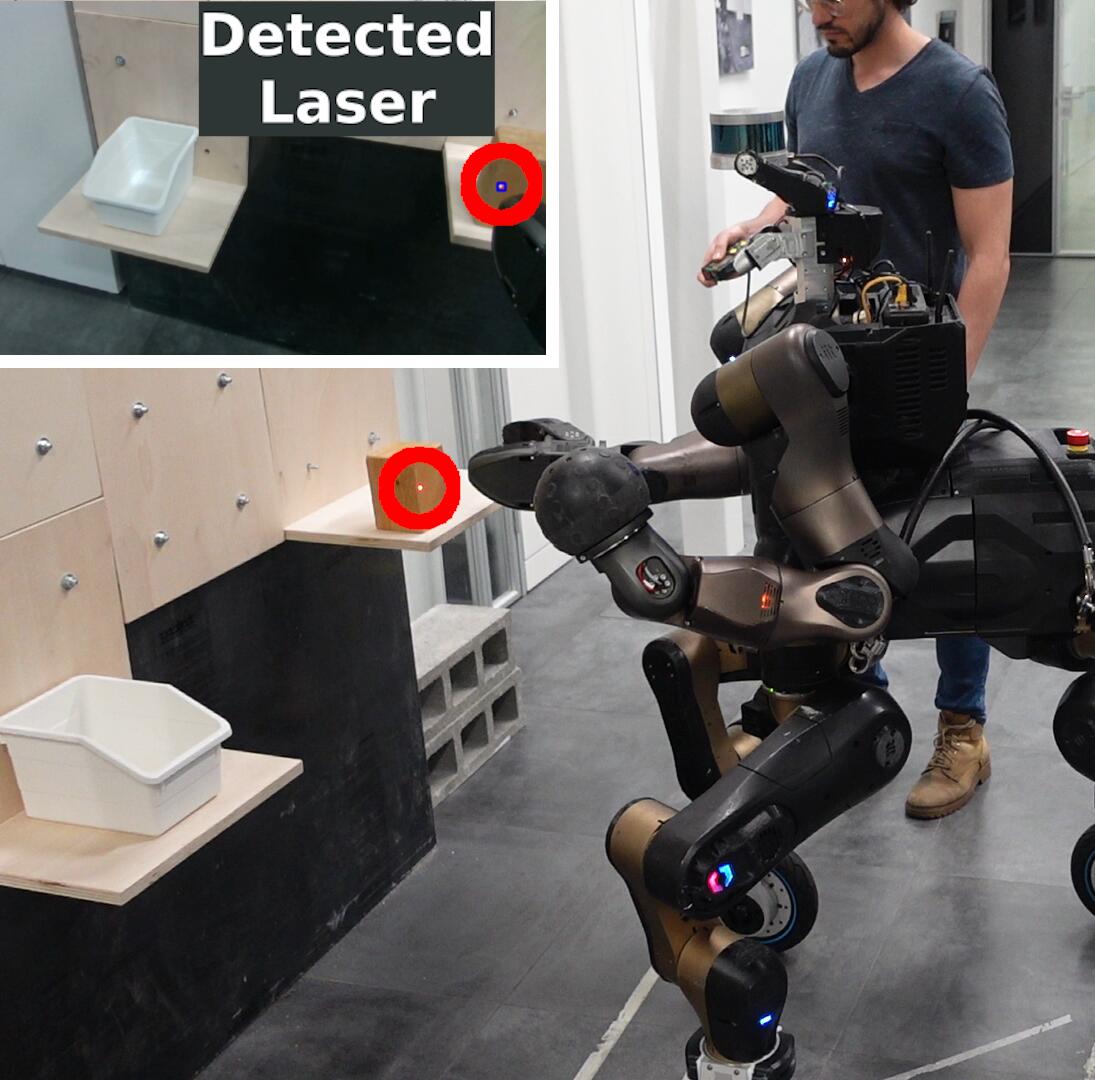}
	\includegraphics[height=0.31\linewidth, keepaspectratio]{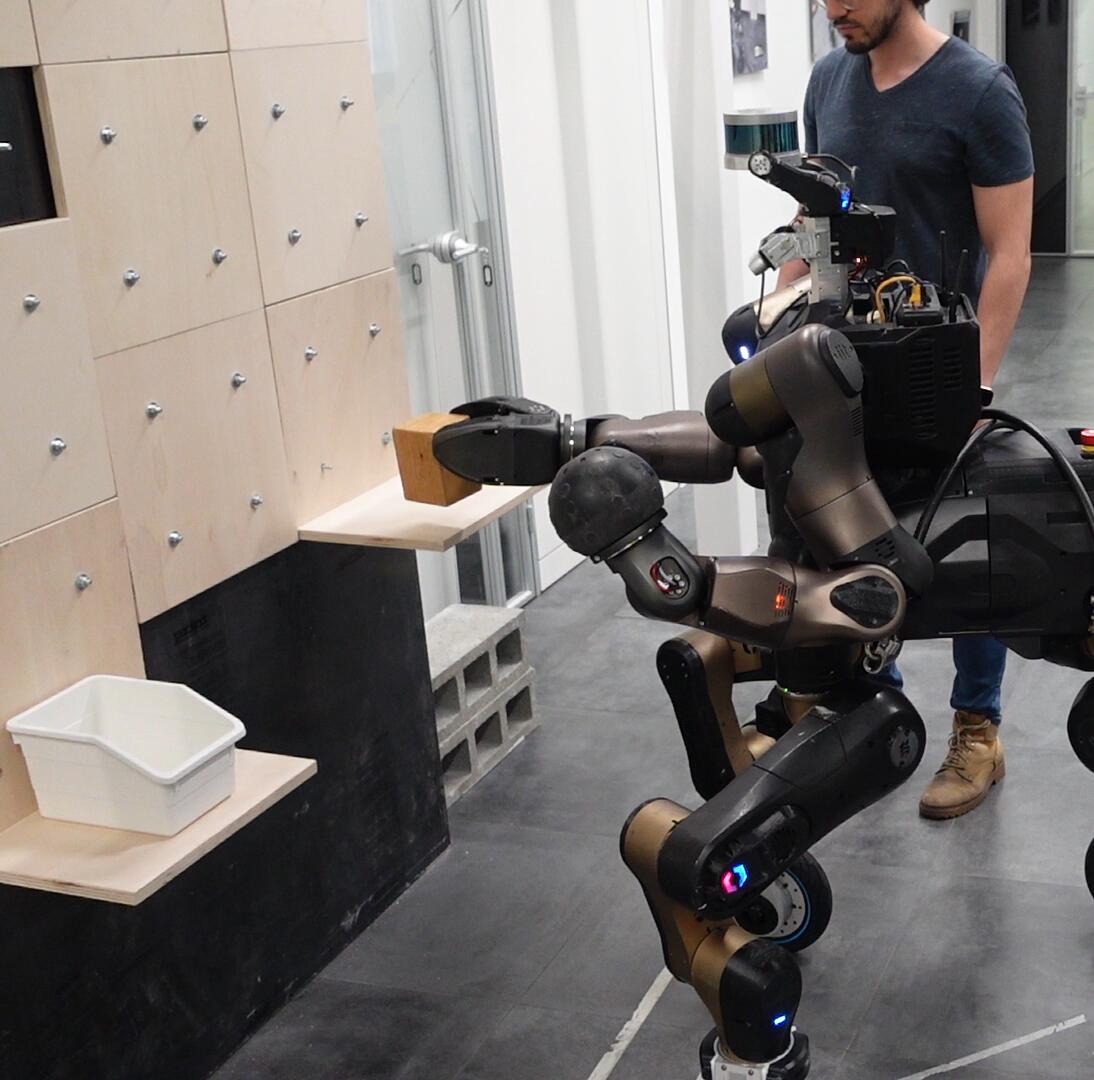}\\
	\vspace{3px}	
	\includegraphics[height=0.35\linewidth, keepaspectratio]{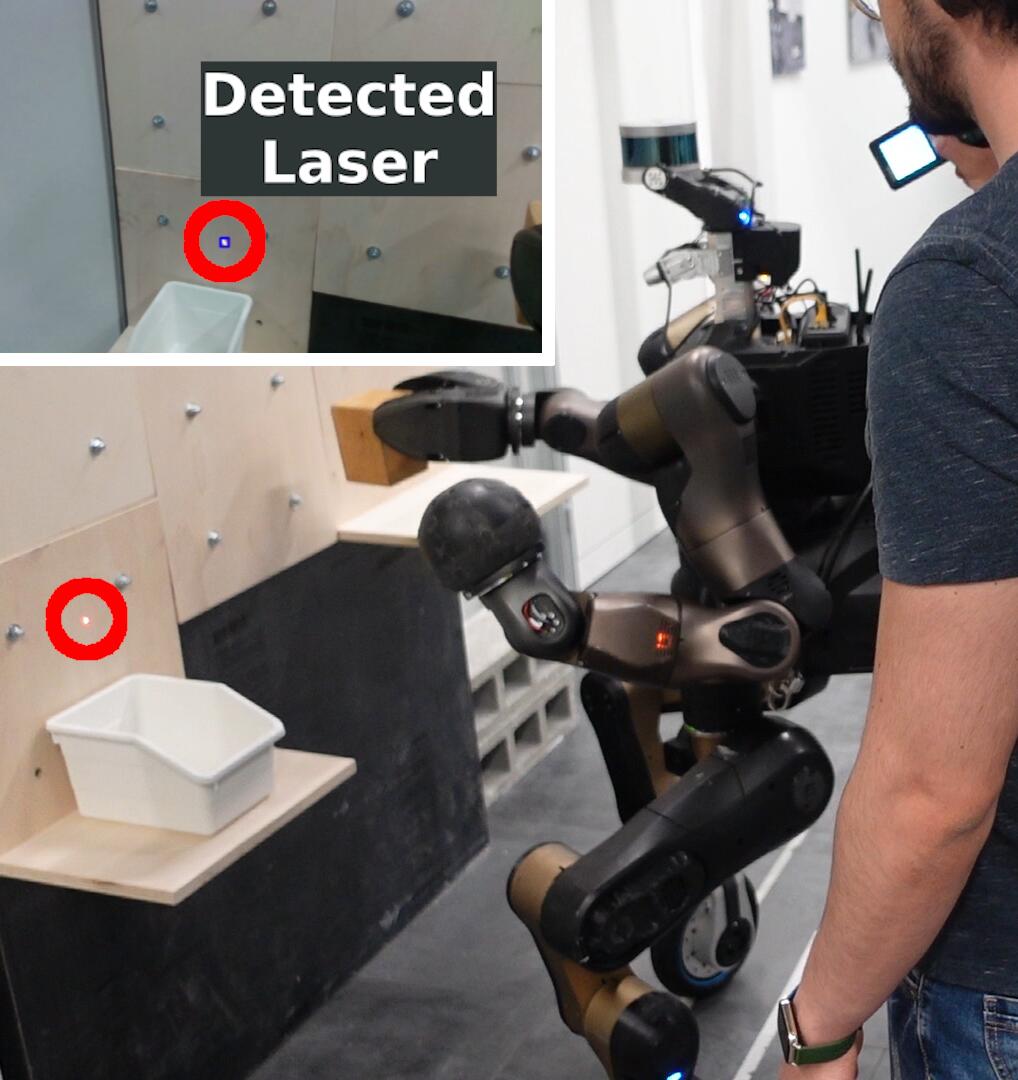}
	\includegraphics[height=0.35\linewidth, keepaspectratio]{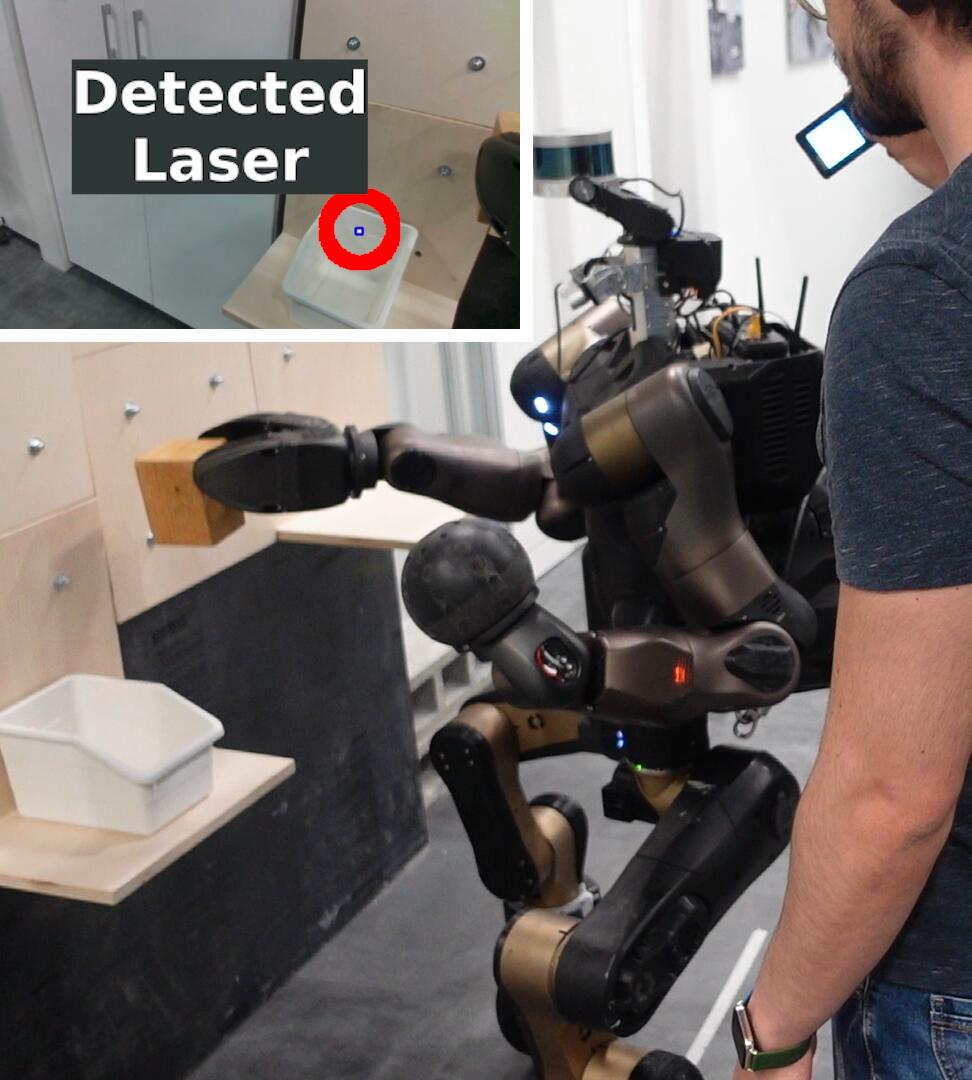}
	\includegraphics[height=0.35\linewidth, keepaspectratio]{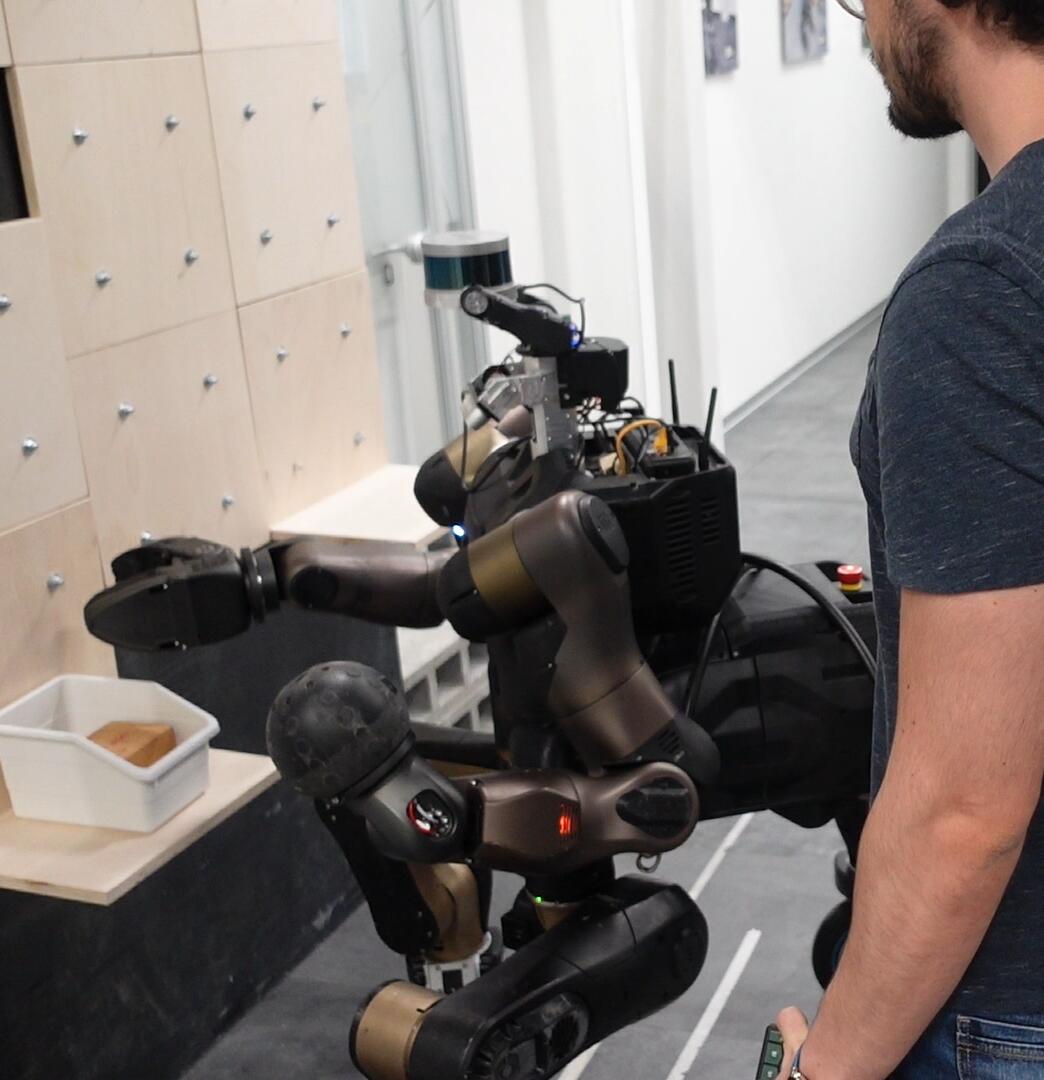}
	\caption[Laser-guided interface: pick-and-place experiment]{Sequences of the pick-and-place experiment. By pointing the laser in the environment, the operator commands the robot to reach the object to grasp it. Then, he guides the arm toward the container to place the object inside it.}
	\label{fig:laser:pickAndPlacePhoto}	
\end{figure}

\begin{figure}
	\centering
	\includegraphics[width=1\linewidth]{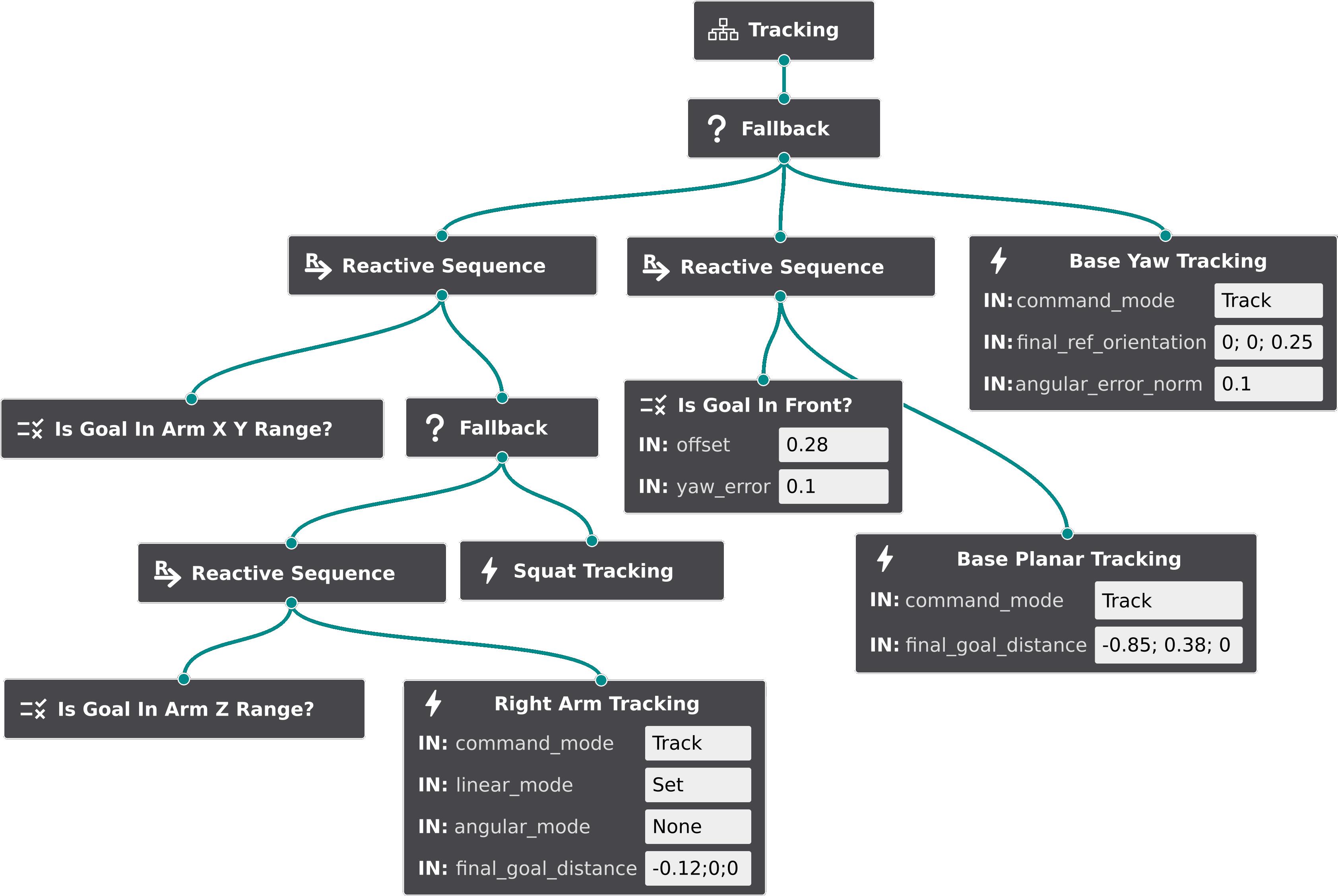}
	\caption[Laser-guided interface: pick-and-place experiment \acrshort{bt}]{The Tracking subtree for the pick-and-place experiment. This is the same \acrshort{bt} of \figurename{}~\ref{fig:laser:trackingBT}, but in this image the relevant parameters of the nodes are shown. According to them, the robot moves its base to keep the laser on its right side, to be better positioned for reaching it with the right end-effector.}
	\label{fig:laser:pickPlaceBt}
\end{figure}

This experiment involves the locomotion, manipulation and grasping abilities of the CENTAURO robot equipped with a custom gripper, the DAGANA, on the right arm to perform a pick-and-place task. 
The mission involves guiding the robot toward an object to grasp and transport it inside a container, as visible in \figurename{}~\ref{fig:laser:pickAndPlacePhoto}.
In this scenario, the complete tree presented in Section~\ref{sec:laser:btflow} is exploited. The Tracking subtree of \figurename{}~\ref{fig:laser:trackingBT} is detailed further in \figurename{}~\ref{fig:laser:pickPlaceBt}, showing the relevant parameters of the action modules. 

Since the mission characteristics, the \acrshort{bt} models the robot behavior in such a way to make the body follow the laser spot by keeping an offset that allows a good placement for the right end-effector. In particular, the laser spot is tracked keeping it slightly on the right side of the robot. Such behavior can be easily configurable using the parameters of the action modules.
Indeed, the \textit{Base Yaw Tracking} node has a \texttt{final\_ref\_orientation} of \SI{0.25}{\radian} for the yaw angle, and the \textit{Is Goal In Front?}\ condition node includes a desired \texttt{offset} and a \texttt{yaw\_error}. This condition returns \texttt{success} when the laser is not exactly in front of the robot, but it is at an angle of \SI[separate-uncertainty = true]{0.28(10)}{\radian}, i.e., slightly on the right.
In line with this strategy, the \textit{Base Planar Tracking} node has a \texttt{final\_goal\_distance} value of \SI{0.38}{\meter} along the $\hat{y}$-axis. An offset of \SI{-0.85}{\meter} along the $\hat{x}$-axis is also present to allow enough room for the arm to reach the goal with the end-effector. 

As told previously, this Tracking subtree is part of the main tree of \figurename{}~\ref{fig:laser:mainBT}. Accordingly to the main tree, the user can request additional discrete commands related to the gripper actions, at any time. Indeed, in the experiment, once the gripper is near enough the object, the user requests a Gripper Grasp, causing the control flow of the \acrshort{bt} to responsively halts the tracking of the laser spot to execute this other subtree. Once grasped the object, the Tracking subtree is activated again with another request.
Similarly, the tracking is halted once the gripper is near the container, following the user's request of a Gripper Open Action, to successfully release the object inside the container.

\begin{figure}
	\centering
	\includegraphics[width=1\linewidth]{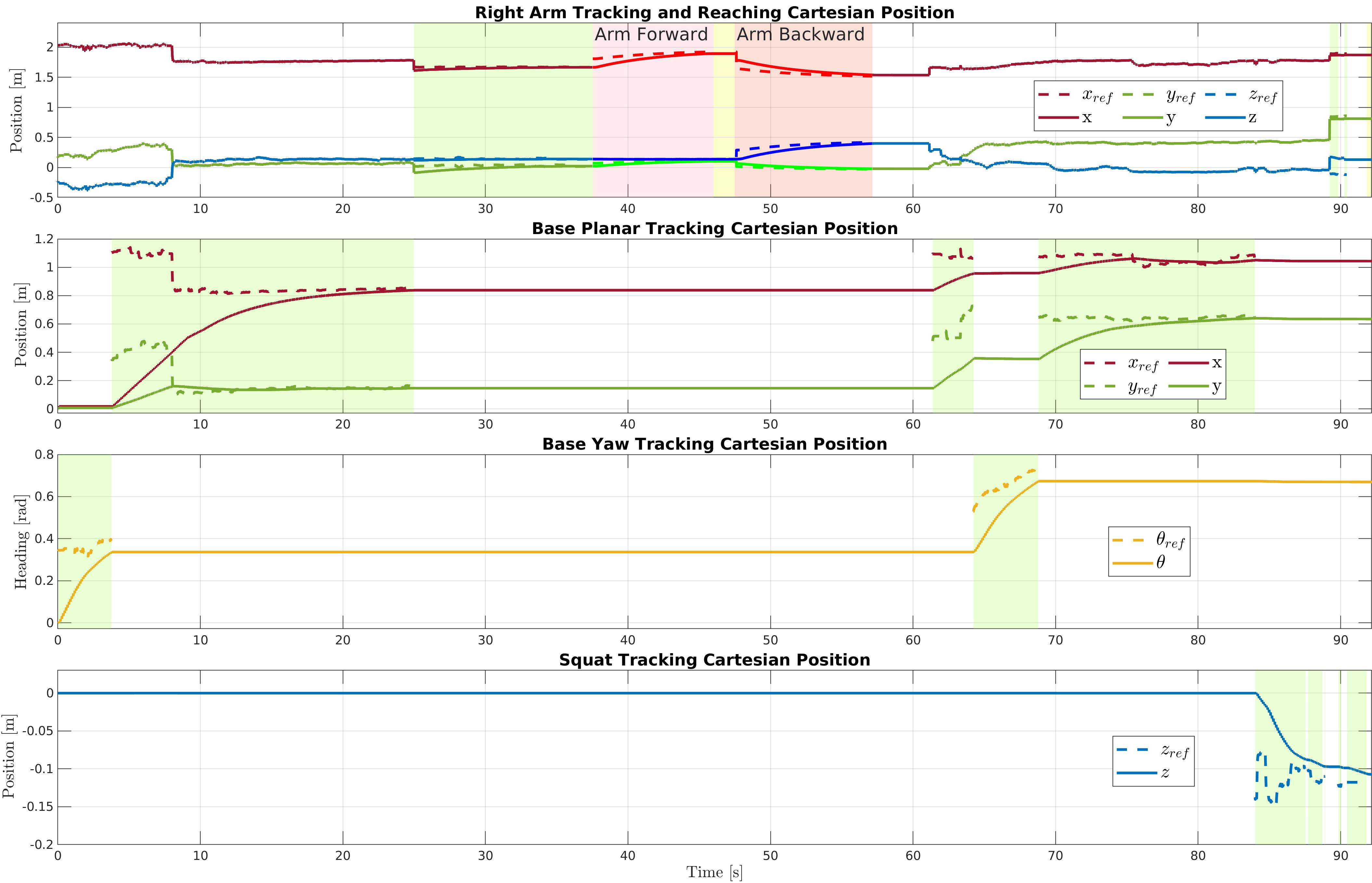}
	\caption[Laser-guided interface: pick-and-place experiment plots]{Plots for the pick-and-place experiment. The colored areas highlight the alternating activation of the action modules, which are, from top plot to bottom plot: Right Arm, Base Planar, Base Yaw, and Squat. Dashed lines represent the reference positions (with eventual offsets included), while continuous lines represent the position of the controlled robot frame of the task (right end-effector for the first task, robot pelvis frame for the others).}
	\label{fig:pickPlacePlot}
\end{figure}

In \figurename{}~\ref{fig:pickPlacePlot}, plots of are shown. Similarly to the previous experiments plots, highlighted areas
show the alternating activation of each action module. Different colors are used in the top plot to highlight different action module related to the right arm.
The dashed lines represent the references to track (including the offsets), and the continuous lines represent the positions of the controlled robot frame of the task (i.e., right end-effector for the Arm task and robot pelvis for the Base Planar Base Yaw, and Squat tasks). 
As modeled by the \acrshort{bt}, the Gaze Action is always active, but no data is shown.

In the time interval [$t=37s$, $t=57s$], the grasping phase is visible, running according to the Gripper Grasp subtree of \figurename{}~\ref{fig:laser:graspBT}. The control flow of this subtree makes the gripper to move forward before closing and backward and upward after. 
Here, the end-effector is not tracking the laser. Instead, in the indicated Arm Forward and Arm Backward sections, the reference to track is a predefined position with respect to the actual one, configured with \texttt{command\_mode == Reach} and \texttt{final\_goal\_distance} parameters of the action module. Also, \texttt{angular\_mode} is set to \texttt{Keep} to maintain the actual end-effector orientation (note that such parameters are not shown in the \figurename{}~\ref{fig:laser:graspBT}). 
Between the Arm Forward and Arm Backward, the Close Gripper node is executing. 

As in the previous experiment of Section~\ref{sec:laser:loco}, a continuous tracking of the laser spot is enabled thanks to the interface developed. In this experiment, more robot capabilities are explored.
The reactive nature of the BT allows the operator to exploit such capabilities agnostically and intuitively, simply by pointing the laser, without the need to command  which robot ability to use (locomotion, arm, squatting).
Apart from the laser, the only other user's inputs required involves grasping and releasing the object. However, these requests are handled flexibly by the \acrshort{bt}, which shows reactiveness by promptly adapting the control flow to accommodate them.

\section{Conclusions}\label{sec:laser:conclusions}

This chapter has presented a laser-guided human-robot interface for commanding even highly-redundant robot in an intuitive manner.
The interface relies on a low-cost laser emitter that the operator project in the environment as he/she would naturally point to location of interest when communicating with other people. Since this is a kind of supervisory control~\cite{Selvaggio2021}, the robot autonomy is intrinsically necessary to generate the robot motions (Section~\ref{sec:laser:intro}).   

To facilitate a seamless human-robot interaction, the reactivity in tracking the laser spot and in generating accordingly specific robot motions plays a key role in the development of the architecture. This is composed by a perception layer that utilizes a neural network model for the laser spot detection, and a motion generation layer based on Behavior Trees (BTs) for planning and executing the robot motions (Section~\ref{sec:laser:concept}).

The perception layer is capable of real-time laser spot tracking, allowing for swift reactions to changes in laser spot position.
The solution adopted is precise and robust across different surfaces, but it can be further specialized  by training the neural network to adapt to specific environmental conditions (Section~\ref{sec:laser:perception}).

The generation of robot's motion through the plan defined by the BT enables the robot behavior to be highly reactive to changes in conditions, such as variations in the position of the goal or discrete user requests. This allows the robot to rapidly adapt its movement style as required. 
Furthermore, the modular architecture developed exhibits the flexibility to be tailored to different missions and robots. Each individual action module can be customized using the available parameters, and their relationships can be easily configured by the logic of the \acrshort{bt} (Section~\ref{sec:laser:control}).

The experiments have demonstrated that the proposed interface is effective in controlling highly-redundant robots with minimal effort for the user. The architecture allows guiding the robot both to discrete waypoints and along a continuous path. The robot can track the goal by leveraging its locomanipulation abilities, following a predetermined logic that is easy to configure according to the task's needs and to the robot in use (Section~\ref{sec:laser:exp}).

In the upcoming Chapter~\ref{chap:Laser2}, it will be shown how the laser-guided interface concept can be applied to a robotic manipulation assistive scenario.

There is room for further improvements in both the perception and motion generation layers. 
Implementing object recognition techniques can enable the robot to distinguish between different types of objects and the environment. 
In this way, the robot will be able to understand what it is being pointed, and choose the appropriate strategy. This can be combined with the possibility to issue specific commands by drawing patterns with the laser projections, to augment the range of possible user's inputs always following a very effortless and intuitive communication.
Additionally, more robot capabilities can be considered in the motion planning, like dual-arm manipulation and specific leg movements. This will enable to accomplish more complex tasks and to navigate in more challenging environments.

\chapter{A Laser-guided Interface for Robotic Assistance}\label{chap:Laser2}

\begin{figure}[H]
	\centering
	\includegraphics[width=\linewidth]{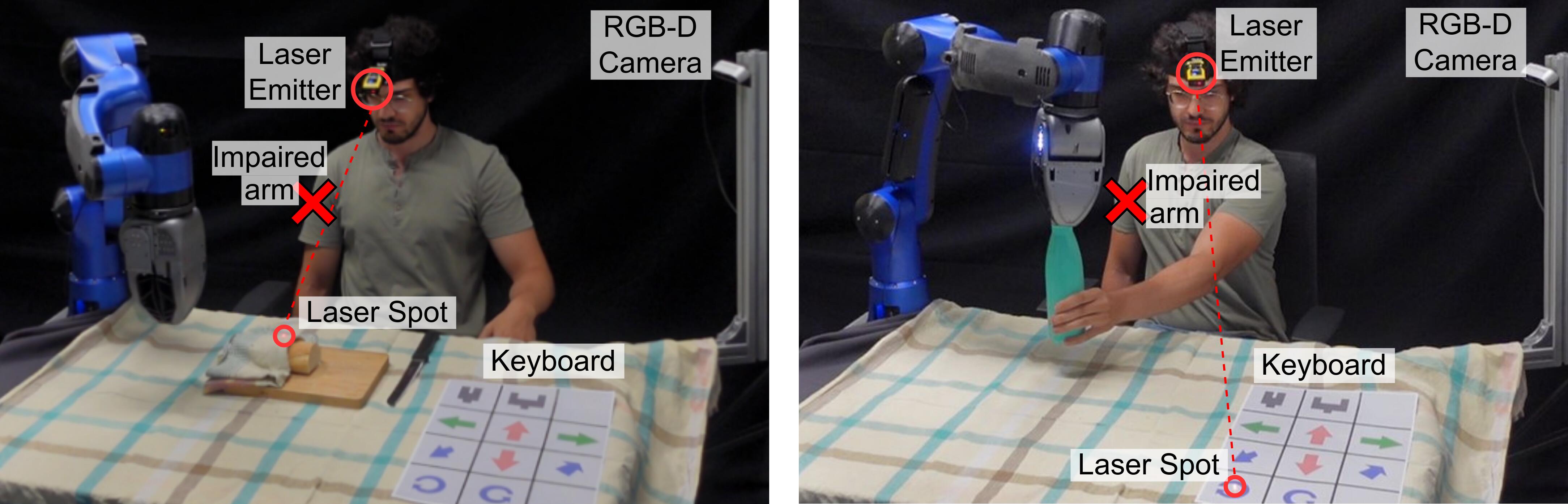}
	\caption[Laser assistive interface use cases]{The developed laser-guided interface provides assistance to users with arm impairments, by providing an intuitive interaction to control an assistive manipulator. The images show some of the validations presented in Section~\ref{sec:laserAssistive:exp}.}
	\label{fig:laserAssistive:firstPhoto}
\end{figure}

\lettrine{I}{n} the previous chapter it has been presented an intuitive human-robot interaction interface built upon the employment of a laser emitter device to let the user to effortlessly command the robot by pointing the laser to locations of interest in the environment.
In this chapter, such interface is realized for an assistive scenario, to help impaired arm users in manipulating objects controlling a robotic arm (\figurename{}~\ref{fig:laserAssistive:firstPhoto}).\\

\noindent This chapter is based on the following article:\\
\fullcite{LaserRal}~\cite{LaserRal}

\noindent The described laser interface has been employed also in the following work:\\
\fullcite{Bertoni2023}~\cite{Bertoni2023}

\section{Laser-guided Assistive Interaction Interface}\label{sec:laserAssistive:system}

In the previous chapters, the management of the complexity of the robot, in the sense of handling their wide range of motion capabilities, was one of the challenges addressed by the human-robot interface. In this chapter, the scenario is different, since the challenge is more about the limited amount of input possibilities that injured people may have to control the robot. 
In assistive scenarios, the key features of intuitiveness, situational awareness and robot autonomy are anyway of paramount importance, to let users be able to control robots without mental and physical effort, hence facilitating the application of robots in people's everyday lives.
 
The human-robot interface presented in this chapter relies on the laser-guided interaction of the previous Chapter~\ref{chap:Laser1}, which allows users to effortlessly indicate point of interest to the robot with a simple laser emitter device. 
Specifically, an interface is presented aimed to assist individuals with upper limbs motion impairments in reaching locations of interest and manipulate objects with a robotic manipulator.
Considering the particular application, the lightweight and compact laser pointing device is conveniently worn on the user's head, permitting to command the robot with head movements that result in pointing the laser. 

The system offers two distinct modalities to interact with the assistive robot based on the detected laser spot position. 
In the first modality, the robot end-effector is commanded to reach the specific point indicated by the laser. If the point is physically reachable by the manipulator, the system generates and executes a collision-free trajectory to reach the indicated goal. This way of interacting with the robot is very intuitive and natural as it resembles the action of people to orient their head and look toward an object of interest to interact with.
In the second modality, a more direct robot control is enabled employing a paper keyboard, placed in a designated area, whose buttons can be virtually pressed by projecting the laser onto them. Some buttons are mapped to Cartesian directions, enabling the user to command the robot through Cartesian end-effector velocities, while other buttons give control to the functionalities of the gripper mounted on the arm.

Previous works on assistive human-robot interfaces that relies on particular body movements, like head~\cite{Kyrarini2019, Rudigkeit2020}, eyes~\cite{Sunny2021, Sharma2022} or tongue~\cite{Mohammadi2021}, necessitate the user to learn and deal with a certain mapping between the particular body movement and the input provided to the robot. 
Instead, with the presented interface, the head movements result in pointing the laser in the environment or in the keyboard buttons, intuitively commanding the robot without any prior knowledge of the system. 
Furthermore, to exploit the movements of the user's head as a command input, it is not necessary to have a specific tracking device. Instead, a \acrshort{rgbd} camera looking at the scene is employed to track the projected laser point.
Additionally, the user has a real-time visual feedback about the command given to the robot through the perception of the laser spot, without the necessity of any additional communication means like a monitor, an augmented reality device, or some auditory feedback.

The advantages of laser-based approaches for assistive human-robot interfaces are shown in the previous works mentioned in Section~\ref{chap2:soa:assistive}. Usually, the laser was used to select an object to grasp~\cite{Nguyen2008, Choi2008, Gualtieri2017, Wilkinson2021, Zhong2019} or to manipulate~\cite{Liu2021, Liu2023}. 
Differently, the functionality of our interface does not focus on the selection of an object considering its automatic grasping. %
Instead, the interface aims to facilitate the regulation of the robot motion and interaction toward any point of interest, still maintaining the intuitiveness of the laser guidance approach.
By combining seamlessly the two control modalities, we provide a combination of robot autonomy and low level Cartesian control that provide flexibility for various tasks. This allows to not only face grasping tasks, but also perform co-manipulation tasks with the robot, to replace the functionality of the impaired limb. For example, the robot can be guided to hold the bread in a point chosen by the user, such that he/she has enough space to cut it (left image of \figurename{}~\ref{fig:laserAssistive:firstPhoto}).
Furthermore, allowing the control of the end-effector after the grasping, the user can move the object where he/she prefers, or it is more convenient, like near his/her healthy arm to open a bottle (right image of \figurename{}~\ref{fig:laserAssistive:firstPhoto}). 
In addition, as demonstrated in other experiments (Section~\ref{sec:laserAssistive:exp}), pick-and-place tasks can be still achieved even without an automatic grasping system utilized in other works.
Compared to the previous works that made use of a laser based interface, apart from the added flexibility in the control of the robot motion, providing such functionalities greatly increases the interaction of the human with the robot. This augmentation enhances the sense of involvement of the human subject in the regulation of the execution of task. Hence, impaired users can play an active part in the task, which is important to help them regain his/her sense of autonomy~\cite{Kim2012, Bhattacharjee2020}.

Another advantage of the proposed interface is that no additional input is required to command manipulation actions on the object, since the user can manipulate it by commanding the robot with the paper keyboard using the same laser pointer device. This relaxes the need for employing additional devices that may require the engagement of the healthy upper limb (e.g.\ pressing additional buttons, making gestures, etc.).
For example, in \cite{Liu2023} the authors show how flashing the laser can be employed to command different manipulation actions, but this requires switching on and off the laser, hence employing the user hand.
Indeed, the user's arm should be left free from other motion activities in order to be fully available to collaborate with the robot arm in bimanual tasks.
Other input systems that do not require engaging the healthy upper limb, like vocal recognition, would complicate the system, may lack robustness especially in noisy environments, and, in some cases, may be not feasible if impaired users have speech difficulties due to their condition. 

In conclusion, the two modalities well combine to provide a variety of possibilities, from commanding a target location, to controlling the full end-effector pose and activating the gripper. The keyboard modality also allows reaching locations that cannot be indicated by the laser (e.g.\ because of occlusions or the absence of a surface to project the laser), and to account for any potential errors in the end-effector pose achieved with the other modality. In the end, recognizing also the importance of offering different levels of autonomy~\cite{Kim2012, Bhattacharjee2020}, the interface permits the user to choose the best control modality based on the task and their preferences, switching between them simply by adjusting the laser position.

\noindent In summary, the main features and contributions of this work are:

\begin{itemize}
	
	\item A highly intuitive interface for people with upper limbs impairments is introduced. The interface explores a head-wearable laser pointing device to indicate to the assistive robot the locations of interest related to a task to be executed within the robot workspace. With this interface, head movements in the direction of a specific location of interest result in the laser to point to it. This is a very intuitive way to indicate a target position to the robot end-effector, for example to approach an object to grasp. 
	\item In addition to commanding the robot to a target position, the interface integrates an additional keyboard control modality. Indeed, the laser can be used to virtually press the buttons of a paper keyboard located in an area of the environment, with each button of the keyboard mapped to gripper actions and specific end-effector directions.
	This allows the user to utilize the laser to command the gripper of the robot and the robot itself through Cartesian end-effector velocities. 
	\item The laser point projection is detected by employing the neural-network based vision system described in Section~\ref{sec:laser:perception}, which utilizes images from an RGB-D (Red Green Blue-Depth) camera directed at the robot workspace. The detection pipeline is fast enough to allow for a reactive response to laser position changes. The code for the laser spot detection is available \href{https://github.com/ADVRHumanoids/nn_laser_spot_tracking}{https://github.com/ADVRHumanoids/nn\_laser\_spot\_tracking}.
	
\end{itemize}

\section{System Architecture}\label{sec:laserAssistive:methods}
\begin{figure}[H]
	\centering
	\includegraphics[width=1\linewidth]{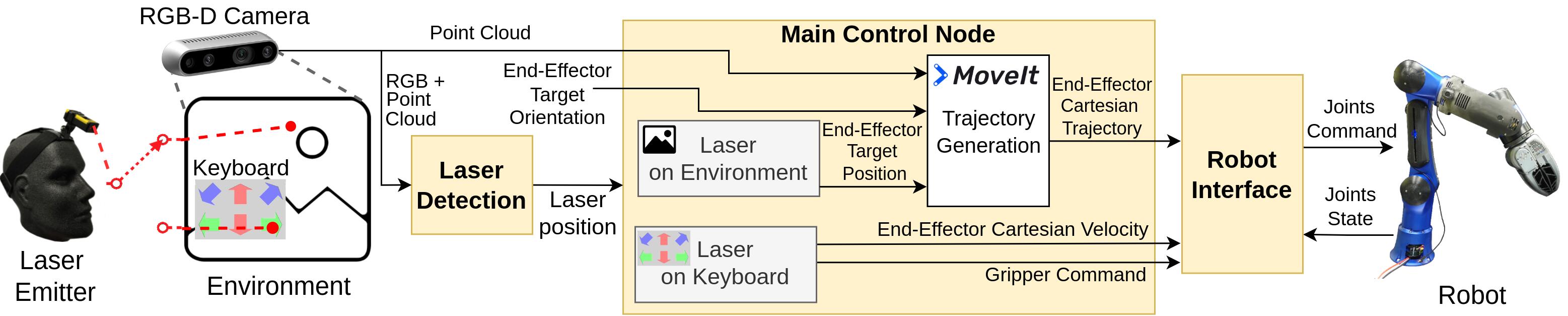}
	\caption[Laser assistive interface scheme]{Scheme representing the overall interface. Depending on where the laser is pointed (environment or keyboard), the relative control modality is exploited to generate robot commands.}
	\label{fig:laserAssistive:controlScheme}
\end{figure}

An overall scheme of the architecture is shown in \figurename{}~\ref{fig:laserAssistive:controlScheme}. The Laser Detection block, in charge of detecting the laser spot projection, has already been detailed in Section~\ref{sec:laser:perception}. 
In the Main Control Node, based on the position of the laser projection, one of the two control modalities is exploited, generating different commands for the robot.
At the rightmost part of the scheme, the Robot Interface is in charge of performing the inverse kinematic, and of communicating with the robot, exploiting the tools presented in Section~\ref{sec:intro:framework}.

\subsection{Environment Control Mode}
When the user points a generic location in the environment, the system interprets the laser spot position as an end-effector goal position $\boldsymbol{x} \in \mathbb{R}^3$. To avoid selecting unwanted locations, the user must keep the laser spot sufficiently still in a position for a certain amount of time, to make the system accept the laser spot as an end-effector goal. 
The target position $\boldsymbol{x}$ is merged with an end-effector target orientation set accordingly to the task necessity. Considering the experimental setup chosen, the end-effector target orientation is parallel to the table surface, to eventually permit to easily grasp an object placed on the table. 
Future interface developments will include the automatic computation of the target orientation depending on the properties of the object pointed.

The end-effector target pose is used to compute a Cartesian trajectory by using MoveIt~\cite{moveit}, the de facto standard motion planner for \acrshort{ros}~\cite{ROS}. MoveIt takes care of generating a trajectory avoiding singularities, self-collisions, obstacles detected by the same camera used to detect the laser, and restricted regions. In the experiments conducted, such regions include fixed bounding boxes drawn around the user and the table where the manipulator is mounted.
It is worth mentioning that in a parallel work additional proximity sensors mounted on the arm are explored to avoid dynamic obstacles during the execution of a task~\cite{Bertoni2023}.

\subsection{Keyboard Control Mode}
\begin{figure}[H]
	\centering
	\includegraphics[width=0.58\linewidth]{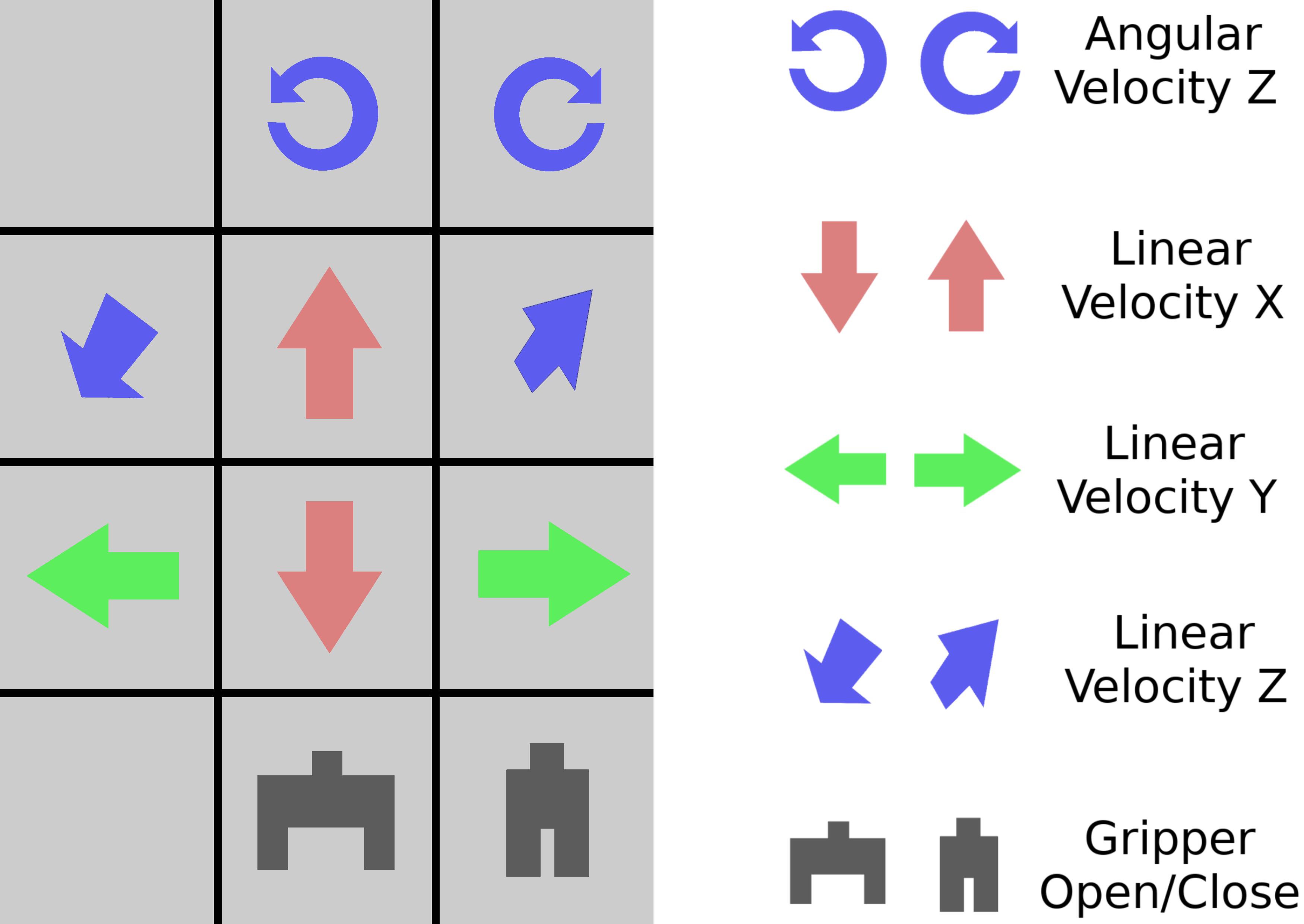}
	\caption[Laser assistive interface keyboard]{The configuration chosen for the paper keyboard, with a legend of the included buttons to be virtually pressed by the laser.}
	\label{fig:keyboard}
\end{figure}

The keyboard modality is a more direct way of controlling the robot, since the user, by pointing the laser to the buttons of the paper keyboard, directly commands robot's functions such as end-effector Cartesian velocities and gripper actions. 

In the keyboard's layout chosen, shown in \figurename{}~\ref{fig:keyboard}, the keyboard has six buttons for commanding the linear Cartesian velocities (one positive-verse and one negative-verse for each of the three axis $\hat{x}$, $\hat{y}$, $\hat{z}$), two buttons for commanding the angular Cartesian velocities along the $\hat{z}$ axis of the end-effector (positive and negative), and two buttons for opening and closing the gripper. 
The directions are relative to a reference frame which has the $\hat{x}$-axis along the width of the table, the $\hat{y}$-axis along the length, and the $\hat{z}$-axis perpendicular to its surface, pointing up. Regarding the angular velocity, the reference frame has the $\hat{z}$-axis pointing out from the gripper.
Each button of the keyboard has a known size (a rectangle of $0.105$m x $0.099$m), so the relative position from the keyboard center is known, and the keyboard itself is fixed in a certain position with respect to the robot. 
Hence, a button is considered virtually pressed when the laser spot is detected in the button's specific area of the environment. Similarly to the environment control mode, a button is considered pressed after the laser is kept in its area for a certain amount of time.
When the user selects a keyboard button related to the end-effector velocities, the wanted direction and verse are commanded with a certain magnitude, while the gripper is commanded in a discrete open/close mode.

The buttons configuration and sizes as detailed here have been chosen according to the tasks' needs. 
In general, different choices can be made, for example the keyboard may be placed elsewhere with respect to the robot, and may include additional buttons for controlling other robot functions, like additional end-effector directions, or the speed of the motions.
Eventually, also the buttons can be placed separately and not fixed: in this case a calibration will be unpractical. A solution is to train the neural network model to detect the laser projected on the specific buttons to directly recognize the selection of the button.

\section{Laser-guided Assistive Interface Experimental Validations}\label{sec:laserAssistive:exp}

The presented assistive human-robot interface is evaluated with a series of \acrshort{adl} tasks where healthy users, simulating different upper limb impairments, collaborates with the robotic manipulator presented in Section~\ref{sec:intro:arm}, equipped the DAGANA gripper introduced in Section~\ref{sec:intro:dagana}.
Two sets of experiments are considered, each involving different kinds of upper limb disabilities.

In the first set, it is considered the case of individuals affected by stroke or other similar impairments that limit the motion capacity of a single arm, while the other arm is fully functional. In this scenario, the impaired arm is substituted by the robotic arm, which, guided by the user through the proposed interface, executes human-robot collaborative bimanual \acrshort{adl} tasks. These tasks consist in cutting some bread (Section~\ref{sec:laserAssistive:bread}) and opening a bottle (Section~\ref{sec:laserAssistive:bottle}).

In the second set, it is considered the case of people with impairments on both arms. Individuals with such disabilities require assistance in more \acrlong{adl}, including object transportation. In this scenario, pick-and-place tasks are accomplished by different users, where the assistive robot executes all the actions as instructed by the user through the control modalities provided by the proposed interface (Section~\ref{sec:laserAssistive:pickandplace}). 

All the experiments described have been recorded and shown in the video available at \href{https://youtu.be/abHbEAOdSMs}{https://youtu.be/abHbEAOdSMs}.

\subsection{Experimental Setup}\label{sec:laserAssistive:expSetup}

\begin{figure}[H]
	\centering
	\includegraphics[width=0.95\linewidth]{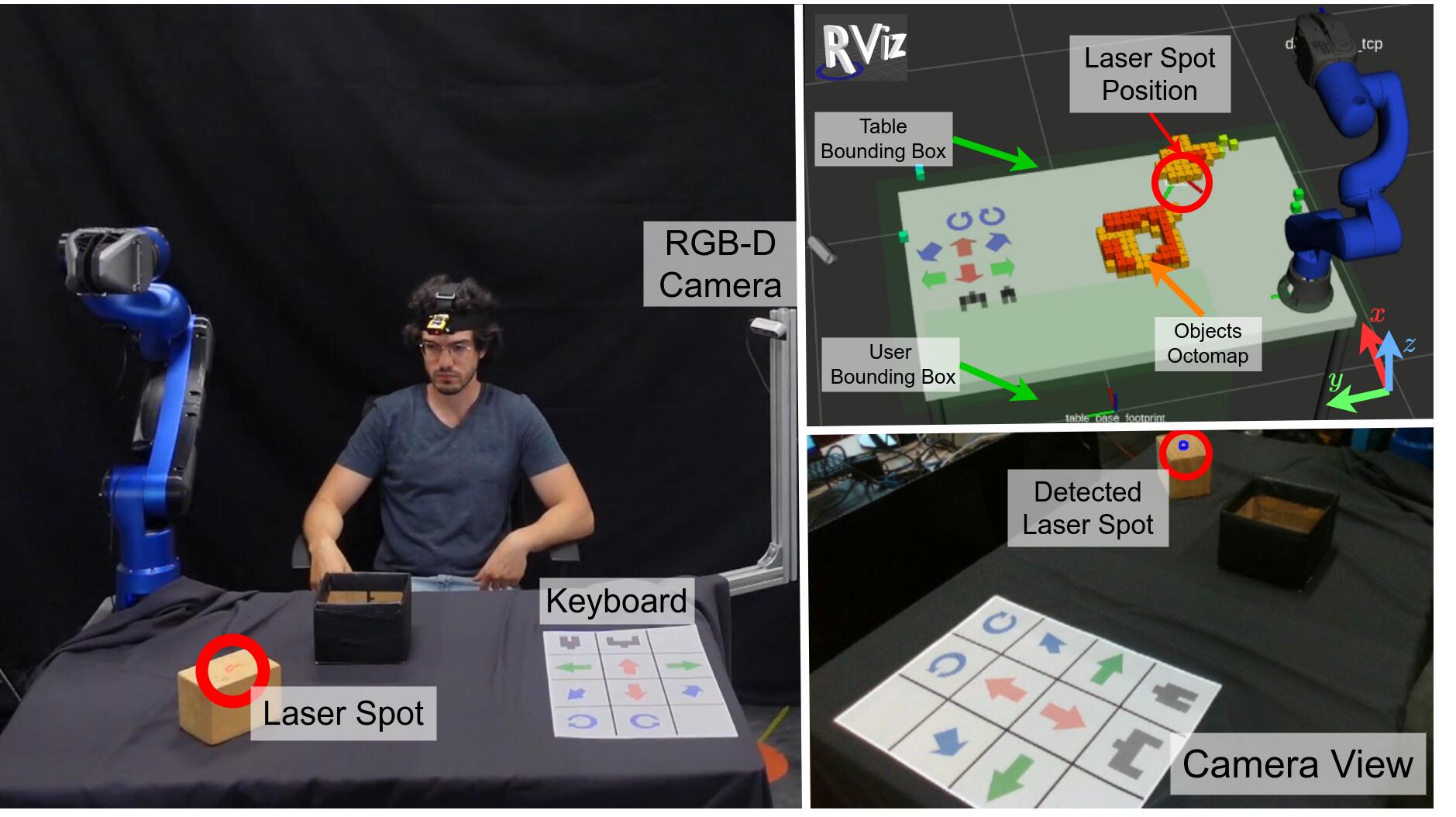}
	\caption[Laser assistive interface experimental setup]{Setup of the experiments with the robot visualization. On the left, the user commands the robot with the laser worn on the head. On the top-right, RViz displays the elements (the laser spot position and the obstacles) considered for generating the trajectory towards the goal. On the bottom-right, the laser spot is detected by the system.}
	\label{fig:laserAssistive:setup}
\end{figure}

\begin{figure}[H]
	\centering
	\includegraphics[width=0.75\linewidth]{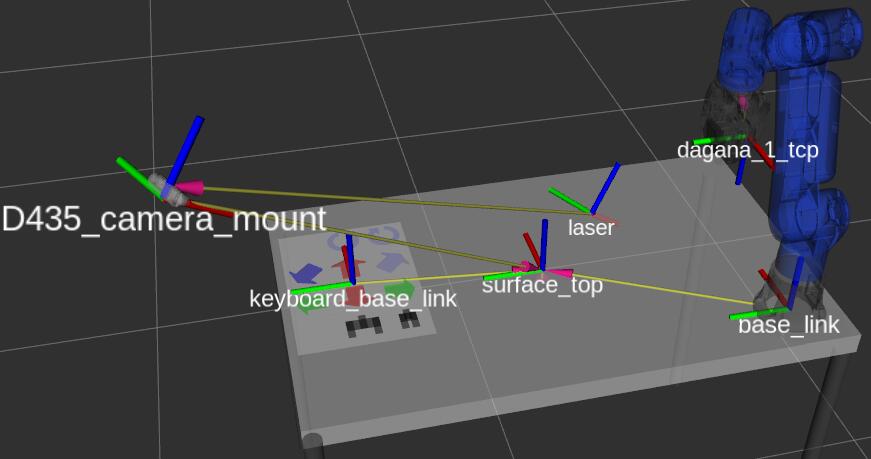}
	\caption[Laser assistive interface frames]{The frames of the relevant elements of the system. Please note that the laser orientation has no meaning and its frame is used solely as a marker to highlight its detected position.}
	\label{fig:laserAssistive:tfScheme}
\end{figure}

\figurename{}~\ref{fig:laserAssistive:setup} illustrates the setup of one the experiment which shares similar elements with all the others. 
In the image on the left, the user stays sit in front of a table, where the manipulator, the paper keyboard, and the camera are fixed on the surface. The camera installed in the scene, an Intel Realsense D435 RGB-D, is pointed at the table to provide images to track the laser spot and to detect potential obstacles.
To calibrate the camera position with respect to the robot, in a preliminary phase it is utilized an \textit{ArUco} marker\footnote{\href{http://wiki.ros.org/aruco_detect}{http://wiki.ros.org/aruco\_detect}} fixed in a known position above the table.
The frames of the relevant elements involved in the setup are shown in \figurename{}~\ref{fig:laserAssistive:tfScheme}.

The user, by moving the head, directs the laser to command the robot with one of the two control modalities (environment and keyboard) to perform the requested manipulation tasks. For example, in the figure shown, the user is utilizing the environment control mode to command the robot to reach the wooden block as indicated by the laser spot. 
The laser spot in the \acrshort{rgb} image is detected by the neural network, as shown in the bottom-right window of \figurename{}~\ref{fig:laserAssistive:setup}. The pixel coordinates are then matched against the depth images to retrieve the position of the laser, shown as a frame in the RViz windows (top-right area of \figurename{}~\ref{fig:laserAssistive:setup} in \figurename{}~\ref{fig:laserAssistive:tfScheme}). 
The RViz window of \figurename{}~\ref{fig:laserAssistive:setup} displays also two green bounding boxes around the user and the table, and an octomap generated by MoveIt from the \acrshort{rgbd} camera data.
These elements are the obstacles taken into consideration when generating a collision-free trajectory toward the detected laser spot.

For the experiments, the interface parameters are adjusted according to the necessity. With the environment control mode a particular pointed location is accepted as a goal after the spot has been kept in the location for at least $\SI{3}{\second}$, with a tolerance of $\SI{0.04}{\meter}$ radius.
For the keyboard control mode, the time is set to $\SI{1}{\second}$. This reduced duration is chosen to consider the placement of the keyboard aside, thus having a lower likelihood of accidentally pointing the laser in this region with respect to the other modality. The linear and angular velocities commanded with the keyboard have been set to $\SI{0.025}{\meter/\second}$ and $\SI{0.25}{\radian/\second}$, for the linear and angular part, respectively.

\subsubsection{Plots' Layout}\label{sec:laserAssistive:plotLayout}
In the following sections, most of the figures' plots adhere to similar conventions, here described. 

The first row displays the detected laser position with respect to the reference frame \textit{base\_link}. 

The second and third rows show the Cartesian pose (position and orientation) of the end-effector. Please note that the orientation is expressed in quaternions. These rows also show the Cartesian pose reference (dashed lines) which represents the target goal pose when operating in the environment control mode (gray areas).

The fourth and fifth rows display the Cartesian velocity (linear and angular) of the end-effector, referring as before to the \textit{base\_link} frame. These quantities have been computed from the derivation of the sensed robot joint positions, hence they have been post-processed with a moving average filter to improve the visualization. 

The last row presents the state of the gripper, showing its joint position (with position $0$ corresponding to the closed gripper) and joint effort.

The colored areas represent the time intervals during which the user is commanding the robot, corresponding either to the use of the environment control mode or to the activation of a specific button of the paper keyboard. In the top row plot, all the commands issued are shown as such areas. In the other plots, only the correspondent commands are shown (e.g., gripper commands are displayed in the plots relative to the gripper state).

\subsection{Cutting Bread Experiment}\label{sec:laserAssistive:bread}
\begin{figure}[H]
	\centering
	\includegraphics[width=0.48\linewidth]{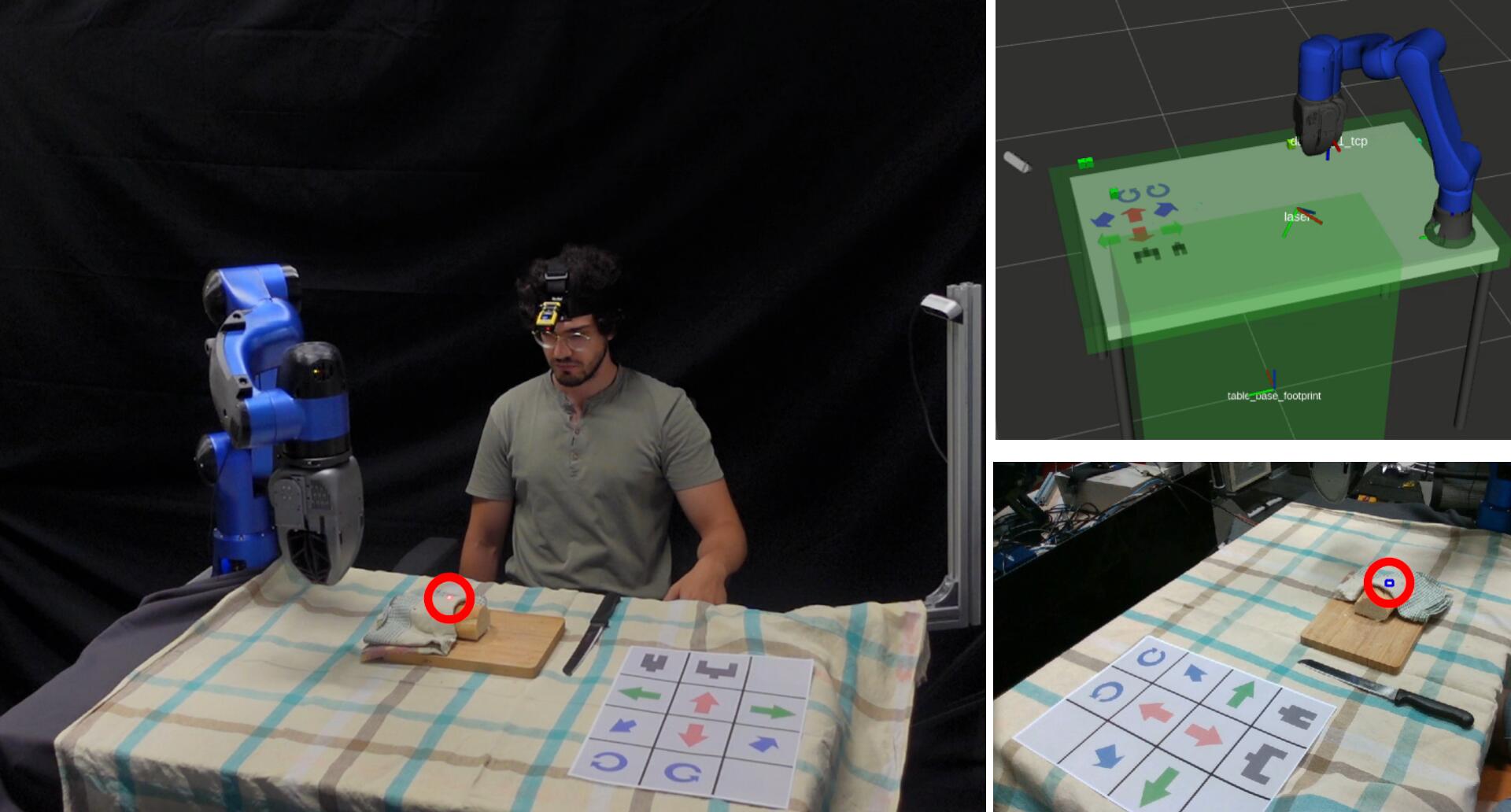}	\hspace{4px}
	\includegraphics[width=0.48\linewidth]{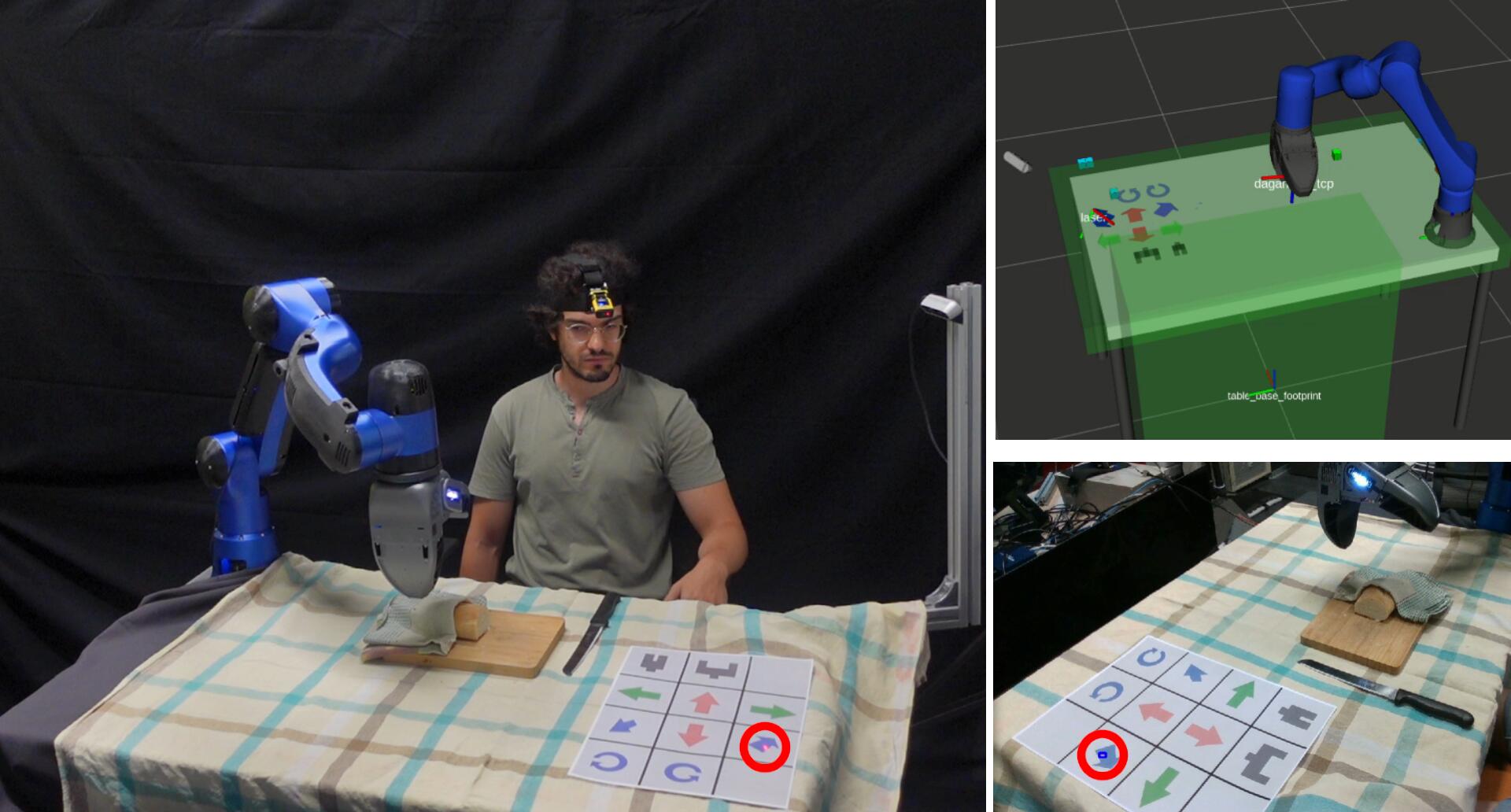}\\
		\vspace{6px}
	\includegraphics[width=0.48\linewidth]{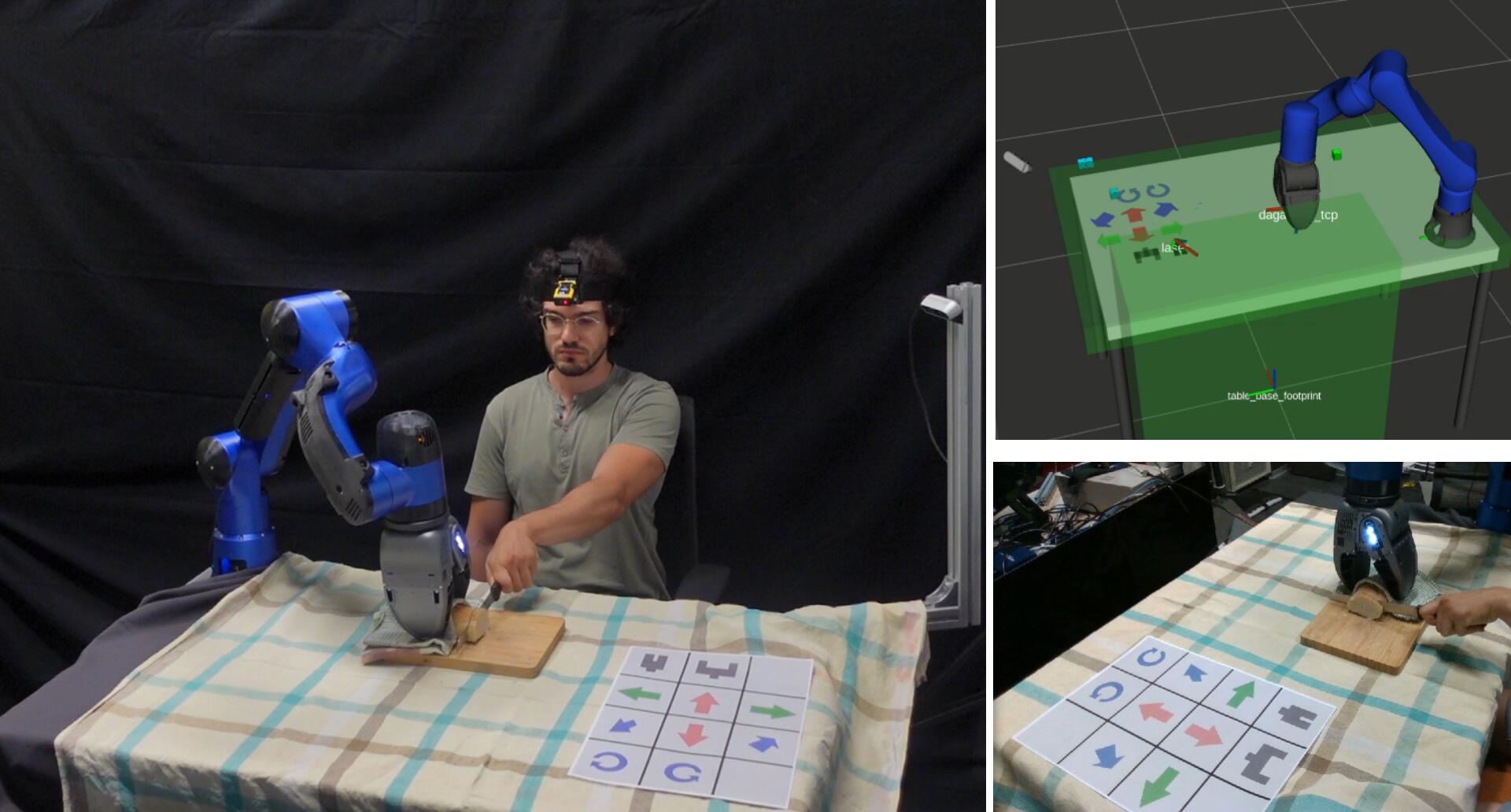} \hspace{4px}
	\includegraphics[width=0.48\linewidth]{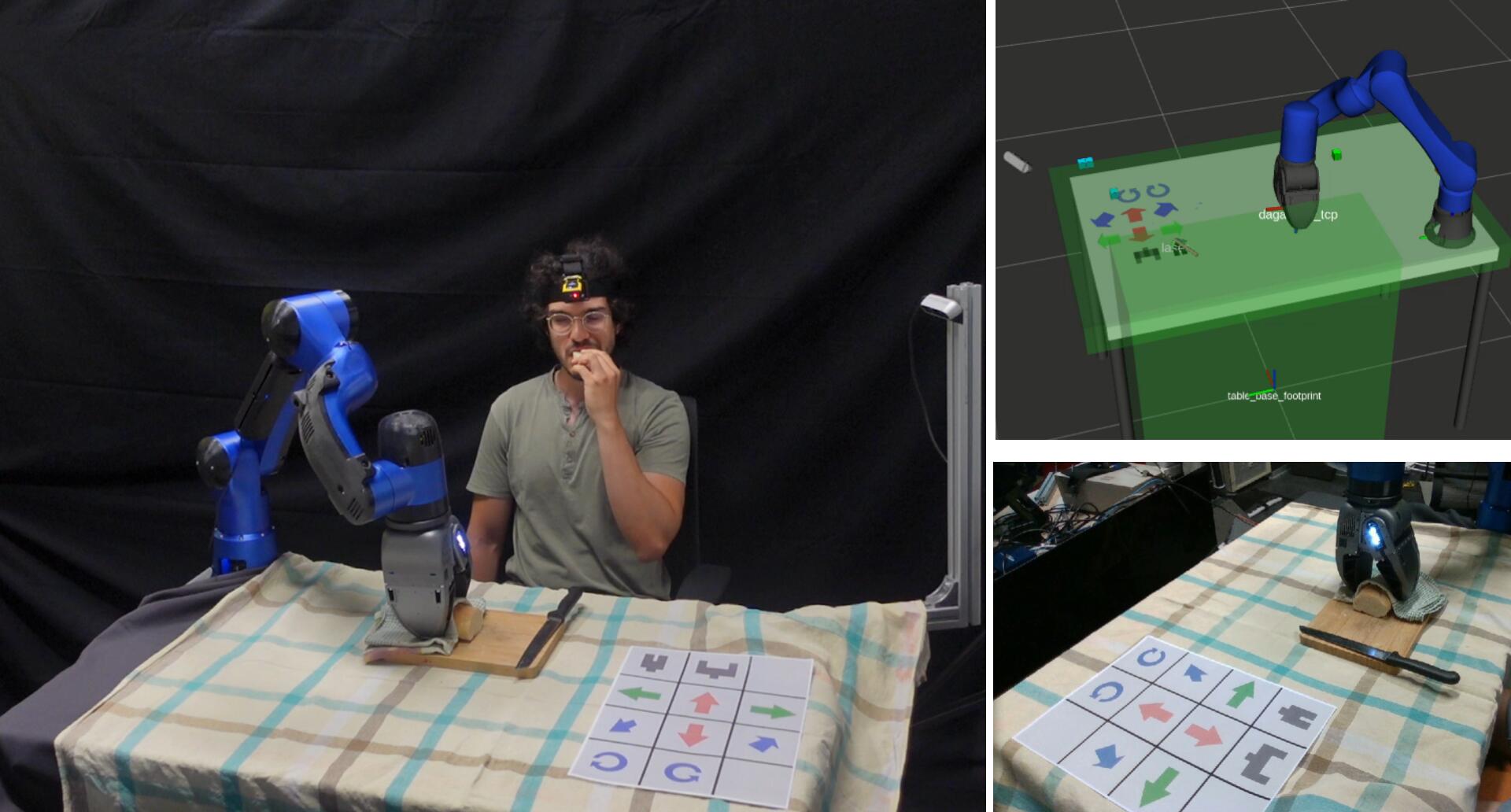}
	\caption[Laser assistive interface: cutting bread experiment]{Sequences of the cutting bread experiment, where the single impaired arm user collaborates with the robot to cut the bread. In each image, on the top-right the RViz window is shown, while on the bottom-right the camera view is added showing the detected laser spot.}
	\label{fig:pane3-frames}
\end{figure}

This experiment involves the \acrshort{adl} task of cutting some bread, as shown in \figurename{}~\ref{fig:pane3-frames}. At the beginning, the user commands the robot to reach the bread by directing the head at it which results in pointing the laser on it. Subsequently, by pointing at the paper keyboard, the user is able to control the robot end-effector commanding it to hold the bread. With the bread steadily held in the gripper, the user proceeds to cut it using his healthy arm.

\begin{figure}
	\centering
	\includegraphics[width=1\linewidth]{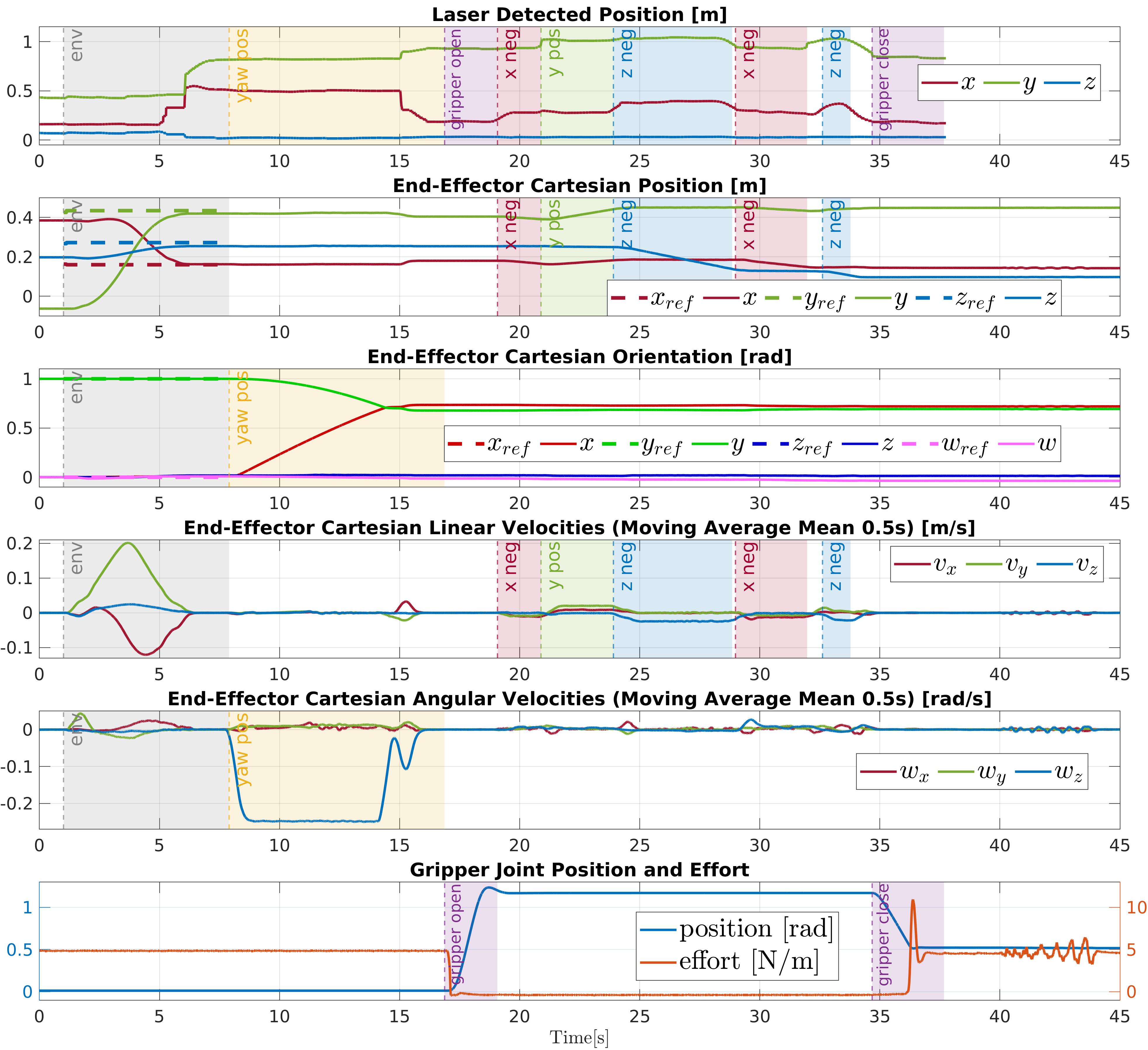}
	\caption[Laser assistive interface: cutting bread experiment plots]{\enquote{Cutting bread} experiment plots, highlighting the intervals when the user is commanding the robot with a specific modality. Plot's layout is explained in Section~\ref{sec:laserAssistive:plotLayout}.}
	\label{fig:pane3-plot-all}
\end{figure}

The corresponding plots are shown in \figurename{}~\ref{fig:pane3-plot-all}, whose layout is explained in Section~\ref{sec:laserAssistive:plotLayout}. At the end of the experiment, the laser position plot (first row) shows no data because the laser is outside the camera view since the user does not need to point anything while he is cutting the bread. It is also worth noticing that the gripper effort (bottom row) in this time interval exhibits chatter, attributed to the user interaction with the bread during the cutting phase.

\subsection{Bottle Experiment}\label{sec:laserAssistive:bottle}

\begin{figure}[H]
	\centering
	\includegraphics[width=0.48\linewidth]{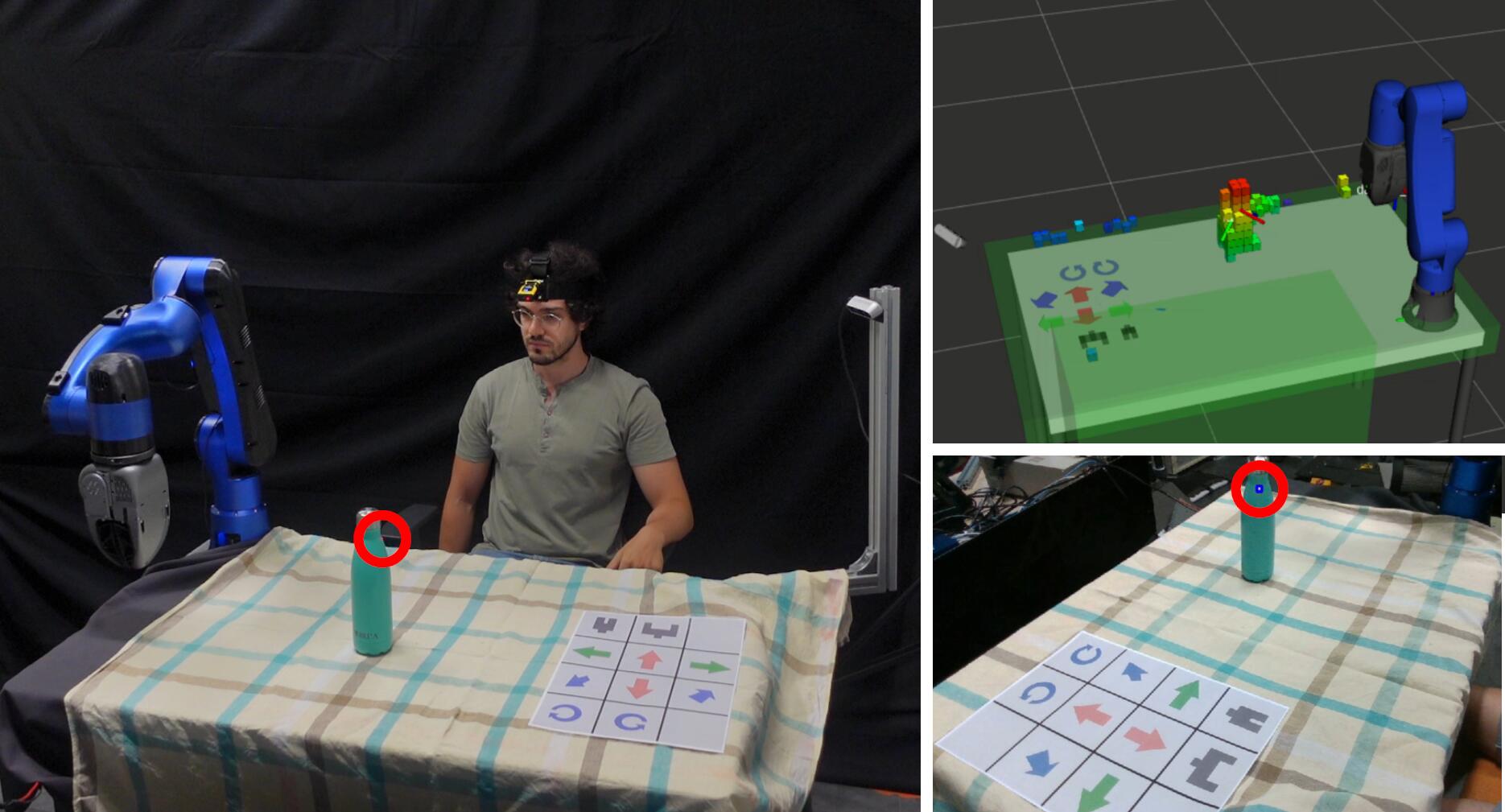}\hspace{5px}
	\includegraphics[width=0.48\linewidth]{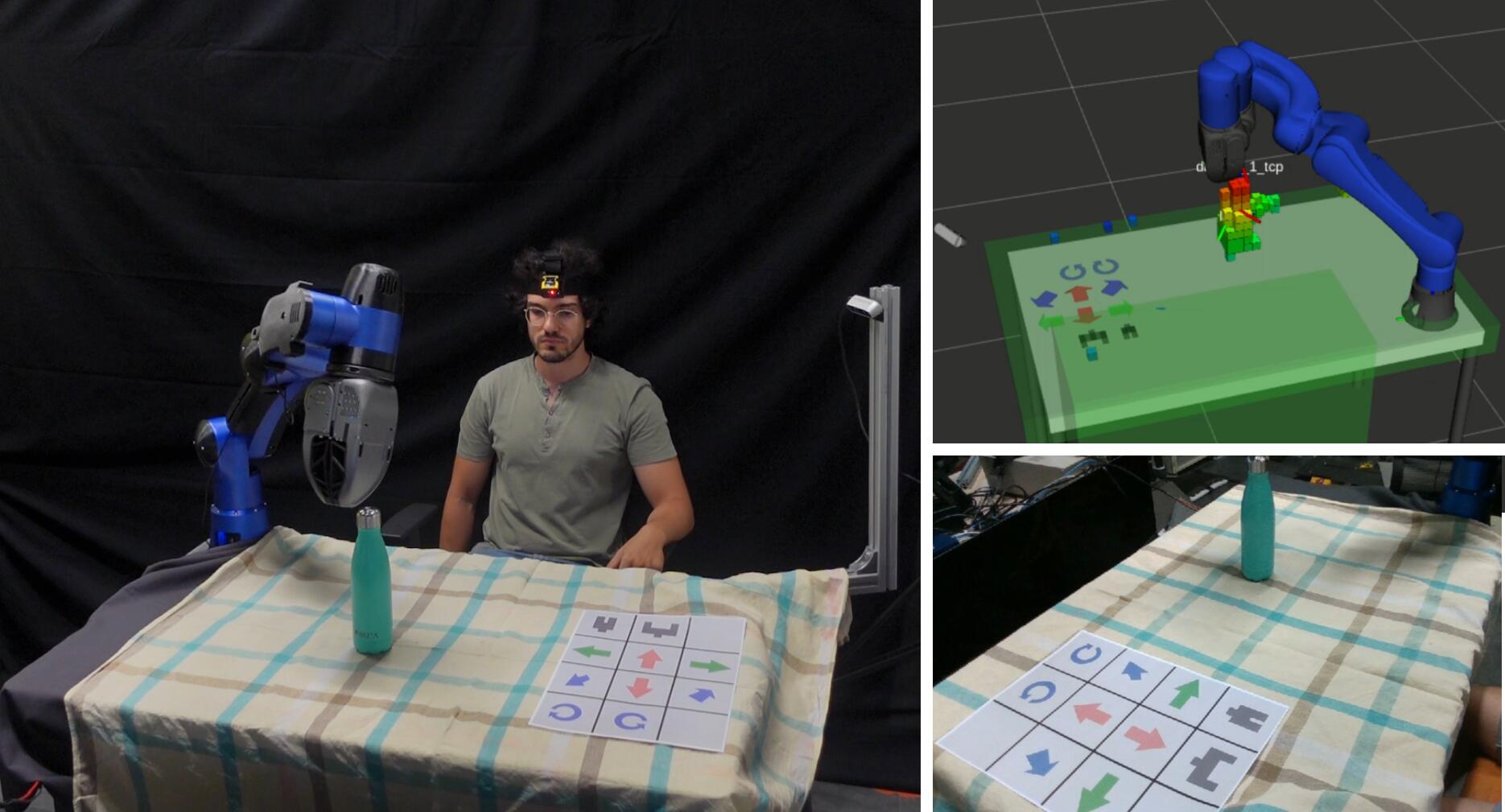}\\
	\vspace{5px}
	\includegraphics[width=0.48\linewidth]{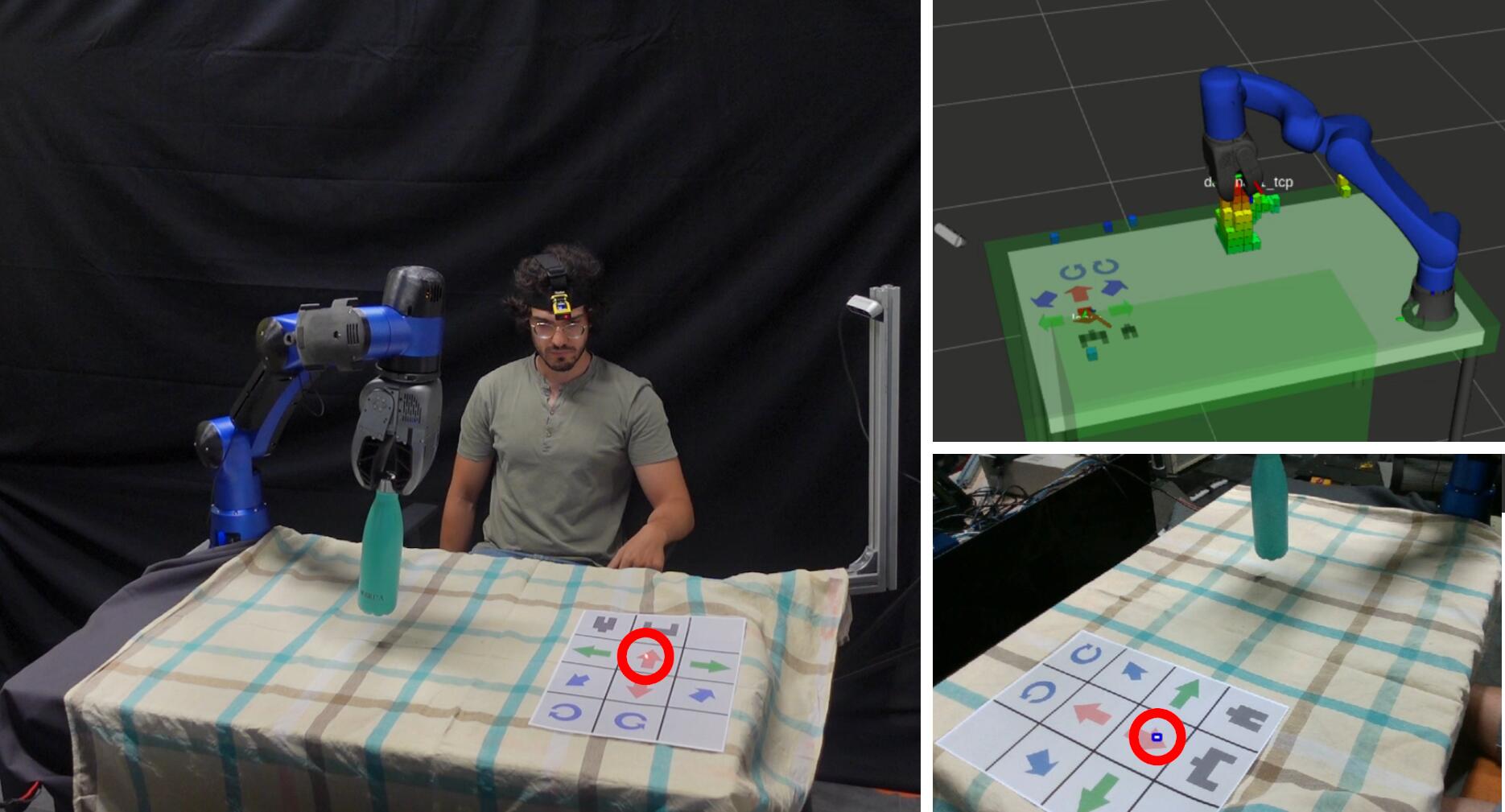}\hspace{5px}
	\includegraphics[width=0.48\linewidth]{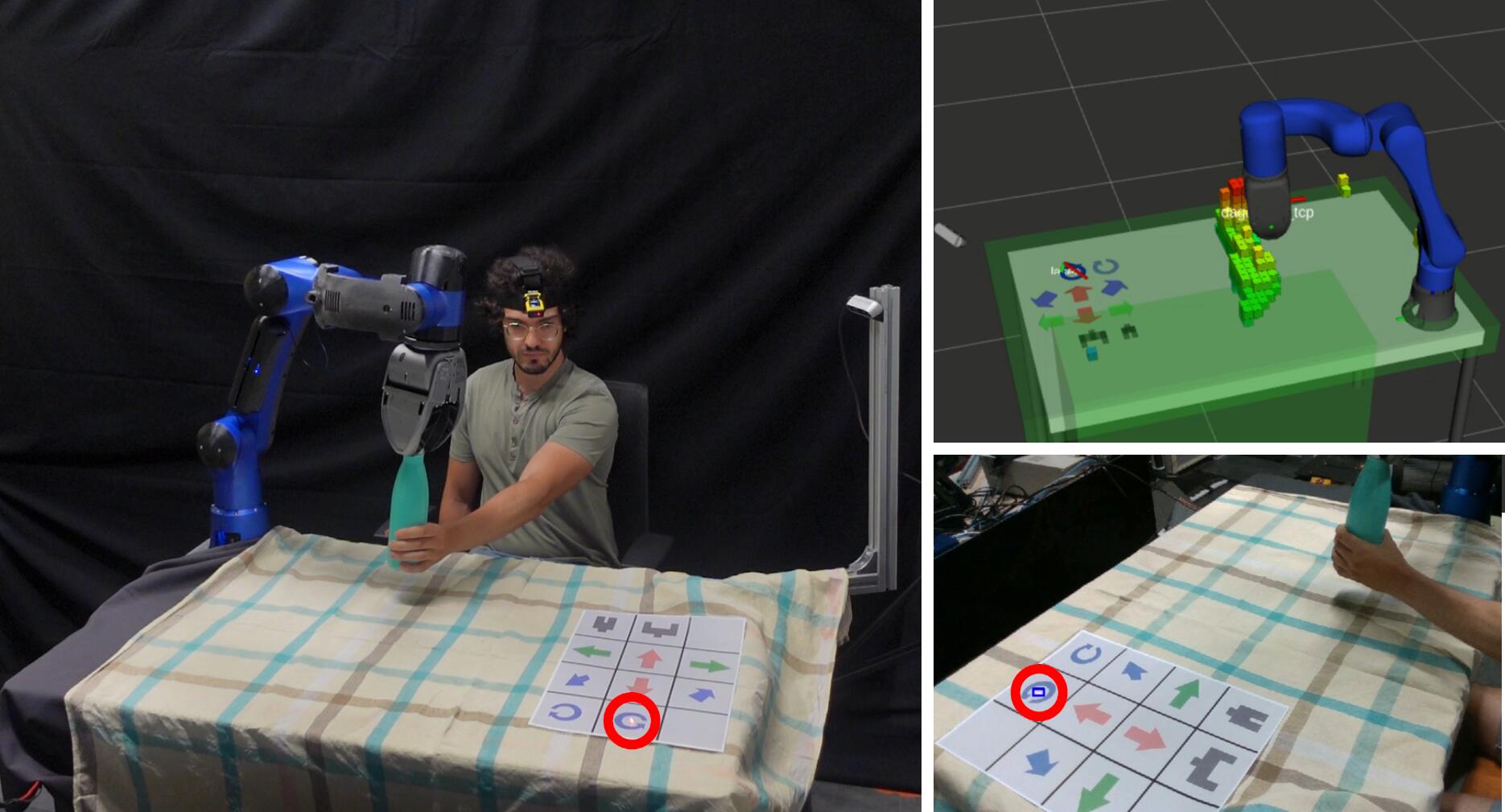}\\
	\vspace{5px}
	\includegraphics[width=0.48\linewidth]{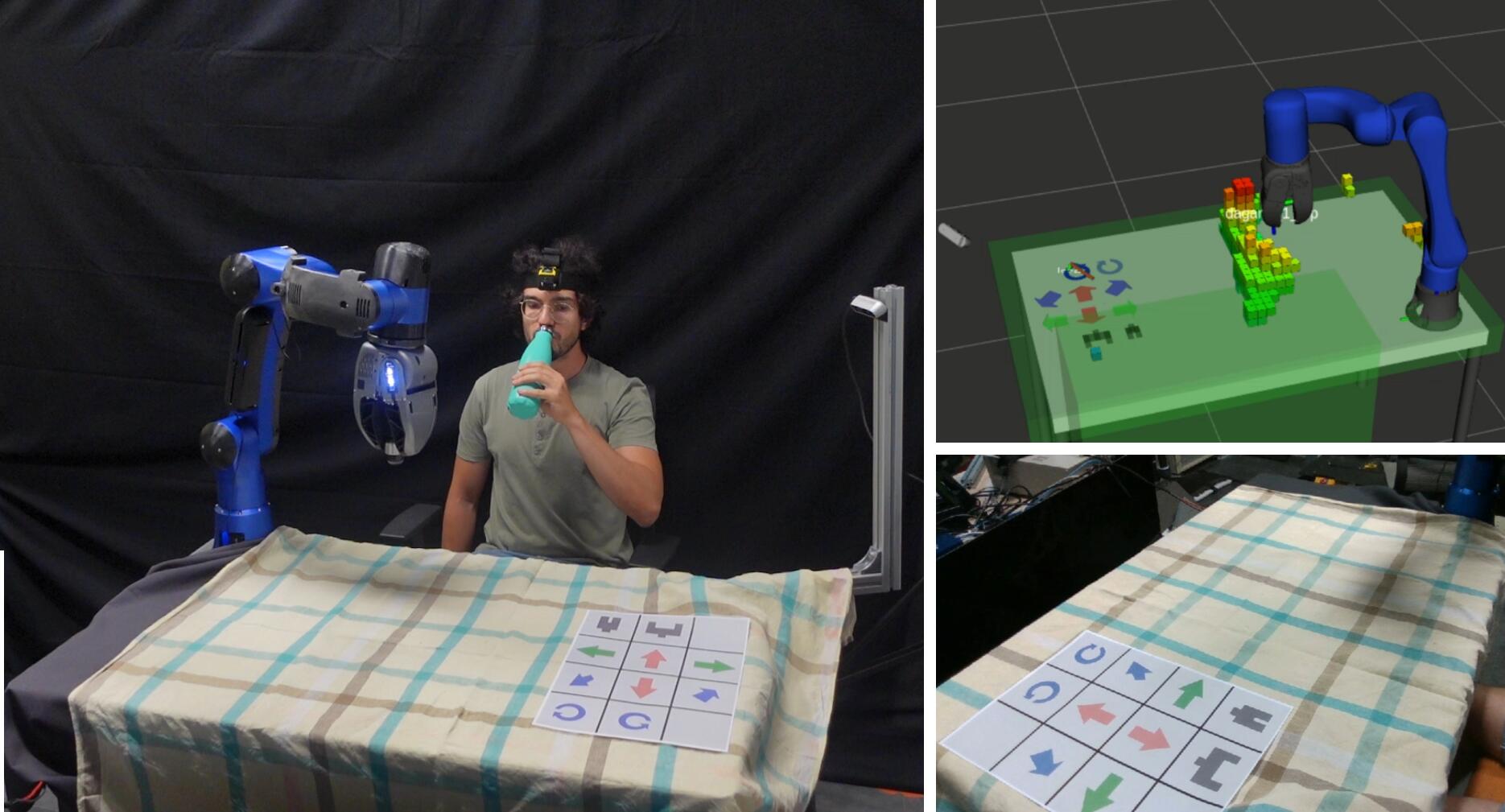}\hspace{5px}
	\includegraphics[width=0.48\linewidth]{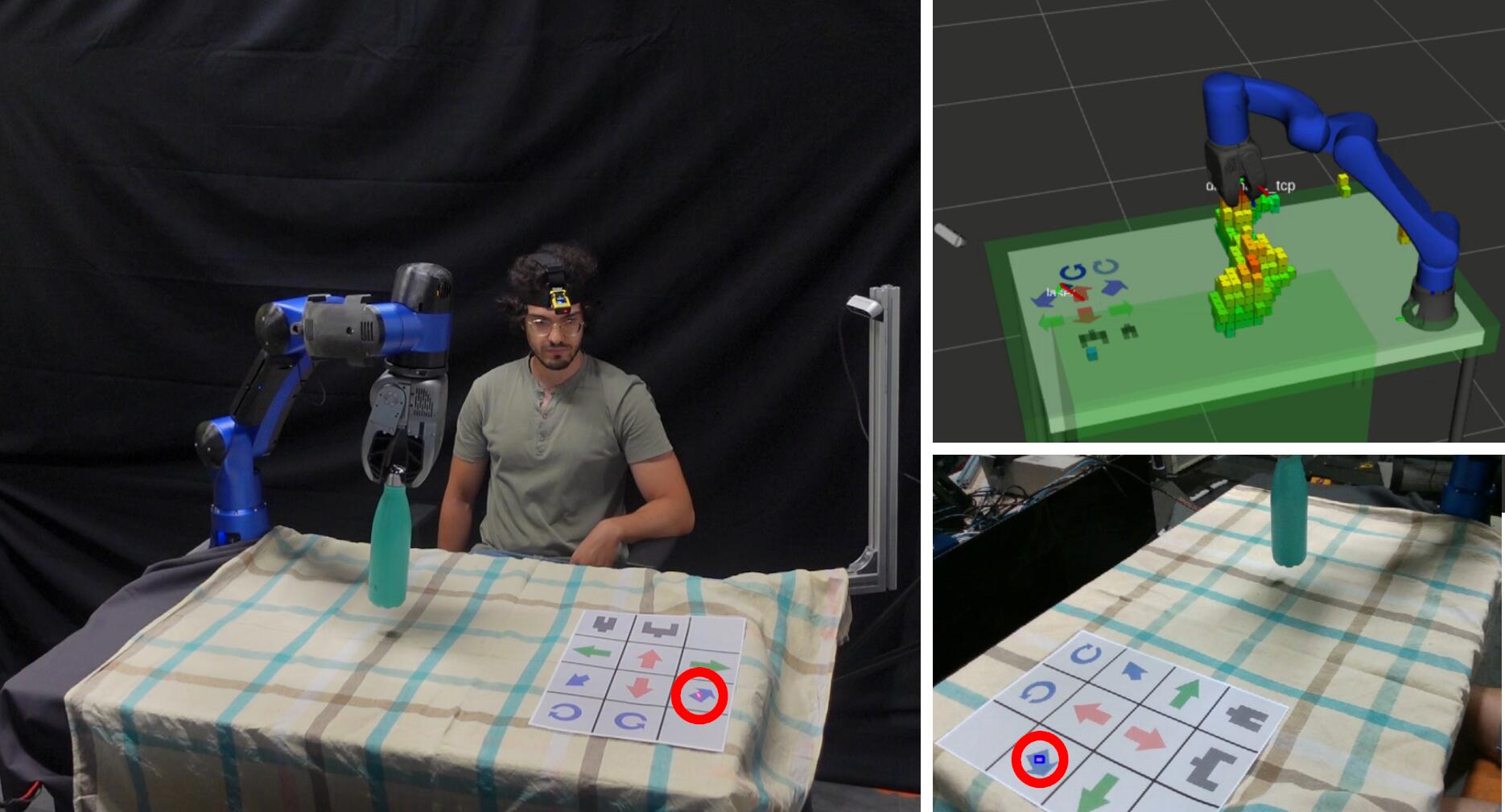}
	\caption[Laser assistive interface: bottle experiment]{Sequences of the bottle experiment, where the user collaborates with the robot to take a bottle and manipulate it to screw and unscrew the cap.}
	\label{fig:laserAssistance:bottiglia-frames}
\end{figure}

In this experiment, illustrated in \figurename{}~\ref{fig:laserAssistance:bottiglia-frames}, the user collaborates with the robot to open and close a bottle. Initially, the user points the laser at the bottle to command the robot to reach it. Subsequently, he employs the paper keyboard to grasp the bottle from the cap and bring it closer. With the robot holding the bottle from the cap, the user uses his healthy arm to grasp the body of the bottle while simultaneously commanding an end-effector yaw rotation pointing the laser at the correspondent keyboard button to unscrew the cap. After drinking, the subject brings the bottle back to the robot, commands an opposite end-effector yaw rotation to screw the cap, and then moves the robot to place the bottle back on the table.

\begin{figure}[H]
	\centering
	\includegraphics[width=1\linewidth]{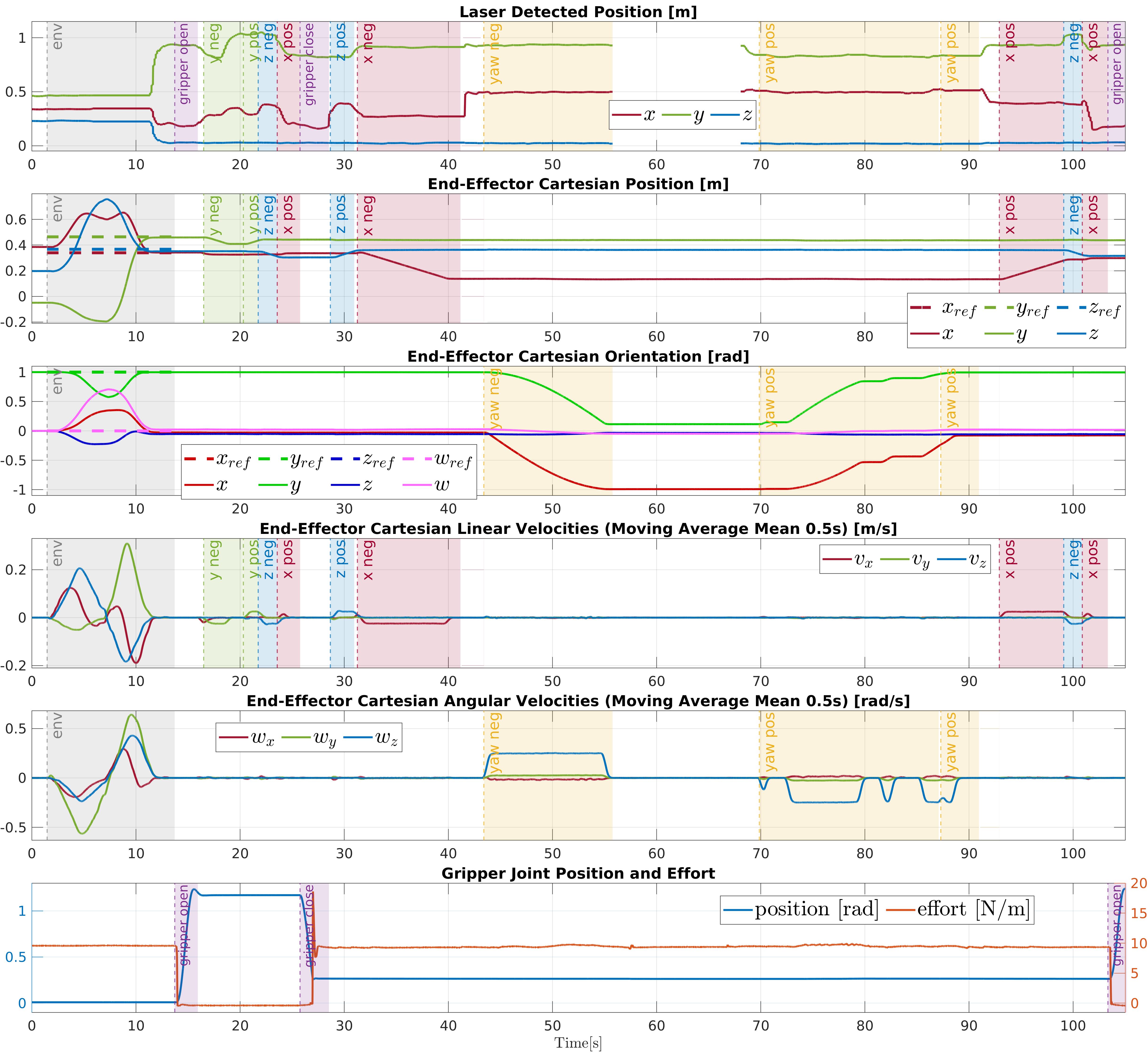}
	\caption[Laser assistive interface: bottle experiment plots]{Bottle experiment plots, highlighting the intervals when the user is commanding the robot with a specific modality. Plot's layout is explained in Section~\ref{sec:laserAssistive:plotLayout}.}
	\label{fig:laserAssistance:bottiglia-plot-all}
\end{figure}

Plots relevant to the bottle experiment are displayed in \figurename{}~\ref{fig:laserAssistance:bottiglia-plot-all}, with the same disposition as in the previous experiment's plots, and as explained in Section~\ref{sec:laserAssistive:plotLayout}. In the laser position plot at the top row, the interval around $\SI{60}{\second}$ shows no data because the laser is outside the camera view, which happens while the user is drinking.

\subsection{Pick-and-place Experiments}\label{sec:laserAssistive:pickandplace}
\begin{figure}[H]
	\centering
	\includegraphics[width=0.48\linewidth]{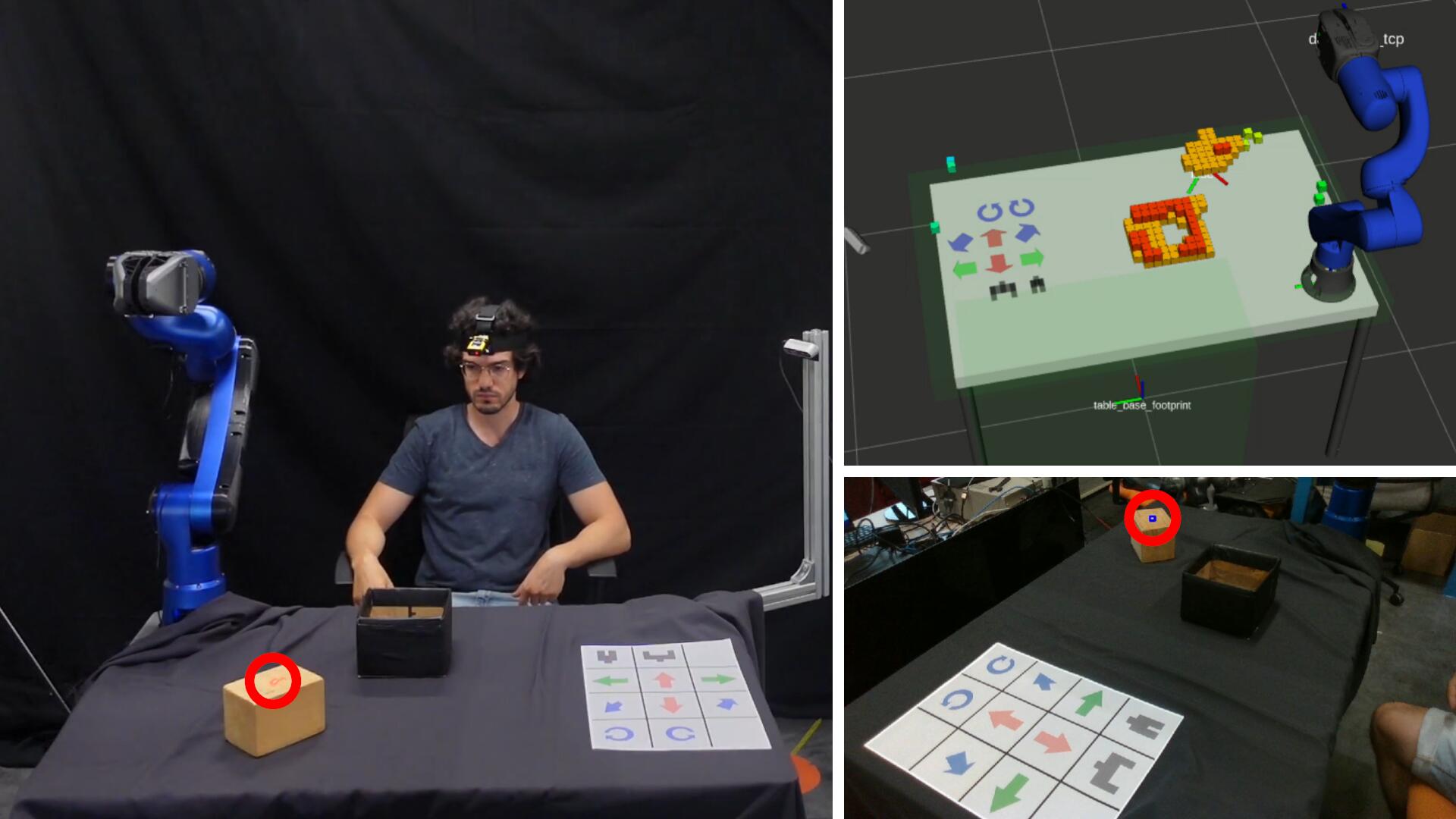}\hspace{5px}
	\includegraphics[width=0.48\linewidth]{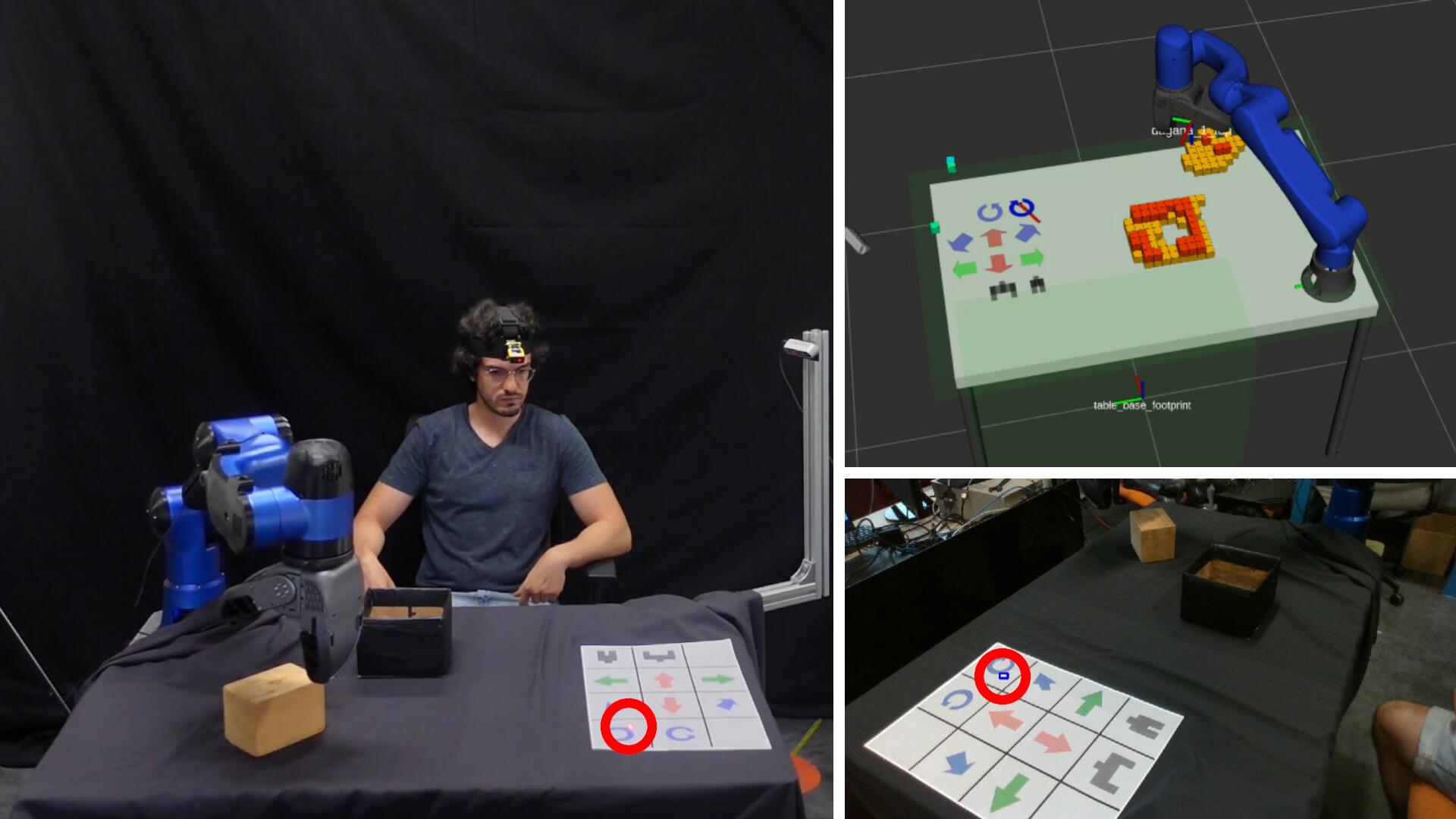}\\	\vspace{5px}
	\includegraphics[width=0.48\linewidth]{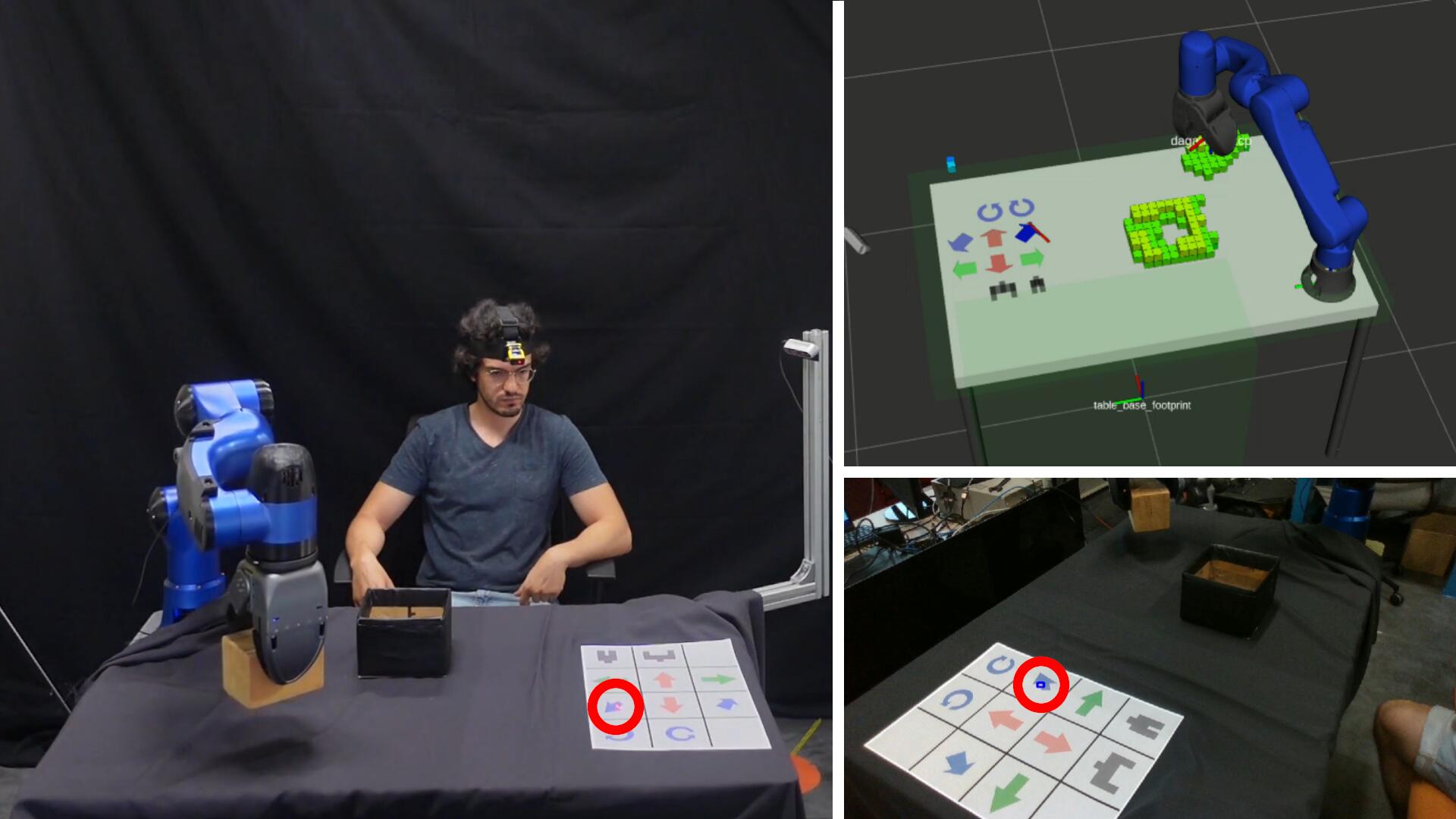}\hspace{5px}
	\includegraphics[width=0.48\linewidth]{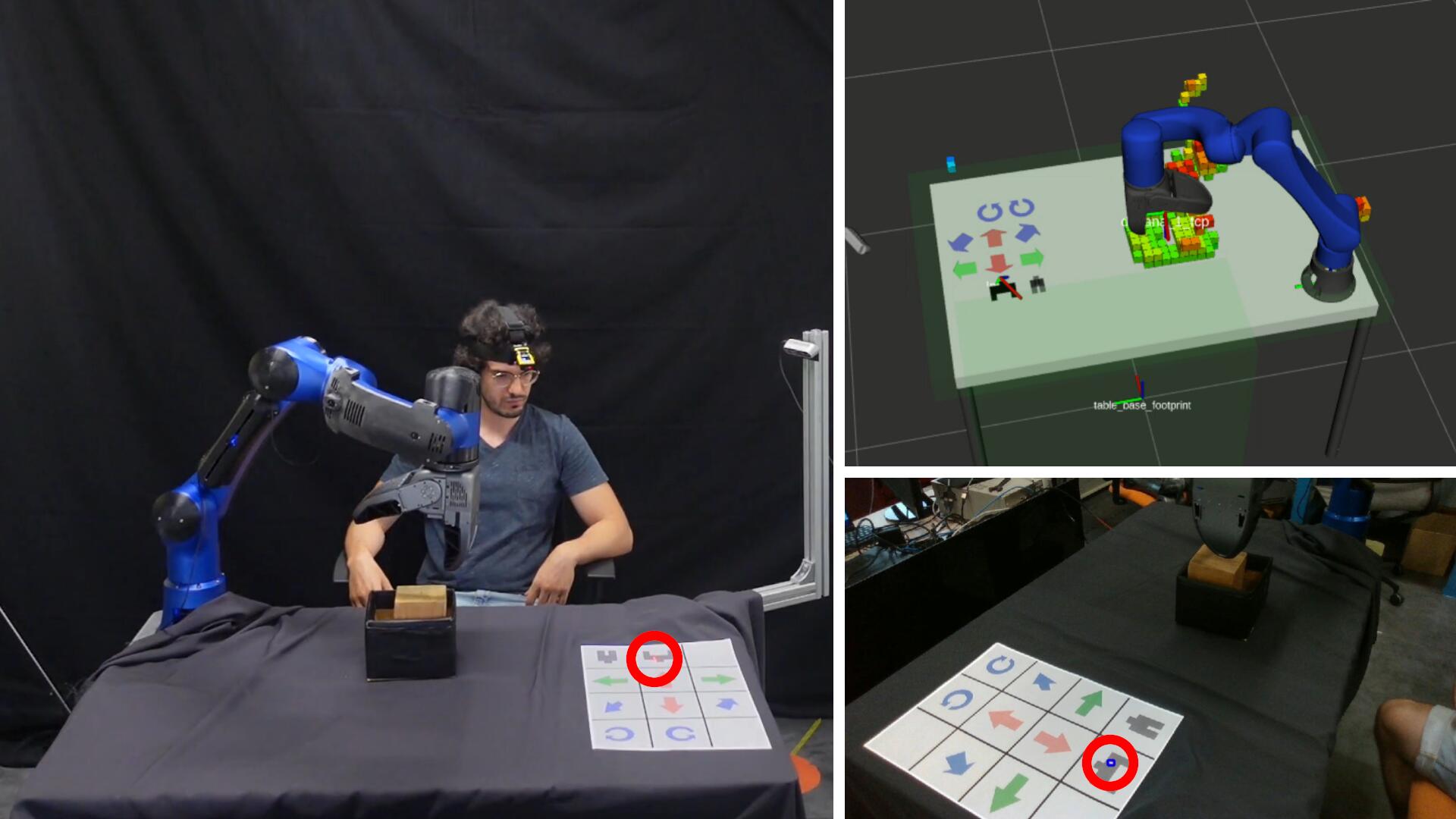}
	\caption[Laser assistive interface: wooden block experiment]{The wooden block pick-and-place experiment, where the impaired arms user commands the robot with the laser to transport the block into the container.}
	\label{fig:exp2-frames}
\end{figure}
\begin{figure}[H]
	\centering
	\includegraphics[width=0.32\linewidth]{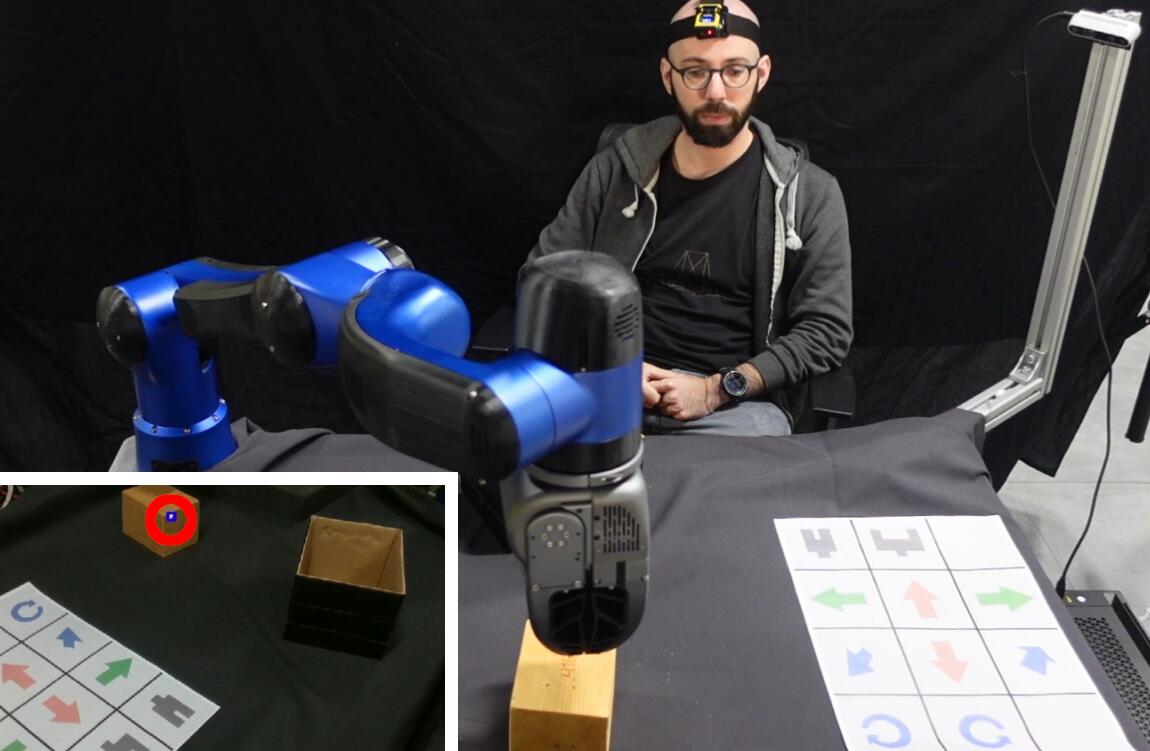}
	\includegraphics[width=0.32\linewidth]{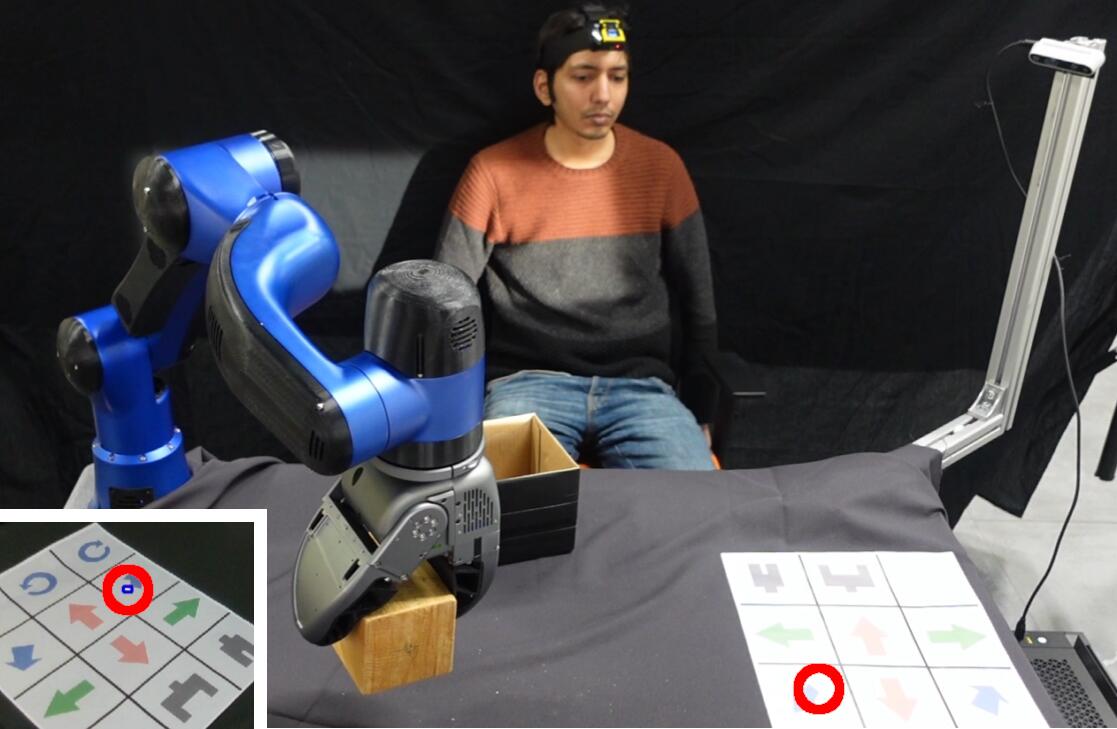}
	\includegraphics[width=0.32\linewidth]{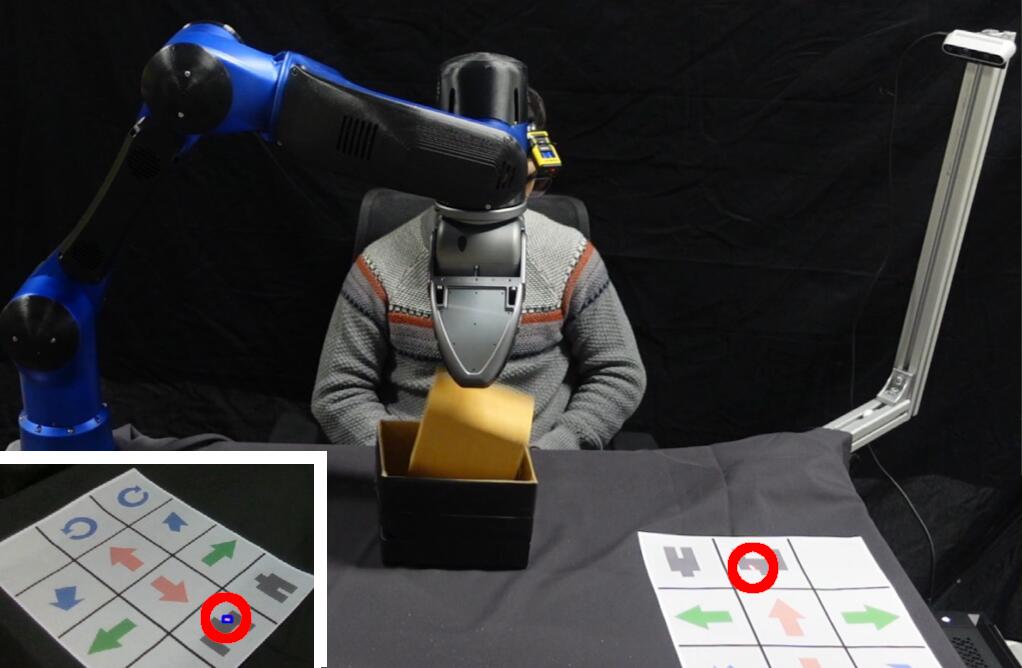}
	\caption{The \enquote{wooden block} pick-and-place experiment, executed by different users.}
	\label{fig:exp2-frames-people}
\end{figure}

\begin{figure}[H]
	\centering
	\includegraphics[width=0.48\linewidth]{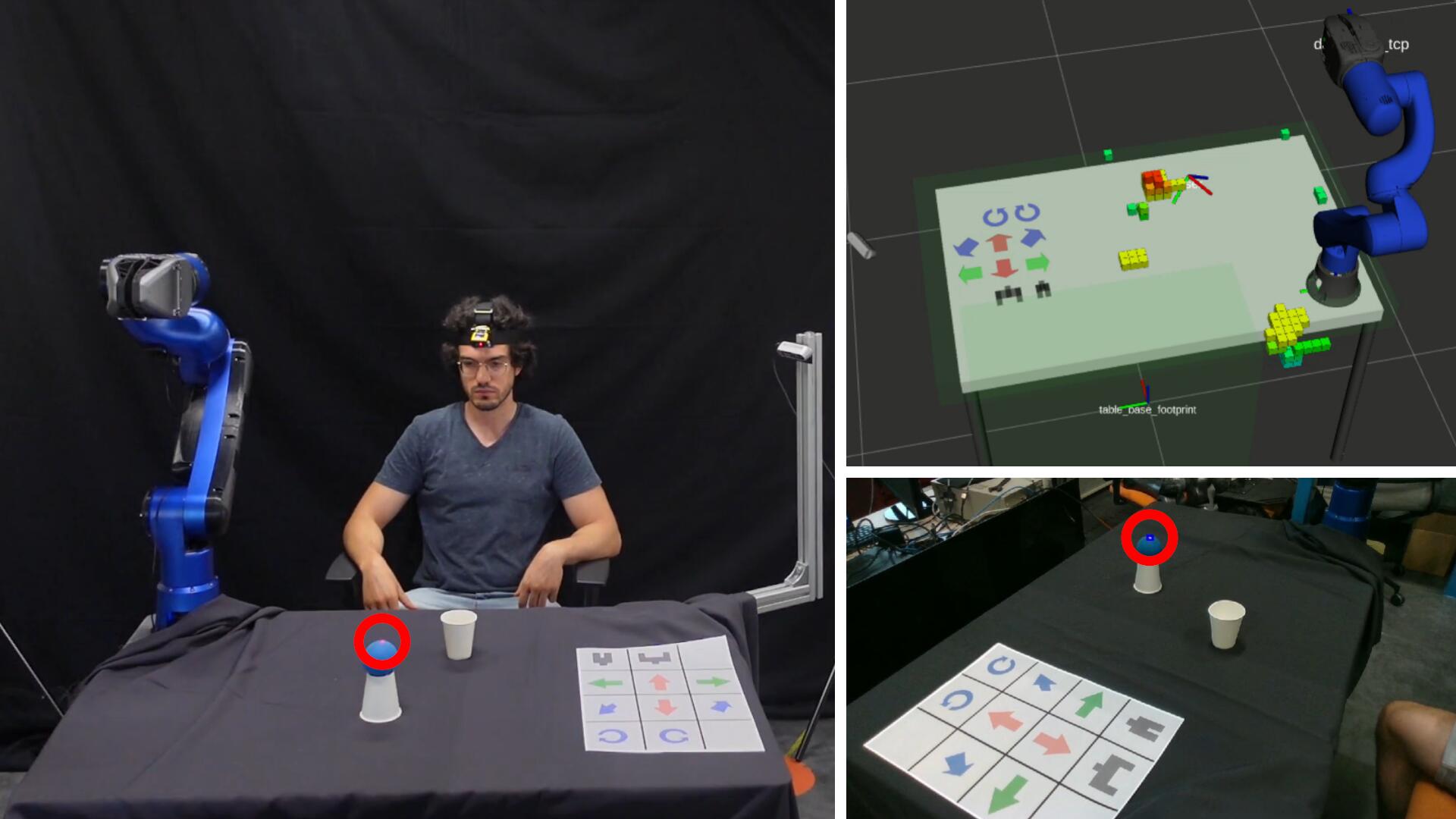}\hspace{5px}
	\includegraphics[width=0.48\linewidth]{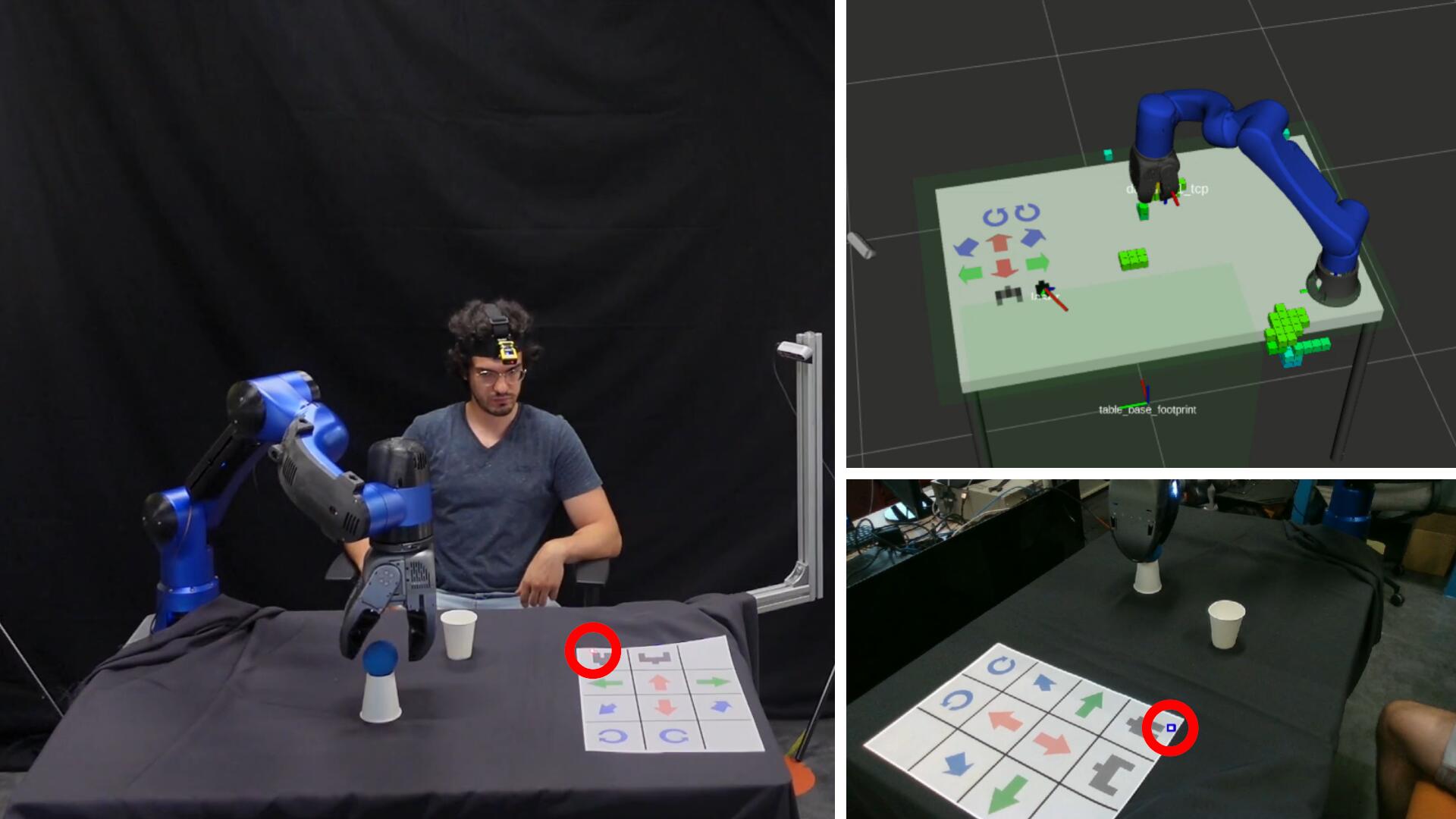}\\	\vspace{5px}
	\includegraphics[width=0.48\linewidth]{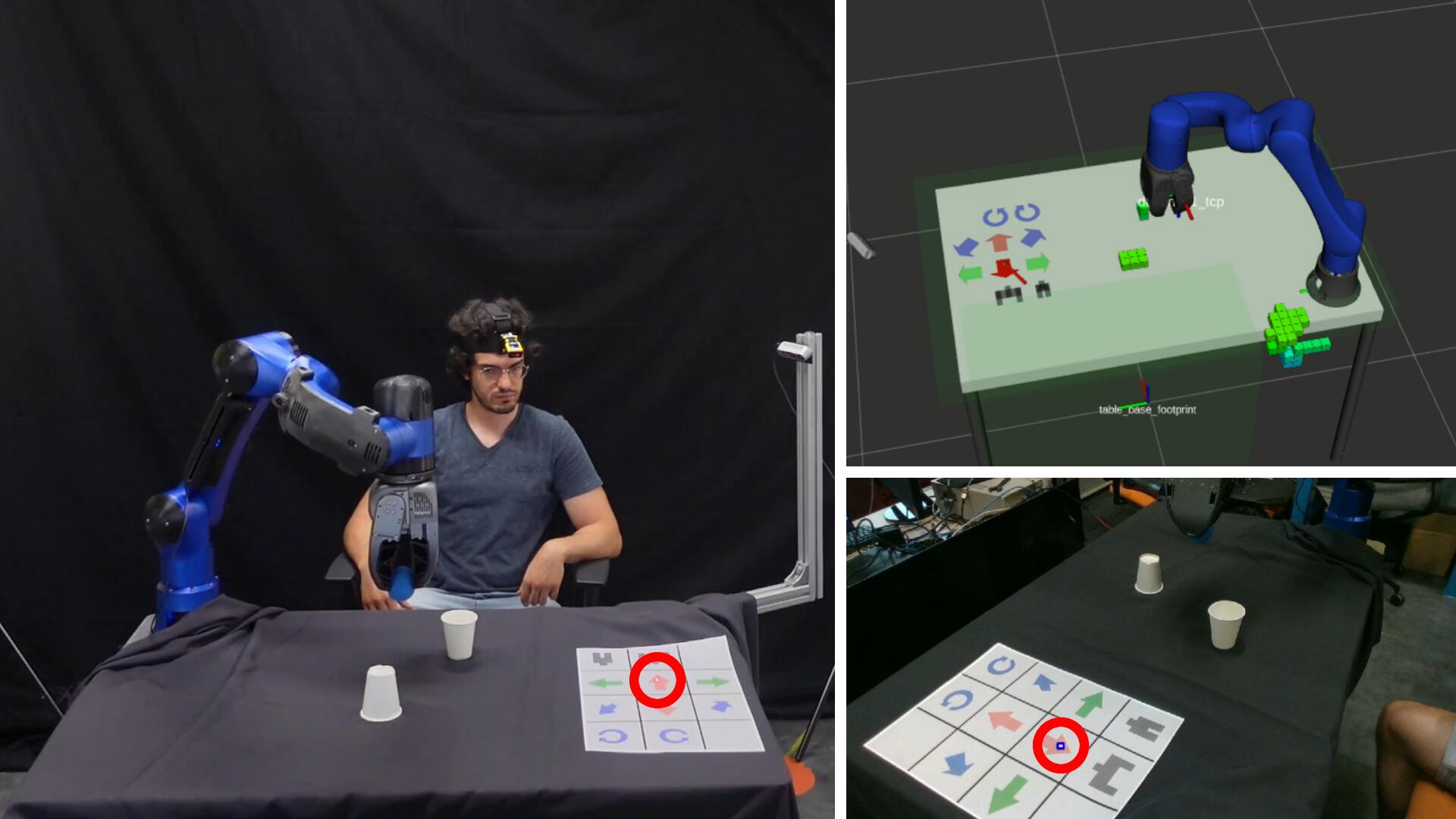}\hspace{5px}
	\includegraphics[width=0.48\linewidth]{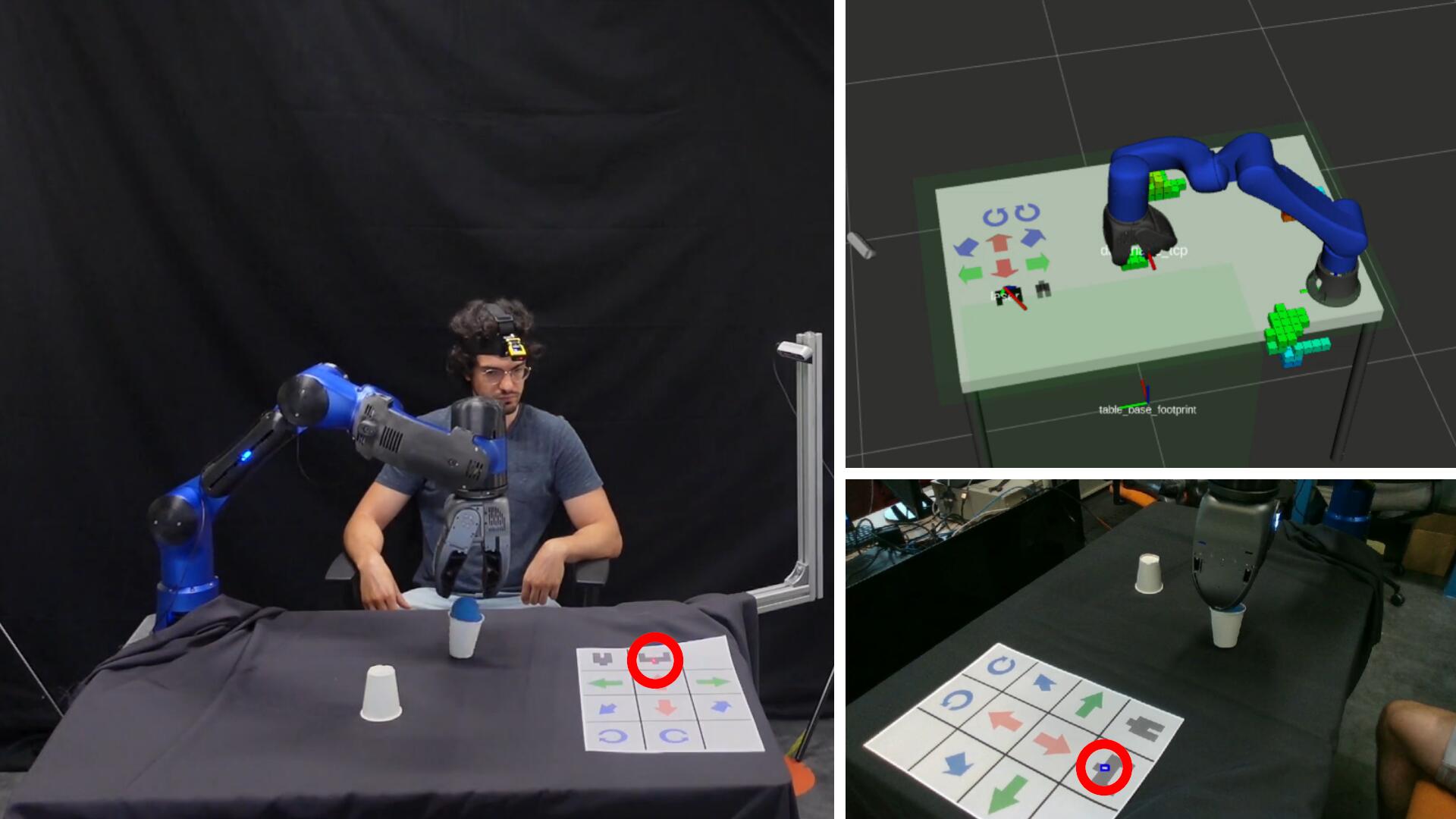}
	\caption[Laser assistive interface: soft ball experiment]{The soft ball pick-and-place experiment, where a precise transportation of the ball inside the plastic cup is requested.}
	\label{fig:exp3-frames}
\end{figure}

\begin{figure}[H]
	\centering
	\includegraphics[width=0.32\linewidth]{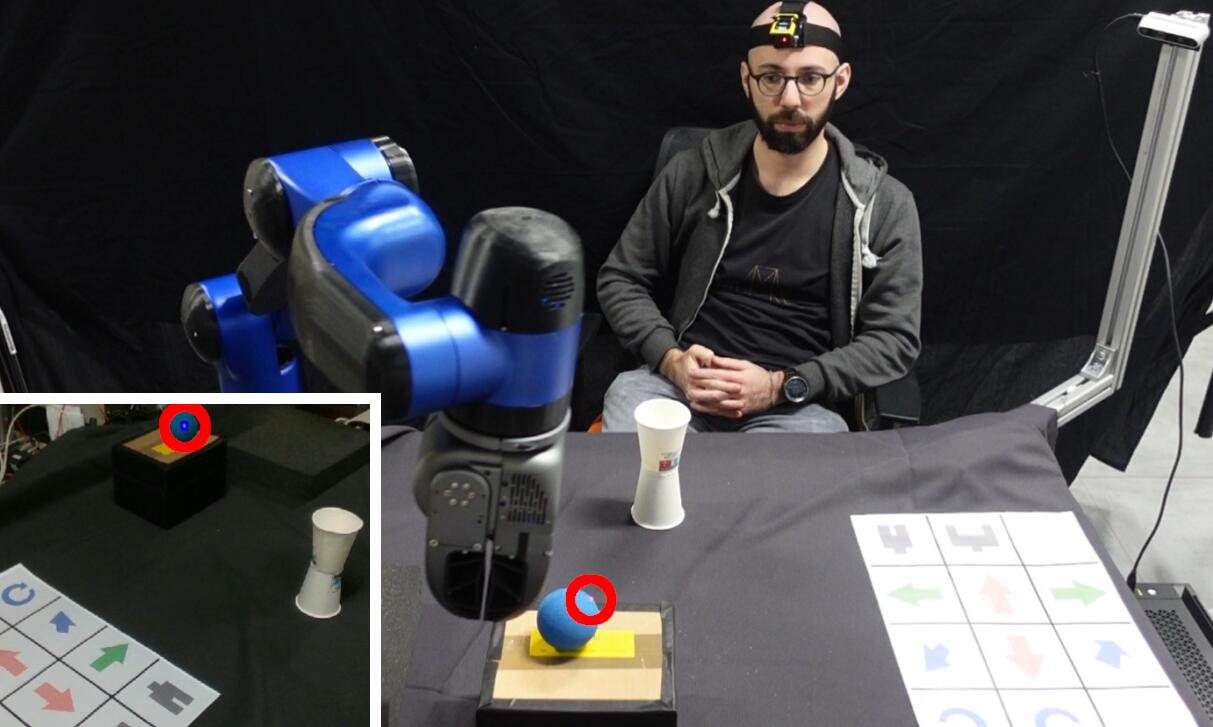}
	\includegraphics[width=0.32\linewidth]{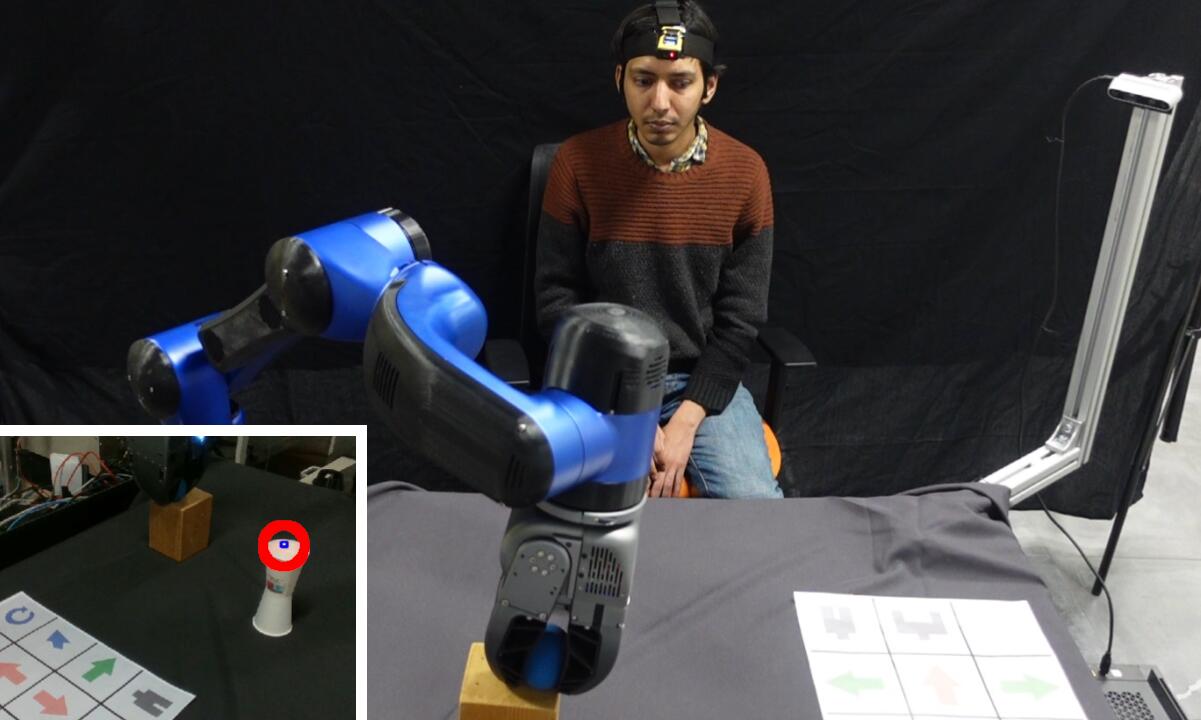}
	\includegraphics[width=0.32\linewidth]{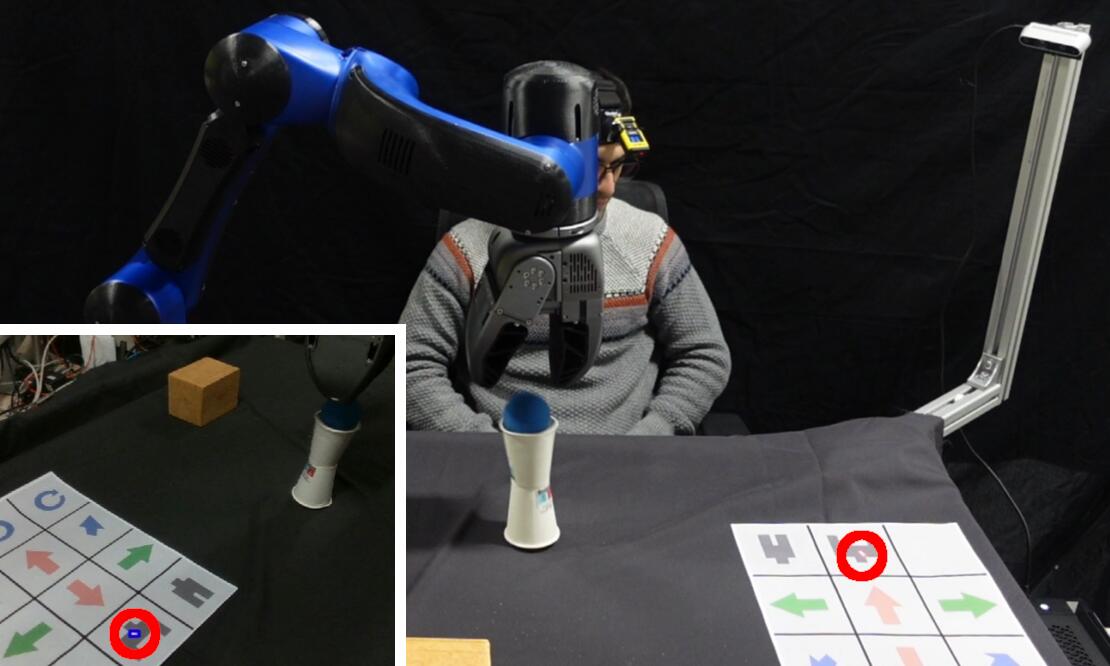}
	\caption{The \enquote{soft ball} pick-and-place experiment, executed by different users.}
	\label{fig:exp3-frames-people}
\end{figure}

In the second set of experiments four different healthy subjects have been involved, simulating both upper limbs impairments. Hence, they can not participate physically in the tasks, but they can only interact with the robot through the interaction control modalities of the proposed interface to command the manipulator to accomplish the tasks.
The first pick-and-place experiment involves transporting a wooden block inside a container, while the other one requires putting a soft ball inside a small cardboard glass. Images of the first subject executing the wooden block experiment are shown in \figurename{}~\ref{fig:exp2-frames}, while the other three subjects are displayed in different moments of the task in \figurename{}~\ref{fig:exp2-frames-people}. The same subjects accomplishing the soft ball experiment are shown in \figurename{}~\ref{fig:exp3-frames} and \figurename{}~\ref{fig:exp3-frames-people}.

\begin{figure}[H]
	\centering
	\includegraphics[width=1\linewidth]{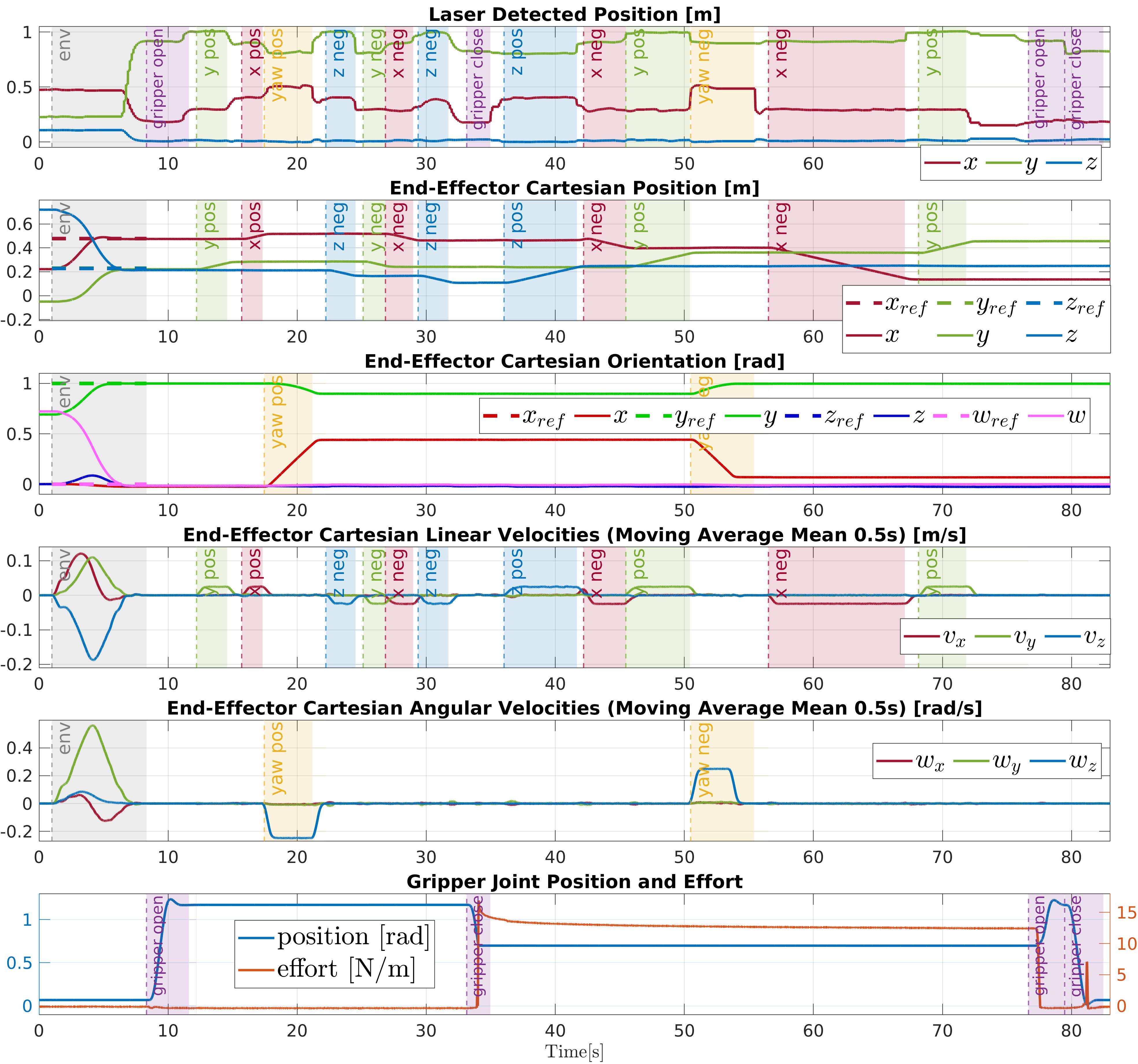}
	\caption[Laser assistive interface: wooden block experiment plot]{Wooden block pick-and-place experiment plots for subject 1, highlighting the intervals when the user is commanding the robot with a specific modality. Plot's layout is explained in Section~\ref{sec:laserAssistive:plotLayout}.}
	\label{fig:exp2-all}
\end{figure}

\begin{figure} [H]
	\centering
	\includegraphics[width=1\linewidth]{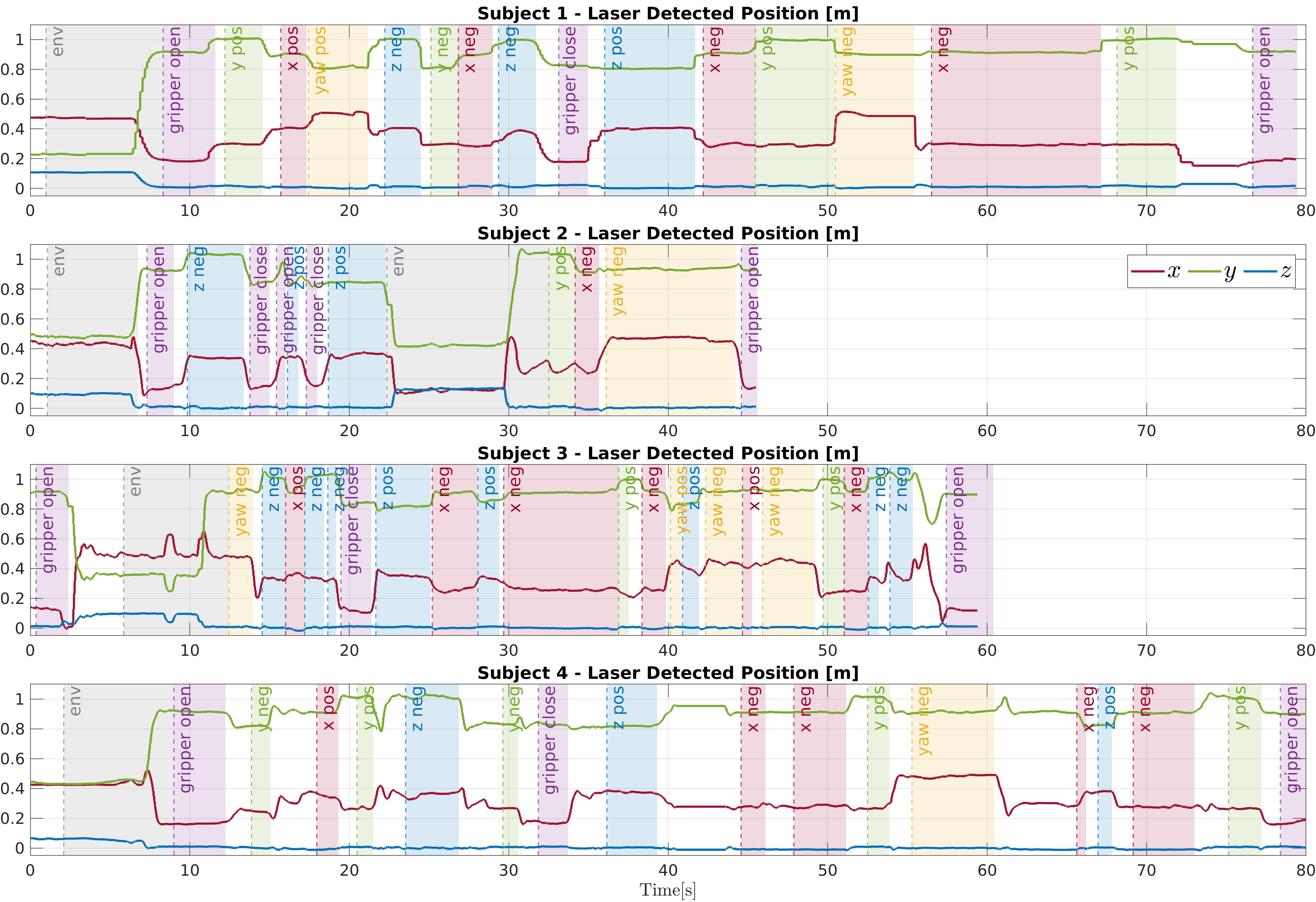}
	\caption{\enquote{Wooden block} pick-and-place experiments plots for all subjects. Each row represents the laser spot position in the task accomplished by a specific subject. The activated robot commands are highlighted by the colored areas.}
	\label{fig:exp2-all-people}
\end{figure}

\begin{figure}[H]
	\centering
	\includegraphics[width=1\linewidth]{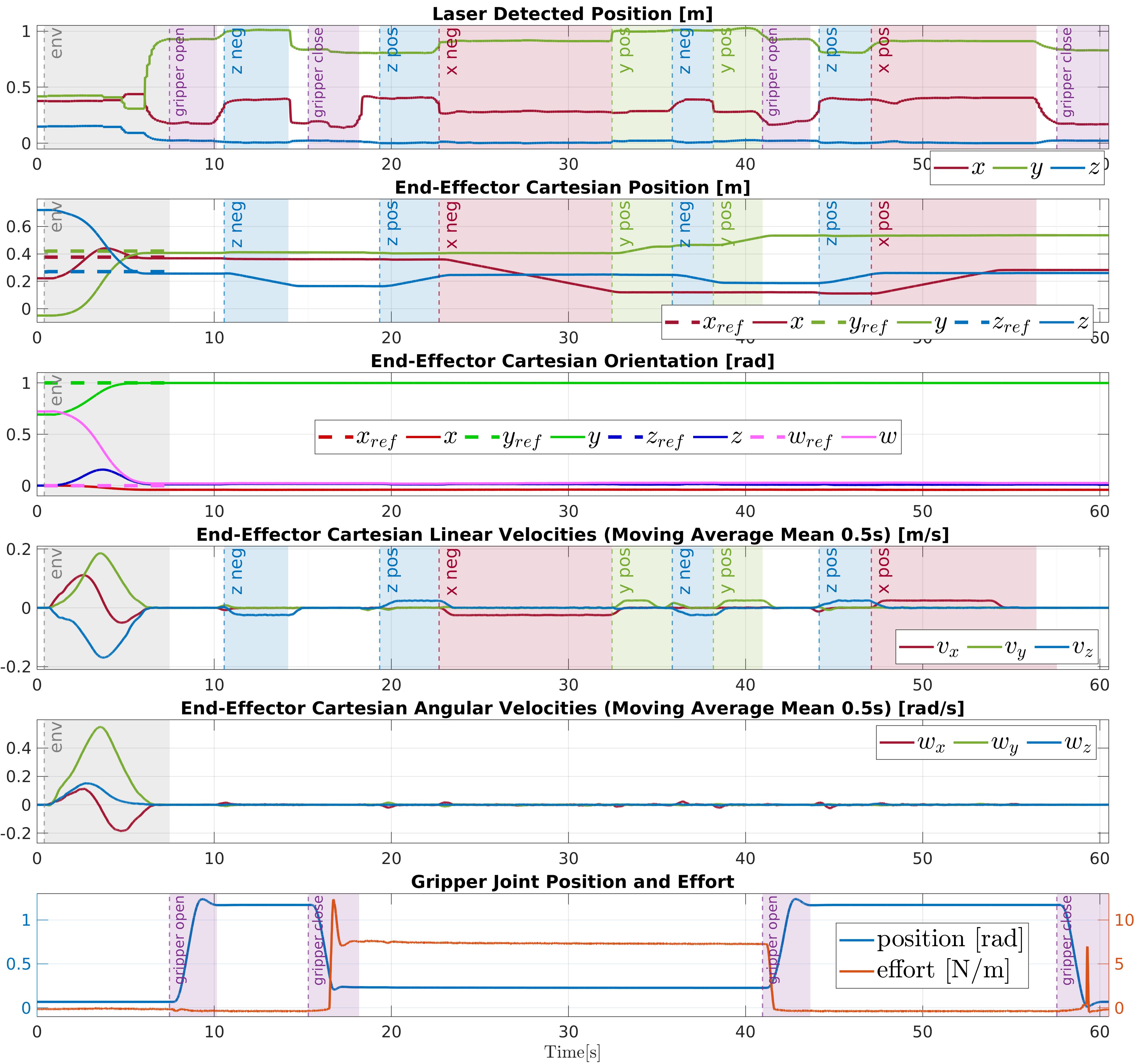}
	\caption[Laser assistive interface: soft ball experiment plot]{Soft ball pick-and-place experiment plots for subject 1, highlighting the intervals when the user is commanding the robot with a specific modality. Plot's layout is explained in Section~\ref{sec:laserAssistive:plotLayout}.}
	\label{fig:exp3-all}
\end{figure}

\begin{figure}[H]
	\centering
	\includegraphics[width=1\linewidth]{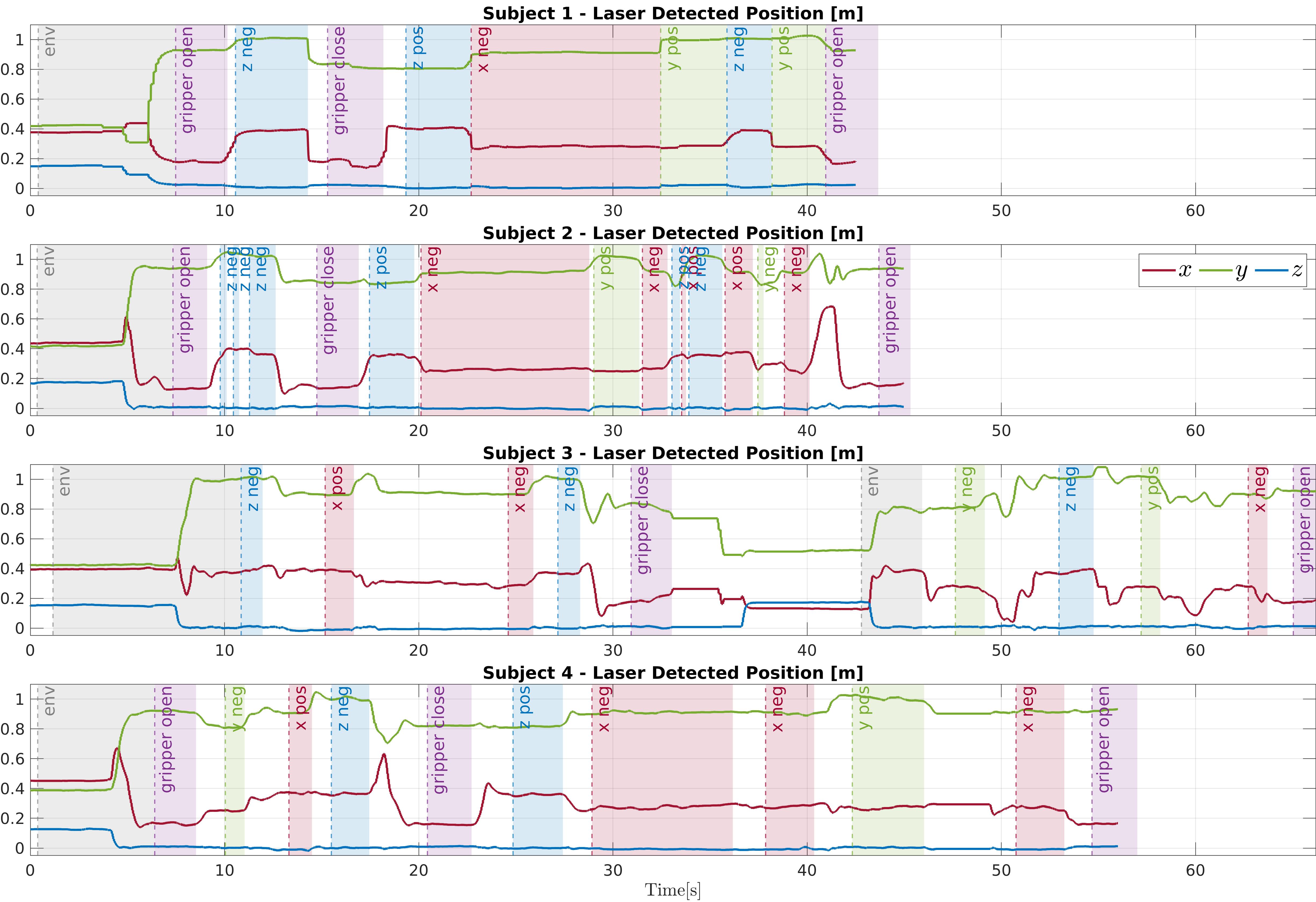}
	\caption{\enquote{Soft ball} pick-and-place experiments plots for all subjects. Each row represents the laser spot position in the task accomplished by a specific subject. The activated robot commands are highlighted by the colored areas.}
	\label{fig:exp3-all-people}
\end{figure}

The plots related to the first subject performing the wood block experiment are shown in \figurename{}~\ref{fig:exp2-all}, with the same layout as in the previous experiments, as explained in Section~\ref{sec:laserAssistive:plotLayout}. Instead, \figurename{}~\ref{fig:exp2-all-people} shows, on each row, the laser spot position during the tasks performed by the four subjects. As in other plots, the highlighted areas represent the command triggered by the laser spot position.
Similarly, the plots related to the first subject performing the soft ball experiment are shown in \figurename{}~\ref{fig:exp3-all}, while \figurename{}~\ref{fig:exp3-all-people} shows, on each row, the laser spot position during the tasks performed by the four participants. 

These pick-and-place experiments share similar characteristics, but they have different kinds of challenges. In the first one, end-effector yaw rotations are necessary to align the gripper with the object to grasp and with the container. Instead, the second one demands more precision due to the small size of the ball and of the plastic cup.
In all the trials, users effectively combine the environment control mode and the keyboard control mode to grasp and transport the objects successfully into their respective containers. It can be observed how the subjects chose to employ the two control modalities differently according to their preferences. For example, subject 2, with the wooden block, and subject 3, with the soft ball, utilized the environment control mode not only at the beginning of the task to reach the object to grasp, but also after, to command the robot toward the container.
Furthermore, the availability of the two control modalities along with their effortless interchangeability allowed users to complete the tasks within a reasonable amount of time, as shown in the plots.
In general, the experiments highlighted the efficacy of the system in intuitively controlling the robot using only head movements to direct the spot of the laser emitter. 
\section{Conclusions}\label{sec:laserAssistive:conclusions}

In the previous Chapter~\ref{chap:Laser1}, a laser-guided human-robot interface has been introduced to allow users to control a robot by intuitively command points of interest with a laser emitter device. 
In this chapter, this concept has been applied and realized in an assistive scenario. 
The interface has been designed to assist individuals with upper limbs disabilities, thus, the laser emitter is conveniently worn on the head of the person allowing to guide the robot solely with head movements.
The interface incorporates two interchangeable control modalities that offer flexibility to the user. In the first modality, pointing the laser to a specific location commands the robot to move to that location.
In the second modality, a paper keyboard is utilized, with buttons that can be virtually selected by directing the laser onto them (Section~\ref{sec:laserAssistive:system}).  

With the first modality, the robot autonomy is employed to generate a collision-free trajectory to the target, allowing the user to control the robot in a supervisory control mode~\cite{Gaofeng2023}. This feels very natural since users simply needs to direct their head toward the target, resulting in the laser to be pointed in the wanted location.
The keyboard of the second modality enables a more direct control of the robot, allowing the user to command more specific robot abilities, such as end-effector Cartesian velocities and gripper actions. 
By integrating these two modalities in the proposed assistive laser-based interface, impaired arm users can control the robot accomplishing both co-manipulation and grasping tasks. Furthermore, they have the flexibility to choose the strategy that best fits their motion capabilities, preferences, and task needs (Section~\ref{sec:laserAssistive:methods}).

Several validation experiments have been conducted, involving a robotic manipulator equipped with the DAGANA gripper.
The experiments showcased the versatility of the interface, its effectiveness, and the intuitive control it provides to single and double-arm impaired users, enabling them to interact with the assistive robot arm and collaboratively execute \acrshort{adl} tasks (Section~\ref{sec:laserAssistive:exp}).

In the future, more autonomous robot capabilities will be developed and integrated, such as autonomous grasping of the pointed object. 
Further experiments will involve individuals with the targeted disabilities, to conduct user studies and to explore more the capability of the interface. Efforts will focus on investigating any specific residual motions that users may have in their impaired arm or arms to further extend the control functionalities of the introduced interface.
These advancements will contribute to the development of more sophisticated assistive robotic systems and their application in diverse home-care settings.

\part*{}

\chapter{Conclusions}\label{chap:conclusion}

\lettrine{A}{s} introduced in Chapter~\ref{chap:intro}, this thesis has focused on the development of innovative human-robot interfaces, addressing the challenges in operating modern robotic systems. 
The aim of these interfaces is to seamlessly connect humans to robots, enabling intuitive control over the various robot capabilities. 
As discussed in the state-of-the-art Chapter~\ref{chap:soa}, such capabilities are promising because they are fostering the deployment of robots in various domains, but at the same time it is evident that new human-robot communication paradigms are necessary to effectively exploit such systems.

Human-robot interfaces can cover a wide range of technologies and can be developed for solving a variety of challenges. In this thesis, as presented in Chapter~\ref{chap:chap3}, the key themes addressed are the intuitiveness of the interface, the improvement of user's situation awareness, especially through haptic feedback, and the implementation of different levels of robot autonomy.

Developing human-robot interfaces that are intuitive means offering users with a natural, easy-to-use interaction that allows them to focus on \textit{what} command to provide to the robot rather than \textit{how} to communicate it. The developed solutions have achieved this by empowering users to utilize their bodies, allowing for intuitive communication of their commands without requiring the user to deal with a complex mapping between the input and the command delivered. 

To let the operator understand what it is happening during the human-robot interaction, it is important to equip the interface with feedback channels for user's situation awareness. 
These channels should help the operator with relevant information, without burdening him/her with unnecessary details or overwhelming him/her with complex feedback means.
While the direct visual perception of the command issued to the robot and the developed \acrshort{gui} modules serve as simple situation awareness tools, the most interesting developed work in this field involves the exploration of a wearable vibrotactile device, designed to complement the human-robot interfaces with haptic clues. 

Incorporating robot autonomy is crucial for enhancing efficiency and user-friendliness of the human-robot interaction. While full autonomy remains an ideal, modern robots can operate semi-autonomously, reducing the cognitive load on human operators and allowing them to focus on the most important aspect of the task. By exploring shared control and supervisory control techniques, the presented works have furbished the robot with a level of autonomy adequate to the task to be accomplished.

\section{Summary of the Developed Human-Robot Interfaces}\label{sec:con:dev}

Two main categories of interfaces have been developed and have been presented in this thesis. The first category is related to the \acrfull{tpo} concept (Part~\ref{part:one}), while the second one to laser-guided interfaces (Part~\ref{part:two}).

The \acrshort{tpo} interfaces allow for a direct control of the robot, while maintaining the intuitiveness and harnessing the potential of highly-capable robots through various implemented autonomy features. 
Laser-guided interfaces leverage on supervisory control, reducing the operator's effort and facing increased robot autonomy requirements with appropriate autonomous motion planning solutions. 

The proposed concepts have been realized with the development of appropriate software frameworks, ensuring versatility across a various range of robotic systems and applications.

\subsection{TelePhysicalOperation}
\acrlong{tpo}, introduced in Chapter~\ref{chap:TPO}, enables direct and intuitive control of robots through the operator's arm movements. By adhering to the \enquote{Marionette} metaphor, the \acrshort{tpo} concept facilitates the remote robot control by allowing the user to apply virtual forces to the robot's body parts, shaping its kinematic chains to exploit redundancy. To realize such paradigm, the \acrshort{tpo} Suit has been designed, equipped with lightweight tracking cameras, and providing a comfortable, wireless, and simple interface.

In Chapter~\ref{chap:TPOH}, the exploration of a haptic feedback channel has been conducted, to provide the operator with tactile stimuli when manipulating the virtual ropes of the \enquote{Marionette}. The wearable vibrotactile devices that deliver such stimuli augment the user's perception of what it is happening, also by incorporating feedback related to the forces sensed at the robot's end-effectors. Additionally, the incorporated buttons expand the user input possibilities. The devices, integrated in the \acrshort{tpo} Suit, are lightweight, simple, and unobtrusive, aligning with the principles of the entire suit.

In Chapter~\ref{chap:tpoAuto}, some autonomy features have been incorporated in the \acrlong{tpo} interface, adopting a shared control approach that facilitates the control of highly-redundant robots for the execution of complex tasks. 
With the manipulability-aware shared locomanipulation motion generation, \acrshort{tpo} virtual forces applied on the end-effector of a mobile manipulator are automatically distributed to the arm and to the mobile base. Hence, the operator can exclusively control the end-effector, while the underlying architecture generates the mobile platform commands depending on the end-effector manipulability level. 
Furthermore, with the bimanual object transportation interface, objects of unknown mass are reached and grasped autonomously by the robot. During the teleoperated transportation, the robot follows object directions imposed by the operator while autonomously adjusting the grasping forces according to the estimated weight of the object. 

\subsection{Laser-guided Interface}
Laser-guided interfaces, introduced in Chapter~\ref{chap:Laser1}, allow the operator to guide the robot by targeting locations of interest with an inexpensive laser emitter device. 
This mode of commanding the robot is inherently intuitive, mimicking the way people naturally target locations or objects of interest, i.e., by pointing at them. 
The user has also a direct visual feedback of the command given to the robot since he/she can directly see the laser projection.

The implemented vision system, based on a neural network, robustly detects the laser spot, allowing for a fast and robust tracking. 
With respect to the more direct control of the \acrlong{tpo}, this supervisory control interface demands a higher level of robot independence, an issue addressed in diverse ways.

Chapter~\ref{chap:Laser1} has detailed a solution based on Behavior Trees (BTs) for the robot's motion generation, especially useful for highly-redundant robots. The BT planner offers reactivity, enabling the robot to promptly respond to goal changes, making it possible to track the laser spot in real-time switching among the various robot capabilities, allowing the operator to command not only a static goal but also a path to follow.
The BT solution also offers modularity, to easily modify the robot behavior according to the task's needs and robot's features.

Chapter~\ref{chap:Laser2} has realized the laser-guided interaction concept for an assistive scenario, where impaired arm(s) users utilize a head-worn laser emitter to control the robot through head movements, enabling a natural and intuitive human-robot interaction.
Two control modalities are provided, with the possibility to seamless switching between them just by directing the laser, without requiring any additional input device. 
With the first modality, the user can indicate a goal with the laser in the robot's workspace, with the robot generating and executing a collision-free path toward it. With the second modality, a more direct control of the robot capacities is enabled to command, for example, the end-effector Cartesian velocities and gripper actions. This is achieved by projecting the laser onto a paper keyboard placed in the environment.

\section{Future works}
In the pursuit of addressing the evolving challenges of robotics, and pushing the boundaries of human-robot interfaces, directions for future works emerge to further reduce the gap which limits the spread of robotics technologies in real scenarios.

This thesis has demonstrated promising advancements in human-robot interfaces, showcased through practical applications. The experiments have resembled real-world contexts, but they have been validated in laboratory settings. Indeed, use cases in real scenarios, where uncertainties are higher, have not been performed yet. 
For the TelePhysicalOperation interface, a user study has been conducted, but more appropriated studies are necessary to further understand what can be improved, and how much people are willing to accept robots and the associated way of interacting with them.
Further trials should encompass more complex tasks, such as the bimanual transportation, not yet evaluated with multiple users. Further validations are also necessary on the haptic side, to understand precisely what kind of stimuli is the most effective for a specific application. 

In assistive scenarios, it is fundamental to test robotic technologies with the patients, since they are the ultimate end-users that must evaluate the interface.
Furthermore, understanding the adaptability of these systems to diverse individuals with varying disabilities is paramount, as injuries and impairments can manifest in unique ways, requiring tailored support. 
  
Exploring new theoretical and practical developments opens up new horizons across all the key themes addressed in this thesis. 
The introduction of multimodal interfaces opens doors to novel input possibilities, but it is crucial to strike a balance, ensuring they remain intuitive and do not overwhelm the operator.
This entails the exploitation of the natural ways people communicate in everyday life, integrating input possibilities like natural gestures. For instance, in the context of the laser-guided interface, the potential exists in employing the laser not only to point at locations, but also to draw particular shapes to command what the robot should do at the indicated location, like removing an obstacle or using a tool for a particular task.

As stated, it is important to keep the operator informed of what it is happening, that can be done conveying haptic feedback to surpass the limitations of visual feedback. With the exploration of more elaborated haptic devices and the conduction of more user studies, it can be defined what it is necessary to enable a more intuitive human-robot bilateral communication during complex tasks.

To address complex tasks, and to prevent overwhelming the operator, it is also crucial to increase the independence of the robot. One practical example can be enhancing the robot's vision system for laser detection. Currently, the laser spot is perceived by the robot merely as a point in the space to be reached. 
With more cognitive capabilities, the robot can comprehend \textit{what} it is being pointed, and interact with it accordingly. For instance, the robot can recognize that an object pointed can be manipulated with a single gripper, or that it must be handled bimanually because of the size.
A field where robot intelligence is particularly important is the assistive one. Future works will also focus on this scenario to enable the utilization of more complex systems, such multiple manipulators, thereby expanding the range of achievable \acrshort{adl} tasks. In this context, since the user interface must be kept simple, further explorations about the robot autonomy are necessary to generate motions for multiple robots, allowing them to coordinate and potentially cooperate with the user.

In conclusion, by consistently exploring intuitiveness, enhancing user situation awareness, and advancing robot autonomy, along with the rigorous validation of these technologies in real-world scenarios through extensive user studies, it will be possible to provide innovative human-robot interfaces to unlock the full potential of robots and expand the horizons of their applications.

\backmatter

\begin{spacing}{0.9}

\printbibliography[heading=bibintoc]

\end{spacing}

\printthesisindex %

\end{document}